%% file: StatisticalRL-nowplain.tex
\PassOptionsToClass {plain,biber}{nowfnt}
\documentclass[examplefnt,biber]{nowfnt} % creates the journal version, needs biber version
\usepackage{amsmath,amsbsy,amstext,amssymb,amsthm,mathtools,comment,dsfont,mathrsfs,cleveref}
\usepackage{color}
\usepackage{bbm}
\usepackage{bm}

\usepackage[linesnumbered,ruled,vlined]{algorithm2e}
\numberwithin{algocf}{chapter}
\usepackage{algorithmic}

\usepackage{appendix}

\input{chapters/commands.tex}

\usepackage[utf8]{inputenc}

\newcommand{\vertiii}[1]{{\left\vert\kern-0.25ex\left\vert\kern-0.25ex\left\vert #1 
    \right\vert\kern-0.25ex\right\vert\kern-0.25ex\right\vert}}

\newcommand{\vertiiibig}[1]{{\big\vert\kern-0.25ex\big\vert\kern-0.25ex\big\vert #1 
    \big\vert\kern-0.25ex\big\vert\kern-0.25ex\big\vert}}

\newcommand{\vertiiiplain}[1]{{\vert\kern-0.25ex\vert\kern-0.25ex\vert #1 
    \vert\kern-0.25ex\vert\kern-0.25ex\vert}}

\definecolor{yxc}{RGB}{255,0,0}
\definecolor{yjc}{RGB}{125,0,0}
\definecolor{ytw}{RGB}{255,105,180}
\definecolor{yly}{RGB}{0,150,0}

\newcommand{\yxc}[1]{\textcolor{yxc}{[Yuxin: #1]}}

\newcommand{\arxiv}[1]{}
\newcommand{\book}[1]{#1}

\newtheorem{assumption}{\textbf{Assumption}}[chapter]
\newtheorem{claim}{\textbf{Claim}}[chapter]
 
\newcommand{\mynewline}{}

\title{Statistical Foundations of Reinforcement Learning: \\ A Non-Asymptotic Perspective}
\maintitleauthorlist{
Yuxin Chen \\
University of Pennsylvania \\
yuxinc@wharton.upenn.edu \\
\and
Yuejie Chi \\
Yale University \\
yuejie.chi@yale.edu \\
\and
Jianqing Fan\\
Princeton University \\
jqfan@princeton.edu \\
\and
Gen Li\\
Chinese University of Hong Kong \\
genli@cuhk.edu.hk \\
\and
Yuting Wei\\
University of Pennsylvania \\
ytwei@wharton.upenn.edu \\
\and
Yuling Yan\\
University of Wisconsin-Madison \\
yuling.yan@wisc.edu 
}

\issuesetup
{%
 copyrightowner={A.~Heezemans and M.~Casey},
 pubyear       = 2026,
 }

\usepackage{mwe}

\author[1]{Chen,Yuxin}
\author[2]{Chi,Yuejie}
\author[3]{Fan,Jianqing}
\author[4]{Li,Gen}
\author[5]{Wei,Yuting}
\author[6]{Yan,Yuling}

\affil[1]{University of Pennsylvania; yuxinc@wharton.upenn.edu}
\affil[2]{Yale University; yuejie.chi@yale.edu}
\affil[3]{Princeton University; jqfan@princeton.edu}
\affil[4]{Chinese University of Hong Kong; genli@cuhk.edu.hk}
\affil[5]{University of Pennsylvania; ytwei@wharton.upenn.edu}
\affil[6]{University of Wisconsin-Madison; yuling.yan@wisc.edu}

\makeatletter
\renewcommand{\nowfntstandardcitation}{%
  \nowfnt@authorlist@toc\ (\nowfnt@issuepubyear),
  ``\nowfnt@articletitle'',
  \nowfnt@journaltitle: Vol. \nowfnt@issuevolume, No. \nowfnt@issueissue,
  pp \nowfnt@issuefirstpage--\nowfnt@issuelastpage. DOI: \nowfnt@issuedoi.}
\makeatother
\articledatabox{\nowfntstandardcitation}

\makeatletter
\providecommand\Hy@colorlink[1]{\begingroup}
\providecommand\Hy@endcolorlink{\endgroup}
\makeatother
\hypersetup{pdfborder={0 0 0},pdfborderstyle={/S/U/W 0}}

\begin{document}

\makeabstracttitle
\renewcommand{\mynewline}{}

\begin{abstract}

\input{chapters/abstract.tex}

\end{abstract}

\newpage

\input{chapters/intro.tex}

%
\newpage

\input{chapters/prelim.tex}

\newpage

\input{chapters/generative_model_model_based.tex}

\newpage
\input{chapters/generative_model_model_free.tex}
%%%

%
\newpage

\input{chapters/onlineRL.tex}
\newpage

\input{chapters/offlineRL.tex}

\newpage
\input{chapters/cousins.tex}

\newpage
\input{chapters/conclusion.tex}

\newpage
%
% % end of main matter

\input{chapters/ack.tex}

%\appendix 

%BACKMATTER SEE DOCUMENTATION
\backmatter  % references, restarts sample

\printbibliography

\end{document}

%% file: chapters/commands.tex
\allowdisplaybreaks

\usepackage{subcaption}

\DeclareMathOperator{\ind}{\mathds{1}}  % Indicator

\newcommand{\mycommentbegin}{\begin{comment}} 
\newcommand{\mycommentend}{\end{comment}}

\newcommand{\rewabs}{r^{\mathsf{abs}}}

\newcommand{\Pv}{P_{\pi}}
\newcommand{\Phatv}{\widehat{P}_{\pi}}

\newcommand{\Phat}{\widehat{P}}
\newcommand{\abPbm}{P^{\mathsf{abs}}}

\newcommand{\Vstar}{V^\star}
\newcommand{\Vhat}{\widehat{V}}
\newcommand{\Vhatstar}{\widehat{V}^\star}

\newcommand{\ucbq}{\texttt{Q-UCB}\xspace}
\newcommand{\ucbvi}{\texttt{UCB-VI}\xspace}
\newcommand{\vilcb}{\texttt{VI-LCB}\xspace}
\newcommand{\ucrl}{\texttt{UCRL}\xspace}
\newcommand{\ucrltwo}{\texttt{UCRL2}\xspace}
\newcommand{\mvp}{\texttt{MVP}\xspace}
\newcommand{\ucb}{\textsf{ucb}\xspace}
\newcommand{\low}{\textsf{low}\xspace}
\newcommand{\iotab}{\iota_{\mathrm{b}}\xspace}

\newcommand{\prnbig}[1]{\big({#1}\big)} % parentheses
\newcommand{\KL}[2]{\mathsf{KL}\prnbig{{#1}\,\|\,{#2}}}

\newcommand{\cD}{\mathcal{D}}

\newcommand{\innprod}[1]{\big\langle{#1} \big\rangle} % inner product
\newcommand{\dnorm}{\mathcal{\pi}_{\mathsf{cal}}}

\newcommand{\fedsyncq}{{\texttt{FedSynQ}}\xspace}
\newcommand{\starQ}{Q^{\star}}

\newcommand{\cC}{\mathcal{C}}

\newcommand{\pistar}{{\pi}^\star}

\newcommand{\pihatstar}{\widehat{\pi}^\star}

\newcommand{\ceta}{c_\eta}
\newcommand{\pib}{\pi^{\mathsf{b}}}
\newcommand{\tmix}{t_{\mathsf{mix}}}
\newcommand{\mumin}{\mu_{\mathsf{min}}}

\newcommand{\that}{\widehat{\cT}}

\newcommand{\NEgap}{\mathsf{gap}}

\newcommand{\DRVI}{\texttt{DRVI}\xspace}

\newcommand{\cA}{{\mathcal{A}}}

\newcommand{\cS}{{\mathcal{S}}}
\newcommand{\cM}{{\mathcal{M}}}

\newcommand{\cT}{{\mathcal{T}}}
\newcommand{\cU}{{\mathcal{U}}}

\newcommand{\bP}{P}

\newcommand{\argmin}{\mathop{\rm argmin}}

\newcommand{\sA}{\mathscr{A}}

\newcommand{\defn}{\coloneqq}
\newcommand{\MDP}{\mathcal{M}}

\newcommand{\net}[1]{\mathcal{N}_{#1}}

\newcommand{\one}{1}

\newcommand{\rhob}{\rho^{\mathsf{b}}}
\newcommand{\taumax}{\tau_{\mathsf{\max}}}

\newcommand{\myrho}{d^{\mathsf{b}}}
 
\newcommand{\infs}{}

\newcommand{\mymid}{\,|\,}
\newcommand{\cb}{c_{\mathsf{b}}}
\newcommand{\nominal}{\textsf{nom}\xspace}
\newcommand{\DPO}{\texttt{DPO}\xspace}
\newcommand{\VPO}{\texttt{VPO}\xspace}
\newcommand{\UCB}{\texttt{UCB}\xspace}

\newcommand{\LCB}{\texttt{LCB}\xspace}
\newcommand{\LCBQ}{\texttt{LCB-Q}\xspace}
\newcommand{\lcb}{{\sf lcb}\xspace}

\newcommand{\Cstar}{C^{\star}_{\rho}}

\newcommand{\rob}{{\mathsf{rob}}}

\newcommand{\no}{{\textsf{nom}\xspace}} % nominal transition kernel superscript
\newcommand{\ror}{\sigma} % uncertainty set radius
\newcommand{\unb}{\cU} % uncertainty set
\newcommand{\Tpess}{\widehat{\mathcal{T}}_{\mathsf{pe}}}
\newcommand{\tildeTpess}{\widetilde{\mathcal{T}}_{\mathsf{pe}}}

\newcommand{\bPi}{{\Pi}}
\newcommand{\Pq}{P^\pi}

\newcommand{\real}{\mathbb{R}}
\newcommand{\Var}{\mathsf{Var}}

%% file: chapters/abstract.tex
Reinforcement learning (RL) embodies a foundational paradigm for sequential decision-making in the face of uncertainty. The rapid growth of model complexity in emerging applications has significantly amplified the challenges of achieving data-efficient RL, particularly in sample-starved scenarios where data collection may be costly, time-consuming, or even high-stakes. Consequently, understanding and improving the data efficiency of RL algorithms has become a central problem of both fundamental theoretical interest and practical importance.

Despite decades of research, the statistical underpinnings of RL remained remarkably underdeveloped until relatively recently, particularly in the finite-sample regimes most relevant to modern practice. Over the past few years, there has been a surge of activity advancing the frontiers of non-asymptotic statistical RL theory, drawing on ideas from high-dimensional statistics and probability, information theory, and statistical learning theory. In this monograph, we present a non-asymptotic treatment for RL based on tabular Markov decision processes, focusing on three canonical sampling mechanisms --- RL with a generative model, online RL, and offline RL --- together with two algorithmic paradigms: model-based and model-free approaches. Centered around non-asymptotic sample complexity analysis, our treatment emphasizes minimax-optimal algorithm design, particularly methods that attain statistical optimality without significant burn-in costs, as well as sharp statistical characterizations of several widely used RL algorithms. Whenever possible, we provide largely self-contained analyses together with intuitive explanations. We also briefly discuss several extensions, such as robust RL, competitive multi-agent RL, federated RL, uncertainty quantification, and RL from human feedback. We hope this monograph equips aspiring researchers with essential tools from modern statistical RL and inspires further advances in both theory and algorithm design.

%, as encountered in clinical trials, autonomous driving, and online advertising, among others

%% file: chapters/intro.tex
\chapter{Introduction}

Reinforcement learning (RL) \citep{sutton2018reinforcement,bertsekas2019reinforcement}, commonly formulated as sequential decision-making and learning in a Markov decision process (MDP), has garnered immense attention in recent years due to its striking empirical success across a wide range of practical domains. By enabling agents to collect data and learn through repeated, trial-and-error interaction with their environments, RL has led to major breakthroughs in applications such as competitive gaming, robotics and control, autonomous driving, e-commerce, power systems, and healthcare, often achieving state-of-the-art performance in complex decision-making tasks \citep{silver2017mastering,mnih2015human,tang2025deep,kiran2021deep,hu2018reinforcement,bastani2021efficient,chen2022reinforcement,yu2021reinforcement,bai2025review}. Beyond these traditional application areas, RL has recently emerged as a cornerstone for fine-tuning and aligning large language models (LLMs). In particular, post-training techniques such as Reinforcement Learning from Human Feedback (RLHF) and Reinforcement Learning with Verifiable Rewards (RLVR) \citep{guo2025deepseek,jaech2024openai} have proven highly effective in enhancing instruction following, safety, and sophisticated reasoning capabilities, among other things, thereby positioning RL as a key enabler in the rapid advancement of modern generative artificial intelligence (AI).

A central objective of RL is to learn, from a collection of noisy data samples, a decision-making strategy (commonly referred to as a {\em policy} in RL) that approximately maximizes long-term rewards in an environment (often quantified through a {\em value function}), without direct access to an accurate description of the environment itself.  In many modern applications, the state space (i.e., the collection of possible situations an agent may encounter in the environment), the action space (i.e., the set of actions available to the agent at each decision point), and the planning horizon can all be extraordinarily large. As a result, data collection must often be carried out at a substantial scale, which may be prohibitively expensive, time-consuming, or even high-stakes, as encountered in domains such as clinical trials, online advertising, and autonomous systems. These challenges are particularly pronounced in the ``sample-starved'' regime, where the intrinsic complexity of the environment vastly exceeds the amount of available data, making accurate model estimation infeasible.\footnote{Here and throughout, the ``model'' refers collectively to the transition kernels and reward functions of the Markov decision process describing the environment, as formalized in Chapter~\ref{chap:mdp}.} Consequently, a fundamental goal of modern RL theory is to understand how reliable learning can be achieved in such data-limited scenarios. Accomplishing this goal requires developing a precise quantitative understanding of the trade-off between statistical accuracy and sample complexity, as well as designing computationally efficient algorithms that provably attain optimal or near-optimal performance.

%Here and throughout, the ``model'' refers collectively to the transition kernel and the reward function of the MDP.

%A core objective of RL is to search for a policy --- based on a collection of noisy data samples --- that approximately maximizes expected cumulative rewards in an MDP, without direct access to a precise description of the underlying model.\footnote{Here and throughout, the ``model'' refers to the transition kernel and the rewards of the MDP taken collectively.}    
%In contemporary applications, it is increasingly more common to encounter environments with prohibitively large state and action space, thus exacerbating the challenge of collecting enough samples, where data collection is expensive, time-consuming, or even high-stake (such as clinical trials, online advertisements, autonomous systems, to name just a few). To enable faithful policy learning in the sample-starved regime (i.e.~the regime where the model complexity overwhelms the sample size), it is crucial to obtain a quantitative picture of the fundamental trade-off between sample complexity and statistical accuracy, and to design efficient algorithms that provably achieve the optimal trade-off. 

%, such that the resulting performance is at most $\varepsilon$ away from optimal. Throughout the monograph, such a guarantee will be referred to as achieving ``$\varepsilon$-accuracy.''

% In this monograph, we seek to enrich the statistical foundations of reinforcement learning, aiming to address 

 \section{Statistical framework and scope}
 \label{sec:scope}

To systematically address the data-efficiency challenges outlined above, this monograph develops a coherent and rigorous statistical framework that accommodates a diverse range of settings. To help orient the reader, we briefly describe three key pillars underlying the topics studied throughout this monograph: the sampling mechanisms, the core algorithmic paradigms, and the performance metrics used to evaluate learning performance.

\paragraph{Data collection mechanisms.} 
The statistical challenges of RL depend fundamentally on how data are collected and made available to the agent. We emphasize three canonical settings that are widely studied in the literature.
\begin{itemize}
\item {\em RL with a generative model}. When equipped with access to a generative model or simulator, the agent can query the environment in a fully controlled manner. More specifically, the agent may sample transitions from arbitrary state-action pairs at will, effectively eliminating the need for exploration and isolating the statistical challenges inherent to policy optimization and value estimation. See Chapters~\ref{chapter:generative-model} and \ref{chapter:generative-model-MF}. 

\item {\em Online RL}. 
In online RL, the agent interacts with the environment sequentially in real time and must adaptively determine its actions based on past observations. Since data can only be gathered through the agent's own interactions with the environment, one must balance \emph{exploration} (i.e., acquiring new information about the unknown environment) and \emph{exploitation} (i.e., leveraging current knowledge of the environment to improve rewards). This fundamental tension leads to the classical exploration-exploitation trade-off, which lies at the heart of online RL. See Chapter~\ref{chapter:online-RL}. 

\item {\em Offline RL}. 
In offline RL, the agent is given access only to a static dataset of historical samples --- typically logged by one or more (possibly unknown) behavior policies --- and needs to learn without any further interaction with the environment. As a result, the agent faces fundamental challenges arising from issues such as limited data coverage and distributional mismatch between the available data and the target policy to be learned. See Chapter~\ref{chapter:offline-RL}.

\end{itemize}

\paragraph{Algorithmic paradigms.}
Broadly speaking, there are at least two principal algorithmic paradigms in RL: \emph{model-based} approaches and \emph{model-free} approaches.
\begin{itemize}
\item Model-based algorithms typically decouple model estimation from policy learning. Namely, one first leverages the available data samples to estimate the unknown environment model and subsequently performs policy optimization by treating the learned model as if it were accurate and applying classical dynamic programming techniques dating back to Richard Bellman~\citep{bellman1952theory,bertsekas2017dynamic}. A notable advantage of model-based methods is their flexibility: once a model has been learned, it can often be reused to solve new downstream tasks without requiring additional interaction with the environment or revisiting the original data samples. 

\item   

In contrast, model-free algorithms seek to learn the optimal value function or policy directly from data, bypassing explicit estimation of the environment's dynamics and rewards. These algorithms are particularly appealing when constructing or storing an environment model is difficult. Model-free methods are also often more memory-efficient because they avoid storing an explicit, potentially large-dimensional representation of the full model.
\end{itemize}

\noindent 
Characterizing the sample efficiency of model-based and model-free approaches is a primary theme of this monograph. As a preview, despite their higher memory and computational costs, model-based algorithms have been shown to attain statistically optimal performance across all three canonical sampling mechanisms outlined above, even in the most sample-starved regimes. By contrast, many classical model-free methods, including standard variants of Q-learning, fail to attain optimal sample complexity in some regimes. Whether model-free methods can achieve statistical optimality across the full spectrum of problem regimes without sacrificing their memory advantages remains an important open question.

\paragraph{Evaluation metrics.} 
In this monograph, we evaluate the performance of RL algorithms primarily through the lens of \emph{sample complexity}, defined within the classical \emph{Probably Approximately Correct (PAC)} learning framework as the minimum number of samples required to guarantee, with high probability, that the statistical error falls below a prescribed threshold. Depending on the problem goal, the notion of statistical error can be tailored to the specific object of interest. For value estimation, we assess the accuracy of the estimated value function, often measured under the $\ell_{\infty}$ norm to ensure uniform precision across all initial states. For policy learning, we instead measure the performance gap between the value function induced by the learned policy and that induced by the target policy. More broadly, such sample complexity guarantees provide a precise, quantitative characterization of how statistical accuracy improves as additional data become available.

Another performance metric widely used in online settings is (cumulative) \emph{regret}, which measures the gap between the cumulative reward that could have been achieved by following an optimal policy and the cumulative reward actually obtained by the learning algorithm over time. Under suitable conditions, regret guarantees can be converted into PAC-style sample complexity guarantees, a connection that we will elaborate on in Chapter~\ref{sec:formulation-online}.

Finally, we also briefly discuss the task of uncertainty quantification (see  Chapter~\ref{sec:UQ-RL}), with the goal of assessing the uncertainty associated with estimated quantities. A common objective is to construct confidence intervals that are both statistically valid and as tight as possible.

\section{Towards non-asymptotic statistical foundations} 
Despite decades of extensive research, the statistical foundations of RL in finite-sample and finite-time regimes --- which carry high operational value in modern practice --- 
remained significantly underdeveloped until the past one to two decades. Early theoretical analyses of core RL algorithms --- such as temporal difference learning and Q-learning --- were conducted primarily in the asymptotic regime, where the number of samples or iterations tends to infinity while all other problem parameters are held fixed. Although this line of work provided important insights into the convergence behavior of RL algorithms, it offers limited guidance on their data efficiency in modern large-scale applications. As discussed previously, many contemporary RL problems have to deal with enormous state-action spaces and long planning horizons. Treating these problem parameters as fixed constants can therefore fail to capture the statistical bottlenecks in high dimensions. 

To overcome these limitations, the past several years have witnessed a flurry of activity aimed at strengthening finite-sample, finite-time theory for RL. These advances draw heavily on ideas from modern development in high-dimensional statistics, high-dimensional probability, information theory, and statistical learning theory. Yet, despite this shift toward non-asymptotic analysis, a complete understanding of the fundamental sample complexity limits remained largely elusive, even for the three canonical data collection mechanisms introduced above. As we will discuss in subsequent chapters, much of the prior literature was, at best, only able to establish sharp or optimal statistical guarantees when the available sample size exceeded an enormous threshold, leaving the sample-starved regime poorly understood until fairly recently. Indeed, even today, sharp and complete statistical characterizations of the performance of RL algorithms are available only in relatively basic settings, while many important problems and settings remain far from fully resolved. 

Notably, RL in high-dimensional, sample-limited settings poses significant technical hurdles that call for both refined analysis techniques and new algorithmic ideas. In some sense, the paradigm shift from asymptotic to non-asymptotic RL theory is reminiscent of the rise of high-dimensional statistics in recent decades, where classical large-sample theory proved insufficient and ultimately motivated the development of fundamentally new mathematical techniques and algorithmic paradigms throughout modern statistics.

\section{Roadmap of this monograph}

In this monograph, we aim to present a coherent, non-asymptotic framework that covers core algorithmic principles as well as cutting-edge theoretical developments. 
In particular, we seek to provide sharp sample complexity characterizations for several RL algorithms, benchmarked against minimax lower bounds to assess their statistical optimality or suboptimality.

The bulk of this monograph is organized around the three sampling mechanisms introduced in Section~\ref{sec:scope}. We devote Chapters~\ref{chapter:generative-model} and \ref{chapter:generative-model-MF} to the generative-model setting, Chapter~\ref{chapter:online-RL} to online RL, and Chapter~\ref{chapter:offline-RL} to offline RL. In each setting, we present optimal or widely used algorithms alongside sharp statistical guarantees and minimax lower bounds. For the major results, we provide largely self-contained proofs supplemented by intermediate arguments that build intuition. Chapter~\ref{chapter:extensions} moves beyond these canonical single-agent formulations and briefly introduces several extensions of growing importance, including distributionally robust RL, multi-agent RL, federated RL, statistical inference, and uncertainty quantification.

To develop these statistical foundations in the cleanest possible settings, we focus predominantly on tabular MDPs, where the state and action spaces are finite, and no low-dimensional function approximation is assumed. Notably, despite decades of study, the fundamental minimax sample complexity limits for tabular MDPs under the three canonical sampling mechanisms remained significantly unresolved for a long time and have only been settled --- in several major, though not all, MDP models --- very recently. We believe that obtaining a comprehensive understanding of these tabular fundamental limits constitutes an indispensable first step toward elucidating and ultimately achieving optimal statistical efficiency in RL at large, particularly in settings involving function approximation.

%In addition to developing algorithms that achieve these fundamental limits, this monograph also establishes tight sample complexity bounds for several widely deployed classical algorithms, including TD learning and Q-learning. Crucially, characterizing the finite-sample behavior of these classical methods also remained an open problem until recently. 

\section{Relation to existing RL texts}

Several existing books and monographs are particularly relevant to, or complementary with, the present monograph, and we briefly discuss a few of them here.

For introductory material, the foundational textbooks \citet{sutton2018reinforcement,bertsekas2019reinforcement,szepesvari2010algorithms} provide excellent and comprehensive introductions to RL for readers with little or no prior background, albeit with much less emphasis on modern statistical foundations. 
The special cases of multi-armed bandits and contextual bandits have been covered in several books (e.g., \citet{bubeck2012regret,slivkins2019introduction,lattimore2020bandit}), with \citet{lattimore2020bandit} offering perhaps the most comprehensive and up-to-date treatment of modern bandit theory. 
The monograph \citet{agarwal2019reinforcement} has become a widely used resource for researchers entering RL theory, owing to its broad coverage of contemporary topics. A substantial portion of the present monograph instead focuses on advances in the statistical theory of tabular RL that appeared after \citet{agarwal2019reinforcement} first became available. These include substantial progress in characterizing and attaining sharp statistical limits for RL with a generative model, online RL, and offline RL, as well as in deriving sharp sample complexity guarantees for Q-learning.

We also note the monograph by \citet{foster2023foundations}, which develops a statistical perspective on interactive decision-making, with a primary focus on the general framework of function approximation. In addition, \citet{zhang2023mathematical} contains chapters on bandits and RL that briefly discuss several modern topics, including linear mixture MDPs and the Bellman--Eluder dimension. 
The recent book \citet{guo2026introduction} offers a comprehensive introduction to modern development of continuous RL from the perspective of stochastic control. The emphasis of these three texts, however, differs drastically from that of the present monograph, as none of them centers on the state-of-the-art statistical development in tabular RL.

Moreover, the present monograph grew out of several tutorials and courses taught by some of the authors in recent years, which also led to lecture notes such as \citet{chi2025statistical}. In comparison, \citet{chi2025statistical} offers a bird's-eye view of the recent statistical and algorithmic advances in RL, including developments in policy optimization, whereas the present monograph focuses more squarely on the statistical foundations of RL, offering a thorough and in-depth treatment accompanied by largely self-contained mathematical proofs.

\section{Notation}
\label{sec:notation}

For clarity and consistency, we introduce below a set of notation and conventions to be used throughout this monograph. Additional notation specific to individual chapters will be introduced later as needed.

\paragraph{Asymptotic notation.}
Let $\mathcal{X}\coloneqq \big( S, A, \frac{1}{1-\gamma}, H, \frac{1}{\varepsilon}, \frac{1}{\delta} \big)$, where the meanings of these parameters will become clear shortly. The notation $f(\mathcal{X}) = O(g(\mathcal{X}))$ or $f(\mathcal{X}) \lesssim g(\mathcal{X})$ means there exists a universal constant $C_1>0$ such that $f(\mathcal{X}) \leq C_1 g(\mathcal{X})$. Similarly, $f(\mathcal{X}) = \Omega(g(\mathcal{X}))$ or $f(\mathcal{X}) \gtrsim g(\mathcal{X})$ means that $g(\mathcal{X}) = O(f(\mathcal{X}))$, while $f(\mathcal{X}) = \Theta(g(\mathcal{X}))$ or $f(\mathcal{X}) \asymp g(\mathcal{X})$ indicates that both $f(\mathcal{X}) = O(g(\mathcal{X}))$ and $g(\mathcal{X}) = O(f(\mathcal{X}))$ hold simultaneously. Additionally, the notation $\widetilde{O}(\cdot)$ (resp.\ $\widetilde{\Omega}(\cdot)$ and $\widetilde{\Theta}(\cdot)$) is defined analogously to $O(\cdot)$ (resp.\ $\Omega(\cdot)$ and $\Theta(\cdot)$) but with logarithmic factors suppressed.

\paragraph{Vectors and matrices.}
For any vector \(a=[a_i]_{1\le i\le n}\in \mathbb{R}^n\), we overload several scalar operators in an entrywise manner. For instance, we write
$$\sqrt{a}\coloneqq[\sqrt{a_i}]_{1\leq i\leq n},\quad  |a|\coloneqq[|a_i|]_{1\leq i\leq n}, \quad \text{and}\quad  a^2 \coloneqq [a_i^2]_{1\leq i\leq n}.$$
For vectors $a=[a_i]_{1\leq i\leq n}$ and $b=[b_i]_{1\leq i\leq n}$, the notation $a \geq b$ (resp.~$a \leq b$) means that $a_i \geq b_i$ (resp.~$a_i\leq b_i$) for all $1\leq i\leq n$. We use $$a \circ b \coloneqq [a_ib_i]_{1\leq i\leq n}$$ to represent the Hadamard product of $a$ and $b$.  We also let $\|A\|$ and $\mathsf{Tr}(A)$ denote the spectral norm and the trace of a matrix $A$, respectively. 

For any finite set $\mathcal{S}=\{1,\ldots,S\}$, we let $\Delta(\mathcal{S})=\{p\in \mathbb{R}^S \mid p \geq 0, 1^{\top}p = 1\}$ represent the set of probability vectors (i.e., the probability simplex) over $\mathcal{S}$. 
For any matrix $P$, we define its $\ell_1$ norm (i.e., the maximum row-wise $\ell_1$ norm) as 
\begin{align}
    \| P \|_1 \coloneqq \max_{i} \sum\nolimits_{j} |P_{i,j}|.
    \label{eq:defn-L1-norm}
\end{align}

\paragraph{Probability divergences.}
For any two probability measures $P$ and $Q$ defined on the same measurable space $(\Omega,\mathcal{F})$, the total-variation (TV) distance between them is defined as
\begin{subequations}
\label{eq:defn-TV-KL-chi2}
\begin{equation} 
\label{eq:TV}
	\mathsf{TV} (P, Q) \coloneqq 
    %\frac{1}{2} \sum_{x \in \mathcal{X}} \big|\mu(x) - \nu(x) \big| =
	 \sup_{A\in \mathcal{F}} \big| P(A) - Q(A) \big|.
\end{equation}
When $P$ is absolutely continuous w.r.t.~$Q$,  the Kullback-Leibler (KL) divergence and $\chi^2$ divergence between them are defined respectively by
\begin{align}
\mathsf{KL}(P\,\|\,Q) &\coloneqq \displaystyle \int \bigg(\log \frac{\mathrm{d}P}{\mathrm{d}Q} \bigg) \mathrm{d} P, \label{eq:defn-KL}\\
\chi^2(P\,\|\,Q) &\coloneqq \displaystyle \int \bigg( \frac{\mathrm{d}P}{\mathrm{d}Q} -1\bigg)^2 \mathrm{d} Q.
\label{eq:defn-chi2}
\end{align}
\end{subequations}
These divergences play a central role in mathematical statistics, information theory, and probability theory as measures of the ``statistical distance'' between probability distributions; see, e.g., \citet{tsybakov2009introduction,polyanskiy2025information}, for in-depth discussions. These divergence metrics are also related through standard inequalities; for instance, it is well known that \citep{tsybakov2009introduction}
\begin{align}
2\,\mathsf{TV}^2(P,Q)
\leq
\mathsf{KL}(P\,\|\,Q) 
\leq \chi^2(P\,\|\,Q). 
\label{eq:KL-ub-chi2}
\end{align}

\paragraph{Miscellaneous.}
For any positive integer $n$, we denote $[n]\coloneqq \{1,\dots,n\}$. 
We use $e_i$ to represent the $i$-th standard basis vector. 
The notation $\ind\{\cdot\}$ stands for the indicator function.

%For any probability vector $q\in \mathbb{R}^{1\times S}$ (which is a row vector) and any vector $V \in \mathbb{R}^S$,   define
%
%\begin{align}
%	\mathsf{Var}_{q} (V)\coloneqq q  \big(V \circ V \big)-\left(q V \right)^{2} \,\in \mathbb{R}
%	\label{eq:defn-Var-q-V}
%\end{align}
%
%with $qV=\sum_i q_iV_i$, which corresponds to the variance of $V_{s}$ when $s$ is sampled according to the distribution specified by $q$.

%Finally, let us introduce the concept of uniform ergodicity for Markov chains. 
%Consider any Markov chain $(X_0, X_1, X_2, \cdots)$ with transition kernel $P$, finite state space $\mathcal{X}$ and stationary distribution $\mu$, 
%	and denote by $P^t(\cdot\,|\,x)$  the distribution of $X_t$ conditioned on $X_0=x \in \mathcal{X}$. 
%	This Markov chain is said to be {\em uniformly ergodic} if, for some $\rho < 1$ and $M < \infty$, one has
%
%\begin{equation}
%\sup_{x \in \mathcal{X}} d_{\mathsf{TV}} \big(\mu, P^t(\cdot\,|\,x) \big) \le M\rho^t,
%\end{equation}
%
%	where $d_{\mathsf{TV}} (\mu, \nu)$ stands for the total variation distance between two distributions $\mu$ and $\nu$ \citep{tsybakov2009introduction}:  

%% file: chapters/prelim.tex
\chapter{Preliminaries}
\label{chap:mdp}

We begin by reviewing the fundamentals of Markov decision processes (MDPs), the canonical framework for developing and analyzing reinforcement learning algorithms. 
This framework sets the stage for the discussions in the chapters that follow. We focus on two widely studied formulations: discounted infinite-horizon MDPs and finite-horizon episodic MDPs. Additionally, we present core concentration inequalities and techniques for deriving minimax lower bounds for multi-hypothesis testing, which serve as the primary analytical tools for this work.

\section{Discounted infinite-horizon MDPs}
\label{sec:prelim-inf-horizon-MDP}

We first introduce the basics and notation for discounted infinite-horizon MDPs. 
More comprehensive introductions can be found in classical references such as \citet{puterman2014markov,bertsekas2017dynamic}.

\subsection{Basic components}

%0.5\textwidth

\begin{figure}[t]
\begin{center} 
\includegraphics[width=0.35\textwidth]{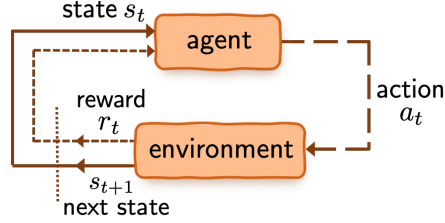}  
\end{center}	
	\caption{An illustration of an MDP, where an agent interacts sequentially with the environment. At time step $t$, the agent observes state $s_t$ of the environment, selects action $a_t$, and receives an immediate reward $r_t$. The environment then transitions to state $s_{t+1}$ according to the transition kernel $P$. 	\label{fig:MDP}}

\end{figure}

A discounted infinite-horizon MDP can be represented by a tuple $\mathcal{M} = \{\cS, \cA, P, r, \gamma\}$, which models an agent's sequential interaction with its environment over an unbounded time horizon (see \Cref{fig:MDP} for an illustration). Its key components are: 
\begin{itemize}
    \item[(i)] $\cS = \{1,\dots, S\}$: a finite state space consisting of $S$ distinct states. A state can be used to describe, for example, the agent's current situation (e.g., the position of a self-driving vehicle, the health condition of a patient in a treatment model). 
    
    \item[(ii)] $\cA = \{1, \dots, A\}$:  an action space consisting of $A$ distinct actions available to the agent (e.g., the direction of movement in a navigation task, the choice of medication to prescribe to a patient).  
    
    \item[(iii)] $P: \cS\times \cA \rightarrow \Delta(\cS)$: the transition kernel of the MDP, where $P (\cdot \mymid s,a)$ specifies the probability distribution over the next state when action $a$ is executed in state $s$.

    \item[(iv)] $r: \cS\times \cA \rightarrow [0,1]$: a deterministic reward function, where $r(s,a)$ indicates the immediate reward received by the agent for taking action $a$ in state $s$. Rewards are normalized to lie within $[0,1]$ without loss of generality. To streamline presentation, we assume throughout this monograph that the reward function is fully known to the learner {\em a priori}. Note, however, that this assumption is not essential and can often be substantially relaxed.

    \item[(v)] $\gamma \in [0,1)$: the discount factor, which guarantees that the infinite-horizon cumulative discounted reward is always finite (as we shall see shortly). The quantity $\frac{1}{1-\gamma}$ is commonly referred to as the {\em effective horizon} of $\mathcal{M}$: it is proportional to the number of steps needed for a reward to be discounted to half its value, and hence reflects the time scale over which future rewards remain significant.

\end{itemize}

\noindent 
Here and throughout, we assume the discounted infinite-horizon MDP to be \textit{stationary} (or \textit{time-homogeneous}), meaning that both the transition kernel $P$ and the reward function $r$ are time-invariant. In addition, we introduce the following notation for convenience:
\begin{align}
	P_{s,a} \coloneqq P(\cdot \mymid s,a) \in \mathbb{R}^{1\times S}. 
\end{align}

Throughout this monograph, we focus on regimes in which the number of states $S$, the number of actions $A$, and the effective horizon $\frac{1}{1-\gamma}$ are all large. This emphasis reflects a common challenge arising in modern RL practice, where algorithms must contend with vast state–action spaces and long horizons.

\subsection{Policy, value function, Q-function, and occupancy distribution}

At each time step, the agent selects an action based on the current state of the environment. Such an action-selection rule is called a policy. Formally, a stationary policy $\pi: \cS \rightarrow \Delta(\cA)$ is a mapping that assigns to each state $s \in \cS$ a probability distribution over actions; namely,  $\pi(a \mymid s)$ indicates the probability of selecting action $a$ in state $s$. Clearly, this definition allows policies to be randomized.  
When $\pi$ is a deterministic policy, we often abuse the notation by letting $\pi(s)$ represent the action selected by policy $\pi$ in state $s$. 
We further denote $P_{\pi}: \mathcal{S}\rightarrow \Delta(\mathcal{S})$ the equivalent transition probability when action $a$ is drawn from $\pi(\cdot\mymid s)$, namely, 
\begin{align}
\forall s \in \cS:\qquad
P_{\pi}(\cdot \mymid s) \coloneqq \mathop{\mathbb{E}}_{a\sim \pi(\cdot\mymid s)}[P(\cdot \mymid s, a)]. 
\label{eq:defn-P-s-pi-inf}
\end{align}
As a remark, for the stationary infinite-horizon MDPs considered here, it is typically sufficient to restrict attention to time-invariant policies.

%0.78\textwidth

\begin{figure}[t]
	\begin{center}	\includegraphics[width=0.55\textwidth]{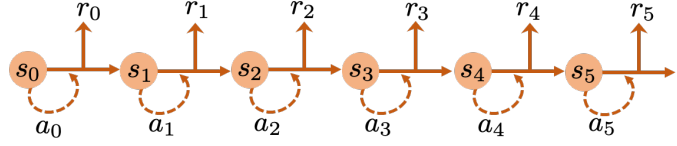} 
	\end{center}	
	\caption{An illustration of a trajectory rolled out by an MDP, where $r_t = r(s_t, a_t)$ is the immediate reward gained at time step $t$. 
		\label{fig:MDP_trajectories}}
\end{figure}

To evaluate a policy, we introduce two central performance measures (or ``utility functions'') in RL: the value function and the Q-function. Consider a sample trajectory $\{(s_t,a_t)\}_{t\geq 0}$ generated by the MDP under policy  $\pi$, which selects action $a_t$ after observing the state $s_t$ at time step $t$ (see Figure~\ref{fig:MDP_trajectories} for an illustration).  Note that the trajectory depends on the policy $\pi$ as the choice of action $a_t$ affects the next state $s_{t+1}$.  We now define the value function and the Q-function associated with this policy.
\begin{itemize}

\item {\em Value function.}
The value function $V^\pi: \cS \rightarrow \mathbb{R}$ of policy $\pi$ is defined as the expected discounted cumulative reward:
\begin{align}
	\forall s\in\cS: \quad 
	V^\pi(s) \coloneqq \mathbb{E} \left[\sum_{t=0}^\infty \gamma^t r(s_t, a_t) \mid s_0 = s \right],
	\label{eq:defn-Vpi-inf}
\end{align}
where the expectation is over the sample trajectory $\{(s_t, a_t)\}_{t\geq 0}$ generated under policy $\pi$, with $a_t \sim \pi(\cdot \mymid s_t)$ and $s_{t+1} \sim P(\cdot \mymid s_t, a_t)$ for all $t\geq 0$, starting from an initial state $s_0 = s$. 
Intuitively, $V^\pi(s)$ represents the long-term reward an agent can expect when starting from state $s$ and following policy $\pi$, with future rewards geometrically discounted over time. 
Since immediate rewards are assumed to lie in $[0,1]$, it follows that 
$$0\leq V^\pi(s) \leq \frac{1}{1-\gamma}$$ 
for every policy $\pi$. 
 Moreover, if the initial state is randomly drawn from some state distribution $\rho \in \Delta(\cS)$, then we can also define the value function of policy $\pi$ under initial state distribution $\rho$ as:
\begin{align}
	V^\pi(\rho) \coloneqq \mathop{\mathbb{E}}\limits_{s\sim \rho}\big[V^\pi(s)\big].
\end{align}

\item {\em Q-function.}
Analogously, the Q-function (or action-value function) $Q^\pi: \cS\times \cA \rightarrow \mathbb{R}$ of policy $\pi$ can be defined such that: for every $(s,a)\in \cS\times \cA$, 
\begin{align}
	%\forall (s,a)\in\cS \times \cA: \quad 
	Q^\pi(s,a) \coloneqq \mathbb{E} \left[\sum_{t=0}^\infty \gamma^t r(s_t, a_t) \mid s_0 = s, a_0 = a \right].
    \label{eq:defn-Qpi-inf-horizon}
\end{align}
In contrast to the value function, the Q-function conditions on not only the initial state $s_0 = s$, but also on the initial action $a_0 = a$, with subsequent actions drawn from $\pi$. It therefore quantifies the expected long-term return by first taking action $a$ in state $s$ and then following policy $\pi$ thereafter. This function provides a natural way to compare the relative benefits of different actions in a given state. 
The intimate connection between the Q-function and the value function is reflected in the following identity: 
\begin{equation}
V^{\pi}(s) = \mathop{\mathbb{E}}\limits_{a \sim \pi(\cdot\mymid s) }  \left[Q^{\pi}(s, a) \right].
\label{eq:Vpi-a-function-of-Qpi-inf}
\end{equation} 

\end{itemize}

As is well known, there exists at least one deterministic policy, denoted by $\pi^\star$ throughout, that is optimal in the sense that it simultaneously maximizes $V^\pi(s)$ and $Q^\pi(s,a)$ for all state-action pairs $(s,a)\in \cS\times \cA$ (see, e.g., \citet{bertsekas2017dynamic}). We adopt the following shorthand notation for the corresponding optimal value function and optimal Q-function:
\begin{align}
	\forall (s,a) \in \cS\times \cA: ~~ V^\star(s) \coloneqq V^{\pi^\star}(s) \; \text{ and } \; Q^\star(s,a) \coloneqq Q^{\pi^\star}(s,a).
    \label{eq:optimal-Vstar-Q-star-defn}
\end{align}
%

%\paragraph{Discounted occupancy distribution.} 
Next, we find it helpful to also introduce the {\em discounted occupancy distributions} associated with policy $\pi$, 
quantifying the frequencies with which each state (or state-action pair) is visited under policy $\pi$. Given an initial state distribution $\rho \in \Delta(\mathcal{S})$, define the state occupancy distribution $d^{\pi}(s;\rho)$ and the state-action occupancy distribution $d^{\pi}(s,a; \rho)$ such that: for every $(s,a)\in \cS\times \cA$, 
\begin{subequations}
\label{eq:defn-dpi-occupancy-discounted-dist}
\begin{align}
	%\forall s\in \cS: \quad 
    d^\pi(s; \rho) &\coloneqq (1-\gamma) \sum_{t=0}^\infty \gamma^t \mathbb{P}(s_t = s \mid s_0 \sim \rho),\\
	%\forall (s,a)\in \cS \times \cA: \quad 
    d^\pi(s,a; \rho) &\coloneqq (1-\gamma) \sum_{t=0}^\infty \gamma^t \mathbb{P}(s_t = s, a_t =a \mid s_0 \sim \rho),
\end{align}
\end{subequations}
where the sample trajectory starts from  $s_0 \sim \rho$ and  follows policy $\pi$ (i.e., $a_t \sim \pi(\cdot \mymid s_t)$ and $s_{t+1} \sim P(\cdot \mymid s_t, a_t)$ for all $t\geq 0$). 
In words, these functions describe the visitation frequencies of states and state-action pairs --- with each visit discounted geometrically over time --- under a policy over an infinite horizon.  The prefactor $1 - \gamma$ is the reciprocal of the total weight, ensuring that these distributions are normalized. 

Correspondingly, we adopt the notation of the discounted occupancy distributions associated with the optimal deterministic policy $\pi^\star$, denoted by $d^{\star}$, as follows:
\begin{subequations}
\label{eq:defn-dpi-occupancy-discounted-dist-opt}
\begin{align}
	%\forall (s,a) \in \cS\times \cA: \quad 
    d^\star(s; \rho) & \coloneqq d^{\pi^\star}(s; \rho) 
    \qquad\qquad \text{ and } \\
 d^\star (s,a; \rho) & \coloneqq d^{\pi^\star}(s,a; \rho) = d^{\star}(s; \rho) \ind\big( a = \pi^\star(s) \big) \label{eq:infinite-d-star-def}
\end{align}
\end{subequations}
for every $(s,a) \in \cS\times \cA$,
where the last equality in \eqref{eq:infinite-d-star-def} is valid since $\pi^\star$ is assumed to be deterministic. Here, we dropped the dependency on $\rho$ for notational simplicity.

\subsection{Bellman operators and Bellman equations}

Central to the design and analysis of reinforcement learning algorithms are the Bellman operators and the Bellman equations, named after Richard Bellman \citep{bellman1952theory},  which we now introduce.

\paragraph{Bellman consistency operator.} 
Given a policy $\pi$, 
the Bellman consistency operator $\mathcal{T}^\pi: \mathbb{R}^{ S \times A} \rightarrow \mathbb{R}^{ S \times A}$  is defined such that, for any input $Q\in \mathbb{R}^{S\times A}$, the $(s,a)$-th entry of the output  $\mathcal{T}^{\pi}(Q)$ is given by
\begin{equation}\label{eq:bellman_consistency_operator}
	\mathcal{T}^{\pi}(Q)(s,a) \coloneqq r(s,a) + \gamma \mathop{\mathbb{E}}_{s^{\prime} \sim P(\cdot\mymid s,a)}  \Big[ \mathop{\mathbb{E}}\limits_{a^{\prime} \sim \pi(\cdot \mymid  s^{\prime}) }  \big[Q(s^{\prime}, a^{\prime}) \big] \Big].
\end{equation}
Intuitively, the operator $\mathcal{T}^\pi$ performs a one-step lookahead: it evaluates a state–action pair $(s,a)$ by adding the immediate reward $r(s,a)$ to the discounted value of the next state–action pair, assuming that future actions are drawn from policy $\pi$.

It follows directly that the Q-function of policy $\pi$ (cf.~\eqref{eq:defn-Qpi-inf-horizon}) is a fixed point of this Bellman consistency operator, satisfying
\begin{equation}\label{eq:bellman_consistency}
  Q^{\pi} = \mathcal{T}^{\pi}(Q^{\pi}).
\end{equation}
The solution is also unique  (see \citet{bertsekas2017dynamic} and also \eqref{eq:gamma-contraction-preliminary-inf}).
Combined with the relation \eqref{eq:Vpi-a-function-of-Qpi-inf} between $V^{\pi}$ and $Q^{\pi}$, this further implies that the value function satisfies
\begin{subequations}
\label{eq:bellman_consistency-V-all}
\begin{equation}
\label{eq:bellman_consistency-V}
   V^{\pi} = \mathcal{T}_{\pi}(V^{\pi}) , 
\end{equation}
where the operator $\mathcal{T}_{\pi}: \mathbb{R}^S\rightarrow \mathbb{R}^S$ is defined such that for any $V\in \mathbb{R}^S$, the $s$-th entry of $\mathcal{T}_{\pi}(V)$ is given by
\begin{equation}
    \mathcal{T}_{\pi}(V)(s) = \mathop{\mathbb{E}}\limits_{a\sim \pi(\cdot\mid s)}[r(s,a)] + \gamma \mathop{\mathbb{E}}\limits_{s'\sim P(\cdot \mid s,\pi)}\big[ V(s') \big].
\end{equation}
\end{subequations}
Since the Bellman equations in \eqref{eq:bellman_consistency} and
\eqref{eq:bellman_consistency-V} form systems of linear equations in
$Q^\pi$ and $V^\pi$, respectively, they can be solved to evaluate a
given policy, provided that the transition kernel $P$ and reward
function $r$ are known {\em a priori}.

\paragraph{Bellman optimality operator.} 
We now introduce another fundamental operator, the Bellman optimality operator $\mathcal{T}$, which is also a mapping from  $\mathbb{R}^{S \times A}$ to itself.  Concretely, given any input  $Q\in \mathbb{R}^{S\times A}$, the $(s,a)$-th entry of the output $\mathcal{T}(Q)$ is defined as
\begin{equation}\label{eq:bellman_optimality_operator}
	\mathcal{T}(Q)(s,a) \coloneqq r(s,a) + \gamma \mathop{\mathbb{E}}\limits_{s^{\prime} \sim P(\cdot\mymid s,a)}  \Big[ \max_{a^{\prime}\in \cA} Q(s^{\prime}, a^{\prime}) \Big].
\end{equation}
In words, $\mathcal{T}(Q)(s,a)$ combines the immediate reward with the discounted expected return of the best possible action in the next state $s'$. 
A fundamental property of the Bellman optimality  operator is that the optimal Q-function $Q^{\star}$ (cf.~\eqref{eq:optimal-Vstar-Q-star-defn}) is its unique fixed point \citep{bertsekas2017dynamic}, satisfying
\begin{equation}\label{eq:bellman_fix_point}
  Q^{\star} = \mathcal{T}(Q^{\star}).
\end{equation}
As we shall see momentarily, this celebrated result underpins the design of various algorithms like Q-learning.

\paragraph{Contraction of Bellman operators.}
A remarkable property of the above Bellman operators is $\gamma$-contraction: specifically, for any $Q_1, Q_2\in\mathbb{R}^{S\times A}$, it holds that \citep{bertsekas2017dynamic}
\begin{subequations}
\begin{align}
\| \mathcal{T}^{\pi}(Q_1) -\mathcal{T}^{\pi}(Q_2) \|_{\infty} & \leq \gamma \, \|  Q_1 - Q_2 \|_{\infty} , \\
\| \mathcal{T}(Q_1) -\mathcal{T}(Q_2) \|_{\infty} & \leq \gamma \,  \|  Q_1 - Q_2 \|_{\infty}.
\end{align}
\label{eq:gamma-contraction-preliminary-inf}
\end{subequations}
\noindent  
In other words, applying either the Bellman consistency operator or the Bellman optimality operator brings two Q-functions strictly closer under the $\ell_{\infty}$ norm. This contraction property is fundamental: it ensures the uniqueness of their fixed points and enables rapid convergence of many iterative algorithms in dynamic programming and RL.

\begin{proof}[Proof of $\gamma$-contraction \eqref{eq:gamma-contraction-preliminary-inf}]
Regarding $\mathcal{T}^{\pi}$, recall the basic fact $$\big|\mathbb{E}[Q_1(s,a)-Q_2(s,a)] \big|\leq \max_{s',a'}|Q_1(s',a')-Q_2(s',a')\big|$$
for any $(s,a)$. This combined with the definition \eqref{eq:bellman_consistency_operator} yields
\begin{align*}
\big|\mathcal{T}^{\pi}(Q_{1})(s,a)-\mathcal{T}^{\pi}(Q_{2})(s,a)\big| & =\gamma\big|\mathbb{E}\big[Q_{1}(s^{\prime},a^{\prime})-Q_{2}(s^{\prime},a^{\prime})\big]\big| \\
&\leq\gamma\big\| Q_{1}-Q_{2}\big\|_{\infty}
\end{align*}
for any $(s,a)$ pair, where the expectation is over $s^{\prime}\sim P(\cdot\mymid s,a)$ and $a^{\prime}\sim\pi(\cdot\mymid s^{\prime})$. This establishes the $\gamma$-contraction property of $\mathcal{T}^{\pi}$. 

The contraction for $\mathcal{T}$ can be proven by repeating the above arguments along with the basic fact that 
$$ \big|\max_aQ_1(s,a)- \max_aQ_2(s,a) \big| \leq \max_a \big|Q_1(s,a)-Q_2(s,a)\big|.$$
Details are omitted here for brevity. 
\end{proof}

\subsection{Dynamic programming under full model specification}
\label{sec:prelim-DP-inf}

With the Bellman operators in place, we are now ready to introduce the classical dynamic programming algorithms for solving discounted infinite-horizon MDPs, assuming a fully known model (i.e., both the transition kernel $P$ and the rewards  $r$ are available to the learner). The interested reader is referred to \citet{bertsekas2017dynamic} for in-depth discussions.

\begin{figure}[t]
	\begin{center} 
		\includegraphics[width=0.55\textwidth]{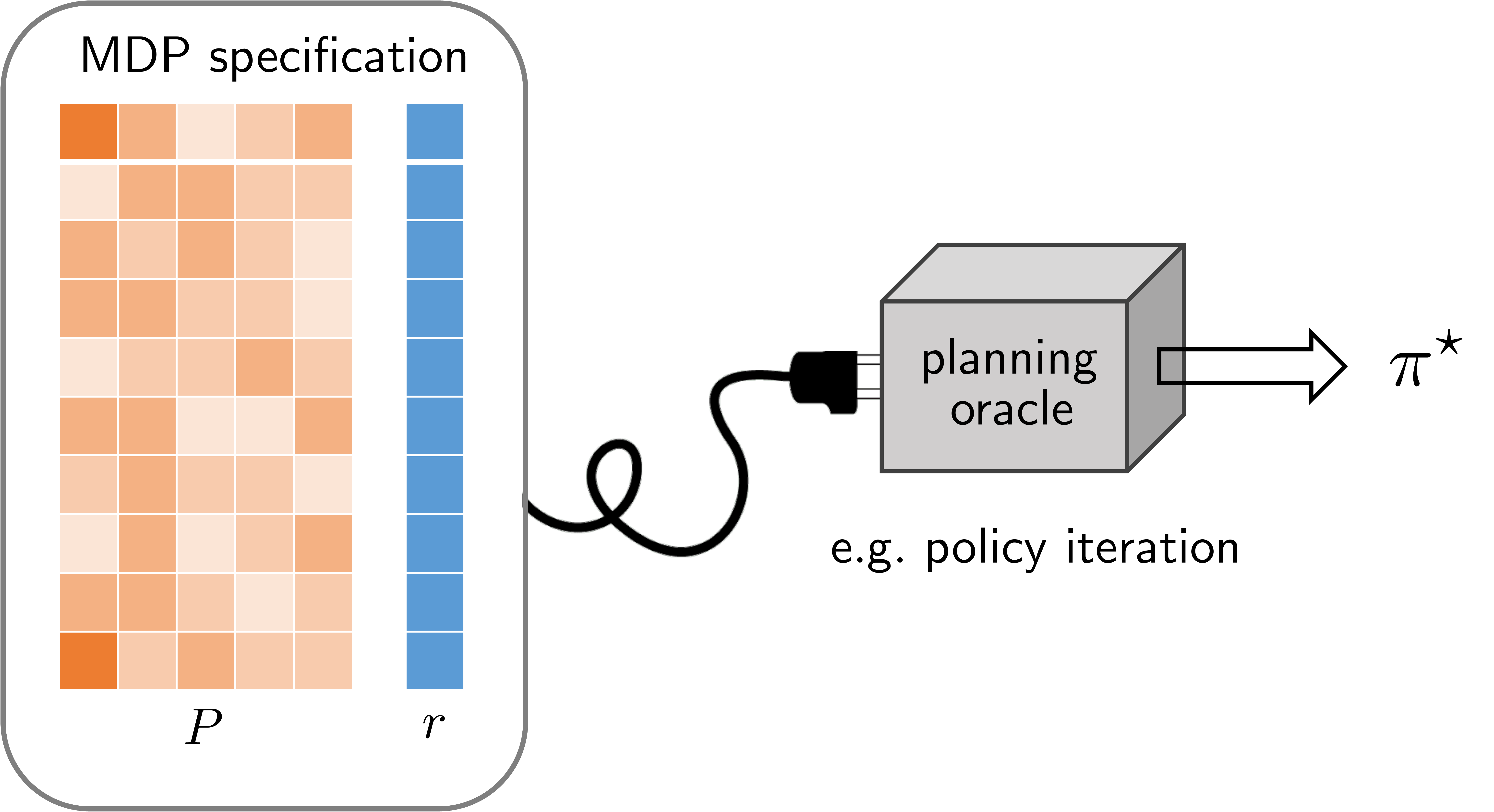}  
	\end{center}	
	\caption{Planning in an MDP with a fully specified model. Classical dynamic algorithms, such as policy iteration and value iteration,  can efficiently compute the optimal policy $\pi^{\star}$ and the associated optimal value given complete knowledge of $P$ and $r$. 	\label{fig:planning}}
	
\end{figure}

\paragraph{Planning.} 
Assuming access to exact model specification (i.e., $P$ and $r$), we describe two classical algorithms for computing the optimal Q-function $Q^{\star}$ and the optimal policy $\pi^{\star}$, a task commonly referred to as {\em planning} (see Figure~\ref{fig:planning}).

%0.6\textwidth

\begin{itemize}

\item {\em Value iteration (\texttt{VI}).} Initialize the Q-function estimate $Q_0$. For each iteration $k=1,2,\ldots$, update the Q-function as
\begin{equation} \label{eq:value_iteration}
Q_k = \mathcal{T}(Q_{k-1}),
\end{equation}
with $\mathcal{T}$ the Bellman optimality operator (cf.~\eqref{eq:bellman_optimality_operator}). 
In words,  the Q-function estimate is refined iteratively by repeatedly applying the Bellman optimality operator to the latest Q-function estimate.

\item {\em Policy iteration (\texttt{PI}).} Initialize the policy estimate $\pi_0$. For each iteration $k=1,2,\ldots$, 
%evaluate $Q^{k} = Q^{\pi_k}$,  and 
update the policy as
\begin{equation} \label{eq:policy_iteration}
\forall s\in \cS: \quad \pi_{k} (s) = \arg\max_{a \in \cA} Q^{\pi_{k-1}} (s,a).
\end{equation}
Namely, the policy is updated by acting greedily w.r.t.~the Q-function of the current policy estimate.

\end{itemize}

Encouragingly, both algorithms achieve fast linear convergence to the optimal Q-function, as asserted by the following lemma.
\begin{lemma}[Linear convergence of \texttt{VI} and \texttt{PI}]
Both value iteration (cf.~\eqref{eq:value_iteration}) and policy iteration (cf.~\eqref{eq:policy_iteration}) achieve
$$ \| Q_k - Q^{\star}\|_{\infty} \leq \gamma^k  \| Q_0 - Q^{\star}\|_{\infty} $$
for every $k \geq 1$, whereas in policy iteration we take $Q_k=Q^{\pi_k}$. 
\end{lemma} 
\begin{proof} 
%
%Here, we only provide the proof for VI. 
In light of the fixed-point property~\eqref{eq:bellman_fix_point} and $\gamma$-contraction~\eqref{eq:gamma-contraction-preliminary-inf} of $\mathcal{T}$, the \texttt{VI} iterates obey
\[
\big\| Q_{k}-Q^{\star}\big\|_{\infty}=\big\|\mathcal{T}(Q_{k-1})-\mathcal{T}(Q^{\star})\big\|_{\infty}\leq\gamma\big\| Q_{k-1}-Q^{\star}\big\|_{\infty}.
%\leq\dots\leq\gamma^{k}\big\| Q_{0}-Q^{\star}\big\|_{\infty}
\]
Applying this inequality iteratively completes the proof for \texttt{VI}. The proof for \texttt{PI} is omitted here and can be found in  \cite{agarwal2019reinforcement}. 
\end{proof}

This lemma reveals that, to achieve $\varepsilon$-accuracy in the sense that $\| Q_k - Q^{\star}\|_{\infty}\leq \varepsilon $, it takes no more than
$$  \frac{1}{1-\gamma}  \log \left( \frac{\| Q_{0} - Q^{\star}\|_{\infty}}{\varepsilon }\right) $$
iterations. Importantly, this iteration complexity is on the order of the effective horizon $\frac{1}{1-\gamma} $ (up to some logarithmic factor), irrespective of the dimension of the state-action space.

\subsection{Vector and matrix notation} \label{sec:vector-matrix-notation}

It is also convenient to introduce several pieces of vector and matrix notation, which allow our results and analysis to be presented more compactly.
%
% \paragraph{Vector and matrix notation.}
% Let us introduce a set of vector and matrix notation below. 
%
Here and throughout, we allow a slight abuse of notation by letting the same symbols represent both vectors/matrices and functions, whenever it is clear from context  (e.g., the notation $r$ may denote either the reward function or the corresponding reward vector).

\begin{itemize}
	\item $r \in \mathbb{R}^{SA}$: a vector representing the reward function, so that its $(s,a)$-th entry is the corresponding immediate reward.
    
	\item $V^{\pi}, V^{\star} \in \mathbb{R}^{S}$:  vectors representing respectively the value function of policy $\pi$ (cf.~\eqref{eq:defn-Vpi-inf}) and the optimal value function (cf.~\eqref{eq:optimal-Vstar-Q-star-defn}).
    
	\item $Q^{\pi}, Q^{\star} \in \mathbb{R}^{SA}$: vectors representing respectively the Q-function $Q^{\pi}$ of policy $\pi$ (cf.~\eqref{eq:defn-Qpi-inf-horizon}) and the optimal Q-function (cf.~\eqref{eq:optimal-Vstar-Q-star-defn}).

	%\item  $V^{\star} \in \mathbb{R}^{S}$ and $Q^{\star} \in \mathbb{R}^{SA}$: representing the optimal value function $V^{\star}$ and optimal Q-function $Q^{\star}$.
    
	\item $P\in \mathbb{R}^{SA\times S}$: a matrix representing the transition kernel, where its $(s,a)$-th row  is a probability vector representing $P(\cdot \mymid s,a)$. 

    \item $\pi(\cdot \mymid s) \in \mathbb{R}^{1\times A}$: a row vector $[\pi(a\mymid s)]_{1\leq a\leq A}$ representing the action distribution induced by policy $\pi$ in state $s$.
    
	\item $\Pi^{\pi} \in \mathbb{R}^{S \times SA}$: a matrix associated with  policy $\pi$ taking the following form
\begin{align}
\label{eqn:bigpi}
	\bPi^{\pi} = 
	\begin{pmatrix}
		\pi(\cdot\mymid 1) &            &  \\
		           & \pi(\cdot\mymid 2) &  \\
		           &            & \ddots \\	
		           &            &         & \pi(\cdot\mymid S)
	\end{pmatrix}  .
\end{align}
%
   % where $\pi(\cdot\mymid s) \in \mathbb{R}^{1\times A}$ denotes the row vector $[\pi(a\mymid s)]_{1\leq a\leq A}$. 
    
	\item $P^{\pi} \in \mathbb{R}^{SA\times SA}$ and $P_{\pi} \in \mathbb{R}^{S \times S }$: two {\em square} probability transition matrices induced by policy $\pi$ over the state-action pairs and the states, respectively, defined by
	\begin{equation}
	\label{eqn:ppivq}
		\Pq \defn \bP \bPi^{\pi} \qquad \text{and} \qquad \Pv \defn \bPi^{\pi} \bP .
	\end{equation}

\item $r_{\pi} \in \mathbb{R}^{S}$: a reward vector associated with policy $\pi$; namely,  
\begin{align}
\label{eq:defn-rpi-inf}
r_{\pi}(s) = \mathop{\mathbb{E}}_{a\sim \pi(\cdot\mymid s)}[r(s,a)]
\qquad \text{for all }s\in \cS,
\end{align}
%
%for all $s\in \cS$, 
or simply, $r_{\pi}= \Pi^{\pi}r$. 
\end{itemize}

We further introduce several variance-related objects, which shall be used frequently in our statistical analysis. 
For any vector $V=[V_i]_{1\leq i\leq S}\in\real^{S}$, we define the vector $\Var_{P}(V) \in \real^{SA}$, whose $(s,a)$-th entry is given by 
\begin{align}
	%\forall (s,a)\in \cS \times \cA, \qquad 
    \big[\Var_{P}(V)\big]_{(s,a)}
    &= \Var_{P_{s,a}}(V) \notag\\
    &= \sum_{s'\in \cS} P(s'\mymid s,a) (V_{s'})^2 
	-  \bigg( \sum_{s'\in \cS}P(s'\mymid s,a) V_{s'} \bigg)^2;
    \label{eq:defn-Var-Psa-V}
\end{align}
that is, the variance of $V_{s'}$ when $s'$ is sampled from the probability vector $P(\cdot\mymid s,a)$. This can be expressed compactly via matrix notation:
\begin{equation}
\Var_{P}(V) = P (V \circ V)-(P V)\circ(P V),
	\label{eq:matrix-VarP-V-expression}
\end{equation} 
where we recall that $a\circ b$ denotes the Hadamard product $[a_ib_i]_{1\leq i\leq S}$ (see Section~\ref{sec:notation}). 
Similarly, we define, for any given policy $\pi$,  
\begin{equation}
	\Var_{\bP_{\pi}}(V) 
    %=  \Pi^{\pi} \Var_{\bP}(V) 
    = P_{\pi} (V \circ V)-(P_{\pi} V)\circ(P_{\pi} V) \in \mathbb{R}^{S}.
	\label{eq:matrix-VarPpi-V-expression}
\end{equation} 
The following lemma provides a perturbation bound for $\mathsf{Var}_P(V)$. 
\begin{lemma}
\label{lem:variance-perturbation-var-P-V}
For any vectors $V,\widehat{V}\in\mathbb{R}^{S}$ obeying $0\leq V\leq \frac{1}{1-\gamma}1_{S}$
and $0\leq\widehat{V}\leq \frac{1}{1-\gamma}1_{S}$, one has
\begin{align*}
\big\|\mathsf{Var}_{P}(\widehat{V})-\mathsf{Var}_{P}(V)\big\|_{\infty} & \leq\frac{4}{1-\gamma}\|\widehat{V}-V\|_{\infty}.
\end{align*}
\end{lemma}
\begin{proof}
From \eqref{eq:matrix-VarP-V-expression}, we have the decomposition
\begin{align*}
\arxiv{&} \mathsf{Var}_{P}(\widehat{V})-\mathsf{Var}_{P}(V) \arxiv{\\} 
 & =P(\widehat{V}\circ\widehat{V}-V\circ V)-(P\widehat{V})\circ(P\widehat{V})+(PV)\circ(PV)\\
 & =P\big((\widehat{V}-V)\circ(\widehat{V}+V)\big)-\big(P(\widehat{V}-V)\big)\circ\big(P(\widehat{V}+V)\big).
\end{align*}
Hence, it follows that
\begin{align*}
\big\|\mathsf{Var}_{P}(\widehat{V})-\mathsf{Var}_{P}(V)\big\|_{\infty} & \leq\|P\|_{1}\|\widehat{V}-V\|_{\infty}\|\widehat{V}+V\|_{\infty}  \arxiv{\notag\\}
\arxiv{&\qquad } +\|P\|_{1}^{2}\|\widehat{V}-V\|_{\infty}\|\widehat{V}+V\|_{\infty}\\
 & \leq\frac{4}{1-\gamma}\|\widehat{V}-V\|_{\infty},
\end{align*}
where the last line is valid since  $\|\widehat{V}\|_{\infty},\|V\|_{\infty}\leq\frac{1}{1-\gamma}$ and $\|P\|_{1}=1$.
\end{proof}

\subsection{Additional basic properties}

Equipped with the above notation, we present a couple of basic properties that will be useful throughout.

To begin, the Bellman consistency equation \eqref{eq:bellman_consistency} can be expressed in vector/matrix form as
\begin{subequations}
\label{eq:Qpi-Vpi-matrix-notation-inf}
\begin{align}
    Q^{\pi} = r + \gamma PV^{\pi},
    \label{eq:Q-pi-bellman-vector-inf}
\end{align}
which, when combined with \eqref{eq:Vpi-a-function-of-Qpi-inf},  yields
\begin{align}
    V^{\pi} = \Pi^{\pi} Q^{\pi} =  r_{\pi} + \gamma P_{\pi}V^{\pi}. 
    \label{eq:Vpi-Bellman-inf-matrix}
\end{align}
Solving \eqref{eq:Vpi-Bellman-inf-matrix} for $V^{\pi}$ immediately gives
\begin{align}
    V^{\pi}  = (I-\gamma  P_{\pi})^{-1}  r_{\pi} . 
    \label{eq:Vpi-Bellman-inversion-matrix-inf}
\end{align}
Substituting \eqref{eq:Vpi-Bellman-inf-matrix} into \eqref{eq:Q-pi-bellman-vector-inf} and solving for $Q^{\pi}$ yield
\begin{align}
    Q^{\pi} &= r + \gamma P \Pi^{\pi} Q^{\pi} = r + \gamma P^{\pi} Q^{\pi} \arxiv{\notag \\}
    \arxiv{&} \Longrightarrow \quad~~ 
     Q^{\pi}  = (I-\gamma  P^{\pi})^{-1}  r .
     \label{eq:Qpi-Bellman-inversion-matrix-inf}
\end{align}
\end{subequations}
In addition, the Bellman optimality condition \eqref{eq:bellman_fix_point} admits a vector/matrix form as
\begin{align}
    Q^\star = r + \gamma P V^\star.
    \label{eq:Q-opt-bellman-vector-inf}
\end{align}

The careful reader would immediately remark that,  
for \eqref{eq:Vpi-Bellman-inversion-matrix-inf} to hold, the matrix $I-\gamma P_{\pi}$ must be invertible. Fortunately, this requirement is automatically satisfied whenever $0\leq \gamma < 1$. A few basic properties of this inversion are provided in the lemma below. 
\begin{lemma}
    \label{lem:basic-property-matrix-notation-inf}
    For any $0\leq \gamma < 1$ and any policy $\pi$, it holds that
    \begin{subequations}
    \label{eq:sum-I-minu-gammaP-expand-prelim} 
    \begin{align}
    (I-\gamma P_{\pi})^{-1}&=I+\sum_{i=1}^{\infty}(\gamma P_{\pi})^{i},
	\label{eq:sum-I-minu-gammaP-expand0-prelim} \\
   % (I-\gamma P^{\pi})^{-1} &=I+\sum_{i=1}^{\infty}(\gamma P^{\pi})^{i},
%	\label{eq:sum-I-minu-gammaP-expand1-prelim} \\
    	(I-\gamma P_{\pi})^{-1}1_{S} &=\frac{1}{1-\gamma}1_{S},
	\label{eq:sum-I-minu-gammaP-expand3-prelim} 
    \\
    \big\|(I-\gamma P_{\pi})^{-1}\big\|_1 &=\frac{1}{1-\gamma},
	\label{eq:sum-I-minu-gammaP-expand5-prelim} 
       % 	(I-\gamma P^{\pi})^{-1}1_{SA} &=\frac{1}{1-\gamma}1_{SA},
	%\label{eq:sum-I-minu-gammaP-expand4-prelim} 
    \end{align}
    \end{subequations}
where $1_n$ denotes the $n$-dimensional all-one vector. 
It also follows directly from \eqref{eq:sum-I-minu-gammaP-expand0-prelim} 
%and \eqref{eq:sum-I-minu-gammaP-expand1-prelim} 
that $(I-\gamma P_{\pi})^{-1}$ is a nonnegative matrix. 

Additionally, the above results continue to hold if $P_{\pi}$ is replaced with $P^{\pi}$ and $1_S$ is replaced with $1_{SA}$. 
%and $(I-\gamma P^{\pi})^{-1}$ are 
%
\end{lemma}
\begin{proof}[Proof of Lemma~\ref{lem:basic-property-matrix-notation-inf}]
With regard to \eqref{eq:sum-I-minu-gammaP-expand0-prelim}, it suffices to note that 
$$
    \lim_{i\rightarrow \infty} \big\| \gamma^i (P_{\pi})^i \big\|_1 = \lim_{i\rightarrow \infty} \gamma^i = 0,
$$
where we have used the fact that $(P_{\pi})^i$ is also a probability transition matrix (so that $\|(P_{\pi})^i\|_1=1$). 
This implies that the matrix geometric series in \eqref{eq:sum-I-minu-gammaP-expand0-prelim} is convergent, thus justifying the validity of the Neumann series expansion in \eqref{eq:sum-I-minu-gammaP-expand0-prelim}.  
Regarding \eqref{eq:sum-I-minu-gammaP-expand3-prelim}, we can see from \eqref{eq:sum-I-minu-gammaP-expand0-prelim} that 
\begin{align*}
	(I-\gamma P_{\pi})^{-1}1_{S} &=\bigg(I+\sum_{i=1}^{\infty}(\gamma P_{\pi})^{i}\bigg)1_{S}=\bigg(1+\sum_{i=1}^{\infty}\gamma^{i}\bigg)1_{S} =\frac{1}{1-\gamma}1_{S},
\end{align*}
where the penultimate relation follows since  $(P_{\pi})^i$ is  a probability transition matrix.  This combined with the nonnegativity of $(I-\gamma P_{\pi})^{-1}$ and the definition of $\|\cdot\|_1$ (cf.~\eqref{eq:defn-L1-norm}) immediately justifies \eqref{eq:sum-I-minu-gammaP-expand5-prelim}. 
%Properties~\eqref{eq:sum-I-minu-gammaP-expand1-prelim} and \eqref{eq:sum-I-minu-gammaP-expand4-prelim} follow from the same arguments. 
\end{proof}

%\begin{lemma}
%   \label{lemma:V-property-matrix-inf}
%    \begin{subequations}
%    \label{eq:V-property-matrix-inf}
%    %
%    \begin{align}
%        V^{\pi}=(I-\gamma P_{\pi})^{-1}r_{\pi},\label{eq:V-property-matrix-inf-1}
%    \end{align}
%    %
%    \end{subequations}
% \end{lemma}

Moreover, combining Lemma~\ref{lem:basic-property-matrix-notation-inf} with the Bellman equations in \eqref{eq:Qpi-Vpi-matrix-notation-inf} reveals basic perturbation bounds. Consider two discounted infinite-horizon MDPs $\mathcal{M}=(\cS,\cA,P,r,\gamma)$ and $\widetilde{\mathcal{M}}=(\cS,\cA,P,\widetilde{r},\gamma)$, which differ only in their reward functions. Denote by $\widetilde{Q}^{\pi}$ (resp.~$\widetilde{V}^{\pi}$) the Q-function (resp.~value function) of policy $\pi$ in $\widetilde{\mathcal{M}}$, and similarly, let $\widetilde{Q}^{\star}$ (resp.~$\widetilde{V}^{\star}$) represent the optimal Q-function (resp.~value function) in $\widetilde{\mathcal{M}}$. 
For any policy $\pi$, it holds that
\begin{subequations}
\label{eq:QV-pi-perturbabtion-r-inf}
\begin{align}
    \big\| Q^{\pi}-\widetilde{Q}^{\pi} \big\|_{\infty} &= \big\| \big(I-\gamma P^{\pi}\big)^{-1} r - \big(I-\gamma P^{\pi}\big)^{-1} \widetilde{r} \big\|_{\infty} \notag\\
    &\leq  \big\| \big(I-\gamma P^{\pi}\big)^{-1}   \big\|_{1} \| r - \widetilde{r} \|_{\infty} 
    = \frac{1}{1-\gamma }\| r - \widetilde{r} \|_{\infty},
    \label{eq:Q-pi-perturbabtion-r-inf}
\end{align}
and similarly, 
\begin{align}
    \big\| V^{\pi}-\widetilde{V}^{\pi}  \big\|_{\infty}
    \leq \frac{1}{1-\gamma }\| r - \widetilde{r} \|_{\infty}. 
        \label{eq:V-pi-perturbabtion-r-inf}
\end{align}
\end{subequations}
Given the validity of \eqref{eq:QV-pi-perturbabtion-r-inf} for every policy $\pi$, we immediately reach
\begin{subequations}
\label{eq:QV-star-perturbabtion-r-inf}
\begin{align}
\big\| Q^{\star}-\widetilde{Q}^{\star}  \big\|_{\infty} 
    & = 
\Big\| \max_{\pi}Q^{\pi}-\max_{\pi}\widetilde{Q}^{\pi}  \Big\|_{\infty}   
\leq \max_{\pi}\big\| Q^{\pi}-\widetilde{Q}^{\pi}  \big\|_{\infty} \notag\\
    &\leq \frac{1}{1-\gamma }\| r - \widetilde{r} \|_{\infty}, 
\end{align}
and similarly,
\begin{align}
    \big\| V^{\star}-\widetilde{V}^{\star}  \big\|_{\infty}
    &\leq \frac{1}{1-\gamma }\| r - \widetilde{r} \|_{\infty}. 
    \label{eq:V-star-perturbabtion-r-inf}
\end{align}
\end{subequations}

\section{Finite-horizon episodic MDPs}

The second class of MDPs studied extensively in this monograph is that of finite-horizon episodic MDPs, which we now introduce. In contrast to infinite-horizon MDPs, where the time step can continue indefinitely, finite-horizon episodic MDPs model problems with a predetermined number of steps in each episode.

\subsection{Basic components}  
A finite-horizon Markov decision process, denoted by $\mathcal{M}=\{\cS, \cA, H, P, r\}$, consists of the following key components.  
\begin{itemize}
\item[(i)] $\cS=\{1,\dots,S\}$: a state space containing $S$ different states, akin to the infinite-horizon counterpart. 

\item[(ii)] $\cA=\{1,\dots,A\}$: an action space comprised of $A$ distinct actions, similar to the infinite-horizon counterpart. 

\item[(iii)] $H$: the horizon length, which specifies the number of time steps in each episode of the MDP. This parameter dictates the planning horizon for decision-making. 

\item[(iv)] $P=\{P_h\}_{1\leq h\leq H}$, with $P_h: \cS\times \cA\rightarrow \Delta(\cS)$ denoting the  transition kernel at step $h$ (namely,  $P_h(\cdot \mymid s,a)$ stands for the transition probability of the MDP at step $h$ when the current state-action pair is $(s,a)$). Conveniently, we introduce the following $S$-dimensional row vector
\begin{equation} \label{eq:transition_vector}
	P_{s,a,h} \coloneqq P_h(\cdot \mymid s,a )  
    \,\in\,\mathbb{R}^{1\times S}
\end{equation}
for any $(s,a,h)\in \cS\times \cA \times [H]$ to indicate the corresponding transition probability.

\item[(v)] $r = \{r_h\}_{1\leq h\leq H}$, with $r_h: \cS\times \cA \rightarrow [0,1]$ denoting the reward function at step $h$ (namely, $r_h(s,a)$ indicates the immediate reward gained at step $h$ when taking action $a$ in state $s$). 
Akin to the infinite-horizon setting, we assume throughout that all immediate rewards are deterministic, fully known to the learner, and lie within the interval $[0, 1]$.
\end{itemize}

\noindent 
 Hereafter, we focus on {\em nonstationary} (i.e., \textit{time-inhomogeneous}) MDPs within the finite-horizon framework, where both the transition probabilities and reward functions are allowed to vary across time steps. We are particularly interested in regimes in which the size $S$ of the state space, the size $A$ of the action space, and the horizon length $H$ are all large, reflecting the challenges inherent in high-dimensional, long-horizon RL problems.

\subsection{Policy, value function, Q-function and occupancy distribution}

We next introduce the counterparts of the preceding concepts for finite-horizon episodic MDPs. 
Let us begin with the notion of a policy. A (possibly randomized) policy, denoted by $\pi=\{\pi_h\}_{1\leq h\leq H}$ with $\pi_h: \cS \rightarrow \Delta (\cA)$, is an action selection rule,   
such that $\pi_h(a \mymid s)$ specifies the probability of choosing action $a$ at step $h$ when in state $s$.  When $\pi$ is a deterministic policy, we again overload the notation and write $\pi_h(s)$ to denote the action selected by $\pi$ in state $s$ at step $h$.

Next, we define the value function and Q-function for finite-horizon episodic MDPs. To this end, consider an episode $\{(s_h,a_h)\}_{1\leq h\leq H}$ with length $H$, 
with $s_h$ and $a_h$ denoting, respectively, the state and the action at step $h$, which serve as the basis for the definitions that follow.

\begin{itemize}
\item {\em Value function.} 
The value function of policy $\pi$ is denoted by $V^{\pi}=\{V_h^{\pi}\}_{1\leq h\leq H}$, 
where for each step $1\leq h\leq H$, $V_h^{\pi}: \cS\rightarrow \mathbb{R}$ quantifies the expected cumulative reward from a given state when starting at step $h$ and following $\pi$ until the end of the episode. More precisely, we define 
\begin{align}
	\forall s\in \cS: \quad 
	V_h^{\pi}(s) \coloneqq \mathbb{E} \left[ \sum_{t=h}^H r_t(s_t, a_t) \mid s_h = s \right],
	\label{eq:defn-V-pi-finite}
\end{align}
where the expectation is over the randomness of the  trajectory $\{(s_t, a_t)\}_{t=h}^H$ when policy $\pi$ is implemented
(i.e., $a_t\sim \pi_t(\cdot\mymid s_t)$ and $s_{t+1}\sim P_t(\cdot\mymid s_t, a_t)$ for all $t\geq h$).   Moreover, 
if the initial state $s_1$ is drawn from some state distribution $\rho \in \Delta(\cS)$, we find it convenient to define the value function of policy $\pi$ under $\rho$ as 
\begin{equation}
	V_1^{\pi}(\rho) \coloneqq \mathop{\mathbb{E}}_{s\sim \rho} \big[ V^{\pi}_1(s) \big].
	\label{eq:defn-V-rho-finite}
\end{equation}

\item {\em Q-function.} 
Analogously, the Q-function of policy $\pi$ is denoted by $Q^{\pi}=\{Q_h^{\pi}\}_{1\leq h\leq H}$, which extends the definition of the value function by also conditioning on the first action to be taken. To be precise, for each $1\leq h\leq H$ and every $(s,a)\in \cS \times \cA$, we define
\begin{align}
	%\forall (s,a)\in \cS \times \cA: \; 
	Q_h^{\pi}(s,a) \coloneqq \mathbb{E} \left[ \sum_{t=h}^H r_t(s_t, a_t) \mid s_h = s, a_h=a \right],
	\label{defn:Q-pi-finite}
\end{align}
where the trajectory $\{(s_t, a_t)\}_{t=h}^H$ is sampled in a way such that $s_{t+1}\sim P_t(\cdot\mymid s_t, a_t)$ and $a_{t+1}\sim \pi_{t+1}(\cdot\mymid s_{t+1})$ and  for all $t\geq h$.  In words, $Q_h^{\pi}(s,a)$ quantifies the remaining expected reward of taking action $a$ in state $s$ at step $h$ and then following policy $\pi$ thereafter. 

\end{itemize}

\noindent 
No discount factor is needed because the value function and Q-function are always finite over a finite horizon. As in the infinite-horizon setting, there always exists at least one deterministic policy that simultaneously maximizes the value function and the Q-function for all $(s,a,h)\in \cS\times \cA\times [H]$ \citep{bertsekas2017dynamic}. 
In light of this, we shall denote by $\pi^{\star}=\{\pi^{\star}_h\}_{1\leq h\leq H}$ an {\em optimal deterministic} policy throughout this monograph, which allows us to use $\pi^{\star}_h(s)$ to indicate the corresponding optimal action chosen in state $s$ at step $h$. 
The resulting optimal value function and optimal Q-function are denoted respectively by $V^{\star}=\{V^{\star}_h\}_{1\leq h\leq H}$ and $Q^{\star}=\{Q^{\star}_h\}_{1\leq h\leq H}$: 
\begin{align}
	\forall h \in [H]:
	\quad V_{h}^{\star} \coloneqq V_h^{\pi^{\star}} \quad \text{and} \quad Q_{h}^{\star} \coloneqq Q_h^{\pi^{\star}} .
    \label{eq:defn-optimalV-optimalQ-finite-prelim}
\end{align}

Furthermore, we introduce the {\em occupancy distributions} associated with policy $\pi = \{\pi_h\}_{1\leq h\leq H}$, describing how often states or state-action pairs are visited at a given step. Specifically, for any step $1\leq h\leq H$, define 
\begin{subequations} 	\label{eq:dh-pi-defn}
\begin{align}
	d_{h}^{\pi}(s; \rho) & \coloneqq\mathbb{P}\big(s_{h}=s\mid s_{1}\sim\rho;\pi\big), \\
	d_{h}^{\pi}(s,a; \rho) & \coloneqq \mathbb{P}\big(s_{h}=s,a_{h}=a\mid s_{1}\sim\rho;\pi\big) 
	= d_{h}^{\pi}(s; \rho) \pi_h( a \mymid s),
\end{align}
\end{subequations}%
which are conditioned on the initial state distribution $s_1\sim \rho$ and the event that all actions are sampled according to $\pi$. As a sanity check, note that at the beginning of the episode, the occupancy state distribution simply coincides with the initial state distribution, i.e.,  
\begin{equation}
	d_1^{\pi}(s; \rho) = \rho(s) \qquad \text{for any policy }\pi\text{ and any state }s\in \cS.
	\label{eqn:defn-d1rho-finite}
\end{equation}
For convenience, we introduce shorthand notation for the occupancy distributions induced by the optimal deterministic policy $\pi^{\star} = \{\pi_h^{\star}\}_{1\leq h\leq H}$: 
\begin{align}
	d_h^{\star}(s) &\coloneqq  d_h^{\pi^{\star}}(s; \rho) , \\
	d_h^{\star}(s, a) &\coloneqq d_h^{\pi^{\star}}(s,a; \rho) = d_h^{\star}(s) \mathds{1} \{a= \pi_h^{\star}(s)\}
	\label{eq:dhstar-finite}
\end{align}
for all $(s,a,h)\in \cS\times \cA \times [H]$, 
where the last identity holds given that policy $\pi^{\star}$ is assumed to be deterministic.

\subsection{Bellman equations and dynamic programming}

Next, we present the Bellman equations for finite-horizon MDPs, which underlie the design of classical dynamic programming algorithms. 

% \citep{bertsekas2017dynamic}

Denote the terminal value functions as $V_{H+1}^{\pi} =0$ and $Q_{H+1}^{\pi}=0$ for any policy $\pi=\{\pi_h\}_{1\leq h\leq H}$. For a given policy $\pi$, its associated value function and Q-function can be computed through backward recursion starting from the terminal states as follows: for each $h=H,\dots,1$, 
\begin{subequations}
\label{eq:Bellman-finite-pi}
\begin{align}
Q_{h}^{\pi} (s,a)  & = r_h(s,a) + \mathop{\mathbb{E}}_{s^{\prime}\sim P_h(\cdot \mymid s,a)} \big[V_{h+1}^{\pi}(s^{\prime})\big], \\
V_{h}^{\pi} (s) & = \mathop{\mathbb{E}}_{a\sim \pi_h(\cdot|s)} \big[Q_h^{\pi} (s,a)\big]
\end{align}
\end{subequations}
for every $(s,a)\in \cS\times \cA$. Intuitively, at each step $h$, the Q-value consists of two components: the immediate reward received at that step, and the expected remaining value of proceeding according to $\pi$ from the next state at step $h+1$.  
%
%each step adds the immediate reward at time $h$ to the expected value at step $h + 1$. 
%
Analogously, given that $V_{H+1}^{\star} =0$ and $Q_{H+1}^{\star}=0$, the Bellman optimality equations assert that: for every $h= H, \ldots, 1$,  
\begin{subequations}
\label{eq:Bellman-finite-opt}
\begin{align}
Q_{h}^{\star} (s,a)  & = r_h(s,a) + \mathbb{E}_{s^{\prime}\sim P_h(\cdot | s,a)} \big[V_{h+1}^{\star}(s^{\prime})\big], \\
V_{h}^{\star}(s) & = \max_{a\in \cA} Q_h^{\star}(s,a)
\end{align}
\end{subequations}
for every $(s,a)\in \cS\times \cA$.

In summary, given full knowledge of the model, the above recursions --- often referred to as {\em value iteration (\texttt{VI})} --- allow the finite-horizon value functions and Q-functions to be computed step by step via backward induction, starting from the terminal step and proceeding backward to $h = 1$. This procedure embodies the dynamic programming principle:  optimal decisions at each step can be determined by reasoning backward from the end of the episode.

\subsection{Vector and matrix notation}
%, and some basic facts

As in the discounted infinite-horizon setting, the following vector and matrix notation will streamline the presentation (with a slight abuse of notation).
\begin{itemize}
\item
	$r_h\in\mathbb{R}^{SA}$: a vector representing the reward function at step $h$, so that its $(s,a)$-th entry is the corresponding immediate reward at step $h$. 
		
\item 
	$V_h^{\pi},V_h^{\star}\in\mathbb{R}^{S}$: vectors representing, respectively, the value function of policy $\pi$ at step $h$ (cf.~\eqref{eq:defn-V-pi-finite}) and the optimal value function at step $h$ (cf.~\eqref{eq:defn-optimalV-optimalQ-finite-prelim}).

\item
$Q_h^{\pi},Q_h^{\star}\in\mathbb{R}^{SA}$: vectors representing, respectively, the Q-function of policy $\pi$ at step $h$ (cf.~\eqref{defn:Q-pi-finite}) and the optimal Q-function at step $h$ (cf.~\eqref{eq:defn-optimalV-optimalQ-finite-prelim}).

\item
	$P_h\in\mathbb{R}^{SA\times S}$: a matrix representing the probability transition kernel $P_h$ at step $h$, whose $(s,a)$-th row is the probability vector corresponding to $P_h(\cdot\mymid s,a)$.

%\item $P_{h,\pi}\in\mathbb{R}^{S\times S}$: a submatrix of $P_h$,  which consists of the rows with indices coming from $\{(s,\pi_h(s))\mid s\in \cS\}$. 

%\item $r_h^{\pi} \in\mathbb{R}^{S}$: a subvector of $r_h$, which consists of the rows with indices coming from $\{(s,\pi_h(s))\mid s\in \cS\}$.

\end{itemize}
The Bellman equations (cf.~\eqref{eq:Bellman-finite-pi} and \eqref{eq:Bellman-finite-opt}) can therefore be expressed compactly using the above vector and matrix notation as
\begin{subequations}
	\label{eq:Bellman-matrix-finite}
\begin{align}
	Q^{\pi}_h &=  r_h +  P_h  V^{\pi}_{h+1}, \qquad &&1\leq h\leq H,  \label{eq:Bellman-matrix-finite-pi}\\
	Q^{\star}_h &=  r_h +  P_h  V^{\star}_{h+1}, \qquad &&1\leq h\leq H.
    \label{eq:Bellman-matrix-finite-opt}
\end{align}
\end{subequations}

In addition, we introduce the notation for variances.  
For any vector $V=[V_i]_{1\leq i\leq S}\in\real^{S}$, we define the scalar quantity $\Var_{P_{s,a,h}}(V) \in \real$ and the vector  $\Var_{P_{h}}(V) \in \real^{SA}$ as
\begin{subequations}
\begin{align}
	\Var_{P_{s,a,h}}(V)
    &= \sum_{s'\in \cS} P_h(s'\mymid s,a) (V_{s'})^2 
	-  \bigg( \sum_{s'\in \cS}P_h(s'\mymid s,a) V_{s'} \bigg)^2,
    \label{eq:defn-Var-Psah-V-finite} \\
	\Var_{P_h}(V) &= \big[ \Var_{P_{s,a,h}}(V) \big]_{(s,a)\in \cS\times \cA} \notag\\
	&= P_h (V \circ V)-(P_h V)\circ(P_h V) ,
	\label{eq:defn-Var-Ph-V-finite}
\end{align}
\end{subequations}
representing the variance of $V(s')$ when $s'$ is drawn according to either a single distribution $P_{s,a,h}$ or, collectively, all rows of the transition matrix $P_h$.

\section{Model-based vs.~model-free approaches}

Thus far, our discussion has centered on the fundamentals of MDPs and on dynamic programming algorithms given complete model knowledge. While classical dynamic programming provides powerful and efficient methods for identifying optimal policies when the model is fully specified, reinforcement learning often has to contend with more challenging scenarios in which the model is unknown {\em a priori} and can only be learned from data --- either pre-collected or gathered in real time. The agent must therefore make decisions under uncertainty. 

%0.8\textwidth

\begin{figure}[t]
\begin{center}
\includegraphics[width=0.6\textwidth]{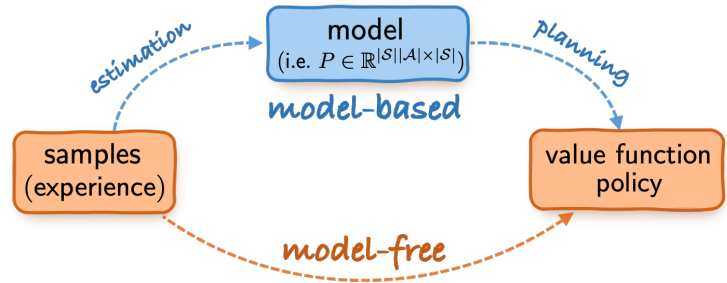} 
\end{center}	
	\caption{The model-based approach vs.~the model-free approach. A key distinction is that model-free methods dispense with explicit model estimation and instead learn the value and policy of interest directly without the aid of an estimated model. 
	\label{fig:model_based_free}}
\end{figure}

At a high level, there are at least two principal approaches in RL, which we describe below and illustrate in Figure~\ref{fig:model_based_free}. 
\begin{itemize}
    \item {\em Model-based RL}, which decouples the learning process into two stages: a model estimation stage that estimates the transition kernel and rewards using available data, and a subsequent planning stage that applies planning algorithms on the learned model to compute near-optimal policies and values. Given the use of the meta-task of model estimation, it requires maintaining and updating the learned model from collected data. 
    
    \item {\em Model-free RL}, which circumvents the model estimation stage and seeks to learn the optimal values and policies directly. Prominent examples include Q-learning \citep{watkins1992q} and its many variants. Compared to the model-based counterpart, it holds the promise of lower space complexity, as it avoids the need to store and update a full description of the model. Following the convention in RL theory (e.g., \citet{strehl2006pac,jin2018q}), we classify an algorithm as model-free if its space complexity is substantially smaller than that required to represent the full model, irrespective of the number of samples collected.
\end{itemize}

\noindent 
In this work, we investigate both model-based and model-free approaches, with particular emphasis on their statistical efficiency. As the subsequent chapters will illustrate, the model-based approach is often able to achieve the best possible sample complexity, albeit at the expense of higher memory usage and computational overhead.

\section{Concentration inequalities}
\label{sec:concentration}

Given that RL relies on learning from data, 
non-asymptotic concentration inequalities developed in the recent literature on high-dimensional probability and statistics \citep{boucheron2013concentration,vershynin2018high,wainwright2019high} play a central role in establishing tight sample complexity bounds.  For this reason, we briefly review a few commonly used concentration results.

Consider a probability transition kernel $P: \cS\times \cA\rightarrow \Delta(\cS)$. Suppose that for each state-action pair $(s,a)$, we have access to $N$ independent samples drawn from $P(\cdot \mymid s,a)$, namely, 
\[
    s^{(i)}(s,a) \overset{\mathrm{i.i.d.}}{\sim} 
    P(\cdot \mymid s,a),\qquad i=1,\dots,N.  
\]
From these samples, we can naturally form an empirical estimate $\widehat{P}(\cdot\mymid s,a)$ of the transition probability $P(\cdot\mymid s,a)$ as follows:
\begin{align}
    \widehat{P}(s'\mymid s,a) = \frac{1}{N} \sum_{i=1}^N \ind \big\{s^{(i)}(s,a) = s' \big\} \quad \text{for all } s'\in \cS.
\end{align}
Taking these estimates together yields an estimate $\widehat{P} \in \mathbb{R}^{SA\times S}$ of the probability transition matrix $P$, namely, the $(s,a)$-th row of $\widehat{P}$ is given by
\begin{align}
   \widehat{P}_{(s,a),s'} =  \widehat{P}(s'\mymid s,a)
   \qquad \text{for all }(s,a,s')\in \cS\times \cA\times \cS.
\end{align}
The following concentration bound is a direct consequence of the classical Hoeffding inequality \citep{vershynin2018high} combined with the union bound. 
\begin{lemma}[Hoeffding inequality]
\label{lem:Hoeffding-prelim}
Consider a given state-action pair $(s,a)$ and a finite set of vectors $\mathcal{V}\subset \mathbb{R}_+^S$. Fix any $0<\delta<1$. 
If $\mathcal{V}$ is independent of $\widehat{P}_{s,a}$,  then with probability greater than $1-\delta$,  
\begin{align}
\big|\big(\widehat{P}_{s,a}-P_{s,a})V\big| & \leq\sqrt{\frac{\|V\|_{\infty}^2\log\big(\frac{2|\mathcal{V}|}{\delta}\big)}{2N}}
	\label{eq:Hoeffding-Psa-prelim}
\end{align}
holds simultaneously for all $V\in \mathcal{V}$. 
\end{lemma}
In particular, the Hoeffding bound \eqref{eq:Hoeffding-Psa-prelim} depends on $V$ only through the maximum magnitude of its entries, without relying on any finer structural properties of $V$, such as its variance.  

Despite its simplicity and broad applicability, Hoeffding's inequality is often conservative, especially when the underlying standard deviation is substantially smaller than the worst-case range. In such situations, Bernstein-type inequalities can provide significantly sharper concentration bounds by explicitly incorporating variance information. To this end, we record below concentration inequalities obtained directly from the Bernstein inequality \citep{vershynin2018high} in conjunction with a union bound; see also \citet[Lemma~6]{agarwal2020model}. 
\begin{lemma}[Bernstein inequality]
\label{lem:Bernstein-prelim}
Consider a given state-action pair $(s,a)$ and a finite set of vectors $\mathcal{V}\subset \mathbb{R}_+^S$. Fix any $0<\delta<1$. 
If $\mathcal{V}$ is independent of $\widehat{P}_{s,a}$,  then with probability greater than $1-\delta$,  
\begin{align}
\big|\big(\widehat{P}_{s,a}-P_{s,a})V\big| & \leq\sqrt{\frac{2\log\big(\frac{4|\mathcal{V}|}{\delta}\big)}{N}}\sqrt{\mathsf{Var}_{P_{s,a}}(V)}+\frac{2\|V\|_{\infty}\log\big(\frac{4|\mathcal{V}|}{\delta}\big)}{3N}
\label{eq:Bernstein-Psa-prelim}
\end{align}
holds simultaneously for all $V\in \mathcal{V}$. 
%\yly{over $\mathcal{V}$?}
%
Moreover, if $\mathcal{V}$ is statistically independent of $\widehat{P}$, then with probability greater than $1-\delta$, one has
\begin{align}
\big|\big(\widehat{P}-P)V\big| & \leq\sqrt{\frac{2\log\big(\frac{4SA|\mathcal{V}|}{\delta}\big)}{N}}\sqrt{\mathsf{Var}_{P}(V)}+\frac{2\|V\|_{\infty}\log\big(\frac{4SA|\mathcal{V}|}{\delta}\big)}{3N} 1_{SA}.
\label{eq:Bernstein-P-whole-prelim}
\end{align}
Here, we recall the definitions of $\mathsf{Var}_{P_{s,a}}(V)$ and $\mathsf{Var}_{P}(V)$ in \eqref{eq:defn-Var-Psa-V}. We use $1_{SA}$ to denote the all-one vector, and for any nonnegative vector $v$, $\sqrt{v}$ denotes the entrywise square root.  
\end{lemma}

In words, the Bernstein-style inequalities in Lemma~\ref{lem:Bernstein-prelim} reveal that the uncertainty of $\widehat{P}_{s,a}V$ is governed by both its variance and the largest entrywise magnitude of $V$. When the sample size is large enough, the variance term typically dominates, a phenomenon that will become evident in subsequent chapters.

A technical challenge arises when Bernstein's inequality is used to construct confidence intervals for quantities such as $P_{s,a}V$ based on the empirical estimate $\widehat{P}_{s,a}V$: the variance term $\mathsf{Var}_{P_{s,a}}(V)$ depends on the true transition kernel $P_{s,a}$, which is unavailable in practice. To circumvent this, it is useful to resort to the following empirical Bernstein inequality --- quoted from \citet[Theorem~4]{maurer2009empirical} --- which replaces the population variance with its empirical estimate, making the confidence interval computable from collected data.
\begin{lemma}[Empirical Bernstein inequality]\label{empirical bernstein}
Consider a given state-action pair $(s,a)$ and a finite set of vectors $\mathcal{V}\subset \mathbb{R}_+^S$. Fix any $0<\delta<1$, and suppose that $N\geq 2$. 
If $\mathcal{V}$ is independent of $\widehat{P}_{s,a}$,  then with probability greater than $1-\delta$,  
\begin{align*}
	\big|\big(\widehat{P}_{s,a}-P_{s,a})V\big| & \leq\sqrt{\frac{2\log\big(\frac{4|\mathcal{V}|}{\delta}\big)}{N-1}}\sqrt{\mathsf{Var}_{\widehat{P}_{s,a}}(V)}+\frac{7\|V\|_{\infty}\log\big(\frac{4|\mathcal{V}|}{\delta}\big)}{3(N-1)}
%\label{eq:Bernstein-Psa-prelim}
\end{align*}
holds simultaneously for all $V\in \mathcal{V}$. 
\end{lemma}

In addition, when coping with sums of i.i.d.~Bernoulli variables,  the following exponential tail bound becomes useful, which follows from standard Chernoff-type arguments.
\begin{lemma}
    \label{lem:tail-bound-sum-Bernoulli}
    Let $Z_i \overset{\mathrm{i.i.d.}}{\sim} \mathsf{Bern}(p)$ $(1\leq i\leq n)$ be a set of independent Bernoulli random variables with mean $0<p\leq 1$. Then for any 
$a>0$, one has
\begin{equation*}
    \mathbb{P}\bigg( \sum_{i=1}^n \big(Z_i - \mathbb{E}[Z_i]\big)  < -a n \bigg)\leq \exp\bigg(-\frac{a^2n }{2p} \bigg).
\end{equation*}
In addition, for any $\tau\geq 2 pn$, one has
\begin{equation*}
    \mathbb{P}\bigg( \sum_{i=1}^n Z_i    > \tau \bigg)\leq 2\exp(-0.19\tau ).
\end{equation*}
\end{lemma}
\begin{proof}
The first inequality can be found in \citet[Theorem~A.1.13]{Alon2008}, while the second inequality is an immediate consequence of \citet[Corollary~A.1.14]{Alon2008} with $\epsilon\geq 1$ therein. 
\end{proof}

%For instance, it naturally appears in exponential tail bounds via the Chernoff-type techniques. The following lemma provides a standard large-deviation bound of this kind; see, e.g., \citet{hagerup1990guided} for a proof.
%
%\begin{lemma}
%\label{lem:sum-bernoulli-KL}
%Let $Z_i \overset{\mathrm{i.i.d.}}{\sim} \mathsf{Bern}(p)$ $(1\leq i\leq n)$ be a collection of independent Bernoulli random variables with mean $p$. Then for any 
%$0<p\leq a < 1$, one has
%
%\begin{equation}
%    \mathbb{P}\bigg( \sum_{i=1}^n Z_n \geq a n \bigg)\leq \exp\big(-n \mathsf{KL}( a\,\|\,p )\big).
%\end{equation}
%%
%\end{lemma}

Moving beyond independent data, concentration inequalities for martingales are often needed to handle sequentially collected observations. Among the most useful are the Azuma-Hoeffding inequality \citep{azuma1967weighted,dubhashi2009concentration}, a martingale counterpart of Hoeffding's inequality, and Freedman's inequality \citep{freedman1975tail,tropp2011freedman}, which serves as a martingale analogue of Bernstein's inequality. Below, we record the standard Azuma-Hoeffding inequality and a user-friendly version of Freedman's
inequality (see \citet[Theorem~EC.1.]{li2024q}). 
\begin{lemma}[Azuma-Hoeffding inequality]\label{lem:azuma-hoeffding}
Consider any $0<\delta<1$. 
Suppose that $Y_{n}=\sum_{k=1}^{n}X_{k}\in\mathbb{R}$,
where $\{X_{k}\}$ is a real-valued scalar sequence obeying 
\[
\left|X_{k}\right|\leq B_k\qquad\text{and}\qquad\mathbb{E}\left[X_{k}\mid\left\{ X_{j}\right\} _{j:j<k}\right]=0\quad\quad\text{for all }k\geq1
\]
for some deterministic quantities $\{B_k\}$. Then with probability exceeding $1-\delta$, one has
\begin{equation}
    |Y_n| \leq \sqrt{2\sum_{k=1}^nB_k^2 \log \frac{2}{\delta}}.
\end{equation}
\end{lemma}
In brief, the Azuma–Hoeffding inequality shows that the fluctuations of a martingale sum can be controlled by the sum of the squared magnitudes of its martingale difference terms. This can be viewed as a direct extension of the Hoeffding inequality \citep{boucheron2013concentration} from independent data to weakly dependent martingale sequences.
\begin{lemma}[Freedman inequality]\label{lem:Freedman-prelim}
Consider any $0<\delta<1$. Suppose that $Y_{n}=\sum_{k=1}^{n}X_{k}\in\mathbb{R}$,
where $\{X_{k}\}$ is a real-valued scalar sequence obeying 
\[
\left|X_{k}\right|\leq B\qquad\text{and}\qquad\mathbb{E}\left[X_{k}\mid\left\{ X_{j}\right\} _{j:j<k}\right]=0\quad\quad\text{for all }k\geq1.
\]
Define
\[
W_{n}\coloneqq\sum_{k=1}^{n}\mathbb{E}
%_{k-1}
\left[X_{k}^{2} \mid\left\{ X_{j}\right\} _{j:j<k}\right],
\]
and suppose that $W_{n}\leq\sigma^{2}$ holds deterministically.
Then for any positive integer $K_0\geq1$, with probability at least $1-\delta$
one has
\begin{equation}
\left|Y_{n}\right|\leq\sqrt{8\max\bigg\{ W_{n},\frac{\sigma^{2}}{2^{K_0}}\bigg\}\log\frac{2K_0}{\delta}}+\frac{4}{3}B\log\frac{2K_0}{\delta}.\label{eq:Freedman-random}
\end{equation}
\end{lemma}

Informally, the Freedman inequality in Lemma~\ref{lem:Freedman-prelim} asserts that the deviation of a martingale sum is controlled by the cumulative conditional variance (i.e., $W_n$) and a uniform upper bound on the magnitude of each summand (i.e., $B$). The resulting concentration bound closely resembles the Bernstein inequality, with a key distinction that the total variance is now replaced by the cumulative conditional variance of the martingale sequence.

\begin{comment}

\section{Other preliminaries: divergence metrics}
\label{sec:divergence}

We also gather a few basic properties about several fundamental divergence metrics.

The above KL divergence between Bernoulli distributions is instrumental in deriving tail bounds for a sum of independent Bernoulli random variables. For instance, it naturally appears in exponential tail bounds via the Chernoff-type techniques. The following lemma provides a standard large-deviation bound of this kind; see, e.g., \citet{hagerup1990guided} for a proof.
%
\begin{lemma}
\label{lem:sum-bernoulli-KL}
Let $Z_i \overset{\mathrm{i.i.d.}}{\sim} \mathsf{Bern}(p)$ $(1\leq i\leq n)$ be a collection of independent Bernoulli random variables with mean $p$. Then for any 
$0<p\leq a < 1$, one has
%
\begin{equation}
    \mathbb{P}\bigg( \sum_{i=1}^n Z_i \geq a n \bigg)\leq \exp\big(-n \mathsf{KL}( a\,\|\,p )\big).
\end{equation}
%
\end{lemma}

\end{comment}

\section{Minimax lower bounds for multi-hypothesis testing}

To assess the statistical optimality of an RL algorithm, one typically derives a statistical lower bound and then determines whether the algorithm attains it. Arguments for statistical or information-theoretic lower bounds often differ substantially from those used for upper bounds. While lower-bound techniques constitute a central topic in statistics and information theory (see, e.g., \citet{tsybakov2009introduction,polyanskiy2025information}), a comprehensive treatment lies beyond the scope of this monograph. Instead, we single out a simple yet remarkably powerful and versatile tool that appears frequently in lower-bound analyses: Fano's lemma. Rooted in classical information theory and originally introduced by Robert Fano, this lemma provides a principled way to quantify the intrinsic difficulty of statistical problems through information-theoretic measures.

More specifically, suppose that there are $M+1$ probability distributions  $P_0,P_1,\dots,P_M$ on a common measurable space $(\mathcal{X},\mathcal{A})$. Consider any test $\psi$ that maps from $\mathcal{X}$ to $\{0,1,\dots,M\}$, with the goal of identifying the index of the distribution generating the observed samples. The performance of such a test can be measured by its error probability under each hypothesis, namely, $P_i(\psi \neq i)$ for $i=0,\dots,M$. 
The minimax probability of error is then defined as
\begin{align}
\label{eq:defn-minimax-prob-error-prelim}
    p_{\mathrm{e}} \coloneqq \inf_{\psi} \max_{0\leq i\leq M} P_i(\psi \neq i),
\end{align}
with the infimum taken over all possible tests. In words, $p_{\mathrm{e}}$ captures the smallest achievable worst-case error probability across all hypotheses. It is also useful to define the average probability of error as follows: 
\begin{align}
\label{eq:defn-avg-prob-error-prelim}
\overline{p}_{\mathrm{e}} \coloneqq \inf_{\psi} \frac{1}{M+1}\sum_{i=0}^M P_i(\psi \neq i),
\end{align}
which measures the error probability averaged over all  $M+1$ hypotheses.   
The Fano lemma provides a fundamental lower bound on the  probability of error for testing among $M+1$ hypotheses; the version below follows from a straightforward adaptation of the results in \citet[Section~2.7.1]{tsybakov2009introduction}.

 %Here and throughout, $\mathsf{KL}(P\,\|\,Q)$ represents the Kullback-Leibler (KL) divergence between two distributions $P$ and $Q$ defined as
%%
%\begin{align*}
%\mathsf{KL}(P\,\|\,Q) \coloneqq \displaystyle \int \bigg(\log \frac{\mathrm{d}P}{\mathrm{d}Q} \bigg) \mathrm{d} P,
%\end{align*}
%
%provided that $P$ is absolutely continuous w.r.t.~$Q$. 
%
\begin{lemma}[Fano's lemma]
    \label{lem:fano}
    Let $P_0,P_1,\dots,P_M$ be probability distributions on a common measurable space. Recall that $p_{\mathrm{e}}$ (resp.~$\overline{p}_{\mathrm{e}}$) is the minimax (resp.~average) probability of error defined in \eqref{eq:defn-minimax-prob-error-prelim} (resp.~\eqref{eq:defn-avg-prob-error-prelim}). 
    \begin{itemize}
        \item[(i)] Suppose that $M=1$. If $\mathsf{KL}(P_1\,\|\,P_0)\leq \beta<\infty$, then
        \begin{align}
            p_{\mathrm{e}}
            \geq 
            \overline{p}_{\mathrm{e}}\geq \max \bigg\{ \frac{1}{4}\exp(-\beta), \frac{1-\sqrt{\beta/2}}{2}\bigg\}.
            \label{eq:fano-binary}
        \end{align}

        \item[(ii)] Suppose that $M\geq 2$. If 
        \begin{subequations}
        \label{eq:fano-multi}
        \begin{align}
            \frac{1}{M+1} \sum_{j=1}^M \mathsf{KL}(P_j\,\|\,P_0)\leq \beta \log M 
        \end{align}
        for some $0<\beta<1$, then 
        \begin{align}
            p_{\mathrm{e}}\geq \overline{p}_{\mathrm{e}}\geq \frac{\log(M+1)-\log 2}{\log M}-\beta.
        \end{align}      
        \end{subequations}

%\item[(iii)] Suppose that $M\geq 2$ and that
%%
%$ \left\|
%        \frac{\mathrm{d}P_j}{\mathrm{d}P_0}
%    \right\|_{\infty}
%    \leq \kappa
%  %  \label{eq:fano-density-ratio-condition}
%$ 
% ($1\leq j\leq M$)
%%
%for some $\kappa\geq1$. If
%%
%\begin{align}
%    \frac{2\kappa}{M+1}
%    \sum_{j=1}^{M}
%    \mathsf{KL}(P_0\,\|\,P_j)
%    \leq \beta\log M
%    \label{eq:fano-reverse-KL-condition}
%\end{align}
%%
%for some $0<\beta<1$, then
%%
%\begin{align}
%    p_{\mathrm{e}}
%    \geq \overline{p}_{\mathrm{e}}
%    \geq
%    \frac{\log(M+1)-\log 2}{\log M}-\beta.
%    \label{eq:fano-reverse-KL-density-ratio}
%\end{align}
        
    \end{itemize}
\end{lemma}

In some applications, it might be more convenient to control
$\mathsf{KL}(P_0\,\|\,P_j)$, with the two distributions appearing in the
opposite order from that in Part~(ii) of Lemma~\ref{lem:fano}. The following bound, which follows from Pinsker's inequality, handles this situation directly.
\begin{lemma}
\label{lem:reverse-Fano-bound}
Let $P_0,P_1,\dots,P_M$ be probability distributions on a common
measurable space. Then
\begin{align}
    p_{\mathrm{e}}
    \geq
    \overline{p}_{\mathrm{e}}
    \geq
    1-\frac{1}{M+1}
    -
    \sqrt{
        \frac{M}{2(M+1)^2}
        \sum_{j=1}^{M}
        \mathsf{KL}(P_0\,\|\,P_j)
    }.
    \label{eq:reverse-Fano-bound}
\end{align}
\end{lemma}
\begin{remark}
The Pinsker-based lower bound in Lemma~\ref{lem:reverse-Fano-bound}  generally loses a factor of $\log M$ relative to Fano's
lemma and may therefore be loose when $M$ is exponentially large.
Nevertheless, it suffices for our purposes in this monograph.
\end{remark}
\begin{proof}[Proof of Lemma~\ref{lem:reverse-Fano-bound}]
Consider any statistical test $\psi$ taking values in $\{0,1,\dots,M\}$, and define the event 
$A_j\coloneqq\{\psi=j\}$. By the definition of the TV distance,
\begin{align*}
    P_j(A_j)-P_0(A_j)
    \leq
    \left|P_j(A_j)-P_0(A_j)\right|
    \leq \sup_A \left|P_j(A)-P_0(A)\right|
    =
    \mathsf{TV}(P_0,P_j).
\end{align*}
Consequently, for every $1\leq j\leq M$,
\begin{align*}
    P_j(\psi=j)
    \leq
    P_0(\psi=j)+\mathsf{TV}(P_0,P_j).
\end{align*} 
Summing over $j$ leads to
\begin{align*}
    \sum_{j=0}^{M}P_j(\psi=j)
    &\leq
    P_0(\psi=0)
    +\sum_{j=1}^{M}P_0(\psi=j)
    +\sum_{j=1}^{M}\mathsf{TV}(P_0,P_j)\\
    &=
    1+\sum_{j=1}^{M}\mathsf{TV}(P_0,P_j)\\
    &\leq
    1+\sum_{j=1}^{M}
    \sqrt{\frac{1}{2}\mathsf{KL}(P_0\,\|\,P_j)}\\
    &\leq
    1+
    \sqrt{
        \frac{M}{2}
        \sum_{j=1}^{M}
        \mathsf{KL}(P_0\,\|\,P_j)
    },
\end{align*}
where the penultimate inequality follows from Pinsker's inequality (see~\eqref{eq:KL-ub-chi2}, or \citet[Lemma~2.5]{tsybakov2009introduction}) and
the last from Cauchy-Schwarz. Consequently,
\begin{align*}
    \frac{1}{M+1}
    \sum_{j=0}^{M}P_j(\psi\neq j)
    &=
    1-
    \frac{1}{M+1}
    \sum_{j=0}^{M}P_j(\psi=j)
    \\
    &\geq
    1-\frac{1}{M+1}
    -
    \sqrt{
        \frac{M}{2(M+1)^2}
        \sum_{j=1}^{M}
        \mathsf{KL}(P_0\,\|\,P_j)
    }.
\end{align*}
Taking the infimum over $\psi$ proves the lower bound on
$\overline{p}_{\mathrm{e}}$, which in turn completes the proof.
\end{proof}

At a high level, Fano's lemma and Lemma~\ref{lem:reverse-Fano-bound} assert that if the candidate distributions are close to one another in terms of KL divergence, then no test can reliably distinguish among them, and consequently, both the minimax error probability and the average error probability must remain bounded away from zero. In this way, Fano's lemma unearths an intimate connection between the performance limit of any statistical test and the information-theoretic separation between distributions (as measured by KL divergence). A standard strategy for proving minimax lower bounds is therefore to construct a carefully chosen collection of nearby distributions and then control their KL divergences.

Given the central role of KL divergence in Fano's lemma, we find it useful to record several of its key properties. We begin with the tensorization (i.e., additivity) property, which characterizes how KL divergence decomposes under independent data; see, e.g., \citet[Section~2.4]{tsybakov2009introduction}. This simple property highlights a key advantage of KL divergence: when samples are independent, analysis of the overall KL divergence reduces to computing the KL divergence for individual samples.

\begin{lemma}\label{lem:KL-tensorize}
For two product distributions
$P=P_{1}\otimes\cdots\otimes P_{n}$ and $Q=Q_{1}\otimes\cdots\otimes Q_{n}$, the KL divergence satisfies
\[
\mathsf{KL}(P\,\|\,Q)=\sum_{i=1}^{n}\mathsf{KL}(P_{i}\,\|\,Q_{i}).
\]
\end{lemma}

 Additionally, when $P$ and $Q$ are both Bernoulli distributions, we adopt a slight abuse of notation and write
 %
% \begin{subequations}
 \begin{align}
 \mathsf{KL}(p\,\|\,q) & \coloneqq p\log \frac{p}{q} + (1-p)\log \frac{1-p}{1-q},   
 \label{eq:defn-KL-bern} 
 \end{align}
% \end{subequations}
 to denote the KL divergence between $\mathsf{Bern}(p)$ and $\mathsf{Bern}(q)$, where $\mathsf{Bern}(p)$ represents the Bernoulli distribution with mean $p$.  The following lemma establishes a quadratic upper bound on this divergence, which is sometimes more convenient to work with than the original expression.
 %of the KL divergence.

\begin{lemma}
\label{lem:KL-quadratic-lower-bound}
For any $0<p,q<1$,
\[
\mathsf{KL}(p\,\|\,q)\le \frac{(p-q)^2}{q(1-q)}.
\]
Moreover, if $1/2\le q<p<1$, then
\[
\mathsf{KL}(p\,\|\,q)
\le \mathsf{KL}(q\,\|\,p) \le \frac{(p-q)^2}{p(1-p)}.
\]
\end{lemma}

\begin{proof}
%[Proof of Lemma~\ref{lem:KL-quadratic-lower-bound}]
%
\begin{comment}
By symmetry, it suffices to focus on the case with $1/2\leq q<p<1$. 
Define a function $f(x)=\mathsf{KL}(x\,\|\,q)$. Then for any $x\in [q,p]$, a direct calculation yields
%
\begin{align*}
f'(x) & =\log x-\log q-\log(1-x)+\log(1-q),\\
f''(x) & =\frac{1}{x}+\frac{1}{1-x}=\frac{1}{x(1-x)}\geq \frac{1}{q(1-q)},
\end{align*}
%
where the last inequality holds since $1/2\le q\le x$ and $x(1-x)$ is decreasing on $[1/2,1]$.  The Taylor expansion then reveals the existence of some $x\in [q,p]$ such that
%
\[
    \mathsf{KL}(p\,\|\,q) = f(p) = f(q)+ f'(q)(p-q)+\frac{1}{2} f''(x) (p-q)^2 \geq \frac{(p-q)^2}{2q(1-q)},
\]
%
where $f(q)=f'(q)=0$. 

In addition, i
\end{comment}
We begin with the first relation. It is known that the KL divergence can be upper bounded by the $\chi^2$ divergence (see \citet[Lemma~2.7]{tsybakov2009introduction}), namely,  
\begin{align*}
\mathsf{KL}(p\,\|\,q) \leq \chi^2(p\,\|\,q) \coloneqq \frac{(p-q)^2}{q}+\frac{(1-p-(1-q))^2}{1-q}=\frac{(p-q)^2}{q(1-q)}.
\end{align*}

We now move to the second relation. Let
\[
a=\frac{p+q}{2}\in\Big[\frac{1}{2},1\Big]
\qquad \text{and} \qquad
b=\frac{p-q}{2}\in\Big[0,\frac{1}{4}\Big],
\]
so that $p=a+b$ and $q=a-b$. 
By the definition in \eqref{eq:defn-KL-bern} and some elementary algebra, we obtain
\begin{align*}
	\mathsf{KL}(p\parallel q)-\mathsf{KL}(q\parallel p) 
	& =(p+q)\log\frac{p}{q}+(2-p-q)\log\frac{1-p}{1-q}\\
	& =2a\log\left(\frac{a+b}{a-b}\right)
	+2(1-a)\log\frac{1-a-b}{1-a+b}\\
	&\eqqcolon g(a,b).
\end{align*}
Taking the derivative of $g(a,b)$ with respect to $b$ gives
\begin{align*}
	\frac{\partial g(a,b)}{\partial b}
	&=2a\left\{ \frac{1}{a+b}+\frac{1}{a-b}\right\} 
	-2\left(1-a\right)\left\{ \frac{1}{1-a+b}+\frac{1}{1-a-b}\right\} \\
	&= f(a)- f(1-a) \leq 0, \quad \text{where} \quad f(x)\coloneqq\frac{2x}{x+b}+\frac{2x}{x-b}.
\end{align*}
The last relation holds since $f$ is decreasing on $(b,\infty)$ and $a\geq 1-a$. 
Thus, for each fixed $a$, the function $g(a,b)$ is non-increasing in $b\geq 0$. Consequently,
\[
\mathsf{KL}(p\parallel q)-\mathsf{KL}(q\parallel p)
= g(a,b)\leq g(a,0)=0,
\]
which proves the desired claim.
\end{proof}

\section{Notes}

While discounted infinite-horizon stationary MDPs and finite-horizon nonstationary MDPs are arguably the two most extensively studied models, several other MDP formulations have also received considerable attention in the RL literature. 
Examples include (i) {\em infinite-horizon average-reward MDPs}, where discounted cumulative rewards are replaced by long-run average rewards \citep{puterman2014markov,wan2021learning,jin2021towards}, and (ii) {\em finite-horizon stationary MDPs}, in which the transition kernels and reward functions are both time-homogeneous \citep{ren2021nearly,li2022settling,zhang2022horizon}.  Each of these models bears similarities with the ones under exposition here, yet also brings its own distinctive challenges. For further details, we refer the interested reader to the aforementioned papers and the references therein. 

For broader background, we refer the interested reader to \citet{wainwright2019high,fan2020statistical,samworth2026modern} for comprehensive treatments of high-dimensional statistics and to \citet{vershynin2018high,boucheron2013concentration} for concentration inequalities. The texts by \citet{tsybakov2009introduction,polyanskiy2025information,gamal2026information} provide excellent introductions to minimax lower bounds based on information-theoretic techniques.

%% file: chapters/generative_model_model_based.tex
\chapter{RL with a generative model: the model-based approach}
\label{chapter:generative-model}

In this chapter, we begin our study of sample complexity in RL by first examining an idealized sampling mechanism, commonly referred to as the generative model or simulator. While this sampling mechanism is a substantial simplification of practical scenarios, it suggests a clean and tractable framework that serves as a natural first step towards understanding the fundamental statistical limits of RL. 
Our focus in this chapter is on model-based RL with the generative model, with the discussion of model-free algorithms deferred to Chapter~\ref{chapter:generative-model-MF}.

\section{Generative model or simulator} 
Consider first the discounted infinite-horizon stationary MDP introduced in Section~\ref{sec:prelim-inf-horizon-MDP}. 
Initially studied by \citet{kearns1999finite}, 
a generative model---also referred to as a simulator---allows one to query any state-action pair and sample an independent state transition from the ground-truth transition kernel. More precisely, each query to the generative model with a pair $(s,a)\in \mathcal{S}\times \mathcal{A}$ returns an independent sample: 
\begin{align}
    \label{eq:generative-model-homogeneous}
    s(s,a)  \overset{\mathrm{ind.}}{\sim} P(\cdot \mymid s,a).  
\end{align}
Similarly, in the case of a finite-horizon nonstationary MDP, each call to the generative model with an arbitrary state-action-step tuple $(s,a,h)\in \mathcal{S}\times \mathcal{A} \times [H]$ returns an independent sample:  
\begin{align}
    \label{eq:generative-model-inhomogeneous}
    s(s,a,h)  \overset{\mathrm{ind.}}{\sim} P_h(\cdot \mymid s,a).  
    %$$	s_{s,a}
\end{align}
Unless otherwise noted, we assume throughout that the reward function is fully known to the agent. This assumption can, in fact, be substantially relaxed without affecting the core insights underlying algorithm design and analysis. 

Studying RL with a generative model offers several key advantages.  To begin with,  simulators have been developed in several applications to simulate complex, real-world environments, such as for autonomous driving \citep{shah2018airsim}, robotic systems \citep{kumar2019offworld,yang2023learning}, and game playing \citep{bellemare2013arcade}. 
Beyond their practical utility, generative models provide a clean abstraction: by allowing direct sampling from arbitrary state–action pairs, they effectively decouple learning from the need for exploration.  This separation simplifies the problem and makes it easier to isolate the fundamental information bottlenecks inherent to the learning part. Overall, investigating RL under this idealistic sampling mechanism helps us distill core design principles, establish theoretical benchmarks, and understand the fundamental statistical limits of RL algorithms, which in turn guide us towards more practical and robust solutions.

\section{Model-based policy evaluation}
\label{sec:model-based-policy-eval-sim}

%
%\footnote{For simplicity of exposition, we assume throughout this subsection that $\mathcal{A}$ is a singleton. In fact, when a (possibly random) policy 
%$\pi$ is fixed, the resulting transition dynamics can be represented by an $S\times S$ matrix obtained by marginalizing over the actions, which is effectively equivalent to the case with $A=1$.} This often reflects the scenario with a pre-determined policy $\pi$  (e.g., for any state $s$, a policy dictates which action to execute, effectively reducing the action space to a single option). 
%

We begin with the fundamental task of {\em policy evaluation}, which seeks to estimate, based on the collected samples, the value function $V^{\pi}$ of~a given policy $\pi$. Our focus throughout this section is on $\ell_{\infty}$-accurate estimation: given a target accuracy level $\varepsilon>0$, the aim is to find an estimate $\widehat{V}^{\pi}$ that is $\varepsilon$-accurate uniformly over every state $s$, namely,   
\begin{align*}
	\forall s\in\mathcal{S}: \qquad \big| \widehat{V}^{\pi}(s) - {V}^{\pi}(s) \big| \leq \varepsilon. 
\end{align*}
Recall our assumption that all immediate rewards reside within $[0, 1]$, and hence the value function falls within the range $\big(0,\frac{1}{1-\gamma}\big]$. This implies that the range of the target accuracy level $\varepsilon$ should be set to $\varepsilon \in \big[0, \frac{1}{1-\gamma}\big]$.

% We don't need this for the policy evaluation development below.

%We assume throughout this section that the action set $\mathcal{A}$ contains a single action. This is without loss of generality because, for any fixed (possibly random) policy $\pi$, the induced transition dynamics can be represented by an $S\times S$ matrix obtained by marginalizing over the actions, making this effectively equivalent to a single-action setting. 

\subsection{A plug-in approach}

In view of classical Bellman equations (cf.~\eqref{eq:bellman_consistency-V-all}), the value function of policy $\pi$ satisfies (see \eqref{eq:Vpi-Bellman-inversion-matrix-inf})
\begin{align*}
    V^{\pi} = (I-\gamma P_{\pi})^{-1} r_{\pi},  
\end{align*}
which depends upon the transition kernel $P_{\pi}$ induced by $\pi$ over the state space $\cS$ (see \eqref{eqn:ppivq}), 
%, i.e.,
%$$ P_{\pi}(s^\prime \mymid s) = \mathbb{E}_{a\sim \pi(\cdot \mymid s)} \big[P(s^\prime \mymid s,a) \big] ,$$
the reward function $r_{\pi}  $ induced by $\pi$ (see \eqref{eq:defn-rpi-inf}), and the discount factor $\gamma$. 
Since the only unknown component of the model in our setting is the transition kernel $P_{\pi}\in \mathbb{R}^{S\times S}$, 
a natural strategy for policy evaluation is the ``plug-in'' approach, which starts by estimating $P_{\pi}$, 
followed by computation of the value function associated with the learned transition kernel $\widehat{P}_{\pi}$. 
This two-stage paradigm, which separates model estimation from value estimation, is dubbed as the model-based approach.  More formally, the model-based policy evaluation algorithm in the presence of a generative model can be described as: 
\begin{itemize}
    \item[1)] {\em Model estimation.} For each $s\in \cS$, draw $N$ independent samples from the generative model:
\begin{align*}
        \forall s\in\mathcal{S}: \qquad
	s^{(i)}(s) \overset{\mathrm{ind.}}{\sim} P_{\pi}(\cdot \mymid s), \qquad i=1,\ldots, N
\end{align*} 
with $P_{\pi}(\cdot \mymid s )$  defined in \eqref{eq:defn-P-s-pi-inf}, 
and compute the empirical transition kernel from state $s$ as:
\begin{equation}
	\widehat{P}_{\pi}(s^\prime \mymid s) = \frac{1}{N}\sum_{i=1}^N \ind \big\{s^{\prime} = s^{(i)} (s)  
	\big\}
    \quad \text{for all }s'\in \cS.
	\label{eq:defn-empirical-P-policy-eval}
\end{equation}
%
%where $\ind\{\cdot\}$ is the indicator function. 
In words, $\widehat{P}_{\pi}(s^\prime \mymid s)$  counts the empirical frequency of transitions from state $s$ to state $s^\prime$. 

    \item[2)] {\em Value estimation.} Compute the value function estimate as
\begin{align}
    \label{eq:value-estimation-MB-inf-vhat}
	\widehat{V}^{\pi} = (I-\gamma \widehat{P}_{\pi})^{-1} r_{\pi}.   
\end{align}
\end{itemize}

\noindent 
It is self-evident that the total number of samples drawn in the above model-based algorithm is $N^{\mathsf{total}} \coloneqq NS$.

\begin{algorithm}[t]
	%\begin{algorithmic}[1] 
    \DontPrintSemicolon
	\textbf{input:} sample size per state $N$, discount factor $\gamma$, reward function $r_{\pi}$ induced by policy $\pi$. \\
   	\For{$s \in \cS$}{
        {\color{blue}\tcc{model estimation based on samples.}}
		draw $N$ independent samples $s^{(i)}(s) \sim  P_{\pi}(\cdot \mymid s)$ (cf.~\eqref{eq:defn-P-s-pi-inf}). \\
		compute $\widehat{P}_{\pi}(\cdot \mymid s)$ based on \eqref{eq:defn-empirical-P-policy-eval}.
	%\ENDFOR
    }

       {\color{blue}\tcc{value estimation using the empirical model.}}
	\textbf{output:}  $\widehat{V}^{\pi} = (I-\gamma \widehat{P}_{\pi})^{-1} r_{\pi}$.  

   %\end{algorithmic} 
    \caption{Model-based policy evaluation with a simulator}
 \label{alg:policy-eval-model-based}
\end{algorithm}

\subsection{Theoretical guarantees}

Despite its conceptual and algorithmic simplicity, the model-based policy evaluation algorithm described in Algorithm~\ref{alg:policy-eval-model-based} turns out to achieve a form of statistical optimality in the generative model setting; namely, no other algorithms can outperform its sample complexity in a minimax sense, except possibly up to logarithmic factors. To formalize this, we first present an upper bound on the sample complexity of Algorithm~\ref{alg:policy-eval-model-based} in the theorem below, and defer the discussion of a matching lower bound to Section~\ref{sec:minimax-lower-bound-generative}. 

\begin{theorem}[\cite{li2020breaking,li2024breaking}]
    \label{thm:model-based-policy-eval}
    For any $0< \varepsilon \leq \frac{1}{1-\gamma}$ and any $0<\delta <1$, Algorithm~\ref{alg:policy-eval-model-based} achieves 
    $$
        \forall s\in \cS:\qquad | \widehat{V}^{\pi} (s) - V^{\pi} (s) | \leq \varepsilon
    $$
    with probability at least $1-\delta$, 
    provided that the sample size per state exceeds 
    \begin{equation}
	    N  \geq \frac{c_0 \log \big( \frac{S}{\delta (1-\gamma)} \big)}{(1-\gamma)^3\varepsilon^2} 
    \end{equation}
    for some sufficiently large constant $c_0>0$. 
\end{theorem}

In brief, Theorem~\ref{thm:model-based-policy-eval} demonstrates that the sample complexity of the model-based policy evaluation algorithm in Algorithm~\ref{alg:policy-eval-model-based} is no greater than 
\begin{align*}
    \widetilde{O}\bigg(\frac{S}{(1-\gamma)^3\varepsilon^2}\bigg)
   % \label{eq:sample-complexity-MB-policy-eval-inf}
\end{align*}
for any meaningful target precision level $\varepsilon$. Notably, this requirement is often significantly lower than the sample size needed to learn the transition kernel $P_{\pi}$.  For instance, when both $\frac{1}{1-\gamma}$ and $\varepsilon$ are on the constant order, this sample complexity simplifies to $\widetilde{O}(S)$, which can be $S$ times smaller than what is required to learn $P_{\pi}\in \mathbb{R}^{S\times S}$ --- an object with  $O(S^2)$ degrees of freedom. Put differently, reliable value estimation becomes statistically feasible even in regimes where faithful model estimation is not. This unveils a crucial insight: while model-based algorithms conceptually involve model estimation as a meta-task, accomplishing this meta-task with high precision is not essential for policy evaluation to succeed.

\subsection{Analysis: a first attempt}
\label{sec:first-attempt-mb-pe}

We begin with an initial attempt at bounding the value estimation error for Algorithm~\ref{alg:policy-eval-model-based}. This analysis strategy is straightforward yet yields a suboptimal bound.

Concretely, let us begin by deriving a simple characterization of the value estimation error. 
The Bellman equation~\eqref{eq:Vpi-Bellman-inf-matrix} tells us that
\[
	r_{\pi}=V^{\pi}- \gamma P_{\pi}V^{\pi} 
    =(I- \gamma P_{\pi})V^{\pi},
    %\qquad\text{and}\qquad\widehat{V}^{\pi}=r_{\pi}+\gamma\widehat{P}_{\pi} \widehat{V}^{\pi},
\]
which taken together with \eqref{eq:value-estimation-MB-inf-vhat} implies that
\begin{equation}
	\widehat{V}^{\pi}=(I-\gamma\widehat{P}_{\pi})^{-1}r_{\pi}
	 =(I-\gamma\widehat{P}_{\pi})^{-1} (I-\gamma P_{\pi})V^{\pi}.\label{eq:Vpi-Vpihat-inv-mb-pe}
\end{equation}
Invoke \eqref{eq:Vpi-Vpihat-inv-mb-pe} to decompose the value estimation error as 
\begin{align}
\widehat{V}^{\pi}-V^{\pi} 
%& =(I-\gamma\widehat{P}_{\pi})^{-1}r_{\pi}-V^{\pi}\nonumber \\
 & =(I-\gamma\widehat{P}_{\pi})^{-1}(I-\gamma P_{\pi})V^{\pi}-(I-\gamma\widehat{P}_{\pi})^{-1}(I-\gamma\widehat{P}_{\pi})V^{\pi}\nonumber \\
 & =\gamma(I-\gamma\widehat{P}_{\pi})^{-1}(\widehat{P}_{\pi}-P_{\pi})V^{\pi}.\label{eq:Vhatpi-Vpi-difference}
\end{align}

While controlling $(I-\gamma\widehat{P}_{\pi})^{-1}$ can be somewhat challenging, the other term $(\widehat{P}_{\pi}-P_{\pi})V^{\pi}$ can be readily handled by standard concentration inequalities. Specifically, 
applying the Bernstein-style inequality in Lemma~\ref{lem:Bernstein-prelim} with $A=1$ and a single vector fixed $V^{\pi}$ shows that,  with probability at least $1-\delta$, 
%recall that for any given $s \in [S]$, the $s$-th row of $\widehat{P}_{\pi}$ can be viewed as the sum of $N$ independent vectors: 
%
%\[
%	\big(\widehat{P}_{\pi}\big)_{s,:} = \frac{1}{N} \sum_{j=1}^N e_{s^{(i)}(s)}
%\]
%
%with $e_j$ being the $j$-th standard basis vector and $s^{(i)}(s)\sim P_\pi(\cdot \mid s)$. The Bernstein inequality asserts that with probability at least $1-\delta/S$,   
%
%\[
%	\big| \big(\widehat{P}_{\pi} - P_{\pi}\big)_{s,:} V^{\pi} \big| \leq C_0 \Bigg( \sqrt{\frac{\mathsf{Var}_{P_{\pi}(\cdot | s)}\big(V^{\pi}\big)\log\frac{S}{\delta}}{N}}+\frac{\|V^{\pi}\|_{\infty}\log\frac{S}{\delta}}{N} \Bigg)
%\]
%
%holds for some universal constant $C_0>0$. Take the union bound over all $s\in \mathcal{S}$ to arrive at
%
\begin{align}
\big|(\widehat{P}_{\pi}-P_{\pi})V^{\pi}\big| &  \leq C_0 \Bigg( \sqrt{\frac{\mathsf{Var}_{P_{\pi}}\big(V^{\pi}\big)\log\big(\frac{S}{\delta}\big)}{N}}+\frac{\|V^{\pi}\|_{\infty}\log\big(\frac{S}{\delta}\big)}{N}1_{S}
	\Bigg)
	\label{eq:Bernstein-Phat-P-V}	
\end{align}
%
%with probability at least $1-\delta$, 
for some universal constant $C_0>0$, 
where $1_{S}\in \mathbb{R}^S$ is the all-one vector.

Regarding the term $(I-\gamma\widehat{P}_{\pi})^{-1}$, 
it follows from Lemma~\ref{lem:basic-property-matrix-notation-inf} that
$(I-\gamma\widehat{P}_{\pi})^{-1}$ is a nonnegative matrix and obeys
\begin{align}
	(I-\gamma\widehat{P}_{\pi})^{-1}1_{S} %&=\bigg(I+\sum_{i=1}^{\infty}(\gamma\widehat{P}_{\pi})^{i}\bigg)1_{S}=\bigg(1+\sum_{i=1}^{\infty}\gamma^{i}\bigg)1_{S} \notag\\
	&=\frac{1}{1-\gamma}1_{S}.
	\label{eq:sum-I-minu-gammaP-expand}
\end{align}
Combining these properties with \eqref{eq:Vhatpi-Vpi-difference} and \eqref{eq:Bernstein-Phat-P-V} shows that
\begin{align}
 & \big|(I-\gamma\widehat{P}_{\pi})^{-1}(\widehat{P}_{\pi}-P_{\pi})V^{\pi}\big|\leq(I-\gamma\widehat{P}_{\pi})^{-1}\big|(\widehat{P}_{\pi}-P_{\pi})V^{\pi}\big|\nonumber \\
 & \leq C_{0}\bigg\{\frac{(I-\gamma\widehat{P}_{\pi})^{-1}\sqrt{\mathsf{Var}_{P_{\pi}}(V^{\pi})\log\big(\frac{S}{\delta}\big)}}{\sqrt{N}}+\frac{\|V^{\pi}\|_{\infty}\log\big(\frac{S}{\delta}\big)}{N}(I-\gamma\widehat{P}_{\pi})^{-1}1_{S}\bigg\}	\notag \\
 &  = C_{0}\bigg\{\frac{(I-\gamma\widehat{P}_{\pi})^{-1}\sqrt{\mathsf{Var}_{P_{\pi}}(V^{\pi})\log\big(\frac{S}{\delta}\big)}}{\sqrt{N}}+\frac{\|V^{\pi}\|_{\infty}\log\big(\frac{S}{\delta}\big)}{(1-\gamma)N}1_{S}\bigg\}
	\label{eq:I-gammaP-prodct-UB1}
\end{align}
occurs with probability exceeding $1-\delta$.

We then attempt to bound $(I-\gamma\widehat{P}_{\pi})^{-1}$ and $\sqrt{\mathsf{Var}_{P_{\pi}}(V^{\pi})}$ separately:  
\begin{align}
	\Big\|(I-\gamma\widehat{P}_{\pi})^{-1}\sqrt{\mathsf{Var}_{P_{\pi}}(V^{\pi})} \Big\|_{\infty}
	& \leq \big\|(I-\gamma\widehat{P}_{\pi})^{-1} \big\|_{1} \sqrt{\big\|\mathsf{Var}_{P_{\pi}}(V^{\pi}) \big\|_{\infty}} \notag\\
	&= \frac{1}{1-\gamma} \sqrt{\big\|\mathsf{Var}_{P_{\pi}}(V^{\pi}) \big\|_{\infty}} \notag\\
	& \leq \frac{1}{(1-\gamma)^2},
	\label{eq:Vhatpi-Vpi-difference-decoupled}
\end{align}
where the penultimate relation arises from Lemma~\ref{lem:basic-property-matrix-notation-inf}, 
and the last line is valid since $\mathsf{Var}_{q}(V^{\pi})\leq\mathbb{E}_{q}[V^{\pi}\circ V^{\pi}]\leq (1-\gamma)^{-2}$ for any probability distribution $q$.  
Substituting the above result into \eqref{eq:Vhatpi-Vpi-difference} and \eqref{eq:I-gammaP-prodct-UB1} and applying the basic inequality $\|V^{\pi}\|_{\infty} \leq (1-\gamma)^{-1}$ 
give 
\begin{align}
\|\widehat{V}^{\pi}-V^{\pi}\|_{\infty} & =\gamma\big\|(I-\gamma\widehat{P}_{\pi})^{-1}(\widehat{P}_{\pi}-P_{\pi})V^{\pi}\big\|_{\infty}\nonumber \\
 & \leq C_{0}\Bigg\{\sqrt{\frac{\log\big(\frac{S}{\delta}\big)}{(1-\gamma)^{4}N}}+\frac{\log\big(\frac{S}{\delta}\big)}{(1-\gamma)^{2}N}\Bigg\}
	\label{eq:eq:Vhatpi-Vpi-difference-decoupled-Hoeffding}
\end{align}
with probability exceeding $1-\delta$.

The high-probability bound \eqref{eq:eq:Vhatpi-Vpi-difference-decoupled-Hoeffding} reveals that: in order to attain $\varepsilon$-accurate value estimation (in an $\ell_{\infty}$ sense), 
it suffices to draw $$N=\widetilde{O}\bigg(\frac{1}{(1-\gamma)^4\varepsilon^2}\bigg)$$ samples per state from the generative model.  
Nevertheless, this simple analysis incurs an extra multiplicative factor of $\frac{1}{1-\gamma}$ compared to the result stated in Theorem~\ref{thm:model-based-policy-eval}, thus falling short of tightness for problems with long effective horizons. 
Upon closer inspection,  the separate treatment of $(I-\gamma\widehat{P}_{\pi})^{-1}$ and $\sqrt{\mathsf{Var}_{P_{\pi}}(V^{\pi})}$ overlooks their intimate connection that could allow for tighter control. 
Furthermore, we have not fully exploited the variance structure in \eqref{eq:Bernstein-Phat-P-V}; only crude upper bounds are applied. 
These factors together account for the suboptimality of the above analysis.

\subsection{Analysis: a second attempt}
\label{sec:analysis-mb-pe-2nd-attempt}
 
We now refine our analysis for the value estimation error of interest, leveraging the interaction between $(I-\gamma\widehat{P}_{\pi})^{-1}$ and $\sqrt{\mathsf{Var}_{P_{\pi}}(V^{\pi})}$, 
as well as the underlying variance structure. 
To simplify the presentation and concentrate our discussion on the main term, 
we first note --- without a formal proof --- that, for $N$ large enough, 
\[
	(I-\gamma\widehat{P}_{\pi})^{-1}\sqrt{\mathsf{Var}_{P_{\pi}}(V^{\pi})} \approx (I-\gamma P_{\pi})^{-1}\sqrt{\mathsf{Var}_{P_{\pi}}(V^{\pi})}.
\]
This motivates us to control $(I-\gamma P_{\pi})^{-1}\sqrt{\mathsf{Var}_{P_{\pi}}(V^{\pi})}$.

Recall from Lemma~\ref{lem:basic-property-matrix-notation-inf} that $(I-\gamma P_{\pi})^{-1}$ is a nonnegative matrix with each row summing to $\frac{1}{1-\gamma}$, thus indicating that $(1-\gamma)(I-\gamma P_{\pi})^{-1}$ is a valid probability transition matrix. 
This inspires one to push $(I-\gamma P_{\pi})^{-1}$ into the square root operator, yielding 
\begin{align}
(I-\gamma P_{\pi})^{-1}\sqrt{\mathsf{Var}_{P_{\pi}}(V^{\pi})}
	& = \frac{1}{1-\gamma}\Big\{ (1-\gamma) (I-\gamma P_{\pi})^{-1}\sqrt{\mathsf{Var}_{P_{\pi}}(V^{\pi})} \Big\}
\notag \\
	& \leq \frac{1}{1-\gamma} \sqrt{(1-\gamma)(I-\gamma P_{\pi})^{-1}\mathsf{Var}_{P_{\pi}}(V^{\pi})} 
\notag \\
	&=  \sqrt{\frac{1}{1-\gamma}(I-\gamma P_{\pi})^{-1}\mathsf{Var}_{P_{\pi}}(V^{\pi})}.
	\label{eq:Jensen-I-gamma-P-Var}
\end{align}
Here, we apply Jensen's inequality: for any probability vector $q$ and any nonnegative random variable $v$, it holds that  $\mathbb{E}_q[\sqrt{v}]\leq \sqrt{\mathbb{E}_q[v]}$.

To better exploit the underlying variance structure, rewrite and upper-bound $\mathsf{Var}_{P_{\pi}}(V^{\pi})$ (cf.~\eqref{eq:matrix-VarPpi-V-expression}) as follows: 
\begin{align*}
\mathsf{Var}_{P_{\pi}}(V^{\pi}) & =P_{\pi}(V^{\pi}\circ V^{\pi})-(P_{\pi}V^{\pi})\circ(P_{\pi}V^{\pi})\\
 & =P_{\pi}(V^{\pi}\circ V^{\pi})-\frac{1}{\gamma^{2}}(V^{\pi}-r_{\pi})\circ(V^{\pi}-r_{\pi})\\
 & =\frac{1}{\gamma^{2}}(\gamma^{2}P_{\pi}-I)(V^{\pi}\circ V^{\pi})+\frac{2}{\gamma^{2}}V^{\pi}\circ r_{\pi}-\frac{1}{\gamma^{2}}r_{\pi}\circ r_{\pi}\\
 & \leq\frac{1}{\gamma^{2}}(\gamma P_{\pi}-I)(V^{\pi}\circ V^{\pi})+\frac{2}{\gamma^{2}}V^{\pi}\circ r_{\pi},
\end{align*}
where the second line applies the Bellman equation $V^{\pi}=r_{\pi}+\gamma P_{\pi}V^{\pi}$ (cf.~\eqref{eq:Vpi-Bellman-inf-matrix}).
This yields a simple upper bound on  $(I-\gamma P_{\pi})^{-1}\mathsf{Var}_{P_{\pi}}(V^{\pi})$: 
\begin{align}
	0 & \leq(I-\gamma P_{\pi})^{-1}\mathsf{Var}_{P_{\pi}}(V^{\pi})\notag\\
 & \leq\frac{1}{\gamma^{2}}(I-\gamma P_{\pi})^{-1}(\gamma P_{\pi}-I)(V^{\pi}\circ V^{\pi})+\frac{2}{\gamma^{2}}(I-\gamma P_{\pi})^{-1}(V^{\pi}\circ r_{\pi})\notag\\
 & =-\frac{1}{\gamma^{2}}V^{\pi}\circ V^{\pi}+\frac{2}{\gamma^{2}}(I-\gamma P_{\pi})^{-1}(V^{\pi}\circ r_{\pi})\notag\\
 & \leq\frac{2}{\gamma^{2}}(I-\gamma P_{\pi})^{-1}(V^{\pi}\circ r_{\pi})\notag\\
 & \leq\frac{2\|V^{\pi}\|_{\infty}}{\gamma^{2}}(I-\gamma P_{\pi})^{-1}r_{\pi}\notag\\
 & =\frac{2\|V^{\pi}\|_{\infty}}{\gamma^{2}}V^{\pi}\leq\frac{2\|V^{\pi}\|_{\infty}^{2}}{\gamma^{2}}1_{S},\label{eq:I-gamma-P-Var-bound}
\end{align}
where the last line uses \eqref{eq:Vpi-Bellman-inversion-matrix-inf}. 
%the penultimate relation uses $\|r_{\pi}\|_{\infty}\leq1$. 
Taking this together with \eqref{eq:Jensen-I-gamma-P-Var} yields
\begin{align}
	0 \leq \Big\| (I-\gamma P_{\pi})^{-1}\sqrt{\mathsf{Var}_{P_{\pi}}(V^{\pi})} \Big\|_{\infty }\leq
	\sqrt{\frac{2}{1-\gamma}}\frac{\|V^{\pi}\|_{\infty}}{\gamma} . 
	\label{eq:lemma-VarP-V-bound}
\end{align}
Substituting this into \eqref{eq:Vhatpi-Vpi-difference} and \eqref{eq:I-gammaP-prodct-UB1}, and 
using the basic fact that $\|V^{\pi}\|_{\infty}\leq\frac{1}{1-\gamma}$, we arrive at
\begin{align*}
\|\widehat{V}^{\pi}-V^{\pi}\|_{\infty} & \leq C_{0}\Bigg\{\frac{\gamma}{\sqrt{N}}\Big\|(I-\gamma\widehat{P}_{\pi})^{-1}\sqrt{\mathsf{Var}_{P_{\pi}}(V^{\pi})}\Big\|_{\infty}+\frac{\gamma}{(1-\gamma)^{2}N}\Bigg\}\\
 & \approx C_{0}\Bigg\{\frac{\gamma}{\sqrt{N}}\Big\|(I-\gamma P_{\pi})^{-1}\sqrt{\mathsf{Var}_{P_{\pi}}(V^{\pi})}\Big\|_{\infty}+\frac{\gamma}{(1-\gamma)^{2}N}\Bigg\}\\
 & \leq C_{0}\Bigg\{\sqrt{\frac{2}{N(1-\gamma)^{3}}}+\frac{1}{(1-\gamma)^{2}N}\Bigg\}\\
 & \leq2C_{0}\sqrt{\frac{2}{N(1-\gamma)^{3}}},
\end{align*}
provided that $N$ is sufficiently large. This reveals that $\varepsilon$-accurate value estimation is attainable --- at least in the asymptotic regime with $N$ (resp.~$\varepsilon$) approaching infinity (resp.~0) --- when the sample size per state exceeds the following order 
\[
	N = \widetilde{O}\bigg( \frac{1}{(1-\gamma)^3\varepsilon^2} \bigg),
\]
thereby matching the advertised result in Theorem~\ref{thm:model-based-policy-eval}.

However, an obvious limitation of the above analysis stems from the stringent requirement in $N$: the sharpness of the above analysis is only guaranteed for very large $N$. Additional efforts are needed to extend the claimed results to the data-limited regime, i.e., the situation where $\varepsilon$ is not exceedingly small.

\subsection{Sharp analysis: proof of Theorem~\ref{thm:model-based-policy-eval}}

In order to further improve the analysis strategy outlined in Section~\ref{sec:analysis-mb-pe-2nd-attempt}, a key step lies in controlling the following term:
\begin{equation}
	\Big((I-\gamma\widehat{P}_{\pi})^{-1}-(I-\gamma P_{\pi})^{-1}\Big) \sqrt{\mathsf{Var}_{P_{\pi}}(V^{\pi})},
	\label{eq:key-quantity-sharp-analysis-MB-PE}
\end{equation}
as it determines the range of $N$ within which optimality can be achieved. 
Interestingly, the analyses presented in Sections~\ref{sec:first-attempt-mb-pe} and \ref{sec:analysis-mb-pe-2nd-attempt} already shed light on how to handle this term. 
More specifically, if we express the value estimation error as
\begin{align*}
	\widehat{V}^{\pi} - V^{\pi} = \Big((I-\gamma\widehat{P}_{\pi})^{-1}-(I-\gamma P_{\pi})^{-1}\Big)r_{\pi}
\end{align*}
according to the Bellman equation~\eqref{eq:Vpi-Bellman-inversion-matrix-inf}, then the term \eqref{eq:key-quantity-sharp-analysis-MB-PE} can be interpreted as a sort of value estimation error with the reward function taken to be $r^{(1)}=\sqrt{\mathsf{Var}_{P_{\pi}}(V^{\pi})}$, which can be tackled using arguments in Section~\ref{sec:analysis-mb-pe-2nd-attempt}. 
However, applying this approach results in a similar issue:  the need to bound an even ``higher-order'' term
\[
\Big((I-\gamma\widehat{P}_{\pi})^{-1}-(I-\gamma P_{\pi})^{-1}\Big) \sqrt{\mathsf{Var}_{P_{\pi}}(V^{(1)})}
\]
with $V^{(1)}\coloneqq(I-\gamma\Pv)^{-1}r^{(1)}
%\sqrt{\mathsf{Var}_{\Pv}\big[V^{\pi}\big]}
$. 
Motivated by this line of thought, we propose a recursive analysis approach, with the aim of controlling the residual terms in an increasingly more refined fashion.

\paragraph{Step 1: recursive construction of auxiliary rewards and values.} 
To apply the above recursive arguments, 
 construct a sequence of auxiliary ``reward'' and ``value'' functions recursively: 
\begin{subequations}
	\label{eqn:superV}
\begin{align}
	\label{eqn:superV-true}
\begin{array}{lll}
	r^{(0)} \coloneqq r_{\pi}, & V^{(0)} \coloneqq (I-\gamma\Pv)^{-1}r^{(0)}, \\
	r^{(l)}\coloneqq\sqrt{\mathsf{Var}_{\Pv}\big[V^{(l-1)}\big]}, & V^{(l)} \coloneqq (I-\gamma\Pv)^{-1}r^{(l)}, & l\geq 1. \\
\end{array}
\end{align}
Similarly, we find it convenient to introduce the auxiliary ``value estimates'' as follows:
\begin{align}
	\label{eqn:superV-hat}
	\begin{array}{lll}
		& \widehat{V}^{(l)}\coloneqq(I-\gamma\Phatv)^{-1}r^{(l)}, \qquad l\geq 0. 
	\end{array}
\end{align}
\end{subequations}

As an immediate consequence of inequality~\eqref{eq:lemma-VarP-V-bound}, 
we can readily establish an entrywise bound on $V^{(l)}$ for any $l>1$: 
\begin{align}
	\big\|V^{(l)}\big\|_{\infty} & = \big\|(I-\gamma\Pv)^{-1}r^{(l)}\big\|_{\infty}  =\Big\|(I-\gamma\Pv)^{-1}\sqrt{\mathsf{Var}_{\Pv}\big[V^{(l-1)}\big]}\Big\|_{\infty} \notag\\
	& \leq \frac{\sqrt{2}}{\gamma\sqrt{1-\gamma}}\big\|V^{(l-1)}\big\|_{\infty}
	 \leq \cdots \leq \bigg( \frac{\sqrt{2}}{\gamma\sqrt{1-\gamma}} \bigg)^l \big\|V^{(0)}\big\|_{\infty}\label{eq:Vl-inf-norm-bound-0} \\
	 &= \bigg( \frac{\sqrt{2}}{\gamma\sqrt{1-\gamma}} \bigg)^l \big\|V^{\pi}\big\|_{\infty} 
	  \leq \frac{1}{1-\gamma} \bigg( \frac{\sqrt{2}}{\gamma\sqrt{1-\gamma}} \bigg)^l,
	  \label{eq:Vl-inf-norm-bound}
\end{align}
where the last line holds since $\|V^{\pi}\|_{\infty}\leq \frac{1}{1-\gamma}$.

\paragraph*{Step 2: establishing a recursion for $\widehat{V}^{(l)}-V^{(l)}$.}
Equipped with the above notation, we observe that, for any $l \geq 0$, 
\begin{align}
\widehat{V}^{(l)}-V^{(l)} & =(I-\gamma\Phatv)^{-1}r^{(l)}-V^{(l)}\nonumber \\
 & =\big(I-\gamma\Phatv\big)^{-1} \big(I-\gamma\Pv\big)V^{(l)} -\big(I-\gamma\Phatv\big)^{-1}\big(I-\gamma\Phatv\big)V^{(l)}\nonumber \\
 & =\gamma\big(I-\gamma\Phatv\big)^{-1}\big(\Phatv-\Pv\big)V^{(l)},\label{eq:Vl-Vhat-l-perturbation}
\end{align}
where the second line follows from the identity 
$\big(I-\gamma\Pv\big)V^{(l)}=r^{(l)}$ (see the construction \eqref{eqn:superV-true}).

The Bernstein inequality in Lemma~\ref{lem:Bernstein-prelim} together with the union bound asserts that, with probability exceeding $1-\delta$, one has 
\begin{equation}
	\big|\big(\Phatv-\Pv\big)V^{(l)}\big|\leq\sqrt{\frac{\beta_1}{N}}\sqrt{\mathsf{Var}_{\Pv}\big[V^{(l)}\big]}+\frac{\big\|V^{(l)}\big\|_{\infty}\beta_1}{N} 1_S
	\label{eq:Bernstein-Vl}
\end{equation}
uniformly for all $0\leq l\leq m$, where $m$ is some large enough integer to be specified shortly, and $\beta_1 = C_0 \log \frac{Sm}{\delta}$ for some large enough constant $C_0>0$.  
Combining this result with (\ref{eq:Vl-Vhat-l-perturbation}) reveals that 
\begin{align}
\arxiv{&}	 \big\|\widehat{V}^{(l)}-V^{(l)}\big\|_{\infty} \book{&} =\gamma\big\|\big(I-\gamma\Phatv\big)^{-1}\big(\Phatv-\Pv\big)V^{(l)}\big\|_{\infty} \nonumber\\
 & \arxiv{\qquad}  \overset{(\mathrm{i})}{\leq}\gamma \Big\| \big(I-\gamma\Phatv\big)^{-1}\big|\big(\Phatv-\Pv\big)V^{(l)} \big|\Big\|_{\infty}\nonumber\\
	&  \arxiv{\qquad}  \overset{(\mathrm{ii})}{\leq}\gamma\sqrt{\frac{\beta_{1}}{N}} \Big\| \big(I-\gamma\Phatv\big)^{-1}\sqrt{\mathsf{Var}_{\Pv}\big[V^{(l)}\big]} \Big\|_{\infty}
	+\frac{\gamma\big\| V^{(l)}\big\|_{\infty}\beta_{1}}{(1-\gamma)N} \label{eq:upper-bound-Vl-Vhatl-00}\\
	&  \arxiv{\qquad}  \overset{(\mathrm{iii})}{\leq}\gamma\sqrt{\frac{\beta_{1}}{N}}\big\|\widehat{V}^{(l+1)}-V^{(l+1)}\big\|_{\infty}+\gamma\sqrt{\frac{\beta_{1}}{N}} \big\|V^{(l+1)}\big\|_{\infty}+\frac{\gamma\big\| V^{(l)}\big\|_{\infty}\beta_{1}}{(1-\gamma)N}	\notag\\
 &  \arxiv{\qquad}  \overset{(\mathrm{iv})}{\leq}\gamma\sqrt{\frac{\beta_{1}}{N}}\big\|\widehat{V}^{(l+1)}-V^{(l+1)}\big\|_{\infty}+\bigg(\sqrt{\frac{2\beta_{1}}{(1-\gamma)N}}+\frac{\beta_{1}}{(1-\gamma)N}\bigg)\big\| V^{(l)}\big\|_{\infty}	
\label{eq:upper-bound-Vl-Vhatl}
\end{align}
with probability exceeding $1-\delta$.  
Here, (i) follows since $\big(I-\gamma\Phatv\big)^{-1}$
is a nonnegative matrix, (ii) comes from (\ref{eq:Bernstein-Vl})
and the triangle inequality,  (iii) arises from  the definitions of $r^{(l)}$, $V^{(l)}$
and $\widehat{V}^{(l)}$, while (iv) is a consequence of \eqref{eq:Vl-inf-norm-bound-0}. 
Crucially, inequality \eqref{eq:upper-bound-Vl-Vhatl} establishes a recursive relation that enables a sharp upper bound analysis.

\paragraph*{Step 3: upper bounding via the above recursion.}
Define 
\begin{align}
b_{1} \coloneqq \gamma\sqrt{\frac{\beta_{1}}{N}},
~~ 
b_{2} \coloneqq \sqrt{\frac{2\beta_1}{(1-\gamma)N}} + \frac{\beta_1}{(1-\gamma)N},
	~~ b_{3} \coloneqq \frac{\sqrt{2}}{\gamma\sqrt{1-\gamma}}
	\label{eq:defn-b123-mb-pe}
\end{align}
for notational simplicity. 
Combining inequalities (\ref{eq:Vl-inf-norm-bound}) and (\ref{eq:upper-bound-Vl-Vhatl}) shows that with probability at least $1-\delta$,  
%
%\begin{align*}
%  \big\|\widehat{V}^{(0)}-V^{(0)}\big\|_{\infty} 
%	& \leq  
% 	b_1 \big\|\widehat{V}^{(1)}-V^{(1)}\big\|_{\infty} + b_2\big\|V^{(0)}\big\|_{\infty}	,
%   \end{align*}
%   %
%and for any $1\leq l < m$,
%
\begin{align*}  
	\big\|\widehat{V}^{(l)}-V^{(l)}\big\|_{\infty} 
  %& \leq\gamma\sqrt{\frac{\beta_{1}}{N}}\big\|\widehat{V}^{(l+1)}-V^{(l+1)}\big\|_{\infty}
 % +
 % \left(4\sqrt{\frac{\beta_1}{(1-\gamma)N}} + \frac{\gamma\beta_1}{(1-\gamma)N}\right)
  % \Big(\frac{4}{\gamma}\Big)^{l+1}\left\{ \sqrt{\frac{16\gamma^{2}\beta_{1}}{N(1-\gamma)}}+\frac{\gamma^{2}\beta_{1}}{(1-\gamma)N}\right\} 
 % \left(\frac{4}{\gamma\sqrt{1-\gamma}}\right)^{l-1}\|\bm V^{(1)}\|_{\infty}\\
 %& =: 
&\leq b_{1}\big\|\widehat{V}^{(l+1)}-V^{(l+1)}\big\|_{\infty}+b_{2}\| V^{(l)}\|_{\infty} \notag\\	
 &\leq b_{1}\big\|\widehat{V}^{(l+1)}-V^{(l+1)}\big\|_{\infty}+b_{2}b_{3}^{l}\| V^{(0)}\|_{\infty}
\end{align*}
holds for all $0\leq l < m$. 
Apply this relation recursively and combine terms to arrive at
\begin{align}
%\big\|\widehat{V}^{(0)}-V^{(0)}\big\|_{\infty} & \leq\underbrace{b_{1}^{m}\big\|\widehat{V}^{(m)}-V^{(m)}\big\|_{\infty}}_{\eqqcolon \alpha_{1}}+
%\underbrace{\Big( b_1b_{2}\sum_{l=0}^{m-2}(b_{1}b_{3})^{l} + b_1 \Big)}_{\eqqcolon \alpha_{2}}\big\|V^{(1)}\big\|_{\infty}
%   + \frac{\gamma\beta_{1}}{(1-\gamma)N}\big\|V^{(0)}\big\|_{\infty}.
\big\|\widehat{V}^{(0)}-V^{(0)}\big\|_{\infty} & \leq b_{1}\big\|\widehat{V}^{(1)}-V^{(1)}\big\|_{\infty}+b_{2}\|V^{(0)}\|_{\infty} \notag\\
 & \leq b_{1}\Big\{ b_{1}\big\|\widehat{V}^{(2)}-V^{(2)}\big\|_{\infty}+b_{2}b_{3}\|V^{(0)}\|_{\infty}\Big\}+b_{2}\|V^{(0)}\|_{\infty} \notag\\
 & \leq\cdots \notag\\
 & \leq\underbrace{b_{1}^{m}\big\|\widehat{V}^{(m)}-V^{(m)}\big\|_{\infty}}_{\eqqcolon\,\alpha_{1}}+\underbrace{b_{2}\sum_{l=0}^{m-1}(b_{1}b_{3})^{l}\big\| V^{(0)}\big\|_{\infty}}_{\eqqcolon\,\alpha_{2}},	
\label{eqn:Chopin}
\end{align}
leaving us with two terms to control.

\paragraph*{Step 4: controlling quantity $\alpha_{1}$.} 
To bound quantity $\alpha_{1}$, we invoke identity \eqref{eq:upper-bound-Vl-Vhatl-00} to obtain
\begin{align*}
\arxiv{&} \big\|\widehat{V}^{(m)}-V^{(m)}\big\|_{\infty} \book{&} \leq\gamma\sqrt{\frac{\beta_{1}}{N}}\Big\|\big(I-\gamma\Phatv\big)^{-1}\sqrt{\mathsf{Var}_{\Pv}\big[V^{(m)}\big]}\Big\|_{\infty}+\frac{\gamma\beta_{1}\big\| V^{(m)}\big\|_{\infty}}{(1-\gamma)N}\\
 & \arxiv{\qquad}\leq\gamma\sqrt{\frac{\beta_{1}}{N}}\Big\|\big(I-\gamma\Phatv\big)^{-1}\Big\|_{1}\Big\|\sqrt{\mathsf{Var}_{\Pv}\big[V^{(m)}\big]}\Big\|_{\infty}+\frac{\gamma\beta_{1}\big\| V^{(m)}\big\|_{\infty}}{(1-\gamma)N}\\
 & \arxiv{\qquad}\leq\frac{\gamma}{1-\gamma}\sqrt{\frac{\beta_{1}}{N}}\big\| V^{(m)}\big\|_{\infty}+\frac{\gamma\beta_{1}\big\| V^{(m)}\big\|_{\infty}}{(1-\gamma)N},
\end{align*}
where the last inequality uses the basic fact $\|\big(I-\gamma\Phatv\big)^{-1}\|_{1}=\frac{1}{1-\gamma}$ (cf.~\eqref{eq:sum-I-minu-gammaP-expand5-prelim})
as well as the elementary relation $\mathsf{Var}_{\Pv}\big[V^{(m)}\big]\leq\big\|V^{(m)}\big\|^2_{\infty}$. 
This combined with inequality~\eqref{eq:Vl-inf-norm-bound} yields
\begin{align*}
\big\|\widehat{V}^{(m)}-V^{(m)}\big\|_{\infty}\leq\frac{\gamma}{(1-\gamma)^{2}}\Bigg(\sqrt{\frac{\beta_{1}}{N}}+\frac{\beta_{1}}{N}\Bigg)\Bigg(\frac{\sqrt{2}}{\gamma\sqrt{1-\gamma}}\Bigg)^{m}.
 % & =\Bigg(\sqrt{\frac{\beta_{1}}{N}}+\frac{\beta_{1}}{N}\Bigg)\frac{4^{m+1}}{\gamma^{m}(\sqrt{1-\gamma})^{m+4}}\|r\|_{\infty}.\nonumber
\end{align*}
As a consequence, we arrive at 
\begin{align}
\alpha_{1} & =b_{1}^{m}\big\|\widehat{V}^{(m)}-V^{(m)}\big\|_{\infty}\notag\\
 & \leq\left(\gamma\sqrt{\frac{\beta_{1}}{N}}\right)^{m}\frac{\gamma}{(1-\gamma)^{2}}\Bigg(\sqrt{\frac{\beta_{1}}{N}}+\frac{\beta_{1}}{N}\Bigg)\Big(\frac{\sqrt{2}}{\gamma\sqrt{1-\gamma}}\Big)^{m}\nonumber\\
 & \leq\left(\sqrt{\frac{2\beta_{1}}{(1-\gamma)N}}\right)^{m}\frac{\gamma}{(1-\gamma)^{2}}\Bigg(\sqrt{\frac{\beta_{1}}{N}}+\frac{\beta_{1}}{N}\Bigg)\nonumber\\
 & \leq\left(\frac{1}{e}\right)^{m}\frac{2}{(1-\gamma)^{2}}\sqrt{\frac{\beta_{1}}{N}},
%	\leq\left(\frac{1}{e}\right)^{m}\frac{1}{\sqrt{(1-\gamma)^{3}}}\sqrt{\frac{4\beta_{1}}{(1-\gamma)N}},
 \label{eq:T1-bound}
\end{align}
where the last line holds as long as $N>\frac{2e^{2}}{1-\gamma}\beta_{1}$.

\paragraph*{Step 5: controlling quantity $\alpha_{2}$.}
Regarding quantity $\alpha_{2}$, 
we note that $b_{1}b_{3}\leq1/2$ holds as long as $N\geq 64\beta_1/(1-\gamma)$. 
Summing the terms of the geometric sequence leads to 
\begin{align*}
	 b_{2}\sum_{l=0}^{m-1}(b_{1}b_{3})^{l}  
\leq \frac{b_{2}}{1-b_{1}b_{3}}  
\leq \sqrt{\frac{8\beta_{1}}{(1-\gamma)N}}+\frac{2\beta_{1}}{(1-\gamma)N},
\end{align*}
which taken collectively with the bound $\|V^{(0)}\|_{\infty}\leq \frac{1}{1-\gamma}$ gives
\begin{align}
	\alpha_{2} &=  \bigg\{ b_{2}\sum_{l=0}^{m-1}(b_{1}b_{3})^{l}  \bigg\} \|V^{(0)}\|_{\infty}
	\leq \sqrt{\frac{8\beta_{1}}{(1-\gamma)^{3}N}}+\frac{2\beta_{1}}{(1-\gamma)^{2}N},
	\label{eq:T2-bound}
\end{align}
provided that $N\geq\frac{64\beta_{1}}{1-\gamma}$.

\paragraph*{Step 6: putting all this together.}
Taking inequalities~\eqref{eqn:Chopin},  \eqref{eq:T1-bound} and \eqref{eq:T2-bound} together shows that, with probability at least $1-\delta$, 
\begin{align*}
  \big\|\widehat{V}^{(0)}-V^{(0)}\big\|_{\infty}
	\leq \left(\frac{1}{e}\right)^{m}\frac{2}{(1-\gamma)^{2}}\sqrt{\frac{\beta_{1}}{N}} +
	\sqrt{\frac{8\beta_{1}}{(1-\gamma)^{3}N}}+\frac{2\beta_{1}}{(1-\gamma)^{2}N}
	%\left(\frac{1}{e}\right)^{m-1} \frac{1.1\gamma^2\beta_{1}}{(1-\gamma)N} \right\} \| V^{(1)}\|_{\infty} . 
\end{align*}
holds as long as $N\geq\frac{64\beta_{1}}{1-\gamma}$.  
To finish up, set $m=\lceil\log (\frac{e}{1-\gamma})\rceil$ to derive
\begin{align}
	\big\|\widehat{V}^{\pi}-V^{\pi}\big\|_{\infty} & =\big\|\widehat{V}^{(0)}-V^{(0)}\big\|_{\infty}\leq2\sqrt{\frac{8\beta_{1}}{(1-\gamma)^{3}N}}+\frac{2\beta_{1}}{(1-\gamma)^{2}N} \notag\\
 & \leq 3\sqrt{\frac{8\beta_{1}}{(1-\gamma)^{3}N}}
 	  \label{eqn:ice-cream}
\end{align}
with probability exceeding $1-\delta$, 
provided that $N\geq\frac{ \beta_{1}}{1-\gamma}$.  
This concludes the proof by recalling that $\beta_1 = C_0 \log (\frac{Sm}{\delta})$ and $m=\lceil\log (\frac{e}{1-\gamma})\rceil$.

\section{Model-based policy learning}
\label{sec:model-based-learning-sim}

Having established the effectiveness of the model-based approach for policy evaluation, 
we now turn attention to model-based policy learning, 
which seeks to identify a policy that approximately maximizes the value function based on the data samples at hand. 
To be precise, suppose that the action space $\cA$ contains $A\geq 2$ actions, then for any target accuracy level $\varepsilon>0$, the aim of policy learning is to compute an $\varepsilon$-accurate policy ${\pi}_{\mathsf{est}}$ obeying
\begin{align*}
	\forall s\in\cS: \qquad V^{{\pi}_{\mathsf{est}}}(s) \geq V^{\star}(s) -\varepsilon.
	%, \quad  Q^{{\pi}_{\mathsf{est}}}(s,a) \geq Q^{\star}(s,a) -\varepsilon.
\end{align*}
Akin to the policy evaluation case, the meaningful range of target precision level $\varepsilon$ is $\varepsilon \in \big(0,\frac{1}{1-\gamma}\big]$. 
Naturally, one would hope to achieve policy learning using as few samples as possible.

\subsection{Algorithms}

A model-based policy learning algorithm typically starts by constructing an empirical MDP $\widehat{\mathcal{M}}$ based on all collected data samples, mirroring what is used in model-based policy evaluation. The empirical model is then fed into the Bellman recursion --- via methods such as policy iteration and value iteration introduced in Section~\ref{sec:prelim-DP-inf} --- to derive the optimal policy for the empirical model.

\begin{figure}
\begin{center}
\includegraphics[width=0.6\textwidth]{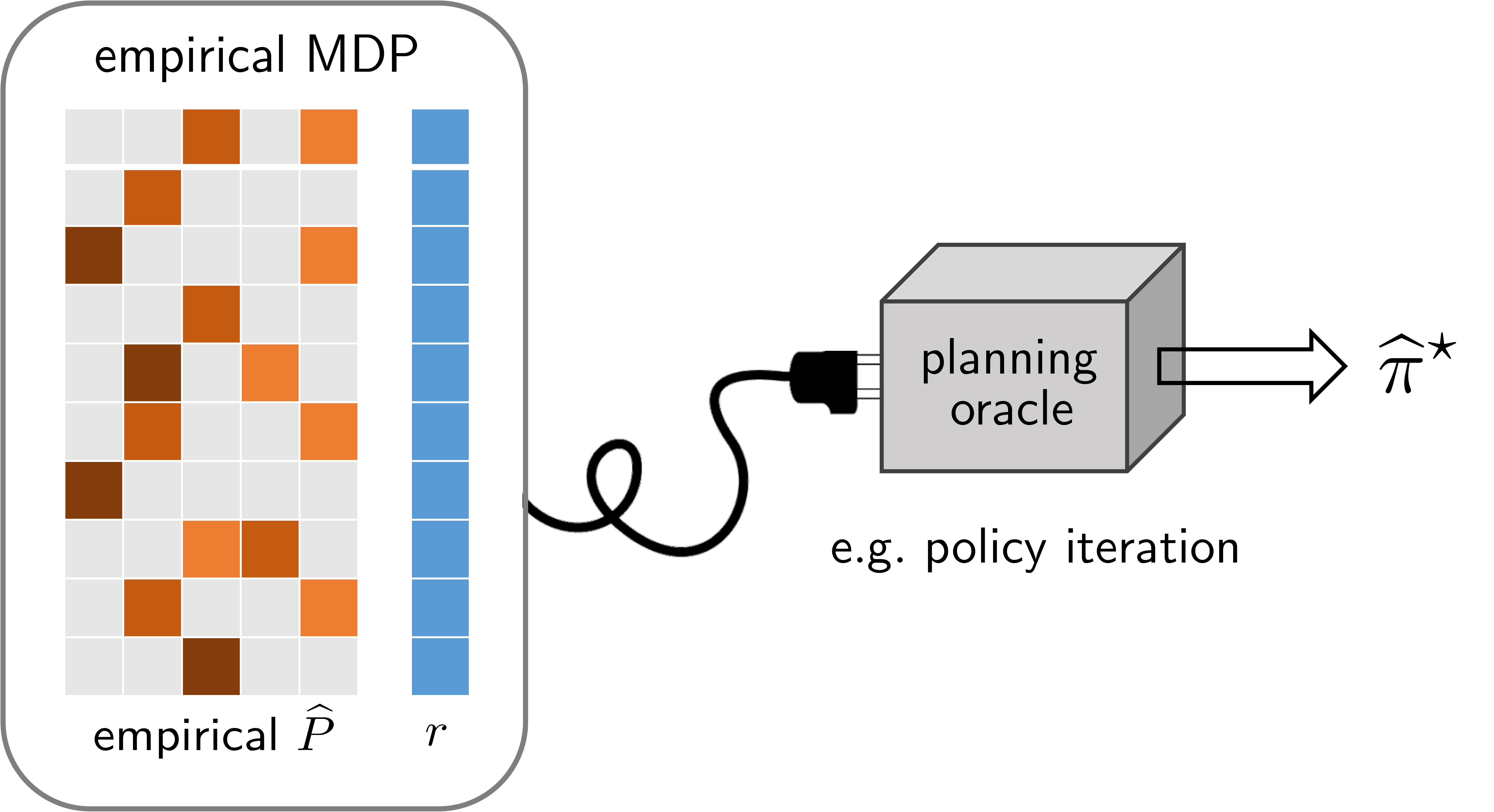} 
\end{center}	
	\caption{An illustration of the model-based policy learning algorithm, which first computes an empirical model and then plugs it into a planning algorithm. 
	\label{fig:model_based}}
\end{figure}

\paragraph{A model-based approach.} For concreteness, let us describe below the most natural version of the model-based policy learning algorithm. 
\begin{itemize}
	\item[1)] {\em Model estimation.} 
		For each state-action pair  $(s,a)\in \cS\times \cA$, draw $N$ independent samples from the generative model: 
		$$
			s^{(i)}(s,a) \overset{\mathrm{i.i.d.}}{\sim} P(\cdot \mymid s,a), \qquad i=1,\ldots, N,
		$$ 
		and construct an empirical transition kernel $\widehat{P}$ as follows
		\begin{equation}
			\forall s' \in \mathcal{S}, \qquad \widehat{P}(s^\prime \mymid s, a) = \frac{1}{N}\sum_{i=1}^N \ind \big\{  s^{\prime} = s^{(i)}(s,a) \big\}.
			\label{eq:defn-empirical-P}
		\end{equation}
		In words, $\widehat{P}(s^\prime \mymid s,a)$ records the empirical frequency of transitions from $(s,a)$ to state $s^\prime$.  
		This leads to an empirical MDP $\widehat{\mathcal{M}} = (\cS,\cA, \widehat{P}, r,\gamma)$ constructed from the data samples. 
		%We shall denote by $\widehat{V}^{\pi}$ the value function of policy $\pi$ in the empirical MDP $\widehat{\mathcal{M}}$.  

	\item[2)] {\em Planning.} Compute an optimal (deterministic) policy
\begin{align}
	\widehat{\pi}^{\star}\in \arg\max_\pi \widehat{{V}}^{\pi}.
	\label{defn:pi-p-star-noperturb}
\end{align}
 of the empirical MDP $\widehat{\mathcal{M}}$, using, e.g., policy iteration (see Section~\ref{sec:prelim-DP-inf}). Here and below, $\widehat{V}^{\pi}$ denotes the value function of policy $\pi$ in the empirical MDP $\widehat{\mathcal{M}}$. 
	
\end{itemize}
See Algorithm~\ref{alg:policy-learn-model-based} for the full procedure, and Figure~\ref{fig:model_based} for an illustration. Clearly, the total number of samples collected in the above algorithm is given by $N^{\mathsf{total}} \coloneqq NSA$.

\begin{algorithm}[t]
	%\begin{algorithmic}[1] 
    \DontPrintSemicolon
	\textbf{input:} discount factor $\gamma$, reward function $r$, sample size per state-action pair $N$. \\
	\For{$(s,a) \in \cS \times \cA$}{
        {\color{blue}\tcc{model estimation based on samples.}}
		draw $N$ independent samples $s^{(i)}(s,a) \overset{\mathrm{i.i.d.}}{\sim} P(\cdot \mymid s, a)$. \\
		compute $\widehat{P}(\cdot \mymid s,a)$ based on \eqref{eq:defn-empirical-P}.
	}
{\color{blue}\tcc{planning w.r.t.~the empirical model.}}
	\textbf{output}  $\widehat{\pi}^{\star}$
    %\coloneqq \arg\max_\pi \widehat{{V}}^{\pi}
    as an optimal (deterministic) policy of the MDP $\widehat{\mathcal{M}} = (\cS,\cA, \widehat{P}, r,\gamma)$.
    %, where $\widehat{{V}}^{\pi}$ represents the value function of policy $\pi$ in $\widehat{\mathcal{M}}$.  

   %\end{algorithmic} 
    \caption{Model-based policy learning with a simulator}
 \label{alg:policy-learn-model-based}
\end{algorithm}

\begin{comment}
\begin{algorithm}[t]
	\begin{algorithmic}[1] 
	\STATE\textbf{input:} discount factor $\gamma$, reward function $r$, sample size per state-action pair $N$. 
		\FOR{$(s,a) \in \cS \times \cA$}
		\STATE{ draw $N$ independent samples $s^{(i)}(s,a) \overset{\mathrm{i.i.d.}}{\sim} P(\cdot \mymid s, a)$. }
		\STATE{ compute $\widehat{P}(\cdot \mymid s,a)$ based on \eqref{eq:defn-empirical-P-policy-eval}.}
	\ENDFOR

	\STATE compute $\widehat{\pi}\coloneqq \arg\max_\pi \widehat{{V}}^{\pi}$ of the MDP $\widehat{\mathcal{M}} = (\cS,\cA, \widehat{P}, r,\gamma)$, 
		where $\widehat{{V}}^{\pi}$ represents the value function of policy $\pi$ in $\widehat{\mathcal{M}}$.  

   \end{algorithmic} 
    \caption{Model-based policy learning.}
 \label{alg:policy-learn-model-based}
\end{algorithm}
\end{comment}

\subsection{Theoretical guarantees}
We now state the sample complexity guarantee for the model-based method in Algorithm~\ref{alg:policy-learn-model-based}. Sharp analyses for model-based policy learning under the generative model that cover the full range of \(\varepsilon\) were first developed by \citet{li2020breaking,li2024breaking};  their analyses apply to a perturbed version of Algorithm~\ref{alg:policy-learn-model-based}, in which the reward function is randomly perturbed. More recently, \citet{zurek2024plug} showed that reward perturbation is unnecessary. The guarantee below is established through a refinement of the analysis in \citet{li2020breaking,li2024breaking}. 
\begin{theorem}[\citet{li2020breaking,li2024breaking,zurek2024plug}]
\label{Thm:sample-compl-main}
	There exists some universal constant $c_0>0$ such that: for any $0<\delta<1$ and any $0<\varepsilon \leq \frac{1}{1-\gamma}$, any optimal deterministic policy $\widehat{\pi}^{\star}$ returned by Algorithm~\ref{alg:policy-learn-model-based} achieves
    \begin{subequations}
	\begin{align}
		\forall s \in \mathcal{S} ,~~\,\, \qquad 
		V^{\pihatstar}(s)  &\geq  V^\star(s) - \varepsilon 
		  \label{eq:V-epsilon-pertubed} \\
		 \forall (s,a) \in \mathcal{S}\times \mathcal{A}, \qquad
		 Q^{\pihatstar}(s,a) &  \geq  Q^\star(s,a)  - \gamma \varepsilon  
         \label{eq:Q-epsilon-pertubed}
	\end{align}
    \end{subequations}
	with probability at least $1-\delta$, provided that 
	the sample size per state-action pair exceeds 
	\begin{align}
	\label{EqnSamples-main}
		N \geq \frac{c_{0}\log\big(\frac{SAN}{(1-\gamma)^2\varepsilon\delta}\big)}{(1-\gamma)^{3}\varepsilon^{2}}  . 
	\end{align}
\end{theorem}

%\begin{remark}
%	Notably, 
%\end{remark}

In a nutshell, the above theorem demonstrates that the model-based approach succeeds in finding an $\varepsilon$-optimal policy  
as soon as the total sample complexity exceeds the order of $\frac{SA}{(1-\gamma)^{3}\varepsilon^2}$ (modulo some log factor). This sample complexity is provably minimax optimal, as we will elaborate in Section~\ref{sec:minimax-lower-bound-generative}. Importantly, this result imposes no restriction on the range of $\varepsilon$ and, in particular, the accuracy level $\varepsilon$ is permitted to be as large as $\frac{1}{1-\gamma}$. 
As a consequence, this theorem is applicable even in regimes with small-to-moderate sample sizes, since its validity is guaranteed as long as 
\begin{align}
	N \geq \widetilde{\Omega}\Big(\frac{1}{1-\gamma}\Big) . 
	\label{eq:sample-size-range-planning}
\end{align}
Noteworthily, Condition~\eqref{eq:sample-size-range-planning} is necessary for learning a policy that performs strictly better than a random guess, and hence characterizes the full range of ``meaningful'' sample sizes. %The optimality of the (perturbed) model-based approach will be established in Section~\ref{sec:minimax-lower-bound-generative}. 

From a computational standpoint, it suffices to compute a deterministic policy $\widehat{\pi}$ satisfying $\|\widehat{V}^{\star}-\widehat{V}^{\widehat{\pi}}\|_{\infty}\leq \frac{1}{N(1-\gamma)}$: the same sample complexity guarantee continues to hold after increasing the universal constant $c_0$. Empirical VI and PI achieves this accuracy within $O\big(\frac{1}{1-\gamma}\log(\frac{N}{1-\gamma})\big)$ iterations, thereby yielding a computationally lightweight solution.

\subsection{Analysis: a first attempt}

We now proceed to establish performance guarantees for policy learning. Note that it suffices to analyze the suboptimality \eqref{eq:V-epsilon-pertubed} of $\pihatstar$ w.r.t.~value functions.  Clearly, if \eqref{eq:V-epsilon-pertubed} can be established, then the Bellman equation gives
\begin{align*}
Q^{\pihatstar}(s,a)-Q^{\star}(s,a) & =r(s,a)+\gamma\mathbb{E}_{s'\sim P(\cdot|s,a)}\big[V^{\pihatstar}(s')\big] \arxiv{\\}
\arxiv{&\quad} -\big(r(s,a)+\gamma\mathbb{E}_{s'\sim P(\cdot|s,a)}\big[V^{\star}(s')\big]\big)\\
 & =\gamma\mathbb{E}_{s'\sim P(\cdot|s,a)}\big[V^{\pihatstar}(s')-V^{\star}(s')\big]\geq-\gamma\varepsilon,
\end{align*}
as claimed in \eqref{eq:Q-epsilon-pertubed}.

In this subsection, we begin by examining the performance of the model-based policy learning algorithm in Algorithm~\ref{alg:policy-learn-model-based}. We first present a simple approach to establishing asymptotically optimal statistical guarantees (for large enough $N$) for Algorithm~\ref{alg:policy-learn-model-based}, albeit at the cost of a significant burn-in requirement. 
For simplicity, in this subsection we assume  (without providing a formal proof) that
\begin{align}
    \widehat{P} \approx P, 
    \label{eq:assumption-Phat-P-135}
\end{align}
which holds in the asymptotic regime as $N\rightarrow \infty$, but still leaves substantial room for improvement from a non-asymptotic perspective. In Section~\ref{sec:sharp-analysis-thm:sample-compl-MB-generative-inf}, we will relax this assumption (and the large-$N$ requirement) and extend the analysis all the way to the most sample-limited regime. 
Before proceeding, let us introduce some useful notation: 
\begin{itemize}
\item $\widehat{V}^{\pi}$ and $\widehat{Q}^{\pi}$: the value function and Q-function of policy $\pi$ in the empirical MDP  $\widehat{\mathcal{M}}$, respectively;

\item $\widehat{\pi}^{\star}$: an optimal {\em deterministic} policy of $\widehat{\mathcal{M}}$ (returned by Algorithm~\ref{alg:policy-learn-model-based}); 

\item $\widehat{V}^{\star} \coloneqq \widehat{V}^{\widehat{\pi}^\star}$ and $\widehat{Q}^{\star} \coloneqq \widehat{Q}^{\widehat{\pi}^\star}$: the optimal value function and optimal Q-function in $\widehat{\mathcal{M}}$.
\end{itemize}
%

%(with $\pihatstar$ the policy returned by Algorithm~\ref{alg:policy-learn-model-based}), 

To bound the suboptimality of $\pihatstar$ in terms of the value function, we start with the following elementary decomposition:
\begin{align}
	0\leq \Vstar - V^{\pihatstar} 
	&= V^{\pistar} - V^{\pihatstar} \notag\\
	&= \big( V^{\pistar} - \Vhat^{\pistar} \big) + \big( \Vhat^{\pistar}- \Vhat^{\pihatstar} \big) + \big( \Vhat^{\pihatstar} - V^{\pihatstar}\big) 
	 \nonumber\\
	&\leq  \big( V^{\pistar} - \Vhat^{\pistar}  \big) + \big(\Vhat^{\pihatstar} - V^{\pihatstar} \big) , 
	\label{eq:Vhat-Vstar-pistar-tot}
\end{align}
where the last line follows since $\pihatstar=\arg\max_{\pi}\widehat{V}^{\pi}$ and hence $\Vhat^{\pistar} \leq \Vhat^{\pihatstar}$. 
We are thus left with two terms to control (see \eqref{eq:Vhat-Vstar-pistar-tot}), both of which are concerned with certain policy evaluation errors (i.e., the policy evaluation errors w.r.t.~$\pistar$ and $\pihatstar$).

\paragraph{Step 1: controlling $V^{\pistar} - \Vhat^{\pistar}$.} 
Given that $\pi^{\star}$ is the optimal policy of the true MDP and hence independent of the data, 
we can readily accomplish this step by invoking the policy evaluation result presented in Theorem~\ref{thm:model-based-policy-eval}. Specifically, taking $\pi=\pi^{\star}$ in the analysis of Theorem~\ref{thm:model-based-policy-eval} (see \eqref{eqn:ice-cream}) shows that, with probability at least $1-\delta$, 
\begin{align}
	\big\|\widehat{V}^{\pi^{\star}}-V^{\pi^{\star}}\big\|_{\infty} & \leq c_1  \sqrt{\frac{\log\big(\frac{S}{\delta(1-\gamma)}\big)}{N(1-\gamma)^3}}
	\label{eq:optimal-Vpistar-bound}
\end{align}
as long as $N\geq \frac{c_2\log (\frac{S}{\delta(1-\gamma)}) }{1-\gamma}$, 
where $c_1,c_2>0$ are universal constants.

\paragraph{Step 2: controlling $\Vhat^{\pihatstar}-V^{\pihatstar}$.} 
Bounding the policy evaluation error $\Vhat^{\pihatstar}-V^{\pihatstar}$ is, however, substantially more complex than bounding $V^{\pistar} - \Vhat^{\pistar}$. The complexity stems from the intricate statistical dependency between $\widehat{\pi}^{\star}$ and the data samples, rendering Theorem~\ref{thm:model-based-policy-eval} inapplicable. 
We present here an argument valid in the asymptotic regime as $N\rightarrow \infty$. Since this approach is relatively loose in the non-asymptotic regime --- and will be sharpened in Section~\ref{sec:sharp-analysis-thm:sample-compl-MB-generative-inf} --- we omit the control of second-order (or higher-order) terms here in order to streamline the presentation and better highlight the main ideas.

Note that the Bellman equation \eqref{eq:Vpi-Bellman-inversion-matrix-inf} implies that
\begin{align}
\Vhat^{\pihatstar}-V^{\pihatstar} & =\big(I-\gamma\widehat{P}_{\pihatstar}\big)^{-1}r_{\pihatstar}-V^{\pihatstar}\nonumber \\
 & =\big(I-\gamma\widehat{P}_{\pihatstar}\big)^{-1}\left\{ \big(I-\gamma P_{\pihatstar}\big)V^{\pihatstar}-\big(I-\gamma\widehat{P}_{\pihatstar}\big)V^{\pihatstar}\right\} \nonumber \\
 & =\gamma\big(I-\gamma\widehat{P}_{\pihatstar}\big)^{-1}\big(\widehat{P}_{\pihatstar}-P_{\pihatstar}\big)V^{\pihatstar}\label{eq:Vhat-V-pihatstar-identity} \\
 & =\gamma\big(I-\gamma\widehat{P}_{\pihatstar}\big)^{-1}\big(\widehat{P}_{\pihatstar}-P_{\pihatstar}\big)V^{\pistar}  \arxiv{\notag\\}
 \arxiv{&\qquad} +\gamma\big(I-\gamma\widehat{P}_{\pihatstar}\big)^{-1}\big(\widehat{P}_{\pihatstar}-P_{\pihatstar}\big)(V^{\pihatstar}-V^{\pistar}) 
 \label{eq:Vhat-V-pihatstar-identity-246} \\
 & \approx \gamma\big(I-\gamma\widehat{P}_{\pihatstar}\big)^{-1}\big(\widehat{P}_{\pihatstar}-P_{\pihatstar}\big)V^{\pistar}, 
 \label{eq:Vhat-V-pihatstar-identity-2468}
\end{align}
where we remind the readers of the matrix notation $P_{\pi}$ and $r_{\pi}$ in \eqref{eqn:ppivq} and \eqref{eq:defn-rpi-inf}, respectively. 
Informally, the first term in \eqref{eq:Vhat-V-pihatstar-identity-246} is easier to cope than its counterpart in \eqref{eq:Vhat-V-pihatstar-identity}, since $V^{\pistar}$ is a fixed vector independent from $\widehat{P}$;  meanwhile, 
the last term in \eqref{eq:Vhat-V-pihatstar-identity-246} can be regarded as a second-order term (as it involves both $\widehat{P}-P$ and $V^{\pihatstar}-V^{\pistar}$), and is therefore expected to be asymptotically negligible as $N\rightarrow \infty$, a key observation that yields \eqref{eq:Vhat-V-pihatstar-identity-2468}.

To bound \eqref{eq:Vhat-V-pihatstar-identity-2468}, it is first seen from \eqref{eqn:ppivq} and the Bernstein inequality in Lemma~\ref{lem:Bernstein-prelim} that, with probability greater than $1-\delta$, 
\begin{align}
\arxiv{&} \big|\big(\widehat{P}_{\pihatstar}-P_{\pihatstar}\big)V^{\pistar}\big|  \book{&}=\Pi_{\pihatstar}\big|\big(\widehat{P}-P\big)V^{\pistar}\big| \notag\\
 & \arxiv{\qquad} \leq\Pi_{\pihatstar}\bigg(\sqrt{\frac{2\log\big(\frac{4SA}{\delta}\big)}{N}}\sqrt{\mathsf{Var}_{P}(V^{\pistar})}+\frac{2\|V^{\pistar}\|_{\infty}\log\big(\frac{4SA}{\delta}\big)}{3N}1_{SA}\bigg) \notag\\
 & \arxiv{\qquad} \leq\sqrt{\frac{2\log\big(\frac{4SA}{\delta}\big)}{N}}\sqrt{\mathsf{Var}_{P_{\pihatstar}}(V^{\pistar})}+\frac{2\log\big(\frac{4SA}{\delta}\big)}{3N(1-\gamma)}1_{S} \notag\\
 & \arxiv{\qquad} \approx \sqrt{\frac{2\log\big(\frac{4SA}{\delta}\big)}{N}}\sqrt{\mathsf{Var}_{P_{\pihatstar}}(V^{\pistar})}, 
 \label{eq:Phat-pistarhat-bernstein}
\end{align}
where the penultimate line holds since $\|V^{\pistar}\|_{\infty}\leq\frac{1}{1-\gamma}$, and in the last line we omit the higher-order term proportional to $1/N$ (as the other term scales with $1/\sqrt{N}$). In view of the non-negativity of $\big(I-\gamma\widehat{P}_{\pihatstar}\big)^{-1}$ (cf.~Lemma~\ref{lem:basic-property-matrix-notation-inf}), it follows from \eqref{eq:Phat-pistarhat-bernstein} that
\begin{align}
 \arxiv{&} \gamma \big|\big(I-\gamma\widehat{P}_{\pihatstar}\big)^{-1}\big(\widehat{P}_{\pihatstar}-P_{\pihatstar}\big)V^{\pistar}\big|\book{&}\leq\gamma\big(I-\gamma\widehat{P}_{\pihatstar}\big)^{-1}\big|\big(\widehat{P}_{\pihatstar}-P_{\pihatstar}\big)V^{\pistar}\big| \notag\\
 & \approx \gamma\big(I-\gamma P_{\pihatstar}\big)^{-1}\big|\big(\widehat{P}_{\pihatstar}-P_{\pihatstar}\big)V^{\pistar}\big| \notag\\
 & \lessapprox \gamma\sqrt{\frac{2\log\big(\frac{4SA}{\delta}\big)}{N}}\big(I-\gamma P_{\pihatstar}\big)^{-1}\sqrt{\mathsf{Var}_{P_{\pihatstar}}(V^{\pistar})} \notag\\
  & \approx \gamma\sqrt{\frac{2\log\big(\frac{4SA}{\delta}\big)}{N}}\big(I-\gamma P_{\pihatstar}\big)^{-1}\sqrt{\mathsf{Var}_{P_{\pihatstar}}(V^{\pihatstar})}  \label{eq:V-pihat-policy-eval-error-1357}\\
  & \leq 
  \sqrt{\frac{4\log\big(\frac{4SA}{\delta}\big)}{(1-\gamma)N}}\|V^{\pihatstar}\|_{\infty}1_S 
  \leq\sqrt{\frac{4\log\big(\frac{4SA}{\delta}\big)}{(1-\gamma)^{3}N}} 1_S
  \label{eq:V-pihat-policy-eval-error-135}
\end{align}
with probability exceeding $1-\delta$, 
where the approximation in the second and the fourth lines relies on the assumption~\eqref{eq:assumption-Phat-P-135} (and hence $V^{\pihatstar}\approx V^{\pistar}$), and the last line invokes \eqref{eq:lemma-VarP-V-bound} and the basic property $\|V^{\pihatstar}\|_{\infty}\leq \frac{1}{1-\gamma}$. We remark that the approximation in \eqref{eq:V-pihat-policy-eval-error-1357} serves to align the policy associated with the value function and the one associated with the transition kernel, thereby enabling the application of~\eqref{eq:lemma-VarP-V-bound}. 
It is worth emphasizing that the use of property~\eqref{eq:lemma-VarP-V-bound} here is crucial for obtaining the asymptotically sharp bound, as it carefully leverages the underlying variance structure to sharpen the horizon dependency. 

\paragraph{Step 3: putting all this together.}  
To finish up, substituting \eqref{eq:optimal-Vpistar-bound}, \eqref{eq:Vhat-V-pihatstar-identity-2468} and \eqref{eq:V-pihat-policy-eval-error-135} into \eqref{eq:Vhat-Vstar-pistar-tot}, we can conclude that 
\[
\|\Vstar - V^{\pihatstar} \|_{\infty} \lesssim 
\sqrt{\frac{\log\big(\frac{SA}{\delta(1-\gamma)}\big)}{(1-\gamma)^{3}N}}
\]
holds with high probability when $N\rightarrow \infty$.

Nevertheless, while the arguments above --- similar in spirit to those used in \citet{azar2013minimax} --- 
suffice to establish the desirable statistical bound for the model-based algorithm asymptotically, they rely on an excessively large sample size requirement on $N$; for instance, working out the full details may require $N$ to be at least polynomial in $S$.  To overcome this limitation, a substantially more refined analysis strategy is needed, which we introduce next.

\subsection{Sharp analysis for Theorem~\ref{Thm:sample-compl-main}: leave-one-out analysis}
\label{sec:sharp-analysis-thm:sample-compl-MB-generative-inf}

We now present a refined analysis for establishing sharp statistical bounds for $\big\|\Vstar-V^{\pihatstar}\big\|_{\infty}$.
In light of the decomposition \eqref{eq:Vhat-Vstar-pistar-tot} and the bound \eqref{eq:optimal-Vpistar-bound}, it suffices to derive a tight bound on $\|\Vhat^{\pihatstar} - V^{\pihatstar}\|_{\infty}$ that remains valid even when $N$ is small or moderately sized.

\paragraph{Leave-one-out analysis: key ideas.}
As alluded to previously, the most challenging part for bounding $\|\Vhat^{\pihatstar} - V^{\pihatstar}\|_{\infty}$ lies in coping with the intricate statistical dependency between ${\pihatstar}$ and $\widehat{P}$. To better illustrate this issue, we use the following alternative form of
the value-estimation identity in \eqref{eq:Vhat-V-pihatstar-identity}:
\begin{align}
\widehat{V}^{\pihatstar}-V^{\pihatstar}
&=
\gamma\bigl(I-\gamma P_{\pihatstar}\bigr)^{-1}
\bigl(\widehat{P}_{\pihatstar}-P_{\pihatstar}\bigr)
\widehat{V}^{\star}.
\label{eq:Vhat-V-pihatstar-identity-temp}
\end{align}
\begin{proof}[Proof of~\eqref{eq:Vhat-V-pihatstar-identity-temp}]
We begin with the Bellman equations:
\[
\widehat{V}^{\pihatstar}
=
r_{\pihatstar}
+\gamma\widehat{P}_{\pihatstar}\widehat{V}^{\pihatstar},
\qquad
V^{\pihatstar}
=
r_{\pihatstar}
+\gamma P_{\pihatstar}V^{\pihatstar}.
\]
Subtracting the two Bellman equations gives
\begin{align*}
\widehat{V}^{\pihatstar}-V^{\pihatstar}
&=
\gamma\widehat{P}_{\pihatstar}\widehat{V}^{\pihatstar}
-\gamma P_{\pihatstar}V^{\pihatstar}\arxiv{\\
&}=
\gamma\bigl(\widehat{P}_{\pihatstar}-P_{\pihatstar}\bigr)
\widehat{V}^{\pihatstar}
+
\gamma P_{\pihatstar}
\bigl(\widehat{V}^{\pihatstar}-V^{\pihatstar}\bigr).
\end{align*}
Rearranging terms, we are left with
\[
\bigl(I-\gamma P_{\pihatstar}\bigr)
\bigl(\widehat{V}^{\pihatstar}-V^{\pihatstar}\bigr)
=
\gamma\bigl(\widehat{P}_{\pihatstar}-P_{\pihatstar}\bigr)
\widehat{V}^{\pihatstar}.
\]
This together with $\widehat{V}^{\pihatstar}=\widehat{V}^{\star}$ concludes the proof. 
\end{proof}
\begin{remark}
The decomposition~\eqref{eq:Vhat-V-pihatstar-identity-temp} differs from the one used in~\citet{li2024breaking}, namely,~\eqref{eq:Vhat-V-pihatstar-identity}. The alternative form adopted here is particularly well suited to the leave-one-out argument developed below and, together with a self-bounding variance analysis, avoids the separation and reward-perturbation arguments used in~\citet{li2024breaking}. 
\end{remark}

Notably, given that both $\pihatstar$ and $\widehat{V}^{\star}$ depend on the empirical
transition kernel $\widehat{P}$, one cannot directly invoke the Bernstein inequality (or other standard concentration inequalities) to control the relevant quantity
\[
\bigl(\widehat{P}_{\pihatstar}-P_{\pihatstar}\bigr)\widehat{V}^{\star}.
\]

To remedy this issue, we resort to a powerful decoupling technique known as the ``{\em leave-one-out analysis}''. Originally developed in the high-dimensional probability and statistics literature (e.g., \citet{el2018impact,ma2020implicit,chen2021spectral}), the leave-one-out analysis was first introduced to RL theory by \citet{agarwal2019optimality}, and later on refined by \citet{li2020breaking,li2024breaking} to sharpen the sample complexity analysis. At a high level, this approach consists of the following key ingredients:
\begin{itemize}
    \item {\em Construction of leave-one-out auxiliary MDPs.} For each state $s$, construct a finite collection of auxiliary MDPs whose transition kernels coincide with the original empirical kernel
    $\widehat{P}$, except that all transition rows associated with state $s$ are replaced by absorbing transitions. The reward at state $s$ is
    adjusted through a scalar parameter $u$. We refer to these MDPs as ``leave-one-out MDPs'', as each of them effectively drops one block of data samples  associated with state $s$ when estimating the transition kernel. 

    \item {\em Stability of optimal value functions.} Since leave-one-out auxiliary MDPs represent a slight perturbation of the original empirical MDP, at least one of the resulting optimal value functions is often expected to be exceedingly close to the optimal value function of the original empirical MDP. This reflects a form of ``stability'' in leave-one-out perturbations, a property typically required in leave-one-out decoupling arguments. 

    \item {\em Harnessing local statistical independence in leave-one-out MDPs.} Since a leave-one-out MDP is built without samples collected for, say, state $s$,  it is independent of the empirical transition vector $\widehat{P}_{s,a}$ for every $a\in \cA$. As a result, the $(s,a)$-th component of the term $\big(\widehat{P}-P\big)\Vhatstar $ can be bounded using the following observation
    \begin{align}
    \label{eq:LOO-intuition-generative}
    \big(\widehat{P}_{s,a}-P_{s,a}\big)\Vhatstar
    \approx \big(\widehat{P}_{s,a}-P_{s,a}\big)\Vhatstar_{\mathsf{loo}} ,
    \end{align}
  Such independence enables the application of standard concentration inequalities to control \eqref{eq:LOO-intuition-generative}, which is crucial in analyzing \eqref{eq:Vhat-V-pihatstar-identity-temp} despite complicated statistical dependency.  
\end{itemize}

In what follows, we present a proof that formalizes the above high-level ideas.

\paragraph{Preliminaries: absorbing MDPs with identical optimal value functions.}
Before diving into the main analysis, we first explain how, given any MDP $\mathcal{M}=(\cS,\cA,P,r,\gamma)$, one can construct an absorbing MDP while preserving the same optimal value function.

Concretely, consider any state $s$ and any scalar $u\in[0, 1/(1-\gamma)]$. We construct an auxiliary MDP with the same state space $\cS$, action space $\cA$ and discount factor $\gamma$ --- denoted by $\MDP_{s,u}$ --- that resembles the original MDP $\MDP$ except that state $s$ becomes absorbing under every action. More precisely, its transition kernel is taken to be
\begin{align}
\label{eqn:transition-kernel-M-su-inf}
\begin{array}{rll}
P_{\MDP_{s,u}}(s\mid s,a)&=1, &\text{for all }a\in\cA,\\
P_{\MDP_{s,u}}(s'\mid s,a)&=0, &\text{for all }s'\neq s,\ a\in\cA,\\
P_{\MDP_{s,u}}(\cdot\mid s',a)&=P(\cdot\mid s',a), &\text{for all }s'\neq s,\ a\in\cA,
\end{array}
\end{align}
where the first two lines in \eqref{eqn:transition-kernel-M-su-inf} ensure that state $s$ is absorbing under every action $a$. Meanwhile, the reward function of $\MDP_{s,u}$ --- denoted by $r_{\MDP_{s,u}}$ --- is given by
\begin{align}
\label{eqn:reward-M-su-inf}
\begin{array}{rll}
r_{\MDP_{s,u}}(s,a)&=(1-\gamma)u, &\text{for all }a\in\cA,\\
r_{\MDP_{s,u}}(s',a)&=r(s',a), &\text{for all }s'\neq s,\ a\in\cA,
\end{array}
\end{align}
so that the immediate rewards coincide with those of $\mathcal{M}$ everywhere except at state $s$.  
For notational convenience, let $\Vstar_{s,u}$ represent the optimal value function in $\MDP_{s,u}$. Since the immediate rewards remain in $[0,1]$, one has $$0\leq\Vstar_{s,u}\leq\frac{1}{1-\gamma}1_S.$$ 
By properly choosing $u$, we can ensure that $\MDP_{s,u}$ and $\MDP$ share the same optimal value function, as formalized in the following lemma.
\begin{lemma}
\label{LemAbsToNormalV}
Setting $u^{\star}=V^{\star}(s)$, one has
\begin{align}
\label{eqn:relation}
\Vstar_{s,u^{\star}}=\Vstar.
\end{align}
\end{lemma}
The proof of Lemma~\ref{LemAbsToNormalV} consists of comparing the Bellman optimality equations of $\MDP$ and $\MDP_{s,u^{\star}}$, and is deferred to Section~\ref{sec:proof-LemAbsToNormalV}. In summary, for each state $s$, we can form an $s$-absorbing MDP $\MDP_{s,u^{\star}}$ that shares the same optimal value function as the original MDP $\MDP$. Moreover, it decouples the randomness arising from state $s$, as we will soon demonstrate.

\paragraph{Step 1: constructing leave-one-out auxiliary MDPs.}
As alluded to previously, the leave-one-out analysis employed here requires constructing a collection of leave-one-out auxiliary MDPs. Concretely, for each state $s$ and each scalar $u\in \big[0,\frac{1}{1-\gamma}\big]$, we can apply the method described above (see \eqref{eqn:transition-kernel-M-su-inf} and \eqref{eqn:reward-M-su-inf}) to construct an $s$-absorbing MDP --- denoted by $\widehat{\MDP}_{s,u}$ --- 
associated with the empirical MDP $\widehat{\MDP}$. In other words, the transition kernel of $\widehat{\MDP}_{s,u}$ coincides with $\widehat{P}$ except that all rows associated with state $s$ are made absorbing, while the rewards at state $s$ are replaced by $(1-\gamma)u$.

Recall that each row $\widehat{P}_{s,a}$ is derived from samples associated with $(s,a)$, whereas in $\widehat{\MDP}_{s,u}$, this row is replaced by the $s$-th standard basis vector. Consequently, for any fixed $u$, the MDP $\widehat{\MDP}_{s,u}$ and its optimal value function $\Vhatstar_{s,u}$ are independent of $\widehat{P}_{s,a}$ for every $a\in\cA$. Moreover, it follows from Lemma~\ref{LemAbsToNormalV} that $\Vhatstar_{s,\Vhatstar(s)}=\Vhatstar$.

\paragraph{Step 2: linking $\Vhatstar$ with a small set of value functions independent of $\widehat{P}_{s,a}$.}
With the above construction in place, one might be tempted to use $\widehat{\MDP}_{s,\Vhatstar(s)}$ as a surrogate for $\widehat{\MDP}$, since the two share identical optimal value functions. However, a subtle issue arises: $\widehat{\mathcal{M}}_{s,\Vhatstar(s)}$ is no longer statistically independent of $\widehat{P}_{s,a}$, because the choice of $\widehat{u}^{\star}=\Vhatstar(s)$ explicitly depends on $\widehat{P}_{s,a}$. Unfortunately, this dependence undermines the key advantage of the leave-one-out construction described earlier.

Fortunately, this statistical dependency is encapsulated in a single scalar parameter $\widehat{u}^{\star}$. Inspired by techniques from high-dimensional probability \citep{vershynin2018high}, a plausible strategy is to invoke a standard covering argument: construct a finite, yet sufficiently dense, cover of the feasible range of $\widehat{u}^{\star}$ (i.e., $\big[0,\frac{1}{1-\gamma}\big]$), and then approximate $\widehat{\mathcal{M}}_{s,\widehat{u}^{\star}}$
using a group of auxiliary MDPs $\{\widehat{\mathcal{M}}_{s,u}\}$ with $u$ selected from this cover. 

To formalize this idea, form an epsilon-net of $\big[0,\frac{1}{1-\gamma}\big]$ with precision $\epsilon=\frac{1}{N(1-\gamma)}$:
\begin{align}
\label{eqn:net-eps-LOO-inf}
\net{\epsilon}\defn\{0,\epsilon,2\epsilon,\ldots,N\epsilon\},
\qquad \epsilon=\frac{1}{N(1-\gamma)}.
\end{align}
This epsilon-net has cardinality $N+1$ and is independent of the samples. For each $u\in \mathcal{N}_{\epsilon}$, we construct an $s$-absorbing MDP $\widehat{\MDP}_{s,u}$ as described previously. Since $\mathcal{N}_{\epsilon}$ is statistically independent of $\widehat{P}_{s,a}$ for any fixed $\epsilon$,  each $\widehat{\MDP}_{s,u}$ with $u\in \mathcal{N}_{\epsilon}$ remains statistically independent of $\widehat{P}_{s,a}$. Notably, one can find a point $u\in \mathcal{N}_{\epsilon}$ such that the optimal value function of its associated absorbing MDP is $\epsilon$-close to that of the original empirical MDP, as asserted by the lemma below. 
\begin{lemma}
\label{LemValueCover}
For each state $s\in\cS$, there exists a point $u_0\in\net{\epsilon}$ such that
\begin{align}
\label{eq:loo-value-cover}
\big\|\Vhatstar-\Vhatstar_{s,u_0}\big\|_{\infty}\leq\epsilon.
\end{align}
\end{lemma}
The proof of this lemma is postponed to Section~\ref{sec:proof-LemValueCover}. Importantly, this lemma unveils that, analyzing $\Vhatstar$ reduces to studying a small collection of leave-one-out MDPs that are statistically independent of $\widehat{P}_{s,a}$. The entire statistical dependence of $\Vhatstar$ on $\widehat{P}_{s,a}$ is thus encapsulated in the choice of an auxiliary MDP from this small collection. This insight paves the way for effective decoupling of statistical dependencies.  

\paragraph{Step 3: establishing optimal value estimation accuracy.} Armed with these leave-one-out auxiliary MDPs, we are ready to derive the desired statistical bound. More concretely, we intend to establish the following lemma.
\begin{lemma}
\label{lem:policy-learning-evaluation}
Given $0<\delta<1$, let
\begin{align}
\label{eq:beta-policy-learning}
\beta_1=2\log\bigg(\frac{16SAN}{(1-\gamma)\delta}\bigg).
\end{align}
With probability at least $1-\delta$, it holds that
\begin{subequations}
\label{eqn:V-diff-fabulous-all-inf-generative}
\begin{align}
\big\|\widehat{V}^{\pihatstar}-V^{\pihatstar}\big\|_{\infty}
&\leq c_1\sqrt{\frac{2\beta_1}{N(1-\gamma)^3}},
\label{eqn:V-diff-fabulous-0-inf-generative}\\
0\leq\Vstar-V^{\pihatstar}
&\leq 2c_1\sqrt{\frac{\beta_1}{N(1-\gamma)^3}}\,1_S,
\label{eqn:V-diff-fabulous-1-inf-generative}
\end{align}
\end{subequations}
provided that $N\geq\frac{c_0\beta_1}{1-\gamma}$, where $c_0,c_1>0$ are sufficiently large universal constants.  
\end{lemma}

\begin{proof}[Proof of Lemma~\ref{lem:policy-learning-evaluation}]
Once the first bound~\eqref{eqn:V-diff-fabulous-0-inf-generative} is proven, then the second bound follows by combining it with \eqref{eq:optimal-Vpistar-bound} and the decomposition \eqref{eq:Vhat-Vstar-pistar-tot}, where we apply \eqref{eq:optimal-Vpistar-bound} with failure probability $\delta/2$ and note that $\beta_1\geq\log\big(\frac{2S}{\delta(1-\gamma)}\big)$. Thus, it suffices to establish that the first bound~\eqref{eqn:V-diff-fabulous-0-inf-generative} holds with probability at least $1-\delta/2$.

To begin, for each state-action pair $(s,a)$, the set $\{\Vhatstar_{s,u}\}_{u\in\net{\epsilon}}$ is independent of $\widehat{P}_{s,a}$. The Bernstein inequality (see Lemma~\ref{lem:Bernstein-prelim}) taken together with a union bound implies that, with probability at least $1-\delta/2$,
\begin{align}
\label{eqn:bernstein-union}
\big|(\widehat{P}_{s,a}-P_{s,a})\Vhatstar_{s,u}\big|
\leq\sqrt{\frac{\beta_1}{N}}\sqrt{\mathsf{Var}_{P_{s,a}}\big[\Vhatstar_{s,u}\big]}
+\frac{\beta_1}{3N(1-\gamma)}
\end{align}
holds simultaneously for all $(s,a)\in\cS\times\cA$ and all $u\in\net{\epsilon}$. Here, we have used $\|\Vhatstar_{s,u}\|_{\infty}\leq\frac{1}{1-\gamma}$ and $|\net{\epsilon}|=N+1$.

For each $s$, Lemma~\ref{LemValueCover} guarantees that there exists a $u_0\in\net{\epsilon}$ such that $\big\|\Vhatstar-\Vhatstar_{s,u_0}\big\|_{\infty}\leq\epsilon$. 
Invoking this important fact and the triangle inequality, we obtain
\begin{align}
\big|(\widehat{P}_{s,a}-P_{s,a})\Vhatstar\big|
&\leq \big|(\widehat{P}_{s,a}-P_{s,a})\Vhatstar_{s,u_0} \big| + \big|(\widehat{P}_{s,a}-P_{s,a})(\Vhatstar- \Vhatstar_{s,u_0} )\big| \notag \\
& \leq  \big|(\widehat{P}_{s,a}-P_{s,a})\Vhatstar_{s,u_0} \big| + 2\epsilon \notag \\
&\leq \sqrt{\frac{\beta_1}{N}}\sqrt{\mathsf{Var}_{P_{s,a}}\big[\Vhatstar_{s,u_0}\big]}
+\frac{\beta_1}{3N(1-\gamma)} + 2\epsilon \notag \\
&\leq \sqrt{\frac{\beta_1}{N}}\sqrt{\mathsf{Var}_{P_{s,a}}\big[\Vhatstar\big]}
+\bigg(2+\sqrt{\frac{\beta_1}{N}}\bigg)\epsilon
+\frac{\beta_1}{3N(1-\gamma)}\notag\\
&\leq\sqrt{\frac{\beta_1}{N}}\sqrt{\mathsf{Var}_{P_{s,a}}\big[\Vhatstar\big]}
+\frac{2\beta_1}{N(1-\gamma)}. 
\label{eq:Phat-sa-Vhat-star-concentration-proof-mb-tight}
\end{align}
Here, the second line follows from $\big|(\widehat{P}_{s,a}-P_{s,a})(\Vhatstar-\Vhatstar_{s,u_0})\big|\leq 2\big\|\Vhatstar-\Vhatstar_{s,u_0}\big\|_{\infty}\leq2\epsilon$, the third line follows from \eqref{eqn:bernstein-union},  the penultimate line follows from the triangle inequality
 \begin{align*}
\sqrt{\mathsf{Var}_{P_{s,a}}\big[\Vhatstar_{s,u_0}\big]}
&\leq\sqrt{\mathsf{Var}_{P_{s,a}}\big[\Vhatstar\big]}
+\sqrt{\mathsf{Var}_{P_{s,a}}\big[\Vhatstar_{s,u_0}-\Vhatstar\big]} \arxiv{\\
&}\leq\sqrt{\mathsf{Var}_{P_{s,a}}\big[\Vhatstar\big]}+\epsilon,
\end{align*}
while the last step uses $\epsilon=\frac{1}{N(1-\gamma)}$, $\beta_1\geq4$, and $N\geq1$. Since this result~\eqref{eq:Phat-sa-Vhat-star-concentration-proof-mb-tight} holds for every action, we may select $a=\pihatstar(s)$ for every $(s,a)\in \cS\times \cA$ to demonstrates that 
\begin{align} \label{eq:Bernstein-empirical-optimal-value}
	\Big| \big(\Phat_{\pihatstar}-\bP_{\pihatstar}\big)\Vhatstar \Big| &\leq \sqrt{\frac{\beta_1}{N}}\sqrt{\mathsf{Var}_{P_{\pihatstar}}\big[\Vhatstar\big]}
+\frac{2\beta_1}{N(1-\gamma)} \one_S
\end{align}
with probability exceeding $1 - \delta/2$.

It remains to convert \eqref{eq:Bernstein-empirical-optimal-value} into the advertised value estimation bound. Akin to Section~\ref{sec:analysis-mb-pe-2nd-attempt}, the key is to exploit the interaction between the inverse matrix $\big(I-\gamma P_{\pihatstar}\big)^{-1}$ and the variance term $\sqrt{\mathsf{Var}_{P_{s,a}}\big[\Vhatstar\big]}$. To begin, recall that $(I-\gamma P_{\pihatstar})^{-1}$ is nonnegative and has each row summing to $\frac{1}{1-\gamma}$ (see Lemma~\ref{lem:basic-property-matrix-notation-inf}). These properties, taken together with \eqref{eq:Vhat-V-pihatstar-identity-temp} and \eqref{eq:Bernstein-empirical-optimal-value}, yield
\begin{align}
\arxiv{&} \big\|\Vhat^{\pihatstar}-V^{\pihatstar}\big\|_{\infty}\arxiv{\notag\\}
&\arxiv{\quad}\leq\gamma\Big\|(I-\gamma P_{\pihatstar})^{-1}
\big|(\widehat{P}_{\pihatstar}-P_{\pihatstar})\Vhatstar\big|\Big\|_{\infty}\notag\\
&\arxiv{\quad}\leq\gamma\sqrt{\frac{\beta_1}{N}}\Big\|(I-\gamma P_{\pihatstar})^{-1}
\sqrt{\mathsf{Var}_{P_{\pihatstar}}\big[\Vhatstar\big]}\Big\|_{\infty}
+\frac{2\beta_1}{N(1-\gamma)^2}.
\label{eq:empirical-value-residual-bound}
\end{align}
We are thus left with the task of controlling the variance-dependent term in~\eqref{eq:empirical-value-residual-bound}. Towards this end, we develop the following key lemma, derived through a self-bounding argument; the proof can be found in Section~\ref{sec:proof:lem:empirical-value-residual}. 
\begin{lemma}
\label{lem:empirical-value-residual}
On the event where~\eqref{eq:Bernstein-empirical-optimal-value} holds, one has
\begin{align}
\gamma\Big\|(I-\gamma P_{\pihatstar})^{-1}
\big|(\widehat{P}_{\pihatstar}-P_{\pihatstar})\Vhatstar\big|
\Big\|_{\infty}
\leq
2\sqrt{\frac{2\beta_1}{N(1-\gamma)^3}}
+
\frac{6\beta_1}{N(1-\gamma)^2}.
\label{eq:empirical-value-residual-bound-final}
\end{align}
\end{lemma}
\begin{remark}
Lemma~\ref{lem:empirical-value-residual} is another key ingredient in
the present analysis that helps bypass the separation and
reward-perturbation arguments used in~\citet{li2024breaking}. It is
closely related to the earlier bound~\eqref{eq:lemma-VarP-V-bound}, but
the latter applies to the true value function rather than the
data-dependent empirical value function. 
\end{remark}
Combining Lemma~\ref{lem:empirical-value-residual} with inequality~\eqref{eq:empirical-value-residual-bound} readily establishes \eqref{eqn:V-diff-fabulous-0-inf-generative}, thus completing the proof of Lemma~\ref{lem:policy-learning-evaluation}.
\end{proof}

\paragraph{Step 4: putting everything together.}
Lemma~\ref{lem:policy-learning-evaluation} implies the desired result as soon as
\begin{align*}
N\geq\frac{c_0\beta_1}{(1-\gamma)^3\varepsilon^2}
\end{align*}
for a sufficiently large universal constant $c_0>0$. Since
$\varepsilon\leq \frac{1}{1-\gamma}$, this condition also guarantees
$N\geq c_0\beta_1/(1-\gamma)$. This completes the proof.

\section{Minimax lower bounds}
\label{sec:minimax-lower-bound-generative}
 
To confirm the optimality of model-based algorithms, we present a matching minimax lower bound under the generative model, originally established in the seminal work of \cite{azar2013minimax}. This result demonstrates that the performance guarantees achieved by the model-based approach are, in fact, statistically optimal (up to logarithmic factors).

\begin{theorem}
\label{theorem:azar}
Consider any $0<\varepsilon<\frac{1}{32(1-\gamma)}$ and $\gamma \in [2/3,1)$, and suppose $SA\geq 16$. Assume that $N$ independent samples are drawn for each state-action pair from the generative model, where 
$$ N \leq  \frac{c_N\log (SA)}{(1-\gamma)^3 \varepsilon^2} $$
for some small universal constant $c_N>0$. 
\begin{itemize}
\item[(i)] 
For any (deterministic) policy estimator $\widehat{\pi}$, there is a $\gamma$-discounted infinite-horizon MDP $\mathcal{M}$ with $S$ states and $A$ actions such that 
$$
\mathbb{P}_{\mathcal{M}}\big( \|Q^{\widehat{\pi}} - Q^{\star}\|_\infty > \varepsilon \big) \geq 1/4.$$  

\item[(ii)] 
For any Q-function estimator $\widehat{Q}$, there exists a $\gamma$-discounted infinite-horizon MDP $\mathcal{M}$ with $S$ states and $A$ actions such that 
$$
\mathbb{P}_{\mathcal{M}}\big( \|\widehat{Q} - Q^{\star}\|_\infty > \varepsilon \big) \geq 1/4.$$ 
\end{itemize}
Here, $\mathbb{P}_{\mathcal{M}}(\cdot)$ denotes the probability distribution under $\mathcal{M}$. 
\end{theorem}
%
%\begin{remark}

%\end{remark} 

As asserted by Theorem~\ref{theorem:azar},  no algorithm can reliably learn an $\varepsilon$-optimal policy, or an $\varepsilon$-accurate optimal Q-function, for every $\gamma$-discounted infinite-horizon MDP, unless the total sample size exceeds $\widetilde{\Omega}\big(\frac{SA}{(1-\gamma)^3 \varepsilon^2}\big)$. 
By comparing this minimax lower bound with the performance guarantees of Algorithm~\ref{alg:policy-eval-model-based} (see Theorem~\ref{thm:model-based-policy-eval},  corresponding the policy evaluation setting with $A=1$), as well as Algorithm~\ref{alg:policy-learn-model-based} (see Theorem~\ref{Thm:sample-compl-main}), we conclude that the model-based approach is information-theoretically optimal (up to logarithmic factors) in sample complexity for both policy evaluation and policy learning, across the entire range of precision levels $\varepsilon$. 

Finally, we remark that \citet{azar2013minimax} established the first tight minimax lower bound on the sample complexity, applicable to both generative model and online sampling settings. Although the statement of \citet[Theorem~3]{azar2013minimax} focuses on estimating optimal Q-functions and restricts the precision level to $\varepsilon \leq O(1)$, the lower bound developed therein actually extends to policy learning and holds for the broader range $\varepsilon\leq O\big(\frac{1}{1-\gamma}\big)$. The original proof in \citet{azar2013minimax} relies on a change-of-measure argument by analyzing the likelihood ratio between two carefully constructed MDPs. In contrast, the proof to be presented shortly in this monograph adopts an information-theoretic approach based on Fano's inequality (see Lemma~\ref{lem:fano}).

%We remark that the work \cite{azar2013minimax} established a minimax lower bound on the order of $\widetilde{\Omega}\Big( \frac{SA}{(1-\gamma)^3\varepsilon^2} \Big)$, which actually holds both for policy evaluation (for a fixed policy $\pi$), finding the optimal Q-function, and finding the optimal policy.

%In view of the minimax lower bound in \cite{azar2013minimax} --- which we shall elaborate on in Theorem~\ref{theorem:azar} --- it corroborates the {\em minimax optimality} of the perturbed model-based approach for the full $\varepsilon$-range (which is previously unavailable), and demonstrates the information-theoretic infeasibility to learn a policy strictly better than a random guess if  $N\leq \widetilde{O}\big( \frac{1}{1-\gamma} \big)$. Put another way, the condition \eqref{eq:sample-size-range-planning} contains the full range of  ``meaningful'' sample sizes.  

\subsection{Analysis: lower bound for a simpler small-scale case}
\label{sec:analysis-lower-bound-Q-generative}

\begin{figure}
\begin{center} 

    \begin{subfigure}{0.4\textwidth}
        \centering
        \includegraphics[height=0.39\textwidth]{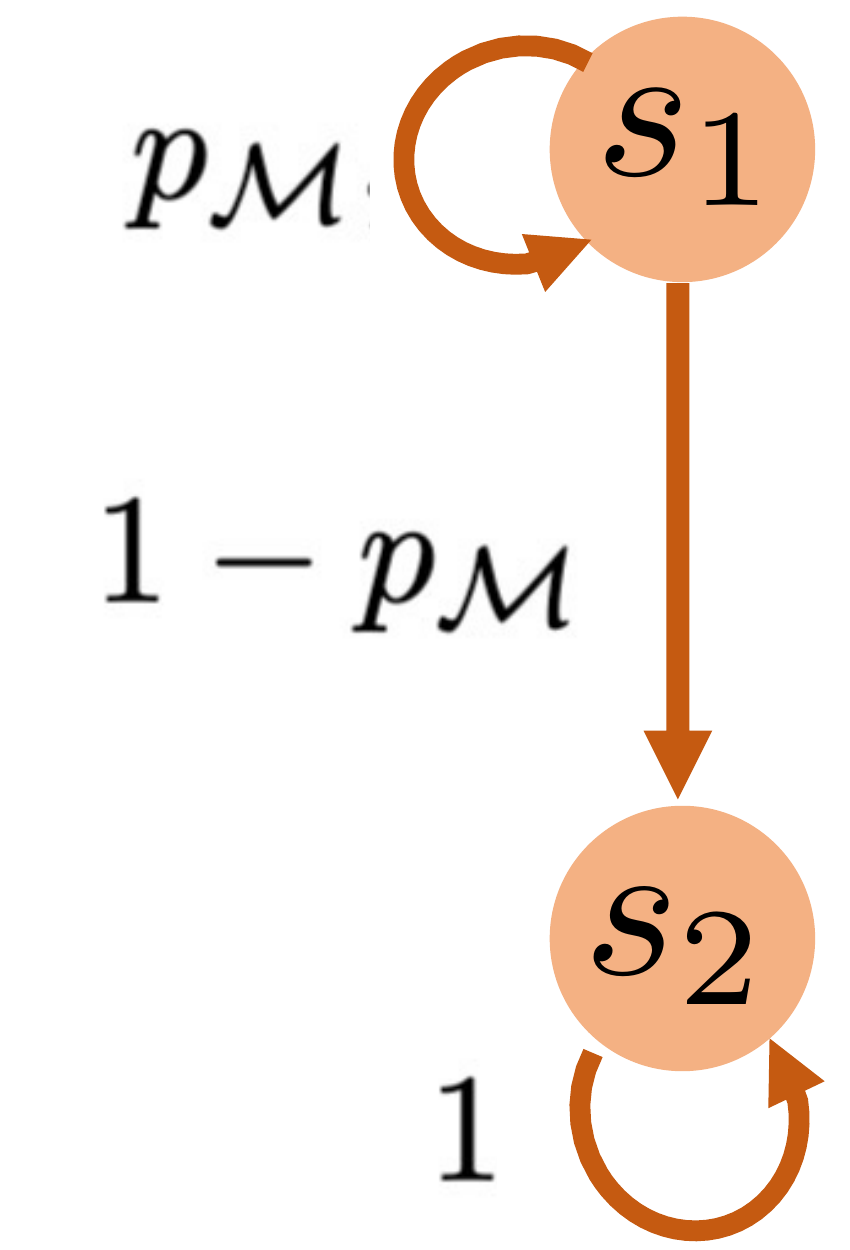}
        \caption{constructed MDP in Section~\ref{sec:analysis-lower-bound-Q-generative}}
        \label{fig:LB-MDP-generative-Q}
    \end{subfigure}
    \hfill
    \begin{subfigure}{0.55\textwidth}
        \centering
        \includegraphics[height=0.35\textwidth]{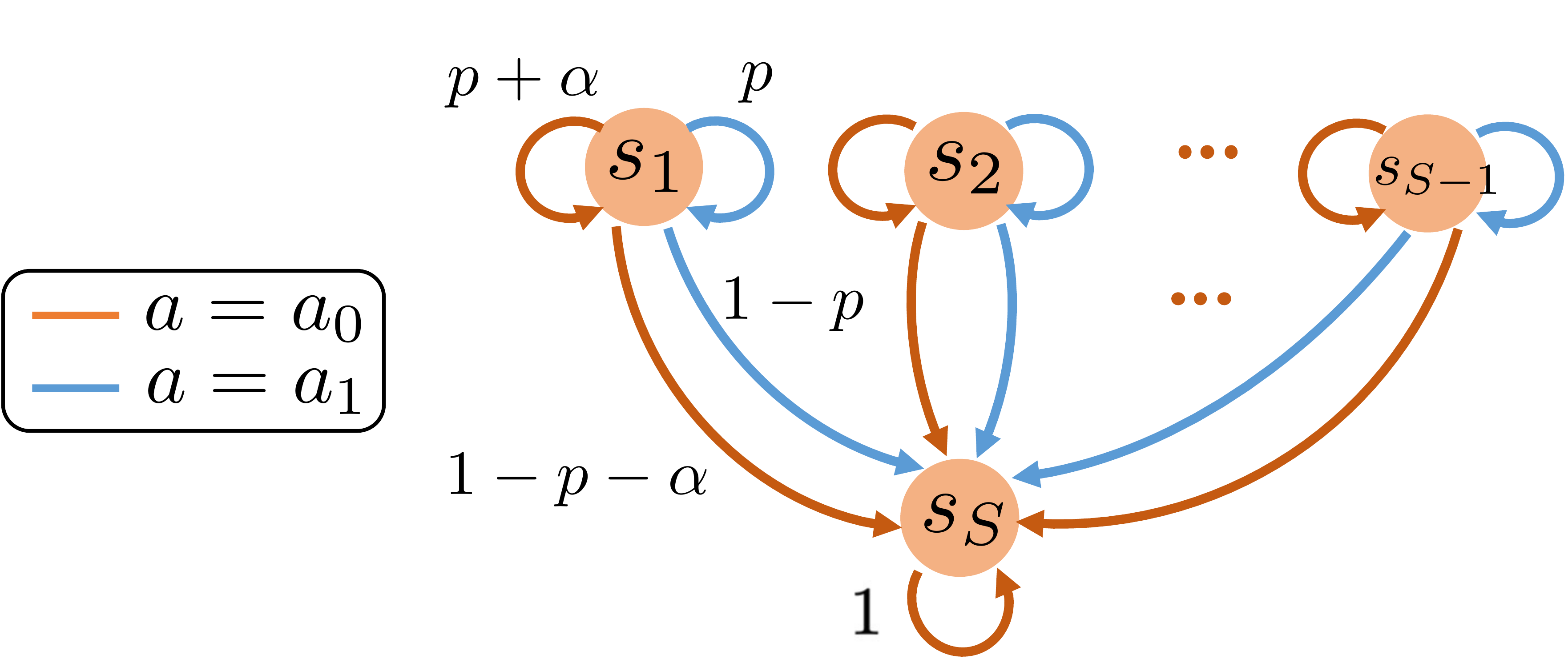}
        \caption{constructed MDP $\mathcal{M}_{0,0}$ in Section~\ref{sec:analysis-lower-bound-policy-generative} ($A=2$)}
        \label{fig:LB-MDP-generative-policy}
    \end{subfigure}

\end{center}
    \caption{Illustration of MDP construction for the minimax lower bounds.}

    \label{fig:LB-MDP-generative}

\end{figure}

%state-action space of cardinality $N = 3L$, where $L$ is a positive integer.

%with their relationships shown in Figure~\ref{fig:LB-MDP-generative}. 

%In the rest of the proof, we show that it takes at least 
%$$ \frac{N}{c_1 (1-\gamma)^3 \varepsilon^2} \log \left(\frac{N}{c_2 \delta} \right) ,$$
%many samples to distinguish these two MDPs with probability $1 - \delta.$ 

%By construction of these MDPs, learning $Q_{\mathcal{M}}^\star(z)$ for each $z \in s_0 \times \mathcal{A}$ is independent of each other. Then it boils down to studying the sample complexity for each $Q_{\mathcal{M}}^\star(z)$ on the class of MDP $\mathcal{M}^\star$.
%\yxc{TODO}

Before presenting the proof of Theorem~\ref{theorem:azar}, we first consider a related problem of Q-function estimation in a family of small-scale MDPs containing two states and one action. As it turns out, such a simplified setting suffices to illustrate the key ideas underlying the application of Fano's lemma for lower bound analyses, while also revealing the fundamental bottleneck associated with the effective horizon that arises in more general policy learning and value estimation tasks.

Consider any $0<\varepsilon<\frac{1}{32(1-\gamma)}$ and any $\gamma \in [2/3,1)$. 
Our goal in this subsection is to show that one can construct two MDPs $\mathcal{M}_0$ and $\mathcal{M}_1$ --- each with two states and a single action --- such that if $N$ independent samples are drawn from the generative model for each state-action pair and  
\begin{align}
\label{eq:N-UB-test-lower-Q-generative-overall}
    N\leq \frac{0.003}{(1-\gamma)^3\varepsilon^2}, 
\end{align}
then any estimator $\widehat{Q}$ must satisfy 
\begin{align}
\mathbb{P}_{\mathcal{M}_i}\Big(\big|\widehat{Q}(s_1,a)-Q^{\star}_{\mathcal{M}_i}(s_1,a)\big|> \varepsilon\Big)> 0.1 
\quad \text{for some } i\in \{0,1\}.
\label{eq:P-Mi-Q-inaccurate-lower-generative}
\end{align}
Here, we denote by $\mathbb{P}_{\mathcal{M}_0}(\cdot)$ (resp.~$\mathbb{P}_{\mathcal{M}_1}(\cdot)$) the probability measure under $\mathcal{M}_0$ (resp.~$\mathcal{M}_1$). 
In other words, no algorithm can simultaneously estimate the optimal Q-function for both $\mathcal{M}_0$ and $\mathcal{M}_1$ to the desired level of precision $\varepsilon$, unless the sample size per state-action pair exceeds the order of $\frac{1}{(1-\gamma)^3\varepsilon^2}$.  
The main idea of the proof is as follows: we identify two simple MDPs that are difficult to distinguish, formulate a binary hypothesis testing problem to determine which of these two MDPs generates the observed samples, and then relate the resulting minimax testing error (obtained via Fano's lemma) to the estimation risk.

\paragraph{Step 1: constructing hard MDP instances.} 

The first step is to construct two MDP instances,  deliberately designed to be nearly indistinguishable unless sufficiently many samples are available. We begin by specifying the common structure and properties shared by the two MDPs under consideration.
%
%The first step lies in constructing a collection $\mathcal{C}_{\mathsf{mdp}}$ of MDP instances. 
%
%Take $L$ to be a positive integer; 
Specifically, each MDP $\mathcal{M}$ of interest is described as follows; see Figure~\ref{fig:LB-MDP-generative}(a) for an illustration. 
%s
\begin{itemize}

\item {\em State space.}
The state space $\mathcal{S}$ contains $2$ states, given by
\begin{align}
    \mathcal{S}= \{s_1,s_2\}.
    \label{eq:state-space-hard-instance-generative}
\end{align}

\item {\em Action space.}  In each state $s\in \mathcal{S}$, the action space is a singleton set $\mathcal{A}=\{a\}$.

\item {\em Probability transition kernel.} We now specify the transition kernel of the MDP $\mathcal{M}$, denoted by $P_{\mathcal{M}}$. 
\begin{itemize}
%    \item In state $s_0$, taking action $a_i$ leads deterministically to state $s_i$, i.e., 
%    %
%    \begin{equation}
%        P_{\mathcal{M}}( s_i \mymid s_0, a_i) = 1,
%        \qquad \forall 1\leq i\leq 2. 
%    \end{equation}

    \item For state $s_1$, set
\begin{subequations}
    \begin{equation}
        P_{\mathcal{M}}( s \mymid s_1, a) =
        \begin{cases}
           p_{\mathcal{M}}, & \text{if }s= s_1 \\
           1-p_{\mathcal{M}}, & \text{if }s=s_{2} 
        \end{cases}
        \label{eq:PM-hard-instance-generative}
    \end{equation}  
    for some parameter $0<p_{\mathcal{M}}<1$.  
    That is, when the current state is $s_1$, the process either remains in the same state or transitions to $s_2$.

    \item State $s_2$ is set to be an absorbing state, i.e., 
    \begin{equation}
        P_{\mathcal{M}}( s_2 \mymid s_2, a) =1.
    \end{equation}  

    \end{subequations}
\end{itemize}

\item {\em Immediate rewards.} The reward function  is chosen to be
\begin{equation}
    r(s,a)= \begin{cases}
    1, & \text{if }s=s_1, \\
    0, & \text{else}. 
    \end{cases}
\end{equation}

\end{itemize}
Clearly, a key parameter governing the optimal Q-function in the above construction of $\mathcal{M}$ is $p_{\mathcal{M}}$. One can readily calculate the optimal Q-function in $\mathcal{M}$, denoted by $Q^{\star}_{\mathcal{M}}$, through the Bellman equation: 
\begin{align}
    Q_{\mathcal{M}}^\star(s_1, a) = 
    %\gamma V^\star(s_0) =
    \frac{1}{1 - \gamma p_{\mathcal{M}}}.
    %, \qquad \forall a_i \in \mathcal{A}.
    \label{eq:Qstar-hard-instance-generative}
\end{align}
\begin{proof}[Proof of \eqref{eq:Qstar-hard-instance-generative}]
%Fo notational simplicity, we suppress the subscript $\mathcal{M}$ in $Q_{\mathcal{M}}^{\star}$ and $V_{\mathcal{M}}^{\star}$ as long as it is clear from the context. 
%
First, since $s_2$ is an absorbing state with 0 immediate reward, it follows that $V_{\mathcal{M}}^\star(s_2) = 0$.
Second, the Bellman equation gives
\begin{align*}
    V_{\mathcal{M}}^\star(s_1) &= Q_{\mathcal{M}}^\star(s_1, a) =r(s_1, a)+\gamma \big\{ p_{\mathcal{M}} V_{\mathcal{M}}^\star(s_1) + (1-p_{\mathcal{M}}) V_{\mathcal{M}}^\star(s_{2})\big\} \\
    &= 1 + \gamma  p_{\mathcal{M}} V_{\mathcal{M}}^\star(s_1),
    %\\
    %&\qquad\qquad \Longrightarrow \qquad V^\star(s_j) = \frac{1}{1-\gamma p_{\mathcal{M}}}.
\end{align*}
and hence $Q_{\mathcal{M}}^\star(s_1, a)=V_{\mathcal{M}}^\star(s_1) = \frac{1}{1-\gamma p_{\mathcal{M}}}$. 
%Finally, it follows again from the Bellman equation that
%
%\begin{align*}
%    Q^\star(s_0, a_i)=\gamma V^\star(s_i) = \frac{\gamma}{1-\gamma p_{\mathcal{M}}},\qquad \forall a_i\in \mathcal{A}
%\end{align*}
%
%as claimed. 
\end{proof}

  In the sequel, we focus on two MDPs constructed as above, denoted by $\mathcal{M}_0$ and $\mathcal{M}_1$, which differ only in their choice of $p_{\mathcal{M}}$ (cf.~\eqref{eq:PM-hard-instance-generative}): 
\begin{align}
    p_{\mathcal{M}} = \begin{cases}
p, &\qquad \text{if }\mathcal{M} = \mathcal{M}_0,\\[3pt]
p+\alpha, &\qquad \text{if }\mathcal{M} = \mathcal{M}_1,
\end{cases}
\label{eq:choice-pM-lower-bound-generative}
\end{align}
where $p$ and $\alpha$ are parameters to be specified shortly, satisfying $1/2\leq p<p+\alpha <1$. 
In what follows, we denote by $\mathbb{E}_{\mathcal{M}_0}[\cdot]$ (resp.~$\mathbb{E}_{\mathcal{M}_1}[\cdot]$) the expectation under $\mathcal{M}_0$ (resp.~$\mathcal{M}_1$). The main step of the proof is to analyze the (in)distinguishability of these two MDPs given samples drawn from the generative model.

%Focusing on the two MDPs $\mathcal{M}_0$ and $\mathcal{M}_1$, we seek to analyze the accuracy of estimating $Q_{\mathcal{M}_0}^{\star}$ and $Q_{\mathcal{M}_1}^{\star}$.

\paragraph{Step 2: applying Fano's lemma.} 
Suppose that for each state-pair, $N$ independent samples are drawn from the generative model. In this setting, all useful information is contained in whether the process transitions to $s_2$ after executing action $a$ in state $s_1$.

Let $(s_1,a, s^{(i)}_{\mathsf{next}})$ ($1\leq i\leq N$) denote the  $N$ independent samples drawn w.r.t.~the state–action pair $(s_1,a)$. We introduce a set of random variables $\iota_1, \ldots, \iota_N$, where $\iota_i = \mathds{1}\{s^{(i)}_{\mathsf{next}}=s_1\}$ is an indicator of staying in state $s_1$ in the $i$-th sample, and is hence an independent Bernoulli random variable.  %
Let $P_{0}$ and $P_1$ denote the distributions of all observed samples under $\mathcal{M}_0$ and $\mathcal{M}_1$, respectively, and denote by $P_{0,\iota_i}$ and $P_{1,\iota_i}$ the distributions of $\iota_i$ under $\mathcal{M}_0$ and $\mathcal{M}_1$, respectively. In view of the tensorization property of KL divergence (cf.~Lemma~\ref{lem:KL-tensorize}), we have
\begin{align}
\mathsf{KL}(P_{1}\,\|\,P_{0})&=\sum_{i=1}^{N}\mathsf{KL}(P_{1,\iota_{i}}\,\|\,P_{0, \iota_{i}})=N\,\mathsf{KL}(p+\alpha\,\|\,p) \notag\\
&\leq \frac{N\alpha^2}{p(1-p)} \leq \frac{2N\alpha^2}{1-p} \eqqcolon \beta,
%\frac{N\alpha^2}{2p(1-p)} \geq \frac{N\alpha^2}{2(1-p)}, 
\label{eq:KL-P1-P0-lower-generative-Q}
\end{align}
where  $\mathsf{KL}(p+\alpha\,\|\,p)$ denotes the KL divergence between $\mathsf{Bern}(p+\alpha)$ and $\mathsf{Bern}(p)$, and the last line arises from Lemma~\ref{lem:KL-quadratic-lower-bound} and the property $p\geq 1/2$.  As a consequence, Fano's lemma (see Lemma~\ref{lem:fano}) implies that the minimax probability of error necessarily exceeds
\begin{align}
    p_{\mathrm{e}}\coloneqq \inf_{\psi}\max_{0\leq i\leq 1}\mathbb{P}_{\mathcal{M}_i}(\psi \neq i) \geq \max \bigg\{ \frac{1}{4}\exp(-\beta), \frac{1-\sqrt{\beta/2}}{2} \bigg\}, 
    \label{eq:lb-fano-pe-simple-Q-generative}
\end{align}
where the infimum is taken over all possible tests $\psi$. In particular, 
\begin{align}
    p_{\mathrm{e}} \geq \max \bigg\{ \frac{1}{4}\exp(-\beta), \frac{1-\sqrt{\beta/2}}{2} \bigg\} \geq 0.1
    \qquad \text{if }\beta \leq 0.9.
    \label{eq:lb-fano-pe-simple-Q-generative-simpler}
\end{align}

\paragraph{Step 3: choosing parameters $p$ and $\alpha$.}
In light of \eqref{eq:lb-fano-pe-simple-Q-generative-simpler}, to guarantee that $p_{\mathrm{e}}$ is lower bounded by 0.1, it suffices to choose the parameters $p$ and $\alpha$ to ensure that $\beta\leq 0.9$. Recall that $N$ needs to be no larger than the order of $\frac{1}{(1-\gamma)^3\varepsilon^2}$ (see \eqref{eq:N-UB-test-lower-Q-generative-overall}). Motivated by this, we select $p,\alpha$ as
\begin{align}
p = \frac{4\gamma -1}{3\gamma} 
\qquad \text{and} 
\qquad \alpha= \frac{2(1-\gamma p)^2\varepsilon}{\gamma^2}=\frac{32(1-\gamma)^2\varepsilon}{9\gamma^2}.
\label{eq:choise-p-alpha-lower-bound-generative}
\end{align}
It is readily seen that as $\gamma\rightarrow 1$, we have $p\rightarrow 1$ and $\alpha\rightarrow 0$, so that $p_{\mathcal{M}_0}$ and  $p_{\mathcal{M}_1}$ become increasingly similar and harder to differentiate. Moreover, one can verify that
$1-p \asymp 1-\gamma$, $\alpha \asymp (1-\gamma)^2\varepsilon$, and 
 hence $\alpha^2/(1-p)\asymp (1-\gamma)^3\varepsilon^2$, which matches the inverse of the upper bound~\eqref{eq:N-UB-test-lower-Q-generative-overall} on $N$.  
 %hence when $\gamma\rightarrow 1$, we have $p\rightarrow 1$ and $\alpha\rightarrow 0$.  
%In other words, $p_{\mathcal{M}_0}$ and  $p_{\mathcal{M}_1}$ are constructed to be increasingly similar as $\gamma\rightarrow 1$, thereby making them difficult to differentiate.
%

A few simple yet useful properties are summarized below.
\begin{lemma}\label{lem:properties-M01-lower-generative}
Consider the above MDPs with parameters~\eqref{eq:choise-p-alpha-lower-bound-generative}. Suppose that $0<\varepsilon<\frac{1}{32(1-\gamma)}$ and $\gamma \in [2/3,1)$. Then one has
\begin{itemize}
\item[(i)] $ 1/2\leq p< p+\alpha < 1 $ and $1-p\geq 2\alpha$;

\item[(ii)]  
$
   Q^\star_{{\mathcal{M}}_1}(s_1,a) - Q^\star_{{\mathcal{M}}_0}(s_1,a)  > 2\varepsilon; 
%   \label{eq:Qstar-M10-gap-lower-bound-generative}
$
%for all $a\in \mathcal{A}$. 
%

%\item[(ii)] $\alpha^2 \leq p^2 / 2$ and $\alpha^2\leq p(1-p)$. 

\item[(iii)] $\frac{\alpha^{2}}{1-p}=\frac{32^{2}(1-\gamma)^{3}\varepsilon^{2}}{27\gamma^{3}}\leq128(1-\gamma)^{3}\varepsilon^{2}$. 
\end{itemize}
\end{lemma}
\begin{proof}[Proof of Lemma~\ref{lem:properties-M01-lower-generative}]
Regarding (i), it suffices to see that
\[
    1-p = \frac{1-\gamma}{3\gamma} > \frac{64(1-\gamma)^2\varepsilon}{9\gamma^2}=2\alpha
    \quad \text{and} \quad
    p=\frac{4\gamma-1}{3\gamma}> \frac{1}{2}
\]
under our conditions on $\varepsilon$ and $\gamma$. 
Regarding (ii), \eqref{eq:Qstar-hard-instance-generative} tells us that
\begin{align}
   Q^\star_{{\mathcal{M}}_1}(s_1,a) - Q^\star_{{\mathcal{M}}_0}(s_1,a) &= \frac{1}{1 - \gamma (p+\alpha)} - \frac{1}{1 - \gamma p} \notag\\
   &= \frac{\gamma \alpha}{(1-\gamma p)(1-\gamma(p+\alpha))} > 2\varepsilon, 
   \label{eq:Qstar-M10-gap-lower-bound-generative}
\end{align}
where the last inequality uses the choice \eqref{eq:choise-p-alpha-lower-bound-generative} of $\alpha$. 
Property (iii) also follows from direct calculation under our conditions on $\gamma$. 
\end{proof}
Combine Lemma~\ref{lem:properties-M01-lower-generative} with \eqref{eq:lb-fano-pe-simple-Q-generative-simpler} and the definition \eqref{eq:KL-P1-P0-lower-generative-Q} of $\beta$ to demonstrate that the minimax probability of testing error obeys 
\begin{align}
\label{eq:N-UB-test-lower-Q-generative}
p_{\mathrm{e}}\geq 0.1
\qquad \text{as long as }
    N\leq \frac{0.003}{(1-\gamma)^3\varepsilon^2}. 
\end{align}

\paragraph{Step 4: connecting testing error with estimation risk.} 
Finally, we show how the minimax probability of testing error can be leveraged to bound the minimax estimation risk.  Recall from Lemma~\ref{lem:properties-M01-lower-generative}(ii) that $Q^\star_{{\mathcal{M}}_1}(s_1,a) - Q^\star_{{\mathcal{M}}_0}(s_1,a)  > 2\varepsilon$. Suppose instead that there exists an estimator $\widehat{Q}$ such that 
\begin{align}
\mathbb{P}_{\mathcal{M}_i}\Big(\big|\widehat{Q}(s_1,a)-Q^{\star}_{\mathcal{M}_i}(s_1,a)\big|\leq \varepsilon\Big)> 0.9
,
\qquad \forall i\in \{0,1\}.
\label{eq:assump-Qhat-good-for-01-lower-generative}
\end{align}
Based on this estimator, consider the test
\[
    \psi = \begin{cases}
    0, &\text{if }\widehat{Q}(s_1,a)\leq \frac{1}{1-\gamma p}+\varepsilon \\
    1, &\text{else}.
    \end{cases}
\]
By construction, under $\mathcal{M}_0$ we have
\begin{align*}
    \mathbb{P}_{\mathcal{M}_0}(\psi = 0) &= \mathbb{P}_{\mathcal{M}_0}\bigg(\widehat{Q}(s_1,a)\leq \frac{1}{1-\gamma p}+\varepsilon\bigg) \notag\\
    &= \mathbb{P}_{\mathcal{M}_0}\big(\widehat{Q}(s_1,a)\leq Q^{\star}_{\mathcal{M}_0}(s_1,a)+\varepsilon\big)
    > 0.9,
\end{align*}
where the last inequality arises from the assumption \eqref{eq:assump-Qhat-good-for-01-lower-generative}. An analogous argument shows that $\mathbb{P}_{\mathcal{M}_1}(\psi=1)>0.9$. As a result, the testing error of $\psi$ satisfies
\[
    p_{\mathrm{e}}(\psi) \coloneqq \max \big\{ \mathbb{P}_{\mathcal{M}_0}(\psi\neq 0), \mathbb{P}_{\mathcal{M}_1}(\psi\neq 1)\big\} < 0.1.
\]
This contradicts the lower bound \eqref{eq:N-UB-test-lower-Q-generative} on the minimax error probability,  thereby establishing the claimed result \eqref{eq:P-Mi-Q-inaccurate-lower-generative}.

\subsection{Proof of Theorem~\ref{theorem:azar}}
\label{sec:analysis-lower-bound-policy-generative}

We now turn to establishing the lower bound in  Theorem~\ref{theorem:azar}, covering both policy learning and estimation of optimal Q-functions. Our construction and analysis largely build upon those developed in Section~\ref{sec:analysis-lower-bound-Q-generative}, with modifications to expand the state-action space to include $S$ states and $A$ actions. At a high level, we again construct a family of similar MDPs and formulate a multi-hypothesis testing problem aimed at identifying the underlying data-generating MDP; we then connect the minimax error probability of this testing problem with the performance of any RL algorithm.

\paragraph{Step 1: constructing hard MDP instances.} 
Let us construct a family of $\gamma$-discounted infinite-horizon MDPs with $S$ states and $A$ actions. We begin by presenting their common structure before describing their differences in the transition kernels. 
\begin{itemize}

\item {\em State space.}
The state space $\mathcal{S}$ contains $S$ states, given by
\begin{align*}
    \mathcal{S}= \{s_1,\cdots, s_S\}. 
    %\label{eq:state-space-hard-instance-generative-policy}
\end{align*}
The state space $\mathcal{S}$ is partitioned into the following subsets:
\begin{equation*}
\mathcal{S}_1 = \{s_{1},\dots,s_{S-1}\},
\qquad 
\mathcal{S}_2 = \{s_{S}\},
%\label{eq:defn-S1-S2-lower-bound-generative-policy}
\end{equation*}
where each subset is associated with distinct transition probabilities, to be specified shortly.

\item {\em Action space.}  For ease of exposition, we allow the action space to vary across states. For any state in $\mathcal{S}_1$, the action space contains $A$ actions, i.e., 
    $$\mathcal{A} \coloneqq \{a_0,a_1,\dots,a_{A-1}\}.$$
In contrast, 
%for any state in $\mathcal{S}_2$,
the action space associated with $s_S$ is the singleton set $\{a_0\}$. 

\item {\em Immediate rewards.} The reward function  is chosen to be
\begin{equation*}
    r(s,a)= \begin{cases}
    1, & \text{if }s\in \mathcal{S}_1,\\
    0, & \text{else}. 
    \end{cases}
\end{equation*}
Since only states in $\mathcal{S}_1$ yield positive rewards, this structure favors policies that induce a higher probability of remaining within $\mathcal{S}_1$.

%As a result, a higher probability of staying in $\mathcal{S}_1$ often results in higher values. 
\end{itemize}

Having described the common structure shared by this family of MDPs, we now specify the optimal policy and transition kernel for each instance. In particular, we construct a collection of 
\begin{align}
M=1 + (S-1)(A-1)
\end{align}
MDP instances:
\begin{equation}
\mathcal{C}_{\mathsf{mdp}} = \{\mathcal{M}_{0,0}\} \cup 
\big\{\mathcal{M}_{j,k}\mid 1 \leq j <S, 1 \leq k < A\big\}. 
\label{eq:defn-Cmdp-lower-bound-generative}
\end{equation}
%
%denoted by $\mathcal{M}_{0,0}$ and $\{\mathcal{M}_{j,k}\}_{1 \leq j \leq S/2, 1 \leq k < A}$.
As in Section~\ref{sec:analysis-lower-bound-Q-generative}, the construction is parameterized by two scalars $p$ and $\alpha$ satisfying $1/2 \leq p < p + \alpha < 1$ and $1-p\geq 2\alpha$. See Figure~\ref{fig:LB-MDP-generative}(b) for an illustration of $\mathcal{M}_{0,0}$ with $A=2$. 

\begin{itemize}

\item {\em Optimal policy.} The optimal policy of $\mathcal{M}_{j,k}$, denoted by $\pi^{\star}_{j,k}$, is chosen such that
\begin{subequations}
\label{eq:optimal-policy-lower-bound-policy-generative}
\begin{align}
\pi_{0,0}^{\star}(s)=a_{0}\qquad\forall s\in\mathcal{S}_{1}
\end{align}
and for any $1\leq j<S$ and $1\leq k<A$, 
\begin{align}
\pi_{j,k}^{\star}(s)=\begin{cases}
a_{k}, & \text{if }s=s_{j},\\
a_{0}, & \text{if }s\in\mathcal{S}_{1}\text{ and }s\neq s_{j}.
\end{cases}
\end{align}

\end{subequations}
%
%We do not specify the optimal policies on $\mathcal{S}_2$ here, as they are not needed for the subsequent analysis. 
The transition probability kernels will be constructed to ensure consistency with these optimal policies.

\item {\em Probability transition kernel.}  
The probability transition kernel of $\mathcal{M}_{j,k}$, denoted by $P_{\mathcal{M}_{j,k}}$, is defined below. 
\begin{itemize}
    \item In any state $s_i$ with $1\leq i< S$ (so that $s_i\in \mathcal{S}_1$), set
    \begin{equation*}
        P_{\mathcal{M}_{j,k}}( s \mymid s_i, a) =
        \begin{cases}
           p+\alpha, & \text{if }a=\pi_{j,k}^{\star}(s_i)\text{ and }s=s_i,\\
           1-p-\alpha, & \text{if }a=\pi_{j,k}^{\star}(s_i)\text{ and }s=s_{S}, \\
                p, & \text{if }a\neq \pi_{j,k}^{\star}(s_i)\text{ and }s=s_i, \\
           1-p, & \text{if }a\neq \pi_{j,k}^{\star}(s_i)\text{ and }s=s_{S}, \\  
           0 & \text{else}.
        \end{cases}
    \end{equation*}  
    In words, executing the action $\pi_{j,k}^{\star}(s_i)$ in state $s_i \in \mathcal{S}_1$ has a higher probability of remaining in the same state. Given that states in $\mathcal{S}_1$ are associated with higher immediate rewards, this design ensures that taking action $\pi_{j,k}^{\star}(s_i)$ in state $s_i$ yields a larger Q-value.

    \item 
    %Any state $s_i$ with $i>S/2$ (so that $s_i\in \mathcal{S}_2$) 
    The last state $s_S$ is set to be an absorbing state, i.e., 
    \begin{equation*}
        P_{\mathcal{M}_{j,k}}( s_S \mymid s_S, a_0) =1.
        %,\qquad \forall a\in \mathcal{A}.
    \end{equation*}  
\end{itemize}

\end{itemize}
\noindent 
Evidently, the transition probabilities of these MDP instances differ only in those corresponding to the optimal actions; all non-optimal actions share the same transition probabilities.

Repeating the same arguments as for \eqref{eq:Qstar-hard-instance-generative} immediately reveals that: for each $s\in \mathcal{S}_1$,  
\begin{align}   Q_{\mathcal{M}_{j,k}}^\star(s, a) = 
\begin{cases}
    \frac{1}{1 - \gamma (p+\alpha)} & \text{if } a = \pi_{j,k}^{\star}(s),\\
    \frac{1}{1 - \gamma p} & \text{if } a \neq  \pi_{j,k}^{\star}(s).
\end{cases}
    \label{eq:Qstar-hard-instance-generative-policy-learning}
\end{align}
This characterization is clearly consistent with the optimal policy defined in \eqref{eq:optimal-policy-lower-bound-policy-generative}.

\paragraph{Step 2: applying Fano's lemma.} To invoke Fano's lemma stated in Lemma~\ref{lem:fano}, we need to examine the pairwise KL divergences among the constructed MDP instances.   
Recall that $N$ independent samples are drawn from the generative model for each state-action pair. 
Let $P_{\mathcal{M}}$ and $\pi_{\mathcal{M}}^{\star}$ denote the distribution of the observed samples and the optimal policy under $\mathcal{M}$, respectively. By construction, the optimal policies of $\mathcal{M}_{j,k}$ (for $j \neq 0$) and $\mathcal{M}_{0,0}$ differ in exactly one state, leading to two state-action pairs with different transition probabilities in $P_{\mathcal{M}_{j,k}}$ and $P_{\mathcal{M}_{0,0}}$. Consequently, the corresponding sample distributions differ only in the $2N$ samples associated with these two state-action pairs. 

Among these $2N$ samples, $N$ of them contribute a KL divergence of $\mathsf{KL}(p+\alpha \,\|\, p)$ each, while the remaining $N$ contribute $\mathsf{KL}(p \,\|\, p+\alpha)$ each. Therefore, the overall KL divergence between $P_{\mathcal{M}_{j,k}}$ and $P_{\mathcal{M}_{0,0}}$ can be expressed as
\[
\mathsf{KL}\big(P_{\mathcal{M}_{j,k}}\,\|\,P_{\mathcal{M}_{0,0}}\big)=N\,\mathsf{KL}(p+\alpha\,\|\,p)+N\,\mathsf{KL}(p\,\|\,p+\alpha)
\]
for $j\neq 0$. 
Repeating the arguments for
\eqref{eq:KL-P1-P0-lower-generative-Q}, we further arrive at
\begin{align}
\mathsf{KL}\big(P_{\mathcal{M}_{j,k}}\,\|\,P_{\mathcal{M}_{0,0}}\big) &\leq \frac{N\alpha^2}{p(1-p)}+\frac{N\alpha^2}{(p+\alpha)(1-p-\alpha)} \notag\\
&\leq \frac{2N\alpha^2}{(p+\alpha)(1-p-\alpha)} 
\leq \frac{8N\alpha^2}{1-p},
\label{eq:KL-upper-bound-1357-generative}
\end{align}
with the proviso that $1/2\leq p < p+\alpha <1$ 
and $1-p-\alpha \geq (1-p)/2$ (a consequence of $1-p\geq 2\alpha$).  

Define the minimax probability of testing error as
\begin{equation}
p_{\mathrm{e}} \coloneqq \inf_{\psi} \max_{\mathcal{M}\in \mathcal{C}_{\mathsf{mdp}}} \mathbb{P}_{\mathcal{M}} (\psi \neq \mathcal{M}),
\end{equation}
where the infimum is taken over all tests that map the observed data to an MDP in $\mathcal{C}_{\mathsf{mdp}}$ (cf.~\eqref{eq:defn-Cmdp-lower-bound-generative}).  
Fano's lemma (cf.~Lemma~\ref{lem:fano}) gives
\begin{align}
p_{\mathrm{e}}\geq \frac{\log(M+1)-\log 2}{\log M} - \frac{1}{4} \geq \frac{1}{4},
\label{eq:pe-lower-bound-policy-generative}
\end{align}
provided that $M\geq 4$ and
\begin{align}
\frac{1}{M+1} \sum_{\mathcal{M}\in \mathcal{C}_{\mathsf{mdp}}} \mathsf{KL}( P_{\mathcal{M}}\,\|\, P_{\mathcal{M}_{0,0}}) \leq \frac{1}{4} \log M.
\label{eq:pe-lower-bound-policy-generative-234}
\end{align}
With \eqref{eq:KL-upper-bound-1357-generative} in place, this KL condition \eqref{eq:pe-lower-bound-policy-generative-234} --- and hence \eqref{eq:pe-lower-bound-policy-generative} --- can be guaranteed if
\begin{align}
\frac{8N\alpha^2}{1-p} \leq 
\frac{1}{4} \log \frac{SA}{4}
\leq 
\frac{1}{4} \log M.
\label{eq:KL-simplified-UB-generative}
\end{align}

\paragraph{Step 3: choosing parameters $p$ and $\alpha$.} 
It then comes down to choosing the parameters 
$p$ and $\alpha$ so as to satisfy \eqref{eq:KL-simplified-UB-generative}. 
Akin to Section~\ref{sec:analysis-lower-bound-Q-generative}, 
we set
\begin{align*}
p = \frac{4\gamma -1}{3\gamma} 
\qquad \text{and} 
\qquad \alpha= \frac{2(1-\gamma p)^2\varepsilon}{\gamma^2}=\frac{32(1-\gamma)^2\varepsilon}{9\gamma^2},
%\label{eq:choise-p-alpha-lower-bound-generative-policy}
\end{align*}
From Lemma~\ref{lem:properties-M01-lower-generative}(iii), it follows that $\frac{\alpha^2}{1-p}\leq 128(1-\gamma)^{3}\varepsilon^{2}$, and as a result, Condition \eqref{eq:KL-simplified-UB-generative} is satisfied provided that
\begin{equation*}
N\leq\frac{1}{4096}\frac{1}{(1-\gamma)^{3}\varepsilon^{2}}\log\frac{SA}{4}.
\end{equation*}

\paragraph{Step 4: connecting testing error with performance of policy learning.} 
We now relate the minimax probability of testing error $p_{\mathrm{e}}$ to the statistical risk of policy learning. As shown in \eqref{eq:Qstar-hard-instance-generative-policy-learning} and Lemma~\ref{lem:properties-M01-lower-generative}(ii), for any $\mathcal{M}\in \mathcal{C}_{\mathsf{mdp}}$ and any state $s\in \mathcal{S}_1$, the optimal action is separated from all suboptimal actions by the following gap:
\begin{align}
Q^{\star}_{\mathcal{M}}\big(s, \pi_{\mathcal{M}}^{\star}(s) \big) -  Q^{\star}_{\mathcal{M}}\big(s, a \big) > 2\varepsilon,
\qquad \forall a\neq \pi_{\mathcal{M}}^{\star}(s).
\label{eq:Q-sep-lower-bound-policy-generative}
\end{align}

Now suppose that there exists a (deterministic) policy estimate $\widehat{\pi}$ achieving
\begin{align}
    \mathbb{P}_{\mathcal{M}}( \| Q^{\widehat{\pi}} - Q^{\star}_{\mathcal{M}} \|_{\infty} > \varepsilon ) < 1/4,
    \qquad \forall \mathcal{M}\in \mathcal{C}_{\mathsf{mdp}}. 
    \label{eq:assumption-pihat-good-estimate-lower-generative}
\end{align}
In view of \eqref{eq:Q-sep-lower-bound-policy-generative}, the only policy that can satisfy  $\| Q^{\widehat{\pi}} - Q^{\star}_{\mathcal{M}} \|_{\infty} \leq  \varepsilon$ is $\pi^{\star}_{\mathcal{M}}$. Hence,  \eqref{eq:assumption-pihat-good-estimate-lower-generative} is equivalent to
\begin{align}
    \mathbb{P}_{\mathcal{M}}( \widehat{\pi} \neq \pi_{\mathcal{M}}^{\star} ) < 1/4,
    \qquad \forall \mathcal{M}\in \mathcal{C}_{\mathsf{mdp}}. 
    \label{eq:assumption-pihat-good-estimate-lower-generative-2}
\end{align}
By construction, the optimal policy $\pi_{\mathcal{M}}^{\star}$ uniquely identifies
 $\mathcal{M}$ within the collection $\mathcal{C}_{\mathsf{mdp}}$. Define the test
\[
\psi_{\mathsf{policy}} = \mathcal{M}_{\widehat{\pi}},
\]
with $\mathcal{M}_{\widehat{\pi}}$ indicating the MDP in $\mathcal{C}_{\mathsf{mdp}}$ whose optimal policy coincides with $\widehat{\pi}$.  Then it follows that
\begin{align*}
    \mathbb{P}_{\mathcal{M}}( \psi_{\mathsf{policy}} \neq \mathcal{M} )
    = \mathbb{P}_{\mathcal{M}}( \widehat{\pi} \neq \pi_{\mathcal{M}}^{\star} ) < 1/4,
    \qquad \forall \mathcal{M}\in \mathcal{C}_{\mathsf{mdp}},
%    \label{eq:assumption-pihat-good-estimate-lower-generative-2}
\end{align*}
which contradicts the lower bound \eqref{eq:pe-lower-bound-policy-generative}. We have thus established that for any policy estimator $\widehat{\pi}$,  there exists some $\mathcal{M}\in \mathcal{C}_{\mathsf{mdp}}$ such that 
\begin{align*}
    \mathbb{P}_{\mathcal{M}}( \| Q^{\widehat{\pi}} - Q^{\star}_{\mathcal{M}} \|_{\infty} > \varepsilon ) \geq 1/4.
\end{align*}

\paragraph{Step 5: connecting testing error with Q-function estimation risk.} 
It remains to link the minimax probability of testing error $p_{\mathrm{e}}$ with the estimation risk for learning optimal Q-functions. Suppose for the moment that there exists an estimator $\widehat{Q}$ such that
\begin{align}
    \mathbb{P}_{\mathcal{M}}( \| \widehat{Q} - Q^{\star}_{\mathcal{M}} \|_{\infty} > \varepsilon ) < 1/4,
    \qquad \forall \mathcal{M}\in \mathcal{C}_{\mathsf{mdp}}. 
    \label{eq:assumption-Qhat-good-estimate-lower-generative}
\end{align}
Combined with \eqref{eq:Q-sep-lower-bound-policy-generative}, this reveals that: 
under every $\mathcal{M}\in \mathcal{C}_{\mathsf{mdp}}$, the following event occurs with probability at least $3/4$: 
\begin{align}
\widehat{Q}\big(s,\pi_{\mathcal{M}}^{\star}(s)\big)-\widehat{Q}\big(s,a\big) & \geq Q_{\mathcal{M}}^{\star}\big(s,\pi_{\mathcal{M}}^{\star}(s)\big)-Q_{\mathcal{M}}^{\star}\big(s,a\big)-2\|\widehat{Q}-Q_{\mathcal{M}}^{\star}\|_{\infty} \notag\\
 & >2\varepsilon-2\varepsilon=0
 \label{eq:Q-hat-gap-lower-generative-135}
\end{align}
simultaneously for all $s\in \mathcal{S}$ and all $a\neq \pi_{\mathcal{M}}^{\star}(s)$. As an immediate consequence, if we define $\pi_{\mathsf{est}}$ to be the greedy policy w.r.t.~$\widehat{Q}$ (namely, $\pi_{\mathsf{est}}(s)=\arg\max_a \widehat{Q}(s,a)$ for all $s\in \cS$), then under $\mathcal{M}$ one must have $\pi_{\mathsf{est}}=\pi^{\star}_{\mathcal{M}}$. 

In light of this observation, consider the following test:
\[
    \psi_{\mathsf{est}} = \mathcal{M}_{\pi_{\mathsf{est}}},
\]
where $\mathcal{M}_{\pi_{\mathsf{est}}}$ denotes the MDP in $\mathcal{C}_{
\mathsf{mdp}
}$ whose optimal policy coincides with $ \pi_{\mathsf{est}}$. It is then readily seen that
\begin{align*}
    \mathbb{P}_{\mathcal{M}}( \psi_{\mathsf{est}} \neq \mathcal{M} )
    = \mathbb{P}_{\mathcal{M}}( \pi_{\mathsf{est}} \neq \pi_{\mathcal{M}}^{\star} ) < 1/4,
    \qquad \forall \mathcal{M}\in \mathcal{C}_{\mathsf{mdp}},
%    \label{eq:assumption-pihat-good-estimate-lower-generative-2}
\end{align*}
resulting in a contradiction with the minimax lower bound \eqref{eq:pe-lower-bound-policy-generative}. Therefore, we conclude that 
for any estimator $\widehat{Q}$,  there exists some $\mathcal{M}\in \mathcal{C}_{\mathsf{mdp}}$ such that 
\begin{align*}
    \mathbb{P}_{\mathcal{M}}( \| \widehat{Q} - Q^{\star}_{\mathcal{M}} \|_{\infty} > \varepsilon ) \geq 1/4.
\end{align*}

\section{Extension: finite-horizon nonstationary MDPs}
\label{sec:finite-horizon-generative}

%For $1\le h\le H$, let $\widehat{\bm Q}_h^{\star}$, $\widehat{\bm V}_h^{\star}$, and $\widehat{\pi}_h^{\star}$ represent the optimal action value function, optimal value function, and optimal policy of $\widehat{\mathcal{M}}$ w.r.t. empirical MDP $\widehat{\mathcal{M}}=(\cS,\cA, \{\widehat{ P}_h\}_{1\le h\le H}, \{ r_h\}_{1\le h\le H}, H)$ with the probability transition kernel $\{\widehat{ P}_h\}_{1\le h\le H}$, respectively.
%Assume that $\widehat{\pi}$ satisfies

%For any pre-specified order of $\mathcal{A}$, let $\widehat{\pi}$ be the policy given by
%\begin{align} \label{eq:policy-Q-finite}
%\forall (s,h) \in \mathcal{S} \times [H], \qquad \widehat{\pi}(s, h) = \min \{a \in \mathcal{A}: \widehat{Q}_h^{\star}(s,a) > \widehat{V}_h^{\star}(s) - \newnoise\},
%\end{align}
%where $\widehat{V}_h^{\star}(s) = \max_a \widehat{Q}_h^{\star}(s,a)$ and $\newnoise \sim \mathsf{Unif}(0,\xi)$.

%that obeys
%%
%\begin{align} \label{eq:policy-Q-finite}
%\forall (s,h) \in \mathcal{S} \times [H], \qquad \widehat{V}_h^{\widehat{\pi}}(s) > \widehat{V}_h^{\star}(s) - \xi. 
%\end{align}
%%

Moving beyond discounted infinite-horizon stationary MDPs, we now briefly discuss model-based policy learning for finite-horizon nonstationary MDPs. The proof techniques are similar in spirit to --- and in fact simpler than --- those for the discounted infinite-horizon setting, owing in part to the availability of independent samples across different steps $h$ over the entire horizon. For brevity, we refer interested readers to \citet{li2024breaking} for the full proof details. 
% conference version li2020breaking doesn't have finite-horizon contents.

\paragraph{A model-based approach.} %
We begin by describing what is arguably the most natural version of the model-based policy learning algorithm under the generative model. In essence, the algorithm performs model-based planning using the empirical model, just as its discounted infinite-horizon counterpart in Algorithm~\ref{alg:policy-learn-model-based}. 
\begin{itemize}
	\item[1)] {\em Model estimation.} For each state-action-step tuple $(s,a,h)\in \cS\times \cA\times [H]$, draw $N$ independent samples from the generative model:
\[
	s^{(i)}(s,a,h) \overset{\mathrm{i.i.d.}}{\sim} P_h(\cdot \mymid s,a), \qquad i=1,\ldots, N,
\]
and construct the empirical transition kernels $\{\widehat{P}_h\}_{h=1}^H$ via
\begin{equation}
	\forall s' \in \mathcal{S}, \qquad \widehat{P}_h(s^\prime \mymid s, a) = \frac{1}{N}\sum\nolimits_{i=1}^N \ind \{s^{\prime}  = s^{(i)}(s,a,h) \}.
	\label{eq:defn-empirical-P-finite}
\end{equation}
Namely, $\widehat{P}_h(s^\prime \mymid s, a)$ stands for the empirical frequency of transitions from $(s,a)$ to state $s^\prime$ at step $h$. 

\item[2)] {\em Planning.} Compute an optimal policy $\widehat{\pi}^{\star} = \{ \widehat{\pi}^{\star}_h\}_{h=1}^H$ for the empirical MDP $\widehat{\mathcal{M}} = (\cS, \cA, \widehat{P}, r, H)$. 
\end{itemize}
%
% let us consider the empirical MDP $\widehat{\mathcal{M}}=(\cS,\cA, \{\widehat{P}_h\}_{1\le h\le H}, \{r_h\}_{1\le h\le H}, H)$ constructed based on the data samples in hand (see \eqref{eq:defn-empirical-P-finite} for the definition of the empirical transition kernel $\widehat{P}_h$). We then run a classical dynamic programming algorithm \citep{bertsekas2017dynamic} to find an optimal policy $\widehat{\pi}^{\star}$ of the empirical MDP $\widehat{\mathcal{M}}$.  

Note that the optimal policy $\widehat{\pi}^{\star}_h$ of the empirical MDP $\widehat{\mathcal{M}}$ at step $h$ is computed via backward induction starting from $h=H$. This backward recursion ensures that $\widehat{\pi}^{\star}_h$ depends solely on the empirical transition kernels at or after step $h$, and as a result, $\widehat{\pi}^{\star}_h$ is statistically independent of the empirical transitions $\{\widehat{P}_j\}_{1\leq j < h}$. Consequently, the statistical dependencies across steps are naturally decoupled. In contrast to the discounted infinite-horizon setting, this obviates the need for the leave-one-out construction used there to decouple the statistical dependencies. As the following theorem shows, this simple model-based approach already achieves the desirable sample complexity. 
\begin{theorem}[\cite{li2024breaking}]
\label{Thm:sample-compl-finite}
	There exists some universal constant $c_0>0$ such that: for any $\delta > 0$ and $0<\varepsilon \leq H$, the policy $\widehat{\pi}^{\star}$ returned by the aforementioned  model-based algorithm obeys
	\begin{align*}
		\forall (s,h) \in \mathcal{S}\times [H] ,~~\,\,\, \qquad
		V_h^{\widehat{\pi}^{\star}}(s)  & \geq  V_h^\star(s) - \varepsilon \\
		\forall (s,a,h) \in \mathcal{S} \times \mathcal{A} \times [H], \qquad  Q_h^{\widehat{\pi}^{\star}}(s,a)  & \geq  Q_h^\star(s,a)  - \varepsilon 
	\end{align*}
	with probability at least $1-\delta$, provided that 
	%we set $\xi = c_1\varepsilon$ and that 
	the sample size for every tuple $(s,a,h)$ exceeds 
	\begin{align}
	\label{EqnSamples-finite}
		N \geq \frac{c_{0}H^3\log\big(\frac{HSA}{ 
		\delta}\big)}{\varepsilon^{2}}  . 
	\end{align}
\end{theorem}

Akin to the discounted infinite-horizon setting, the model-based approach achieves $\varepsilon$-accuracy as long as the sample size per state-action-step tuple $(s,a,h)$ exceeds the order 
\[
  \frac{H^3}{\varepsilon^2}  
\]
up to logarithmic factors. This guarantee, which holds for the full range $\varepsilon \in (0,H]$, mirrors the bound \eqref{EqnSamples-main} for the discounted infinite-horizon setting, with the effective horizon $\frac{1}{1-\gamma}$ now replaced by the finite horizon length $H$. Since there are $SAH$ distinct $(s,a,h)$-tuples in total, the overall sample complexity scales as 
\begin{equation}
	\widetilde{O} \bigg( \frac{SAH^4}{\varepsilon^2} \bigg).
	\label{eq:sample-size-finite-horizon-generative}
\end{equation}
The quadruple horizon dependence $H^4$ in this bound --- as opposed to the cubic scaling $1/(1-\gamma)^3$ seen in the discounted infinite-horizon case --- stems from time inhomogeneity. Because the transition kernels $P_h$ may vary across steps, one incurs an additional factor of $H$ in the sample complexity. Encouragingly, the sample complexity \eqref{eq:sample-size-finite-horizon-generative} is provably minimax optimal in view of the matching lower bound \citep{domingues2021episodic}.

\section{Appendix: proof of auxiliary lemmas in Section~\ref{sec:sharp-analysis-thm:sample-compl-MB-generative-inf}}

\subsection{Proof of Lemma~\ref{LemAbsToNormalV}}
\label{sec:proof-LemAbsToNormalV}

To prove this lemma, it suffices to verify that $\Vstar$ satisfies the Bellman optimality equation w.r.t.~$\MDP_{s,u^{\star}}$. For notational simplicity, let $\abPbm$ and $\rewabs$ denote its probability transition kernel and reward function, respectively.

At the absorbing state $s$, every action $a\in\cA$ satisfies
\begin{align*}
\rewabs(s,a)+\gamma(\abPbm\Vstar)_{(s,a)}
=(1-\gamma)u^{\star}+\gamma V^{\star}(s)=V^{\star}(s),
\end{align*}
where the last identity follows from $u^{\star}=V^{\star}(s)$. Next, for any state $s'\neq s$, the transition kernel and reward function remain unchanged, so that
\begin{align*}
\max_{a\in\cA}\big\{\rewabs(s',a)+\gamma(\abPbm\Vstar)_{(s',a)}\big\}
&=\max_{a\in\cA}\big\{r(s',a)+\gamma(\bP\Vstar)_{(s',a)}\big\}\\
&=V^{\star}(s').
\end{align*}
Combining the above two cases shows that $\Vstar$ satisfies the Bellman optimality equation for $\MDP_{s,u^{\star}}$. This taken together with the uniqueness of the value-function fixed point completes the proof.

\subsection{Proof of Lemma~\ref{LemValueCover}}
\label{sec:proof-LemValueCover}

In view of Lemma~\ref{LemAbsToNormalV}, setting $\widehat{u}^{\star}=\Vhatstar(s)$ gives $\Vhatstar_{s,\widehat{u}^{\star}}=\Vhatstar$. By construction of the net \eqref{eqn:net-eps-LOO-inf}, there exists $u_0\in\net{\epsilon}$ such that $|\widehat{u}^{\star}-u_0|\leq\epsilon$. The two auxiliary MDPs have identical transition kernels, and their rewards differ by at most $(1-\gamma)|\widehat{u}^{\star}-u_0|$. The basic property~\eqref{eq:V-star-perturbabtion-r-inf} therefore gives
\begin{align*}
\big\|\Vhatstar-\Vhatstar_{s,u_0}\big\|_{\infty}
=\big\|\Vhatstar_{s,\widehat{u}^{\star}}-\Vhatstar_{s,u_0}\big\|_{\infty}
\leq|\widehat{u}^{\star}-u_0|\leq\epsilon.
\end{align*}

\subsection{Proof of Lemma~\ref{lem:empirical-value-residual}}
\label{sec:proof:lem:empirical-value-residual}

It is first seen from the Bellman equation $\Vhatstar=r_{\pihatstar}+\gamma\widehat{P}_{\pihatstar}\Vhatstar$ that
\begin{align*}
\gamma P_{\pihatstar}\Vhatstar
&=\Vhatstar-r_{\pihatstar}
-\gamma(\widehat{P}_{\pihatstar}-P_{\pihatstar})\Vhatstar,\\
0\leq\Vhatstar-r_{\pihatstar}
&=\gamma\widehat{P}_{\pihatstar}\Vhatstar
\leq\frac{\gamma}{1-\gamma}1_S.
\end{align*}
Using the first identity to expand the variance (cf.~\eqref{eq:matrix-VarPpi-V-expression}), we obtain
\begin{align}
\gamma^2\mathsf{Var}_{P_{\pihatstar}}\big[\Vhatstar\big]
&=(\gamma^2P_{\pihatstar}-I)(\Vhatstar\circ\Vhatstar)
+2\Vhatstar\circ r_{\pihatstar}-r_{\pihatstar}\circ r_{\pihatstar}\notag\\
&\qquad+2\gamma(\Vhatstar-r_{\pihatstar})\circ
\big[(\widehat{P}_{\pihatstar}-P_{\pihatstar})\Vhatstar\big] \arxiv{\notag\\}
\arxiv{&\qquad} -\gamma^2\big[(\widehat{P}_{\pihatstar}-P_{\pihatstar})\Vhatstar\big]
\circ\big[(\widehat{P}_{\pihatstar}-P_{\pihatstar})\Vhatstar\big]\notag\\
&\leq(\gamma P_{\pihatstar}-I)(\Vhatstar\circ\Vhatstar)
+\frac{2}{1-\gamma}1_S \arxiv{ \notag\\}
\arxiv{&\quad} +\frac{2\gamma^2}{1-\gamma}
\big|(\widehat{P}_{\pihatstar}-P_{\pihatstar})\Vhatstar\big|.
\label{eq:empirical-value-variance-identity}
\end{align}
Here, the last inequality discards the two entrywise square terms and makes use of $0\leq r_{\pihatstar}\leq1_S$, $0\leq\Vhatstar\leq\frac{1}{1-\gamma}1_S$, the preceding bound on $\Vhatstar-r_{\pihatstar}$, and $\gamma^2\leq\gamma$.

The usefulness of this expansion becomes apparent after multiplying by $(I-\gamma P_{\pihatstar})^{-1}$, which together with $(I-\gamma P_{\pihatstar})^{-1}1_S=\frac{1}{1-\gamma}1_S$ gives
\begin{align*}
\arxiv{&}\gamma^2(I-\gamma P_{\pihatstar})^{-1}\mathsf{Var}_{P_{\pihatstar}}\big[\Vhatstar\big] \arxiv{\\}
\arxiv{&\quad} &\leq - \Vhatstar\circ\Vhatstar + \frac{2}{(1-\gamma)^2}1_S
+\frac{2\gamma^2}{1-\gamma}(I-\gamma P_{\pihatstar})^{-1}
\big|(\widehat{P}_{\pihatstar}-P_{\pihatstar})\Vhatstar\big| \notag\\
\arxiv{\\}
\arxiv{&\quad} &\leq  \frac{2}{(1-\gamma)^2}1_S
+\frac{2\gamma^2}{1-\gamma}(I-\gamma P_{\pihatstar})^{-1}
\big|(\widehat{P}_{\pihatstar}-P_{\pihatstar})\Vhatstar\big|.
\end{align*}
We can now apply the same argument as in \eqref{eq:Jensen-I-gamma-P-Var} to move the inverse matrix inside the square root. Applying Jensen's inequality, and taking the $\ell_{\infty}$ norm and squaring, we arrive at
\begin{align}
\arxiv{&} \gamma^2\Big\|(I-\gamma P_{\pihatstar})^{-1}
\sqrt{\mathsf{Var}_{P_{\pihatstar}}\big[\Vhatstar\big]}\Big\|_{\infty}^2 \arxiv{\notag\\}
&\arxiv{\quad}\leq\frac{\gamma^2}{1-\gamma}\Big\|(I-\gamma P_{\pihatstar})^{-1}
\mathsf{Var}_{P_{\pihatstar}}\big[\Vhatstar\big]\Big\|_{\infty}\notag\\
&\arxiv{\quad}\leq\frac{2}{(1-\gamma)^3}
+\frac{2\gamma}{(1-\gamma)^2}\Big\|(I-\gamma P_{\pihatstar})^{-1}
\big|(\widehat{P}_{\pihatstar}-P_{\pihatstar})\Vhatstar\big|\Big\|_{\infty},
\label{eq:empirical-value-variance-bound}
\end{align}
where the last line also uses $\gamma\leq1$. 

Notice that the final term in~\eqref{eq:empirical-value-variance-bound} is proportional to the same nonnegative quantity that upper bounds the value estimation error in \eqref{eq:empirical-value-residual-bound}. 
Substituting \eqref{eq:empirical-value-variance-bound} into \eqref{eq:empirical-value-residual-bound} and using $\sqrt{x+y}\leq\sqrt{x}+\sqrt{y}$ for $x,y\geq0$, we reach
\begin{align*}
&\gamma\Big\|(I-\gamma P_{\pihatstar})^{-1}
\big|(\widehat{P}_{\pihatstar}-P_{\pihatstar})\Vhatstar\big|\Big\|_{\infty}\\
&\quad\leq\sqrt{\frac{2\beta_1}{N(1-\gamma)^3}}
+\frac{2\beta_1}{N(1-\gamma)^2} \arxiv{\\}
\arxiv{&\qquad\quad}+\sqrt{\frac{2\gamma\beta_1}{N(1-\gamma)^2}}
\Big\|(I-\gamma P_{\pihatstar})^{-1}
\big|(\widehat{P}_{\pihatstar}-P_{\pihatstar})\Vhatstar\big|\Big\|_{\infty}^{1/2}\\
&\quad\leq\sqrt{\frac{2\beta_1}{N(1-\gamma)^3}}
+\frac{3\beta_1}{N(1-\gamma)^2}+\frac{\gamma}{2}\Big\|(I-\gamma P_{\pihatstar})^{-1}
\big|(\widehat{P}_{\pihatstar}-P_{\pihatstar})\Vhatstar\big|\Big\|_{\infty},
\end{align*}
where the last step follows from the AM-GM inequality. Solving the above inequality yields
\[
\gamma\Big\|(I-\gamma P_{\pihatstar})^{-1}
\big|(\widehat{P}_{\pihatstar}-P_{\pihatstar})\Vhatstar\big|\Big\|_{\infty} \leq 2\sqrt{\frac{2\beta_1}{N(1-\gamma)^3}}+\frac{6\beta_1}{N(1-\gamma)^2}.
\]

 \section{Notes}

 \paragraph{Sample complexity under a generative model.} 
The generative model (or simulator) adopted in this chapter was first introduced by \citet{kearns1999finite}. Since then, it has  become a standard framework in the theoretical studies of RL
and has been extensively investigated in subsequent work aimed at understanding and comparing the data efficiency of RL algorithms  (e.g.,  \citet{kearns2002sparse,kakade2003sample,lattimore2012pac,azar2013minimax,sidford2018near,sidford2018variance,wang2017randomized,wainwright2019variance,pananjady2019value,agarwal2019optimality,yang2019sample,mou2020linear,wang2021sample,khamaru2021instance,sidford2020solving,li2022minimax,vaswani2022near,shi2024curious,zhang2025estimation}).

%\paragraph{Model-based RL.}
Several prior works have investigated the effectiveness of model-based RL for discounted infinite-horizon MDPs under a generative model. The seminal work \citet{azar2013minimax} was the first to establish the asymptotic minimax optimality of model-based RL, albeit under the requirement that the total sample size be at least on the order of $\frac{S^2A}{(1-\gamma)^2}$. Subsequently,   \citet{agarwal2019optimality} significantly broadened the regime in which model-based methods are provably optimal.
 However, the results therein restrict the accuracy level to the range $\varepsilon \in \big(0, \frac{1}{\sqrt{1-\gamma}}\big]$, and therefore do not cover the full $\varepsilon$-regime of interest; in other words, the optimality guarantees in \citet{agarwal2019optimality}  hold only when the total sample size surpasses the order of $\frac{SA}{(1-\gamma)^2}$. Similar technical barriers had already appeared in the simpler setting of model-based policy evaluation \citep{pananjady2019value, agarwal2019optimality}, and had also arisen in the analysis of model-free algorithms \citep{sidford2018variance,wainwright2019variance} and in the studies of finite-horizon nonstationary MDPs \citep{yin2021near}. In stark contrast, however, existing lower bounds (e.g., \citet{azar2013minimax}) do not rule out the possibility of achieving nontrivial statistical accuracy below such sample size thresholds, giving rise to a gap between upper and lower bounds in the most sample-starved regime.

The works of \citet{li2020breaking,li2024breaking} established the first full-range minimax-optimality guarantees for model-based policy-learning algorithms under a generative model, covering both discounted infinite-horizon and finite-horizon nonstationary MDPs. Their analyses consider a randomly perturbed variant of the model-based algorithm, in which the reward function is modified by an additional perturbation step. These results close the gap between achievable upper bounds and minimax lower bounds, thereby pinning down the fundamental sample complexity limit across the entire \(\varepsilon\)-range. More recently, \citet{zurek2024plug} showed, through a higher-order recursive analysis, that the reward-perturbation step is unnecessary; their result retains the original, unperturbed model-based algorithm. The analysis presented in this chapter builds upon the main proof ideas of \citet{li2020breaking,li2024breaking}, while employing a different error decomposition and a new self-bounding argument to eliminate the need for reward perturbation.

\paragraph{Leave-one-out analyses.} Finally, we remark that the construction of state-absorbing MDPs or state-action-absorbing MDPs falls under the category of ``leave-one-out'' type analysis, which is particularly effective in decoupling complicated statistical dependency in various high-dimensional statistical problems \citep{chen2021spectral}, such as high-dimensional M-estimation \citep{el2013robust,el2015impact,javanmard2018debiasing}, phase retrieval \citep{chen2019gradient,ma2020implicit}, 
synchronization \citep{zhong2017near,abbe2020entrywise,ling2022improved}, ranking from pairwise comparisons~\citep{chen2019spectral,chen2022partial,fan2024uncertainty,fan2025ranking,fan2026spectral}, matrix completion and inference \citep{ma2020implicit,Chen22931,chen2020noisy},  tensor completion \citep{cai2022nonconvex}, 
network and clustering analysis \citep{abbe2020entrywise, bhattacharya2026inferences,zhang2024leave,agterberg2025estimating,ke2025optimal}. The first application of this powerful analysis framework to MDPs should be attributed to \citet{agarwal2019optimality}, which has since been applied and refined by \citet{pananjady2019value,li2020breaking,li2024breaking,li2024settling,yan2022model,shi2024curious} to establish the optimality of model-based RL in various settings.

%% file: chapters/generative_model_model_free.tex
\chapter{RL with a generative model: the model-free approach}
\label{chapter:generative-model-MF}

Thus far, we have established the statistical optimality of model-based RL algorithms under a generative model. Another widely studied, yet fundamentally different, paradigm is the model-free approach (see Figure~\ref{fig:model_based_free}), 
    which avoids explicit model estimation and instead seeks to learn the desired value function and/or policy directly from data. In this chapter, we discuss some of the canonical model-free algorithms, once again assuming access to a generative model.

\section{Prelude: stochastic approximation}

The algorithmic and theoretical foundations of many model-free algorithms --- such as temporal difference learning and Q-learning --- are deeply rooted in the stochastic approximation framework, originally developed by \citet{robbins1951stochastic}. In this section, we briefly review the key ideas underlying stochastic approximation.

Consider the problem of finding a fixed point of a given mapping $F: \mathbb{R}^n \rightarrow \mathbb{R}^n$, that is, 
\begin{equation}
    \text{find }x\in\mathbb{R}^n \qquad \text{ such that }F(x)=x.
    \label{eq:goal-stoc-approx}
\end{equation}
In many practical settings, however, the mapping $F(\cdot)$ is not directly accessible: one cannot evaluate 
$F(z)$ exactly at an arbitrary point $z\in \mathbb{R}^n$. Instead, for any given query point $z\in \mathbb{R}^n$, one is permitted to draw a random sample $f(z; \zeta)$ --- with $\zeta$ capturing the underlying randomness --- that provides an unbiased, noisy estimate of $ F(z)$.

To address the challenge arising from the lack of direct access to $F(\cdot)$,  \citet{robbins1951stochastic} proposed the following iterative scheme for solving \eqref{eq:goal-stoc-approx} based on successive draws of random samples: 
\begin{align}
\label{eq:general-stoc-approx}
    x_t = x_{t-1} + \eta_t  \big(f(x_{t-1}; \zeta_t)-x_{t-1}\big),
    \qquad t= 1,2,\cdots
\end{align}
where $x_0\in \mathbb{R}^n$ indicates an initial guess, $0<\eta_t<1$ denotes the learning rate at iteration $t$, and $f(x_{t-1}; \zeta_t)$ is a fresh sample satisfying 
\begin{equation*}
    \mathbb{E}[f(x_{t-1}; \zeta_t) \mid x_{t-1}]=F(x_{t-1}).
\end{equation*}
Under suitable conditions on the mapping $F(\cdot)$ (e.g., monotonicity and quasi-contractivity w.r.t.~appropriate cones), the data generating process (e.g., bounded noise variance), and the learning rate sequence $\{\eta_t\}_{t\geq 0}$ (e.g., linearly or polynomial decaying stepsizes), this remarkably simple procedure \eqref{eq:general-stoc-approx} is guaranteed to converge to a solution of \eqref{eq:goal-stoc-approx} \citep{lai2003stochastic,borkar2008stochastic}. See, for example, \citet[Section~2.2]{wainwright2019stochastic} for a non-asymptotic treatment. 

The stochastic approximation paradigm has had far-reaching impacts across statistics, optimization, machine learning, and many other fields. A particularly influential descendant is {\em stochastic gradient descent}, now a cornerstone of modern machine learning \citep{bottou2010large,bottou2018optimization}.

\section{Temporal difference learning}
\label{sec:model-free-policy-eval-sim}

Equipped with the stochastic approximation framework, we are now ready to introduce several canonical model-free RL algorithms. 
As in Chapter~\ref{chapter:generative-model}, we begin with the simpler task of policy evaluation, previously introduced in Section~\ref{sec:model-based-policy-eval-sim}. Given a policy $\pi$ and a target precision level $\varepsilon>0$, the objective is to compute an estimate of the corresponding value function $V^{\pi}$, denoted by $\widehat{V}^{\pi} \in \mathbb{R}^{S}$, such that
\begin{equation*}
    \big\| \widehat{V}^{\pi} - V^{\pi} \big\|_{\infty} \leq \varepsilon .
\end{equation*}
In sharp contrast to the model-based policy evaluation procedure \eqref{eq:value-estimation-MB-inf-vhat}, which solves the Bellman equation using an explicit estimate of the probability transition kernel $\widehat{P}_{\pi}$,  we now consider an alternative approach that entirely avoids estimating $\widehat{P}_{\pi}$. 

% \ref{sec:model-based-policy-eval-sim}

\subsection{Algorithm}
\label{sec:model-free-algorithm}

Arguably, the most well-known model-free policy evaluation algorithm is temporal-difference (TD) learning, originally formalized by \citet{sutton1988learning} and extensively analyzed in the literature (e.g., \citet{tesauro1995temporal,tsitsiklis1997analysis,bhandari2021finite,srikant2019finite,li2024high}).  
The core idea of TD learning is to apply stochastic approximation to solve the Bellman equation (or equivalently, to find a fixed point of the Bellman operator).

More concretely, recall that the value function of a policy $\pi$ is characterized by the Bellman consistency equation \eqref{eq:bellman_consistency-V-all}. 
Consequently, computing $V^{\pi}$ amounts to finding a fixed point of the mapping  
\begin{equation*}
    F(V) = \mathcal{T}_{\pi}(V) \coloneqq r_{\pi} + \gamma P_{\pi }V. 
\end{equation*}
This reformulation situates policy evaluation within the stochastic approximation framework. In particular, it naturally suggests an iterative scheme: maintain an estimate $V_t: \mathcal{S} \rightarrow \mathbb{R}$ of the value function in each iteration $t$,
and update it according to
\begin{subequations}
\label{eq:TD-general-sampling}
\begin{align}
    V_t &= V_{t-1} + \eta_t \big(   \mathcal{T}_{\pi, t}(V_{t-1}) -V_{t-1}  \big) ,
    %\notag\\
    %&= (1-\eta_t) V_{t-1} + \eta_t \mathcal{T}_{\pi, t}(V_{t-1})  , 
    \qquad t=1,2,\cdots
\end{align}
provided that, at each iteration, one can construct an unbiased estimate $\mathcal{T}_{\pi,t}(V_{t-1})$ of the Bellman operator applied to the current iterate, satisfying
\begin{equation}
    \mathbb{E}\big[ \mathcal{T}_{\pi,t}(V_{t-1}) \mid V_{t-1} \big] = \mathcal{T}_{\pi}(V_{t-1}). 
\end{equation}
\end{subequations}

When a generative model is available, this scheme \eqref{eq:TD-general-sampling} admits a particularly simple implementation:  at each iteration $t$,  
\begin{itemize}
    \item for each $s\in \mathcal{S}$, 
    draw an independent next-state sample $s_t(s)\sim P_{\pi}(\cdot \mid s)$ (cf.~\eqref{eq:defn-P-s-pi-inf}) by querying the generative model; 

    \item update the value estimate $V_t$ for each state $s\in\mathcal{S}$ according to
    \begin{subequations}
    \begin{align}
	V_t (s) &= V_{t-1} (s) + \eta_t \big\{ r_{\pi}(s) + \gamma   V_{t-1}(s_t)-V_{t-1} (s) \big\} \label{eqn:td-learning-0}\\
    &= (1- \eta_t ) V_{t-1} (s) + \eta_t \big( r_{\pi}(s) + \gamma   V_{t-1}(s_t) \big),   
    \label{eqn:td-learning}
	%\qquad s_t \equiv s_t(s)  
\end{align}
\end{subequations}
with $s_t$ taken to be $s_t(s)$. 
\end{itemize}
A straightforward calculation shows that this update is unbiased, i.e., 
\[
\mathbb{E}[V_t \mid V_{t-1}] =  (1- \eta_t ) V_{t-1}  + \eta_t \big( r_{\pi} + \gamma  P_{\pi} V_{t-1} \big),
\]
so that \eqref{eqn:td-learning} indeed implements a stochastic approximation procedure for solving the Bellman equation. 
 Notably, the term 
 \[
    r_{\pi}(s) + \gamma   V_{t-1}(s_t) - V_{t-1}(s)
 \]
in \eqref{eqn:td-learning-0} is often referred to as the ``temporal difference'', as it reflects the discrepancy between the current value estimate and a one-step look-ahead estimate obtained by bootstrapping one step forward in the MDP. The complete procedure is summarized in Algorithm~\ref{alg:td-infinite}. This variant of TD learning is commonly termed {\em synchronous} TD learning, since each iteration updates all coordinates of the value function estimate simultaneously.

In contrast to the model-based approach in Algorithm~\ref{alg:policy-eval-model-based}, TD learning maintains only $S$-dimensional value iterates and does not require storing an explicit $S^2$-dimensional estimate of the transition kernel induced by $\pi$. This substantially reduces memory requirements, highlighting a central appeal of model-free algorithms.

 \begin{algorithm}[t]
    \DontPrintSemicolon
    \textbf{input:}  reward function $r_{\pi}$ induced by policy $\pi$, discount factor $\gamma$, learning rates $\{\eta_t\}$, number of iterations $T$, initialization $V_0$. \\
   \For{$t=1,2,\cdots,T$} {
	%\STATE{ 
    %\For{$s\in \mathcal{S}$} {
    draw a sample $s_t(s) \overset{\mathrm{ind.}}{\sim}  P_{\pi}(\cdot \mymid s)$ (cf.~\eqref{eq:defn-P-s-pi-inf}) for every $s\in \cS$.  \\
    {\color{blue}\tcc{stochastic approximation.}}
   for every $s\in \cS$, update $V_t(s)$ as
    \begin{align*}
	V_t (s) &= (1- \eta_t ) V_{t-1} (s) + \eta_t \big\{ r_{\pi}(s) + \gamma   V_{t-1}\big(s_t(s)\big) \big\}.  
\end{align*}
    }
%	\textbf{output} $V_T$.
  %  }
    \caption{TD learning with a generative model}
 \label{alg:td-infinite}
\end{algorithm}

\subsection{Theoretical guarantees}
\label{sec:TD-learning-theory}

%\begin{subequations}
%\label{eq:thm-td}

We now analyze the efficacy of TD learning through the lens of sample complexity. To this end, we need to specify the learning rate schedule $\{\eta_t\}_{t\geq 1}$, as it plays a crucial role in determining the algorithm's performance. 
In this section, we choose the learning rates to satisfy 
\begin{equation}
\frac{1}{1+\frac{c_1 (1-\gamma) T}{\log^2 T}}\le\eta_{t}\le\frac{1}{1+\frac{c_2 (1-\gamma) }{\log^2 T} t},
\qquad 1\leq t\leq T,\label{eq:thm:eta-TD}
\end{equation}
where $c_{1}\geq c_{2}>0$ are some universal constants. Clearly, Condition~\eqref{eq:thm:eta-TD} encompasses two commonly used learning-rate choices as special cases:
\begin{itemize}
    \item[(i)]  {\em constant learning rates}, where $\eta_t\equiv \frac{1}{1+\frac{c_1 (1-\gamma) T}{\log^2 T}}$ for all $t\leq T$; 
    
    \item[(ii)] {\em rescaled linear learning rates}, where 
    $\eta_{t}=\frac{1}{1+\frac{c_2 (1-\gamma) }{\log^2 T}t}$ for all $t\leq T$. 
\end{itemize}

The following theorem characterizes the $\ell_{\infty}$-based sample complexity of TD learning under a generative model, which sharpens the sample complexity rate  first established by \citet{li2024q}. It is worth noting that the guarantees in
\citet{li2024q} were established only for $\varepsilon\in(0,1]$.
In this chapter, we develop a refined analysis that extends the range
of $\varepsilon$ to the full range $\varepsilon\in\big(0,\frac{1}{1-\gamma}\big]$.

\begin{theorem}[Extension of~\citet{li2024q}]
\label{thm:policy-evaluation-TD}
	Consider any $\delta\in(0,1)$, $\varepsilon\in\big(0,\frac{1}{1-\gamma}\big]$, and $\gamma\in [1/2,1)$. Suppose that the learning rates satisfy \eqref{eq:thm:eta-TD} for some small enough universal constants $c_{1}\geq c_{2}>0$, and that the total number of iterations $T\geq2$ obeys
\begin{equation}
	T\ge\frac{c_{3}\big(\log^{3}T\big)\big(\log\frac{ST}{\delta}\big)}{(1-\gamma)^{3}\varepsilon^{2}}
	\label{eq:thm:sample-size-TD-thm}
\end{equation}
for some sufficiently large universal constant $c_{3}>0$. 
%\end{subequations}
%
If $0\leq {V}_{0}(s) \leq\frac{1}{1-\gamma}$ for all $s\in \cS$, 
then  with probability at least $1-\delta$, TD learning (see Algorithm~\ref{alg:td-infinite}) achieves
\begin{align*}
	\forall s\in \cS:\qquad  \big| {V}_{T}(s) -  {V}^{\pi}(s) \big| \le\varepsilon.
	%\label{eq:VT-entrywise-TD-thm}
\end{align*}
\end{theorem}
\begin{remark}[Mean estimation guarantee]
In view of the trivial upper bound $\big| {V}_{T}(s) -  {V}^{\pi}(s) \big| \le \frac{1}{1-\gamma}$, the high-probability result in Theorem~\ref{thm:policy-evaluation-TD} immediately implies an error bound in expectation. Specifically, choosing $\delta = \varepsilon(1-\gamma)/2\in(0,1)$ in Theorem~\ref{thm:policy-evaluation-TD} and decomposing according to the high-probability event, we obtain
\begin{equation*}
    \mathbb{E}\Big[ \max_{s\in \cS}\big| V_{T}(s) -  V^{\pi}(s)\big| \Big] \leq \varepsilon (1-\delta)+ \frac{1}{1-\gamma} \delta \leq 2\varepsilon, 
\end{equation*}
as long as the total number of iterations exceeds
\begin{equation*}
T\ge\frac{c_{3}\big(\log^{3}T\big)\big(\log\frac{2ST}{\varepsilon (1-\gamma)}\big)}{(1-\gamma)^{3}\varepsilon^{2}}.
\end{equation*}
\end{remark}

Given that each iteration of TD learning draws $S$ fresh samples (one per state),  the total number of samples used by Algorithm~\ref{alg:td-infinite} is $ST$. Consequently, 
Theorem~\ref{thm:policy-evaluation-TD} immediately implies that, to yield an accuracy level $\varepsilon \in \big(0,\frac{1}{1-\gamma}\big]$, TD learning needs no more than
\begin{equation*}
	%\label{eq:TD-sample-complexity-simple}
	\widetilde{O} \bigg( \frac{S}{(1-\gamma)^3 \varepsilon^2} \bigg)
\end{equation*}
samples, and moreover, the {\em last iterate} of this iterative procedure already attains the desired accuracy.  This non-asymptotic result is valid provided 
the learning rates are properly chosen --- either as a suitable constant or according to a rescaled linear schedule (see~\eqref{eq:thm:eta-TD}). In comparison with the minimax lower bound presented in Theorem~\ref{theorem:azar} (by setting $A=1$ therein), Theorem~\ref{thm:policy-evaluation-TD} demonstrates that plain TD learning --- without any additional algorithmic trick like iterate averaging or variance reduction --- attains minimax-optimal sample complexity (up to logarithmic factors) when one draws samples from the generative model, for the full accuracy range $\varepsilon\in\big(0,\frac{1}{1-\gamma}\big]$.

% \begin{remark} 	It is also noteworthy that while the last iterate of plain TD learning is shown to be minimax optimal (which concerns worst-case optimality), it may not necessarily enjoy local optimality.  	As discussed by \citet{khamaru2021instance}, 	additional algorithmic tricks like variance reduction may sometimes be needed to ensure local optimality. 
%\end{remark}

\subsection{Analysis: a first attempt} \label{sec:TD-analysis-first}

As a starting point, we present an initial --- albeit suboptimal --- proof strategy that captures some of the core ideas for analyzing the performance of TD learning; this approach will be refined shortly in Section~\ref{sec:TD-analysis}.

Let us adopt the vector and matrix notation introduced in Section~\ref{sec:vector-matrix-notation}, viewing $V_t, V^{\pi}$ and $r_{\pi}$ as vectors in 
$\mathbb{R}^S$, among others. 
By defining an empirical transition matrix $P_t\in \{0,1\}^{S\times S}$ such that
\begin{align}
\label{eq:defn-Pt-TD-proof-generative}
\forall (s,s')\in \cS,\qquad
	P_t(s,s') \coloneqq \begin{cases} 1, \quad &\text{if }s'=s_t(s), \\ 0, & \text{otherwise}, \end{cases} 
\end{align}
where we recall that $s_t(s)\overset{\mathrm{ind.}}{\sim} P_{\pi}(\cdot \mid s)$, we can express the TD learning update rule \eqref{eqn:td-learning} compactly in vector-matrix form as follows:
\begin{align} 
\label{eq:iteration-rule-td}
	V_t = (1-\eta_t) V_{t-1} + \eta_t( r_{\pi} + \gamma P_t V_{t-1}), 
	\qquad  t\geq 1.
\end{align}

\paragraph{Preliminary facts.} 
Before proceeding, we gather a few elementary facts that will be useful throughout the analysis. The following lemma asserts that the TD iterates always fall within the admissible range of value functions. The proof, based on induction, is deferred to Section~\ref{appendix:proof-lemma-TV-bounded}. 
\begin{lemma} \label{lemma:TD-bounded}
If $0\leq V_0(s)\leq \frac{1}{1-\gamma}$ for all $s\in \mathcal{S}$, then
the TD learning iterates \eqref{eq:iteration-rule-td} obey 
\[
0\le V_{t}\le\frac{1}{1-\gamma} 1_S \quad \text{and} \quad \| V_{t}- V^{\pi}\|_{\infty}\le\frac{1}{1-\gamma} 
\]
for all $0 \leq t\leq T$, 
provided that $0\leq \eta_t \leq 1$ for all $0\leq t\leq T$. 
\end{lemma}

Next, given the iterative nature of TD learning, it is convenient to introduce the following quantities that, as we shall see momentarily, capture the cumulative effect of the learning rates $\{\eta_t\}$:
\begin{equation}
\eta_{k}^{(t)}\coloneqq\begin{cases}
\prod_{i=1}^{t}\big(1-\eta_{i}(1-\gamma) \big), & \text{if }k=0,\\
\eta_{k}\prod_{i=k+1}^{t} \big( 1-\eta_{i}(1-\gamma) \big), & \text{if }0<k<t,\\
\eta_{t}, & \text{if }k=t. 
\end{cases}\label{def:eta-k-t}
\end{equation}
Some basic properties of $\eta_k^{(t)}$ are stated below. In particular, $\eta_k^{(t)}$ can be very small when $k$ is small. The proof is elementary and deferred to Section~\ref{appendix:proof-lemma-TD-step-size}.
\begin{lemma} \label{lemma:TD-step-size}
Recall that the learning rates are chosen to satisfy \eqref{eq:thm:eta-TD}, and 
suppose $c_1 c_2 \leq 1/8$. Then for any $t \geq T / (c_2 \log T)$, one has
\[
\eta_{k}^{(t)} \leq \begin{cases}
    \frac{1}{T^2} , & \text{if } 0 \leq k \leq t/2, \\
  \le\eta_{k}\leq  \frac{2\log^3 T}{(1-\gamma)T} , & \text{if }  t/2 < k \leq t.
\end{cases}
\]
\end{lemma}

\paragraph{Step 1: decomposing the error $V_t - V^{\pi}$. }

At a high level, this step unrolls the TD recursion and decompose the
estimation error into an initialization term and a martingale term. For notational convenience, define the error sequence
\begin{equation}
\Delta_t \coloneqq V_t - V^{\pi}.
\label{eq:defn-Delta-t-TD}
\end{equation}
In view of the TD learning update rule \eqref{eq:iteration-rule-td}, we have the following decomposition:
\begin{align}
\Delta_t 
&= (1-\eta_t) (V_{t-1} - V^\pi) + \eta_t( r_{\pi} + \gamma P_t V_{t-1} - V^\pi) \notag\\
&= (1-\eta_t) \Delta_{t-1} + \eta_t \gamma ( P_t V_{t-1} - P_{\pi} V^{\pi}) \notag\\
&= (1-\eta_t) \Delta_{t-1} + \eta_t \gamma \big( P_{\pi} \Delta_{t-1} + ( P_t -  P_{\pi}) V_{t-1} \big) \label{eq:Delta-t-recursive-decomp-TD-2}\\
&= \big( I-\eta_t( I-\gamma P_{\pi})\big) \Delta_{t-1} + \eta_t\gamma  ( P_t -  P_{\pi}) V_{t-1} ,
\label{eq:Delta-t-recursive-decomp-TD}
\end{align}
where $P_{\pi}$ is defined in \eqref{eqn:ppivq}, and the second line arises from the Bellman equation $ V^{\pi} =  r_{\pi} + \gamma  P_{\pi} V^{\pi}$ (cf.~\eqref{eq:Vpi-Bellman-inf-matrix}). 
Iterating the recursion \eqref{eq:Delta-t-recursive-decomp-TD} yields
\begin{align}
 \Delta_t 
 & = \prod_{i = 1}^t \big( I-\eta_i( I-\gamma P_{\pi})\big) \Delta_0 + \sum_{k = 1}^t Z_k, \label{eq:defn-xit-TD}
\end{align}
where the process $\{Z_k:1\leq k \leq t\}$ is defined as
\begin{equation}
Z_k \coloneqq \gamma\eta_k \Bigg( \prod_{i = k+1}^t \big( I-\eta_i( I-\gamma P_{\pi})\big)\Bigg) ( P_k -  P_{\pi}) V_{k-1}. \label{eq:TD-Zk-defn}
\end{equation}
In words, the error at iteration $t$ decomposes into a suitably contracted version of the initial error and a martingale sum, since each $Z_k$ has zero conditional mean given the history up to iteration $k-1$ and hence $\{Z_k\}$ forms a martingale difference sequence (this is an immediate consequence of the definition \eqref{eq:defn-Pt-TD-proof-generative} of $P_k$). In what follows, we shall look at the two terms on the right-hand side of \eqref{eq:defn-xit-TD} separately.

\paragraph{Step 2: bounding the first term in \eqref{eq:defn-xit-TD}.}
We first show that the first term on the right-hand side of \eqref{eq:defn-xit-TD} is negligible. To this end, it is observed that
\begin{align}
\bigg\Vert \prod_{i=1}^{t}\big( I-\eta_{i}( I-\gamma P_{\pi})\big)\Delta_{0}\bigg\Vert _{\infty} 
& \le \bigg\Vert \prod_{i=1}^{t}\big( I-\eta_{i}( I-\gamma P_{\pi})\big)\bigg\Vert _{1}\left\Vert \Delta_{0}\right\Vert _{\infty} \notag\\
 & \overset{\text{(i)}}{=}\Bigg\{ \prod_{i=1}^{t}\big(1-\eta_{i}(1-\gamma)\big)\Bigg\} \left\Vert \Delta_{0}\right\Vert _{\infty} \notag\\
 & = \eta_{0}^{(t)}\left\Vert \Delta_{0}\right\Vert _{\infty}\overset{\text{(ii)}}{\leq}\frac{1}{(1-\gamma)T^{2}}.\label{eq:TD-first-term} 
\end{align}
Here, (i) invokes the basic fact that,  for any $0\leq k < t$,
\begin{equation} \label{eq:TD-L1-norm}
    \bigg\Vert \prod_{i = k+1}^t \big( I-\eta_i( I-\gamma P_{\pi})\big)\bigg\Vert_1 = \prod_{i=k+1}^{t}\big(1-\eta_{i}(1-\gamma)\big),
\end{equation}
given that $P_{\pi}$ is a probability transition matrix; 
and (ii) results from Lemma~\ref{lemma:TD-step-size} and the assumption that $\|\Delta_0\|_{\infty}\leq \frac{1}{1-\gamma}$, provided that $t \ge T / (c_{2}\log T)$. As will be shown shortly, this upper bound \eqref{eq:TD-first-term} is negligible relative to the other term.

%adapted to the filtration $\mathcal{F}_k=\sigma(V_{0},V_{1},\ldots,V_{k})$, namely,
%\[
%\mathbb{E} [Z_k \mid \mathcal{F}_{k-1}] = 0, \quad \text{for } 1\leq k \leq t.
%\]
%

\paragraph{Step 3: bounding the martingale sum in \eqref{eq:defn-xit-TD}.} We now turn to the second term on the right-hand side of \eqref{eq:defn-xit-TD}. Recall that $\{Z_k\}$ forms a martingale difference sequence.  
In order to control the martingale sum $\sum_k Z_k$, a natural approach is to apply standard martingale concentration inequalities, such as the Azuma–Hoeffding inequality (cf.~Lemma~\ref{lem:azuma-hoeffding}). To this end, we first note that
\begin{align}
\|Z_k\|_{\infty} & = \gamma\bigg\|\eta_k\prod_{i = k+1}^t \big( I-\eta_i( I-\gamma P_{\pi})\big)( P_k -  P_{\pi}) V_{k-1} \bigg\|_{\infty} \nonumber\\
 & \leq \gamma\bigg\|\eta_k\prod_{i = k+1}^t \big( I-\eta_i( I-\gamma P_{\pi})\big) \bigg\|_{1}  \big\| ( P_k -  P_{\pi}) V_{k-1} \big\|_{\infty} \nonumber\\
	& \overset{\text{(i)}}{=} \gamma\Bigg\{ \eta_k\prod_{i = k+1}^t \big(1-\eta_i(1-\gamma)\big) \Bigg\} 
	\big\| ( P_k -  P_{\pi}) V_{k-1} \big\|_{\infty} \nonumber\\	
 & \leq \gamma\eta_{k}^{(t)}\big(\| P_{k}\|_{1}+\| P_{\pi}\|_{1}\big)\| V_{k-1}\|_{\infty} \notag\\
 &\overset{\text{(ii)}}{\leq} 
 \begin{cases}
 \frac{4\gamma\log^{3}T}{(1-\gamma)^2T}, & \text{if } k > t/2\\
   \frac{2\gamma}{(1-\gamma)T^2}, &\text{if } 0\leq k\leq t/2
 \end{cases}
 ~\eqqcolon B_{Z_k},
 \label{eq:TD_B_azuma}
\end{align}
provided that $t \geq T / (c_2 \log T)$. Here, 
(i) follows from \eqref{eq:TD-L1-norm}, 
whereas (ii) arises from the facts $\| P_{k}\|_{1}=\| P_{\pi}\|_{1}=1$ (recall the definition \eqref{eq:defn-Pt-TD-proof-generative} of $P_k$), as well as Lemmas~\ref{lemma:TD-bounded}~and~\ref{lemma:TD-step-size}.

Take $t=T$. Armed with the bound \eqref{eq:TD_B_azuma}, invoking the Azuma–Hoeffding inequality in Lemma~\ref{lem:azuma-hoeffding} and applying the union bound over all coordinates of $\sum_{k}Z_k$, we arrive at 
\begin{align}
    \bigg\| \sum_{k=1}^T Z_k \bigg \|_{\infty} &\leq 
    \bigg( 2 \sum_{k=1}^T B_{Z_k}^2  \log\bigg(\frac{2S}{\delta} \bigg) \bigg)^{1/2} \notag\\ 
    &\leq \gamma 
    \bigg( \frac{16\log^6T\log\big(\frac{2S}{\delta}\big)}{(1-\gamma)^4T} + \frac{4\log\big(\frac{2S}{\delta}\big)}{(1-\gamma)^2T^3}    \bigg)^{1/2} \notag\\
    &\leq \widetilde{O} \bigg( \frac{1}{(1-\gamma)^2\sqrt{T}}    \bigg)
    \label{eq:sum-Zk-Asuma-TD}
\end{align}
with probability greater than $1-\delta$. 

\paragraph{Step 4: putting all pieces together.} Substituting the above results \eqref{eq:TD-first-term} and \eqref{eq:sum-Zk-Asuma-TD} into \eqref{eq:defn-xit-TD}, we arrive at
\begin{align*}
\|\Delta_T\|_{\infty} 
\leq \widetilde{O} \bigg( \frac{1}{(1-\gamma)^2\sqrt{T}}    \bigg)
\end{align*}
with high probability. In other words, this result tells us that, in order to yield $\varepsilon$ accuracy (i.e., $\|\Delta_T\|_{\infty}\leq \varepsilon$), it suffices to take $T$ on the order of
\begin{equation*}
    \widetilde{O}\bigg( \frac{1}{(1-\gamma)^4\varepsilon^2} \bigg)
   \qquad \text{iterations.}
\end{equation*}

\paragraph{Discussion.} While the above proof is simple and fully non-asymptotic, it yields an iteration complexity that exceeds the bound in Theorem~\ref{thm:policy-evaluation-TD} by a multiplicative factor of $\frac{1}{1-\gamma}$, and therefore falls short of establishing the minimax optimality of synchronous TD learning. A closer inspection reveals that this suboptimality stems primarily from the use of the Azuma–Hoeffding inequality (cf.~Lemma~\ref{lem:azuma-hoeffding}), which fails to exploit the underlying variance structure --- an ingredient that is crucial for shaving the extra $\frac{1}{1-\gamma}$ factor. To sharpen the analysis, we would need to invoke the Freedman inequality (cf.~Lemma~\ref{lem:Freedman-prelim}) and leverage the intimate connection between the cumulative conditional covariance and the intrinsic variance structure of the discounted infinite-horizon MDP, as developed in the next subsection.

\subsection{Sharp analysis: proof of Theorem~\ref{thm:policy-evaluation-TD}} \label{sec:TD-analysis}

We now present a refined analysis of TD learning that, as noted above, leverages the Freedman inequality for martingales in conjunction with the intrinsic variance structure of the discounted infinite-horizon MDP.

\paragraph{Step 1: decomposing the error as in Section~\ref{sec:TD-analysis-first}.} 
As already shown in Section~\ref{sec:TD-analysis-first} (see~\eqref{eq:defn-xit-TD} and \eqref{eq:TD-first-term}), we can decompose and bound the estimation error $\Delta_t=V_t-V^{\pi}$ at iteration $t$ as
\begin{align}
\label{eq:defn-xit-TD-sharp}
 \| \Delta_t \|_{\infty}
 & \leq \frac{1}{(1-\gamma)T^2}
 %R_1 
 + \bigg\| \sum_{k = 1}^t Z_k \bigg\|_{\infty}, 
\end{align}
where $\{Z_k:1\leq k \leq t\}$ is defined in \eqref{eq:TD-Zk-defn}.  
It thus boils down to bounding the entrywise magnitude of the martingale sum $\sum_kZ_k$.

\paragraph{Step 2: bounding the martingale sum via Freedman's inequality.}
We intend to apply the Freedman inequality in Lemma~\ref{lem:Freedman-prelim} to control each coordinate of the sum $\sum_{k = 1}^t Z_k$. This requires examining the following two quantities appearing in the Freedman inequality:
\begin{equation}
 \max_{1\le k\leq t} \|Z_{k}\|_{\infty} \qquad \text{and} \qquad W_{t} \coloneqq\sum_{k=1}^{t}\mathsf{Var}_{k-1}(Z_{k} ),
\label{eq:defn-Wt-TD}
\end{equation}
where $\mathsf{Var}_{k-1}(Z_{k})$ represents an $S$-dimensional vector whose $j$-th entry is the variance of the $j$-th entry of $Z_k$ {\em conditioned} on what happens up to iteration $k-1$. Throughout this step, we assume $t\geq T/(c_2\log T)$. Note that a crude bound on the variance proxy is sufficient for now; a
sharper bound exploiting the variance structure of the TD iterates is
developed in the next two steps.
\begin{enumerate}
\item \textit{Bounding the maximum magnitude.} 
Recalling inequality \eqref{eq:TD_B_azuma}, one immediately sees that
\begin{align}
 \max_{k: 1\leq k\leq t} \|Z_k\|_{\infty}
\le \frac{4\gamma \log^{3}T}{(1-\gamma)^2T} \eqqcolon B.
\label{eq:TD_B}
\end{align}

\item \textit{Bounding the conditional entrywise variance of $Z_k$.} 
Next, we turn to $W_t$ by first examining each conditional variance term $\mathsf{Var}_{k-1}(Z_{k})$. 
As can be seen from the definition \eqref{eq:TD-Zk-defn} of $Z_k$, each coordinate of $Z_k$ is a bilinear form of $P_k-P_{\pi}$ (recall the definition of $P_k$ in \eqref{eq:defn-Pt-TD-proof-generative}).  To analyze such a bilinear form, we first make note of a useful elementary fact:
for any fixed vectors $u,v\in \mathbb{R}^S$ obeying $u\geq 0$, one has
\begin{align}
\mathsf{Var}\big(u^{\top}( P_{k}- P_{\pi}) v\big) 
& =\sum_{i=1}^{S}u_{i}^{2}\mathsf{Var}\big(( P_{k}- P_{\pi})_{i,\cdot} v\big) \arxiv{\nonumber \\
&} \leq \max_{i}|u_{i}| \cdot u^\top \mathsf{Var}_{P_{\pi}} (v)  \nonumber \\
& \leq \|u\|_1 u^{\top}\mathsf{Var}_{ P_{\pi}}( v),\label{eq:Var-u-Pk-v-bound}
\end{align}
where we remind the reader of the notation $\mathsf{Var}_{ P}( V)$ in \eqref{eq:matrix-VarPpi-V-expression}. 
Applying this relation to each entry of $Z_k$ defined in \eqref{eq:TD-Zk-defn} gives
\begin{align}
\mathsf{Var}_{k-1}(Z_{k}) 
& \leq\gamma^{2}\bigg\Vert \eta_{k}\prod_{i=k+1}^{t}\big( I-\eta_{i}( I-\gamma P_{\pi})\big)\bigg\Vert_{1} \arxiv{\notag \\}
\arxiv{& \qquad\quad \cdot}  \bigg\{ \eta_{k}\prod_{i=k+1}^{t}\big( I-\eta_{i}( I-\gamma P_{\pi})\big)\bigg\} \mathsf{Var}_{ P_{\pi}}\left( V_{k-1}\right) \notag\\
 & = \gamma^{2} \eta_{k}^{(t)} \eta_{k} \prod_{i=k+1}^{t}\big( I-\eta_{i}( I-\gamma P_{\pi})\big)\mathsf{Var}_{ P_{\pi}}\left( V_{k-1}\right),
 \label{eq:TD-var-Zk-upper-bound-each-term}
\end{align}
where the first inequality is a consequence of  \eqref{eq:Var-u-Pk-v-bound}, and the second relation results from \eqref{eq:TD-L1-norm} and the definition of $\eta_k^{(t)}$ (cf.~\eqref{def:eta-k-t}). 
Although the upper bound in \eqref{eq:TD-var-Zk-upper-bound-each-term} appears somewhat complicated, it admits a more tractable upper bound once summed as in the next step.
\end{enumerate}

To control $W_t$, we bound two partial sums of $\mathsf{Var}_{k-1}(Z_{k})$ separately, where the first is negligible and the second constitutes the dominant contribution.
\begin{itemize}
    \item Consider first those $k$'s obeying $ k \leq t/2$. It follows from \eqref{eq:TD-var-Zk-upper-bound-each-term} that
    \begin{align}
\sum_{k=1}^{\lfloor t/2\rfloor} \mathsf{Var}_{k-1}(Z_{k})  
        &\leq \gamma^{2}\sum_{k=1}^{\lfloor t/2\rfloor} \eta_{k}^{(t)}\bigg\Vert \eta_{k}\prod_{i=k+1}^{t}\big( I-\eta_{i}( I-\gamma P_{\pi})\big)\bigg\Vert _{1}\| V_{k-1}\|_{\infty}^{2} 1_S \notag \\
        & = \gamma^{2} \sum_{k=1}^{\lfloor t/2\rfloor} \big(\eta_{k}^{(t)}\big)^{2}\| V_{k-1}\|_{\infty}^{2}  1_S \leq \frac{\gamma^{2}}{2(1-\gamma)^{2}T^{3}}  1_S,
        \label{eq:partial-sum-Var-Zk-TD-small-k}
    \end{align}
    where the penultimate step uses \eqref{eq:TD-L1-norm} and \eqref{def:eta-k-t}, and the last relation follows from Lemmas~\ref{lemma:TD-bounded}-\ref{lemma:TD-step-size} and $t\leq T$. 
    As will be made clear shortly, this partial sum \eqref{eq:partial-sum-Var-Zk-TD-small-k} becomes negligible compared to the remaining part.

    \item Next, let us turn to those $k$'s obeying $t/2 < k\leq t$, which account for the dominant contribution to $W_t$. Invoke Lemma~\ref{lemma:TD-step-size} and \eqref{eq:TD-var-Zk-upper-bound-each-term} to first demonstrate that
    \begin{align}
        \mathsf{Var}_{k-1}(Z_{k}) 
        &\leq \frac{2\gamma^2\log^3 T}{(1-\gamma)T} \eta_{k}\prod_{i=k+1}^{t}\big( I-\eta_{i}( I-\gamma P_{\pi})\big)\mathsf{Var}_{ P_{\pi}}( V_{k-1}).
        \label{eq:var-k-Zk-ub-TD}
    \end{align}
    To obtain a  deterministic bound on $W_t$, Lemma~\ref{lemma:TD-bounded} gives 
    \begin{align*}
    \mathsf{Var}_{P_\pi}(V_{k-1})\leq\frac{1}{(1-\gamma)^2}1_S.
    \end{align*}
    In addition, one observes the following telescoping property for the matrix sum of interest:
    \begin{align}
    &\sum_{k=\lfloor t/2\rfloor+1}^{t}\eta_{k}\prod_{i=k+1}^{t}\big( I-\eta_{i}( I-\gamma P_{\pi})\big)  \notag\\
    & = ( I-\gamma P_{\pi})^{-1} \hspace{-1ex} \sum_{k=\lfloor t/2\rfloor+1}^{t}\bigg[\prod_{i=k+1}^{t} \big( I-\eta_{i}( I-\gamma P_{\pi})\big) \arxiv{\notag\\}
    \arxiv{&\hspace{16em} } - \prod_{i=k}^{t} \big( I-\eta_{i}( I-\gamma P_{\pi})\big) \bigg] \notag\\
    &= ( I-\gamma P_{\pi})^{-1} - ( I-\gamma  P_{\pi})^{-1}\prod_{i=\lfloor t/2\rfloor+1}^{t} \big( I-\eta_{i}( I-\gamma P_{\pi})\big) \notag\\
    &\le ( I-\gamma P_{\pi})^{-1}, \label{eq:telescoping_td}
    \end{align}
    where the inequality in the last line holds entrywise since $(I-\gamma P_\pi)^{-1}$ and $I-\eta_i(I-\gamma P_\pi)$ are both nonnegative matrices. Applying the preceding two bounds to \eqref{eq:var-k-Zk-ub-TD} and using $(I-\gamma P_\pi)^{-1}1_S=\frac{1}{1-\gamma}1_S$, we reach
    \begin{align}
        \sum_{k=\lfloor t/2\rfloor+1}^{t}\mathsf{Var}_{k-1}(Z_k)
        &\leq\frac{2\gamma^2\log^3T}{(1-\gamma)^3T}(I-\gamma P_\pi)^{-1}1_S\arxiv{\notag\\
&}=\frac{2\gamma^2\log^3T}{(1-\gamma)^4T}1_S.
        \label{eq:partial-sum-Var-Zk-TD-large-k}
    \end{align}
\end{itemize}

Combining \eqref{eq:partial-sum-Var-Zk-TD-small-k} and \eqref{eq:partial-sum-Var-Zk-TD-large-k}, and using $\gamma^2\leq1$, we obtain the deterministic bound
\begin{align}
\|W_{t}\|_{\infty} & \leq \frac{1}{2(1-\gamma)^{2}T^{3}} +\frac{2\log^{3}T}{(1-\gamma)^{4}T} 
\leq\frac{3\log^{3}T}{(1-\gamma)^{4}T} \eqqcolon\sigma_w^{2} .
	\label{eq:Wt-trivial-UB-TD}
\end{align}
Equipped with the deterministic bounds \eqref{eq:TD_B} and \eqref{eq:Wt-trivial-UB-TD}, we can first apply Freedman's inequality while retaining the random cumulative variance $W_t$. To this end, take $K=2\lceil\log_2\frac{1}{1-\gamma}\rceil$. Since \eqref{eq:thm:sample-size-TD-thm} and $\varepsilon\leq1/(1-\gamma)$ ensure $T\geq\max\{8,1/(1-\gamma)\}$ for sufficiently large $c_3$, we have
\begin{equation}\label{eq:TD-log-bound}
\frac{\sigma_w^2}{2^K}\leq\frac{3\log^3T}{(1-\gamma)^2T},\qquad
\log\frac{4STK}{\delta}\leq3\log\frac{ST}{\delta}.
\end{equation}
For a fixed $t$, applying Lemma~\ref{lem:Freedman-prelim} to each entry of $\sum_{k=1}^t Z_k$ and taking a union bound over all $S$ entries give, with probability at least $1-\delta/(2T)$,
\begin{equation}\label{eq:TD-freedman}
\bigg\|\sum_{k=1}^t Z_k\bigg\|_\infty
\leq\sqrt{8\max\bigg\{\|W_t\|_\infty,\frac{\sigma_w^2}{2^K}\bigg\}
\log\frac{4STK}{\delta}}+\frac43B\log\frac{4STK}{\delta},
\end{equation}
with $B$ and $\sigma_w^2$ defined in \eqref{eq:TD_B} and \eqref{eq:Wt-trivial-UB-TD}, respectively. 
It then boils down to  deriving a sharper bound on $\|W_t\|_{\infty}$.

\paragraph{Step 3: exploiting the variance structure of $W_t$.}
To further sharpen the bound on $W_t$, the key idea is to relate the conditional variance of each TD update to the decrease in the squared iterates, which, as we shall see shortly, allows the weighted sum of conditional variances to be decomposed into a telescoping term, a martingale term, and a lower-order
residual term. To begin, since $k>t/2$, Lemma~\ref{lemma:TD-step-size} gives
\begin{equation}
\eta_k^{(t)}\leq\eta_k\leq\frac{2\log^3T}{(1-\gamma)T}
\eqqcolon\overline\eta,\qquad t/2<k\leq t.
\label{eq:defn-bar-eta-large-k-TD}
\end{equation}
Combining this bound with \eqref{eq:TD-var-Zk-upper-bound-each-term} yields
\begin{align}
&\sum_{k=\lfloor t/2\rfloor+1}^t\mathsf{Var}_{k-1}(Z_k) \arxiv{\notag\\}
\arxiv{&\quad}\leq\overline\eta\gamma^2\sum_{k=\lfloor t/2\rfloor+1}^t
\eta_k\prod_{i=k+1}^t\big(I-\eta_i(I-\gamma P_\pi)\big)
\mathsf{Var}_{P_\pi}(V_{k-1})
\eqqcolon\overline W_t.
\label{eq:TD-variance-proxy}
\end{align}
It then comes down to controlling $\overline W_t$. Let $\mathbb E_{k-1}$ denote conditional expectation given the history through iteration $k-1$.
We first establish below a one-step relation between the conditional
variance and the evolution of the squared TD iterates. This relation is a key ingredient in our refined analysis, which differs from the proof in \citet{li2024q}. Its proof is deferred to Section~\ref{sec:proof:lemma:TD-square}. 
\begin{lemma}
\label{lemma:TD-square}
For every $k\geq1$,
\begin{align}
\eta_k(1-\eta_k)\gamma^2
\mathsf{Var}_{P_\pi}(V_{k-1})
&\leq
\big(I-\eta_k(I-\gamma P_\pi)\big)
(V_{k-1}\circ V_{k-1}) \arxiv{\notag\\
&\quad}
-\mathbb E_{k-1}[V_k\circ V_k]
+\frac{2\eta_k}{1-\gamma}1_S .
\label{eq:TD-square-energy}
\end{align}
\end{lemma}

We now return to $\overline W_t$ in \eqref{eq:TD-variance-proxy}. The sample-size condition~\eqref{eq:thm:sample-size-TD-thm} and $\varepsilon\leq1/(1-\gamma)$ ensure $\overline\eta\leq1/2$ for sufficiently large $c_3$. Applying Lemma~\ref{lemma:TD-square} to the weighted sum in
\eqref{eq:TD-variance-proxy}, and adding and subtracting $V_k\circ V_k$ inside the weighted sum, we obtain
\begin{align}
\overline W_t
&\leq\frac{\overline\eta}{1-\overline\eta}\Bigg\{
\sum_{k=\lfloor t/2\rfloor+1}^t
\prod_{i=k+1}^t\big(I-\eta_i(I-\gamma P_\pi)\big) \arxiv{\notag\\}
\arxiv{&\qquad\quad\cdot} \Big[\big(I-\eta_k(I-\gamma P_\pi)\big)(V_{k-1}\circ V_{k-1})-V_k\circ V_k\Big] \notag\\
&\book{\qquad\qquad}\qquad+\sum_{k=\lfloor t/2\rfloor+1}^t
\prod_{i=k+1}^t\big(I-\eta_i(I-\gamma P_\pi)\big)
\big(V_k\circ V_k-\mathbb E_{k-1}[V_k\circ V_k]\big)\notag\\
&\book{\qquad\qquad}\qquad+\frac{2}{1-\gamma}\sum_{k=\lfloor t/2\rfloor+1}^t
\eta_k\prod_{i=k+1}^t\big(I-\eta_i(I-\gamma P_\pi)\big)1_S
\Bigg\}.
\label{eq:overline-Wt-three-sums-TD}
\end{align}
Note that the second sum in \eqref{eq:overline-Wt-three-sums-TD} consists of centered fluctuations. To express it as a martingale sum, define
\begin{equation}
\widetilde Z_k\coloneqq
\prod_{i=k+1}^t\big(I-\eta_i(I-\gamma P_\pi)\big)
\big(V_k\circ V_k-\mathbb E_{k-1}[V_k\circ V_k]\big),
\quad k>t/2.
\label{eq:TD-square-martingale}
\end{equation}
The matrix weights are deterministic, so $\mathbb E_{k-1}[\widetilde Z_k]=0$ and $\{\widetilde Z_k\}$ forms a martingale difference sequence. The first sum in \eqref{eq:overline-Wt-three-sums-TD} telescopes, resulting in
\begin{align*}
&\sum_{k=\lfloor t/2\rfloor+1}^t
\prod_{i=k+1}^t\big(I-\eta_i(I-\gamma P_\pi)\big) \arxiv{\\}
\arxiv{&\qquad\cdot} \Big\{\big(I-\eta_k(I-\gamma P_\pi)\big)(V_{k-1}\circ V_{k-1})-V_k\circ V_k\Big\}\\*
&\quad=\prod_{i=\lfloor t/2\rfloor+1}^t\big(I-\eta_i(I-\gamma P_\pi)\big)
(V_{\lfloor t/2\rfloor}\circ V_{\lfloor t/2\rfloor})-V_t\circ V_t\\*
&\quad\leq\frac{1}{(1-\gamma)^2}1_S,
\end{align*}
where the last inequality uses Lemma~\ref{lemma:TD-bounded} and the row-sum identity~\eqref{eq:TD-L1-norm}. The third sum in \eqref{eq:overline-Wt-three-sums-TD} is at most $\frac{2}{(1-\gamma)^2}1_S$, due to the matrix telescoping calculation in \eqref{eq:telescoping_td} and $(I-\gamma P_\pi)^{-1}1_S=\frac{1}{1-\gamma}1_S$. Combining the above bounds with \eqref{eq:TD-square-martingale}, we obtain
\begin{equation}
\overline W_t\leq\frac{\overline\eta}{1-\overline\eta}
\bigg(\frac{3}{(1-\gamma)^2}1_S+
\sum_{k=\lfloor t/2\rfloor+1}^t\widetilde Z_k\bigg).
\label{eq:TD-variance-energy-bound}
\end{equation}
Thus, controlling $\overline W_t$ reduces to controlling the martingale sum
$\sum_k \widetilde Z_k$.

\paragraph{Step 4: applying martingale concentration to bound $\|W_t\|_{\infty}$.} 
The centered term in \eqref{eq:TD-variance-energy-bound} from the preceding step is itself a martingale sum. Applying
Freedman's inequality to this second martingale yields a self-bounding
inequality for the variance proxy $W_t$, which ultimately provides the sharp bound on $W_t$
needed in the original Freedman inequality. More precisely, applying
Freedman's inequality to the martingale sequence
$\{\widetilde Z_k\}$ defined in~\eqref{eq:TD-square-martingale} gives the following result. 
\begin{lemma}
\label{lemma:TD-variance-proxy-bound}
Let $\overline W_t$ be defined in
\eqref{eq:TD-variance-proxy}, and $\overline{\eta}$ be defined in \eqref{eq:defn-bar-eta-large-k-TD}.  Then, with probability at least $1-\delta/(2T)$, 
\begin{align}
\|\overline W_t\|_\infty
&\leq \frac{16\overline\eta}{(1-\gamma)^2},
\qquad
\|W_t\|_\infty
\leq
\frac{33\log^3T}{(1-\gamma)^3T}.
\label{eq:TD-Wt-bound-lemma}
\end{align}
\end{lemma}

The proof of Lemma~\ref{lemma:TD-variance-proxy-bound} is deferred to
Section~\ref{sec:proof:lemma:TD-variance-proxy-bound}. It is noteworthy that the derivation
of this lemma is a key point of departure from the analysis of
\citet{li2024q}, enabling us to extend the guarantee to the full
$\varepsilon$-range. Its main ingredient is a second martingale
decomposition of the squared TD iterates, which exploits the
variance structure of the MDP and yields the sharp dependence on the
effective horizon.

\paragraph{Step 5: putting all pieces together.}
Thus far, we have shown that, for each fixed $t\geq T/(c_2\log T)$,  \eqref{eq:TD-freedman} and \eqref{eq:TD-Wt-bound-lemma} hold simultaneously with probability at least $1-\delta/T$. On this event, substituting the variance bound \eqref{eq:TD-Wt-bound-lemma} into \eqref{eq:TD-freedman}, together with \eqref{eq:TD_B} and \eqref{eq:TD-log-bound}, we see that 
\begin{align}
\bigg\|\sum_{k=1}^t Z_k\bigg\|_\infty
&\leq \frac{C}{2}\sqrt{\frac{(\log^3T)\big(\log\frac{ST}{\delta}\big)}{(1-\gamma)^3T}}
+\frac{C(\log^3T)\big(\log\frac{ST}{\delta}\big)}{2(1-\gamma)^2T}\arxiv{\notag\\
&}\leq C\sqrt{\frac{(\log^3T)\big(\log\frac{ST}{\delta}\big)}{(1-\gamma)^3T}}
\label{eq:TD-second-term}
\end{align}
for some large enough universal constant $C>0$. 
Plugging \eqref{eq:TD-second-term} into the error decomposition~\eqref{eq:defn-xit-TD-sharp} and taking a union bound over all $t$ obeying $T/(c_2\log T)\leq t\leq T$, we obtain
\begin{equation}
\|\Delta_t\|_\infty
\leq\frac{1}{(1-\gamma)T^2}
+C\sqrt{\frac{(\log^3T)\big(\log\frac{ST}{\delta}\big)}{(1-\gamma)^3T}}
\lesssim \sqrt{\frac{(\log^3T)\big(\log\frac{ST}{\delta}\big)}{(1-\gamma)^3T}}
\label{eq:TD-sharp-final}
\end{equation}
with probability at least $1-\delta$. This concludes the proof.

\section{Q-learning}  \label{sec:Q-learning}

Having established the appealing performance of synchronous TD learning for policy evaluation, we now turn our attention to the use of model-free algorithms for learning optimal Q-functions.  In particular, we focus on the Q-learning algorithm introduced by \citet{watkins1989learning,watkins1992q}, one of the most celebrated model-free reinforcement learning methods.

\subsection{Algorithm}

Akin to TD learning, the Q-learning algorithm is designed as a stochastic approximation method for computing a fixed point of the Bellman operator.  A key distinction is that the operator of interest is now the Bellman optimality operator \eqref{eq:bellman_optimality_operator}, rather than the Bellman consistency operator \eqref{eq:bellman_consistency_operator}. Unlike the latter, the Bellman optimality operator is nonlinear and nonsmooth (owing to the presence of the max operation), which introduces additional algorithmic and analytical challenges.

We now describe the algorithmic procedure of Q-learning. Following the stochastic approximation framework \eqref{eq:general-stoc-approx}, we maintain a running estimate of the optimal Q-function, $Q_t: \cS\times \cA \rightarrow \mathbb{R}$, for all $t\geq 0$, and at each iteration $t$, we update the Q-function estimate according to
\begin{subequations} \label{eq:q-learning-update-rule}
\begin{equation}
\label{eqn:q-learning-update}
	Q_t  = (1- \eta_t ) Q_{t-1} + \eta_t \mathcal{T}_t (Q_{t-1}) , 
\end{equation}
where we initialize it such that $0\leq Q_0(s,a) \leq \frac{1}{1-\gamma}$ for every state-action pair $(s,a)$. 
Here, $\eta_t \in (0,1]$ denotes the learning rate at iteration $t$,   
and $\mathcal{T}_t(Q_{t-1})$ is constructed as an unbiased estimator of $\mathcal{T}(Q_{t-1})$, where $\mathcal{T}(\cdot)$ denotes the Bellman optimality operator \eqref{eq:bellman_optimality_operator}. With the generative model in place, $\mathcal{T}_t(Q_{t-1})$ can be constructed as follows:
\begin{itemize}
    \item for every pair $(s,a)\in \cS\times \cA$, draw an independent sample 
$$s_t(s,a) \sim P(\cdot \mymid s,a)$$  
by querying the generative model; 

    \item compute $\mathcal{T}_t(Q_{t-1})$ for every $(s,a)$-th entry as
\begin{align}  \label{defn:q-learning-Bellman}
	\mathcal{T}_t (Q_{t-1}) (s, a ) &= 
		 r(s , a ) + \gamma \max_{a' \in \cA} Q_{t-1}\big(s_t(s,a), a'\big) .
\end{align}
\end{itemize}
\end{subequations}

This version of Q-learning is commonly referred to as {\em synchronous} Q-learning, since all entries of the Q-function are updated simultaneously at each iteration.  In addition, the corresponding estimate for the optimal value function, $V_t: \cS \rightarrow \mathbb{R}$, can be computed as
\begin{align}
	\label{defn:Vt}
	\forall s\in \cS: \qquad V_t(s) :=  \max_{a\in \cA} Q_{t}(s,a) 
\end{align}
 at iteration $t$. The complete description of the Q-learning procedure is summarized in Algorithm~\ref{alg:q-infinite}.

\begin{algorithm}[t]
	%\begin{algorithmic}[1] 
    \DontPrintSemicolon
	\textbf{input:} discount factor $\gamma$, reward function $r$, learning rates $\{\eta_t\}$, number of iterations $T$, initialization $Q_0$. \\
%\STATE  \textbf{initialization:} $Q_0=0$. 
   \For{$t=1,2,\cdots,T$}{
  %  \For{$(s,a)\in \cS\times \cA$}{
	   draw a sample $s_t(s,a) \overset{\mathrm{ind.}}{\sim}  P(\cdot \mymid s,a)$ for each $(s,a)\in \cS\times \cA$. \\
       {\color{blue}\tcc{stochastic approximation.}}
	   update $Q_t$ according to
       %\eqref{eq:q-learning-update-rule}.  
\begin{align}
\label{eqn:q-learning-update-alg}
	Q_t  &= (1- \eta_t ) Q_{t-1} + \eta_t \mathcal{T}_t (Q_{t-1}) , 
    \end{align}
    where for every $(s,a)$ pair, the $(s,a)$-th entry of $\mathcal{T}_t (Q_{t-1})$ is 
    \begin{align}
    	\mathcal{T}_t (Q_{t-1}) (s , a ) &= 
		 r(s , a ) + \gamma \max_{a' \in \cA} Q_{t-1}\big(s_t(s,a), a'\big) .
         %,~~\forall (s,a)\in \cS\times \cA.
\end{align}
    }
   % }
%\textbf{output} $Q_T$.		
   %\textbf{return:} ${Q}_{T}$. 
   %\end{algorithmic} 
    \caption{Q-learning with a generative model}
 \label{alg:q-infinite}
\end{algorithm}

\subsection{Theoretical guarantees} 

Having established the minimax optimality of TD learning over the accuracy range covered by the minimax lower bound, which can be viewed as a special case of Q-learning with number of actions $A=1$, it is natural to conjecture that Q-learning should likewise attain near-optimal performance.  However, this intuition breaks down when $A \geq 2$,  as the multi-action setting fundamentally alters the situation. We make this precise via the following $\ell_{\infty}$-based sample complexity bound for Q-learning.

To state the result, we need to specify the learning rate schedule. In particular, we choose 
deterministic learning rates satisfying
\begin{equation}
\frac{1}{1+\frac{c_1 (1-\gamma) T}{\log^3 T}}\le\eta_{t}\le\frac{1}{1+\frac{c_2 (1-\gamma) t}{\log^3 T}},
\qquad 0\leq t\leq T
\label{eq:thm:infinite-horizaon-eta}
\end{equation}
for some universal constants $c_{1}\geq c_{2}>0$. Similar to the discussions in Section~\ref{sec:TD-learning-theory}, such choices encompass both constant and rescaled linear learning rates (when appropriately chosen). The theorem below presents the state-of-the-art
sample complexity analysis for Q-learning, originally established by \citet{li2024q}. In fact, it further extends the results of \citet{li2024q} by broadening the admissible range of \(\varepsilon\) from \((0,1]\) to the full range through a refined --- and, indeed, simpler --- analysis.
\begin{theorem}[Extension of \cite{li2024q}]
\label{thm:infinite-horizon-simple}
Consider any $\delta\in(0,1)$,  $\varepsilon\in\big(0,\frac{1}{1-\gamma}\big]$, $\gamma\in [1/2,1)$, and $A\geq 2$. Suppose that the deterministic learning rates satisfy \eqref{eq:thm:infinite-horizaon-eta} for some small enough universal constants $c_{1}\geq c_{2}>0$, and that the total number of iterations $T\geq2$ satisfies
\begin{equation}
T\ge\frac{c_{3}\big(\log^{4}T\big)\big(\log\frac{SAT}{\delta}\big)}{(1-\gamma)^{4}\varepsilon^{2}}
	\label{eq:thm:infinite-horizon-T}
\end{equation}
for some large enough universal constant $c_{3}>0$. 
If the initialization obeys $0\leq {Q}_{0}(s,a) \leq\frac{1}{1-\gamma}$ for every $(s,a)\in \cS\times \cA$,  then with probability at least $1-\delta$, 
Q-learning (cf.~Algorithm \ref{alg:q-infinite}) achieves
\begin{align*}
\forall (s,a)\in \cS\times \cA: \qquad 
	 \big| {Q}_{T}(s,a) -  {Q}^{\star}(s,a) \big| \le\varepsilon.
	%\label{eq:QT-entrywise-Q-thm}
\end{align*}
\end{theorem}
\begin{remark}[Mean estimation guarantee]
    The high-probability guarantee in Theorem~\ref{thm:infinite-horizon-simple} can be readily converted into a bound on the mean estimation
error. In light of the crude upper bound $\big| {Q}_{T}(s,a) -  {Q}^{\star}(s,a) \big| \le \frac{1}{1-\gamma}$,  one can take $\delta = \varepsilon (1-\gamma)/2$ to obtain
\begin{equation*}
    \mathbb{E}\bigg[ \max_{(s,a)\in \cS\times \cA}\big| {Q}_{T}(s,a) -  {Q}^{\star}(s,a)\big| \bigg] \leq \varepsilon (1-\delta)+ \frac{1}{1-\gamma} \delta \leq 2\varepsilon, 
\end{equation*}
provided that the total number of iterations obeys 
\begin{equation*}
T\ge\frac{c_{3}\big(\log^{4}T\big)\big(\log\frac{2SAT}{\varepsilon (1-\gamma)}\big)}{(1-\gamma)^{4}\varepsilon^{2}}.
	%\label{eq:thm:infinite-horizon-T-mean}
\end{equation*}
\end{remark}

In essence, Theorem~\ref{thm:infinite-horizon-simple} develops a non-asymptotic upper bound on the iteration complexity of Q-learning in the presence of the generative model.  
Recognizing that $S A$ independent samples are drawn in each iteration --- yielding a total of $SAT$ samples overall ---  
we can readily translate this result into the following sample complexity bound
\begin{equation}
	\widetilde{O}\bigg( \frac{S A}{(1-\gamma)^4 \varepsilon^2} \bigg),
	\label{eq:sample-complexity-sync-Q-learning}
\end{equation}
which suffices for Q-learning to achieve $\varepsilon$-accuracy (for $0 < \varepsilon \leq 1/(1-\gamma)$) in the entrywise sense.

We provide further discussion about the learning rate choices. When $T\asymp \frac{ (\log^{4}T)\big(\log\frac{SAT}{\delta}\big)}{(1-\gamma)^{4}\varepsilon^{2}}$, Condition~\eqref{eq:thm:infinite-horizaon-eta} permits adopting the constant learning rates 
\[
	\eta_t \equiv \frac{1}{1+\frac{c_1 (1-\gamma) T}{\log^3 T}} = 
    \widetilde{O} \big( (1-\gamma)^3 \varepsilon^2 \big),  \qquad 0\leq t\leq T, 
\]
depending almost solely on the discount factor $\gamma$ and the target accuracy $\varepsilon$. 
Interestingly, both constant and rescaled linear learning rates satisfying \eqref{eq:thm:infinite-horizaon-eta} lead to the same sample complexity bound (in an order-wise sense), offering flexibility and robustness for practical use.

Nevertheless, the sample complexity bound in \eqref{eq:sample-complexity-sync-Q-learning}, over the accuracy range covered by Theorem~\ref{theorem:azar}, fails to match the minimax lower bound stated there. More specifically, the sample complexity bound of Q-learning incurs an additional $\widetilde{O}\big(\frac{1}{1-\gamma}\big)$ multiplicative factor compared to what is achievable by its model-based counterpart (cf. Theorem~\ref{Thm:sample-compl-main} \citep{li2024breaking}). In sharp contrast to the $A=1$ regime --- where TD learning attains near-optimal performance --- the presence of multiple actions ($A \ge 2$) appears to degrade the sample optimality of Q-learning.

% The careful reader  remark that there remains a gap between the sample complexity bound for Q-learning (cf.~Theorem~\ref{thm:infinite-horizon-simple})  and the minimax lower bound (cf.~Theorem~\ref{theorem:azar}). More specifically, the minimax lower bound scales on the order of $\frac{S A}{(1-\gamma)^3\varepsilon^2}$ and is achievable---up to some logarithmic factor---by the model-based approach \citep{azar2013minimax,agarwal2019optimality,li2020breaking}.  This raises natural questions regarding

\subsection{Tightness and suboptimality}
\label{sec:Q_learning_suboptimality}
The gap between the upper bound in Theorem~\ref{thm:infinite-horizon-simple} and the minimax lower bound in Theorem~\ref{theorem:azar} naturally raises the following questions: can the sample complexity upper bound for Q-learning be further improved, or is there an inherent bottleneck that prevents plain Q-learning from achieving optimal data efficiency? As it turns out, the horizon dependence in Theorem~\ref{thm:infinite-horizon-simple} is sharp in the regime of the lower bound below, which shows that Q-learning is provably suboptimal when there are at least two actions. This is confirmed by the lower bound below, originally established by  \citet{li2024q}.
\begin{theorem}[\cite{li2024q}]
\label{thm:LB-example-Q-generative}
	Assume that $3/4\leq \gamma < 1$ and  $T\geq \frac{c_3}{(1-\gamma)^2}$ for some sufficiently large constant $c_3>0$. 
	Suppose that the initialization is $Q_0 \equiv 0$, and that the learning rates are taken to be either (i) $\eta_t = \frac{1}{1+\ceta (1-\gamma)t}$ for any constant $c_{\eta}>0$ for all $t\geq 0$, or (ii) $\eta_t \equiv \eta \in (0,1)$ for all $t\geq 0$.  
	There exists a $\gamma$-discounted MDP with $S=4$ and $A=2$ such that Q-learning (cf.~Algorithm \ref{alg:q-infinite}) obeys
\begin{equation}
	\max_{s\in \cS} \mathbb{E} \Big[ \big|V_T(s) - V^{\star}(s)\big|^2 \Big] 
	\geq  \frac{c_{\mathsf{lb}}}{(1-\gamma)^4T\log^2 T} 
	\label{eq:lower-bounding-VT-thm-sync}
\end{equation}
for some universal constant $c_{\mathsf{lb}} > 0$.
\end{theorem}

It is noteworthy that establishing an {\em algorithm-specific} lower bound requires techniques fundamentally different from those used in information-theoretic analyses (e.g., Fano’s lemma is no longer useful here). Although the hard MDP instance constructed in the proof of this theorem is partially inspired by the one in Section~\ref{sec:minimax-lower-bound-generative} for deriving minimax lower bounds, the technical approach is drastically different. We defer the proof of this theorem to Section~\ref{sec:proof-thm-LB-Q-generative}, which requires carefully tracking the dynamics of Q-learning throughout its execution.

We now discuss the implications of the algorithm-dependent lower bound established in Theorem~\ref{thm:LB-example-Q-generative}. This result shows that plain Q-learning cannot achieve $  \varepsilon  $-accuracy, in the sense that $\max_s \mathbb{E} \big[ \big|V_T(s) - V^{\star}(s)\big|^2 \big] \leq \varepsilon^2$, 
unless the number of iterations already exceeds 
\[
	\widetilde{\Omega}\bigg( \frac{1}{(1-\gamma)^{4}\varepsilon^{2}}\bigg),
\]
within the iteration regime of Theorem~\ref{thm:LB-example-Q-generative}, when either constant or rescaled linear learning rates are used. This is a lower bound for the stated mean-square criterion; it does not assert a matching high-probability lower bound over the entire enlarged accuracy interval. As a consequence, the upper bound for Q-learning derived in Theorem~\ref{thm:infinite-horizon-simple} is tight in its dependence on the effective horizon $\frac{1}{1-\gamma}$ when $A\geq 2$.  Crucially, this lower bound reveals an inherent suboptimality of plain Q-learning: its horizon dependence is provably worse than the minimax rate by a multiplicative factor of $\frac{1}{1-\gamma}$ (up to logarithmic factors) once there are at least two actions.

We note that the potential suboptimality of Q-learning has been recognized in the literature since its early development. A commonly cited explanation attributes such limitations to the over-estimation of Q-values arising from the use of the max operator applied to empirical estimates (see, e.g., \citet{thrun1993issues,hasselt2010double,van2016deep}). Informally speaking, even in the idealistic scenario where $Q_{t-1}$ forms a nearly unbiased estimate of $Q^{\star}$, taking the maximum over actions can introduce an inherent positive bias, that is,
\[
    \mathbb{E}\Big[\max_{a'\in \mathcal{A}}Q_{t-1}( s, a') \Big] 
    \geq \max_{a'\in \mathcal{A}} \mathbb{E}\big[Q_{t-1}( s, a') \big] 
    \approx \max_{a'\in \mathcal{A}} \mathbb{E}\big[Q^{\star}( s, a') \big] ,
\]
where $\max_{a'\in \mathcal{A}}Q_{t-1}( s, a')$ is a key term in the empirical Bellman operator $\mathcal{T}_t(Q_{t-1})$. 
In this sense,  Theorems~\ref{thm:infinite-horizon-simple} and \ref{thm:LB-example-Q-generative} taken together can be viewed as quantifying the impact of over-estimation on the sample complexity and convergence rate of plain Q-learning. A more quantitative comparison between over-estimation and under-estimation errors is provided towards the end of Section~\ref{sec:sharp-analysis-Q-generative}.

\subsection{Analysis: a first attempt}
\label{sec:proof-Q-learning-loose-generative}
Before presenting the proof of Theorem~\ref{thm:infinite-horizon-simple}, we first illustrate how to establish a looser upper bound $\widetilde{O}\big(\frac{1}{(1-\gamma)^5\varepsilon^2}\big)$ on the iteration complexity, which exceeds the advertised bound in Theorem~\ref{thm:infinite-horizon-simple} by a multiplicative factor of $\frac{1}{1-\gamma}$. Although suboptimal, this result features a simpler proof and introduces the error decomposition and comparison argument that we shall refine in the next subsection.

Akin to the analysis of TD learning, we adopt the vector and matrix notation introduced in Section~\ref{sec:vector-matrix-notation}, allowing us to express the Q-learning update rule \eqref{eq:q-learning-update-rule} as
\begin{align}
 Q_t=(1-\eta_t)Q_{t-1}+\eta_t\big(r+\gamma P_tV_{t-1}\big).
 \label{eq:iteration-rule-Q-infinite}
\end{align}
Here, the empirical transition matrix $P_t\in\{0,1\}^{SA\times S}$ is defined as
\begin{align}
\label{eq:defn-Pt-Q-learning-sync-generative}
\forall(s,a,s')\in\cS\times\cA\times\cS:\quad
P_t\big((s,a),s'\big)\coloneqq
\begin{cases}
1,&\text{if }s'=s_t(s,a),\\
0,&\text{otherwise},
\end{cases}
\end{align}
and the vector $V_t\in\mathbb{R}^S$ is defined as $V_t(s)=\max_aQ_t(s,a)$ for all $s\in\cS$.

\paragraph{Preliminary facts.}
We begin with a lemma showing that, when initialized properly, each coordinate of both the Q-function and the value function estimates remains within $\big[0,\frac{1}{1-\gamma}\big]$ throughout the execution of the Q-learning algorithm. Its proof is deferred to Section~\ref{appendix:proof-lemma-Q-bounded}.
\begin{lemma}
\label{lemma:non-negativity-Qt-Vt}
Assume that $0\leq Q_0(s,a)\leq\frac{1}{1-\gamma}$ for all $(s,a)\in\cS\times\cA$.
Then for every $t\geq0$, we have
\begin{subequations}
\begin{equation}
0\leq Q_t\leq\frac{1}{1-\gamma}1_{SA}
\qquad\text{and}\qquad
0\leq V_t\leq\frac{1}{1-\gamma}1_S,
\label{eq:Q-i-LB-UB}
\end{equation}
provided that $0\leq\eta_t\leq1$ for all $t\geq0$. As an immediate consequence,
\begin{equation}
\|Q_t-Q^\star\|_\infty\leq\frac{1}{1-\gamma}
\quad\text{for all }t\geq0.
\label{eq:Delta-t-range-infinite}
\end{equation}
\end{subequations}
\end{lemma}
Additionally, with a slight abuse of notation, we define
\begin{equation}
\eta_i^{(t)}\coloneqq
\begin{cases}
\prod_{j=1}^t(1-\eta_j),&\text{if }i=0,\\
\eta_i\prod_{j=i+1}^t(1-\eta_j),&\text{if }0<i<t,\\
\eta_t,&\text{if }i=t,
\end{cases}
\label{def:eta-i-t-simple}
\end{equation}
which captures the cumulative effect of the learning rates over time. Note that this definition differs from the TD counterpart in~\eqref{def:eta-k-t} in that it does not include the factor $1-\gamma$. For the remainder of this subsection and the sharp analysis that follows, we set
\begin{equation}\label{eq:Q-comparison-parameters}
\overline{\eta}\coloneqq\min\left\{1,\frac{2\log^3T}{c_2(1-\gamma)T}\right\}.
\end{equation}
 The lemma below collects some useful properties of $\eta_i^{(t)}$, with its proof deferred to Section~\ref{appendix:proof-lemma-Q-step-size}.
\begin{lemma}\label{lemma:Q-step-size}
Let $\overline{\eta}$ be defined in~\eqref{eq:Q-comparison-parameters}. For $\lfloor T/2\rfloor<i\leq t\leq T$, the weights~\eqref{def:eta-i-t-simple} satisfy
\begin{equation}\label{eq:Q-comparison-weights}
0\leq\eta_i^{(t)}\leq\eta_i\leq\overline{\eta}
\qquad \text{and} \qquad
\sum_{i=\lfloor T/2\rfloor+1}^t\big(\eta_i^{(t)}\big)^2\leq\overline{\eta}.
\end{equation}
Moreover, it holds that
\begin{equation}
\sum_{i=\tau}^t\eta_i^{(t)}=
\begin{cases}
1-\prod_{j=\tau}^t(1-\eta_j),&\text{if }1\leq\tau\leq t,\\
1,&\text{if }\tau=0.
\end{cases}
\label{eq:Q-weight-sum}
\end{equation}
\end{lemma}

\paragraph{Step 1: decomposing the error $Q_t-Q^\star$.}
Define the error sequence
\begin{align}
\Delta_t\coloneqq Q_t-Q^\star.
\label{eq:defn-Delta-t-Q-learning-sync}
\end{align}
By the update rule~\eqref{eq:iteration-rule-Q-infinite} and the Bellman optimality equation $Q^\star=r+\gamma PV^\star$, we can decompose
\begin{align}
\Delta_t
&= (1-\eta_t) Q_{t-1} + \eta_t\big( r + \gamma P_t V_{t-1} \big) - Q^\star \nonumber \\
&= (1-\eta_t)\big( Q_{t-1} - Q^\star \big) + \eta_t\big( r + \gamma P_t V_{t-1} - Q^\star \big) \nonumber \\
&=(1-\eta_t)\Delta_{t-1}+\gamma\eta_t\big(P_tV_{t-1}-PV^\star\big)\nonumber\\
&=(1-\eta_t)\Delta_{t-1}
+\gamma\eta_tP_t(V_{t-1}-V^\star)
+\gamma\eta_t(P_t-P)V^\star.
\label{eq:iteration-infinite}
\end{align}
Unrolling the recursion~\eqref{eq:iteration-infinite} from iteration $\lfloor T/2\rfloor$ gives
\begin{align}
\Delta_t
&=\bigg(\prod_{j=\lfloor T/2\rfloor+1}^t(1-\eta_j)\bigg)\Delta_{\lfloor T/2\rfloor}
+\gamma\sum_{i=\lfloor T/2\rfloor+1}^t\eta_i^{(t)}P_i(V_{i-1}-V^\star) \nonumber \\
& \qquad + \underbrace{ \gamma\sum_{i=\lfloor T/2\rfloor+1}^t\eta_i^{(t)}(P_i-P)V^\star }_{\eqqcolon\, \xi_t} , \qquad \lfloor T/2\rfloor<t\leq T. \label{eq:Q-loose-error-decomposition} 
\end{align}
This identity decomposes the estimation error into three components: the error inherited from the previous $\lfloor T/2\rfloor$ iterations, the estimation error propagated thereafter, and the fluctuation term $\xi_t$, which we tackle separately below.

\paragraph{Step 2: controlling the estimation error without fluctuations.} Given that
\[
\|V_{i-1}-V^\star\|_\infty
\leq\|Q_{i-1}-Q^\star\|_\infty=\|\Delta_{i-1}\|_\infty,
\]
combining~\eqref{eq:Q-loose-error-decomposition} with the basic properties $\|P_i\|_1=1$ and $\|\Delta_{\lfloor T/2\rfloor}\|_\infty \leq \frac{1}{1-\gamma}$ implies that
\begin{align}
\|\Delta_t\|_\infty
&\leq\frac{1}{1-\gamma}\prod_{j=\lfloor T/2\rfloor+1}^t(1-\eta_j)
+\gamma\sum_{i=\lfloor T/2\rfloor+1}^t\eta_i^{(t)}\|\Delta_{i-1}\|_\infty  +\|\xi_t\|_\infty.
\label{eq:Q-loose-norm-recursion}
\end{align}

To analyze this recursion, we introduce an auxiliary sequence that
captures, in some sense, the deterministic part of the estimation error.  Set $c_{\lfloor T/2\rfloor}\coloneqq \frac{1}{1-\gamma}$, and for all $t$ obeying $\lfloor T/2\rfloor<t\leq T$, define recursively that
\begin{equation}
c_t\coloneqq\frac{1}{1-\gamma}\prod_{j=\lfloor T/2\rfloor+1}^t(1-\eta_j)
+\gamma\sum_{i=\lfloor T/2\rfloor+1}^t\eta_i^{(t)}c_{i-1}.
\label{eq:Q-warmup-radius-identity}
\end{equation}
Clearly, $c_t$ satisfies similar recursion as the deterministic part of
\eqref{eq:Q-loose-norm-recursion}, with the fluctuation term
$\|\xi_t\|_\infty$ omitted.  Recalling from~\eqref{def:eta-i-t-simple} that $\eta_i^{(t)}=(1-\eta_t)\eta_i^{(t-1)}$ for $i<t$ and $\eta_t^{(t)}=\eta_t$, one can rewrite the last summand in~\eqref{eq:Q-warmup-radius-identity} to derive
\begin{align*}
c_t&=(1-\eta_t)c_{t-1}+\gamma\eta_tc_{t-1} =\big(1-\eta_t(1-\gamma)\big)c_{t-1}.
\end{align*}
Iterating this scalar relation yields
\begin{equation}
c_t=\frac{1}{1-\gamma}
\prod_{i=\lfloor T/2\rfloor+1}^t\big(1-\eta_i(1-\gamma)\big),\qquad \lfloor T/2\rfloor\leq t\leq T,
\label{eq:Q-warmup-radius}
\end{equation}
where an empty product equals one. Informally speaking, $c_t$ reflects the error obtained by propagating the
initial bound $\|\Delta_{\lfloor T/2\rfloor}\|_\infty\leq(1-\gamma)^{-1}$
through the recursion~\eqref{eq:Q-loose-norm-recursion} after setting
$\|\xi_t\|_\infty=0$.
In particular, the factor $1-\eta_i(1-\gamma)$ captures the contraction
factor governing the propagation of initialization error, analogous to
the factor appearing in the TD bound~\eqref{eq:TD-first-term}.

\paragraph{Step 3: accounting for the effect of fluctuations.}
We still need to take into account the fluctuation terms in~\eqref{eq:Q-loose-norm-recursion}. Towards this end, we now show that these fluctuations affect the estimation error only through their largest magnitude over time. More precisely, we claim that
\begin{equation}
\|\Delta_t\|_\infty\leq c_t+\frac{1}{1-\gamma}
\max_{\lfloor T/2\rfloor<i\leq T}\|\xi_i\|_\infty
\qquad\text{for all }\lfloor T/2\rfloor\leq t\leq T.
\label{eq:Q-loose-comparison}
\end{equation}
We prove this claim~\eqref{eq:Q-loose-comparison} by induction. The base case with $t=\lfloor T/2\rfloor$ follows from Lemma~\ref{lemma:non-negativity-Qt-Vt} and $c_{\lfloor T/2\rfloor}=(1-\gamma)^{-1}$. Suppose the claim holds at all iterations prior to $t$. Substituting this bound into
\eqref{eq:Q-loose-norm-recursion}, and using
\eqref{eq:Q-warmup-radius-identity} and~\eqref{eq:Q-weight-sum}, we arrive at
\begin{align*}
\|\Delta_t\|_\infty
&\leq c_t
+\frac{\gamma}{1-\gamma}
\max_{\lfloor T/2\rfloor<j\leq T}\|\xi_j\|_\infty
\left(1-\prod_{j=\lfloor T/2\rfloor+1}^t(1-\eta_j)\right)
+\|\xi_t\|_\infty\\
&\leq c_t
+\frac{1}{1-\gamma}
\max_{\lfloor T/2\rfloor<j\leq T}\|\xi_j\|_\infty
\left(1-\gamma\prod_{j=\lfloor T/2\rfloor+1}^t(1-\eta_j)\right)\\
&\leq c_t
+\frac{1}{1-\gamma}
\max_{\lfloor T/2\rfloor<j\leq T}\|\xi_j\|_\infty,
\end{align*}
where in the second inequality we use the fact
$
\frac{\gamma}{1-\gamma}(1-C)+1
=
\frac{1-\gamma C}{1-\gamma}
$
for any $C>0$. 
This completes the induction, thus establishing~\eqref{eq:Q-loose-comparison}.

\paragraph{Step 4: bounding the size of fluctuations $\xi_t$.} 
In order to make use of \eqref{eq:Q-loose-comparison}, we need to control the sizes of the fluctuation terms $\{\xi_t\}$.
For each fixed $t$, the summands in $\xi_t = \gamma\sum_{i=\lfloor T/2\rfloor+1}^t\eta_i^{(t)}(P_i-P)V^\star $ form a martingale difference sequence. Since $0\leq V^\star\leq(1-\gamma)^{-1}1_S$, it follows that 
$$ \gamma \big| \eta_i^{(t)}(P_i-P)V^\star \big| \leq  \gamma\frac{\eta_i^{(t)}}{(1-\gamma)}. $$ 
Applying the Azuma-Hoeffding inequality (see Lemma~\ref{lem:azuma-hoeffding}) to each coordinate of $\xi_t$, followed by a union bound over all $SA$ coordinates and all $\lfloor T/2\rfloor<t\leq T$, we can demonstrate that with probability at least $1-\delta$,
\begin{align}
\|\xi_t\|_\infty
&\leq\frac{\gamma}{1-\gamma}
\sqrt{2\sum_{i=\lfloor T/2\rfloor+1}^t\big(\eta_i^{(t)}\big)^2\log\frac{4SAT}{\delta}}\nonumber\\*
&\leq\frac{\gamma\sqrt{2\overline{\eta}\log\frac{4SAT}{\delta}}}{1-\gamma}
\qquad\text{for all }\lfloor T/2\rfloor<t\leq T,
\label{eq:UB-xi-t-analysis-Q-generative}
\end{align}
where the last inequality follows from Lemma~\ref{lemma:Q-step-size}.

\paragraph{Step 5: bounding the term $c_T$.}
To finish the proof, it remains to bound $c_T$ in~\eqref{eq:Q-loose-comparison}. By~\eqref{eq:Q-warmup-radius} and the elementary inequality $1-x\leq e^{-x}$, one has
\begin{align}
c_T\leq\frac{1}{1-\gamma}
\exp\Bigg\{-(1-\gamma)\sum_{i=\lfloor T/2\rfloor+1}^T\eta_i\Bigg\}.
\label{eq:cT-exponential-estimate-Q}
\end{align}
From the sample-size condition~\eqref{eq:thm:infinite-horizon-T}, it is easy to verify that
$\log T\geq4$ and $(1-\gamma)T\geq\log^3T$. 
Consequently, taking $c_1\leq1$, the learning rate bound in~\eqref{eq:thm:infinite-horizaon-eta} implies that
\[
\eta_i\geq\frac{1}{1+c_1(1-\gamma)T/\log^3T}
\geq\frac{\log^3T}{2(1-\gamma)T},
\qquad \lfloor T/2\rfloor<i\leq T,
\]
which in turn yields
\[
(1-\gamma)\sum_{i=\lfloor T/2\rfloor+1}^T\eta_i
\geq(1-\gamma)\frac{T}{2}\cdot\frac{\log^3T}{2(1-\gamma)T}
=\frac14\log^3T\geq4\log T.
\]
Substitution into \eqref{eq:cT-exponential-estimate-Q} leads to
$$c_T\leq \frac{1}{(1-\gamma)T^4}.$$

\paragraph{Step 6: putting all this together.} 
With the above bound on $c_T$ in place, we can combine it with \eqref{eq:UB-xi-t-analysis-Q-generative} and~\eqref{eq:Q-loose-comparison} to conclude that, with probability at least $1-\delta$,
\begin{align}
\|Q_T-Q^\star\|_\infty
&\leq\frac{1}{(1-\gamma)T^4}
+\frac{\gamma\sqrt{2\overline{\eta}\log\frac{4SAT}{\delta}}}{(1-\gamma)^2} \arxiv{\nonumber\\
&}\lesssim\sqrt{\frac{(\log^3T)\big(\log\frac{SAT}{\delta}\big)}{(1-\gamma)^5T}}.
\label{eq:QT-gamma-5-Q-learning-generative}
\end{align}
This in turn establishes the iteration complexity bound $\widetilde{O}\big(\frac{1}{(1-\gamma)^5\varepsilon^2}\big)$.

\paragraph{Discussion.} 
As alluded to previously, the bound \eqref{eq:QT-gamma-5-Q-learning-generative}, while admitting a concise proof, is loose by a factor of $\frac{1}{1-\gamma}$. The gap stems from the use of the Azuma–Hoeffding inequality in controlling $\xi_t$, which relies solely on coarse magnitude bounds on the summands and fails to capture the variance structure inherent in the discounted infinite-horizon MDP.  To obtain sharper guarantees, it is therefore essential to exploit this variance structure through more refined martingale concentration techniques, such as the Freedman inequality (see Lemma~\ref{lem:Freedman-prelim}). However, even after incorporating this refinement, significant challenges remain: the resulting recursive relation becomes considerably more intricate and requires new analytical ideas to handle. These developments are the focus of the next subsection.

\subsection{Sharp analysis: proof of Theorem~\ref{thm:infinite-horizon-simple}}
\label{sec:sharp-analysis-Q-generative}

We now present the full proof of Theorem~\ref{thm:infinite-horizon-simple}, building in part on the ideas developed in the suboptimal analysis in Section~\ref{sec:proof-Q-learning-loose-generative}. A key new idea is
to introduce auxiliary comparison functions inspired by
distributionally robust Bellman operators --- a topic that we will revisit in detail in Section~\ref{sec:DRL} --- with uncertainty sets carefully calibrated to capture the variance of the sampling noise.

\paragraph{Preliminaries: distributional robustness.}
Before embarking on the analysis of Q-learning, we first single out some properties of distributionally robust RL, which may seem unrelated at first glance but will play a pivotal role in our sharp analysis. In contrast to the standard Bellman operator, which uses a single nominal transition kernel, a distributionally robust Bellman operator optimizes the expected value over a prescribed uncertainty set of
plausible transition kernels. The following lemma characterizes how a $\chi^2$-based uncertainty set induces a variance-dependent deviation from the nominal Bellman update. The proof is deferred to Section~\ref{appendix:proof-lemma-Q-comparison-facts}.

\begin{lemma}\label{lemma:Q-comparison-facts}
For any $p\in\Delta(\cS)$ and $\rho>0$, define
\begin{equation}
\mathcal{U}_{\rho}(p)\coloneqq
\left\{q\in\Delta(\cS):\ \chi^2(q\,\|\,p)\leq\rho,~~q(s)=0\text{ whenever }p(s)=0\right\},
\label{eq:Q-comparison-ball}
\end{equation}
where $\chi^2(\cdot\,\|\,\cdot)$ is the $\chi^2$ divergence defined in \eqref{eq:defn-chi2}.
The following properties hold.
\begin{enumerate}
    \item[(i)] For any $V\in\mathbb{R}^S$ satisfying
    $\max_s V(s)-\min_s V(s)\leq 1/(1-\gamma)$, one has
    \begin{equation}\label{eq:Q-comparison-support}
    \sqrt{\rho\mathsf{Var}_p(V)}-\frac{\rho}{1-\gamma}
    \leq\sup_{q\,\in\,\mathcal{U}_{\rho}(p)}q^{\top}V-p^{\top}V
    \leq\sqrt{\rho\mathsf{Var}_p(V)}.
    \end{equation}
    The same bounds hold for $p^{\top}V-\inf_{q\,\in\,\mathcal{U}_{\rho}(p)}q^{\top}V$.

    \item[(ii)] Consider an MDP with the same rewards and discount factor as the original MDP, but with transition kernel $P'$ obeying
    $P'(\cdot\mymid s,a)\in\mathcal{U}_{\rho}\big(P(\cdot\mymid s,a)\big)$
    for every $(s,a)\in\cS\times\cA$.
    Its optimal Q-function, denoted by $Q^{\star,P'}$, satisfies
    \begin{equation}
    \big\|Q^{\star,P'}-Q^{\star}\big\|_{\infty}
    \leq\sqrt{\frac{\rho\gamma}{2(1-\gamma)^3}}.
    \label{eq:Q-comparison-stability}
    \end{equation}
\end{enumerate}
\end{lemma}

In words, part (i) of Lemma~\ref{lemma:Q-comparison-facts} shows that the largest change in the expected value over a $\chi^2$-uncertainty set is controlled by the standard deviation, while part (ii) shows that the corresponding perturbation changes the optimal Q-function by only a controlled amount.  As we will shortly see, these two properties allow us to construct auxiliary comparison functions that
capture the variance-dependent fluctuations while remaining close to
the optimal Q-function $Q^\star$.

\paragraph{Step 1: constructing comparison functions based on robust Bellman operators.}
We now leverage the above types of uncertainty sets to construct auxiliary comparison functions, whose roles will become clear momentarily. As detailed below, we
employ $\chi^2$-based perturbations of the transition kernels to capture
the variance-dependent fluctuations, while using a constant reward shift
to provide additional slack.

More precisely, set 
\begin{equation}
\rho\coloneqq8\overline{\eta}\log\frac{4SAT}{\delta} \qquad \text{and}\qquad b\coloneqq\frac{2\rho}{1-\gamma}, 
\label{eq:defn-rho-b-model-free-proof}
\end{equation}
where $\overline{\eta}$ is defined in~\eqref{eq:Q-comparison-parameters}. 
For any $Q:\cS\times\cA\to\mathbb{R}$, let
$V_Q(s)\coloneqq\max_{a\in\cA}Q(s,a)$ for every $s\in \cS$, and define the operators
$\mathcal{T}^+$ and $\mathcal{T}^-$ such that, for every $(s,a)\in \cS\times \cA$,
\begin{subequations}
\label{eq:Q-comparison-operator}
\begin{align}
\big[\mathcal{T}^+(Q)\big](s,a)
&\coloneqq
r(s,a)+b+
\gamma\sup_{p\,\in\,\mathcal{U}_{\rho}(P_{s,a})}p^\top V_Q,
\label{eq:Q-comparison-operator-plus} \\
\big[\mathcal{T}^-(Q)\big](s,a)
&\coloneqq
r(s,a)-b+
\gamma\inf_{p\,\in\,\mathcal{U}_{\rho}(P_{s,a})}p^\top V_Q. 
\label{eq:Q-comparison-operator-minus}
\end{align}
\end{subequations}
The lemma below shows that the robust Bellman-type operators $\mathcal{T}^+$ and $\mathcal{T}^-$ are both $\gamma$-contractions. 
\begin{lemma}
\label{lemma:robust-Bellman-contraction}
The operators $\mathcal{T}^+$ and $\mathcal{T}^-$ defined in
\eqref{eq:Q-comparison-operator} satisfy, for any
$Q,\widetilde Q:\cS\times\cA\to\mathbb{R}$,
\begin{subequations}
\begin{align}
\big\|\mathcal{T}^+(Q)-\mathcal{T}^+(\widetilde Q)\big\|_\infty
&\leq \gamma\|Q-\widetilde Q\|_\infty, \label{eq:contraction-Tplus-Q}\\
\big\|\mathcal{T}^-(Q)-\mathcal{T}^-(\widetilde Q)\big\|_\infty
&\leq \gamma\|Q-\widetilde Q\|_\infty.
\label{eq:contraction-Tminus-Q}
\end{align}
\end{subequations}
\end{lemma}
\begin{proof}[Proof of Lemma~\ref{lemma:robust-Bellman-contraction}]
For every $p\in\Delta(\cS)$, it is readily seen that
\[
\big|p^\top V_Q-p^\top V_{\widetilde Q}\big|
\leq \big\|V_Q-V_{\widetilde Q}\big\|_\infty
\leq \big\|Q-\widetilde Q\big\|_\infty.
\]
Using the elementary inequality
$$\Big|\sup\nolimits_{p\,\in\,\mathcal U}f(p)-\sup\nolimits_{p\,\in\,\mathcal U}g(p)\Big|
\leq \sup\nolimits_{p\,\in\,\mathcal U}|f(p)-g(p)|
$$
we obtain, for every $(s,a)\in\cS\times\cA$,
\[
\left|
\big[\mathcal T^+(Q)\big](s,a)
-
\big[\mathcal T^+(\widetilde Q)\big](s,a)
\right|
\leq
\gamma \big\|Q-\widetilde Q\big\|_\infty.
\]
Taking the maximum over $(s,a)$ proves~\eqref{eq:contraction-Tplus-Q}. The proof of~\eqref{eq:contraction-Tminus-Q} is analogous and is
therefore omitted for brevity. 
\end{proof}

Given the $\gamma$-contraction property of both $\mathcal{T}^+$ and $\mathcal{T}^-$, 
we can define $Q^+$ (resp.~$Q^-$) to be the {\em unique} fixed point of $\mathcal{T}^+$ (resp.~$\mathcal{T}^-$). 
Writing $V^+(s)\coloneqq\max_{a\in\cA}Q^+(s,a)$ and $V^-(s)\coloneqq\max_{a\in\cA}Q^-(s,a)$ for every $s\in \cS$, we arrive at
\begin{subequations}
\label{eq:Q-comparison-fixed-points}
\begin{align}
Q^+(s,a)
&=
r(s,a)+b+
\gamma\sup_{p\,\in\,\mathcal{U}_{\rho}(P_{s,a})}p^\top V^+, \label{eq:Q-comparison-fixed-points-plus}\\
Q^-(s,a)
&=
r(s,a)-b+
\gamma\inf_{p\,\in\,\mathcal{U}_{\rho}(P_{s,a})}p^\top V^-.
\label{eq:Q-comparison-fixed-minus}
\end{align}
\end{subequations}
As we will shortly see, the fixed points $Q^+$ and $Q^-$ will serve as
upper and lower comparison functions for the Q-learning iterates,
respectively.

Before continuing, let us record a range bound that will be useful later. Subtracting the constant $b/(1-\gamma)$ from~\eqref{eq:Q-comparison-fixed-points-plus} yields, for every $(s,a)\in \cS\times \cA$, 
\begin{align}
 Q^+(s,a)-\frac{b}{1-\gamma}
 &=r(s,a)+\gamma\sup_{p\,\in\,\mathcal{U}_{\rho}(P_{s,a})}
 \!p^\top\bigg(V^+-\frac{b}{1-\gamma}1_S\bigg).
 \label{eq:Qstar-plus-Bellman-b-shift}
\end{align}
Given that each uncertainty set is compact, we can choose a maximizer $P^+_{s,a}\in\mathcal{U}_{\rho}(P_{s,a})$ for each $(s,a)$. Let $P^+$ denote the transition kernel whose $(s,a)$-th row is $P^+_{s,a}$. With this choice,
\eqref{eq:Qstar-plus-Bellman-b-shift} is the Bellman optimality equation
for an MDP with reward function $r$ and transition kernel $P^+$. By
uniqueness of the solution to the Bellman equation, the
corresponding optimal Q-function is
 \begin{equation}
 \label{eq:defn-Q-star-Pplus-model-free}
 Q^{\star,P^+} = Q^+-\frac{b}{1-\gamma}1_{SA}.
 \end{equation}
Since the rewards lie in $[0,1]$, one immediately obtains
\begin{equation}\label{eq:Q-comparison-bounded}
 0\leq Q^+-\frac{b}{1-\gamma}1_{SA}
 \leq\frac{1}{1-\gamma}1_{SA}.
\end{equation}
Moreover, adding a constant does not change the range of a function.
Therefore,
\begin{equation}\label{eq:to_be_verified}
 \max_sV^+(s)-\min_sV^+(s)\leq\frac{1}{1-\gamma}.
\end{equation}
Applying the same argument to $\mathcal{T}^-$ yields the corresponding
range bounds for $Q^-$ and $V^-$.  Importantly, all four functions,
$Q^+$, $Q^-$, $V^+$, and $V^-$, are deterministic and do not depend on the sampled transitions.

\paragraph{Step 2: decomposing the estimation error using the comparison functions.} 
We now explain the role of the comparison function $Q^+$ in~\eqref{eq:Q-comparison-fixed-points}. We focus on
the upper bound; the lower bound follows by an analogous argument using
$Q^-$. Decompose the estimation error as
\begin{equation}
\label{eq:Q-comparison-target-error}
 \Delta_t=(Q_t-Q^+)+(Q^+-Q^\star).
\end{equation}
This decomposition separates the analysis into two parts: controlling
the Q-learning iterates relative to $Q^+$, and controlling the distance
between $Q^+$ and the true optimal Q-function. To control the latter, combining Lemma~\ref{lemma:Q-comparison-facts} with property~\eqref{eq:Q-comparison-bounded} gives
\begin{align}
\big \|Q^+-Q^\star\big\|_\infty
 &\leq\big\|Q^{\star,P^+}-Q^\star\big\|_\infty + \big\|Q^{\star,P^+}-Q^+ \big\|_\infty \nonumber \\
 &\leq\sqrt{\frac{\gamma\rho}{2(1-\gamma)^3}}
       +\frac{2\rho}{(1-\gamma)^2},
 \label{eq:Q-comparison-distance}
\end{align}
where we recall the definition of $Q^{\star,P^+}$ in~\eqref{eq:defn-Q-star-Pplus-model-free} and the choice of $b$ in~\eqref{eq:defn-rho-b-model-free-proof}. 

Now we turn to the first term in~\eqref{eq:Q-comparison-target-error}. Subtracting $Q^+$ from the update~\eqref{eq:iteration-rule-Q-infinite}, and rearranging as in
\eqref{eq:iteration-infinite}, we can demonstrate that
\begin{align*}
 Q_t-Q^+
 &=(1-\eta_t)(Q_{t-1}-Q^+)
   +\gamma\eta_tP_t(V_{t-1}-V^+) \\
 &\qquad+\gamma\eta_t(P_t-P)V^+
   -\eta_t\big(Q^+-r-\gamma PV^+\big).
\end{align*}
The final term can be understood as the Bellman residual of $Q^+$, which we denote as
\begin{equation}
\label{eq:Q-comparison-residuals}
 Q^+_{\mathsf{res}}\coloneqq Q^+-r-\gamma PV^+.
\end{equation}
For $t$ obeying
$\lfloor T/2\rfloor<t\leq T$, 
unrolling the preceding recursion  using \eqref{def:eta-i-t-simple} and~\eqref{eq:Q-weight-sum} results in
\begin{align}
\arxiv{&}  Q_t-Q^+ \arxiv{\nonumber \\}
 &=\bigg(\prod_{j=\lfloor T/2\rfloor+1}^t(1-\eta_j)\bigg)(Q_{\lfloor T/2\rfloor}-Q^+)+\gamma\sum_{i=\lfloor T/2\rfloor+1}^t\eta_i^{(t)}P_i(V_{i-1}-V^+)  \nonumber \\
 &\quad + \underbrace{\gamma\sum_{i=\lfloor T/2\rfloor+1}^t\eta_i^{(t)}(P_i-P)V^+}_{\eqqcolon\, \xi_t^+} -\bigg(1-\prod_{j=\lfloor T/2\rfloor+1}^t(1-\eta_j)\bigg)Q^+_{\mathsf{res}}.
 \label{eq:Q-comparison-decomposition}
\end{align}
Here, $\xi_t^+$ is the martingale fluctuation that reflects the randomness of the
transition samples when the deterministic comparison value function $V^+$ is
used.  Importantly, the comparison function $Q^{+}$ is constructed so that its Bellman
residual can absorb this fluctuation term $\xi_t^+$. In particular, if
\begin{align}
|\xi_t^+|\leq Q^+_{\mathsf{res}}
\label{eq:condition-fluc-Qres-absorb}
\end{align}
holds componentwise for all
$\lfloor T/2\rfloor<t\leq T$, then the last two terms in~\eqref{eq:Q-comparison-decomposition} satisfy
\[
 \xi_t^+-\bigg(1-\prod_{j=\lfloor T/2\rfloor+1}^t(1-\eta_j)\bigg)Q^+_{\mathsf{res}}
 \leq\bigg(\prod_{j=\lfloor T/2\rfloor+1}^t(1-\eta_j)\bigg)Q^+_{\mathsf{res}}.
\]
Thus, these two terms combine into a contribution weighted by the initial contraction factor. The remaining propagated
error can then be controlled by an induction analogous to the one in
Section~\ref{sec:proof-Q-learning-loose-generative}. We verify next
that $Q^+_{\mathsf{res}}$ indeed dominates $\xi_t^+$.

\paragraph{Step 3: controlling the fluctuation term.}
We next verify the condition
\eqref{eq:condition-fluc-Qres-absorb}. The key observation is that:
since $V^+$ is deterministic, the fluctuation term $\xi_t^+$ is a
martingale sum with a variance that can be expressed directly in terms
of $\mathsf{Var}_{P}(V^+)$. We then show that the Bellman residual of
$Q^+$ is large enough to dominate this fluctuation.

More precisely, for each fixed $t$, the quantity $\xi_t^+$ is a sum of martingale differences, which can be bounded via Freedman's inequality (cf.~Lemma~\ref{lem:Freedman-prelim}). For a fixed coordinate $(s,a)$, write
$$
\xi_t^+(s,a)=\sum_{i=\lfloor T/2\rfloor+1}^t Z_i^+,\qquad \text{with }Z_i^+\coloneqq\gamma\eta_i^{(t)}\big[(P_i-P)V^+ \big](s,a).
$$
As before, let $\mathbb{E}_{i}$ denote conditional expectation given the
history through iteration $i$. 
Since $V^+$ is deterministic and $\mathbb E_{i-1}[P_i]=P$, it follows that 
\begin{align*}
\mathbb E_{i-1}[Z_i^+]&=0 , \qquad  |Z_i^+| \leq\frac{\gamma\overline{\eta}}{1-\gamma},\\
 \sum_{i=\lfloor T/2\rfloor+1}^t\mathsf{Var}_{i-1}(Z_i^+)
 &=\gamma^2\sum_{i=\lfloor T/2\rfloor+1}^t\big(\eta_i^{(t)}\big)^2
       \mathsf{Var}_{P_{s,a}}(V^+) \arxiv{\\
&}\leq\gamma^2\overline{\eta}\,\mathsf{Var}_{P_{s,a}}(V^+).
\end{align*}
Here, the bound on $|Z_i^+|$ follows because each coordinate of
$(P_i-P)V^+$ is bounded by the range of $V^+$, which is at most
$\frac{1}{1-\gamma}$ by~\eqref{eq:to_be_verified}. The last inequality arises from~\eqref{eq:Q-comparison-weights}. 
Applying Freedman's inequality (Lemma~\ref{lem:Freedman-prelim}) with
$K_0=1$, followed by a union bound over all $SA$ coordinates and all $\lfloor T/2\rfloor<t\leq T$, we can show that, with probability at least $1-\delta/2$,
\begin{equation}\label{eq:Q-comparison-noise-bound}
 |\xi_t^+|\leq\gamma\sqrt{\rho\,\mathsf{Var}_P(V^+)}
       +\frac{\gamma \rho}{6(1-\gamma)}1_{SA}, \qquad\text{for }\lfloor T/2\rfloor<t\leq T,
\end{equation}
where we recall the choice of $\rho$ in~\eqref{eq:defn-rho-b-model-free-proof}.

It remains to show that $Q^+_{\mathsf{res}}$ dominates the right-hand side of~\eqref{eq:Q-comparison-noise-bound}. Substituting~\eqref{eq:Q-comparison-fixed-points} into the definition of $Q^+_{\mathsf{res}}$ gives, for every $(s,a)$,
\[
 Q^+_{\mathsf{res}}(s,a)=b+\gamma\bigg(
 \sup_{p\,\in\,\mathcal{U}_{\rho}(P_{s,a})}p^\top V^+-P_{s,a}V^+\bigg).
\]
Applying part (i) of Lemma~\ref{lemma:Q-comparison-facts},  and using 
$b=\frac{2\rho}{1-\gamma}$ together with the range bound
\eqref{eq:to_be_verified},  we reach
\begin{align}
\gamma\sqrt{\rho\,\mathsf{Var}_P(V^+)}
 +\frac{(2-\gamma)\rho}{1-\gamma}1_{SA} \leq Q^+_{\mathsf{res}}&\leq\frac{2\rho +\sqrt{\rho}}{1-\gamma}1_{SA},
 \label{eq:Q-comparison-residual-lower-upper}
\end{align}
where for the upper bound, we used $\mathsf{Var}_{P_{s,a}}(V^+)\leq(1-\gamma)^{-2}$ and $\gamma\leq1$. Since $2-\gamma\geq\gamma/6$, combining~\eqref{eq:Q-comparison-noise-bound} and~\eqref{eq:Q-comparison-residual-lower-upper}, we conclude that
\begin{equation}\label{eq:Q-comparison-event}
 |\xi_t^+|\leq Q^+_{\mathsf{res}}\qquad\text{for all }\lfloor T/2\rfloor<t\leq T
\end{equation}
holds with probability at least $1-\delta/2$. This establishes the absorption condition~\eqref{eq:condition-fluc-Qres-absorb} discussed in Step~2.

\paragraph{Step 4: propagating the comparison bound through Q-learning iterates.}
We now use the residual bound from Step~3 to control the Q-learning
iterates relative to the comparison function $Q^+$. Substituting
\eqref{eq:Q-comparison-event} into
\eqref{eq:Q-comparison-decomposition} yields, componentwise, 
\begin{align}
  Q_t-Q^+  
 &\leq\bigg(\prod_{j=\lfloor T/2\rfloor+1}^t(1-\eta_j)\bigg)\big(Q_{\lfloor T/2\rfloor}-Q^{+}+Q^+_{\mathsf{res}}\big) \arxiv{\nonumber \\}
 \arxiv{&\qquad} +\gamma\sum_{i=\lfloor T/2\rfloor+1}^t\eta_i^{(t)}P_i(V_{i-1}-V^+) .
 \label{eq:Q-comparison-absorbed}
\end{align}
The residual has therefore been absorbed into the initial term, leaving
a recursion of the same form as the one analyzed in
Section~\ref{sec:proof-Q-learning-loose-generative}. Moreover, by
Lemma~\ref{lemma:non-negativity-Qt-Vt}, $Q_{\lfloor T/2\rfloor}$ is
bounded above by $(1-\gamma)^{-1}1_{SA}$, while $Q^+\geq0$. Combining
these facts with the upper bound in
\eqref{eq:Q-comparison-residual-lower-upper} gives
\[
Q_{\lfloor T/2\rfloor}-Q^++Q^+_{\mathsf{res}}
\leq
\frac{1+2\rho+\sqrt{\rho}}{1-\gamma}1_{SA}.
\]
To propagate this bound through the remaining recursion, we find it convenient to define
\begin{equation}\label{eq:Q-comparison-radius}
\widetilde c_t
\coloneqq
\frac{1+2\rho+\sqrt{\rho}}{1-\gamma}
\prod_{i=\lfloor T/2\rfloor+1}^t
\big(1-\eta_i(1-\gamma)\big),
\qquad
\lfloor T/2\rfloor\leq t\leq T,
\end{equation}
where an empty product equals one. Applying the same induction used to establish
\eqref{eq:Q-loose-comparison} in
Section~\ref{sec:proof-Q-learning-loose-generative} --- now with
\eqref{eq:Q-comparison-absorbed} as the recursive inequality and
$\widetilde c_t$ as the corresponding radius --- and using
\eqref{eq:Q-weight-sum} together with $\|P_i\|_1=1$, we arrive at
\begin{equation}\label{eq:Q-comparison-sandwich}
Q_t\leq Q^++\widetilde c_t1_{SA},
\qquad
\lfloor T/2\rfloor\leq t\leq T.
\end{equation}
Thus, the Q-learning iterates are controlled by the comparison function $Q^+$ up to a term that contracts at rate
$1-\eta_i(1-\gamma)$. 
Combining \eqref{eq:Q-comparison-distance} and
\eqref{eq:Q-comparison-sandwich} with
\eqref{eq:Q-comparison-target-error} gives
\begin{align}
\Delta_T
&\leq
\frac{1+2\rho+\sqrt{\rho}}{1-\gamma}
\prod_{i=\lfloor T/2\rfloor+1}^T
\big(1-\eta_i(1-\gamma)\big)1_{SA}
\nonumber\\
&\quad+
\left(
\sqrt{\frac{\gamma\rho}{2(1-\gamma)^3}}
+\frac{2\rho}{(1-\gamma)^2}
\right)1_{SA}.
\label{eq:Delta-UB-final}
\end{align}

Additionally, applying the same argument to the lower comparison function $Q^-$ yields
an upper bound on $-\Delta_T$ of the same form. Combining the resulting upper and lower bounds, we conclude that, with probability at least $1-\delta$,
\begin{align}
\|\Delta_T\|_\infty
&\leq
\frac{1+2\rho+\sqrt{\rho}}{1-\gamma}
\prod_{i=\lfloor T/2\rfloor+1}^T
\big(1-\eta_i(1-\gamma)\big)
\nonumber\\
&\quad+
\sqrt{\frac{\gamma\rho}{2(1-\gamma)^3}}
+\frac{2\rho}{(1-\gamma)^2}.
\label{eq:Delta-final}
\end{align}

\paragraph{Step 5: completing the proof.}
We now verify that the right-hand side of
\eqref{eq:Delta-final} is at most $\varepsilon$ under the sample-size
condition~\eqref{eq:thm:infinite-horizon-T}.

First, we bound the two terms involving $\rho$. Since $SAT/\delta\geq2$,
\[
\log\frac{4SAT}{\delta}
\leq 3\log\frac{SAT}{\delta}.
\]
Therefore, the choice~\eqref{eq:defn-rho-b-model-free-proof} of $\rho$, together with
\eqref{eq:Q-comparison-parameters} and the sample-size condition
\eqref{eq:thm:infinite-horizon-T}, gives
\begin{align}
\rho
&\leq
\frac{48(\log^3T)\log\frac{SAT}{\delta}}
{c_2(1-\gamma)T}
\leq
\frac{48\varepsilon^2(1-\gamma)^3}
{c_2c_3\log T}.
\label{eq:Q-comparison-theta-bound}
\end{align}
Consequently,
\begin{align*}
\sqrt{\frac{\gamma\rho}{2(1-\gamma)^3}}
&\leq
2\varepsilon\sqrt{\frac{6}{c_2c_3\log T}}
\leq \frac{\varepsilon}{4},\\
\frac{2\rho}{(1-\gamma)^2}
&\leq
\frac{96\varepsilon^2(1-\gamma)}
{c_2c_3\log T}
\leq \frac{\varepsilon}{4},
\end{align*}
provided that the universal constants are chosen sufficiently large.

It remains to bound the first term on the right-hand side of
\eqref{eq:Delta-final}. Comparing
\eqref{eq:Q-comparison-radius} with~\eqref{eq:Q-warmup-radius} gives
\[
\widetilde c_T
=
(1+2\rho+\sqrt{\rho})c_T.
\]
The preceding estimates imply $\rho\leq1/16$, and hence
\[
1+2\rho+\sqrt{\rho}\leq2.
\]
Using the bound on $c_T$ established at the end of
Section~\ref{sec:proof-Q-learning-loose-generative}, we obtain
\[
\widetilde c_T
\leq 2c_T
\leq \frac{2}{(1-\gamma)T^4}
\leq \frac{\varepsilon}{2}.
\]
Here, the last inequality makes use of $T\geq2$ and
\[
\varepsilon\geq\frac{1}{(1-\gamma)\sqrt{T}},
\]
which follows from the sample-size condition~\eqref{eq:thm:infinite-horizon-T}, provided that the constant
$c_3$ is sufficiently large. Combining the three bounds with
\eqref{eq:Delta-final} yields
\[
\|Q_T-Q^\star\|_\infty\leq\varepsilon
\]
with probability at least $1-\delta$, thereby completing the proof.

\paragraph{Discussion: over-estimation vs.~under-estimation.}
Before concluding this subsection, we highlight an important difference between over-estimation and under-estimation in Q-learning. Consider an auxiliary TD-type iteration $\widetilde Q_t$ that employs the same initialization and transition samples as Q-learning, but follows a  deterministic optimal policy $\pi^\star$ instead of maximizing over actions. Since $Q^{\pi^\star}=Q^\star$, this auxiliary iteration evaluates the optimal policy. Moreover, because Q-learning maximizes over actions, one can anticipate that $\widetilde Q_t\leq Q_t$. Hence, the lower deviation of Q-learning can be bounded below by that of the auxiliary TD-type iteration. The TD-learning analysis gives
\begin{align*}
Q_T-Q^\star \geq \widetilde Q_T-Q^\star
&\geq
-\widetilde{O}\left(
\sqrt{\frac{1}{(1-\gamma)^3T}}
\,\right)1_{SA},
\end{align*}
which is smaller in magnitude than the leading square-root term in the corresponding over-estimation bound by a factor of $1/\sqrt{1-\gamma}$. In other words, over-estimation poses a more significant issue in Q-learning than under-estimation, an asymmetry  reflected both in the proof and in practice.

\section{Extension: asynchronous Q-learning}
\label{sec:async-Q-generative-model}

As noted previously, the Q-learning algorithm presented in Algorithm~\ref{alg:q-infinite} is commonly referred to as {\em synchronous} Q-learning, reflecting the fact that the Q-values for all state–action pairs are updated simultaneously at each iteration. Another important practical variant is {\em asynchronous} Q-learning \citep{tsitsiklis1994asynchronous,jaakkola1993convergence}, in which only a single state–action pair is updated per iteration (typically following a Markovian trajectory, to be made precise shortly). Although the sampling mechanism in asynchronous Q-learning often deviates substantially from that of a generative model, its core algorithmic ideas and theoretical guarantees remain closely aligned with those of the synchronous counterpart. For this reason, we introduce this variant in this chapter as a natural extension of synchronous Q-learning.

%Moving beyond the synchronous setting, another scenario of practical importance is the case where the acquired samples take the form of a single Markovian trajectory \citep{tsitsiklis1994asynchronous}.   In this section, we extend our analysis framework for synchronous Q-learning to accommodate Markovian non-i.i.d.~samples.   

\subsection{Learning from Markovian samples} 

In contrast to the generative model setting, asynchronous Q-learning is most commonly applied in settings where samples are collected along a Markovian trajectory. To be precise, suppose that we have access to a sample trajectory $\{(s_t, a_t,r_t)\}_{t=0}^{\infty}$, generated by the MDP of interest $\mathcal{M} = \{\cS, \cA, P, r, \gamma\}$ under a stationary {\em behavior policy} $\pib$ --- so called because this policy dictates the actual ``behavior'' in generating this sample trajectory. Starting from an initial state $s_0$, this trajectory evolves sequentially as
\begin{equation}
	a_t \sim \pib(\cdot\mymid s_t),  \quad r_t = r(s_t,a_t),\quad s_{t+1} \sim P(\cdot\mymid s_t, a_t),\quad t\geq 0; 
    \label{eq:Markov-chain-behavior-asynQ}
\end{equation}
see Figure~\ref{fig:async_trajectory} for an illustration. 
When $\pib$ is stationary, the process $\{(s_t,a_t)\}_{t=0}^{\infty}$ can be viewed as a sample path of a time-homogeneous Markov chain. Notably, the behavior policy $\pib$ may differ substantially from the target optimal policy $\pi^{\star}$. Throughout this section, we assume that asynchronous Q-learning operates on this single Markovian trajectory, 
with sample complexity naturally measured by the length of the observed trajectory required to achieve a prescribed level of accuracy.

\subsection{Algorithm: asynchronous Q-learning}

%width=0.9\textwidth

\begin{figure}
\begin{center}
\includegraphics[width=0.8\textwidth]{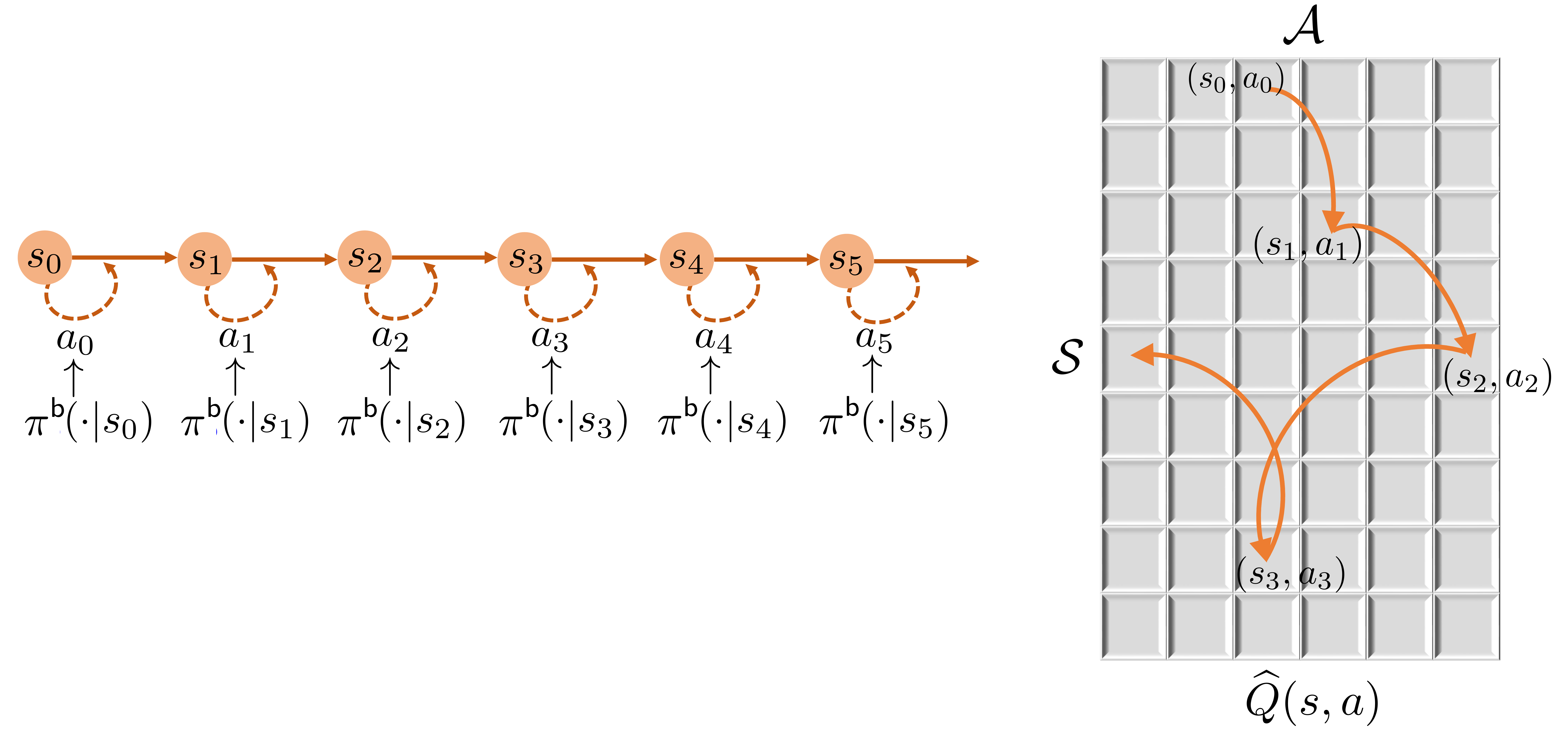} 
\end{center}	
	\caption{In asynchronous Q-learning, one observes a single Markovian trajectory $\{(s_t,a_t)\}_{t\geq 0}$ induced by a fixed behavior policy $\pi^{\mathsf{b}}$, and updates the Q-estimate sequentially according to it.\label{fig:async_trajectory}}
\end{figure}

We now introduce the asynchronous Q-learning algorithm, in which only the state–action pair visited along the Markovian trajectory is updated at each iteration.  Specifically, given the Markovian trajectory $\{s_t, a_t,r_t\}_{t=0}^{\infty}$ generated by the behavior policy $\pi^{\mathsf{b}}$, the asynchronous Q-learning algorithm maintains a Q-function estimate $Q_t: \cS\times \cA \rightarrow \mathbb{R}$ at each time $t\geq 0$ and  updates only one entry at each iterate as follows:
\begin{equation}
\label{eqn:async-q-learning}
\begin{split}
	Q_t(s_{t-1},a_{t-1}) &= (1- \eta_t ) Q_{t-1}(s_{t-1},a_{t-1}) + \eta_t \mathcal{T}_t (Q_{t-1}) (s_{t-1}, a_{t-1}),  \\
Q_t(s ,a) & = Q_{t-1}(s ,a ), \qquad \forall (s,a)\neq (s_{t-1},a_{t-1}),
\end{split}
\end{equation}
whereas $\eta_t\in (0,1)$ represents the learning rate, and $\mathcal{T}_t$ denotes the empirical Bellman optimality operator w.r.t.~the $t$-th sample transition $(s_{t-1},a_{t-1},s_t)$ of the Markovian trajectory, defined as 
\begin{align}
	\mathcal{T}_t (Q) (s_{t-1}, a_{t-1}) \coloneqq  
		r(s_{t-1}, a_{t-1}) + \gamma \max_{a' \in \cA} Q(s_t, a').
		%, \quad & \text{if }(s,a)=(s_{t-1}, a_{t-1}) .
	\label{defn:empirical-Bellman-t}
\end{align}
It is self-evident that the term $\mathcal{T}_t (Q) (s_{t-1}, a_{t-1})$ at iteration $t$ can be viewed as a stochastic, unbiased estimate of the deterministic quantity $ \mathcal{T} (Q_{t-1}) (s_{t-1}, a_{t-1})$, 
where $ \mathcal{T}(\cdot)$ is the Bellman optimality operator defined in \eqref{eq:bellman_optimality_operator}. Consequently, the update of $Q_t(s_{t-1}, a_{t-1})$ follows the paradigm of stochastic approximation, and asynchronous Q-learning can also be interpreted as a stochastic approximation scheme for computing the fixed point of the Bellman optimality operator. The corresponding value function estimate $V_t: \cS \rightarrow \mathbb{R}$  at time $t$ is again given by
\begin{align*}
	\forall s\in \cS: \qquad V_t(s) :=  \max_{a\in \cA} Q_{t}(s,a) .
\end{align*}

It is worth emphasizing that, at each time $t$, only a single entry --- the one corresponding to the sampled state-action pair $(s_{t-1},a_{t-1})$ --- is updated, while all other entries remain unchanged; see Figure~\ref{fig:async_trajectory} for an illustration. At a high level, this procedure resembles coordinate descent in optimization, with coordinates selected according to a Markov chain \citep{sun2020markov}. In addition, although the initial estimate $Q_0$ can be chosen arbitrarily, we often set $Q_0(s,a) = 0$ for all $(s,a)$ unless otherwise noted.  The complete asynchronous Q-learning algorithm is summarized in Algorithm~\ref{alg:async-q}.

% \paragraph{Goal.} Given {\em a single} sample trajectory $\{s_t, a_t,r_t\}_{t=0}^{\infty}$ generated by the behavior policy $\pi_{\mathsf{b}}$, we aim to compute/approximate the optimal Q-function $Q^{\star}$ in an $\ell_\infty$ sense.  This setting --- in which a state-action pair can be updated only when the Markovian trajectory reaches it~--- is commonly referred to as {\em asynchronous} Q-learning \citep{tsitsiklis1994asynchronous} in tabular RL.

%\begin{algorithm}[t]
% \caption{Asynchronous Q-learning} \label{alg:async-vr-q} 
% \begin{algorithmic}[1] 
% \STATE \textbf{Input parameter:} learning rate $\eta$.
% \STATE \textbf{Initialization:} $Q_0(s,a)=0$ for all $(s,a)\in \cS\times \cA$. 
% \FOR{$t=1,2,\dots,T$}
% \STATE Draw an action $a_{t-1} \sim \pi(\cdot |s_{t-1})$ and the next state $s_{t}\sim P(\cdot|s_{t-1},a_{t-1})$ from the MDP;
% \STATE{Update the Q-function $Q_t$ according to \eqref{eqn:q-learning}.}
%  \ENDFOR
% \end{algorithmic} 
% \end{algorithm}

\begin{algorithm}[t]
    \DontPrintSemicolon
	\textbf{input:} discount factor $\gamma$, reward function $r$, learning rates $\{\eta_t\}$, number of iterations $T$, initialization $Q_0$. \\
   \For{$t=1,2,\cdots,T$} {
	draw action $a_{t-1} \sim \pi_{\mathsf{b}}(s_{t-1})$, observe reward $r(s_{t-1},a_{t-1})$, and draw next state $s_{t}\sim P(\cdot\,|\,s_{t-1},a_{t-1})$.\\
		%\blue{O.} \\ 
    {\color{blue}\tcc{stochastic approximation.}}
	 update $Q_t$ according to
 \begin{align*}
	Q_t(s_{t-1},a_{t-1}) &= (1- \eta_t ) Q_{t-1}(s_{t-1},a_{t-1}) + \eta_t \mathcal{T}_t (Q_{t-1}) (s_{t-1}, a_{t-1}),  \\
Q_t(s ,a) & = Q_{t-1}(s ,a ), \qquad \forall (s,a)\neq (s_{t-1},a_{t-1}),
\end{align*}    
where the  $(s_{t-1}, a_{t-1})$-th entry of $\mathcal{T}_t (Q)$ is given by
\begin{align*}
	\mathcal{T}_t (Q) (s_{t-1}, a_{t-1}) =  
		r(s_{t-1}, a_{t-1}) + \gamma \max_{a' \in \cA} Q(s_t, a').
		%, \quad & \text{if }(s,a)=(s_{t-1}, a_{t-1}) .
\end{align*}
}

	%\textbf{output} $Q_T$.

    \caption{Asynchronous Q-learning}
 \label{alg:async-q}
\end{algorithm}

\subsection{Assumptions and key quantities}
In contrast to the generative model setting --- where one can obtain the same number of samples across all state–action pairs --- it is generally unclear whether a Markovian sample trajectory will visit every state–action pair sufficiently often. Since insufficient visitation may result in a lack of critical information, we impose a standard assumption and introduce several key quantities, commonly used in the analysis of asynchronous Q-learning, to ensure adequate coverage of the state–action space by the sample trajectory.

Recall that when the behavior policy $\pi^{\mathsf{b}}$ is stationary, the sequence $\{(s_t, a_t)\}_{t=0}^{\infty}$ forms a time-homogeneous Markov chain; in what follows, we denote by $\mu_{\pib}$ its stationary distribution.  The following assumption concerns the uniform ergodicity of this chain. 
\begin{assumption}
\label{assump:ergodicity-asyncQ}
	The Markov chain $\{(s_t, a_t)\}_{t=0}^{\infty}$ induced by the behavior policy $\pib$ (see \eqref{eq:Markov-chain-behavior-asynQ}) is assumed to be uniformly ergodic. 
\end{assumption}
At a high level, uniform ergodicity ensures that, regardless of the initial state, the Markov chain eventually ``forgets'' its starting point and converges to the stationary distribution $\mu_{\pib}$ at a geometric rate. Consequently, aside from an initial burn-in period, the state–action visitation frequencies can be well approximated by the stationary distribution $\mu_{\pib}$. See \citet[Section~1.2]{paulin2015concentration} for a formal definition of uniform ergodicity and its further implications.

With uniform ergodicity in place, two key quantities characterize, respectively,  the information bottleneck in the stationary distribution and the rate of convergence toward it, both of which are critical in determining the performance of asynchronous Q-learning. 
\begin{itemize}
\item The first key quantity is the {\em minimum state–action occupancy probability} of the sample trajectory, defined formally as
\begin{equation}
	\label{defn:mu-min}
	\mumin \defn \min_{(s,a)\in \cS\times \cA} \mu_{\pib}(s,a). 
\end{equation}
This metric captures the information bottleneck induced by the least-visited state-action pair along the trajectory. In particular, a smaller 
$\mumin$ 
indicates poorer coverage of the state-action space and, consequently, a higher chance that certain state-action pairs are insufficiently visited, which can adversely affect learning performance.

\item The second key quantity is the {\em mixing time} associated with the sample trajectory, defined and denoted by
\begin{equation}
	\label{defn-tmix}
	\tmix 
	\defn \min \bigg\{ t \,\,\Big| \max_{(s,a)\in \cS\times \cA} \mathsf{TV}\big( P^t(\cdot\mymid s,a), \mu_{\pib} \big) \leq \frac{1}{4} \bigg\},
\end{equation}
where we recall that $\mathsf{TV}(\cdot,\cdot)$ indicates the total-variation distance between two probability measures. Here, $P^t(\cdot\mymid s,a)$ represents the distribution of $(s_t,a_t)$ when the sample trajectory is initialized at $(s_0,a_0)=(s,a)$. 
In words, the mixing time characterizes how quickly the Markov chain can ``forget'' its initial state, quantifying (the order of) the time required for the distribution of the trajectory to approach its stationary distribution.

\end{itemize}

\subsection{Theoretical guarantees}

With the above assumption and key quantities in place, we are now ready to present the state-of-the-art theoretical guarantees for asynchronous Q-learning, originally established by \cite{li2024q}. For simplicity, the theorem below adopts a time-invariant learning rate schedule given by
\begin{subequations}
\begin{equation}
\eta_t \equiv \eta = \frac{c_1\log^3 T}{(1-\gamma) T\mumin} ,\qquad 0\leq t\leq T \label{eq:thm:infinite-horizaon-eta-asyn}
\end{equation}
for some universal constants $0<c_{1}\le 1$.

\begin{theorem}[\cite{li2024q}]
\label{thm:infinite-horizon-asyn}
Consider any $\delta\in(0,1)$,  $\varepsilon\in(0,1]$, and $\gamma\in [1/2,1)$. 
\label{eq:thm-infinite-horizon-condition-asyn}
%. 
Suppose that Assumption~\ref{assump:ergodicity-asyncQ} holds, and that the total number of iterations $T$ exceeds
\begin{equation}
	T\ge\frac{c_{2} \log^2\big(\frac{SAT}{\delta}\big)}{\mumin}\max\bigg\{\frac{\log^{3}T}{(1-\gamma)^{4}\varepsilon^{2}}, \frac{t_{\mathsf{mix}}}{1-\gamma}\bigg\}
	\label{eq:thm:infinite-horizon-T-asyn}
\end{equation}
for some sufficiently large universal constant $c_{2}>0$. 
If the initialization obeys $0\leq {Q}_{0}(s,a) \leq\frac{1}{1-\gamma}$ for all $(s,a)\in \cS\times \cA$, 
then asynchronous Q-learning (cf.~Algorithm~\ref{alg:async-q}) with learning rates \eqref{eq:thm:infinite-horizaon-eta-asyn} achieves
\[
	\max_{(s,a)\in \cS\times \cA} \big| {Q}_{T}(s,a) -  {Q}^{\star}(s,a) \big| \le\varepsilon
\]
 with probability at least $1-\delta$. 
\end{theorem}
\begin{remark}
As before, this high-probability result can be readily translated into a bound on the expected $\ell_{\infty}$-based estimation error with essentially the same sample complexity. We omit the details here and refer the reader to \cite{li2024q} for details.
\end{remark}
\end{subequations}

%\begin{remark} \label{remark:expected-error-asynQ}
%	One can immediately translate 	the above high-probability result into the following  mean estimation error bound: 
%	%
%	\begin{equation}
%		\mathbb{E}\Big[ \max_{s,a} \big| {Q}_{T}(s,a) -  {Q}^{\star}(s,a) \big| \Big] \le \varepsilon (1-\delta)+\delta \frac{1}{1-\gamma}
%		\leq 2\varepsilon,
%	\end{equation}
%	%
%	which holds as long as $T\ge\frac{c_{2} \log^2\frac{|\mathcal{S}||\mathcal{A}|T}{\varepsilon(1-\gamma)}}{\mumin}\max\Big\{\frac{\log^{3}T}{(1-\gamma)^{4}\varepsilon^{2}}, \frac{t_{\mathsf{mix}}}{1-\gamma}\Big\}$ for some large enough constant $c_2>0$. 
%\end{remark}

Given that each iteration of asynchronous Q-learning uses a single sample transition,  Theorem~\ref{thm:infinite-horizon-asyn} demonstrates that,  with high probability, the total sample size needed for asynchronous Q-learning to yield entrywise $\varepsilon$ accuracy is 
\begin{align}
	\widetilde{O}\bigg( \frac{1}{\mumin(1-\gamma)^{4}\varepsilon^{2}} + \frac{t_{\mathsf{mix}}}{\mumin(1-\gamma)} \bigg),
	\label{eq:sample-size-async-Q}
\end{align}
provided that the learning rates are suitably chosen (see~\eqref{eq:thm:infinite-horizaon-eta-asyn}). We next provide several remarks on the two terms appearing in this bound.
\begin{itemize}
\item {\em Dependency on the minimum state-action occupancy.}
The first term in \eqref{eq:sample-size-async-Q} mirrors the sample complexity of synchronous Q-learning. More specifically, recall that synchronous Q-learning achieves a sample complexity of $\widetilde{O}\big(\frac{SA}{(1-\gamma)^4\varepsilon^2}\big)$ under a generative model (cf.~\eqref{eq:sample-complexity-sync-Q-learning}). In comparison, the key distinction here is that the total number of state–action pairs (i.e., $SA$) is replaced by 
$1/ \mumin$ for asynchronous Q-learning, in order to account for the non-uniform visitation frequencies induced by the Markovian sampling process. Moreover, when the target accuracy $\varepsilon$ is sufficiently small, this term becomes the dominant contribution to the sample complexity of asynchronous Q-learning. The sharpness of this term has been further confirmed by a matching algorithm-dependent lower bound established in \citet[Theorem~5]{li2024q}. 

\item {\em Dependency on the mixing time.} 
The second term in \eqref{eq:sample-size-async-Q} is nearly independent of the target accuracy (up to logarithmic factors) and can be interpreted as a burn-in time. Roughly speaking, it reflects the time required for asynchronous Q-learning to effectively emulate its synchronous counterpart despite the temporal dependence inherent in Markovian data. Notably, the mixing time of the Markov chain seems to only influence the length of this burn-in phase and becomes largely immaterial once the trajectory is sufficiently close to its stationary distribution.

\end{itemize}

For the sake of brevity, we omit the full proof of  Theorem~\ref{thm:infinite-horizon-asyn}, and refer to the synchronous and asynchronous analyses in \citet{li2024q}. Instead, we highlight a key observation that bridges the analysis of asynchronous and synchronous Q-learning. Specifically, over any time window of length $T_0=\widetilde{\Theta}\big(\frac{t_{\mathsf{mix}}}{\mu_{\mathsf{min}}}\big)$, classical concentration inequalities for Markov chains (e.g., \citet[Theorem~3.14]{paulin2015concentration}) guarantee that each state-action pair $(s,a)$ is visited at least on the order of $$T_0\mu_{\pi^{\mathsf{b}}}(s,a)\geq T_0\mu_{\mathsf{min}}=\widetilde{\Theta}(t_{\mathsf{mix}})$$ times with high probability. Consequently, it can be anticipated that within each such window of length $\widetilde{\Theta}\big(\frac{t_{\mathsf{mix}}}{\mu_{\mathsf{min}}}\big)$, the progress made by asynchronous Q-learning is at least comparable to --- if not exceeding --- that of $\widetilde{\Theta}(t_{\mathsf{mix}})$ iterations of synchronous Q-learning. 
This observation suggests a natural coupling between the two algorithms. When the total number of iterations $T$ is large enough, the trajectory of asynchronous Q-learning can be partitioned into windows of length $\widetilde{\Theta}\big(\frac{t_{\mathsf{mix}}}{\mu_{\mathsf{min}}}\big)$, each of which can be compared to  $\widetilde{\Theta}(t_{\mathsf{mix}})$ iterations of synchronous Q-learning. Once this correspondence is established, the performance guarantees for asynchronous Q-learning follow by leveraging the analysis developed for the synchronous setting. In particular, since synchronous Q-learning requires $\widetilde{O}\big(\frac{1}{(1-\gamma)^4\varepsilon^2}\big)$ iterations to reach $\varepsilon$-accuracy, this argument suggests that asynchronous Q-learning requires $\widetilde{O}\big(\frac{1}{\mu_{\mathsf{min}}(1-\gamma)^4\varepsilon^2}\big)$ iterations to attain the same level of accuracy, provided that $T$ is large enough. We refer the interested reader to \citet{li2024q} for the complete proof.

Before concluding this section, we briefly mention several minor extensions of the above results. First, careful readers may observe that the learning rate schedule in \eqref{eq:thm:infinite-horizaon-eta-asyn} requires prior knowledge of the stopping time $T$ and, more importantly, the minimum state–action occupancy probability 
$\mu_{\mathsf{min}}$. Fortunately, this requirement can be eliminated by adopting a more data-driven, adaptive learning rate schedule without sacrificing performance; see \citet[Section~III.D]{li2021sample} for details. Moreover, it is worth emphasizing that ``mixing'' is not strictly necessary for the success of asynchronous Q-learning. The analysis can be readily extended to settings where the Markov chain underlying the sample trajectory does not mix but instead admits a finite cover time --- namely, that within any time window of length exceeding
$t_{\mathsf{cover}}$, the event that all state–action pairs are visited at least once occurs with probability at least 1/2. In this case, the analysis proceeds by partitioning the dynamics of asynchronous Q-learning into time windows of length $\widetilde{\Theta}(t_{\mathsf{cover}})$ (rather than those determined by the mixing time) and analyzing each window separately through comparisons with the synchronous counterpart. We refer the reader to \citet[Theorem~2]{li2021sample} and the accompanying discussion therein for further details.

\section{Extension: finite-horizon nonstationary MDPs}
\label{sec:finite-horizon-Q-learning}

We now briefly discuss synchronous Q-learning for finite-horizon nonstationary MDPs. Both the algorithm and its analysis closely parallel those in the discounted infinite-horizon setting. We therefore omit the details here and refer interested readers to \citet{li2021tightening-icml} for a complete treatment.

In a nutshell, the algorithm maintains a sequence of Q-function estimates $Q_{t,h}: \cS \times \cA \rightarrow \mathbb{R}$ for each iteration $t$ and time step $h$. At each iteration $t$, the algorithm updates these estimates backward from $h = H$ down to $1$ via
\begin{equation}
\label{eqn:q-learning-finite}
	Q_{t,h}  = (1- \eta_t ) Q_{t-1,h} + \eta_t \mathcal{T}_{t,h} (Q_{t, h+1}),
	%\qquad H \geq h \geq 1. 
\end{equation}
where $0<\eta_t<1$ denotes the learning rate for the $t$-th iteration, and $\mathcal{T}_{t,h}$ is the empirical Bellman operator constructed from samples drawn from the generative model at step $h$ of this iteration. More concretely, $\mathcal{T}_{t,h}$ is defined as follows:
\begin{itemize}
\item  for each tuple $(s,a,h)\in \cS\times \cA\times [H]$, draw an independent next-state sample
\[
	s_{t,h}(s,a) \overset{\mathrm{ind.}}{\sim} P_h(\cdot \mymid s,a);
\]
\item compute  $\mathcal{T}_{t,h}$ entrywise according to
 \begin{align*}
	\mathcal{T}_{t,h} (Q_{t,h+1}) (s , a ) &\coloneqq 
		 r_h(s , a ) + \max_{a' \in \cA} Q_{t,h+1}\big(s_{t,h}(s,a), a'\big) . %\label{defn:empirical-Bellman-t-finite} 
\end{align*}
\end{itemize}   
In addition, the value function estimate $V_{t,h}: \cS \rightarrow \mathbb{R}$ associated with the $t$-th iterate and step $h$ is defined as
\begin{align}
	\label{defn:Vt-finite}
	\forall s\in \cS: \qquad V_{t,h}(s) :=  \max_{a\in \cA} Q_{t,h}(s,a) .
\end{align}
To ensure the desired convergence guarantees, we also assume that the learning rates satisfy
\begin{equation}
	\frac{1}{1+\frac{c_{1}T}{H\log^{2}T}}\le\eta_{t}\le\frac{1}{1+\frac{c_{2}t}{H\log^{2}T}},\quad0\leq t\leq T
	\label{eq:thm:finite-horizaon-eta}
\end{equation}
for sufficiently small universal constants $c_{1}\geq c_{2}>0$.

The next theorem establishes the $\ell_{\infty}$-based sample complexity of this synchronous Q-learning algorithm for finite-horizon MDPs. 
\begin{theorem}[\citet{li2021tightening-icml}]
\label{thm:finite-horizon-simple-Q-generative}
Consider any $\delta\in(0,1)$ and $\varepsilon\in(0,1]$. Suppose that
the learning rates obey \eqref{eq:thm:finite-horizaon-eta} and the total number of iterations obeys
\begin{equation}
T\ge\frac{c_{3}H^{4}\big(\log^{3}T\big)\big(\log\frac{SAT}{\delta}\big)}{\varepsilon^{2}}
	\label{eq:thm:finite-horizon-T}
\end{equation}
for some sufficiently large universal constant $c_{3}>0$.  
If the initialization obeys $0\leq Q_{0,h}(s,a) \leq H+1-h$ for every $(s,a,h)\in \cS\times \cA\times [H]$, then the aforementioned synchronous Q-learning algorithm achieves
\[
	%\max_{(s,a)\in \cS\times \cA, \, 1\leq h\leq H} 
    \forall (s,a,h)\in \cS\times \cA\times [H]:\quad  
    \big| Q_{T,h}(s,a) - Q_h^{\star}(s,a) \big| \le\varepsilon
\]
with probability greater than $1-\delta$. 
\end{theorem}

Recalling that each iteration collects $SAH$ independent samples, 
Theorem~\ref{thm:finite-horizon-simple-Q-generative} indicates that synchronous Q-learning achieves a sample complexity of
\begin{equation*}
	\widetilde{O}\Big( \frac{ SA H^5}{ \varepsilon^2} \Big). 
\end{equation*}
for finite-horizon nonstationary MDPs. Notably, this sample complexity bound is a factor of $H$ larger than the minimax-optimal complexity \eqref{eq:sample-size-finite-horizon-generative} attained by the model-based approach in Section~\ref{sec:finite-horizon-generative}, revealing a suboptimality gap for Q-learning analogous to that observed in the discounted infinite-horizon setting.

%\section{Appendix A: proof of auxiliary lemmas in Section~\ref{sec:TD-analysis}}

%

\section{Appendix A: proof of auxiliary lemmas in Section~\ref{sec:model-free-policy-eval-sim}}

\subsection{Proof of Lemma~\ref{lemma:TD-bounded}} \label{appendix:proof-lemma-TV-bounded}

We would like to apply induction to establish that, for any $0\leq t \leq T$,
\begin{equation} \label{eq:TD-Vt-bounded}
	0\le V_{t}\le\frac{1}{1-\gamma} 1_S.
\end{equation}
Regarding the base case with $t=0$, this property is automatically guaranteed by our assumption. 
Next, suppose that \eqref{eq:TD-Vt-bounded} holds true for the $(t-1)$-th iteration. Recalling
that $0 \leq \eta_t \leq 1$, $0\leq r_{\pi} \leq 1_S$, $P_{t}\geq0$ and $\| P_{t}\|_{1}=1$, one can invoke the update rule \eqref{eq:iteration-rule-td} to show that
\begin{align*} 
	V_t = (1-\eta_t) V_{t-1} + \eta_t( r_{\pi} + \gamma P_t V_{t-1}) \geq 0
\end{align*}
and
\begin{align*}
	V_t &= (1-\eta_t) V_{t-1} + \eta_t( r_{\pi} + \gamma P_t V_{t-1})\\
	& \leq(1-\eta_{t})\,\|V_{t-1}\|_{\infty}1_S+\eta_{t}\big(\|r_{\pi}\|_{\infty}+\gamma\|P_{t}\|_{1}\|V_{t-1}\|_{\infty}\big)1_S\\
	& \leq(1-\eta_{t})\frac{1}{1-\gamma}1_S+\eta_{t}\Big(1+\frac{\gamma}{1-\gamma}\Big)1_S=\frac{1}{1-\gamma}1_S.
\end{align*}
This establishes \eqref{eq:TD-Vt-bounded} for the $t$-th iteration, which completes the induction argument for proving \eqref{eq:TD-Vt-bounded}.   
Moreover, taking together \eqref{eq:TD-Vt-bounded} and the fact that $0 \leq V^{\pi} \leq \frac{1}{1-\gamma}1_S$ immediately reveals that
\[
\| V_{t}- V^{\pi}\|_{\infty}\le\frac{1}{1-\gamma}.
\]

\subsection{Proof of Lemma~\ref{lemma:TD-step-size}} \label{appendix:proof-lemma-TD-step-size}

Suppose that $t \geq T / (c_2 \log T)$. We divide into two cases.
{\setlength\leftmargini {1.1em} 
\begin{itemize}
	\item For any $0\leq k \le t/2$, it follows from the definition~\eqref{def:eta-k-t} of $\eta_k^{(t)}$ as well as the condition $0< \eta_k< 1$ that
\begin{subequations}\label{eq:eta-i-t-UB12-TD}
		\begin{align}
			\eta_{k}^{(t)} & \le 
            \prod_{i=t/2+1}^t \big(1-\eta_{i}(1-\gamma)\big)\overset{\text{(i)}}{\leq}\Big(1-\frac{\log^{2}T}{2c_{1}T}\Big)^{t/2}\overset{\text{(ii)}}{\leq}\Big(1-\frac{\log^{2}T}{2c_{1}T}\Big)^{\frac{T}{2c_{2}\log T}} \notag\\
			& =\left\{ \Big(1-\frac{\log^{2}T}{2c_{1}T}\Big)^{\frac{2c_{1}T}{\log^{2}T}}\right\} ^{\frac{\log T}{4c_{1}c_{2}}}\overset{\text{(iii)}}{\leq}\frac{1}{T^{2}}.
			\label{eq:eta-i-t-UB1-TD}
		\end{align}
		Here, (i) follows from a direct consequence of \eqref{eq:thm:eta-TD}, i.e., 
		\[
			(1-\gamma)\eta_{i}\geq\frac{1-\gamma}{1+\frac{c_{1}(1-\gamma)T}{\log^{2}T}}\geq\frac{1-\gamma}{\frac{2c_{1}(1-\gamma)T}{\log^{2}T}}=\frac{\log^{2}T}{2c_{1}T}
		\]
        as long as $T\geq \frac{\log^2 T}{c_1(1-\gamma)}$ (as ensured by \eqref{eq:thm:sample-size-TD-thm} and $\varepsilon\leq1/(1-\gamma)$ for sufficiently large $c_3$); 
		(ii) is valid provided that $t \geq T / (c_2 \log T)$; and (iii) holds as long as $c_{1}c_{2}\le 1/8$ and $T\geq c_1^{-1} \log^2 T$ (again ensured by \eqref{eq:thm:sample-size-TD-thm}).

		\item When it comes to the case with $k> t/2$, we can apply the definition \eqref{def:eta-k-t} and the  condition~\eqref{eq:thm:eta-TD} to reach
		\begin{equation}
			\eta_{k}^{(t)}\le\eta_{k}\leq\frac{1}{c_{2}(1-\gamma)k/\log^{2}T}<\frac{2\log^{3}T}{(1-\gamma)T},\label{eq:eta-i-t-UB2-TD}
		\end{equation}
        where the last inequality holds since $k>t/2\geq T/(2c_2\log T)$. 
	\end{subequations}
\end{itemize} 
}

\subsection{Proof of Lemma~\ref{lemma:TD-square}}
\label{sec:proof:lemma:TD-square}
For each state $s$, the TD update gives
\[
V_k(s)
=
(1-\eta_k)V_{k-1}(s)
+\eta_k\big(r_\pi(s)+\gamma(P_kV_{k-1})(s)\big).
\]
Taking conditional expectations gives
\begin{align}
\mathbb E_{k-1}[V_k\circ V_k]
&=(1-\eta_k)V_{k-1}\circ V_{k-1}
\arxiv{\notag\\
&\quad}+\eta_k\mathbb E_{k-1}\Big[
\big(r_\pi+\gamma P_kV_{k-1}\big)
\circ
\big(r_\pi+\gamma P_kV_{k-1}\big)
\Big]
\notag\\
&\quad-\eta_k(1-\eta_k)\mathbb E_{k-1}\Big[
\big(r_\pi+\gamma P_kV_{k-1}-V_{k-1}\big)
\circ
\big(r_\pi+\gamma P_kV_{k-1}-V_{k-1}\big)
\Big],
\label{eq:variance_ddd}
\end{align}
where we have used the elementary identity
\[
\big((1-\eta)x+\eta y\big)^2
=
(1-\eta)x^2+\eta y^2-\eta(1-\eta)(y-x)^2.
\]
Since $r_\pi\leq 1_S$ and
$V_{k-1}\leq (1-\gamma)^{-1}1_S$, we have
\begin{align*}
\mathbb E_{k-1}\Big[
\big(r_\pi+\gamma P_kV_{k-1}\big)
\circ
\big(r_\pi+\gamma P_kV_{k-1}\big)
\Big]
&=
r_\pi\circ r_\pi
+2\gamma r_\pi\circ(P_\pi V_{k-1})
+\gamma^2P_\pi(V_{k-1}\circ V_{k-1})
\\
&\leq
\gamma P_\pi(V_{k-1}\circ V_{k-1})
+\frac{2}{1-\gamma}1_S,
\end{align*}
and
\begin{align*}
\mathbb E_{k-1}\Big[
\big(r_\pi+\gamma P_kV_{k-1}-V_{k-1}\big)
\circ
\big(r_\pi+\gamma P_kV_{k-1}-V_{k-1}\big)
\Big]
\geq
\gamma^2\mathsf{Var}_{P_\pi}(V_{k-1}).
\end{align*}
Substituting these bounds into~\eqref{eq:variance_ddd} and rearranging terms, we conclude the proof.

\subsection{Proof of Lemma~\ref{lemma:TD-variance-proxy-bound}}
\label{sec:proof:lemma:TD-variance-proxy-bound}

We intend to apply Freedman's inequality to the martingale sequence
$\{\widetilde Z_k\}$ defined in
\eqref{eq:TD-square-martingale}. Towards this end, we first bound the magnitude and cumulative conditional variance of $\widetilde Z_k$. 
\begin{itemize}
\item We start with bounding the magnitude $\max_{k:t/2<k\leq t}\|\widetilde Z_k\|_\infty$. Since conditionally on the past, each coordinate of $V_k$ lies in an interval of length at most $\eta_k\gamma/(1-\gamma)$, and the squaring operator is $2/(1-\gamma)$-Lipschitz on $[0,1/(1-\gamma)]$, it follows that
\[
\big\|V_k\circ V_k-\mathbb E_{k-1}[V_k\circ V_k]\big\|_\infty
\leq\frac{2\eta_k}{(1-\gamma)^2}   \leq\frac{2\overline\eta}{(1-\gamma)^2}, 
\]
where we recall the definition of $\overline{\eta}$ in \eqref{eq:defn-bar-eta-large-k-TD}. This gives immediately
\[ 
\max_{k:\,t/2<k\leq t}\|\widetilde Z_k\|_\infty
\leq\frac{2\overline\eta}{(1-\gamma)^2}. 
\]

\item We now move to the cumulative conditional variance $\sum_k \mathsf{Var}_{k-1}(\widetilde Z_k)$. First, it is well known that $\mathsf{Var}(X^2)\leq 4C_{\mathsf{max}}^2\mathsf{Var}(X)$ for any random variable $X$ obeying $|X|\leq C_{\mathsf{max}}$
(see, e.g., \citet[Lemma~30]{chen2021implicit}), and hence
\begin{align*}
\mathsf{Var}_{k-1}(V_k\circ V_k)
&\leq\frac{4}{(1-\gamma)^2} \mathsf{Var}_{k-1}(V_k)
 \leq \frac{4\eta_k^2\gamma^2}{(1-\gamma)^2} \mathsf{Var}_{P_\pi}(V_{k-1}),
\end{align*}
where the last line is valid since $ \mathsf{Var}_{k-1}(V_k) = \eta_k^2 \gamma^2 \mathsf{Var}_{P_\pi}(V_{k-1})$. It then follows from \eqref{eq:Var-u-Pk-v-bound} that 
\begin{align*}
\mathsf{Var}_{k-1}(\widetilde Z_k)
&\leq\frac{4\eta_k^2\gamma^2}{(1-\gamma)^2}
\prod_{i=k+1}^t\big(I-\eta_i(I-\gamma P_\pi)\big)
\mathsf{Var}_{P_\pi}(V_{k-1}).
\end{align*}
Using $\eta_k^2\leq\overline\eta\eta_k$ and the definition \eqref{eq:TD-variance-proxy} of $\overline W_t$, we conclude that
\begin{align*}
\sum_{k=\lfloor t/2\rfloor+1}^t\mathsf{Var}_{k-1}(\widetilde Z_k)
\leq\frac{4}{(1-\gamma)^2}\overline W_t. 
\end{align*}

\item In addition, similar to the derivation in \eqref{eq:Wt-trivial-UB-TD}, we record a crude bound of $\|\overline W_t\|_\infty \leq\frac{\overline\eta}{(1-\gamma)^3}$, and hence 
$$  \bigg\| \sum_{k=\lfloor t/2\rfloor+1}^t\mathsf{Var}_{k-1}(\widetilde Z_k)\bigg\|_{\infty} \leq  \frac{4\overline\eta}{(1-\gamma)^5} \eqqcolon \widetilde{\sigma}_w^2. $$
\end{itemize}
Applying Freedman's inequality in Lemma~\ref{lem:Freedman-prelim} to each coordinate of $\sum_{k>t/2}\widetilde Z_k$ with $K=2\lceil\log_2\frac{1}{1-\gamma}\rceil$ chosen as in Step~2, we have
$ \frac{\widetilde{\sigma}_w^2}{ 2^K}
\leq\frac{4\overline\eta}{(1-\gamma)^4}$. Taking a union bound over the $S$ coordinates yields
\begin{align*}
\bigg\|\sum_{k=\lfloor t/2\rfloor+1}^t\widetilde Z_k\bigg\|_\infty
&\leq\sqrt{\frac{32}{(1-\gamma)^2}
\bigg(\|\overline W_t\|_\infty+\frac{\overline\eta}{(1-\gamma)^2}\bigg)
\log\frac{4STK}{\delta}}\arxiv{\\}
\arxiv{&\quad} +\frac{8\overline\eta}{3(1-\gamma)^2}\log\frac{4STK}{\delta} 
\end{align*}
with probability at least $1-\delta/(2T)$. Further, the sample-size condition \eqref{eq:TD-log-bound} and $\varepsilon \leq \frac{1}{1-\gamma}$ give
\begin{equation}
\overline\eta\log\frac{4STK}{\delta}
\leq\frac{6}{c_3}(1-\gamma)^2\varepsilon^2
\leq\frac{1}{64},
\label{eq:TD-small-step}
\end{equation}
Substituting the bound into \eqref{eq:TD-variance-energy-bound}, and using $(1-\overline\eta)^{-1}\leq2$ and \eqref{eq:TD-small-step}, we obtain
\begin{align*}
\|\overline W_t\|_\infty
&\leq\frac{13\overline\eta}{2(1-\gamma)^2}
+\sqrt{\frac{2\overline\eta}{(1-\gamma)^2}
\bigg(\|\overline W_t\|_\infty+\frac{\overline\eta}{(1-\gamma)^2}\bigg)}\\
&\leq\frac12\|\overline W_t\|_\infty
+\frac{8\overline\eta}{(1-\gamma)^2},
\end{align*}
where the last step invokes the AM-GM inequality. Solving the above relation yields 
$$
\|\overline W_t\|_\infty\leq \frac{16\overline\eta}{(1-\gamma)^2}.
$$ 
Combining this with the  bound \eqref{eq:partial-sum-Var-Zk-TD-small-k} and \eqref{eq:TD-variance-proxy}, we arrive at
\begin{equation}
\|W_t\|_\infty\leq\frac{\gamma^2}{2(1-\gamma)^2T^3}
+\frac{16\overline\eta}{(1-\gamma)^2}
\leq \frac{33\log^3T}{(1-\gamma)^3T}
\label{eq:Wt-bound-Var}
\end{equation}
with probability at least $1-\delta/(2T)$.

\section{Appendix B: proof of auxiliary lemmas in Section~\ref{sec:Q-learning}}

\subsection{Proof of Lemma~\ref{lemma:non-negativity-Qt-Vt}} \label{appendix:proof-lemma-Q-bounded}

The proof is similar to that of Lemma~\ref{lemma:TD-bounded}. Specifically, we would like to show by induction that, for every $t\ge 0$,
\begin{equation}
	0\le Q_t \le \frac{1}{1-\gamma}1_{SA} 
	\quad \text{and} \quad 
	0\le V_t\le \frac{1}{1-\gamma}1_S.
	\label{eq:Q-bounded-induction} 
\end{equation}
We focus on proving the bounds for $Q_t$; once these are established, the bounds for $V_t$ follow immediately from the relation $V_t(s)=\max_{a}Q_t(s,a)$.

The base case $t=0$ follows directly from the assumption of the lemma. Now suppose that \eqref{eq:Q-bounded-induction} holds for the $(t-1)$-th iterate. Since $0\le \eta_t\le 1$, $0\le r\le 1_{SA}$, $P_t\ge 0$, and each row of $P_t$ sums to one, the update rule \eqref{eq:iteration-rule-Q-infinite} yields
\[
Q_t=(1-\eta_t)Q_{t-1}+\eta_t\bigl(r+\gamma P_tV_{t-1}\bigr)\ge 0
\]
and
\begin{align*}
	Q_t
	&=(1-\eta_t)Q_{t-1}+\eta_t\bigl(r+\gamma P_tV_{t-1}\bigr) \\
	&\le (1-\eta_t)\|Q_{t-1}\|_{\infty}1_{SA}
	+\eta_t\Bigl(\|r\|_{\infty}+\gamma \|P_t\|_{1}\|V_{t-1}\|_{\infty}\Bigr)1_{SA} \\
	&\le (1-\eta_t)\frac{1}{1-\gamma}1_{SA}
	+\eta_t\bigg(1+\frac{\gamma}{1-\gamma}\bigg)1_{SA}
	= \frac{1}{1-\gamma}1_{SA}.
\end{align*}
This verifies \eqref{eq:Q-bounded-induction} for the $t$-th iterate and establishes \eqref{eq:Q-i-LB-UB}. 

The other claim \eqref{eq:Delta-t-range-infinite} follows by combining \eqref{eq:Q-i-LB-UB} with the fact that $0\leq Q^{\star} \leq  \frac{1}{1-\gamma}1_{SA}$. 

\subsection{Proof of Lemma~\ref{lemma:Q-step-size}}
\label{appendix:proof-lemma-Q-step-size}

From assumption~\eqref{eq:thm:infinite-horizaon-eta}, we have $0\leq\eta_j\leq1$ for all $1\leq j\leq T$. For any $\lfloor T/2\rfloor<i\leq t\leq T$, we invoke the upper bound in that assumption to obtain
\begin{align*}
0\leq\eta_i^{(t)}
&\leq\eta_i
\leq\frac{\log^3T}{c_2(1-\gamma)i}
<\frac{2\log^3T}{c_2(1-\gamma)T},
\end{align*}
where the first two inequalities follow from~\eqref{def:eta-i-t-simple}, and the last inequality uses $i>\lfloor T/2\rfloor$. Combining this with $\eta_i\leq1$ and the definition~\eqref{eq:Q-comparison-parameters} of $\overline{\eta}$ verifies that $0\leq\eta_i^{(t)}\leq\eta_i\leq\overline{\eta}$.

Moreover, a telescoping argument reveals that
\begin{align*}
\sum_{i=0}^{t}\eta_i^{(t)}
&=\prod_{j=1}^{t}(1-\eta_j)
+\eta_1\prod_{j=2}^{t}(1-\eta_j)
+\eta_2\prod_{j=3}^{t}(1-\eta_j) \arxiv{\\
&}\qquad+\cdots+\eta_{t-1}(1-\eta_t)+\eta_t\\
&=\prod_{j=2}^{t}(1-\eta_j)
+\eta_2\prod_{j=3}^{t}(1-\eta_j)
+\cdots+\eta_{t-1}(1-\eta_t)+\eta_t\\
&=\cdots=(1-\eta_t)+\eta_t=1.
\end{align*}
In fact, the same argument starting at any $1\leq\tau\leq t$ yields
\begin{align*}
\prod_{j=\tau}^{t}(1-\eta_j)+\sum_{i=\tau}^{t}\eta_i^{(t)}
&=\prod_{j=\tau+1}^{t}(1-\eta_j)
+\sum_{i=\tau+1}^{t}\eta_i^{(t)}\arxiv{\\
&}=\cdots=1,
\end{align*}
where an empty product equals one and an empty sum equals zero. Rearranging this identity, and taking it together with the preceding calculation for $\tau=0$, we establish~\eqref{eq:Q-weight-sum}.

Finally, it follows from $\eta_i^{(t)}\leq\overline{\eta}$ and~\eqref{eq:Q-weight-sum} that
\begin{align*}
\sum_{i=\lfloor T/2\rfloor+1}^{t}\big(\eta_i^{(t)}\big)^2
&\leq\overline{\eta}\sum_{i=\lfloor T/2\rfloor+1}^{t}\eta_i^{(t)} 
\leq\overline{\eta}.
\end{align*}

\subsection{Proof of Lemma~\ref{lemma:Q-comparison-facts}}
\label{appendix:proof-lemma-Q-comparison-facts}

\paragraph{Proof of part (i).}
For any given $p$ and $V$, write $\sigma^2\coloneqq\mathsf{Var}_p(V)$.
For every $q\in\mathcal{U}_{\rho}(p)$, observe that
\begin{align*}
 \big|q^{\top}V-p^{\top}V\big|
&=\Bigg|\sum_{s:p(s)>0}\big(q(s)-p(s)\big)
\big(V(s)-p^{\top}V\big)\Bigg|\\
&\leq\sqrt{ \Bigg(\sum_{s:p(s)>0}\frac{(q(s)-p(s))^2}{p(s)}\Bigg)  \Bigg(\sum_{s:p(s)>0}p(s)\big(V(s)-p^{\top}V\big)^2\Bigg) } \\
&=\sqrt{\chi^2(q\,\|\,p)}\,\sigma\leq\sqrt{\rho}\,\sigma.
\end{align*}
where the first equality uses $\sum_s q(s) = \sum_s p(s)$ as well as the support restriction in \eqref{eq:Q-comparison-ball}, 
and the second relation follows from the Cauchy-Schwarz inequality.
This proves the upper bound in \eqref{eq:Q-comparison-support}.

To establish the lower bound in \eqref{eq:Q-comparison-support}, suppose first that
$\sigma\geq\sqrt{\rho}/(1-\gamma)$, and consider a specific vector $q_1$ as follows
\[
q_1(s)=p(s)\left(1+\frac{\sqrt{\rho}}{\sigma}
\big(V(s)-p^{\top}V\big)\right),\quad \forall s\in\cS.
\]
Since $|V(s)-p^{\top}V|\leq1/(1-\gamma)$, every entry of $q_1$ is nonnegative. 
Moreover, direct calculation yields
\[
\sum_s q_1(s)=1,\qquad
\chi^2(q_1\,\|\,p)=\frac{\rho}{\sigma^2}\mathsf{Var}_p(V)=\rho,
\qquad q_1^{\top}V-p^{\top}V=\sqrt{\rho}\,\sigma.
\]
Consequently, $q_1$ lies within $\mathcal{U}_{\rho}(p)$ and attains the upper bound.
If instead $0\leq \sigma<\sqrt{\rho}/(1-\gamma)$, then
$\sqrt{\rho}\,\sigma-\rho/(1-\gamma)\leq0$, whereas the choice $q_2=p$ already gives $q_2^{\top}V-p^{\top}V=0$. 
This proves \eqref{eq:Q-comparison-support} in all cases.

Finally, applying the above arguments to $-V$ immediately establishes 
\begin{equation*}
\sqrt{\rho\mathsf{Var}_p(V)}-\frac{\rho}{1-\gamma}
    \leq p^{\top}V-\inf_{q\,\in\,\mathcal{U}_{\rho}(p)}q^{\top}V
\leq\sqrt{\rho\mathsf{Var}_p(V)}.
\end{equation*}

\paragraph{Proof of part (ii).}
Let $Q^{\pi,P'}$ (resp.~$Q^{\pi,P}$) denote the Q-function of policy $\pi$ under the transition kernel $P'$ (resp.~$P$). Let $\{(s_k,a_k)\}_{k\geq 0}$ represent a sample trajectory generated under policy $\pi$, and let $\tau$ be an independent random variable with
\[
\mathbb{P}(\tau=k)=(1-\gamma)\gamma^k,\qquad k=0,1,\ldots.
\]
The Q-function can then be expressed as the expected reward at the random stopping time $\tau$:
\begin{equation}
Q^{\pi,P}(s,a)
=\sum_{k=0}^{\infty}\gamma^k\mathbb{E}_{P}[r(s_k,a_k)]
=\frac{1}{1-\gamma}\mathbb{E}_{P}[r(s_{\tau},a_{\tau})],
\label{eq:Q-comparison-geometric-return}
\end{equation}
with an analogous identity holding under $P'$. Here and below, $\mathbb{E}_P[\cdot]$ denotes expectation w.r.t.~the transition kernel $P$. 

To proceed, we examine the distributions of randomly stopped trajectories under different transition kernels. Conditional on $\tau=k$, let $\mathbb{P}'_k$ and $\mathbb{P}_k$ denote the distributions of the trajectory $(s_0,a_0,\ldots,s_k,a_k)$ under $P'$ and $P$, respectively, assuming that the initial state--action pair and the policy are the same in both cases. 
The chain rule for KL divergence reveals that
\begin{align}
\mathsf{KL}(\mathbb{P}'_k\,\|\,\mathbb{P}_k)
&=\sum_{j=0}^{k-1}\mathbb{E}_{P'}\left[
\mathsf{KL}\big(P'(\cdot\mymid s_j,a_j)\,\|\,P(\cdot\mymid s_j,a_j)\big)\right] \leq k\rho,
\label{eq:KL-ub-k-rho-proof}
\end{align}
where we use the fact that for any $q\in\mathcal{U}_{\rho}(p)$, one has
$
\mathsf{KL}(q\,\|\,p)
\le \chi^2(q\,\|\,p)\leq\rho
$ (see~\eqref{eq:KL-ub-chi2}). 

Let $\mathbb{P}'$ and $\mathbb{P}$ denote the joint distributions of the stopping time and the corresponding randomly stopped trajectory under the transition kernels $P'$ and $P$, respectively. 
Another application of the chain rule combined with~\eqref{eq:KL-ub-k-rho-proof} yields
\[
\mathsf{KL}(\mathbb{P}'\,\|\,\mathbb{P})
=\sum_{k=0}^{\infty}(1-\gamma)\gamma^k
\mathsf{KL}(\mathbb{P}'_k\,\|\,\mathbb{P}_k)
\leq\rho\sum_{k=0}^{\infty}k(1-\gamma)\gamma^k
=\frac{\rho\gamma}{1-\gamma}.
\]
Combining \eqref{eq:Q-comparison-geometric-return} with $0\leq r\leq1$ and Pinsker's inequality (see~\eqref{eq:KL-ub-chi2}, or \citet[Lemma~2.5]{tsybakov2009introduction}), we can demonstrate that
\begin{align*}
\big|Q^{\pi,P'}(s,a)-Q^{\pi,P}(s,a)\big|
&\leq\frac{1}{1-\gamma}\mathsf{TV}(\mathbb{P}',\mathbb{P}) \arxiv{\\
&}\leq\frac{1}{1-\gamma}\sqrt{\frac12\mathsf{KL}(\mathbb{P}'\,\|\,\mathbb{P})}
\leq\sqrt{\frac{\rho\gamma}{2(1-\gamma)^3}},
\end{align*}
which holds uniformly over all $(s,a)$ and all deterministic policy $\pi$. This in turn reveals that
\begin{align*}
\big|Q^{\star,P'}(s,a)-Q^{\star}(s,a)\big|
&=\left|\max_{\pi}Q^{\pi,P'}(s,a)-\max_{\pi}Q^{\pi,P}(s,a)\right|\arxiv{\\
&}\leq\max_{\pi}\big|Q^{\pi,P'}(s,a)-Q^{\pi,P}(s,a)\big|,
\end{align*}
thereby establishing~\eqref{eq:Q-comparison-stability}.

\section{Appendix C: proof of the lower bound in Theorem~\ref{thm:LB-example-Q-generative}}
\label{sec:proof-thm-LB-Q-generative}

We now present the proof of the algorithm-dependent lower bound in Theorem~\ref{thm:LB-example-Q-generative} for Q-learning. For brevity, we focus on the case of time-invariant learning rates, i.e., $$\eta_t\equiv \eta\in (0,1),\qquad t\geq 0$$ throughout the execution of the algorithm. The proof for the case of rescaled linear learning rates is similar and can be found in \citet{li2024q}. Without loss of generality, we assume
\begin{equation} \label{eq:logT_ub}
\log T\leq \frac{1}{1-\gamma}
\end{equation} 
throughout the proof; otherwise the lower bound in Theorem~\ref{thm:LB-example-Q-generative} becomes looser than the minimax lower bound $\frac{1}{(1-\gamma)^3T}$ established in \citet{azar2013minimax} (see Theorem~\ref{theorem:azar}).

 \subsection{Step 1: construction of a hard MDP}

\begin{figure}[t]
\begin{center} 
	\includegraphics[width=0.8\textwidth]{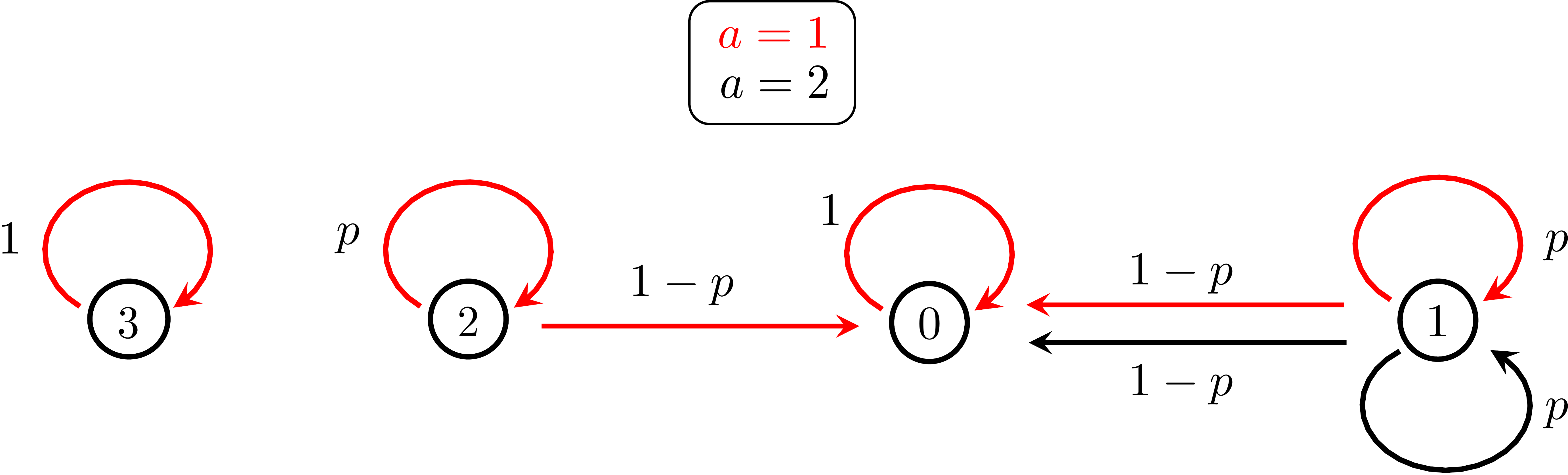}  
\end{center}	
	\caption{An illustration of the hard MDP in the proof of the lower bound for Q-learning. 	\label{fig:hard-MDP-Q}}

\end{figure}

We begin by constructing a hard MDP instance $\mathcal{M}_{\mathsf{hard}}$  with 4 states and 2 actions as follows; see Figure~\ref{fig:hard-MDP-Q} for an illustration. The subsequent analysis will focus on tracking the Q-learning dynamics for this specific construction.
\begin{itemize}
\item {\em State space.} The state space of this MDP contains 4 states, namely, $$\cS=\{0,1,2,3\}.$$ 
\item {\em Action space.}
To streamline the presentation, we allow the action space to vary across states. Let $\cA_s$ represent the action space associated with each state $s\in \cS$, which we specify as follows:
\[
\mathcal{A}_0=\{1\},
\quad \mathcal{A}_1=\{1,2\},
\quad \mathcal{A}_2=\{1\},
\quad \mathcal{A}_3=\{1\}. 
\]
In other words, only state 1 admits more than one action, while all other states are associated with a single action. 

%As we shall see later, the most intricate part of the analysis gravitates around state 1 (which has two actions), with the remaining states are constrcute

\item {\em Probability transition kernel and reward function.}
The probability transition kernel and reward function of $\mathcal{M}_{\mathsf{hard}}$ are chosen as  
%
%\begin{subequations}
%\label{eq:construction-hard-MDP}
\begin{align*}
	%& \mathcal{A}_0=\{1\},  & 
    & \text{from state }0: &
    P(0 \mymid 0, 1) &= 1, &\quad && r(0, 1) = 0, \\
	%& \mathcal{A}_1=\{1,2\},  &
    & \text{from state }1: &
     P(1 \mymid 1, 1) &= p,   & P(0 \mymid 1, 1) = 1-p,   && r(1, 1) = 1,  \\
	& \quad   & 
    P(1 \mymid 1, 2) &= p,  & P(0 \mymid 1, 2) = 1-p,  && r(1, 2) = 1, \\
	%& \mathcal{A}_2=\{1\},  
    %& 
    & \text{from state }2: &
    P(2 \mymid 2, 1) &= p,  & P(0 \mymid 2,1) = 1-p,  && r(2,1) = 1, \\
	%& \mathcal{A}_3=\{1\},  
    %&
    & \text{from state }3: &
    P(3 \mymid 3, 1) &= 1,  &\quad && r(3,1) = 1, 
\end{align*}
%\end{subequations}
%
for some parameter $p\in (0,1)$. States $0$ and $3$ are absorbing, while from states $1$ and $2$ the process either remains in the same state or transitions to state $0$. Moreover, state $0$ yields zero immediate reward; once it is entered, no further rewards can be obtained. 
\end{itemize}
In fact, the MDP construction bears some similarity with that in Section~\ref{sec:analysis-lower-bound-Q-generative} for the minimax lower bounds, with the key distinction being that the action space is now augmented to include at least two actions in order to capture the over-estimation issue of Q-learning. Motivated by the construction in Section~\ref{sec:analysis-lower-bound-Q-generative} (see \eqref{eq:choise-p-alpha-lower-bound-generative}), we choose the parameter $p$ as 
\begin{align}
	\label{defn-parameter-p-Q-lower}
	p = \frac{4\gamma-1}{3\gamma} .
\end{align}
A careful reader may also notice the inclusion of state 
3, which is isolated and absorbing. As we shall see later, this state is introduced to simplify a small part of the analysis.

To facilitate the subsequent convergence analysis of Q-learning on this MDP, we first characterize its optimal value function and Q-function. The proof is deferred to the end of this section.
\begin{lemma}
\label{lem:optimal-V-Q-hardMDP}
The optimal value function and Q-function of the MDP $\mathcal{M}_{\mathsf{hard}}$ constructed above satisfy
\begin{subequations}
\label{eq:optimal-V-Q-hardMDP}
\begin{align*}
&V^{\star}(0) = Q^{\star}(0,1) = 0; \\
&V^{\star}(1) = Q^{\star}(1, 1) = Q^{\star}(1, 2) = V^{\star}(2) = Q^{\star}(2,1) = \frac{1}{1-\gamma p}= \frac{3}{4(1-\gamma)}; \\
&V^{\star}(3) = Q^{\star}(3, 1) = \frac{1}{1-\gamma}.
\end{align*}
\end{subequations}
\end{lemma}
\begin{proof}[Proof of Lemma~\ref{lem:optimal-V-Q-hardMDP}]
%\label{sec:proof-lem:optimal-V-Q-hardMDP}
Since state $0$ is absorbing with zero reward --- so no further rewards are obtained once it is entered --- we have
\[
%V^{\pi}(0)=0\qquad\text{for all }\pi
%\qquad \Longrightarrow \qquad 
V^{\star}(0) = Q^{\star}(0,1)  =0.
\]

We now turn to states 1 and 2. Since actions 1 and 2 in state 1 are equivalent in both rewards and transition probabilities, it is clear that
\begin{equation}
Q^{\star}(1,1)=Q^{\star}(1,2)=V^{\star}(1).\label{eq:Q1-Q2-Vstar}
\end{equation}
Also, it follows from the Bellman optimality equation \eqref{eq:bellman_fix_point} that
\[
	Q^{\star}(1,1)=r(1,1)+\gamma P(0\mymid1,1)V^{\star}(0)+\gamma P(1\mymid1,1)V^{\star}(1).
\]
Combining this with (\ref{eq:Q1-Q2-Vstar}) and rearranging terms, we reach
\[
	V^{\star}(1)=\frac{r(1,1)+\gamma P(0\mymid1,1)V^{\star}(0)}{1-\gamma P(1\mymid1,1)}=\frac{1}{1-\gamma p}=\frac{3}{4(1-\gamma)}.
\]
where the second identity follows since $V^{\star}(0)=0$, and the third identity relies on the choice \eqref{defn-parameter-p-Q-lower} of $p$.  
The derivations for $V^{\star}(2)$ and $Q^{\star}(2,1)$ are omitted as they follow from identical arguments.

When it comes to state $3$, we invoke Bellman's equation to obtain
\[
V^{\star}(3)  = Q^{\star}(3,1)=r(3,1)+\gamma P(3\mymid 3,1)V^{\star}(3) = 1 + \gamma V^{\star}(3)  .
\]
A little algebra then leads to $V^{\star}(3) = \frac{1}{1-\gamma}$.
\end{proof}

\subsection{Step 2: preliminary calculation of Q-learning iterates}
We now embark on the analysis of  Q-learning for the MDP instance constructed in Step 1.  As before, we let $P_t$ encode the empirical transitions such that for any triple $(s,a,s')$,
\begin{align}
	P_t(s'\mymid s,a) \coloneqq \begin{cases}  1, \quad & \text{if } s_t(s,a)=s', \\ 0, & \text{otherwise}, \end{cases}
\end{align}
where $s_t(s,a)$ denotes the sample collected during the $t$-th iteration w.r.t.~the $(s,a)$ pair. Given that state 2 has a singleton action space, we simplify the notation by writing
\[
	P_t(s'\mymid 2) \coloneqq   P_t(s'\mymid 2,1).
\]

In what follows, we first introduce some notation and basic properties concerning the learning rates, and then slightly expand the Q-learning updates for each state. 

%\paragraph{Key quantities related to learning rates}

%\paragraph{Preliminary calculations.}
%Before moving forward, we record several basic relations as a result of the Q-learning update rule.  

%\subsubsection{Basic update rules and expansion}

\paragraph{Preliminary facts about learning rates.}
Let us introduce the following notation about learning rates which will be convenient for analyzing the Q-learning iterations:
%
%\begin{subequations}
%
\begin{align}
\label{eq:defn-eta-kt-hard}
\eta_k^{(t)} &\defn 
\begin{cases}
\eta\big(1-\eta(1-\gamma p)\big)^{t-k}, &\text{if }1 \le k \le t, \\
%\eta_0^{(t)} &\defn 
\big(1-\eta(1-\gamma p)\big)^t, & \text{if }k=0.
\end{cases}
%\eta_t^{(t)} &\defn \eta. 
\end{align}
%
%\end{subequations}
%
A direct geometric-sum calculation gives 
\begin{subequations}
\begin{align}
	\eta_0^{(t)} + (1-\gamma p)\sum_{k = 1}^t \eta_k^{(t)} &= %\big(1-\eta(1-\gamma p)\big)^t \notag\\
    %&\qquad+ \eta(1-\gamma p)\sum_{k = 1}^t \big(1-\eta(1-\gamma p)\big)^{t-k}
    1 
	\label{eq:eta-sum-hat} \\
    \big(1-\eta(1-\gamma p)\big)^{t-\tau} + (1-\gamma p)\sum_{k = \tau+1}^t \eta_k^{(t)} &= 1
    \label{eq:eta-sum-hat-tau}
\end{align}
\end{subequations}
%
%Similarly, 
for any given integer $0\leq \tau < t$.
%one has
%
%\begin{align}
%%\label{eq:eta-sum-hat-tau}	
%	\big(1-\eta(1-\gamma p)\big)^{t-\tau} + (1-\gamma p)\sum_{k = \tau+1}^t \eta_k^{(t)} = 1.
%\end{align}

\paragraph{State 0.} To begin, since $Q_0 = V_0 = 0$ and state 0 is absorbing, the Q-learning update rule implies that, for all $t\geq 1$, 
\begin{align} 
V_t(0) = Q_t(0,1) &= (1-\eta)Q_{t-1}(0,1) + \eta \big(r(0,1)+ \gamma V_{t-1}(0) \big) \notag\\
%= \eta Q_{t-1}(0,1) 
&= \big(1-\eta(1-\gamma)\big)Q_{t-1}(0,1) = \cdots \notag\\
&= \big(1-\eta(1-\gamma)\big)^t Q_0(0,1) = 0.
\label{eq:vt-0-Q-lower}
\end{align}

\paragraph{State 1.}
For state 1, the Q-learning iterates satisfy
\begin{subequations}
\label{eq:Qt-1-Q-lower}
\begin{align}
	Q_t(1, 1) &= (1-\eta)Q_{t-1}(1, 1) + \eta\big\{ 1 + \gamma P_{t}( 1 \mymid 1, 1) V_{t-1}(1)\big\}, \\
	Q_t(1, 2) &= (1-\eta)Q_{t-1}(1, 2) + \eta\big\{ 1 + \gamma P_{t}( 1 \mymid 1, 2) V_{t-1}(1)\big\}.
\end{align}
\end{subequations}
%

%In what follows, we shall first determine a crude range for certain quantities relates to the learning rates $\eta_t$, and then combine this with the above relations to establish the desired result.

\paragraph{State 2.}
The Q-learning update combined with~\eqref{eq:vt-0-Q-lower} gives
\begin{align} 
	\arxiv{&}V_t(2) = Q_t(2,1) \arxiv{\notag\\}
    & \arxiv{~}= \big(1-\eta\big)Q_{t-1}(2,1) + \eta\big\{ r(2,1)+ \gamma P_t(2 \mymid 2)V_{t-1}(2) + \gamma P_t(0 \mymid 2)V_{t-1}(0) \big\} \notag\\
	& \arxiv{~}= \big(1-\eta\big)V_{t-1}(2) + \eta\big\{ 1 + \gamma P_t(2 \mymid 2)V_{t-1}(2)\big\} \notag\\
    &\arxiv{~}= \big(1-\eta(1-\gamma p)\big)V_{t-1}(2) + \eta\big\{ 1+ \gamma \big( P_t(2 \mymid 2) - p \big) V_{t-1}(2)\big\}.\label{eq:vt-2-Q-lower}
\end{align}
Therefore, for any $t$ and $\tau<t$, applying \eqref{eq:vt-2-Q-lower} recursively leads to
\begin{align}
	V_t(2) & = \big(1-\eta(1-\gamma p)\big)^{t-\tau}V_{\tau}(2) \arxiv{ \notag\\}
  \arxiv{  &\qquad} + \sum_{k = \tau+1}^t \eta_k^{(t)}+ \sum_{k = \tau+1}^t \eta_k^{(t)} \gamma \big(P_k(2 \mymid 2) - p \big)V_{k-1}(2) ,
	\label{eq:expansion-Vt2-tau-Q-lower-012}
\end{align}
where $\eta_k^{(t)}$ has been defined in \eqref{eq:defn-eta-kt-hard}. 
Recalling from \eqref{eq:eta-sum-hat-tau} that
\[
    \sum_{k = \tau+1}^t \eta_k^{(t)} = \frac{1 - \big(1-\eta(1-\gamma p)\big)^{t-\tau} }{1-\gamma p}, 
\]
we can rearrange terms in \eqref{eq:expansion-Vt2-tau-Q-lower-012} to reach
\begin{align}
	V_t(2) 
    %&= \big(1-\eta(1-\gamma p)\big)V_{t-1}(2) + \eta\big\{ 1+ \gamma \big( P_t(2 \mymid 2) - p \big) V_{t-1}(2)\big\} \nonumber\\
	%&= \big(1-\eta(1-\gamma p)\big)^{t-\tau}V_{\tau}(2) \notag\\
    %&\qquad + \sum_{k = \tau+1}^t\eta\big(1-\eta(1-\gamma p)\big)^{t-k}\big\{ 1+ \gamma \big(P_k(2 \mymid 2) - p \big)V_{k-1}(2)\big\} \nonumber\\
	%&=  \big(1-\eta(1-\gamma p)\big)^{t-\tau}V_{\tau}(2) + \sum_{k = \tau+1}^t \eta_k^{(t)}  + \sum_{k = \tau+1}^t \eta_k^{(t)}  \gamma \big(P_k(2 \mymid 2) - p \big)V_{k-1}(2) \nonumber\\
	%&=  \big(1-\eta(1-\gamma p)\big)^{t-\tau}V_{\tau}(2) + \frac{1 - \big(1-\eta(1-\gamma p)\big)^{t-\tau} }{1-\gamma p}  + \sum_{k = \tau+1}^t \eta_k^{(t)}  \gamma \big(P_k(2 \mymid 2) - p \big)V_{k-1}(2) \nonumber\\
	&= \frac{1}{1-\gamma p} - \big(1-\eta(1-\gamma p)\big)^{t-\tau}\left[\frac{1}{1-\gamma p} - V_{\tau}(2)\right] \arxiv{ \notag\\}
   \arxiv{  &\qquad } + \sum_{k = \tau+1}^t\eta_k^{(t)}\gamma \big(P_k(2\mymid 2) - p\big) V_{k-1}(2).
	\label{eq:expansion-Vt2-tau-Q-lower}
\end{align}
%
%where the penultimate line arises from \eqref{eq:eta-sum-hat-tau}.
In particular, in the special case where $\tau=0$ (so that $V_{\tau}(2)=V_{0}(2)=0$), this identity simplifies to
\begin{align} 
	V_t(2) = \frac{1-\eta_0^{(t)}}{1-\gamma p} + \sum_{k = 1}^t\eta_k^{(t)}\gamma \big(P_k(2 \mymid 2) - p \big)V_{k-1}(2),  
	\label{eq:Vt2-expansion-case1}
\end{align}
where we recall the definition of $\eta_0^{(t)}$ in \eqref{eq:defn-eta-kt-hard}. 

Moreover, 
%\eqref{eq:construction-hard-MDP}, 
we know that $\mathbb{E}[ P_k(2 \mymid 2) ] = p$ by construction, which together with the identity \eqref{eq:Vt2-expansion-case1} and Lemma~\ref{lem:optimal-V-Q-hardMDP} leads to
\begin{subequations}
	\label{eq:EV2}
\begin{align}    
	\mathbb{E} \big[V_t(2)\big] &= \frac{1-\eta_0^{(t)}}{1-\gamma p}
    =
    \frac{1-\big(1-\frac{4\eta}{3}(1-\gamma)\big)^t}{\frac{4}{3} (1-\gamma)}; 
    \label{eq:EV2-01}\\
	%\quad \text{and}\quad
	V^{\star}(2) - \mathbb{E} \big[V_t(2)\big] &= \frac{\eta_0^{(t)}}{1-\gamma p}=\frac{\big(1-\frac{4\eta}{3}(1-\gamma)\big)^t}{\frac{4}{3} (1-\gamma)}. 
    \label{eq:EV2-02}
\end{align}
\end{subequations}
Similarly, applying the above argument to \eqref{eq:expansion-Vt2-tau-Q-lower} and rearranging terms, we obtain, for any integer $0\leq \tau <T$, 
\begin{align}
	 V^{\star}(2) - \mathbb{E}\big[V_{T}(2)\big] 
	&= \big(1-\eta(1-\gamma p)\big)^{T-\tau}\left(\frac{1}{1-\gamma p} - \mathbb{E}\big[ V_{\tau}(2)\big]\right) 
    \notag\\
    &= \big(1-\eta(1-\gamma p)\big)^{T-\tau}\left(V^{\star}(2) - \mathbb{E}\big[ V_{\tau}(2)\big]\right) .
    %&\leq \big(1-\eta(1-\gamma p)\big)^{T-\tau}\frac{1}{1-\gamma p}
	\label{eq:Vstar2-V2-expansion-Case2}
\end{align}

\paragraph{State 3.}
When it comes to state 3, it holds that
\begin{align} 
	V_t(3) = Q_t(3,1)  & = \big(1-\eta\big)Q_{t-1}(3,1) + \eta\big\{ r(3,1)+ \gamma V_{t-1}(3) \big\} \notag\\
	& = \big(1-\eta(1-\gamma)\big)V_{t-1}(3) + \eta .  \label{eq:vt-3-Q-lower}
\end{align}
Recognizing that $V^{\star}(3) = \frac{1}{1-\gamma}$ (see Lemma~\ref{lem:optimal-V-Q-hardMDP}) obeys
\[
V^{\star}(3)  = \big(1-\eta(1-\gamma)\big)V^{\star}(3) + \eta,
\]
we can combine these two identities to arrive at
\begin{align}
V_t(3) - V^{\star}(3) &=   \big(1-\eta(1-\gamma)\big) \big(V_{t-1}(3)-V^{\star}(3) \big) \notag\\
&=\big(1-\eta(1-\gamma)\big)^T \big(V_{0}(3)-V^{\star}(3) \big) \notag\\
&= -\frac{1}{1-\gamma}\big(1-\eta(1-\gamma)\big)^T.
\label{eq:Vt3-expansion-case3}
\end{align}

%With similar derivation, \eqref{eq:vt-3-Q-lower} leads to
%\begin{align} 
%	V_t(3) = \frac{1}{1-\gamma}\Big[1-\big(1-\eta(1-\gamma)\big)^T\Big] = V^{\star}(3) - \frac{1}{1-\gamma}\big(1-\eta(1-\gamma)\big)^T.	
%\end{align}

\paragraph{Roadmap.} 
In what follows, we divide the analysis into three parts based on the learning rate $\eta$: small ($0 < \eta \le \frac{4}{(1-\gamma)T}$), large ($\eta \ge \frac{1}{(1-\gamma)^2T}$), and moderate ($\frac{4}{(1-\gamma)T} < \eta \le \frac{1}{(1-\gamma)^2T}$). Each regime is tackled in its own subsection below. In particular, the regimes with small and large learning rates will be analyzed by lower bounding the mean and variance of $V^{\star}(2)-V_T(2)$, respectively, for which it suffices to analyze this state that has a single available action. The moderate learning rate regime presents the main technical challenge. For this regime, we analyze the dynamics of state 1, which, by admitting two actions, captures the over-estimation issue; the dynamics of state 3 will also be invoked later to facilitate the analysis.

\subsection{Step 3: the regime with small learning rates}
%
%In order to analyze the suboptimality of the Q-learning iterates, we study the gap $V^{\star}(2) - V_T(2)$, beginning with its expectation.
%
Suppose now the learning rate satisfies $$0< \eta \le \frac{4}{(1-\gamma)T},$$ which, as we shall see, is fairly small. In this regime, it suffices to look at the dynamics associated with state 2 (which has a single action). Our strategy is to show that, under such a small learning rate, the bias $V^{\star}(2) - \mathbb{E} \big[V_t(2)\big]$ does not decay sufficiently quickly and therefore remains large after $T$ iterations.

Towards this end, we first recall from \eqref{eq:EV2} that
\begin{align*}
	%\label{eq:EV2}
	%\mathbb{E} \big[V_T(2)\big] = \frac{1-\eta_0^{(t)}}{1-\gamma p}
	%\quad \text{and}\quad
	V^{\star}(2) - \mathbb{E} \big[V_T(2)\big] = \frac{\eta_0^{(T)}}{1-\gamma p},  
\end{align*}
which reduces the problem to deriving a lower bound on  $\eta_0^{(T)}$. 
Observe that
\begin{subequations}
\begin{align}
\eta(1-\gamma p) & \leq
1-\gamma p=\frac{4(1-\gamma)}{3}\leq\frac{1}{3}, 
\label{eq:eta-prod-gamma-p-13}\\
\eta(1-\gamma p) & = \frac{4\eta(1-\gamma)}{3} < \frac{16}{3T},
\end{align}
\end{subequations}
provided that $\gamma\geq 3/4$ and $\eta \le \frac{4}{(1-\gamma)T}$.  As a consequence, 
\begin{align*}
	\log \eta_0^{(T)} 
	= T \log \big(1-\eta(1-\gamma p)\big) \ge - 1.5T \eta(1-\gamma p) \ge -8,
	%\label{eq:log-eta-0T-LB}
\end{align*}
where the first inequality holds due to the elementary fact $\log(1-x)\geq -1.5x$ for all $0\leq x \leq 0.5$.
Taking the above results collectively yields
\begin{equation*}
%\label{eqn:Vstar-lb-violin}
	%\mathbb{E} \big[ \, \big| V^{\star}(2) - V_T(2) \big| \, \big]  \geq 
	V^{\star}(2) - \mathbb{E} \big[V_T(2)\big]  =  \frac{\eta_0^{(T)}}{1-\gamma p} \ge  \frac{e^{-8}}{1-\gamma p} = \frac{3}{4e^8(1-\gamma)},
\end{equation*}
thus indicating that
\begin{align}
	\mathbb{E}\big[ \big( V^{\star}(2) - V_T(2) \big)^2 \big] 
	&\geq \big( V^{\star}(2) - \mathbb{E} \big[V_T(2)\big] \big)^2 
	\geq \frac{9}{16e^{16}(1-\gamma)^2} \notag\\
    &\geq \frac{9c_3}{16e^{16}(1-\gamma)^4T},
    \label{eq:MSE-V2-case1}
\end{align}
provided that $T\geq c_3/(1-\gamma)^2$. This establishes the advertised lower bound \eqref{eq:lower-bounding-VT-thm-sync} for this regime.

\subsection{Step 4: the regime with large learning rates}

Next, consider the regime with
$$ \eta \geq \frac{1}{(1-\gamma)^2T},$$
corresponding to large learning rates. In this regime, it is again sufficient to look at state 2.  In contrast to the small learning rate regime where we analyze the bias, we now aim to show that the variance of $V_T(2)$ is large.  This is motivated by the intuition that larger learning rates can lead to greater variability in estimating optimal values.

Before proceeding, we first note that if $\eta> \frac{4}{(1-\gamma)T}$ (which covers the range $\eta\geq \frac{1}{(1-\gamma)^2T}$ for $\gamma \geq 3/4$), then one has 
\begin{align}
    \eta_0^{(T)} \arxiv{&}= \bigg(1-\frac{4\eta}{3}(1-\gamma)\bigg)^T
   \book{&} \leq \bigg(1-\frac{4\eta}{3}(1-\gamma)\bigg)^{\frac{3}{4\eta(1-\gamma)}\frac{4(1-\gamma)\eta T}{3}} \notag\\
    &\leq e^{-\frac{4(1-\gamma)\eta T}{3}} < e^{-16/3} < \frac{1}{100},
    \label{eq:EV2-0246}
\end{align}
where the last line makes use of the elementary inequality $(1-x)^{1/x}\leq e^{-1}$ for all $0<x<1$. This combined with \eqref{eq:EV2-02} yields an upper bound on the bias: 
\begin{align}
	\big| V^{\star}(2) - \mathbb{E} \big[V_T(2)\big] \big| &
    = \frac{\eta_0^{(T)}}{\frac{4}{3} (1-\gamma)}
    %=\frac{\big(1-\frac{4\eta}{3}(1-\gamma)\big)^T}{\frac{4}{3} (1-\gamma)}
    %\leq \frac{\big(1-\frac{4\eta}{3}(1-\gamma)\big)^{\frac{3}{4\eta(1-\gamma)}\frac{4(1-\gamma)\eta T}{3}}}{\frac{4}{3} (1-\gamma)} \notag\\
   % &\leq \frac{e^{-\frac{4(1-\gamma)\eta T}{3}}}{\frac{4(1-\gamma)}{3} }
    %\le \frac{e^{-\frac{16}{3}}}{\frac{4(1-\gamma)}{3} } 
    < \frac{1}{100(1-\gamma)}.
    \label{eq:EV2-02468}
\end{align}
%
  %and the assumption $\gamma\geq 3/4$.  

We now turn to the variance of $V_T(2)$. 
 Recalling the decomposition \eqref{eq:expansion-Vt2-tau-Q-lower} of $V_T(2)$, we intend to exploit martingale properties to lower-bound its variance. Specifically, consider first a martingale sequence $\{Z_k\}_{0\leq k\leq T}$ adapted to a filtration $\mathcal{F}_0\subseteq  \mathcal{F}_1 \subseteq \cdots \subseteq \mathcal{F}_T$, 
satisfying  $\mathbb{E} [Z_{k+1} \mymid \mathcal{F}_{k}]=0$ and $\mathbb{E} [Z_k \mymid \mathcal{F}_{k}]=Z_k$ for all $0\leq k\leq T$. 
In addition, consider any $0\leq \tau < T$, and let $W_{0}$ be a random variable such that $\mathbb{E}[W_{0}\mymid \mathcal{F}_{\tau}]=W_0$. 
Then, by the law of total variance and standard martingale properties, we have
\begin{align}
\arxiv{&}\mathsf{Var}\Bigg(W_{0}+\sum_{k=\tau+1}^{T}Z_{k}\Bigg) \arxiv{ \notag} \arxiv{ \\}
&\arxiv{~} =\mathbb{E}\Bigg[\mathsf{Var}\Bigg(W_{0}+\sum_{k=\tau+1}^{T}Z_{k}\,\bigg|\,\mathcal{F}_{T-1}\Bigg)\Bigg]+\mathsf{Var}\Bigg(\mathbb{E}\Bigg[W_{0}+\sum_{k=\tau+1}^{T}Z_{k}\,\bigg|\,\mathcal{F}_{T-1}\Bigg]\Bigg) \notag\\
 &\arxiv{~}   \overset{\mathrm{(i)}}{=}\mathbb{E}\big[\,\mathsf{Var}\left(Z_{T}\mid\mathcal{F}_{T-1}\right)\big]+\mathsf{Var}\Bigg(W_{0}+\sum_{k=\tau+1}^{T-1}Z_{k}\Bigg)=\cdots \notag\\
 &\arxiv{~}  =\sum_{k=\tau+1}^{T}\mathbb{E}\big[\,\mathsf{Var}\left(Z_{k}\mid\mathcal{F}_{k-1}\right)\big]+\mathsf{Var}\left(W_{0}\right) \notag\\
 &\arxiv{~} \geq\sum_{k=\tau+1}^{T}\mathbb{E}\big[\,\mathsf{Var}\left(Z_{k}\mid\mathcal{F}_{k-1}\right)\big],
	\label{eq:variance-expansion-martingale}
\end{align}
where (i) arises from the above assumptions on $W_0$ and $\{Z_k\}$. In words, this decomposes the variance of a martingale sum by summing the expected conditional variances over time.  
Consequently, for any $0\leq \tau < T-1$, applying inequality \eqref{eq:variance-expansion-martingale} to the decomposition \eqref{eq:expansion-Vt2-tau-Q-lower} (with $\tau$ replaced by $\tau+1$) yields
\begin{align}
\mathsf{Var}\big(V_{T}(2)\big) & \ge\mathbb{E}\Bigg[\sum_{k=\tau+2}^{T}\mathsf{Var}\Big(\eta_{k}^{(T)}\gamma\big(P_{k}(2\mymid2)-p\big)V_{k-1}(2)\mymid V_{k-1}(2)\Big)\Bigg] \notag\\
 & =\sum_{k=\tau+2}^{T}\big(\eta_{k}^{(T)}\gamma\big)^{2}p(1-p)\mathbb{E}\big[\big(V_{k-1}(2)\big)^{2}\big] \notag\\
 & \geq\frac{\left(1-\gamma\right)\left(4\gamma-1\right)}{9}
 %\cdot\frac{1}{4(1-\gamma)^{2}}
 \sum_{k=\tau+2}^{T}\big(\eta_{k}^{(T)}\big)^{2} \big(\mathbb{E}\big[V_{k-1}(2)\big]\big)^{2},
	\label{eq:Var-VT-2-LB-769}
\end{align}
where the identity arises from the fact that $P_k(2\mymid 2)$ is a Bernoulli random variable with mean $p$, and the last line uses the choice \eqref{defn-parameter-p-Q-lower} of $p$ and the elementary fact $\mathbb{E}[X^2]\geq (\mathbb{E}[X])^2$.  As we will see shortly, the rationale for replacing $\mathbb{E}\big[\big(V_{k-1}(2)\big)^{2}\big]$ with $\big(\mathbb{E}\big[V_{k-1}(2)\big]\big)^{2}$ is that the mean $\mathbb{E}\big[V_{k-1}(2)\big]$ becomes quite large beyond a certain time threshold and is easier to compute.

To further control \eqref{eq:Var-VT-2-LB-769}, we develop a lower bound on $\mathbb{E}[V_{k-1}(2)]$. Given \eqref{eq:EV2-02468} and the facts that $V_0(2)=0$ and $V^{\star}(2)=\frac{3}{4(1-\gamma)}$ (see Lemma~\ref{lem:optimal-V-Q-hardMDP}), one is allowed to choose $0<\tau\leq T$ such that 
\begin{align}
\mathbb{E}\big[ V_{\tau}(2)\big] \ge \frac{1}{2(1-\gamma)} > \mathbb{E}\big[ V_{\tau-1}(2)\big].
\label{eq:choice-tau-lower-Q-generative}
\end{align}
It is also seen from \eqref{eq:EV2-02} that $\mathbb{E}\big[ V_{k}(2)\big]$ is monotonically increasing in $k$, and hence $$\mathbb{E}\big[ V_{k}(2)\big]\geq \mathbb{E}\big[ V_{\tau}(2)\big]\geq \frac{1}{2(1-\gamma)}\qquad  \text{for all } k\ge \tau; $$
namely, $\mathbb{E}[V_k(2)]$ becomes large once $k$ exceeds the threshold $\tau$. 
Combine this inequality with \eqref{eq:Var-VT-2-LB-769} to yield
\begin{align}
\mathsf{Var}\big(V_{T}(2)\big) 
%& \ge\mathbb{E}\Bigg[\sum_{k=\tau+2}^{T}\mathsf{Var}\Big(\eta_{k}^{(T)}\gamma\big(P_{k}(2\mymid2)-p\big)V_{k-1}(2)\mymid V_{k-1}(2)\Big)\Bigg] \notag\\
% & =\sum_{k=\tau+2}^{T}\big(\eta_{k}^{(T)}\gamma\big)^{2}p(1-p)\mathbb{E}\big[\big(V_{k-1}(2)\big)^{2}\big] \notag\\
 & \geq\frac{\left(1-\gamma\right)\left(4\gamma-1\right)}{9}\cdot\frac{1}{4(1-\gamma)^{2}}\sum_{k=\tau+2}^{T}\big(\eta_{k}^{(T)}\big)^{2} \notag\\
 & =\frac{ 4\gamma-1}{36\left(1-\gamma\right)}\frac{\eta^2(1-(1-\eta(1-\gamma p))^{2(T-\tau-1)})}{1-(1-\eta(1-\gamma p))^2},
	\label{eq:Var-VT-2-LB-123}
\end{align}
where the last line arises from \eqref{eq:defn-eta-kt-hard} and calculates the sum of the geometric sequence.

Recalling that $\eta(1-\gamma p) \le 1/3$ (cf.~\eqref{eq:eta-prod-gamma-p-13}), one can deduce that
\begin{align*}
\big(1-\eta(1-\gamma p)\big)^{2(T-\tau-1)}
&\le \frac{81}{16}\big(1-\eta(1-\gamma p)\big)^{2(T-\tau+1)} \\
&= \frac{81}{16} \frac{\big(V^{\star}(2) - \mathbb{E}\big[V_{T}(2)\big])^2}{\big(V^{\star}(2) - \mathbb{E}\big[V_{\tau-1}(2)\big])^2} \\
&\le \frac{81 (1-\gamma)^2}{4} \big(V^{\star}(2) - \mathbb{E}\big[V_{T}(2)\big])^2
\le \frac12,
\end{align*}
where the second line follows from \eqref{eq:Vstar2-V2-expansion-Case2}, and the two inequalities in the last line arise from 
\eqref{eq:choice-tau-lower-Q-generative} and \eqref{eq:EV2-02468}, respectively. 
Moreover, it is also readily seen that
\begin{align*}
1-(1-\eta(1-\gamma p))^2 \le 2\eta(1-\gamma p) = \frac{8\eta(1-\gamma)}{3}.
\end{align*}
Substitute the above two inequalities into~\eqref{eq:Var-VT-2-LB-123} to yield
\begin{align}
\mathbb{E}\big[\big(V^{\star}(2) - V_T(2)\big)^2\big] &\geq \mathsf{Var}\big(V_{T}(2)\big) \ge \frac{\eta(4\gamma-1)}{192(1-\gamma)^2}
\geq \frac{4\gamma-1}{192(1-\gamma)^4T} \notag\\
&\geq \frac{1}{86(1-\gamma)^4T},
\label{eq:MSE-V2-case2-135}
\end{align}
with the proviso that $\eta \geq \frac{1}{(1-\gamma)^2T}$ and $\gamma \geq 3/4$. This proves the claimed lower bound \eqref{eq:lower-bounding-VT-thm-sync} for this regime.

%Recalling from that
%
%\begin{align*}
%	 V^{\star}(2) - \mathbb{E}\big[V_{T}(2)\big] 
%	> \frac{\big(1-\eta(1-\gamma p)\big)^{T-\tau+1}}{4(1-\gamma)},
%\end{align*}
%

\subsection{Step 5: the regime with moderate learning rates}

We now turn to the regime of moderate learning rates, namely, 
$$\frac{4}{(1-\gamma)T} < \eta < \frac{1}{(1-\gamma)^2T}.$$
This is the regime in which Q-learning is most effective, and at the same time, the most challenging to analyze when developing lower bounds. 
Throughout this case, we make note of the property
\begin{equation}
	\label{eq:eta0_lb_case3}
	\eta_{0}^{(T)}\leq \frac{1}{100},
\end{equation}
which has been justified previously in \eqref{eq:EV2-0246}. 
%
%In fact, if instead $\eta_{0}^{(T)}>1/75$, then the scenario becomes much easier to cope with.  To see this, recall from the previous result~\eqref{eq:EV2-02468} that
%
%\begin{align}
%	V^{\star}(2) - \mathbb{E}\big[V_T(2)\big] 
%    =\frac{\big(1-\frac{4\eta}{3}(1-\gamma)\big)^T}{\frac{4}{3} (1-\gamma)}
%    %\ge \frac{\eta_0^{(T)}}{1-\gamma p} 
%    > \frac{1}{100(1-\gamma)},
%\end{align}
%
%which together with the assumption $T\geq \frac{1}{(1-\gamma)^2}$ yields
%
%\begin{align*}
%	\mathbb{E}\big[ \big( V^{\star}(2) - V_T(2) \big)^2 \big] 
%	&\geq \big( V^{\star}(2) - \mathbb{E} \big[V_T(2)\big] \big)^2 
%	%+ \mathsf{Var}\big(V_{T}(2)\big)  \notag\\
%	 \geq \frac{1}{10000(1-\gamma)^2}
%	 \geq \frac{1}{10000(1-\gamma)^4 T} .
%	\label{eq:MSE-V2-case3-456}
% \end{align*}
%
%
We shall focus attention on the dynamics w.r.t.~state $1$
and its associated value function $V_{t}(1)$. Note that the presence of 2 available actions at this state is deliberately designed to capture the over-estimation issue inherent in Q-learning. 

%under the condition \eqref{eq:eta0_lb_case3}. 

\paragraph{Step 5.1: decoupling the impact of initialization.} 
It is slightly more convenient to work with a Q-learning sequence initialized at the true optimal value. We therefore introduce an auxiliary sequence governed by the following update rule:
\begin{align}
	\label{eqn:Qhat-schumann}
	\widehat{Q}_t(a) = (1-\eta)\widehat{Q}_{t-1}(a) + \eta\big\{ 1 + \gamma P_{t}(1\mymid 1, a)\widehat{V}_{t-1}\big\}
\end{align}
with 
\begin{align*}
	\widehat{V}_{t-1} \coloneqq \max_a \widehat{Q}_{t-1}(a) \qquad \text{and} \qquad \widehat{Q}_0(a) \coloneqq Q^{\star}(1, a) = \frac{1}{1-\gamma p},
\end{align*}
where $ Q^{\star}(1, a)$ has been computed in Lemma~\ref{lem:optimal-V-Q-hardMDP}. Evidently, $\{\widehat{Q}_t(a)\}$ corresponds to a Q-learning sequence initialized at the ground truth. 
To decouple the effect of initialization from the original dynamics of $Q_t(1, a)$, we show that the discrepancy resulting from the different initializations of $\widehat{Q}_t(a)$ and $Q_t(1, a)$ vanishes geometrically, i.e.,
\begin{align} 
\label{eq:Qt-lower-bound-generative}
	Q_t(1, a) \ge \widehat{Q}_t(a) - \frac{1}{1-\gamma} \big(1-\eta(1-\gamma)\big)^t ,
	\qquad  a \in \{1,2\}.
\end{align}
The proof is postponed to Section~\ref{sec:proof-auxiliary-property-lower-bound}. 

In light of this bound \eqref{eq:Qt-lower-bound-generative}, we focus on bounding the auxiliary sequence $\widehat{Q}_t$ in the sequel, and defer translating the result back to $Q_t(1, a)$ until the final sub-step of the proof. 
Before proceeding, we record a crude lower bound on $\widehat{V}_t$:
\begin{align}
	\label{eqn:Tailbound-vhat-rolland}
	\mathbb{P}\bigg\{\widehat{V}_t \ge \frac{1}{4(1-\gamma)} \bigg\} \ge \frac{1}{2}
    \qquad \text{for all } t\leq T.
    %t \ge \tau.  
\end{align}
In other words, it shows that $\widehat{V}_t$ has the correct order with nontrivial probability.  The proof is postponed to Section~\ref{sec:proof-auxiliary-property-lower-bound}.

%In other words, once the iteration count exceeds a certain threshold, then $\widehat{V}_t$ attains large values with nontrivial probability.  

%As a preliminary step, we would like to establish a simple, crude lower bound on $\widehat{V}_t$, showing that it has the correct order with nontrivial probability.  

%\paragraph{Step 5.2: developing a crude lower bound on the auxiliary sequence.}  
%

\paragraph{Step 5.2: deriving recursive relations about estimation errors.}
We now study the over-estimation error of the auxiliary sequence $\widehat{Q}_t$ (cf.~\eqref{eqn:Qhat-schumann}). 
Define the error sequence
\begin{subequations}
\begin{align}
	\Delta_t(a) &\defn \widehat{Q}_t(a) - Q^{\star}(1, a), \qquad a = 1, 2; \\
	\Delta_{t,\mathsf{max}} &\defn \max_a \Delta_t(a). 
\end{align}
\end{subequations}
Our objective is to show that  $\Delta_{T,\mathsf{max}}$ is large in expectation (so that $\widehat{Q}_t(a)$ overestimates $Q^{\star}(1,a)$), thus putting the ``over-estimation'' issue that fundamentally slows convergence on a rigorous footing.

To do so, a key step lies in establishing recursive relations about $\Delta_{t,\mathsf{max}}$. 
According to the update rule \eqref{eqn:Qhat-schumann} and the Bellman optimality equation $Q^{\star}(1, a) = 1 + \gamma pV^{\star}(1)$, we have 
\begin{align*}
\Delta_t(a) 
&= (1-\eta) \widehat{Q}_{t-1}(a) + \eta\big(1 + \gamma P_{t}(1\mymid 1, a)\widehat{V}_{t-1} \big) 
-  Q^{\star}(1,a)\\
&= (1-\eta)\Delta_{t-1}(a) + \eta\big(1 + \gamma P_{t}(1\mymid 1, a)\widehat{V}_{t-1} - Q^{\star}(1, a)\big) \\
&= (1-\eta)\Delta_{t-1}(a) + \eta\gamma\big(P_{t}(1\mymid 1, a)\widehat{V}_{t-1} - pV^{\star}(1)\big) \\
	&= (1-\eta)\Delta_{t-1}(a) + \eta\gamma\big(p \big( \widehat{V}_{t-1} - V^{\star}(1) \big) + \big( P_{t}(1\mymid 1, a) - p \big) \widehat{V}_{t-1}\big) \\
&= (1-\eta)\Delta_{t-1}(a) + \eta\gamma\Big(p\Delta_{t-1,\mathsf{max}} + \big(P_{t}(1\mymid 1, a) - p \big)\widehat{V}_{t-1}\Big),
\end{align*}
where the last line holds since $\widehat{V}_{t-1} - V^{\star}(1) = \max_a \big(\widehat{Q}_{t-1}(a) - V^{\star}(1)\big) = \max_a\Delta_{t-1}(a)$. 
%(as a consequence of the relation~\eqref{eq:optimal-V-Q-hardMDP}). 
Applying this relation recursively yields
\begin{align*}
\Delta_t(a) = \sum_{k = 1}^t\eta\big(1-\eta\big)^{t-k}\gamma\Big(p\Delta_{k-1,\mathsf{max}} + \big(P_{k}(1\mymid 1, a) - p \big)\widehat{V}_{k-1}\Big),
\end{align*}
where we use the initialization $\Delta_0(a) = 0$ (by construction). Letting 
\begin{subequations}
	\label{eqn:defn-xi-t}
\begin{align}
	\xi_t(a) &\defn \sum_{k = 1}^t\eta \big(1-\eta\big)^{t-k}\gamma \big( P_{k}(1\mymid 1, a) - p \big)\widehat{V}_{k-1}, \\
	\xi_{t,\mathsf{max}} &\defn \max_a \xi_t(a),
\end{align}
\end{subequations}
%
%Setting $\Delta_t \defn \max_a \Delta_t(a)$, 
one can take the maximum over $a$ in the above display about $\Delta_t(a)$ to derive the following identity regarding the maximum error:
\begin{align}
\Delta_{t,\mathsf{max}} = \sum_{k = 1}^t\eta \big(1-\eta\big)^{t-k}\gamma p \Delta_{k-1,\mathsf{max}} + \xi_{t,\mathsf{max}}.
\label{eq:recursion-Delta-t-max-Q-lb}
\end{align}

\paragraph{Step 5.3: lower bounding $\mathbb{E} [ \xi_{t,\mathsf{max}} ]$.} 
As can be seen from the recursion \eqref{eq:recursion-Delta-t-max-Q-lb},  a crucial step is to control $\xi_{t,\mathsf{max}}$, which involves the maximum of multiple random quantities. Towards this end, we develop a lower bound on $\mathbb{E} [ \xi_{t,\mathsf{max}} ]$ as follows; the proof is deferred to Section~\ref{sec:proof-property-lb-xi}.  
\begin{lemma}
\label{lem:lower-bound-E-xi-lower-Q}
Define a threshold
\begin{align}
\label{eq:tau-hat-byproduct-duplicate}
\widehat{\tau} \defn \max\bigg\{\tau \,\,\Big|\,\, \big(1-\eta(1-\gamma p)\big)^{T-\tau+1} \le \frac{6}{7}\bigg\}	.
\end{align}
Then there exists some universal constant $c>0$ such that
\begin{align}
\label{eqn:lb-xi}
	\mathbb{E} [ \xi_{t,\mathsf{max}} ] &\ge \frac{c}{\sqrt{(1-\gamma)^2T}\log T}
	\qquad \text{for all}~t \ge \widehat{\tau}.
\end{align}
%
%for some universal constant $c>0$, where $\widehat{\tau}$ is a threshold defined as
%
\end{lemma}
\begin{remark}
 Property \eqref{eq:eta0_lb_case3} guarantees that $\widehat{\tau}$ (cf.~\eqref{eq:tau-hat-byproduct-duplicate}) is well-defined.  
\end{remark}
\begin{remark}
	Note that $\mathbb{E} [ \xi_{t,\mathsf{max}} ]$ is strictly positive due to the presence of the max operator together with the availability of more than one action. In contrast, when there is only a single  action $a$, one readily has $\mathbb{E} [ \xi_{t,\mathsf{max}} ]=\mathbb{E} [ \xi_{t}(a) ]=0$. This distinction gives rise to a fundamental difference in sample complexity between the cases $A=1$ and $A\geq 2$, leading respectively to the horizon dependency $\frac{1}{(1-\gamma)^3}$ and $\frac{1}{(1-\gamma)^4}$. 
\end{remark}
%
%In words, $\tau'$ marks a threshold where the most recent $\tau'$ iterates contribute significantly to the estimation error. 
%
Substituting \eqref{eqn:lb-xi} into \eqref{eq:recursion-Delta-t-max-Q-lb} reveals that, for all $t \ge \widehat{\tau}$,
\begin{align}
\label{eqn:Delta-iterates-exp}
	\mathbb{E} [ \Delta_{t,\mathsf{max}} ] &\ge \sum_{k = 1}^t\eta \big(1-\eta\big)^{t-k}\gamma p \mathbb{E} [ \Delta_{k-1,\mathsf{max}} ] + \frac{c}{\sqrt{(1-\gamma)^2T}\log T},
\end{align}
leaving us with a deterministic recursive relation to solve.

\paragraph{Step 5.4: solving the recursion~\eqref{eqn:Delta-iterates-exp}.}
To solve the above recursion, it is convenient to introduce the following auxiliary scalar sequence:
\begin{align}
\label{eqn:new-sequence}
x_t = (1-\eta)x_{t-1} + \eta \bigg(\gamma p x_{t-1} + \frac{c}{\sqrt{(1-\gamma)^2T}\log T} \bigg)
\end{align}
initialized at $x_{\widehat{\tau}} = 0$, where  $\widehat{\tau}$ has been defined in \eqref{eq:tau-hat-byproduct-duplicate}.  
A direct comparison reveals that
\begin{equation}
	\mathbb{E} \big[ \Delta_{T,\mathsf{max}} \big] \ge x_T,
	\label{eq:E-Delta-Tmax-xT-order}
\end{equation}
which follows readily from the iterative expansion of \eqref{eqn:new-sequence}: 
\begin{align*}
x_t = \sum_{k = \widehat{\tau}}^t\eta \big(1-\eta\big)^{t-k}\gamma p x_{k-1} + 
 \sum_{k = \widehat{\tau}}^T \eta
\big(1- \eta)^{t-k}\frac{c}{\sqrt{(1-\gamma)^2T}\log T}. 
\end{align*}
A straightforward calculation shows that the sequence \eqref{eqn:new-sequence} obeys
\begin{align*}
	x_T &= \frac{c}{\sqrt{(1-\gamma)^2T}\log T} \sum_{k = \widehat{\tau}}^T \eta
\big(1-\eta(1-\gamma p) \big)^{T-k} \\
&= \frac{c\big[1-
\big(1-\eta(1-\gamma p) \big)^{T-\widehat{\tau}+1}\big]}{(1-\gamma p)\sqrt{(1-\gamma)^2T}\log T} \\
&= \frac{3c\big[1-
\big(1-\eta(1-\gamma p) \big)^{T-\widehat{\tau}+1}\big]}{4\sqrt{(1-\gamma)^4 T}\log T}
\ge \frac{3c}{28\sqrt{(1-\gamma)^4T}\log T},
\end{align*}
where the second equality follows by calculating the sum of the geometric sequence,
and the last inequality holds as long as $\big(1-\eta(1-\gamma p)\big)^{T-\widehat{\tau}+1} \le 6/7$ (see \eqref{eq:tau-hat-byproduct-duplicate}). 
Combining this with \eqref{eq:E-Delta-Tmax-xT-order} yields
\begin{align}
	\mathbb{E} \big[ \Delta_{T,\mathsf{max}} \big] \ge x_T \ge \frac{3c}{28\sqrt{(1-\gamma)^4T}\log T} .
    \label{eq:E-Delta-lower-auxiliary-Q-lower}
\end{align}

\paragraph{Step 5.5: putting all this together.}
Thus far, we have established the claimed lower bound for the auxiliary sequence $\widehat{Q}_t$. To complete the argument, it remains to connect it to the original sequence $Q_t$. To do so, combine the lower bound \eqref{eq:E-Delta-lower-auxiliary-Q-lower} with \eqref{eq:Qt-lower-bound-generative} to yield
\begin{align*} 
	\mathbb{E} \big[ V_T(1) - V^{\star}(1) \big] &\ge \mathbb{E} \Big[  \Delta_{T,\mathsf{max}} - \frac{1}{1-\gamma}\big(1-\eta(1-\gamma)\big)^T \Big] \\
	&\ge \frac{3c}{28\sqrt{(1-\gamma)^4T}\log T} - \frac{\big(1-\eta(1-\gamma)\big)^T}{1-\gamma}.
\end{align*}
As it turns out, the last term in the above display can be easily handled with the aid of $V_T(3)$. Specifically, taking this result and \eqref{eq:Vt3-expansion-case3} collectively, we arrive at
\begin{align*} 
	\arxiv{&} \mathbb{E} \big[ V_T(1) - V^{\star}(1) \big] + \mathbb{E} \big[ V^{\star}(3) - V_T(3)  \big] \arxiv{\\}
	&\arxiv{\quad}  = \mathbb{E} \big[ V_T(1) - V^{\star}(1) \big] +  \frac{\big(1-\eta(1-\gamma)\big)^T}{1-\gamma} \book{\\}
\book{&}	\ge \frac{3c}{28\sqrt{(1-\gamma)^4T}\log T} .
\end{align*}
Apply the elementary inequality $\max\{a,b\}\geq \frac{1}{2}(a+b)$ to obtain
\begin{align*} 
	%\max \mathbb{E} \Big[ \big|V_T(s) - V^{\star}(s)\big| \Big] 
\arxiv{&} \max \Big\{ \mathbb{E} \Big[ \big|V_T(1) - V^{\star}(1) \big| \Big]	,\; \mathbb{E} \Big[ \big|V_T(3) - V^{\star}(3) \big| \Big] \Big\}  \arxiv{\\}
	&\arxiv{~~} \ge \frac{1}{2} \Big\{	\mathbb{E} \big[ V_T(1) - V^{\star}(1)   \big] +  \mathbb{E} \big[   V^{\star}(3) - V_T(3) \big] \Big\}  \book{\\}
	%&\quad \ge \max\Bigg\{\frac{\big(1-\eta(1-\gamma)\big)^T}{1-\gamma}, ~\frac{3c}{28\sqrt{(1-\gamma)^4T}\log T} - \frac{\big(1-\eta(1-\gamma)\big)^T}{1-\gamma} \Bigg\} \\
	%&\quad\ge \frac{1}{2} \cdot \frac{\big(1-\eta(1-\gamma)\big)^T}{1-\gamma} + \frac{1}{2}  \Bigg[\frac{3c}{28\sqrt{(1-\gamma)^4T}\log T} - \frac{\big(1-\eta(1-\gamma)\big)^T}{1-\gamma}\Bigg] \\
\book{&}	\geq  \frac{3c}{56\sqrt{(1-\gamma)^4T}\log T},
\end{align*}
%
%where the penultimate line holds since $\max\{a,b\}\geq \frac{1}{2}(a+b)$. 
yielding
\begin{align*} 
	\arxiv{&} \max_s \mathbb{E} \Big[ \big|V_T(s) - V^{\star}(s)\big|^2 \Big] 
  \book{&}  \geq \max_s \Big( \mathbb{E} \Big[ \big|V_T(s) - V^{\star}(s)\big| \Big]  \Big)^2 \notag\\
	&\arxiv{\quad} \geq \bigg( \frac{3c}{56\sqrt{(1-\gamma)^4T}\log T} \bigg)^2 = \frac{9c^2}{56^2 (1-\gamma)^4T \log^2 T}.
\end{align*}
We have thus established the desired lower bound for the moderate learning-rate regime, which, together with the bounds in \eqref{eq:MSE-V2-case1} and \eqref{eq:MSE-V2-case2-135} for the other two regimes, completes the proof of Theorem~\ref{thm:LB-example-Q-generative}.

\subsection{Proof of auxiliary properties}
\label{sec:proof-auxiliary-property-lower-bound}

\paragraph{Proof of property~\eqref{eq:Qt-lower-bound-generative}.}
We prove this claim by induction. The base case $t=0$ follows trivially.
Assume now the hypothesis holds for $t-1$.
By combining the inductive hypothesis \eqref{eq:Qt-lower-bound-generative} for the $(t-1)$-th iteration with the definitions of $V_{t-1}$ and  $\widehat{V}_{t-1}$, we obtain 
\begin{align*} 
	V_{t-1}(1) = \max_{a} Q_{t-1}(1, a) 
	&\ge \max_{a}  \widehat{Q}_{t-1}(a) - \frac{\big(1-\eta(1-\gamma)\big)^{t-1}}{1-\gamma} \\
	&= \widehat{V}_{t-1} - \frac{\big(1-\eta(1-\gamma)\big)^{t-1}}{1-\gamma}.
\end{align*}
Given the update rules of $Q_t(1, a)$ and $\widehat{Q}_t(a)$, their difference evolves as
\begin{align*}
\arxiv{&}Q_t(1, a) - \widehat{Q}_t(a) \arxiv{\notag\\}
&\arxiv{\quad} = (1-\eta)\big(Q_{t-1}(1, a) - \widehat{Q}_{t-1}(a)\big) + \eta\gamma P_{t}(1\mymid 1, a)\big(V_{t-1}(1) - \widehat{V}_{t-1}\big) \\
&\arxiv{\quad}\ge -(1-\eta)\frac{\big(1-\eta(1-\gamma)\big)^{t-1}}{1-\gamma} - \eta\gamma P_{t}(1\mymid 1, a) \frac{\big(1-\eta(1-\gamma)\big)^{t-1}}{1-\gamma} \\
&\arxiv{\quad}\ge -(1-\eta)\frac{\big(1-\eta(1-\gamma)\big)^{t-1}}{1-\gamma} - \eta\gamma  \frac{\big(1-\eta(1-\gamma)\big)^{t-1}}{1-\gamma} \\
&\arxiv{\quad}= -\frac{\big(1-\eta(1-\gamma)\big)^{t}}{1-\gamma}, 
\end{align*}
where the first inequality invokes the induction hypothesis for the $(t-1)$-th iteration, and the second follows since $P_t(1 \mid 1, a) \le 1$.  
This completes the inductive step and, in turn, establishes the result for all $t \ge 0$.
%\end{proof}
%
%As we shall discuss momentarily, the gap $\frac{1}{1-\gamma}\big(1-\eta(1-\gamma)\big)^t$ is sufficiently small for this case. 

\paragraph{Proof of property~\eqref{eqn:Tailbound-vhat-rolland}.}
We find it convenient to introduce a second auxiliary sequence defined by the recursion:
\begin{subequations}
\label{eq:sequence-overline-Qt}
\begin{align}
	\overline{Q}_t &= (1-\eta)\overline{Q}_{t-1} + \eta\big\{ 1 + \gamma P_{t}(1\mymid 1, 1)\overline{Q}_{t-1}\big\},\\ 
	%\qquad \text{and} \qquad 
	\overline{Q}_0 &= V^{\star}(1) = \frac{1}{1-\gamma p}. 
\end{align}
\end{subequations}
In words, this sequence corresponds to the Q-learning iterates in a simplified setting with only a single available action, effectively removing the max operator. 
Given that $\widehat{V}_t =\max_a \widehat{Q}_t(a) \ge \widehat{Q}_t(1)$, it is straightforward to verify that, for all $t\geq 0$, 
\begin{align}
	\widehat{Q}_t(1) \ge (1-\eta)\widehat{Q}_{t-1}(1) + \eta\big\{1 + \gamma P_{t}(1\mymid 1, 1)\widehat{Q}_{t-1}(1)\big\}
	\ge \overline{Q}_t.
	\label{eq:ordering-Qt-hat-overline}
\end{align}
%
%. 
%In words, removing the max operator yields a smaller sequence of Q-estimates. 
%This auxiliary sequence 
%$\{\overline{Q}_t\}$ introduced in \eqref{eq:sequence-overline-Qt} 
%serves as a useful tool for establishing this bound, as its simpler structure allows us to more easily characterize the evolution of the lower envelope of the Q-iterates.

%This property enables us to lower bound $\widehat{V}_t$ through the simpler dynamics of $\overline{Q}_t$. 
%
%\paragraph{A useful lower bound on the auxiliary sequence~\eqref{eqn:Qhat-schumann}.}
%
%We now establish a lower bound on the sequence~\eqref{eqn:Qhat-schumann} introduced above. 
%

%The auxiliary sequence $\{\overline{Q}_t\}$ introduced in \eqref{eq:sequence-overline-Qt} serves as a useful tool for establishing this bound, as its simpler structure allows us to more easily characterize the evolution of the lower envelope of the Q-iterates.

We use the sequence $\overline{Q}_{t}$ to lower-bound $\widehat{V}_t$, as its simpler structure allows us to more easily characterize the evolution of the lower envelope of the Q-iterates.
First, observe that $\overline{Q}_t$ admits the decomposition: %(akin to the derivation in \eqref{eq:Vt2-expansion-case1})
\begin{align}
\notag \overline{Q}_{t} & =\big(1-\eta(1-\gamma p)\big)\overline{Q}_{t-1}+\eta\big\{1+\gamma\big(P_{t}(1\mymid1,1)-p\big)\overline{Q}_{t-1}\big\}\\
\notag & =\big(1-\eta(1-\gamma p)\big)^{t}\overline{Q}_{0} \arxiv{\notag\\}
\arxiv{&\qquad} +\sum_{k=1}^{t}\eta\big(1-\eta(1-\gamma p)\big)^{t-k}\big\{1+\gamma\big(P_{k}(1\mymid1,1)-p\big)\overline{Q}_{k-1}\big\} \notag\\
  & =\eta_{0}^{(t)}  \frac{1}{1-\gamma p} +\sum_{k=1}^{t}\eta_{k}^{(t)}+\sum_{k=1}^{t}\eta_{k}^{(t)}\gamma\big(P_{k}(1\mymid1,1)-p\big)\overline{Q}_{k-1} \notag\\
 & =\frac{1}{1-\gamma p}+\sum_{k=1}^{t}\underset{\eqqcolon\,z_{k}}{\underbrace{\eta_{k}^{(t)}\gamma\big(P_{k}(1\mymid1,1)-p\big)\overline{Q}_{k-1}}},
	\label{eqn:qbar-up-decomp}
\end{align}
where the last line results from \eqref{eq:eta-sum-hat}. It then comes down to controlling $\sum_k z_k$.

By construction, the sequence $\{z_k\}$ forms a martingale satisfying
\begin{align*}
	\mathbb{E}\big[ z_k\mymid &P_{k-1}(1\mymid 1, 1), \ldots, P_{1}(1\mymid 1, 1)\big] = 0 \arxiv{ \\} \book{\qquad}
	\text{and} \arxiv{ &} \qquad |z_k| \le \max_{1 \le k \le t} \eta_k^{(t)} \cdot \frac{\gamma p}{1-\gamma},
\end{align*}
where the last inequality follows from the basic property $0\leq \overline{Q}_{k-1}\leq \frac{1}{1-\gamma}$ (akin to Lemma~\ref{lemma:non-negativity-Qt-Vt}) and the fact that $\big| P_{k}(1\mymid1,1)-p\big|\leq \max \{p, 1-p\}=p$ since $p = \frac{4\gamma-1}{3\gamma}$ and $\gamma \geq 3/4$. In order to invoke Freedman's inequality to control \eqref{eqn:qbar-up-decomp}, we need to bound the conditional variance. 
Given these properties and the fact that $P_k(1\mymid 1,1)$ is a Bernoulli random variable with mean $p$, we obtain
\begin{align*}
&\sum_{k = 1}^t \mathsf{Var}\Big(z_k\mymid P_{k-1}(1\mymid 1, 1), \ldots, P_{1}(1\mymid 1, 1)\Big) = \sum_{k = 1}^t \big( \eta_k^{(t)}\big)^2p(1-p)\big(\gamma\overline{Q}_{k-1}\big)^2 \\
&\le \max_{1 \le k \le t} \eta_k^{(t)} \cdot \sum_{k = 1}^t \eta_k^{(t)} \cdot \frac{1}{3(1-\gamma)} \le \frac{\max_{1 \le k \le t} \eta_k^{(t)}}{4(1-\gamma)^2}.
\end{align*}
Here, the penultimate inequality relies on the fact $0\leq \overline{Q}_{k-1}\leq \frac{1}{1-\gamma}$ (akin to Lemma~\ref{lemma:non-negativity-Qt-Vt}) and the choice of $p$, 
whereas the last inequality results from the following condition (derived through \eqref{eq:eta-sum-hat})
\[
	\sum_{k = 1}^t \eta_k^{(t)} = \big(1 - \eta_0^{(t)}\big)\frac{1}{1-\gamma p} \le \frac{1}{1-\gamma p} = \frac{3}{4(1-\gamma)} .
\]
Freedman's inequality (see Lemma~\ref{lem:Freedman-prelim} with $K_0=1$) then implies that
\begin{align} 
	\label{eq:variance}
\bigg|\sum_{k = 1}^t z_k\bigg| \le \sqrt{\frac{4\max_{1 \le k \le t} \eta_k^{(t)}}{(1-\gamma)^2}\log\frac{2}{\delta}} + \frac{4\max_{1 \le k \le t} \eta_k^{(t)}}{3(1-\gamma)}\log\frac{2}{\delta} 
\end{align}
with probability at least $1-\delta$.

%From this inequality, we see that a key ingredient towards bounding $\sum_{k = 1}^t z_k$ lies in controlling the quantity $\max_{1 \le k \le t} \eta_k^{(t)}$. To this end, notice that 
To continue, observe that
\begin{align}
	\label{eqn:schubert-winterreise}
	\max_{1 \le k \le t} \eta_k^{(t)} \le \eta \le \frac{1}{(1-\gamma)^2T} \le \frac{1}{50},\qquad\text{for all}~t \ge 1,
\end{align} 
provided that $T\geq \frac{50}{(1-\gamma)^2}$. 
Setting $\delta = 1/2$ in \eqref{eq:variance} then yields $$\sum_{k = 1}^t z_k \geq -\frac{1}{2(1-\gamma)}$$ with probability at least $1/2$. Combine this with the decomposition~\eqref{eqn:qbar-up-decomp} and the property \eqref{eq:ordering-Qt-hat-overline} to conclude
\begin{align*}
	\widehat{V}_t \geq \widehat{Q}_t(1) \geq \overline{Q}_t \geq \frac{1}{1-\gamma p} -\frac{1}{2(1-\gamma)}
	=  \frac{1}{4(1-\gamma)}
\end{align*}
with probability at least $1/2$, where the last identity relies on the choice \eqref{defn-parameter-p-Q-lower} of $p$.
This completes the proof of \eqref{eqn:Tailbound-vhat-rolland}. 
%\end{proof}

\subsection{Proof of Lemma~\ref{lem:lower-bound-E-xi-lower-Q}}
\label{sec:proof-property-lb-xi}
In light of the definition~\eqref{eqn:defn-xi-t}, it is convenient to express, for every $t$,  
\begin{align*}
	\mathbb{E} [\xi_{t,\mathsf{max}}] &= \mathbb{E} \left[ \frac{\xi_t(1)+\xi_t(2)+|\xi_t(1)-\xi_t(2)|}{2} \right] 
	= \mathbb{E} \left[ \frac{|\xi_t(1)-\xi_t(2)|}{2}  \right] \\
	&= \frac{1}{2}\mathbb{E} \left[ \, \left|\sum_{k = 1}^t\eta\big(1-\eta\big)^{t-k}\gamma \big( P_{k}(1\mymid 1, 1) -  P_{k}(1\mymid 1, 2) \big)\widehat{V}_{k-1} \,\right| \right] ,
\end{align*}
where we have used the fact that $\mathbb{E}[\xi_t(a)] = 0$. Importantly, the summation within the absolute value operator is a martingale sum. To be more precise, define
\begin{align*}
	\zeta_t &\defn \sum_{k = 1}^t  \underset{\eqqcolon\, z_k}{\underbrace{ \eta\big(1-\eta\big)^{t-k}\gamma \big(P_{k}(1\mymid 1, 1) - P_{k}(1\mymid 1, 2) \big)\widehat{V}_{k-1} }},
\end{align*}
%
%for any $k\geq 1$, 
where $\{z_k\}$ forms a martingale sequence since
\begin{align*}
	\mathbb{E}\Big[ z_k\mymid \big\{ P_{j}(1\mymid 1, 1), P_{j}(1\mymid 1, 2) \big\}_{1\leq j< k} \Big] = 0.
\end{align*}

In order to control $\mathbb{E}[|\zeta_t|]$, we first seek to analyze the tail of $|\zeta_t|$. 
To do so, we claim that, for any $0<\delta<1$,
%invoking Freedman's inequality (Lemma~\ref{lem:Freedman-prelim}) shows that
%
\begin{align}
\label{eqn:tail-bunny}
\mathbb{P}\left\{\big|\zeta_t\big| \ge \sqrt{\frac{8\log\frac{2}{\delta}}{3(1-\gamma)}\sum_{k = 1}^t\eta^2\Big[\big(1-\eta\big)^{t-k}\Big]^2} + \frac{4\eta \log\frac{2}{\delta}}{3(1-\gamma)} \right\} \le \delta.
\end{align}
%
%for any $0<\delta < 1$. 
To validate this bound, we first notice that 
\begin{align}
\label{eqn:unif-bound}
|z_k| \le \max_{k: 1 \le k \le t} \eta\big(1-\eta\big)^{t-k} \cdot \frac{1}{1-\gamma} \le \frac{\eta}{1-\gamma},
\end{align}
where we have used $|\widehat{V}_{k-1}|\leq \frac{1}{1-\gamma}$ according to Lemma~\ref{lemma:non-negativity-Qt-Vt}. In addition, we can further bound the aggregate condition variance as
\begin{align*}
	& \sum_{k = 1}^t \mathsf{Var}\big(z_k\mymid P_{k-1}(1\mymid 1, 1), P_{k-1}(1\mymid 1, 2), \ldots, P_{1}(1\mymid 1, 1), P_{1}(1\mymid 1, 2)\big) \\
	&\qquad = \sum_{k = 1}^t\eta^2\Big[\big(1-\eta\big)^{t-k}\Big]^2\cdot2p(1-p)\cdot\big(\gamma\widehat{V}_{k-1}\big)^2   \\
    &\qquad\le \sum_{k = 1}^t\eta^2\Big[\big(1-\eta\big)^{t-k}\Big]^2\cdot \frac{2}{3(1-\gamma)}, 
\end{align*}
where the last inequality comes from the fact that $|\widehat{V}_{k-1}| \le \frac{1}{1-\gamma}$, the assumption that $\gamma\geq 3/4$ and the choice $p = \frac{4\gamma - 1}{3\gamma}$.
These bounds, taken together with Freedman's inequality (see Lemma~\ref{lem:Freedman-prelim} with $K_0=1$), justify the tail bound \eqref{eqn:tail-bunny}.

Next, we make note of a crude deterministic bound: 
\begin{align}
\big|\zeta_t\big| \le  \sum_{k=1}^t |z_k| \leq  \left[ \sum_{k=1}^t \eta\big(1-\eta\big)^{t-k} \right] \cdot \frac{1}{1-\gamma} \leq \frac{1}{1-\gamma},
\label{eq:zeta-t-UB-crude-Q-lower-generative}
\end{align} 
which is a consequence of the facts that $|\widehat{V}_{k-1}| \le \frac{1}{1-\gamma}$ (see Lemma~\ref{lemma:non-negativity-Qt-Vt}) and $\sum_{k=1}^t \eta (1-\eta)^{t-k}\leq \eta \cdot \frac{1}{1-(1-\eta)}=1$. This immediately implies that 
\[
    \delta_0\coloneqq \frac{(1-\gamma)^2}{2}\mathbb{E} \big[ |\zeta_t|^2 \big] \leq \frac{1}{2}.
\]
Thus, setting $\delta = \delta_0 \in (0,1)$ in 
 \eqref{eqn:tail-bunny}  reveals that, with probability at least $1-\delta_0$, 
\begin{align} \label{eq:B-delta}
|\zeta_t| \le B	\defn 
\sqrt{\frac{8\log\frac{2}{\delta_0}}{3(1-\gamma)} \sum_{k = 1}^t\eta^2\Big[\big(1-\eta\big)^{t-k}\Big]^2} + \frac{4\eta \log\frac{2}{\delta_0}}{3(1-\gamma)}.
\end{align}
%
%with $\delta_0 = \frac{(1-\gamma)^2}{2}\mathbb{E} \big[ |\zeta_t|^2 \big] \in $. 
%
When $T \ge \frac{1}{(1-\gamma)^2}$, one can guarantee that 
\begin{align}
	\mathbb{E} [\xi_{t,\mathsf{max}}]  
	&= \frac{1}{2}\mathbb{E} \big[ |\zeta_t| \big]
	\ge \frac{1}{2}\mathbb{E} \big[ |\zeta_t|\mathds{1}\big(|\zeta_t| \le B\big) \big]  
	\ge \frac{1}{2B}\mathbb{E} \big[ |\zeta_t|^2\mathds{1}\big(|\zeta_t| \le B\big) \big] \nonumber \\
	&= \frac{1}{2B}\Big\{ \mathbb{E} [ |\zeta_t|^2 ] - \mathbb{E} \big[ |\zeta_t|^2\mathds{1}\big(|\zeta_t| > B\big) \big] \Big\} \nonumber \\
	&\overset{(\mathrm{i})}{\geq} \frac{1}{2B}\Big\{ \mathbb{E} [ |\zeta_t|^2 ] - \frac{1}{(1-\gamma)^2} \mathbb{P}\big\{|\zeta_t| > B\big\} \Big\} \nonumber \\
	&\ge \frac{1}{2B}\Big\{ \mathbb{E} [ |\zeta_t|^2 ] - \frac{\delta}{(1-\gamma)^2} \Big\} 
	\overset{(\mathrm{ii})}{\geq} \frac{1}{4B}\mathbb{E} [ |\zeta_t|^2 ]. \label{eq:emax_intermediate_lb}
	% &\stackrel{(i)}{\ge} \frac{c}{\sqrt{(1-\gamma)^2T}\log T},
\end{align}
Here, (i) follows from \eqref{eq:zeta-t-UB-crude-Q-lower-generative}, while (ii) holds due to the choice of $\delta_0$.

It thus suffices to lower bound $\mathbb{E} [|\zeta_t|^2]$ 
for $t \ge \widehat{\tau}$, where we recall the definition
\begin{align}
\label{eq:tau-hat-byproduct}
\widehat{\tau} \defn \max\Bigg\{\tau \,\,\Big|\,\, \big(1-\eta(1-\gamma p)\big)^{T-\tau+1} \le \frac{6}{7}\Bigg\}	.
\end{align}
According to \eqref{eq:eta0_lb_case3}, it satisfies $1\leq \widehat{\tau}\leq T$. First, observe that for any $t \ge \widehat{\tau}$, the martingale properties about $\{z_k\}$ give
\begin{align*}
\mathbb{E}\big[|\zeta_{t}|^{2}\big] & \ge\sum_{k=1}^{t}\mathbb{E}\left[\mathsf{Var}\Big(\eta\big(1-\eta\big)^{t-k}\gamma\big(P_{k}(1\mymid1,1)-P_{k}(1\mymid1,2)\big)\widehat{V}_{k-1}\mymid\widehat{V}_{k-1}\Big)\right]\\
 & \geq\sum_{k=1}^{t}\mathbb{E}\bigg[\mathsf{Var}\Big(\eta\big(1-\eta\big)^{t-k}\gamma\big(P_{k}(1\mymid1,1)-P_{k}(1\mymid1,2)\big)\widehat{V}_{k-1}\mymid\widehat{V}_{k-1}\Big)\\
 & \qquad\qquad\cdot\mathds{1}\bigg\{\widehat{V}_{k-1}\ge\frac{1}{4(1-\gamma)}\bigg\}\bigg] \\
 & \ge \sum_{k = 1}^t\eta^2 \Big[\big(1-\eta\big)^{t-k}\Big]^2 \gamma^2\cdot  \frac{2p(1-p)}{16(1-\gamma)^2} 
    \mathbb{P}\bigg\{\widehat{V}_{k-1}\ge\frac{1}{4(1-\gamma)}\bigg\},
\end{align*}
where the first line relies on \eqref{eq:variance-expansion-martingale}, and the last step makes use of the constraint $\widehat{V}_{k-1} \ge \frac{1}{4(1-\gamma)}$. 
To further control the right-hand side, we take $\tau' \defn \max\big\{t - \frac{1}{\eta}, 1\big\} $ and show that 
%\yxc{TODO}
%
\begin{align}
	\mathbb{E} \big[ |\zeta_t|^2 \big]  
	%&\overset{\text{(i)}}{\ge} \sum_{k = 1}^t\eta^2 \Big[\big(1-\eta\big)^{t-k}\Big]^2 \gamma^2\cdot 2p(1-p) \frac{1}{16(1-\gamma)^2} 
    %\mathbb{P}\bigg\{\widehat{V}_{k-1}\ge\frac{1}{4(1-\gamma)}\bigg\}
	%\nonumber\\
    &\overset{\text{(i)}}{\ge} \frac{1}{2}\sum_{k = 1}^t\eta^2 \Big[\big(1-\eta\big)^{t-k}\Big]^2 \gamma^2\cdot  \frac{2p(1-p)}{16(1-\gamma)^2} 
	\nonumber\\
	&\ge \frac{1}{48(1-\gamma)}\sum_{k = 1}^t\eta^2 \Big[\big(1-\eta\big)^{t-k}\Big]^2 \nonumber\\
&\ge \frac{1}{48(1-\gamma)}
	\sum_{k = \tau'}^t\eta^2 \Big[\big(1-\eta\big)^{t-k}\Big]^2 \overset{\text{(ii)}}{\ge} \frac{\eta}{1200(1-\gamma)}, \label{eq:square-zeta}
\end{align}
%
%and the last step makes use of the fact \eqref{eqn:Tailbound-vhat-rolland}. 
%
where 
%(i) makes use of the constraint $\widehat{V}_{k-1} \ge \frac{1}{4(1-\gamma)}$, (ii) 
(i) arises from \eqref{eqn:Tailbound-vhat-rolland}, 
and (ii) is valid as long as the following property holds for $t\geq \widehat{\tau}$ (which will be proven towards the end of this subsection)
\begin{align}
	\sum_{k = \tau'}^t\eta^2 \Big[\big(1-\eta\big)^{t-k}\Big]^2
	\ge \frac{1}{24} \eta.
	\label{eq:sum-LB-1/8-890}
\end{align}

We are now well equipped to control $\mathbb{E} [\xi_{t,\mathsf{max}}] $
using the property \eqref{eq:emax_intermediate_lb}. Recalling the expression of $B$ in \eqref{eq:B-delta}, we see that bounding $\mathbb{E} [ |\zeta_t|^2 ] / B$ reduces to  controlling the following two quantities: 
\begin{align}
\frac{\mathbb{E} [ |\zeta_t|^2 ]}{\frac{\eta}{1-\gamma}\log\frac{2}{\delta_0}} \qquad \mbox{and}\qquad \frac{\mathbb{E} [ |\zeta_t|^2 ]}{\sqrt{\frac{\log\frac{2}{\delta_0}}{1-\gamma}\sum_{k = 1}^t\eta^2 \big[(1-\eta)^{t-k}\big]^2}} . 
	\label{eq:two-terms-LB-123}
\end{align}
\begin{itemize}
\item
For the first term in \eqref{eq:two-terms-LB-123}, recalling that $\delta_0 = \frac{(1-\gamma)^2}{2}\mathbb{E} \big[ |\zeta_t|^2 \big]$, we can demonstrate that
\begin{align}
\log \frac{1}{\delta_0} &= -\log \frac{(1-\gamma)^2}{2}\mathbb{E} \big[ |\zeta_t|^2 \big] \le -\log \frac{(1-\gamma)\eta}{2400} \notag\\
&\le \log (9600T) ,
%\lesssim \log T ,
	\label{eq:log-delta-T-relation-123}
\end{align}
where the first inequality is due to \eqref{eq:square-zeta}, and the second inequality arises since $\eta \geq \frac{4}{(1-\gamma)T}$ (given the range of the learning rates in this case). 
Combine this with \eqref{eq:square-zeta} to yield 
\[
\frac{\mathbb{E} [ |\zeta_t|^2 ]}{\frac{\eta}{1-\gamma}\log\frac{2}{\delta_0}} \gtrsim \frac{1}{\log T}. 
\]

\item
Turning to the second term in \eqref{eq:two-terms-LB-123}, we observe that
\begin{align*}
\arxiv{&}\frac{\mathbb{E} \big[ |\zeta_t|^2 \big]}{\sqrt{\frac{\log\frac{2}{\delta_0}}{1-\gamma}\sum_{k = 1}^t\eta^2 \big[(1-\eta)^{t-k}\big]^2}}  \book{&} \overset{\mathrm{(i)}}{\gtrsim} \sqrt{\frac{1}{(1-\gamma)\log T}\sum_{k = 1}^t\eta^2 \big[(1-\eta)^{t-k}\big]^2} \\
&\arxiv{ \qquad } \overset{\mathrm{(ii)}}{\gtrsim}  \sqrt{\frac{\eta}{(1-\gamma)\log T}} \overset{\mathrm{(iii)}}{\gtrsim} \frac{1}{\sqrt{(1-\gamma)^2T\log T}}.
\end{align*}
Here, (i) follows from \eqref{eq:square-zeta} and \eqref{eq:log-delta-T-relation-123} since
\[
	\mathbb{E} \big[ |\zeta_t|^2 \big] \gtrsim \frac{1}{1-\gamma}\sum_{k = 1}^t\eta^2 \big[(1-\eta)^{t-k}\big]^2\qquad \text{and} \qquad \log\frac{2}{\delta_0} \lesssim \log T;
\]
(ii) arises from \eqref{eq:sum-LB-1/8-890}; and (iii) relies on the fact $\eta \gtrsim \frac{1}{(1-\gamma)T}$ (given the range of the learning rates considered in this step).
\end{itemize}

Substituting the above relations into  \eqref{eq:emax_intermediate_lb} and using the expression of $B$ in \eqref{eq:B-delta}, we reach 
\begin{align*}
	\mathbb{E} [ \xi_{t, \mathsf{max}} ] \ge \frac{1}{4B}\mathbb{E} \big[ |\zeta_t|^2 \big]
	&\ge \frac{c}{\sqrt{(1-\gamma)^2T}\log T}
\end{align*}
for some numerical constant $c > 0$. Thus, this proves property~\eqref{eqn:lb-xi}, as long as the claim \eqref{eq:sum-LB-1/8-890} can be established. 

\begin{proof}[Proof of the claim \eqref{eq:sum-LB-1/8-890}]
Recall that $\tau' \defn \max\big\{t - \frac{1}{\eta}, 1\big\}$. For any $k$ obeying $\tau'\le k \le t$, one has 
\[
\eta\big(1-\eta\big)^{t-k} \ge \eta\big(1-\eta\big)^{t-\tau'} \ge
\eta\big(1-\eta\big)^{1/\eta} 
\ge \frac{1}{3}\eta,
	%,\qquad\text{for all}~\tau'\le k \le t.
\]
where the last inequality is valid as long as $\eta\leq 0.1$. 
Moreover, it is seen that for any $t\geq \widehat{\tau}$, 
\begin{align*}
	\sum_{k = \tau'}^t\eta\big(1-\eta\big)^{t-k} &= 1 - \big(1-\eta\big)^{t-\tau'+1}
	\ge 1-\max\left\{\big(1-\eta\big)^{1/\eta+1}, \, \big(1-\eta\big)^t\right\} \\
	& \ge 1-\max\Big\{e^{-1}, \, \big(1-\eta\big)^{\widehat{\tau}}\Big\} 
	\ge \frac{1}{8}. 
\end{align*}
Here, the first identity can be derived via the sum of a geometric sequence; to see why the last inequality holds, note that according to the definition \eqref{eq:tau-hat-byproduct} and property~\eqref{eq:eta0_lb_case3},
\begin{align*}
	&\big(1-\eta(1-\gamma p)\big)^{T-\widehat{\tau}} > \frac{6}{7} \\
	\Longrightarrow \quad \big(1-\eta(1-\gamma p)\big)^{\widehat{\tau}} &< \frac{\big(1-\eta(1-\gamma p)\big)^{T}}{6/7} = \frac{\eta_0^{(T)}}{6/7} \leq \frac{1/100}{6/7} \leq \frac{7}{8},
\end{align*}
which in turn implies that
\begin{align}
\big(1-\eta\big)^{\widehat{\tau}} \le \big(1-\eta(1-\gamma p)\big)^{\widehat{\tau}} 
\leq \frac{7}{8}.
\label{eq:prod-etai-tau-T-2345}
\end{align}
Putting the preceding two bounds together yields, for any $t\geq \widehat{\tau}$, 
\begin{align*}
\sum_{k=\tau'}^{t}\eta^{2}\Big[\big(1-\eta\big)^{t-k}\Big]^{2} & \geq\min_{k:\,\tau'\leq k\leq t}\left\{ \eta\big(1-\eta\big)^{t-k}\right\} \sum_{k=\tau'}^{t}\eta\big(1-\eta\big)^{t-k}\\
 & \geq\frac{1}{3}\eta\cdot\frac{1}{8}\geq\frac{1}{24}\eta
\end{align*}
as claimed. 
\end{proof}

%\end{proof}

\section{Notes}

%\paragraph{Model-free RL.}

The asymptotic convergence of TD learning and Q-learning --- namely, 
their convergence behavior as the number of iterations tends to infinity, with all other parameters held fixed 
--- has been extensively studied since its early development, e.g.,  \citet{tsitsiklis1997analysis,tsitsiklis1994asynchronous,jaakkola1993convergence,szepesvari1997asymptotic,borkar2000ode,devraj2021q,li2023statistical}.
Recently, substantial research efforts have shifted toward understanding the non-asymptotic performance of model-free methods,  which elucidates how key problem parameters --- in addition to the number of iterations --- affect algorithmic performance. For TD learning, finite-time, finite-sample analyses have been developed in a growing body of recent work, including but not limited to \citet{bhandari2018finite, lakshminarayanan2018linear, srikant2019finite, xu2019two, doan2019finite, cai2019neural, chen2020finite, mou2020linear, kaledin2020finite,chen2021finite-offpolicy,khamaru2021instance,li2024q,chen2022finite,patil2023finite,li2024high,mou2024optimal,samsonov2024improved,srikant2025rates,chen2025concentration,durmus2025finite}. 
In parallel, finite-time, finite-sample performance guarantees for Q-learning and its variants have been established in, e.g., \citet{kearns1999finite, even2003learning,murphy2005generalization,beck2012error,ghavamzadeh2011speedy,wainwright2019stochastic,wainwright2019variance,chen2020finite,qu2020finite,xu2020finite,li2021sample,chen2021lyapunov,chen2022finite,li2024q,jin2024feasible,jiao2026sample}.

Since TD learning and Q-learning can be viewed as instances of stochastic approximation, a notable portion of the existing theoretical analyses builds upon classical techniques from stochastic approximation theory, along with various modern extensions of these techniques. Prominent examples include the ODE method (e.g., \citet{borkar2000ode,borkar2025ode,meyn2022control}),  Stein's method (e.g., \citet{srikant2019finite,srikant2025rates}), Lyapunov-drift approach (e.g., \citet{chen2021lyapunov,cayci2023sample,chen2025non,chen2026non}), and extensions to two-time-scale settings (e.g., \citet{konda2004convergence}), among others. See also the tutorial \citet{meyn2023stability} for the stability of Q-learning with linear function approximation.  The continued development of these techniques has provided versatile tools for analyzing not only TD learning and Q-learning, but also a range of model-free variants, including TD learning with gradient correction (TDC), gradient TD learning, Zap Q-learning, SARSA, momentum Q-learning, double Q-learning, and so on \citep{borkar2000ode,sutton2009fast,maei2009convergent,gupta2019finite,liu2015finite,devraj2017zap,xu2019two,xu2021sample,zou2019finite,weng2020mean,weng2020momentum,xiong2020finite}. It is worth noting, however, that these stochastic approximation techniques alone may be insufficient to establish the sharp performance guarantees for TD learning and Q-learning in Theorems~\ref{thm:policy-evaluation-TD} and \ref{thm:infinite-horizon-simple}. Achieving these tight guarantees requires additional proof ideas that carefully exploit the intrinsic variance structure of the MDP, a key ingredient underlying the state-of-the-art analysis in \citet{li2024q}.

In addition, while the sample inefficiency and overestimation bias of Q-learning have long been observed in practice --- motivating algorithmic variants such as double Q-learning \citep{hasselt2010double} --- formal lower bounds on the iteration complexity of Q-learning were not established until \citet{li2024q}. Notably, \citet{wainwright2019stochastic} conjectured, based on numerical evidence, that the tight dependence of the sample complexity of Q-learning on the effective horizon scales as $\widetilde{O}\big(\frac{1}{(1-\gamma)^4}\big)$. However, prior to \citet{li2024q}, neither matching upper nor lower bounds confirming this scaling were available.

With the aim of improving the sample complexity of Q-learning while preserving its model-free nature, more sophisticated algorithmic techniques have been proposed and analyzed in the literature. For example, a variance-reduced variant of Q-learning --- drawing inspiration from variance-reduction methods in stochastic optimization, particularly stochastic variance-reduced gradient (SVRG) \citep{johnson2013accelerating} --- has been shown to achieve the minimax-optimal sample complexity for all $\varepsilon \in (0,1]$ \citep{wainwright2019variance}.  At a high level, this approach partitions the iterative process into epochs. At the beginning of each epoch, a batch of samples is used to construct increasingly more accurate reference estimates of the value function. Subsequent stochastic approximation updates incorporate these references by combining them with fresh stochastic samples to reduce variability, thereby accelerating convergence. We refer interested readers to \citet{wainwright2019variance,li2021sample,mou2022optimal,xia2024instance} for further details on both synchronous and asynchronous Q-learning with variance reduction.
Notably, however, no existing model-free algorithm is known to attain minimax optimality when $\varepsilon \gg 1$.  This suggests the inherent statistical suboptimality of model-free approaches in the most sample-hungry regime. Finally, we note that variance-reduction techniques have also been incorporated into TD learning for policy evaluation; see, for example, \citet{khamaru2021temporal,xu2020reanalysis,li2023accelerated}. 

%% file: chapters/onlineRL.tex
\chapter{Online RL}
\label{chapter:online-RL}

We now turn to online reinforcement learning, where an agent interacts with an unknown environment sequentially to collect data and learn a near-optimal policy. In contrast to the generative model setting --- where arbitrary state–action pairs can be queried on demand --- an agent in online RL must collect data by executing policies and observing trajectories in real time.  This restriction couples data collection with decision-making, necessitating carefully designed exploration strategies that acquire informative samples without sacrificing long-term performance. Consequently, a central challenge of online RL lies in optimally balancing exploration and exploitation under a limited sampling budget.

\section{Problem formulation}
\label{sec:formulation-online}

In this chapter, we focus primarily on online episodic RL formulated in finite-horizon nonstationary MDPs, a setting that has been studied most extensively and for which the statistical limits have been characterized even in the most sample-limited regime. Its counterpart in discounted infinite-horizon MDPs remains not fully solved to date (see, e.g., \citet{lattimore2012pac,strehl2006pac,dong2019q,zhang2021model,szita2010model,zhou2021provably,ji2023regret2}) and calls for further investigation.

\paragraph{Sampling protocol.} 
To be precise, we model the unknown environment as a finite-horizon nonstationary MDP $\mathcal{M}=(\mathcal{S},\mathcal{A},H, P, r)$. For ease of exposition, we assume that the reward function $r=\{r_h\}_{h=1}^H$ is fully known to the agent, an assumption that can be relaxed substantially without fundamentally altering the underlying statistical limits and incurs only minor additional technical complications. By contrast, the transition kernel $P=\{P_h\}_{h=1}^H$ is assumed to be unknown {\em a priori}, constituting the primary source of uncertainty in the online RL setting studied in this chapter.  The agent is permitted to interact with the MDP over $K$ episodes,  each of length $H$. 

In each episode $k=1,\ldots, K$, the sampling protocol of an online RL algorithm proceeds as follows: 
\begin{subequations}
\label{eq:generation-s1k-shk-online}
\begin{itemize}
	\item[(i)] {\em Initialization.} The initial state $s_1^k$ of each episode $k$ is drawn independently according to 
    \begin{align}
    \label{eq:generation-s1k-online}
    s_1^k \overset{\mathrm{i.i.d.}}{\sim} \rho, \qquad k=1,\cdots,K,
    \end{align}
    where $\rho \in \Delta(\mathcal{S})$ denotes an (unknown) initial state distribution. 

    \item[(ii)] {\em Policy selection.}  Based on the information collected in episodes $1$ through $k-1$, the agent selects a policy $\pi^k =\{\pi_h^k \}_{h=1}^H$.  

    \item[(iii)] {\em Sampling.} The agent executes policy $\pi^k$ in the MDP $\mathcal{M}$, generating a sample trajectory $\big\{ (s_h^k,a_h^k,r_h^k) \big\}_{1\leq h\leq H}$,  
    where $s_h^k$, $a_h^k$ and $r_h^k=r_h(s_h^k,a_h^k)$ denote the state, action and immediate reward at step $h$ of this episode, respectively. Here, starting from $s_1^k$, the trajectory is rolled out sequentially for $ h= 1,\dots, H$ according to 
    \begin{align}
    	a_{h}^k\sim \pi_h^k (\cdot \mymid s_h^k)
    	\qquad \text{and}\qquad
    	s_{h+1}^k \sim P_h(\cdot \mymid s_h^k, a_h^k).
        \label{eq:generation-shk-online}
    \end{align}

\end{itemize}
\end{subequations}
For notational convenience, we set $r_{H+1}=0$ throughout this chapter. Consequently, learning the terminal transition kernel $P_H$ is unnecessary, as all actions yield zero rewards at step $H+1$. Since each trajectory has length $H$, the total sample size can be understood as   
$$T=KH.$$

\paragraph{Exploration-exploitation tradeoff.}
A fundamental challenge in online RL arises from the exploration–exploitation tradeoff: the agent must collect informative samples to learn the unknown transition kernel, while simultaneously seeking policies that maximize cumulative rewards. Compared to the generative model setting, at least two aspects significantly complicate matters.
\begin{itemize}
    \item {\em Coverage of the state-action space.} Effective learning typically requires adequate coverage of the state–action space, which is difficult to guarantee under online sampling protocols. In particular, the agent does not know {\em a priori} which regions of the state–action space are reachable through sample rollouts, nor how to reach them. This contrasts sharply to the generative model setting, where one can query arbitrary state–action pairs at any step of the MDP.

    \item {\em Complicated statistical dependency across time.} Samples in online RL are collected sequentially and depend on the agent's past decisions. This creates complex statistical dependencies across both samples and time. In contrast, the samples studied in Chapters~\ref{chapter:generative-model} and \ref{chapter:generative-model-MF} under the generative model are statistically independent, a feature that greatly simplifies both algorithm design and theoretical analysis.

\end{itemize}

\paragraph{Performance metric: regret.} 
To evaluate the performance of an online RL algorithm, 
arguably one of the most widely used metrics is the (cumulative) regret over all $K$ episodes:   
\begin{align}
	\mathsf{Regret}(K) \coloneqq \sum_{k=1}^K \left( V^{\star}_{1}(s_1^k) -V^{\pi^k}_1(s_1^k) \right).
	\label{eq:defn-regret}
\end{align}
This metric quantifies the aggregate performance gap --- expressed in terms of the resulting value functions ---  between the agent and an oracle that always acts optimally. Importantly, regret provides a natural way to formalize the exploration–exploitation tradeoff: it captures both the cost of exploration, since exploratory actions may yield lower rewards, and the cost of insufficient exploitation, as inaccurate model estimates arising from limited data can adversely affect decision-making.
Achieving low regret implies that the algorithm can efficiently identify near-optimal policies while keeping the cumulative performance loss under control. 

The main focus of this chapter is the design of online RL algorithms with provably small cumulative regret $\mathsf{Regret}(K)$.

\paragraph{Connection between regret and PAC statistical guarantees.} 
Careful readers may observe that the policies $\{\pi^{k}\}_{1\leq k\leq K}$ deployed across episodes are generally different, which naturally raises the question of whether one can instead learn a single policy that attains near-optimal values. Indeed, this question was the primary focus of Chapters~\ref{chapter:generative-model}--\ref{chapter:generative-model-MF} and falls within the PAC (probably approximately correct) learning framework, which seeks a single policy estimate $\widehat{\pi}$ satisfying
\[
	V_1^{\star}(\rho) - V_1^{\widehat{\pi}}
    (\rho) \leq \varepsilon.
\]
Here, we recall that $\rho$ denotes the state distribution generating the initial states $\{s_1^k\}_{1\leq k\leq K}$ (see \eqref{eq:generation-s1k-online}). 
Fortunately, regret guarantees can often be readily translated into corresponding PAC learning guarantees (see, e.g., \citet{jin2018q}).

To illustrate this connection, consider an online RL algorithm that, for any MDP with $S$ states, $A$ actions and horizon $H$, satisfies the expected regret bound
\begin{subequations}
\label{eq:PAC-from-regret-conversion}
\begin{equation}
	\mathbb{E}\big[\mathsf{Regret}(K)\big]
	%=\sum_{k=1}^K \mathbb{E}\big[ V^{\star}_{1}(s_1^k) -V^{\pi^k}_1(s_1^k) \big] 
	%= \sum_{k=1}^K \mathbb{E}\big[ V^{\star}_{1}(s) -V^{\pi^k}_1(s) \big] 
	\leq C_{S,A,H} K^{1-\alpha}
	\label{eq:PAC-from-regret-conversion-regret}
\end{equation}
for some $\alpha\in [0,1]$ and some quantity $C_{S,A,H}$ that may depend on $(S,A,H)$. 
Here, the expectation is taken over all randomness arising from the initial states, sample transitions, and the learning algorithm's internal randomization. Now construct the policy estimate $\widehat{\pi}$ by selecting one of the deployed policies uniformly at random: 
\begin{align}
		\label{eq:uniform-policy-pihat}
    \widehat{\pi} \sim \mathsf{Unif}\big(\pi^1,\pi^2,\dots,\pi^K \big).
\end{align}
Then with probability at least $1-\delta$, 
\begin{align}
    V_1^{\star}(\rho) - V_1^{\widehat{\pi}}(\rho) 
	&\leq C_{S,A,H+1}K^{-\alpha} \delta^{-1}.
	\label{eq:PAC-from-regret-conversion-pac}
\end{align}
\end{subequations}

\begin{proof}[Proof of property~\eqref{eq:PAC-from-regret-conversion-pac}]
Let us first consider the special case where the initial state distribution $\rho$ is supported on a single state $s$, namely,
$\rho(s)=1$ and $\rho(s')=0$ for all $s'\neq s$. 
Markov's inequality asserts that, with probability at least $1-\delta$, 
\begin{align*}
    V_1^{\star}(s) - V_1^{\widehat{\pi}}(s) 
	& \le \delta^{-1} \mathbb{E}\big[V^{\star}_1(s)-V^{\widehat{\pi}}_1(s)\big]  =  \frac{1}{K\delta}\sum_{k=1}^K \mathbb{E} \big[ V^{\star}_1(s)-V^{\pi^k}_1(s) \big] \notag\\
	& = \frac{1}{K\delta} \mathbb{E} \big[ \mathsf{Regret}(K) \big]
	\leq C_{S,A,H}K^{-\alpha} \delta^{-1},
\end{align*}
where the first equality follows from the definition \eqref{eq:uniform-policy-pihat} of $\widehat{\pi}$. 
%as claimed. 

	We now extend the argument to an arbitrary initial state distribution $\rho$.  To this end, we augment the MDP by introducing an additional step 0, consisting of a single  state $s_0$, a single action $a_0$, zero immediate reward (i.e., $r_0(s_0,a_0)=0$), and transition kernel $P_0(\cdot\mymid s_0,a_0) = \rho$. Since there is only one available action, this additional step involves no decision-making; accordingly, with a slight abuse of notation, we continue to denote by $\pi$ the obvious extension of a policy $\pi$ for the original MDP and write $V_0^\pi$ for its value function starting from step~0.  The online RL algorithm can therefore be applied to the augmented MDP without modification, yielding
%
%\begin{subequations}
%\label{eq:PAC-from-regret-conversion-extended}
\begin{equation}
	\mathbb{E}\big[\mathsf{Regret}(K)\big] = \sum_{k=1}^K \mathbb{E}\big[ V^{\star}_{0}(s_0) -V^{\pi^k}_0(s_0) \big] 
	%=\sum_{k=1}^K \mathbb{E}\big[ V^{\star}_{1}(s_1^k) -V^{\pi^k}_1(s_1^k) \big] 
	%= \sum_{k=1}^K \mathbb{E}\big[ V^{\star}_{1}(s) -V^{\pi^k}_1(s) \big] 
	\leq C_{S,A,H+1} K^{1-\alpha}
	\label{eq:PAC-from-regret-conversion-regret-extended}
\end{equation}
Given that the augmented MDP has a singleton initial state, the preceding argument applies directly, implying that
\begin{align}
    V_0^{\star}(s_0) - V_0^{\widehat{\pi}}(s_0) \leq C_{S,A,H+1}K^{-\alpha} \delta^{-1}
	\label{eq:V0-UB-13579}
\end{align}
with probability at least $1-\delta$. 
To finish up, observe that for every policy $\pi$, one has $V_0^{\pi}(s_0) = V_1^{\pi}(\rho) $, 
as the additional initial step incurs zero reward and admits only a single action. This together with \eqref{eq:V0-UB-13579} completes the proof.  
\end{proof}

	This simple argument demonstrates that sublinear regret naturally yields sample-efficient PAC statistical guarantees. For example, if an online RL algorithm enjoys the regret bound $\mathbb{E}[\mathsf{Regret}(K)]\leq \widetilde{O}(\sqrt{SAH^3K}) $ for any MDP with $S$ states, $A$ actions and horizon length $H$, then the randomly selected policy $\widehat{\pi}$ achieves (by choosing $\delta = \frac{1} {\log (SAH)}$ in \eqref{eq:PAC-from-regret-conversion-pac}) 
\begin{align*}
    V_1^{\star}(s) - V_1^{\widehat{\pi}}(s) 
	&\leq \widetilde{O}\Bigg( \sqrt{\frac{SAH^3}{K}} \,\Bigg)
\end{align*}
with probability $1-o(1)$; consequently, to ensure $V_1^{\star}(s) - V_1^{\widehat{\pi}}(s) \leq \varepsilon$ with probability approaching 1, it suffices to run this online RL algorithm for
\[
	K= \widetilde{\Theta}\bigg( \frac{SAH^3}{\varepsilon^2} \bigg)
	\quad \text{episodes}. 
\]

\input{chapters/bandits.tex}

\section{A model-based online RL algorithm: \ucbvi}
\label{sec:ucbvi-intro}

Recall that in the generative model setting, model-based algorithms have been shown to achieve minimax-optimal sample complexity, even in the most data-hungry regime. Such results provide strong incentives for studying the model-based paradigm in the online RL setting as well. Given the proven efficiency of the optimism principle in multi-armed bandits (see Section~\ref{sec:bandits-UCB}), it is natural to investigate how this principle can be effectively integrated into model-based online RL algorithms.

\subsection{Algorithmic ideas}

To begin with, the model-based policy learning algorithm introduced previously consists of a model estimation stage, where an empirical estimate of the probability transition kernel is constructed from collected data, followed by a planning stage that computes a policy by treating the empirical model as if it were the true environment. This naturally suggests a paradigm that integrates model-based policy learning with online decision-making and sampling, which operates as follows in each episode.
\begin{itemize}
	\item \emph{Model estimation.} Use all previously collected data to construct an empirical estimate $\widehat{P}$ of the transition kernel.

	\item \emph{Policy update.} Compute an optimal policy $\widehat{\pi}$ w.r.t.~the updated empirical model $\widehat{P}$, for instance via dynamic programming. 

	\item \emph{Sampling.} Execute the updated policy $\widehat{\pi}$ to obtain a new sample trajectory at the beginning of the next episode.

\end{itemize}
While conceptually intuitive, this approach lacks one crucial component: exploration. By focusing exclusively on exploitation --- optimizing solely w.r.t.~the current empirical model --- the agent may fail to sufficiently explore regions of the state-action space that are poorly understood yet potentially highly rewarding.

To address this issue, a natural strategy is to incorporate the optimism principle in the face of uncertainty into the above framework, namely, to guide exploration by acting according to an optimistic view of the world. Compared with stochastic bandits (Section~\ref{sec:bandits-UCB}), where uncertainty arises primarily from noisy rewards, the online RL formulation considered herein faces uncertainty stemming from unknown transition kernels. Consequently, there are multiple objects around which one may construct confidence regions, including the transition kernel itself and the corresponding optimal value and Q-functions. The former (i.e., confidence regions for the transition kernel) led to some of the earliest UCB-style algorithms for online RL, most notably \ucrl and \ucrltwo \citep{auer2006logarithmic,jaksch2010near}, whereas the latter  (i.e., optimism over the optimal value and Q-function) led to the development of \ucbvi \citep{azar2017minimax}, the first online RL algorithm achieving asymptotically near-optimal regret. Since the primary focus of this work is sample efficiency, we concentrate on the \ucbvi algorithm and its variants.

In a nutshell, the \ucbvi algorithm --- short for {\em upper confidence bound value iteration} and originally proposed by \citet{azar2017minimax} --- implements the optimism principle by constructing upper confidence bounds on the optimal Q-function in each episode, and subsequently performing policy updates and data collection in accordance with these optimistic Q-value estimates rather than the unadjusted empirical estimates. To be more precise, in each episode $k$, the algorithm proceeds as follows, assuming the Q-value estimates are initialized optimistically as $Q_h^{\ucb}(s,a) = H$ for all $(s,a,h)\in \cS\times \cA\times [H]$. 
\begin{itemize}

	\item {\em Sampling.} Compute the current policy $\pi^{\ucb}=\{\pi^{\ucb}_h\}_{1\leq h\leq H}$ w.r.t.~the optimistic Q-estimates $\{Q^{\ucb}_h\}_{1\leq h\leq H}$ via a greedy rule, i.e.,
		\[
			\pi^{\ucb}_h(s) \leftarrow \arg\max_a Q^{\ucb}_h(s,a),\qquad \forall (s,h)\in \cS\times [H].
		\]
        Armed with the updated policy estimate $\pi^{\ucb}$, the agent executes $\pi^{\ucb}$ in the MDP at the beginning of the episode,  collecting a sample trajectory denoted by  $$\big(s_1^{k},a_1^{k},\dots,s_H^{k},a_H^{k},s_{H+1}^{k}\big) .$$
    
    \item {\em Model estimation.} Compute an empirical estimate $\widehat{P}=\{\widehat{P}_h\in \mathbb{R}^{SA\times S}\}_{1\leq h\leq H}$ of the transition probability kernel using all previously collected samples. More specifically, the empirical model $\widehat{P}$ is constructed such that,  for all $(s,a,h,s')\in \cS\times \cA \times [H] \times \cS$,  
		\begin{equation}
			  \widehat{P}_{h}(s' \mymid s,a) = \frac{N_h(s,a,s')}{N_h(s,a)}, 
		\end{equation}
		where $N_h(s,a)$ denotes the number of visits to the state-action-step tuple $(s,a,h)$ up to the current episode, and $N_h(s,a,s')$ denotes the number of such visits that result in a transition to state $s'$ at step $h+1$. Throughout this section, we often use the notation $\widehat{P}_{s,a,h} \coloneqq [\widehat{P}_{h}(s' \mymid s,a)]_{s'\in \cS}$ to denote the empirical transition vector associated with the state-action pair $(s,a)$ at step $h$. 
        If the tuple $(s,a,h)$ has not been visited, one may simply set $\widehat{P}_{s,a,h}(s')=1/S$ for all $s'\in \cS$, so that $\widehat{P}_{s,a,h}$  remains a valid probability distribution.

        %\yc{do we need to discuss the case $N_h(s,a)=0$?}   

	\item {\em Optimistic Q-function estimation.} 
		Maintain upper confidence estimates of the optimal Q-function and value function. More concretely, working backward for $h=H,\ldots, 1$, one computes, for every state-action-step tuple,  
\begin{align*}
	Q_{h}^{\ucb}(s,a)\, & \leftarrow \,\min\big\{{r}_{h}(s,a)+\big\langle\widehat{P}_{h}(\cdot\mymid s,a),V^{\ucb}_{h+1}\big\rangle+b_{h}(s,a),H\big\}
	\\
V^{\ucb}_{h}(s)\, & \leftarrow \, \max\nolimits_{a}Q^{\ucb}_{h}(s,a) 
\end{align*}
from the initial value $V_{H+1}^{\ucb}=0$, where 
 $Q^{\ucb}_{h}$ (resp.~$V^{\ucb}_h$) indicates the running estimate for the Q-function (resp.~value function), 
		and $b_{h}(s,a)\geq 0$ is some suitably chosen exploration bonus that accounts for statistical uncertainty. 
		Here, the iterate $Q_{h}^{\ucb}(s,a)$ is explicitly capped at $H$ to reflect the known feasible range $[0,H]$. 

\end{itemize}
\noindent 
The full procedure is described in Algorithm~\ref{alg:main-ucbvi}. %Note that the policy update in Algorithm~\ref{alg:main-ucbvi} is presented at the beginning of each episode, rather than at the end, purely for notational convenience. 

At a high level, the \ucbvi algorithm makes decisions adaptively by favoring actions with either large estimated Q-values or high uncertainty in the Q-value estimates, thereby balancing exploitation with exploration in a principled manner. From a computational standpoint, each episode requires running a value-iteration-type procedure, whose complexity is roughly comparable to that of solving a fully specified finite-horizon MDP. Consequently, the overall computational complexity of \ucbvi is polynomial in the problem parameters.

\begin{algorithm}[t]
	\DontPrintSemicolon
\caption{The upper confidence bound value iteration (\ucbvi) algorithm for episodic online RL \label{alg:main-ucbvi}}
%\begin{algorithmic}[ht]

	{\small 

	\textbf{input:}  number of episodes $K$, reward function $r$, confidence parameter $\delta$, constants $c_1, c_2>0$. \\
	%$c_1=\frac{460}{9}$, $c_2 = 2\sqrt{2}$, $c_3=\frac{544}{9}$.  
	
	\textbf{initialize } 
	%set $\delta' \leftarrow \frac{\delta}{200SAH^2K^2}$; and set 
	%$N^{\mathsf{all}}_h(s,a,s')\leftarrow 0$, 
	 $N_h(s,a,s')\leftarrow 0$, $N_h(s,a)\leftarrow 0$, $Q^{\ucb}_h(s,a)\leftarrow H$, $V^{\ucb}_h(s)\leftarrow H$, $V^{\ucb}_{H+1}(s)\leftarrow 0$, $\widehat{P}_{s,a,h}(s')\leftarrow 1/S$ for all $(s,a,s',h)\in \mathcal{S}\times \mathcal{A}\times\mathcal{S}\times [H]$. \\
%\textbf{initialization: } 
	\For{$k=1,\cdots,K$} {
    {\color{blue}\tcc{policy update.}} 
		compute policy $\pi^{\ucb}$ such that $\pi^{\ucb}_h(s) \leftarrow \arg\max_{a}Q^{\ucb}_h(s,a)$ for all $(s,h)\in \mathcal{S}\times  [H]$.         {\label{line:policy-update-ucbvi}} \\
        observe the initial state $s_1 \overset{\mathrm{ind.}}{\sim} \rho$ of a new episode. \\ 
		\For {$h=1,\cdots,H$} {
			%observe $s_{h}^k$, 
            {\color{blue}\tcc{sampling.}} 
			take $ a_h = \pi^{\ucb}_h(s_h)  $, 
			%$ a_h^k= \arg\max_{a}Q_h(s_h^k,a)$, 
			observe $s_{h+1}$; set $(s,a,s')\leftarrow (s_h,a_h,s_{h+1})$.  \label{line:choose_action-ucbvi}
			\\
			
			{\color{blue}\tcc{empirical model update.}}
			 $N_h(s,a) \leftarrow  N_h( s,a )+1$; $N_h(s,a,s') \leftarrow   N_h(s,a,s')+1$. \\
	%	{\color{blue}\tcc{updates using data of this epoch.\label{line:a1-ucbvi}}}
%$N_h(s,a)\leftarrow \sum_{\widetilde{s}}N_h(s,a,\widetilde{s})$.  
		%	{\color{blue}\tcp{number of visits to $(s,a,h)$ in this epoch.} \label{line:Nh-update-ucbvi}}
			$\widehat{P}_{s,a,h}(\widetilde{s}) \leftarrow  \frac{N_h(s,a,\widetilde{s})}{N_h(s,a)}$ for all $\widetilde{s} \in \mathcal{S}$.  \label{line:P-hsa-update-ucbvi}

		}
		{\color{blue}\tcc{optimistic updates of Q-functions \& values.}}
			%set $V^{\ucb}_{H+1}(s)\leftarrow 0$ for all $s\in \mathcal{S}$. \\
			\For{$h=H,\cdots,1$} {
				\For{$(s,a)\in \mathcal{S}\times \mathcal{A}$} {
					compute the bonus $b_h(s,a)$ according to \eqref{eq:bh-defn-Bernstein-ucbvi}; update
					\begin{subequations}
						\label{eq:updateqv-ucbvi}
					\begin{align} 
						Q^{\ucb}_h(s,a) &\leftarrow \min\big\{    r_h(s,a)  +\langle \widehat{P}_{s,a,h}, V^{\ucb}_{h+1} \rangle +b_h(s,a)    ,\,   H\big\}, \label{eq:updateq-ucbvi}\\
						V^{\ucb}_{h}(s) &\leftarrow \max_{a}Q^{\ucb}_h(s,a).
						\label{eq:updatev-ucbvi}
					\end{align}
					\end{subequations}
				}
			}
		}

	}

%\end{algorithmic}
\end{algorithm}

\subsection{Choices of exploration bonuses}
\label{sec:choice-bonus-online}
A crucial design element of \ucbvi is the exploration bonus $b_h(s,a)$, which needs to be carefully constructed to ensure a valid, yet tight, upper confidence bound on the optimal Q-function (i.e., $Q_h^{\ucb}(s,a)\geq Q^{\star}_h(s,a)$). As is evident from the algorithm description, the primary source of uncertainty in Q-function estimation in each iteration of dynamic programming arises from the term $\big\langle\widehat{P}_{h}(\cdot\mymid s,a),V^{\ucb}_{h+1}\big\rangle$ appearing in \eqref{eq:updateq-ucbvi}, whereas the ideal quantity would be the population-level counterpart $\big\langle P_{h}(\cdot\mymid s,a),V^{\ucb}_{h+1}\big\rangle$. 
Two common choices for designing $b_h(s,a)$ are obtained by applying the Hoeffding inequality or the Bernstein inequality to control the deviation 
\begin{align}
	\big\langle\widehat{P}_{h}(\cdot\mymid s,a)-P_{h}(\cdot\mymid s,a),V^{\ucb}_{h+1}\big\rangle,
	\label{eq:stochastic-term-ucbvi}
\end{align}
which we introduce below. Throughout, we let $\delta>0$ denote a confidence parameter (so that $1-\delta$ is the target success probability), and recall that $N_h(s,a)$ is the visitation count of the state-action-step tuple $(s,a,h)$. 
\begin{itemize}
	\item {\em Hoeffding-style bonus.} This type of bonuses is comparatively simple, as it depends mainly on the range of the value function. More specifically, it typically takes the following form: 
	\begin{equation*}
		b_h(s,a) =\widetilde{\Theta} \bigg( \sqrt{\frac{H^2}{N_h(s,a)}} \bigg).
		%\log\frac{SAHK}{\delta}
		%\label{eq:bh-defn-Hoeffding-ucbvi}
	\end{equation*}
	%
	%where  $c_0>0$ some universal constant. 
	This choice of upper confidence width is obtained by bounding the deviation term in \eqref{eq:stochastic-term-ucbvi} via the Hoeffding inequality (see Lemma~\ref{lem:Hoeffding-prelim}) together with the fact that, by construction, the optimistic value estimate $V^{\ucb}_{h+1}(s)$ falls within $[0,H]$.

	\item {\em Bernstein-style bonus.}  In contrast, the Bernstein-type bonus incorporates variance information to form tighter confidence bounds, often leading to improved statistical guarantees compared to their Hoeffding-based counterparts. By applying the Bernstein inequality (see Lemma~\ref{lem:Bernstein-prelim}) and using the fact that $\|V^{\ucb}_{h+1}\|_{\infty}\leq H$, we see that the deviation term \eqref{eq:stochastic-term-ucbvi} can be bounded, with high probability, by
		\[
			\widetilde{O} \bigg( \sqrt{\frac{   \mathsf{Var}_{P_{s,a,h} }(V^{\ucb}_{h+1})   }{ N_h(s,a) }}
						+\frac{H}{ N_h(s,a)   } \bigg),
		\]
		where we recall the definition of the variance term in \eqref{eq:defn-Var-Psah-V-finite}. Nevertheless, since the true transition kernel $P_{s,a,h}$ is unknown, this variance term $\mathsf{Var}_{P_{s,a,h}}(\cdot)$ cannot be evaluated directly in practice. A natural remedy is to replace it with the empirical variance $\mathsf{Var}_{\widehat{P}_{s,a,h}}(\cdot)$ computed under the empirical model. This yields the following Bernstein-style exploration bonus:
\begin{align*}
	b_h(s,a) 
	%\asymp 
	=\widetilde{\Theta} 
	\Bigg( \sqrt{\frac{   \mathsf{Var}_{\widehat{P}_{s,a,h} }(V^{\ucb}_{h+1})   }{ N_h(s,a) }}
						+ \frac{H}{ N_h(s,a)   }  \Bigg)
	%\label{eq:bh-defn-Bernstein-ucbvi-motivate}
\end{align*}
arising from the empirical Bernstein inequality stated in Lemma~\ref{empirical bernstein}.  
\end{itemize}
\noindent 
We shall shortly see --- to be formalized in Lemma~\ref{lemma:opt} --- that the above exploration bonuses suffice to guarantee, with high probability, that $Q_h^{\ucb}(s,a)\geq Q^{\star}_h(s,a)$ holds for every state-action-step tuple $(s,a,h)$, thereby offering valid upper confidence bounds.

Motivated by insights from the generative model setting, which highlight the importance of exploiting variance structure for data-efficient design, this monograph focuses on adopting and analyzing Bernstein-style bonuses within \ucbvi and its variants. 
To facilitate precise discussion and analysis, we shall work with the following concrete choice throughout this chapter: 
\begin{subequations}
	\label{eq:bh-defn-Bernstein-ucbvi}
\begin{align}
	b_h(s,a) = c_1 \sqrt{\frac{   \mathsf{Var}_{\widehat{P}_{s,a,h} }(V^{\ucb}_{h+1}) \log\frac{1}{\delta'}  }{ N_h(s,a) }}
						+c_2\frac{H \log\frac{1}{\delta'} }{ N_h(s,a)   } ,
\end{align}
where  $c_1,c_2>0$ are some universal constants, and $\delta'$ is a suitably adjusted confidence parameter. Throughout the analysis of \ucbvi, we take  
\begin{align} \label{eq:definition_delta_prime}
	\delta' = \frac{\delta}{100 S^2A^2H^2K^2}; 
\end{align}
\end{subequations}
this particular choice of $\delta'$ made solely for technical convenience  and should not be interpreted as a practical recommendation.

\begin{remark}
\label{remark:different-bonus-than-Azar}
We note that the form of \eqref{eq:bh-defn-Bernstein-ucbvi} differs from the Bernstein-style bonus proposed in the original work of \citet{azar2017minimax}; instead, we adopt a simplified choice suggested by \citet{zhang2020reinforcement,zhang2025settling} in their design of \mvp, an algorithm we will shortly introduce in Section~\ref{sec:mvp-intro}. 
\end{remark}

\subsection{Theoretical guarantees}
\label{sec:theory-ucbvi}
As alluded to previously, the development of the \ucbvi algorithm \citep{azar2017minimax} marked a milestone in online RL theory, leading to the first algorithm provably achieving {\em asymptotically} minimax-optimal regret. Its statistical guarantees are formally stated in the following theorem. 
\begin{theorem}[\citet{azar2017minimax}]
	\label{thm-ucbvi-regret}  
Consider any $K \ge 1$ and any $0<\delta<1$. Then with probability at least $1-\delta$, the \ucbvi algorithm in Algorithm~\ref{alg:main-ucbvi} with Bernstein-style bonuses \eqref{eq:bh-defn-Bernstein-ucbvi} achieves
\begin{equation}
	\mathsf{Regret}(K)
	\le \widetilde{O} 
	\big(  \sqrt{SAH^{3}K} + S^2AH^3 \big),
	%\leq \widetilde{O}\left(\min\big\{\sqrt{SAKH^3},\,HK\big\}\right)
	\label{eq:regret-ucbvi-thm}
\end{equation}
 provided that the constants $c_1,c_2>0$ in \eqref{eq:bh-defn-Bernstein-ucbvi} are suitably chosen. Here, in the definition of $\mathsf{Regret}(K)$ (cf.~\eqref{eq:defn-regret}), $\pi^k$ and $s_1^k$ denote, respectively, the sampling policy deployed and the initial state observed in episode $k$ of Algorithm~\ref{alg:main-ucbvi}. 
	%where the regret metric will be formally defined in \eqref{eq:defn-regret}.  
\end{theorem}
\begin{remark}
	It is noteworthy that the regret bound presented in \citet{azar2017minimax} differs from that in Theorem~\ref{thm-ucbvi-regret} in its dependence on the horizon $H$, since the former considers finite-horizon \emph{stationary} MDPs, whereas Theorem~\ref{thm-ucbvi-regret} is stated for finite-horizon \emph{nonstationary} MDPs.  Intriguingly, this also unveils that the \ucbvi algorithm is asymptotically optimal for finite-horizon stationary MDPs.
\end{remark}

Notably, once the number of episodes is sufficiently large, i.e.,
\begin{align}
	%\big(\text{burn-in cost of \citet{azar2017minimax}}\big) \qquad 
	K \gtrsim S^3A H^3, 
    \label{eq:burn-in-ucbvi}
\end{align}
the lower-order term $S^2AH^3$ becomes negligible, and the regret bound in Theorem~\ref{thm-ucbvi-regret} simplifies to
\begin{equation}
	\mathsf{Regret}(K)
	\le \widetilde{O} 
	\big(  \sqrt{SAH^{3}K} \big).
	\label{eq:regret-asymptotic-ucbvi}
\end{equation}
As we will rigorize in Section~\ref{sec:minimax-lower-bound-online}, this matches the minimax lower bound up to logarithmic factors. Consequently, \ucbvi is asymptotically optimal in terms of regret.

Additionally, when $K \gtrsim S^3AH^3$, the above asymptotic regret bound \eqref{eq:regret-asymptotic-ucbvi} can be readily converted into a PAC-style sample complexity guarantee via the regret-to-PAC conversion in \eqref{eq:PAC-from-regret-conversion} from Section~\ref{sec:formulation-online}. Specifically, if one outputs a policy $\widehat{\pi}$ drawn uniformly at random from those sampling policies deployed over the $K$ episodes, then
\[
	V_1^{\star}(\rho) - V_1^{\widehat{\pi}}(\rho) \leq \varepsilon
\]
holds with probability approaching one, provided that the total number of collected samples satisfies
\[
	T=KH \geq \widetilde{\Omega}\bigg(\frac{SAH^4}{\varepsilon^2} \bigg) 
\]
and $\varepsilon$ is sufficiently small. Somewhat surprisingly, this PAC-style sample complexity coincides, up to logarithmic factors, with the minimax-optimal sample complexity limit under the generative model setting (see Section~\ref{sec:finite-horizon-generative}). In some sense, this result suggests that, at least in the asymptotic regime and from a minimax perspective, online exploration via optimism incurs little additional statistical cost compared to sampling from an idealized generative model.

Nevertheless, the statistical optimality of \ucbvi is guaranteed only after a substantial burn-in period of order
$
K \gtrsim S^3AH^3,
$
which can be prohibitively large in practical scenarios where the available number of episodes is limited. This limitation of the existing theoretical guarantees for \ucbvi\ has motivated a line of subsequent work aimed at reducing or eliminating the burn-in cost while retaining optimal regret guarantees.

%In other words, the \ucbvi algorithm is asymptotically optimal in terms of its regret, with a burn-in cost $K \gtrsim S^3A H^3$ needed to guarantee regret optimality. It is worth noting that this burn-in requirement is undesirable, since the sample size available in many practical scenarios might be far below this requirement. 

%A close inspection reveals that this regret matches the minimax lower bound \eqref{eq:minimax-lower-bound-1} if and only if 
%%
%\begin{align}
%	\big(\text{burn-in cost of \citet{azar2017minimax}}\big) \qquad K \gtrsim S^3A H^3, 
%\end{align}
%
%due to the presence of the lower-order term $H^3S^2A$ in the regret bound. 
%This burn-in cost is clearly undesirable, 
%since the sample size available in many practical scenarios might be far below this requirement. 

%where we recall that $\delta$ is the confidence parameter in Algorithm~\ref{alg:main-ucbvi} and $\delta' = \frac{\delta}{200SAH^2K^2}$ \yxc{edit}.

\subsection{Notation and intuition}
\label{sec:intuition-ucbvi}

Before embarking on the proof, we provide some intuition for why the asymptotic regret scales as $\widetilde{O}(\sqrt{SAH^3K})$. 

\paragraph{Notation.}
To facilitate the discussion and subsequent analysis, 
we introduce augmented notation to explicitly index the iterates of Algorithm~\ref{alg:main-ucbvi} by the episode number $k$ as follows. 
\begin{itemize}

    \item $(s_1^k,a_1^k,\cdots,s_H^k,a_H^k,s_{H+1}^k)$: the sample trajectory observed during episode $k$, with $s_h^k$ (resp.~$a_h^k$) representing the state (resp.~action) at step $h$. 

    \item $\pi^k$: the policy $\pi^{\ucb}$ computed and deployed in episode $k$; note that $\pi^k$ is computed at the beginning of episode $k$ and depends only on data collected prior to episode $k$. 

	\item  $\widehat{P}^k_{s,a,h}\in \mathbb{R}^{S}$: the latest empirical transition probability vector $\widehat{P}_{s,a,h}$ computed {\em before} episode $k$. 

	\item  $N_h^k(s,a)$: the visitation count $N_h(s,a)$ for the $(s,a,h)$-tuple {\em before} episode $k$.  
    
	\item  $b_h^k(s,a) \geq 0 $: the latest bonus term $b_h(s,a)$ {\em before} episode $k$.  

	\item  $Q_h^k\in \mathbb{R}^{SA}$: the Q-function estimate $Q_h^{\ucb}$ {\em before} episode $k$.  

	\item  $V_h^k\in \mathbb{R}^{S}$: the value function estimate $V_h^{\ucb}$ {\em before} episode $k$.

\end{itemize}
%
%\noindent It is also worth noting that the policy iterate $\pi^k$ (cf.~line~\ref{line:policy-update-ucbvi} of Algorithm~\ref{alg:main-ucbvi}), computed at the beginning of episode $k$, is a deterministic function of $Q_h^k$, depending only on data collected prior to episode $k$.

\paragraph{Heuristic arguments.} 
Below, we provide a heuristic discussion of the regret analysis. To simplify the exposition, we consider the asymptotic regime with $K\rightarrow \infty$, where we may informally regard $V_h^k\approx V_h^{\star}\approx V_h^{\pi^k}$ and $\widehat{P}^k\approx P$. 
These approximations allow us to focus on the main bottlenecks while ignoring
lower-order error terms.

As in the multi-armed bandit setting in Section~\ref{sec:bandits-UCB}, an appropriate choice of exploration
bonuses guarantees that the value estimates remain optimistic, namely,
$V_1^k\ge V_1^\star$ (which will be rigorized shortly). This
optimism property immediately yields
\begin{align}
\mathsf{Regret}(K) & =\sum_{k=1}^{K}\Big(V_{1}^{\star}(s_{1}^{k})-V_{1}^{\pi^{k}}(s_{1}^{k})\Big)\leq\sum_{k=1}^{K}\Big(V_{1}^{k}(s_{1}^{k})-V_{1}^{\pi^{k}}(s_{1}^{k})\Big),
	\label{eq:regret-UB1-heuristic}
\end{align}
which is oftentimes more convenient to work with because $\pi^k$ is chosen greedily
w.r.t.~the optimistic Q-values; as a result, both $V_1^k$ and $V_1^{\pi^k}$ can be obtained through compatible
Bellman recursions, allowing their difference to be propagated
recursively and conveniently across the horizon.   

To better understand this over-estimation on the right-hand side of \eqref{eq:regret-UB1-heuristic}, recall that in each episode $k$, the value estimate $V_1^k$ is computed via backward induction from step $H$ back to step $1$ (see \eqref{eq:updateqv-ucbvi}). At each step $h$, the Q-estimate is inflated by a Bernstein-style bonus term:
$$
b_h^k(s,a) \asymp \sqrt{ \frac{\mathsf{Var}_{\widehat{P}^k_{s,a,h}} \big( V_{h+1}^k\big) }{N_h^k (s,a)} } \approx \sqrt{ \frac{\mathsf{Var}_{P_{s,a,h}} \big( V_{h+1}^{\pi^{k}}\big) }{N_h^k (s,a)} } \;\; (\text{ignoring log factors}),
$$
where we omit the $O\big(1/N_h^k (s,a)\big)$ lower-order term for simplicity, as the variance-dependent term typically dominates in the asymptotic regime with $K\rightarrow \infty$.  
While this bonus is designed to compensate for model uncertainty, it also
systematically inflates the value estimates. Since the regret is bounded above by
the optimistic over-estimation in
\eqref{eq:regret-UB1-heuristic}, it is natural to quantify the cumulative
amount of optimism introduced by these bonus terms:
\[
\mathcal{E}_{\mathsf{over}} \coloneqq \sum_{k=1}^K \sum_{h=1}^H b_h^k(s_h^k,a_h^k) \asymp \sum_{k=1}^K \sum_{h=1}^H \sqrt{ \frac{\mathsf{Var}_{P_{s_h^k,a_h^k,h}} \big( V_{h+1}^{\pi^{k}}\big) }{N_h^k (s_h^k,a_h^k)} }.
\]

To bound this cumulative over-estimation error, a standard application of the Cauchy-Schwarz inequality separates this
quantity into two components:
\begin{align}
\mathcal{E}_{\mathsf{over}} \lesssim \sqrt{\sum_{k,h}\frac{1}{N_{h}^{k}(s_{h}^{k},a_{h}^{k})}} \cdot \sqrt{\sum_{k,h}\mathsf{Var}_{P_{s_{h}^{k},a_{h}^{k},h}}\big(V_{h+1}^{\pi^{k}}\big)}.
\label{eq:ovr-est-ub2}
\end{align}
As it turns out, these two sums can be readily controlled via the following basic results.
\begin{lemma}
\label{lem:sum-inv-Nhk-ucbvi}
For any $K\geq 2$, Algorithm~\ref{alg:main-ucbvi} satisfies
\begin{align}
\sum_{k=1}^K \sum_{h=1}^H \frac{1}{\max\big\{ N_h^k(s_h^k,a_h^k),1\big\}}\leq 2SAH\log_2 K.
\label{eq:sum-inv-Nhk-ucbvi}
\end{align}
\end{lemma}
\begin{lemma}
\label{lem:cum-var-ub-Hsquare-ucbvi-mean}
Fix a deterministic policy $\pi$. Generate a sample trajectory $(s_1,a_1,\dots,s_H,a_H)$ by executing policy $\pi$ in the MDP. Then one has
\begin{align}
\mathop{\mathbb{E}}\Bigg[\sum_{h=1}^{H}\mathsf{Var}_{P_{s_{h},a_{h},h}}\big(V_{h+1}^{\pi}\big)\Bigg] & \le H^{2},
\label{eq:lem-cum-var-ub-Hsquare-ucbvi-mean}
\end{align}
where the expectation is over the randomness of the sample trajectory. 
\end{lemma}
\begin{remark}
The variance bound in Lemma~\ref{lem:cum-var-ub-Hsquare-ucbvi-mean} parallels the variance-based upper bounds --- such as \eqref{eq:I-gamma-P-Var-bound} in Section~\ref{sec:analysis-mb-pe-2nd-attempt} --- that were previously established for, and play a pivotal role in the analysis of, $\gamma$-discounted infinite-horizon MDPs. 
\end{remark}

Assuming that the cumulative variance term (i.e., the second sum appearing in the upper bound \eqref{eq:ovr-est-ub2}) concentrates around its expectation, the preceding two lemmas imply that, with high probability, 
\begin{align}
\mathcal{E}_{\mathsf{over}} \leq \widetilde{O}\Big( \sqrt{SAH} \cdot \sqrt{KH^2} 
\Big) = \widetilde{O}\Big( \sqrt{SAH^3K} 
\Big). 
\label{eq:ovr-est-ub1}
\end{align}
Furthermore, since the right-hand side of \eqref{eq:regret-UB1-heuristic} is often orderwise dominated by this over-estimation error (as we shall shortly rigorize), the above heuristic discussion suggests that, asymptotically, $$\mathsf{Regret}(K)\lesssim \widetilde{O}\Big( \sqrt{SAH^3K} 
\Big).$$ 

Finally, we note that the cumulative variance bound in Lemma~\ref{lem:cum-var-ub-Hsquare-ucbvi-mean} --- whose statement and proof are both remarkably simple ---  is the key ingredient behind the asymptotically optimal regret guarantee. Indeed, if one were to bound each summand in \eqref{eq:lem-cum-var-ub-Hsquare-ucbvi-mean} naively by $H^2$ without exploiting the dependence structure across the horizon, the resulting cumulative variance  could scale as $H^3$, thereby precluding asymptotically optimal regret guarantees. This observation further highlights the importance of incorporating variance information into the design of exploration bonuses.

\begin{proof}[Proof of Lemma~\ref{lem:sum-inv-Nhk-ucbvi}]
%[Proof of property~\eqref{eq:sum-inv-Nhk-ucbvi}]
%
For any given $(s,a,h)$, it is easily seen that  
\begin{align*}
	\sum_{k=1}^K \frac{1}{\max\{ N_h^k(s_h^k,a_h^k),1\}}  \mathds{1}\Big\{(s,a)=\big(s_h^k,a_h^k \big) \Big\}  
	&\leq 1+ \sum_{k=1}^{K-1} \frac{1}{k} \leq 2\log_2 K .
\end{align*}
	This holds since: (i) when $(s,a)$ is visited for the first time, we have $N_h^k(s,a)=0$ by definition, so the corresponding summand equals 1; (ii) thereafter, each subsequent visit to $(s,a)$ increments its associated visitation count by 1. 
Summing over all $(s,a,h)$ tuples completes the proof. 
\end{proof}
\begin{proof}[Proof of Lemma~\ref{lem:cum-var-ub-Hsquare-ucbvi-mean}]
Consider first any fixed pair $(s,h)\in\mathcal{S}\times[H]$. 
It is easy to verify that
\begin{align}
\mathsf{Var}_{P_{s,\pi_h(s),h}}\big(V_{h+1}^{\pi}\big) & =\sum_{s'\in\mathcal{S}}P_{s,\pi_{h}(s),h}(s')\Big[V_{h+1}^{\pi}(s')-\big\langle P_{s,\pi_{h}(s),h},V_{h+1}^{\pi}\big\rangle\Big]^{2} \notag\\
 & =\sum_{s'\in\mathcal{S}}P_{s,\pi_{h}(s),h}(s')\Big[V_{h+1}^{\pi}(s')-V_{h}^{\pi}(s)+r_h\big(s,\pi_{h}(s)\big)\Big]^{2} \notag\\
	&= \mathop{\mathbb{E}}\limits_{s'\sim P_{s,\pi_{h}(s),h}}\bigg[\Big(V_{h+1}^{\pi}(s')-V_{h}^{\pi}(s)+r_h\big(s,\pi_{h}(s)\big)\Big)^{2} \bigg],
	\label{eq:variance-decompose-12345}
\end{align}
where the second identity results from the Bellman equation. 
The cumulative variance over the horizon then satisfies
\begin{align*}
	\arxiv{ &} \mathop{\mathbb{E}}\limits\Bigg[\sum_{h=1}^{H}\mathsf{Var}_{P_{s_{h},a_{h},h}}\big(V_{h+1}^{\pi}\big)\Bigg] \book{&} \overset{\text{(i)}}{=}\sum_{h=1}^{H}\mathop{\mathbb{E}}\limits\Bigg[\Big(V_{h+1}^{\pi}(s_{h+1})-V_{h}^{\pi}(s_{h})+r_{h}\big(s_{h},a_{h}\big)\Big)^{2}\Bigg] \notag\\
 & \overset{\text{(ii)}}{=}\mathop{\mathbb{E}}\limits\Bigg[\Bigg(\sum_{h=1}^{H}\Big[V_{h+1}^{\pi}(s_{h+1})-V_{h}^{\pi}(s_{h})+r_{h}\big(s_{h},a_{h}\big)\Big]\Bigg)^{2}\Bigg] \notag\\
 & \overset{\text{(iii)}}{=}\mathop{\mathbb{E}}\limits\Bigg[\Bigg(\sum_{h=1}^{H}r_{h}\big(s_{h},a_{h}\big)-V_{1}^{\pi}(s_{1})\Bigg)^{2}\Bigg]  \overset{\text{(iv)}}{\leq}H^{2}.
% \label{eq:cum-var-ub-Hsquare-ucbvi}
\end{align*}
Here, (i) follows by using \eqref{eq:variance-decompose-12345} and recalling that $a_h=\pi_h(s_h)$ in the sample trajectory; (ii) comes from the Markov property of the trajectory, which ensures that the cross terms vanish; (iii) arises from telescoping together with the condition $V_{H+1}^{\pi}=0$; and (iv) holds because both $\sum_{h=1}^{H} r_h(s_h,a_h)$ and $V_1^{\pi}(s_1)$ reside within the interval $[0,H]$.
\end{proof}

\subsection{Analysis: proof of Theorem~\ref{thm-ucbvi-regret}}
\label{sec:proof-of-thm-ucbvi}

In this subsection, we provide the proof of Theorem~\ref{thm-ucbvi-regret}. While the proof below differs somewhat from the original arguments of \citet{azar2017minimax} --- though remaining similar in spirit --- it provides a treatment that is more coherent with the proof of the \mvp algorithm to be presented in the next section. For notational simplicity, we introduce a logarithmic term $\iotab$ as follows: 
%
%\yc{consider changing the notation of $B$ to $\iota$ or $\iota_B$}
%	
\begin{align}
	\iotab & \coloneqq 40000 (\log_2 K)^3 \log(3SAH)\log\frac{1}{\delta'} ,
	\label{eq:assumption-K-proof-B}
\end{align}
where $\delta'$ is defined in  \eqref{eq:definition_delta_prime}.% 
We will also adopt the notation introduced in Section~\ref{sec:intuition-ucbvi}. 

Before embarking on the main proof, we first observe that when $K\leq \iotab SAH^2$, the regret bound in Theorem~\ref{thm-ucbvi-regret} follows trivially. Indeed, since the value function always lies within $[0,H]$, one has
\begin{align*}
\mathsf{Regret}(K)&=\sum_{k=1}^{K}\left(V_{1}^{\star}(s_{1}^{k})-V_{1}^{\pi^{k}}(s_{1}^{k})\right)
\leq HK \leq  SAH^3\iotab \notag\\
	&\leq \widetilde{O}\big( \sqrt{SAH^{3}K}+S^2AH^3 \big),
%\label{eq:regret-when-K-is-small}
\end{align*}
which already satisfies the claimed regret bound in this regime. 
Consequently, throughout the remainder of the proof, it suffices to assume that 
\begin{equation}
	K\geq  SAH^2 \iotab.
	\label{eq:assumption-K-proof-ucbvi}
\end{equation}

\paragraph{Step 1: justifying optimism.} 
We first establish that the running Q-function and value function estimates in Algorithm~\ref{alg:main-ucbvi} serve as valid upper bounds for the true optimal Q-function and value function, respectively. This property provides a formal justification for the optimism principle underlying \ucbvi. Note that the same optimism property continues to hold if the exploration bonuses are replaced by confidence widths derived from alternative concentration inequalities, such as Hoeffding's inequality. 
\begin{lemma}[Optimism]\label{lemma:opt}
Suppose that the constants $c_1,c_2>0$ in \eqref{eq:bh-defn-Bernstein-ucbvi} obey $c_1\geq 4$ and $c_2\geq c_1^2$. 
With probability exceeding $1-2SAHK\delta'$, 
\begin{equation}
	Q_h^k(s,a)\geq Q_h^{\star}(s,a) \qquad  \text{and}  \qquad V^k_h(s)\geq V^{\star}_h(s)
	\label{eq:optimism-Q-V-geq}
\end{equation}
hold simultaneously for all $(s,a,h,k)\in \cS\times \cA\times [H] \times [K]$. 
 \end{lemma}
 Intuitively, this result holds because the exploration bonus added in each iteration is sufficiently large to dominate the uncertainty in the Bellman-type updates~\eqref{eq:updateqv-ucbvi} arising from the use of empirical model estimates. The proof is deferred to Section~\ref{sec:proof-lemma:opt}. Unless stated otherwise, all subsequent analysis is carried out on the high-probability event under which the inequalities in \eqref{eq:optimism-Q-V-geq} hold. 

\paragraph{Step 2: decomposing regret.}
In view of the optimism property established in Lemma~\ref{lemma:opt}, the regret admits the following upper bound
\begin{align}
\mathsf{Regret}(K) & =\sum_{k=1}^{K}\Big(V_{1}^{\star}(s_{1}^{k})-V_{1}^{\pi^{k}}(s_{1}^{k})\Big)\leq\sum_{k=1}^{K}\Big(V_{1}^{k}(s_{1}^{k})-V_{1}^{\pi^{k}}(s_{1}^{k})\Big)
	\label{eq:regret-UB1}
\end{align}
with probability at least $1-2SAHK\delta'$. The remainder of the analysis therefore focuses on upper bounding the right-hand side of~\eqref{eq:regret-UB1}. 
\begin{remark}
To better understand the rationale behind replacing $V_1^{\star}$ with $V_1^{k}$, note that
$
V_1^k(s_1^k)=\max_a Q_1^k(s_1^k,a)
= 
Q_1^k\big(s_1^k,\pi_1^k(s_1^k)\big)
$
due to the greedy policy update (cf.~line~\ref{line:policy-update-ucbvi} of Algorithm~\ref{alg:main-ucbvi}). Consequently, both $V_1^k(s_1^k)$ and $V_1^{\pi^k}(s_1^k)$ are, in some sense, aligned in terms of the selected actions. This alignment proves convenient when recursively expanding the difference
$
V_1^k(s_1^k)-V_1^{\pi^k}(s_1^k)
$
using Bellman-type updates.
\end{remark}

To control the right-hand side of \eqref{eq:regret-UB1}, we begin by deriving the following upper bound on the optimistic value function estimate.
\begin{lemma}\label{lemma:decomdetail}
For every $1\leq k\leq K$, one has
\begin{align}
	V_1^k(s_1^k) &\leq \sum_{h=1}^{H} \Big\{ \Big\langle \widehat{P}^k_{s_h^k,a_h^k,h} - P_{s_h^k,a_h^k,h}, V_{h+1}^k \Big\rangle + b_h^k(s_h^k,a_h^k) + r_h(s_h^k,a_h^k) \arxiv{\notag\\}
  \arxiv{  &\qquad }+ \Big\langle P_{s_h^k,a_h^k,h}-e_{s_{h+1}^k}, V_{h+1}^k \Big\rangle \Big\} .
    \notag
\end{align}
Here and throughout, $e_s$ denotes the $s$-th standard basis vector. 
\end{lemma}
Intuitively, this recurrence relation is derived by recursively unrolling $V_1^k(s_1^k)$ via its update rule through the entire horizon. Crucially, both the empirical model estimation error $\widehat{P}^k_{s_h^k,a_h^k,h} - P_{s_h^k,a_h^k,h}$ and the sampling noise $P_{s_h^k,a_h^k,h} - e_{s_{h+1}^k}$ are zero-mean fluctuations, whose cumulative contributions may potentially be controlled via suitable concentration inequalities. We include the proof here given its central role in the analysis of UCB-type online RL algorithms. 
\begin{proof}[Proof of Lemma~\ref{lemma:decomdetail}]
According to the update rule \eqref{eq:updateqv-ucbvi}, the value function estimates for each $1\leq h\leq H$ and $1\leq k\leq K$ satisfy  
\begin{align}
	V_h^k(s_h^k) & = \max_{a\in \cA}Q_h^k(s_h^k, a) \overset{\mathsf{(i)}}{=} Q_h^k(s_h^k, a_h^k) \notag\\
    & \leq r_h(s_h^k,a_h^k) + \widehat{P}^k_{s_h^k,a_h^k,h}V_{h+1}^k + b_h^k(s_h^k,a_h^k) \nonumber 
\\ 
    &  =  \Big\langle \widehat{P}^k_{s_h^k,a_h^k,h} - P_{s_h^k,a_h^k,h}, V_{h+1}^k \Big\rangle + b_h^k(s_h^k,a_h^k) + r_h(s_h^k,a_h^k) \arxiv{\notag\\}
  \arxiv{  &\qquad } + \Big\langle P_{s_h^k,a_h^k,h}-e_{s_{h+1}^k}, V_{h+1}^k \Big\rangle + V_{h+1}^k(s_{h+1}^k),\nonumber
\end{align}
where (i) is valid due to the greedy rule for selecting $a_h^k$. 
Importantly, this recurrence relation links the value function estimate at step $h$ (i.e., $V_h^k(s_h^k)$) to its counterpart at step $h+1$ (i.e., $V_{h+1}^k(s_{h+1}^k)$). 
By unrolling this relation recursively from $h=1$ to $H$, we arrive at 
\begin{align*}
  V_1^k(s_1^k)  & \leq  \sum_{h=1}^{H} \Big\{ \Big\langle \widehat{P}^k_{s_h^k,a_h^k,h} - P_{s_h^k,a_h^k,h}, V_{h+1}^k \Big\rangle + b_h^k(s_h^k,a_h^k) 
	+ r_h(s_h^k,a_h^k) \arxiv{\notag\\}
   \arxiv{ &\qquad } + \Big\langle P_{s_h^k,a_h^k,h}-e_{s_{h+1}^k}, V_{h+1}^k \Big\rangle \Big\},
\end{align*}
where the recursive expansion terminates at step $H$ since $V_{H+1}^k = 0$. 
	%This concludes the proof.  
\end{proof}

Substituting the regret bound from Lemma~\ref{lemma:decomdetail} into the decomposition~\eqref{eq:regret-UB1} reveals that, with probability at least $1-2SAHK\delta'$, 
\begin{align}
\arxiv{&} 
	%\mathsf{Regret}(K) 
	\eqref{eq:regret-UB1} \book{&}
	\leq\underset{\eqqcolon\,T_{1}}{\underbrace{\sum_{k=1}^{K}\sum_{h=1}^{H}\Big\langle\widehat{P}_{s_{h}^{k},a_{h}^{k},h}^{k}-P_{s_{h}^{k},a_{h}^{k},h},V_{h+1}^{k}\Big\rangle}}+\underset{\eqqcolon\,T_{2}}{\underbrace{\sum_{k=1}^{K}\sum_{h=1}^{H}b_{h}^{k}(s_{h}^{k},a_{h}^{k})}}\nonumber\\
 & \quad+\underset{\eqqcolon\,T_{3}}{\underbrace{\sum_{k=1}^{K}\sum_{h=1}^{H}\Big\langle P_{s_{h}^{k},a_{h}^{k},h}-e_{s_{h+1}^{k}},V_{h+1}^{k}\Big\rangle}}+\underset{\eqqcolon\,T_{4}}{\underbrace{\sum_{k=1}^{K}\left(\sum_{h=1}^{H}r_{h}(s_{h}^{k},a_{h}^{k})-V_{1}^{\pi^{k}}(s_{1}^{k})\right)}}.  
	\label{eq:decomposition}
\end{align}
Therefore, it remains to bound each of the four terms $\{T_i\}_{i=1}^4$. Before proceeding, we briefly comment on each term. 
\begin{itemize}
    \item The term $T_2$, which involves the (Bernstein-style) bonuses, involves variance quantities over time that need to be carefully aggregated in order to attain asymptotically optimal regret.

    \item The term $T_3$ can be controlled relatively directly using martingale concentration inequalities, since $s_{h+1}^k$ is a newly collected sample that is conditionally independent of $V_{h+1}^k$ given what has happened prior to episode $k$. For similar reasons, the term $T_4$ can also be handled via martingale concentration arguments.

    \item The most challenging term is $T_1$, due to the intricate statistical dependence between $\widehat{P}_{s_h^k,a_h^k,h}^k$ and $V_{h+1}^k$. Addressing this dependence requires considerably more delicate analysis. In fact, as we shall see, the difficulty of handling this term in the moderate-to-small $K$ regime also motivates the design of the \mvp algorithm to be introduced in Section~\ref{sec:mvp-intro}.     
	    
\end{itemize}

%We begin by developing bounds for $T_2$, $T_3$ and $T_4$ --- which will also prove useful in the analysis of the \mvp algorithm presented in Chapter~\ref{sec:mvp-intro} --- before turning to the more challenging term $T_1$. 

\noindent 
In what follows, we bound these terms separately. 
As will become clear shortly, the resulting bounds for these terms are inherently intertwined, ultimately leading to a system of interdependent inequalities that are solved jointly in the final step. To facilitate the analysis, we also introduce a few quantities concerning cumulative variance or cumulative second moments across time:
\begin{subequations}
\label{eq:defn-T56-proof}
\begin{align}
	T_5 &\coloneqq \sum_{k=1}^K\sum_{h=1}^H\mathsf{Var}_{\widehat{P}^k_{s_h^k,a_h^k,h}} \big( V_{h+1}^k \big),
	\label{eq:defn-T5-proof} \\
	T_6^{\ucb} &\coloneqq \sum_{k=1}^K \sum_{h=1}^H\mathsf{Var} _{P_{s_h^k,a_h^k,h}}\big( V_{h+1}^k \big) ,
	\label{eq:defn-T6-proof} \\
    	T_{6}^{\mathsf{opt}} &\coloneqq \sum_{k=1}^K \sum_{h=1}^H \mathsf{Var} _{P_{s_h^k,a_h^k,h}}\big( V_{h+1}^{\star} \big) ,
        \label{eq:defn-T6-varstar-proof}
        \\
        T_{6}^{\mathsf{low}} &\coloneqq \sum_{k=1}^K \sum_{h=1}^H\mathsf{Var} _{P_{s_h^k,a_h^k,h}}\big( V_{h+1}^{\pi^k} \big) ,
        \label{eq:defn-T6-varpi-proof}
        \\
        	T_{6}^{\mathsf{diff}} &\coloneqq \sum_{k=1}^K \sum_{h=1}^H \Big\langle P_{s_h^k,a_h^k,h}, \big( V_{h+1}^k-V_{h+1}^{\pi^k} \big)^2\Big\rangle.
            \label{eq:defn-T6-vardif-proof}
\end{align}
\end{subequations}
Here, $T_5$ aggregates {\em empirical} variance terms, while the remaining quantities correspond to cumulative {\em true} variances or second moments associated with objects related to value functions.

\paragraph{Step 3: bounding the term $T_1$ in \eqref{eq:decomposition}.} 

We begin with the term $T_1$, which is particularly challenging to cope with due to the complicated statistical dependency between $\widehat{P}^{k}_{s_h^k,a_h^k,h}$ and $V_{h+1}^k$. 
To see this, note that $\widehat{P}^{k}_{s,a,h}$ is constructed from {\em all} previous samples associated with $(s,a,h)$, while $V_{h+1}^k$ is itself computed based on past samples. Consequently, these two objects exhibit complicated statistical dependence.

One common strategy for circumventing this technical difficulty is to replace $V_{h+1}^k$ with the optimal value function $V_{h+1}^{\star}$, the latter of which is fixed and therefore statistically independent of the sampling process. This strategy has been adopted in a large body of existing analyses of model-based online RL algorithms (e.g., \citet{azar2017minimax,dann2017unifying,zanette2019tighter,zhang2020reinforcement}). To be precise, let us decompose  
\begin{align}
&
    %\sum_{k,h} \Big\langle \widehat{P}^{k}_{s_h^k,a_h^k,h}-P_{s_h^k,a_h^k,h},V_{h+1}^k \Big\rangle
   T_1 =
    \underset{\eqqcolon\, W_1}{\underbrace{ \sum_{k,h} \Big\langle \widehat{P}^{k}_{s_h^k,a_h^k,h}-P_{s_h^k,a_h^k,h},V_{h+1}^{\star} \Big\rangle }}
	\arxiv{ \nonumber\\
	&\qquad\qquad}
	+ \underset{\eqqcolon\, W_2}{\underbrace{\sum_{k,h} \Big\langle\widehat{P}^{k}_{s_h^k,a_h^k,h}-P_{s_h^k,a_h^k,h}, V_{h+1}^k-V_{h+1}^{\star}\Big\rangle }}.
	\label{eq:decompose-past-work}
\end{align}
The rationale behind the decomposition~\eqref{eq:decompose-past-work} is two-fold: the first term $ W_1$ can be conveniently controlled via concentration inequalities given that $V_{h+1}^{\star}$ is fixed, whereas the second term $W_2$ acts as a higher-order residual that would vanish asymptotically as the value estimate $V_{h+1}^{k}$ converges to $V_{h+1}^{\star}$ for large enough $k$.

The lemma below develops upper bounds that connect  $W_1$ and $W_2$ with cumulative variances. The proof is deferred to Section~\ref{sec:proof-lemma:bound-W1-W2-ucbvi-proof}. 
\begin{lemma}
\label{lem:bound-W1-W2-ucbvi-proof}
Recall the definition of $W_1$ and $W_2$ in \eqref{eq:decompose-past-work}, and set 
%
%\begin{subequations}
\begin{align}
W_3 & \coloneqq\sum_{k=1}^{K}\sum_{h=1}^{H}\Big\langle\widehat{P}_{s_{h}^{k},a_{h}^{k},h}^{k}-P_{s_{h}^{k},a_{h}^{k},h},\big(V_{h+1}^{k}\big)^{2}\Big\rangle.
\label{eq:defn-T7-proof}
\end{align}
With probability greater than $1-12S^2AHK\delta'$, one has
\begin{align*}
	W_{1} & \le \frac{1}{8} \sqrt{SAHT_{6}^{\mathsf{low}} \iotab } + \frac{1}{8} \sqrt{ SAHT_{6}^{\mathsf{diff}} \iotab }
	+ \frac{1}{8} SAH^{2} \iotab ,
	%\leq W_3 
%	& \le \sqrt{4SAH(T_6^{\mathsf{low}}+T_{6}^{\mathsf{diff}})(\log_{2}K)\log\frac{1}{\delta'}} +SAH^{2}(\log_{2}K)\log\frac{1}{\delta'},
%\label{eq:W1-ub-ucbvi-T1-lem} 
\notag\\
W_2 	& \le \frac{1}{8} \sqrt{ S^{2}AHT_{6}^{\mathsf{diff}}\iotab}+ \frac{1}{8}  S^{2}AH^{2} \iotab, \notag\\
 %\sqrt{2S^{2}AHT_{6}^{\mathsf{diff}}(\log_{2}K)\log\frac{1}{\delta'}}+S^{2}AH^{2}(\log_{2}K)\log\frac{1}{\delta'},
W_3 & \le \frac{1}{8} \sqrt{ SAH^{3}T_{6}^{\mathsf{low}}\iotab} + \frac{1}{8} \sqrt{ S^{2}AH^{3}T_{6}^{\mathsf{diff}}\iotab} + \frac{1}{8} S^{2}AH^{3} \iotab , \notag
%&\le 4\sqrt{SAH^3T_6^{\mathsf{low}}(\log_{2}K)\log\frac{1}{\delta'}}  
% +8\sqrt{S^{2}AH^3T_6^{\mathsf{diff}}(\log_{2}K)\log\frac{1}{\delta'}}
% \notag\\
% &\qquad +2S^{2}AH^{3}(\log_{2}K)\log\frac{1} {\delta'},
% \notag
\end{align*}
where $T_6^{\mathsf{low}}$ and $T_{6}^{\mathsf{diff}}$ are defined in \eqref{eq:defn-T56-proof}.
%\end{subequations}
%
\end{lemma}
\begin{remark}
While the primary goal of this step is to bound $W_1$ and $W_2$, the same proof techniques can be readily adapted to upper bound $W_3$ as well, which will appear in the subsequent analysis.
\end{remark}
\begin{remark}
A careful reader may notice that, in addition to the difference in the variance terms, the bound on $W_2$ contains an extra factor in $S$ compared to the bound on $W_1$. This discrepancy arises because, in the analysis of $W_1$, the value function $V_{h+1}^{\star}$ is fixed and statistically independent of the data, which substantially simplifies the argument. In contrast, the term $W_2$ involves the difference $V_{h+1}^k - V_{h+1}^{\star}$, which exhibits intricate statistical dependence on the data and consequently incurs an additional factor of $S$. 
\end{remark}
Combining Lemma~\ref{lem:bound-W1-W2-ucbvi-proof} with \eqref{eq:decompose-past-work}, we reach
\begin{align}
   T_1 &= W_1+W_2 \arxiv{ \notag\\}
	\arxiv{&} \leq \frac{1}{4} \sqrt{ SAHT_{6}^{\mathsf{low}}\iotab}+ \frac{1}{4} \sqrt{ S^{2}AHT_{6}^{\mathsf{diff}}\iotab}+ \frac{1}{4}S^{2}AH^{2}  \iotab 
	%\notag\\
	%& \sqrt{4SAHT_6^{\mathsf{low}}(\log_{2}K)\log\frac{1}{\delta'}} 
  % + 4\sqrt{S^2AHT_{6}^{\mathsf{diff}}(\log_{2}K)\log\frac{1}{\delta'}} \notag\\
   %&\qquad + 2S^2AH^{2}(\log_{2}K)\log\frac{1}{\delta'}
   \eqqcolon T_1^{\mathsf{ub}}
	\label{eq:decompose-past-work-T1-bound}
\end{align}
with probability greater than $1-12S^2AHK\delta'$. 
It is also worth noting that the variance terms $T_6^{\mathsf{low}}$ and $T_6^{\mathsf{diff}}$ remain random quantities and, therefore, require additional arguments to control, which we shall accomplish in the ensuing steps.

\paragraph{Step 4: bounding the terms $T_2,T_3$ and $T_4$ in \eqref{eq:decomposition}.}
This step is dedicated to bounding the terms  $T_2,T_3$ and $T_4$ arising in the regret decomposition \eqref{eq:decomposition}, as formalized in the lemma below.
\begin{lemma}\label{lem:bound-T234}
	With probability exceeding $1-4\big(\log_2 (KH)\big)\delta'$,  one has 
\begin{subequations}
\label{eq:boundt234}
\begin{align}
	T_2 &\le \frac{1}{2}\sqrt{ SAHT_{5}\iotab}+\frac{1}{2} SAH^{2}\iotab,
	\label{eq:boundt2} \\
	|T_3|  &\leq \sqrt{ 8 T_6^{\ucb} \log \frac{1}{\delta'}  } + 5H\log \frac{1}{\delta'} \notag\\
    &\leq 4\sqrt{  T_6^{\mathsf{low}} \log \frac{1}{\delta'}  } + 4\sqrt{  T_6^{\mathsf{diff}} \log \frac{1}{\delta'}  } + 5H\log \frac{1}{\delta'}, \label{eq:boundt3} \\
	|T_4| &\leq \sqrt{2KH^2\log \frac{1}{\delta'}},\label{eq:bdt_4f}
\end{align}
\end{subequations}
provided that $\iotab \geq 8\max\{c_1^2,c_2\}(\log_{2}K)\log\frac{1}{\delta'}$.  
Here, $T_5$, $T_6^{\ucb}$, $T_6^{\mathsf{low}}$ and $T_6^{\mathsf{diff}}$ are defined in \eqref{eq:defn-T56-proof}. 
\end{lemma}
At a high level, $T_3$ and $T_4$ are sums of zero-mean fluctuations and are bounded via martingale concentration inequalities, whereas the bound on $T_2$ follows from the Cauchy-Schwarz inequality alongside basic properties of the visitation counts. 
The proof is deferred to Section~\ref{app:pflem:bound-T234}.

\paragraph{Step 5: combining the bounds on $\{T_i\}_{i=1}^4$.}
Taking the preceding bounds~\eqref{eq:decompose-past-work-T1-bound} and \eqref{eq:boundt234} together with \eqref{eq:decomposition}, we obtain
\begin{subequations}
\label{eq:suboptimality-V1-ub1-ucbvi}
\begin{align}
	\arxiv{&}\sum_{k=1}^{K}\Big(V_{1}^{k}(s_{1}^{k})-V_{1}^{\pi^{k}}(s_{1}^{k})\Big)
	\book{&}\le T_1+T_2+T_3+T_4 \notag\\
	\arxiv{&\qquad}\book{&}\le T_1^{\mathsf{ub}} + T_2  
	 +4\sqrt{T_{6}^{\mathsf{low}}\log\frac{1}{\delta'}}+4\sqrt{T_{6}^{\mathsf{diff}}\log\frac{1}{\delta'}} 
	 + 7\sqrt{KH^{2}}\log\frac{1}{\delta'} \notag\\
	\arxiv{&\qquad} \book{&}\le T_0^{\mathsf{ub}}
\end{align}
with probability at least $1-16S^2AHK\delta'$, where we define
\begin{align}
	T_0^{\mathsf{ub}} &\coloneqq 
	\frac{3}{8}\sqrt{ SAHT_{6}^{\mathsf{low}}\iotab}+\frac{3}{8}\sqrt{ S^{2}AHT_{6}^{\mathsf{diff}}\iotab}+\frac{1}{4} S^{2}AH^{2}\iotab \arxiv{\notag\\ &\qquad\qquad} +7\sqrt{KH^{2}}\log\frac{1}{\delta'}+T_{2}.
	\label{eq:defn-T0-ub-ucbvi-135}
\end{align}
\end{subequations}

Before proceeding, we note that the same arguments used to bound $\sum_{k=1}^{K}\big(V_{1}^{k}(s_{1}^{k})-V_{1}^{\pi^{k}}(s_{1}^{k})\big)$ extend directly to $\sum_{k=1}^{K}\big(V_{h}^{k}(s_{h}^{k})-V_{h}^{\pi^{k}}(s_{h}^{k})\big)$ for any $1<h\le H$. Combining this observation with a union bound yields that, with probability at least $1-16S^2AH^2K\delta'$, 
\begin{align}
	&\sum_{k=1}^{K}\Big(V_{h}^{k}(s_{h}^{k})-V_{h}^{\pi^{k}}(s_{h}^{k})\Big)
	\le T_0^{\mathsf{ub}}
	\label{eq:suboptimality-h-ub1-ucbvi}
\end{align}
holds simultaneously for all $1\leq h\leq H$, with $T_0^{\mathsf{ub}}$ defined in \eqref{eq:defn-T0-ub-ucbvi-135}.

\paragraph{Step 6: bounding the second moments in \eqref{eq:defn-T56-proof}.}
Thus far, the bounds on $T_1$, $T_2$ and $T_3$ (see Lemmas~\ref{lem:bound-W1-W2-ucbvi-proof} and \ref{lem:bound-T234}) --- and hence the bound on $T_0^{\mathsf{ub}}$ --- depend explicitly on the cumulative variances $T_5$, $T_6^{\mathsf{low}}$ and $T_6^{\mathsf{diff}}$ (cf.~\eqref{eq:defn-T56-proof}). To obtain the desirable regret bound, it remains to derive asymptotically tight upper bounds for these aggregate second moments. This is accomplished via the following lemma, whose proof is provided in Section~\ref{sec:pflem:bound-T56}.
\begin{lemma}
\label{lem:bound-T56}
With probability at least $1-14S^2AH^2K\delta'$,
\begin{subequations}
\label{eq:boundt56}
\begin{align}
T_{5} & \leq W_3+ 2HT_{2}+2KH^{2}
+\sqrt{32H^{2}T_6^{\ucb}\log\frac{1}{\delta'}}+5H^{2}\log\frac{1}{\delta'}, %\notag\\
%& \leq W_4+ 2HT_{2}+2KH^{2} +8\sqrt{H^{2}\big(T_6^{\mathsf{low}}+T_6^{\mathsf{diff}}\big)\log\frac{1}{\delta'}}+5H^{2}\log\frac{1}{\delta'},
\label{eq:boundt5} \\
    T_6^{\low} &\le 
     KH^2 + \sqrt{2H^6K \log\frac{1}{\delta'}}, 
    \label{eq:boundtpi} \\
    T_6^{\mathsf{diff}} &\le 2H^2T_0^{\mathsf{ub}} + 18 H^2\log\frac{1}{\delta'},
    \label{eq:boundtdiff-12345}
%\\
%|T_{8}|&
%	\leq \sqrt{32H^{2}T_{6}\log\frac{1}{\delta'}}+5H^{2}\log\frac{1}{\delta'}	 . 
%	\label{eq:boundt8}
\end{align}
\end{subequations}
where $W_3$ and $T_0^{\mathsf{ub}}$ are defined in \eqref{eq:defn-T7-proof} and \eqref{eq:suboptimality-V1-ub1-ucbvi}, respectively. 
In addition, under the condition~\eqref{eq:assumption-K-proof-ucbvi}, one further has 
\begin{subequations}\label{eq:T56-low-simpler-ub-ucbvi}\begin{align}
	T_{6}^{\mathsf{low}} &\leq 2KH^{2}, 
	\label{eq:T6-low-simpler-ub-ucbvi} \\
	T_{5} & \leq6KH^{2}+\frac{1}{2}  \sqrt{S^2 AH^{3}T_{6}^{\mathsf{diff}} \iotab }+ 4 S^{2}AH^{3}\iotab	
	\label{eq:T5-low-simpler-ub-ucbvi}
\end{align}
with probability at least $1-30S^2AH^2K\delta'$. 
\end{subequations}
\end{lemma}

\begin{remark}
	Asymptotically (as $K\rightarrow \infty$), it is natural to expect these variance quantities to satisfy 
	$T_5\approx T_6^{\ucb} \approx T_6^{\mathsf{opt}} \approx T_6^{\low}  \lesssim KH^2$, where the last inequality follows from \eqref{eq:T6-low-simpler-ub-ucbvi} (thanks to the Markov structure of the cumulative variance). Importantly, both establishing and exploiting the bound $T_6^{\mathsf{low}}\lesssim KH^2$ are crucial for obtaining the asymptotically optimal regret guarantees. 
    %Indeed, if one were instead to invoke the trivial bound $T_6^{\mathsf{low}}\lesssim KH^3$, then the resulting regret bound would incur an extra factor of $\sqrt{H}$ relative to the minimax lower limit. 
%
\end{remark}

%where the last inequality applies the AM-GM inequality $\frac{1}{4}\sqrt{2BSAH^{5}K}\leq KH^{2} + \frac{1}{2}BSAH^3$.  

%
%In particular, $T_1$ has already been upper bounded in Chapter~\ref{sec:tec1}, and hence we shall describe how to bound $T_2,\ldots,T_4$ in the sequel.    

\paragraph{Step 7: putting all pieces together.}
We now combine the preceding (interdependent) bounds to derive a clean upper bound on the regret.  

First, substituting \eqref{eq:T5-low-simpler-ub-ucbvi} into \eqref{eq:boundt2} in Lemma~\ref{lem:bound-T234}
yields
\begin{align*}
	\arxiv{&} T_{2} \book{&}\leq\frac{1}{2}\sqrt{SAHT_{5}\iotab }+\frac{1}{2} SAH^{2} \iotab\\
 & \leq\frac{1}{2}\sqrt{6 SAH^{3}K \iotab}+\frac{1}{2}\sqrt{ S^{2}AH^{2}\bigg(\frac{1}{2}\sqrt{ AHT_{6}^{\mathsf{diff}}\iotab}+4SAH^{2}\iotab \bigg)\iotab}+\frac{1}{2} SAH^{2} \iotab\\
 & \leq\frac{1}{2}\sqrt{6 SAH^{3}K \iotab}+ \frac{1}{4}S^{2}AH^{2} \iotab +\frac{1}{8}\sqrt{ AHT_{6}^{\mathsf{diff}}\iotab }+ SAH^{2} \iotab +\frac{1}{2} SAH^{2} \iotab\\
 & \leq\frac{1}{2}\sqrt{6 SAH^{3}K\iotab}+\frac{1}{8}\sqrt{ AHT_{6}^{\mathsf{diff}}\iotab}+ 2 S^{2}AH^{2} \iotab ,
\end{align*}
where the second inequality follows from the elementary fact $\sqrt{a+b}\leq \sqrt{a}+\sqrt{b}$, and the penultimate inequality is due to the AM-GM inequality.  
Substituting this into the definition of $T_{0}^{\mathsf{ub}}$ (cf.~\eqref{eq:suboptimality-V1-ub1-ucbvi}) and combining terms, we reach
\begin{align}
	T_{0}^{\mathsf{ub}}  &\leq \frac{3}{8}\sqrt{SAHT_{6}^{\mathsf{low}}\iotab }+ \sqrt{ S^{2}AHT_{6}^{\mathsf{diff}}\iotab}+ 9  \sqrt{SAH^{3}K  } \iotab+ 2  S^{2}AH^{2} \iotab\notag\\
	& \leq  10 \sqrt{SAH^{3}K } \iotab + \sqrt{ S^{2}AHT_{6}^{\mathsf{diff}}\iotab}+ 2 S^{2}AH^{2} \iotab,
	\label{eq:T0ub-upper-bound-simpler-ucbvi}
\end{align}
where the last inequality results from~\eqref{eq:T6-low-simpler-ub-ucbvi}.

To conclude the proof, observe that the AM-GM inequality implies
\[
\sqrt{S^{2}AHT_{6}^{\mathsf{diff}}\iotab }\leq\frac{1}{4H^{2}}T_{6}^{\mathsf{diff}}+ S^{2}AH^{3}\iotab \leq\frac{1}{2}T_{0}^{\mathsf{ub}}+6 S^{2}AH^{3} \iotab,
\]
where the last inequality follows from \eqref{eq:boundtdiff-12345} in Lemma~\ref{lem:bound-T56}. Combining this bound with \eqref{eq:T0ub-upper-bound-simpler-ucbvi} and rearranging terms, we can readily conclude that
\begin{align*}
T_{0}^{\mathsf{ub}} & \lesssim  \sqrt{SAH^{3}K} \iotab+ S^{2}AH^{3} \iotab
\end{align*}
with probability at least $1-50S^2AH^2K\delta'$. Together with the choice of $\delta'$ in~\eqref{eq:assumption-K-proof-B}, this completes the proof of Theorem~\ref{thm-ucbvi-regret}.

\section{An improved model-based online RL algorithm: \mvp}
\label{sec:mvp-intro}

As noted in Section~\ref{sec:theory-ucbvi}, the \ucbvi algorithm achieves asymptotically optimal regret only after an excessively large burn-in phase characterized by \eqref{eq:burn-in-ucbvi}. How to overcome this limitation and obtain regret guarantees that are optimal for all sample sizes has therefore become a fundamentally important research direction, drawing considerable attention in the years following the introduction of \ucbvi. This objective was recently accomplished by \citet{zhang2024settling,zhang2025settling}, which established that a variant of the optimistic model-based algorithms --- called {\em Monotonic Value Propagation} (\mvp), originally proposed by \citet{zhang2020reinforcement} --- attains minimax-optimal regret (up to logarithmic factors) for any number $K$ of sample trajectories.

\subsection{Revisiting \ucbvi in the sample-starved regime}
\label{sec:revisit-ucbvi-sample-limited}

At a high level, the \mvp algorithm can be viewed as a refinement of \ucbvi in Algorithm~\ref{alg:main-ucbvi}, with only a handful of algorithmic modifications. As we shall see, these seemingly modest algorithmic changes lead to substantially stronger regret guarantees in the sample-starved regime. To motivate these algorithmic ideas, we first revisit the main bottleneck in the analysis of \ucbvi that limits its theoretical performance in the most sample-constrained settings.

As discussed in Step 2 of Section~\ref{sec:proof-of-thm-ucbvi}, the regret of \ucbvi admits a decomposition of the form \eqref{eq:decomposition}, consisting of four terms $\{T_i\}_{1\leq i\leq 4}$. As already noted there, the terms $T_3$ and $T_4$ can be readily handled using standard martingale concentration inequalities, while $T_2$ can likewise be controlled through the variance-based inequalities developed there. The main obstacle is the term 
\begin{equation}
T_1 = \sum_{k=1}^{K}\sum_{h=1}^{H}\Big\langle\widehat{P}_{s_{h}^{k},a_{h}^{k},h}^{k}-P_{s_{h}^{k},a_{h}^{k},h},V_{h+1}^{k}\Big\rangle
,
\label{eq:T1-revisit-sec-mvp}
\end{equation}
whose analysis is complicated due to the intricate statistical dependence between the empirical model $\widehat{P}^k$ and the value estimate $V_{h+1}^k$; indeed, both $\widehat{P}^k$ and $V^{k}$ depend on the entire collection of samples gathered prior to episode $k$, precluding a direct application of standard concentration arguments. Below, we discuss two fundamentally different strategies for overcoming this obstacle. 

\paragraph{Strategy 1: replacing $V^k$ with $V^{\star}$.}
Thus far, the analysis of \ucbvi copes with $T_1$ by first replacing $V_{h+1}^k$ in \eqref{eq:T1-revisit-sec-mvp} with the optimal value function $V_{h+1}^\star$ --- thereby enabling the use of standard concentration techniques given that $V_{h+1}^\star$ is fixed and statistically independent from samples --- and then controlling the resulting approximation error induced by the discrepancy between $V_{h+1}^k$ and $V_{h+1}^\star$. Although this approach is asymptotically well justified --- indeed, one anticipates $V_{h+1}^k$ to converge to $V_{h+1}^\star$ as the number of episodes $K \to \infty$ --- the gap between $V_{h+1}^k$ and $V_{h+1}^\star$ can remain substantial and non-negligible in the regime where $K$ is only moderate or small. As a result, the induced approximation error can be difficult to control sharply in this regime, ultimately giving rise to the large burn-in cost appearing in Theorem~\ref{thm-ucbvi-regret}.

\paragraph{Strategy 2: a covering-type argument.}
We now describe an alternative approach for analyzing this term $T_1$, based on a covering-type argument. A brief sketch of this argument is provided below, which serves to motivate both the algorithmic design and analysis of \mvp.

To keep the exposition simple and highlight the main ideas, we keep the argument below informal.  Recall that $N_{h}^k(s,a)$ denotes the visitation count of $(s,a,h)$ prior to episode $k$. Intuitively, once the collection of visitation counts at step $h$ (i.e., $\big\{N_h^k(s,a)\big\}_{(s,a,k)\in\mathcal S\times\mathcal A\times[K]}$) are given, the empirical transition kernel $\widehat P_{s,a,h}^k$ and the value estimate  $V_{h+1}^k$ become conditionally
independent; to see this, it suffices to note that  $\widehat P_{s,a,h}^k$
depends only on the samples collected at step $h$, whereas 
$V_{h+1}^k$ is computed solely from samples collected at later steps via the
backward Bellman recursion \eqref{eq:updateqv-ucbvi}. 
Now suppose, for the sake of intuition, that these visitation counts were fixed in advance and {\em statistically independent} of the empirical transition kernels $\{\widehat{P}_{s,a,h}^k\}$ (a heuristic assumption that is not technically accurate, but useful for an informal discussion). Under this heuristic viewpoint, standard concentration techniques could be applied directly to control
$$\sum_{k,h} \big(\widehat{P}^{k}_{s_h^k,a_h^k,h}-P_{s_h^k,a_h^k,h}\big)V_{h+1}^k.$$

The remaining step would be to remove the artificial independence assumption by taking a union bound over all possible realizations of ${N_h^k(s,a)}$.  
The main obstacle, however, is combinatorial: the number of such visitation-count configurations grows exponentially with $K$, making a naive union bound prohibitively costly and hence significantly weakening the resulting regret guarantee.

Nevertheless, this heuristic argument points to a potentially useful idea: if the effective number of admissible visitation-count configurations could be reduced --- say, to a quantity polynomial in $K$ --- then the covering-based argument outlined above could become viable. This important observation motivates the design of \mvp, which we introduce next.

\subsection{Algorithmic ideas: doubling-based updates}

We now introduce the \mvp algorithm by singling out its key difference from the \ucbvi algorithm presented in Algorithm~\ref{alg:main-ucbvi}. A key distinguishing feature of \mvp is its {\em epoch-based update} scheme for the empirical transition kernels, inspired by the {\em doubling-trick} framework adopted in  \citet{jaksch2010near}. More specifically, the learning process is partitioned into consecutive epochs according to a simple doubling rule: whenever there exists a tuple $(s,a,h)$ whose visitation count reaches a power of two, the current epoch terminates. At that point, \mvp reconstructs the empirical model, recomputes the UCB-based optimistic Q-functions and value functions based on the updated transition kernels, and then initiates a new epoch with the corresponding updated policy. Within each epoch, the sampling policy is determined by the optimistic Q-function estimates computed prior to the beginning of that epoch; consequently, {\em the same policy} is used to collect data throughout all episodes within the epoch.

Clearly, this epoch-based paradigm differs from the original \ucbvi algorithm in its update frequency: whereas \ucbvi updates the empirical transition model and recomputes the Q-estimates and value estimates after every episode, the epoch-based paradigm in \mvp performs these updates only at the end of each epoch. An important property ensured by the careful design of the doubling update rule is that: for each each $(s,a,h)$, the associated empirical transition probability is updated only when its visitation count equals a power of two, and therefore no more than $O(\log_2 K)$ times over the entire learning process. As we shall see, this structural property plays a central role in substantially reducing the effective covering number required in the covering-based argument mentioned in Section~\ref{sec:revisit-ucbvi-sample-limited}. More specifically, it induces a modified version of the visitation-count sequence $\big\{ N^{k}_{h}(s,a)\big\}_{(s,a,k)\in \mathcal{S}\times \mathcal{A}\times [K]}$, which tracks the empirical model currently in use, rather than the most recently collected samples. Owing to the epoch-based update scheme for the empirical models, the number of admissible configurations of this modified visitation-count sequence is dramatically smaller than that arising in the original \ucbvi algorithm.

\begin{algorithm}[t]
	\DontPrintSemicolon
\caption{The monotonic value propagation (\mvp) algorithm for episodic online RL\label{alg:main-mvp}}
%\begin{algorithmic}[ht]

	{\small 

	\textbf{input:}  number of episodes $K$, reward function $r$, confidence parameter $\delta$, constants $c_1, c_2>0$. \\
	%$c_1=\frac{460}{9}$, $c_2 = 2\sqrt{2}$, $c_3=\frac{544}{9}$.  
	
	\textbf{initialize} 
	%$\delta' \leftarrow \frac{\delta}{200SAH^2K^2}$; and 
	%set 
	$N^{\mathsf{all}}_h(s,a)\leftarrow 0$, $N_h(s,a,s')\leftarrow 0$, $N_h(s,a)\leftarrow 0$, $Q^{\ucb}_h(s,a)\leftarrow H$, $V^{\ucb}_h(s)\leftarrow H$, $V^{\ucb}_{H+1}(s)\leftarrow 0$, $\widehat{P}_{s,a,h}(s')\leftarrow 1/S$ for all $(s,a,s',h)\in \mathcal{S}\times \mathcal{A}\times\mathcal{S}\times [H]$. \\
%\textbf{initialization: } 
	\For{$k=1,\cdots,K$} {
		{\color{blue}\tcc{policy update.}}
		update policy $\pi^{\ucb}$ such that $\pi^{\ucb}_h(s) \leftarrow \arg\max_{a}Q^{\ucb}_h(s,a)$ for all $(s,h)\in \mathcal{S}\times [H]$. \\ 
        observe the initial state $s_1 \overset{\mathrm{ind.}}{\sim} \rho$ of a new episode. \\
		\For {$h=1,\cdots,H$} {
{\color{blue}\tcc{sampling.}} 
			take $ a_h= \pi^{\ucb}_h (s_h)$, 
			observe $s_{h+1}$; set $(s,a,s')\leftarrow (s_h,a_h,s_{h+1})$.  \label{line:choose_action-mvp} 
			\\
			 $N^{\mathsf{all}}_h(s,a) \leftarrow  N^{\mathsf{all}}_h( s,a )+1$; $N_h(s,a,s') \leftarrow   N_h(s,a,s')+1$. \\
		{\color{blue}\tcc{updates using data of this epoch.\label{line:a1-mvp}}}
		\If{$N^{\mathsf{all}}_h(s,a)\in \{1,2,\cdots, 2^{\log_2K}\}$ \label{line:rp_update_start-mvp} }   {\label{line:trigger-set-mvp}
			 $N_h(s,a)\leftarrow \sum_{\widetilde{s}}N_h(s,a,\widetilde{s})$.  
			{\color{blue}\tcp{$\#$ visits in this epoch.} \label{line:Nh-update-mvp}}
			$\widehat{P}_{s,a,h}(\widetilde{s}) \leftarrow  \frac{N_h(s,a,\widetilde{s})}{N_h(s,a)}$ for all $\widetilde{s} \in \mathcal{S}$.  \label{line:P-hsa-update-mvp}
			{\color{blue}\tcp{empirical model.}}
			set \textsf{triggered} = \textsf{true}; 
			 $N_h(s,a,\widetilde{s})\leftarrow 0$  for all $\widetilde{s}\in \mathcal{S}$. 
%		\ENDIF \label{line:rp_update_end}
		}
		}
		{\color{blue}\tcc{optimistic Q-estimation for this epoch.}}
		\If {\textsf{triggered} = \textsf{true}} {
			set \textsf{triggered} = \textsf{false}.
			%, and $V_{H+1}(s)\leftarrow 0$ for all $s\in \mathcal{S}$. 
			\\
			\For{$h=H,\cdots,1$} {
				\For{$(s,a)\in \mathcal{S}\times \mathcal{A}$} {
					%\vspace{-3ex}
					set the bonus $b_h(s,a)$ according to \eqref{eq:bh-defn-Bernstein-ucbvi}; compute
					\begin{subequations}
						\label{eq:updateqv-mvp}
					\begin{align} 
					%	\vspace{-3ex}
					%	b_h(s,a) &\leftarrow c_1 \sqrt{\frac{   \mathbb{ V}(\widehat{P}_{s,a,h} ,V_{h+1}) \log \frac{1}{\delta'}  }{ \max\{N_h(s,a),1 \} }}
					%	+c_2\frac{H\log \frac{1}{\delta'}}{ \max\{N_h(s,a) ,1\}  },  \label{eq:update1-mvp}  \\
						Q^{\ucb}_h(s,a) &\leftarrow \min\big\{    r_h(s,a)+\langle \widehat{P}_{s,a,h}, V^{\ucb}_{h+1} \rangle +b_h(s,a)    ,H\big\}, \label{eq:updateq-mvp}\\
						V^{\ucb}_{h}(s) &\leftarrow \max_{a}Q^{\ucb}_h(s,a).
						\label{eq:updatev-mvp}
					\end{align}
					\end{subequations}
					%\vspace{-3ex}
				}
			}
		}
	}

	}

%\end{algorithmic}
\end{algorithm}

The complete procedure of \mvp is presented in Algorithm~\ref{alg:main-mvp}, which adopts the same variance-aware Bernstein-style exploration bonuses as those recommended in Algorithm~\ref{alg:main-ucbvi}.  In terms of notation,  we now let $N_h^{\mathsf{all}}(s,a)$ indicate the total visitation count of $(s,a,h)$ accumulated over all episodes up to the current one, while $N_h(s,a)$ (resp.~$N_h(s,a,\widetilde{s})$) tracks the visitation count of $(s,a,h)$ (resp.~$(s,a,h,\widetilde{s})$) within the current epoch. In addition, the flag \textsf{triggered} is used to indicate whether we have reached the end of the current epoch. 

\begin{remark}
The name {\em Monotonic Value Propagation}, coined by \citet{zhang2020reinforcement}, stems from the fact that the exploration bonuses \eqref{eq:bh-defn-Bernstein-ucbvi} guarantee a certain monotonicity property in value propagation. For instance, the function $f(p,v,n)$ defined in \eqref{eq:monotone-function-f-pvn-ucbvi} in Section~\ref{sec:proof-lemma:opt} --- which resembles a key operation arising in certain optimistic Bellman recursions --- is non-decreasing with respect to $v$, where $v$ typically represents a vector of value estimates. Notably, this monotonicity property has already been built into the version of \ucbvi presented in Algorithm~\ref{alg:main-ucbvi}; this version differs in this respect from the original \ucbvi algorithm of  \citep{azar2017minimax} (see also Remark~\ref{remark:different-bonus-than-Azar}). 
\end{remark}

\subsection{Theoretical guarantees}

The epoch-based design of \mvp, in conjunction with the optimistic model-based framework, turns out to be sufficient to overcome the suboptimal regret guarantees of \ucbvi in the sample-limited regime. As stated in the following theorem, \mvp achieves minimax-optimal regret uniformly over the entire range $K \geq 1$.
\begin{theorem}[\cite{zhang2024settling,zhang2025settling}]\label{thm-mvp-regret}  
Consider any $K \ge 1$ and any $0<\delta<1$. With probability at least $1-\delta$,  
the \mvp algorithm (see Algorithm~\ref{alg:main-mvp}) with Bernstein-style bonuses \eqref{eq:bh-defn-Bernstein-ucbvi} achieves
\begin{equation}
	\mathsf{Regret}(K) 
	\leq \widetilde{O}\Big( \min\big\{\sqrt{SAH^{3}K} ,HK\big\} \Big),
	%\leq \widetilde{O}\left(\min\big\{\sqrt{SAKH^3},\,HK\big\}\right)
	\label{eq:regret-Thm1}
\end{equation}
provided that the constants $c_1,c_2>0$ in \eqref{eq:bh-defn-Bernstein-ucbvi} are suitably chosen. Here, in the definition of $\mathsf{Regret}(K)$ (cf.~\eqref{eq:defn-regret}), $\pi^k$ and $s_1^k$ denote, respectively, the policy deployed and the initial state observed in episode $k$ of Algorithm~\ref{alg:main-mvp}. 
	%where the regret metric will be formally defined in \eqref{eq:defn-regret}.  
\end{theorem}

As we shall confirm shortly in Section~\ref{sec:minimax-lower-bound-online} (see Theorem~\ref{thm:minimax-lower-online}), the above regret guarantee for \mvp matches the fundamental minimax lower bound up to logarithmic factors,  regardless of how large $K$ is.  In contrast to \ucbvi, this means that \mvp essentially eliminates the burn-in phase required to attain statistically optimal performance, with only logarithmic factors remaining. 
By converting this regret guarantee into a PAC-style sample complexity bound (using arguments in Section~\ref{sec:formulation-online}), we immediately obtain the following implication: for an {\em arbitrary} accuracy level $\varepsilon \in (0,H]$, the \mvp algorithm is guaranteed to yield an $\varepsilon$-optimal policy with high probability using at most 
\begin{equation}
	%\text{(sample complexity)}\qquad 
	\widetilde{O}\bigg(\frac{SAH^3}{\varepsilon^2}\bigg) \quad \text{episodes},
\end{equation}
where the output policy can be drawn, for instance, uniformly at random from the policy sequence $\{\pi^k\}_{k=1}^{K}$. 
Equivalently, this corresponds to  $\widetilde{O}\big(\frac{SAH^4}{\varepsilon^2}\big)$ sample transitions, since each sample trajectory has length $H$.     
It is noteworthy that this sample complexity is information-theoretically unimprovable even when one has access to an idealized generative model (cf.~Section~\ref{sec:finite-horizon-generative}).

Taken together, the above results further underscore the statistical optimality of model-based methods in yet another fundamental RL setting: online RL. To date, \mvp remains the only online RL algorithm known to provably attain minimax-optimal regret without incurring any burn-in cost.

A note of caution is warranted, however. At present, it remains unclear whether \ucbvi is fundamentally suboptimal or whether its large burn-in cost is merely an artifact of existing proof techniques, given the absence of algorithm-specific lower bounds for \ucbvi. Having said that, \mvp enjoys an additional practical advantage over \ucbvi: a substantially lower switching cost. Owing to the doubling-based update rule, the sampling policy in \mvp is no longer updated at every episode and may remain fixed for extended periods of time. This low-switching property can significantly reduce deployment and operational costs in applications where changing sampling policies is expensive.

Finally, it is noteworthy that, among the canonical MDP settings commonly studied in online RL, finite-horizon nonstationary MDPs remain the only setting for which minimax-optimal regret without a burn-in period is currently known to be achievable. How to establish similar results in other prominent settings, such as discounted infinite-horizon MDPs and finite-horizon stationary MDPs, remains open.  Resolving these questions will likely require new ideas in both theory and algorithm design.

\subsection{Analysis: proof of Theorem~\ref{thm-mvp-regret}}
\label{sec:proof-of-thm-mvp}

Following the analysis of \ucbvi, it is convenient to introduce the following logarithmic factor:
\begin{align}
	\iotab & \coloneqq 40000 (\log_2 K)^3 \log(3SAH)\log\frac{1}{\delta'} \qquad \text{with}
	\quad \delta' =  \frac{100\delta}{S^2A^2H^2K^2}.
	\label{eq:assumption-K-proof-B-mvp}
\end{align}
As in Section~\ref{sec:proof-of-thm-ucbvi}, we first note that the desired regret bound in Theorem~\ref{thm-mvp-regret} holds trivially whenever $K \leq SAH \iotab$, since
\[
\mathsf{Regret}(K)=\sum_{k=1}^{K}\left(V_{1}^{\star}(s_{1}^{k})-V_{1}^{\pi^{k}}(s_{1}^{k})\right)
\leq HK 
= \min\left\{ \sqrt{ SAH^{3}K\iotab},HK\right\}.
\]
It therefore suffices, for the remainder of this subsection, to focus on the regime where
\begin{equation}
	K\geq  SAH\iotab. 
	\label{eq:assumption-K-proof-mvp}
\end{equation}
% 

%\begin{remark}[Coupling between $\mathcal{D}^{\mathsf{expand}}$ and the original data $\mathcal{D}^{\mathsf{orig}}$] \label{remark:coupling-expand-orig}

 %As will become clear shortly, this auxiliary mechanism often helps simplify the exposition of the analysis.

\paragraph{Preparation: an expanded set of samples from a generative model.} 
We now introduce a seemingly different, yet closely related, sampling mechanism in which independent samples are generated from a generative model.   Specifically, before online learning begins, we use a generative model to draw a collection of independent samples, as defined below.
\begin{definition}[Expanded samples]\label{def:filt2}
	Let $\mathcal{D}^{\mathsf{expand}}$ be a set of $SAHK$ independent samples generated as follows: 
	for each $(s,a,h)\in \mathcal{S}\times \mathcal{A}\times [H]$, draw $K$ independent samples $(s,a,h,s'^{,(i)})$ obeying 
	$$s'^{,(i)} \overset{\mathrm{ind.}}{\sim} P_{s,a,h}
    \qquad \text{for all }1\leq i\leq K.$$
\end{definition}
Importantly, $\mathcal{D}^{\mathsf{expand}}$ can be coupled with the original set of samples, denoted by  $\mathcal{D}^{\mathsf{orig}}$. Specifically, one first generates $\mathcal{D}^{\mathsf{expand}}$ before online learning begins, and thereafter, whenever the online process queries $(s,a,h)$, one returns an unused sample from $\mathcal{D}^{\mathsf{expand}}$,  without replacement. 
This coupling allows us to  analyze the online process through the view of $\mathcal{D}^{\mathsf{expand}}$, which is sometimes more convenient. Unless stated otherwise, all subsequent proofs are carried out under this coupling between $\mathcal{D}^{\mathsf{orig}}$ and $\mathcal{D}^{\mathsf{expand}}$.

\paragraph{Notation.}
As in the analysis of \ucbvi, it is convenient to introduce the following notation to index the iterates in Algorithm~\ref{alg:main-mvp} explicitly by the episode number $k$. Some of the notation defined below is analogous to that previously used in the analysis of \ucbvi in Section~\ref{sec:proof-of-thm-ucbvi}.
\begin{itemize}

	%\item  $\pi^k$: the policy employed at the beginning of the $k$-th episode.  

	\item  $\widehat{P}^k_{s,a,h}\in \mathbb{R}^{S}$: the latest update of the empirical transition probability vector $\widehat{P}_{s,a,h}$ before episode $k$.

	\item  $b_h^k(s,a) \geq 0 $: the latest update of the exploration bonus term $b_h(s,a)$ before episode $k$.

	\item  $N^{k,\mathsf{all}}_{h}(s,a)$: the total visitation count $N^{\mathsf{all}}_{h}(s,a)$ of the $(s,a,h)$-tuple before episode $k$.

	\item  $N_h^k(s,a)$: the visitation count $N_h(s,a)$ of the $(s,a,h)$-tuple of the latest doubling batch used to compute $\widehat{P}_{s,a,h}$ before episode $k$.   
		When $N^{k,\mathsf{all}}_{h}(s,a)=0$, set $N_h^k(s,a)=1$ for ease of presentation.  %\yxc{check}

	\item  $V_h^k\in \mathbb{R}^{S}$: the latest value estimate $V_h^{\ucb}$ before episode $k$. 
    
	\item  $Q_h^k\in \mathbb{R}^{SA}$: the latest Q-estimate $Q_h^{\ucb}$ before episode $k$. 
	
\end{itemize}

\noindent 
We shall also make use of the following notation for the empirical transition probability vector indexed by the batch number.
\begin{itemize}
	\item For any $j\geq 2$ (resp.~$j=1$), let  $\widehat{P}^{(j)}_{s,a,h}$ denote the empirical transition probability vector for $(s,a,h)$ computed from the $j$-th batch  (resp.~the first batch) of data, namely, the $\{2^{j-2}+i\}_{i=1}^{2^{j-2}}$-th samples (resp.~the 1st sample) associated with $(s,a,h)$.  
	For completeness, we set the empirical transition probability vector for the $0$-th batch to be   $\widehat{P}^{(0)}_{s,a,h}(s') = 1/S $ for every $s'\in \cS$.  
	
\item 
With the coupling between $\mathcal{D}^{\mathsf{expand}}$ and $\mathcal{D}^{\mathsf{orig}}$ in mind, 
we will use $\widehat{P}^{(j)}_{s,a,h}$ interchangeably to denote the empirical probability vector constructed from the $j$-th batch of data from either $\mathcal{D}^{\mathsf{orig}}$ or $\mathcal{D}^{\mathsf{expand}}$, as long as no confusion arises. 

\end{itemize}
%

%\end{remark}

\paragraph{Recap: a basic regret decomposition.} 
Repeating the arguments in Steps 1-2 of Section~\ref{sec:proof-of-thm-ucbvi} for \ucbvi yields the same optimism property \eqref{eq:optimism-Q-V-geq} and regret decomposition \eqref{eq:decomposition} for \mvp; specifically, with probability at least $1-2SAHK\delta'$, we have 
\begin{equation}
	Q_h^k(s,a)\geq Q_h^{\star}(s,a) \qquad  \text{and}  \qquad V^k_h(s)\geq V^{\star}_h(s)
	\label{eq:optimism-Q-V-geq-mvp}
\end{equation}
for all $(s,a,h,k)\in \cS\times \cA \times [H]  \times [K]$, and
\begin{align}
\arxiv{&} \mathsf{Regret}(K) \book{&} \leq\underset{\eqqcolon\,T_{1}}{\underbrace{\sum_{k=1}^{K}\sum_{h=1}^{H}\big\langle\widehat{P}_{s_{h}^{k},a_{h}^{k},h}^{k}-P_{s_{h}^{k},a_{h}^{k},h},V_{h+1}^{k}\big\rangle}}+\underset{\eqqcolon\,T_{2}}{\underbrace{\sum_{k=1}^{K}\sum_{h=1}^{H}b_{h}^{k}(s_{h}^{k},a_{h}^{k})}}\nonumber\\
 & \quad+\underset{\eqqcolon\,T_{3}}{\underbrace{\sum_{k=1}^{K}\sum_{h=1}^{H}\big\langle P_{s_{h}^{k},a_{h}^{k},h}-e_{s_{h+1}^{k}},V_{h+1}^{k}\big\rangle}}+\underset{\eqqcolon\,T_{4}}{\underbrace{\sum_{k=1}^{K}\left(\sum_{h=1}^{H}r_{h}(s_{h}^{k},a_{h}^{k})-V_{1}^{\pi^{k}}(s_{1}^{k})\right)}} .
	\label{eq:decomposition-mvp}
\end{align}
The terms $T_2,\ldots,T_4$ can be handled in a similar manner as in the analysis of \ucbvi in Section~\ref{sec:proof-of-thm-ucbvi}. As alluded to previously, obtaining a sharp bound on $T_1$ constitutes one of the main objectives of the analysis that follows.

\paragraph{Step 1: a compressed representation of visitation counts $\{N_{h}^{k}(s,a)\}$.} 
As discussed in Section~\ref{sec:revisit-ucbvi-sample-limited}, a potential covering-based analysis strategy (see Strategy 2 therein) is to first condition on the visitation counts $\{N_{h}^{k}(s,a)\}$ when carrying out the concentration-of-measure analysis, and then take a union bound over all possible configurations of these visitation counts. A crucial step in implementing this strategy is therefore to control the total number of admissible configurations --- namely, an appropriate covering number --- for $\{N_{h}^{k}(s,a)\}$. To this end, we introduce the notion of a {\em profile}, a term coined by \citet{zhang2025settling}, which enables a compact representation of the visitation counts. 
\begin{definition}[Profile]
	\label{defn:profile}
Consider any collection $\{N_{h}^{k,\mathsf{all}}(s,a)\}$. 
For any $(s,a,h,k)\in \cS\times \cA\times [H] \times [K]$, define
\begin{align*}
	I^k_{s,a,h} \coloneqq \begin{cases} \max\big\{ j\in \mathbb{Z}: 2^{j-1} \leq N_{h}^{k,\mathsf{all}}(s,a) \big\} , & \text{if }N_{h}^{k,\mathsf{all}}(s,a)>0, \\
	0, & \text{if }N_{h}^{k,\mathsf{all}}(s,a)=0. \end{cases}
\end{align*}
The profile of the $k$-th episode $(1\leq k\leq K)$ and the corresponding total profile are then defined, respectively, as 
\begin{align*}
	\mathcal{I}^k &\coloneqq \big\{I^k_{s,a,h} \big\}_{(s,a,h)\in \mathcal{S}\times \mathcal{A}\times [H]} \qquad
	\text{and}\qquad \mathcal{I} \coloneqq \{\mathcal{I}^k\}_{k=1}^K.
	%\label{eq:defn-profile-Ik}
\end{align*}
\end{definition}
By construction, given a total profile $\mathcal{I}$ associated with $\{N_h^{k,\mathsf{all}}(s,a)\}$, the empirical transition kernels satisfy
\begin{equation}
	\widehat{P}^{k}_{s,a,h} = \widehat{P}^{(I^k_{s,a,h})}_{s,a,h},\qquad \forall (s,a,h,k)\in \mathcal{S}\times \mathcal{A}\times [H]\times [K];
	\label{eq:relation-profile-k}
\end{equation}
here, $\widehat{P}^{(j)}_{s,a,h}$ is as defined earlier in this subsection.  
Thus, a total profile fully determines when and where the empirical transition models are updated.  
Moreover, since each visitation count $N_h^k(s,a)$ is recomputed only when the corresponding empirical model is updated (see line~\ref{line:Nh-update-mvp} of Algorithm~\ref{alg:main-mvp}), the total profile serves as a compact representation of the entire collection $\{N_h^k(s,a)\}$.

To quantify the extent of the resulting compression (which stems from the doubling update mechanism), we next bound the total number of admissible profiles.
\begin{lemma}\label{lemma:key2} 
	Suppose that $K\geq SAH\log_2K$. Then the number of all possible total profiles induced by Algorithm~\ref{alg:main-mvp} is at most $$(4SAHK)^{SAH\log_2K +1}.$$ 
\end{lemma}
\begin{proof}
The key idea is to consider the following set:
\begin{align}
\mathcal{C} \coloneqq 
	\arxiv{&} \Big\{ \mathcal{I}=\{\mathcal{I}^1,\ldots,\mathcal{I}^K\}\book{&}  \,\Big|\, \mathcal{I}^1\leq \mathcal{I}^2\leq \cdots \leq \mathcal{I}^K, \notag\\
    &\qquad \qquad
	\mathcal{I}^{k}\in \big\{0, 1,\cdots,\log_2K\big\}^{SAH} \text{ for all } 1\leq k\leq K \Big\}.
	\label{eq:defn-C-choice}
\end{align}
Since the visitation counts $\{N_h^{k,\mathsf{all}}(s,a)\}$ are non-decreasing in $k$, every total profile induced by Algorithm~\ref{alg:main-mvp} satisfies the monotonicity constraint
 $\mathcal{I}^1\leq \mathcal{I}^2\leq \cdots \leq \mathcal{I}^K$, and hence belongs to $\mathcal{C}$. It therefore suffices to show that
\begin{align}
|\mathcal{C}|\leq (4SAHK)^{SAH\log_2K +1}.
\label{eq:cardinality-C-ub-mvp}
\end{align}
The proof of this cardinality bound is deferred to Section~\ref{app:pfkey2}. \end{proof}

In contrast to the uncompressed setting (such as that arising in \ucbvi), where the visitation counts may admit exponentially many configurations in $K$, the doubling update mechanism adopted in Algorithm~\ref{alg:main-mvp} leads to a substantial compression: by Lemma~\ref{lemma:key2}, the exponent governing the number of admissible configurations grows only logarithmically with $K$. This compression is a key ingredient that makes the covering-based analysis strategy effective.

Before proceeding, we record another useful property of the visitation counts. This result is analogous to Lemma~\ref{lem:sum-inv-Nhk-ucbvi}, but concerns the modified visitation counts associated with the epoch-based updates in \mvp.
\begin{lemma}
\label{lem:sum-inv-Nhk-mvp}
For any $K\geq 2$, Algorithm~\ref{alg:main-mvp} satisfies
\begin{align*}
\sum_{k=1}^K \sum_{h=1}^H \frac{1}{\max\{ N_h^k(s_h^k,a_h^k),1\}}\leq 4SAH\log_2 K.
\end{align*}
\end{lemma}
\begin{proof}
Fix any $(s,a,h)\in\mathcal{S}\times\mathcal{A}\times[H]$. By the doubling rule in Algorithm~\ref{alg:main-mvp}, the visitation count $N_h^k(s,a)$ is within a factor of two of $N_h^{k,\mathsf{all}}(s,a)$. Consequently,
\begin{align*}
	\arxiv{&} \sum_{k=1}^K \frac{1}{\max\{ N_h^k(s_h^k,a_h^k),1\}}  \mathds{1}\big\{(s,a)=\big(s_h^k,a_h^k \big) \big\} \arxiv{ \notag\\}
    & \arxiv{\quad}  \leq \sum_{k=1}^K \frac{2}{\max\{ N_h^{k,\mathsf{all}}(s_h^k,a_h^k),1\}}  \mathds{1}\big\{(s,a)=\big(s_h^k,a_h^k \big) \big\} \book{\\}
 \book{&}    \leq 4\log_2 K,
\end{align*}
where the last inequality follows from the same argument as in the proof of Lemma~\ref{lem:sum-inv-Nhk-ucbvi}.  
\end{proof}

\paragraph{Step 2: decomposing $T_1$ based on the doubling updates.} 
With the notion of profiles in place, 
we now turn to the term $T_1$ in \eqref{eq:decomposition-mvp}. To facilitate the analysis, we first introduce some additional notation.
\begin{itemize}
    \item $\mathcal{I}^{\mathsf{true}}=\{\mathcal{I}^{1,\mathsf{true}},\cdots,\mathcal{I}^{K,\mathsf{true}}\}$, 
with $\mathcal{I}^{k,\mathsf{true}}=\{I_{s,a,h}^{k,\mathsf{true}}\}$,   
denotes the total profile induced by the actual visitation counts $\{N_{h}^{k,\mathsf{all}}(s,a)\}$ generated during the online learning process; 

\item $k_{l,j,s,a,h}$: the episode index corresponding to the $(2^{l-1}+j)$-th visit 
to 
$(s,a,h)$ during the online learning process.
\end{itemize}
\noindent 
In view of~\eqref{eq:relation-profile-k}, we can reorganize $T_1$ according to the doubling updates as follows
\begin{align}
 T_1 &=  \sum_{k=1}^{K}\sum_{h=1}^{H}\Big\langle\widehat{P}_{s_{h}^{k},a_{h}^{k},h}^{k}-P_{s_{h}^{k},a_{h}^{k},h},V_{h+1}^{k}\Big\rangle \notag\\
 &=\sum_{k=1}^{K}\sum_{h=1}^{H}\Big\langle\widehat{P}_{s_{h}^{k},a_{h}^{k},h}^{(I_{s_{h}^{k},a_{h}^{k},h}^{k,\mathsf{true}})}-P_{s_{h}^{k},a_{h}^{k},h},V_{h+1}^{k}\Big\rangle\nonumber\nonumber\\
 & =\sum_{l=0}^{\log_{2}K}\sum_{s,a,h}\bigg\langle\widehat{P}_{s,a,h}^{(l)}-P_{s,a,h},\,\sum_{k=1}^{K} V_{h+1}^{k}\mathds1\left\{ (s_{h}^{k},a_{h}^{k})=(s,a),I_{s,a,h}^{k,\mathsf{true}}=l\right\} \bigg\rangle\nonumber\nonumber\\
 & \leq\sum_{l=1}^{\log_{2}K}\sum_{s,a,h}\bigg\langle\widehat{P}_{s,a,h}^{(l)}-P_{s,a,h},\,\sum_{k=1}^{K}V_{h+1}^{k}\mathds1\left\{ (s_{h}^{k},a_{h}^{k})=(s,a),I_{s,a,h}^{k,\mathsf{true}}=l\right\} \bigg\rangle \arxiv {\notag\\ }
 \arxiv{ &\qquad\qquad } + SAH^{2}\nonumber\nonumber\\
 & =\sum_{l=1}^{\log_{2}K}\sum_{j=1}^{2^{l-1}}\sum_{s,a,h}\Big\langle\widehat{P}_{s,a,h}^{(l)}-P_{s,a,h},V_{h+1}^{k_{l,j,s,a,h}}\Big\rangle+SAH^{2},
	\label{eq:PV-sum-decompose}
\end{align}
where, for notational convenience, we set $V_{h+1}^k=0$ for all $k>K$.  Here, the second line results from \eqref{eq:relation-profile-k},   the penultimate line follows from the bound $0\leq V_{h+1}^{k}(s)\leq H$ for all $s\in\mathcal{S}$, while the last identity follows from the fact that, for each $(s,a,h)$ and $l\geq1$, the episodes with $I_{s,a,h}^{k,\mathsf{true}}=l$ correspond to the $(2^{l-1}+j)$-th visits to $(s,a,h)$, for $1\leq j\leq2^{l-1}$.  

The key point of \eqref{eq:PV-sum-decompose} is that it reorganizes the original sum by the doubling level $l$, grouping together all terms associated with the same empirical transition estimate $\widehat{P}_{s,a,h}^{(l)}$. It therefore suggests first controlling the term 
\begin{align}
\label{eq:sum-variation-Vhk-batch}
\sum_{s,a,h}\big\langle\widehat{P}_{s,a,h}^{(l)}-P_{s,a,h},V_{h+1}^{k_{l,j,s,a,h}}\big\rangle.
\end{align}
As before, the main challenge is the intricate statistical dependence between $\widehat{P}_{s,a,h}^{(l)}$ and $V_{h+1}^{k_{l,j,s,a,h}}$. We address this dependence issue next.

\paragraph{Step 3: bounding the term~\eqref{eq:sum-variation-Vhk-batch} via coupling and covering.}
To obtain a sharp bound on \eqref{eq:sum-variation-Vhk-batch}, we employ a covering argument based on the profiles introduced above. We first describe the main idea, before providing the details.
\begin{itemize}
	\item[i)] For any fixed total profile $\mathcal{I}$ and fixed $ l\leq \log_2K$, 
		derive a high-probability bound for weighted sums of the form
		\begin{equation}
			\sum_{s,a,h} \Big(\widehat{P}^{(l)}_{s,a,h}-P_{s,a,h} \Big) X_{h+1,s,a},
			\label{eq:P-X-term}
		\end{equation}
		where each vector $X_{h+1,s,a}$ is a deterministic function of the (fixed) profile $\mathcal{I}$ and the samples collected at steps $h'\geq h+1$. Here, under the view of the expanded sample set $\mathcal{D}^{\mathsf{expand}}$ (cf.~Definition~\ref{def:filt2}), $\widehat{P}_{s,a,h}^{(l)}$ is constructed from the $l$-th batch of samples associated with $(s,a,h)$. These samples are independent of those determining $X_{h+1,s,a}$ once $\mathcal{I}$ is fixed, allowing us to invoke standard martingale concentration inequalities.

	\item[ii)] Take a union bound over all admissible total profiles $\mathcal{I}$. In view of the profile-counting bound in Lemma~\ref{lemma:key2}, this yields uniform control of \eqref{eq:P-X-term} simultaneously over all profiles $\mathcal{I}$ and all associated sequences $\{X_{h+1,s,a}\}$.

	\item[iii)] Show that the above uniform bound can be applied to the decomposition in \eqref{eq:PV-sum-decompose} to obtain the desired bound.

\end{itemize}

\paragraph{Step 3.1: establishing uniform bounds on~\eqref{eq:P-X-term}.} 
We first carry out steps i) and ii). To formalize the class of vectors $\{X_{h,s,a}\}$ appearing in \eqref{eq:P-X-term}, we introduce the following assumption.
\begin{assumption}
\label{assumption-X-mvp}
 For each total profile $\mathcal{I} \in \mathcal{C}$ (cf.~\eqref{eq:defn-C-choice}), 
suppose the set $\big\{ \mathcal{X}_{h,\mathcal{I}} \big\}_{1\leq h\leq H}$ satisfies the following properties for every $1\leq h\leq H$:
\begin{itemize}

	\item $\mathcal{X}_{h,\mathcal{I}}$  is a set of no more than $K+1$ nonnegative vectors in $\mathbb{R}^S$, including the all-zero vector $0$, and each vector $X\in \mathcal{X}_{h,\mathcal{I}}$ obeys $\|X\|_{\infty}\leq H$;

	\item    $\mathcal{X}_{h+1,\mathcal{I}}$ is a collection determined by a {\em deterministic} mapping of $\mathcal{I}$, the reward function $r$, and the empirical distribution vectors (from the view of the expanded sample set  $\mathcal{D}^{\mathsf{expand}}$) 
		\[
			\Big\{\widehat{P}^{(I^k_{s,a,h'})}_{s,a,h'} \Big\}_{ h'\in (h, H],\, (s,a,k)\in \mathcal{S}\times \mathcal{A} \times [K]}.
		\]

	%\item $\|X\|_{\infty}\leq H$ holds for each vector $X\in \mathcal{X}_{h,\mathcal{I}}$;

\end{itemize}
\end{assumption}
\noindent 
For each collection $\big\{ \mathcal{X}_{h,\mathcal{I}} \big\}$ satisfying the above assumptions, 
we can readily establish a uniform concentration bound by combining Freedman's martingale concentration inequality with a union bound; the proof is postponed to Section~\ref{app:pfuniform}.  
\begin{lemma}\label{lemma:uniform}
	Suppose that $K\geq SAH\log_2K$. For each $\mathcal{I}\in \mathcal{C}$ (cf.~\eqref{eq:defn-C-choice}), consider a set $\big\{ \mathcal{X}_{h,\mathcal{I}} \big\}_{1\leq h\leq H}$ satisfying Assumption~\ref{assumption-X-mvp}. Then, with probability at least $1- \delta'$, 
%
%\book{\begingroup\small} 
\begin{align}
	& \sum_{(s,a,h)\,\in\, \mathcal{S}\times \mathcal{A}\times [H]}\big\langle \widehat{P}_{s,a,h}^{(l)}-P_{s,a,h}, X_{h+1,s,a} \big\rangle 
	 \arxiv{\notag\\ 
    &  \qquad }  \leq \sum_{(s,a,h)\,\in\, \mathcal{S}\times \mathcal{A}\times [H]} \max\Big\{ \big\langle \widehat{P}_{s,a,h}^{(l)}-P_{s,a,h}, X_{h+1,s,a} \big\rangle, 0 \Big\}
	\nonumber\\
	&\qquad  \leq \sqrt{\frac{SAH\iotab}{2^{l}\log_2K}\sum_{s,a,h}\mathsf{Var}_{P_{s,a,h}}\big(X_{h+1,s,a}\big)}  \arxiv{\notag\\}
	\arxiv{ &\qquad\qquad\qquad\qquad } +\frac{SAH^2\iotab}{2^{l}\log_2K}\label{eq:xx1-aux-1}
\end{align}
%\book{\endgroup} 
%
holds simultaneously for all $\mathcal{I} \in \mathcal{C}$, all $2\leq l\leq\log_{2}K+1$, and all sequences $\{X_{h,s,a}\}_{(s,a,h)\in\mathcal{S}\times\mathcal{A}\times[H]}$
obeying $X_{h,s,a}\in\mathcal{X}_{h,\mathcal{I}},$ $\forall (s,a,h)\in \mathcal{S}\times\mathcal{A}\times[H]$.  
\end{lemma}

\paragraph{Step 3.2: coupling back to the online process.} 
Next, we carry out iii) through the following lemma. This result is proven by combining the uniform bound in Lemma~\ref{lemma:uniform} with the decomposition~\eqref{eq:PV-sum-decompose}, which allows one to rigorously couple $\mathcal{D}^{\mathsf{expand}}$ with the online learning process.  The proof is deferred to Section~\ref{app:pfdecouple}. 

\begin{lemma}\label{lemma:decouple}
Suppose that $K\geq SAH\log_2K$, and recall the definition of $\iotab$ in \eqref{eq:assumption-K-proof-B-mvp}. With probability exceeding $1-\delta'$, we have
%
%\book{\begingroup\small} 
\begin{align*} 
	\arxiv{&} \sum_{k=1}^{K}\sum_{h=1}^{H}\Big\langle\widehat{P}_{s_{h}^{k},a_{h}^{k},h}^{k}-P_{s_{h}^{k},a_{h}^{k},h},V_{h+1}^{k}\Big\rangle
	\arxiv {\notag\\}
     &\leq \sum_{k=1}^{K}\sum_{h=1}^{H}\max\Big\{\Big\langle\widehat{P}_{s_{h}^{k},a_{h}^{k},h}^{k}-P_{s_{h}^{k},a_{h}^{k},h},V_{h+1}^{k}\Big\rangle,0\Big\} \\
	& \arxiv{\qquad}\leq \sqrt{SAH\iotab\sum_{k=1}^{K}\sum_{h=1}^{H}\mathsf{Var}_{P_{s_{h}^{k},a_{h}^{k},h}}\big(V_{h+1}^{k}\big)}  
    +SAH^{2}\iotab
\end{align*}
%\book{\endgroup}
%
and
%
%\book{\begingroup\small} 
\begin{align*} 
	& \sum_{k=1}^{K}\sum_{h=1}^{H}\Big\langle\widehat{P}_{s_{h}^{k},a_{h}^{k},h}^{k}-P_{s_{h}^{k},a_{h}^{k},h},\big(V_{h+1}^{k}\big)^{2}\Big\rangle \arxiv{\\
 &} \arxiv{\qquad}\leq \sqrt{SAH^3\iotab \sum_{k=1}^{K}\sum_{h=1}^{H}\mathsf{Var}_{P_{s_{h}^{k},a_{h}^{k},h}}\big(V_{h+1}^{k}\big)}+SAH^{3}\iotab.
\end{align*}
%\book{\endgroup}
% 
\end{lemma}
\begin{remark}
In addition to controlling $T_1$, Lemma~\ref{lemma:decouple} provides bounds on the two related quantities: $$\sum_{k,h}\max\Big\{\big\langle\widehat{P}_{s_{h}^{k},a_{h}^{k},h}^{k}-P_{s_{h}^{k},a_{h}^{k},h},V_{h+1}^{k}\big\rangle,0\Big\}
\quad \text{and} \quad
\sum_{k,h}\Big\langle\widehat{P}_{s_{h}^{k},a_{h}^{k},h}^{k}-P_{s_{h}^{k},a_{h}^{k},h},\big(V_{h+1}^{k}\big)^{2}\Big\rangle.
$$
These bounds follow from the same argument and will be useful in the subsequent analysis. 
\end{remark}

\paragraph{Step 4: combining  and solving the coupled  inequalities.}
We now assemble the estimates (e.g., bounds on the quantities $T_1,\ldots,T_4$ in \eqref{eq:decomposition-mvp}) established in the previous steps and complete
the regret analysis.
As in the proof of \ucbvi
(Section~\ref{sec:proof-of-thm-ucbvi}), it is convenient to introduce the
following auxiliary quantities:  
\begin{subequations}
\label{eq:defn-T56-proof-mvp}
\begin{align}
	T_5 &\coloneqq \sum_{k=1}^K\sum_{h=1}^H\mathsf{Var}_{\widehat{P}^k_{s_h^k,a_h^k,h}} \big( V_{h+1}^k \big),
	\label{eq:defn-T5-proof-mvp} \\
	T_6^{\ucb} &\coloneqq \sum_{k=1}^K \sum_{h=1}^H\mathsf{Var} _{P_{s_h^k,a_h^k,h}}\big( V_{h+1}^k \big) ,
	\label{eq:defn-T6-proof-mvp} \\
    W_3 & \coloneqq\sum_{k=1}^{K}\sum_{h=1}^{H}\Big\langle\widehat{P}_{s_{h}^{k},a_{h}^{k},h}^{k}-P_{s_{h}^{k},a_{h}^{k},h},\big(V_{h+1}^{k}\big)^{2}\Big\rangle.
    \label{eq:defn-W_3-proof-mvp} \\
    W_4 & \coloneqq \sum_{k=1}^{K}\sum_{h=1}^{H}\max\Big\{\Big\langle\widehat{P}_{s_{h}^{k},a_{h}^{k},h}^{k}-P_{s_{h}^{k},a_{h}^{k},h},V_{h+1}^{k}\Big\rangle,0\Big\}, 
    \label{eq:defn-W_4-proof-mvp}
\end{align}
\end{subequations}
These quantities naturally arise in the variance decomposition and allow the
different error terms to be coupled through a small collection of inequalities.

We begin by bounding $T_1$ (cf.~\eqref{eq:decomposition-mvp}), $W_3$ and $W_4$ (cf.~\eqref{eq:defn-T56-proof-mvp}) in terms of $T_6^{\ucb}$ (cf.~\eqref{eq:defn-T6-proof-mvp}). In view of Lemma~\ref{lemma:decouple}, with probability at least $1-\delta'$,
\begin{subequations}
\label{eq:T123456-ub-analysis-mvp}
\begin{align}
T_{1}  \leq W_{4}&\leq\sqrt{T_{6}^{\ucb}SAH\iotab}+SAH^{2}\iotab;
\label{eq:T1-W4-ub-analysis-mvp}\\
W_{3} & \leq \sqrt{T_{6}^{\ucb}SAH^3\iotab}+SAH^{3}\iotab.
\label{eq:W3-ub-analysis-mvp}
\end{align}
The other terms $T_2,T_3$, and $T_4$ (cf.~\eqref{eq:decomposition-mvp}) are controlled exactly as in the
analysis of \ucbvi (see Lemma~\ref{lem:bound-T234}), yielding
\begin{align}
T_{2} & \le\frac{1}{2}\sqrt{SAHT_{5}\iotab}+\frac{1}{2}SAH^{2}\iotab\label{eq:T2-ub-analysis-mvp}\\
|T_{3}| & \leq\sqrt{T_{6}^{\ucb}\iotab}+H\iotab\label{eq:T3-ub-analysis-mvp}\\
|T_{4}| & \leq\sqrt{KH^{2}\iotab}\label{eq:T4-ub-analysis-mvp}
\end{align}
with probability at least $1-(4\log_2K)\delta'$. 
In addition, repeating the same arguments as in the proof of Lemma~\ref{lem:bound-T56} reveals that, 
with probability at least $1-2\delta'\log(KH)$, 
\begin{align}
T_{5} & \leq W_3+ 2HT_{2}
+\sqrt{H^{2}T_6^{\ucb}\iotab
}+KH^{2}\iotab, 
\label{eq:T5-ub-analysis-mvp}
\end{align}

At this point, the only term that has not yet been controlled is $T_6^{\ucb}$, which is still intertwined with the other quantities. In the analysis of \ucbvi
(Section~\ref{sec:proof-of-thm-ucbvi}),
this quantity was bounded indirectly by first controlling
$T_6^{\mathsf{low}}$ and $T_6^{\mathsf{diff}}$ separately. Here, we instead bound $T_6^{\ucb}$ directly in terms of
$T_2$ and $W_4$, thereby closing the system of inequalities. This result is formalized in the following lemma; the proof is provided in Section~\ref{sec:pflem:bound-T6-ucb-mvp}. 
\begin{lemma}
\label{lem:bound-T6-ucb-mvp}
With probability at least $1-4SAHK\delta'$,
\begin{align}
T_{6}^{\ucb} & \leq 4HT_{2}
+ 4HW_4
    + KH^{2} \iotab.
    \label{eq:T6-ub-analysis-mvp}
\end{align}
\end{lemma}
%

%As an immediate consequence of this lemma, we have
%%
%\begin{align}
%T_{6}^{\ucb} & \leq 4HT_{2}
%+ 4HW_4
%    + KH^{2} \iotab, 
%    \label{eq:T6-ub-analysis-mvp}
%\end{align}
%
%thereby bounding $T_{6}^{\ucb}$ in terms of the quantities $T_2$ and $W_4$, both of which have already been controlled.   

\end{subequations}
Thus, $T_6^{\ucb}$ is now bounded entirely in terms of quantities that have
already been bounded, completing the chain of recursive estimates.

We now have a closed system of inequalities.  
Solving this system of inequalities in \eqref{eq:T123456-ub-analysis-mvp} and combining the resulting estimates with the regret
decomposition \eqref{eq:decomposition-mvp}, we can establish that
\begin{align}
   \mathsf{Regret}(K)\le   T_1+T_2+T_3+T_4 \lesssim \sqrt{SAH^3K}\iotab 
   \label{eq:final-mvp-solve-inequalities}
\end{align}
as claimed. Since solving the coupled inequalities involves only routine algebraic manipulations, we defer the details to Section~\ref{sec:solve-all-inequalities-mvp}.

\section{Model-free online RL}
\label{sec:model-free-onlineRL}

Having established the statistical efficiency of model-based methods for online RL, we now shift our attention to model-free algorithms. In contrast to their model-based counterparts, the best currently known model-free approaches still fail to achieve minimax-optimal regret guarantees over the full range of sample sizes. For this reason, rather than providing a treatment as comprehensive as that for model-based methods, we restrict ourselves to a brief overview of the main ideas underlying optimistic Q-learning algorithms. 

Broadly speaking, the key question is how to incorporate the principle of optimism into Q-learning in order to encourage efficient exploration, for instance by constructing suitable upper confidence estimates for the optimal Q-values. One of the most notable developments along this direction is the optimistic Q-learning algorithm --- commonly referred to as \ucbq --- developed by \citet{jin2018q}, whose complete procedure is summarized in Algorithm~\ref{alg:main-q-ucb}. Compared to the vanilla Q-learning algorithm introduced in Chapter~\ref{chapter:generative-model-MF}, the update rule of \ucbq incorporates two key modifications: 
\begin{itemize}

\item First, the Q-learning update now incorporates an exploration bonus $b_n$, analogous to the bonus used in the \ucbvi algorithm, in order to ensure that the learned Q-function remains an optimistic estimate (i.e., an upper confidence bound) of the optimal Q-function. More specifically, the optimistic Q-learning update takes the form
\begin{align*} 
					Q_h^{\ucb}(s_h,a_h) &\leftarrow ( 1- \eta_n) Q^{\ucb}_h(s_h,a_h) \arxiv{  \notag\\}
                \arxiv{    &\qquad} + \eta_n \big\{ r_h(s_h,a_h)+ V^{\ucb}_{h+1}(s_{h+1}) + b_n  \big\}, \\
						V^{\ucb}_{h}(s_h) &\leftarrow \min\Big\{ \max_{a'\in \cA}Q^{\ucb}_h(s_h,a'),H\Big\},
					\end{align*}
			%\end{subequations}
%
where $Q_h^{\ucb}$ (resp.~$V_h^{\ucb}$) denotes the running estimate of the optimal Q-function (resp.~value function), and $b_n$ denotes the exploration bonus depending on the visitation count $n$ described above. 
A simple and natural choice of the bonus term can be derived from Hoeffding-type concentration inequalities:
\begin{equation} \label{eq:hoeffding_ucbq}
b_n = c_{\mathsf{b}} \sqrt{\frac{H^3 \log(SAT/\delta)} {n}},
\end{equation}
where $c_{\mathsf{b}}>0$ is a suitable universal constant, and $\delta \in (0,1)$ is the usual confidence parameter.

\item Second, the learning rate for each visited entry $(s_h,a_h)$ is chosen adaptively according to the following linearly rescaled schedule:
\begin{equation}\label{eq:learning_rate_ucbq}
\eta_n = \frac{H+1}{H+n},
\end{equation}
where $n$ denotes the number of visits accumulated thus far for $(s_h,a_h)$ at step $h$. This carefully designed learning-rate schedule plays an important role in controlling error propagation across the horizon and, consequently, in achieving a regret bound with polynomial dependence on $H$.

\end{itemize}

The resulting regret guarantee for this model-free online RL algorithm, originally established by \citet{jin2018q}, is summarized below. 
\begin{theorem}[Jin et al. (2018)]\label{thm-ucbq-regret}  
For any $K \ge 1$ and any $0<\delta<1$, the \ucbq algorithm in Algorithm~\ref{alg:main-q-ucb} with Hoeffding-style bonuses \eqref{eq:hoeffding_ucbq} achieves
\begin{equation}
	\mathsf{Regret}(K) 
	\leq \widetilde{O}\Big( \sqrt{SAH^{5}K} \Big)
	%\leq \widetilde{O}\left(\min\big\{\sqrt{SAKH^3},\,HK\big\}\right)
	\label{eq:regret-ucbq}
\end{equation}
with probability at least $1-\delta$, provided that the constant $c_{\mathsf{b}}>0$ in \eqref{eq:hoeffding_ucbq} is suitably chosen.  
\end{theorem}

In comparison with the statistical guarantees achieved by model-based methods (see, e.g., Theorem~\ref{thm-mvp-regret}), the regret guarantee in Theorem~\ref{thm-ucbq-regret} remains statistically suboptimal in its dependence on the horizon length $H$. This dependence can be improved through more refined uncertainty quantification for the exploration bonus. For instance, incorporating Bernstein-style exploration bonuses reduces the regret bound to $\widetilde{O}(\sqrt{SAH^4K})$ for sufficiently large $K$ \citep{jin2018q}.

Moreover, additional variance-reduction techniques --- often referred to as {\em reference-advantage decomposition} in the online model-free RL literature \citep{zhang2020almost,li2021breaking} --- can further improve the regret guarantee to become asymptotically minimax optimal. Nevertheless, these refined guarantees still rely on an extremely large burn-in phase before statistically optimal behavior is attained. We refer interested readers to the aforementioned work for further details.

\begin{algorithm}[t]
	\DontPrintSemicolon
\caption{Q-learning with upper confidence bounds  (\ucbq) for episodic online RL\label{alg:main-q-ucb}}
%\begin{algorithmic}[ht]

	{\small 

	\textbf{input:}  number of episodes $K$, reward function $r$, confidence parameter $\delta$. \\
	%$c_1=\frac{460}{9}$, $c_2 = 2\sqrt{2}$, $c_3=\frac{544}{9}$.  
	
	\textbf{initialize} 
	%set $\delta' \leftarrow \frac{\delta}{200SAH^2K^2}$; and set 
	%$N^{\mathsf{all}}_h(s,a,s')\leftarrow 0$, 
	 $N_h(s,a)\leftarrow 0$, $Q_h^{\ucb}(s,a)\leftarrow H$, and $V_{H+1}(s)\leftarrow 0$ for all $(s,a,h)\in\mathcal{S}\times \cA\times [H]$. \\
%\textbf{initialization: } 
	\For{$k=1,\cdots,K$} {
	{\color{blue}\tcc{policy update.}} 
    update policy $\pi$ such that $\pi_h(s) \leftarrow \arg\max_{a}Q^{\ucb}_h(s,a)$ for all $(s,h)\in \mathcal{S}\times [H]$.  \\
    
        observe the initial state  $s_1\overset{\mathrm{ind.}}{\sim} \rho$ from a new episode. \\
		\For {$h=1,\cdots,H$} {
			{\color{blue}\tcc{sampling.}}
			take $ a_h= \pi_h (s_h)$, 
			observe $s_{h+1}$; set $(s,a,s')\leftarrow (s_h,a_h,s_{h+1})$.  \label{line:choose_action-ucbq} 
			\\
			
			{\color{blue}\tcc{ update of visitation counts.}}
			 $  N_h(s,a) \leftarrow  N_h( s,a )+1$; $n\leftarrow N_h(s,a)$.

		{\color{blue}\tcc{optimistic update of Q-functions \& values.}}
			%set $V^{\ucb}_{H+1}(s)\leftarrow 0$ for all $s\in \mathcal{S}$. \\

		compute $\eta_n$ and $b_n$ according to \eqref{eq:learning_rate_ucbq} and \eqref{eq:hoeffding_ucbq}. \\
        update
			\begin{subequations}
				\label{eq:updateqv-ucbq}
				\begin{align} 
					Q_h^{\ucb}(s,a) &\leftarrow ( 1- \eta_n) Q^{\ucb}_h(s,a)+ \eta_n \big\{ r_h(s,a)+ V^{\ucb}_{h+1}(s') + b_n  \big\}, \label{eq:updateq-ucbq}\\
						V^{\ucb}_{h}(s) &\leftarrow \min\Big\{ \max_{a'\in \cA}Q^{\ucb}_h(s,a'),H\Big\}.
						\label{eq:updatev-ucbq}
					\end{align}
			\end{subequations}
            }
		}
	}
%\end{algorithmic}
\end{algorithm}

\subsection{Analysis: proof of Theorem~\ref{thm-ucbq-regret}}
\label{sec:proof-thm-ucbq-regret}

In this subsection, we provide the proof of Theorem~\ref{thm-ucbq-regret}, 
following the proof strategy developed in \citet[Section~4]{jin2018q}.

\paragraph{Preliminary properties of the learning rates.}
We begin by recording several useful properties of the learning rates \eqref{eq:learning_rate_ucbq}. For notational convenience, we also introduce two related sequences, defined for integers $N\geq 0$ and $n\geq 1$:  
\begin{subequations}
\label{eq:learning-rate-notation-all-ucbq}
\begin{equation}
	\label{equ:learning_rate_notation}
	\eta_n^N \defn \begin{cases} \eta_n \prod_{i = n+1}^N(1-\eta_i), & \text{if }N>n, \\ \eta_n,  & \text{if }N=n, \\
	0, &\text{if } N < n, \end{cases}
	%\qquad \text{and} \qquad    
\end{equation}
and
    \begin{equation}
    \label{equ:learning_rate_notation-eta_0n}
    \eta_0^N \defn \begin{cases} \prod_{i=1}^N(1-\eta_i) =0 , & \text{if }N>0, \\ 1, & \text{if }N=0. \end{cases}	
\end{equation}
\end{subequations}
As can be easily verified, it holds that
\begin{align}
	\sum_{n=1}^N \eta_n^N =  \begin{cases} 1,\qquad  & \text{if }N > 0, \\ 0, & \text{if }N=0.  \end{cases}
	\label{eq:sum-eta-n-N}
\end{align}
The following properties play an important role in the analysis; the proof is deferred to Section~\ref{sec:proof-properties-learning-rates}. 
\begin{lemma}
	\label{lemma:property_lr} 
	For the learning rates in \eqref{eq:learning_rate_ucbq}, we have, for every integer $N\geq 1$,
\begin{subequations}
\label{eq:properties-learning-rates}
\begin{align}
	 \frac{1}{\sqrt{N}}  \le\sum_{n=1}^{N}\frac{\eta_{n}^{N}}{\sqrt n}&\le\frac{2}{\sqrt{N}},
	\label{eq:properties-learning-rates-sqrt}  \\
 \sum_{n=1}^{N}(\eta_{n}^{N})^{2} &\le\frac{2H}{N}.
	\label{eq:properties-learning-rates-square}
\end{align}
In addition, for every integer $n\geq 1$, one has
\begin{equation}
\sum_{N=n}^{\infty}\eta_{n}^{N}=1+\frac{1}{H}.
	 \label{eq:properties-learning-rates-tail}
\end{equation}
\end{subequations}
\end{lemma}
\begin{remark}
The tail identity~\eqref{eq:properties-learning-rates-tail} will be used
below to control how an estimation error generated at one visit propagates
to subsequent visits. 
\end{remark}

\paragraph{Notation.}
 In the sequel, we adopt the following convenient notation, indexed explicitly by the episode number $k$. 
\begin{itemize}
	\item $(s_h^k, a_h^k)$: the state-action pair encountered and chosen at  step $h$ in the $k$-th episode;
    
    \item  $N_h^k(s,a)$: the visitation count $N_h(s,a)$ for the $(s,a,h)$-tuple before episode $k$;  
    
	\item  $b_h^k(s,a) \geq 0 $: the latest bonus term $b_h(s,a)$ before episode $k$;  
	
	\item $P_h^k \in \{0,1\}^{1\times S}$: the empirical transition at time step $h$ in the $k$-th episode, namely, 
		\begin{equation}
			\forall s\in \cS: \qquad P_h^k (s) = \ind\big\{ s = s_{h+1}^k \big\}.
			\label{eq:P-hk-defn-s}
		\end{equation}

	\item $ Q_h^k(s,a)$ and $V_h^k(s)$: the latest $Q_h^{\ucb}(s,a)$ and $V^{\ucb}_h(s)$ before episode $k$.  
\end{itemize}
\noindent 
Also, throughout this analysis, we denote
\begin{align}
\iota \coloneqq \log(SAKH/\delta).
\end{align}

\paragraph{Step 1: a recursive expression for Q-estimation errors.}
We first unroll the Q-learning updates to derive a recursive expression for the Q iterates. For any $(s,a,h,k)\in \cS\times \cA\times [H]\times [K]$, 
suppose that $(s,a)$ was previously visited $t$ times at step $h$, and 
denote by
$ 
k_1<\cdots<k_t<k
$
these previous episodes. Armed with the notation \eqref{eq:learning-rate-notation-all-ucbq}, unrolling the corresponding updates then yields
\begin{equation}
Q_h^k(s,a)=\eta_0^tH+
\sum_{i=1}^t\eta_i^t\Big\{r_h(s,a)+V_{h+1}^{k_i}(s_{h+1}^{k_i})+b_h^{k_i}(s,a)\Big\},
\label{eq:unroll-q-ucbq}
\end{equation}
where we use the initialization $Q_h^0=H$.
Combining this with the Bellman equation $Q_h^{\star}(s,a)=r_h(s,a)+P_{s,a,h}V_{h+1}^{\star}$ results in
\begin{align}
& Q_h^k(s,a)-Q_h^{\star}(s,a) \nonumber \\
&=\eta_0^t\big(H-Q_h^{\star}(s,a)\big)  \arxiv{  \nonumber\\}
\arxiv{&\quad } + \sum_{i=1}^t \eta_i^t \Big\{ \big(V_{h+1}^{k_i}-V_{h+1}^{\star}\big)(s_{h+1}^{k_i})
+ (P_h^{k_i}-P_{s,a,h})V_{h+1}^{\star} + b_h^{k_i}(s,a)\Big\}, 
\label{eq:q-recursion-model-free}
\end{align}
where we have used the identity $\eta_0^t+\sum_{i=1}^t\eta_i^t=1$ (cf.~\eqref{eq:sum-eta-n-N}).  This expression decomposes the estimation error into a weighted
combination of several components: the influence of the initial
estimate, the propagated error from previously updated value functions,
the statistical fluctuations of the observed transitions, and the
exploration bonuses.

\paragraph{Step 2: optimism.} 
We next establish that the Q-estimates are ``optimistic,''  namely, they can upper bound the optimal Q-function. This property, as stated in the lemma below,  parallels Lemma~\ref{lemma:opt} for the model-based approach. Here, $t$ and the indices $k_1<\cdots<k_t$ are defined as in
Step~1.

\begin{lemma}\label{lemma:opt_model_free}
Suppose that the  constant $c_{\mathsf{b}}$ in the bonus choice \eqref{eq:hoeffding_ucbq} is large enough. Then with probability at least $1-\delta$,
\begin{subequations}
\label{eq:opt-upper-q-bound}
\begin{align}
0\le \big(Q_h^k-Q_h^{\star}\big)(s,a)
&\le \eta_0^t H+
\sum_{i=1}^t \eta_i^t \big(V_{h+1}^{k_i}-V_{h+1}^{\star}\big)(s_{h+1}^{k_i}) + \beta_t,
\end{align}
holds simultaneously for all $(s,a,h,k)\in \cS\times \cA\times [H] \times [K]$, where $\beta_t$ is some residual term obeying 
\begin{equation}
\beta_0 = 0,\qquad 
    \beta_t \leq C\sqrt{\frac{H^3\iota}{t}}
    ~\text{ for }t\geq 1
    \label{eq:beta-t-upper-bound-ucbq}
\end{equation}
\end{subequations}
for some absolute constant $C>0$.
Then with probability exceeding $1-\delta$, for all $(s,a,h)$ we have 
\begin{equation} \label{eq:Q-optimism}
	Q_h^k(s,a)\geq Q_h^{\star}(s,a) \qquad  \text{and}  \qquad V^k_h(s)\geq V^{\star}_h(s).
\end{equation}
\end{lemma}

The proof of this lemma is deferred to
Section~\ref{sec:proof-lemma:opt_model_free}. 
Intuitively, when the
exploration bonus is chosen sufficiently large, it compensates for the
statistical uncertainty in the Bellman updates and thereby ensures that
the resulting Q-estimates remain optimistic. The bound on $\beta_t$ in \eqref{eq:beta-t-upper-bound-ucbq} is tailored to the Hoeffding-style exploration bonus in \eqref{eq:hoeffding_ucbq} and can be improved by adopting more refined bonuses, such as Bernstein-style bonuses.

\paragraph{Step 3: regret decomposition.}
We now proceed with the regret analysis. Similar to the analysis of \ucbvi, we first replace the unknown optimal value by the optimistic value estimate: in view of Lemma~\ref{lemma:opt_model_free} (see~\eqref{eq:Q-optimism}), with probability at least $1 - \delta$, 
\begin{equation} 	
	\label{eq:regret_optimism}
\mathsf{Regret}(K) : = \sum_{k=1}^{K}\big(V^{\star}_{1}(s_1^k)-V^{\pi^{k}}_{1}(s_1^k)\big) \leq \sum_{k=1}^{K}\big( V_1^k(s^k_1) - V_1^{\pi^k}(s^k_1) \big). 
\end{equation}

Now, with a slight abuse of notation, let
$t = N_h^k(s_h^k,a_h^k)$ represent
the number of times \((s_h^k,a_h^k)\) has previously been visited at step \(h\), and let
$k_1 < \cdots < k_t < k$
denote the indices of the corresponding episodes. 
Since policy $\pi^k$ is chosen greedily w.r.t.~$Q_h^k$, we can use \eqref{eq:updatev-ucbq} to show that
\begin{align}
\big(V_h^k-V_h^{\pi^k}\big)(s_h^k) 
&
\leq 
 \max_a Q_h^k(s_h^k,a) -V_h^{\pi^k}(s_h^k)
=
 Q_h^k(s_h^k,a_h^k) -V_h^{\pi^k}(s_h^k) 
\nonumber\\ 
&\le \big(Q_h^k-Q_h^{\pi^k}\big)(s_h^k,a_h^k) 
=\big(Q_h^k-Q_h^{\star}\big)(s_h^k,a_h^k)+P_{s_h^k,a_h^k,h}\big(V_{h+1}^{\star}-V_{h+1}^{\pi^k}\big),
\end{align}
where the last relation is valid since, according to the Bellman equations,  
\[
Q_h^{\star}(s,a)-Q_h^{\pi}(s,a)=P_{s,a,h}V_{h+1}^{\star} - P_{s,a,h}V_{h+1}^{\pi}.
\]
Combine this with Lemma~\ref{lemma:opt_model_free} to show that
\begin{align}
\arxiv{&} \big(V_h^k-V_h^{\pi^k}\big)(s_h^k)  \arxiv{ \nonumber\\ }
&\le \eta_0^{t} H+
\sum_{i=1}^{t} \eta_i^{t}
\big(V_{h+1}^{k_i}-V_{h+1}^{\star}\big)(s_{h+1}^{k_i})
+\beta_{t}  +P_{s_h^k,a_h^k,h}\big(V_{h+1}^{\star}-V_{h+1}^{\pi^k}\big) \nonumber\\
&= \eta_0^{t} H+
\sum_{i=1}^{t} \eta_i^{t}
\big(V_{h+1}^{k_i}-V_{h+1}^{\star}\big)(s_{h+1}^{k_i})
+\beta_{t} \arxiv {\nonumber\\}
\arxiv{&\quad } 
%-\big(V_{h+1}^k-V_{h+1}^{\star}\big)(s_{h+1}^k) 
\nonumber\\
&\quad +\big(V_{h+1}^{\star}-V_{h+1}^{\pi^k}\big)(s_{h+1}^k)  + (P_{s_h^k,a_h^k,h}-P_h^k)\big(V_{h+1}^{\star}-V_{h+1}^{\pi^k}\big),
\label{eq:one-step-regret-recursion-ucbq}
\end{align}
where the last term in \eqref{eq:one-step-regret-recursion-ucbq} is a martingale difference term. Importantly, the upper bound in \eqref{eq:one-step-regret-recursion-ucbq} converts the regret at step $h$ into a one-step propagation of value estimation error from step $h+1$, together with a martingale fluctuation caused by the sampled transition.

\paragraph{Step 4: summing one-step recursion.}
The next step is to aggregate the one-step recursion~\eqref{eq:one-step-regret-recursion-ucbq} over episodes. At a high level, we control the bonus terms by summing their learning-rate-weighted contributions and regroup the propagated value estimation errors according to when they were generated. As it turns out, the tail-sum property~\eqref{eq:properties-learning-rates-tail} of the learning rates is crucial, as it ensures that each past error has only a bounded total influence on future updates.

To be precise, we introduce more notation. For each $(k,h)$,  denote by $t(k,h)=N_h^k(s_h^k, a_h^k)$ the number of episodes prior to episode $k$ in which $(s_h^k,a_h^k)$ was visited at step $h$,   and let 
$k_1(k,h) < \cdots < k_t(k,h) < k$
denote the indices of these episodes.  We now handle the first three terms in~\eqref{eq:one-step-regret-recursion-ucbq} separately. 

\begin{itemize}
\item 
The aggregate contribution from the first term on the right-hand side of~\eqref{eq:one-step-regret-recursion-ucbq} is particularly simple to handle. Using the definition~\eqref{equ:learning_rate_notation-eta_0n}, we arrive at
\begin{align}
\sum_{k=1}^K \eta_0^{t(k,h)}H =
    \sum_{k=1}^K \eta_0^{N_h^k(s_h^k,a_h^k)}H
    =\sum_{k=1}^K H\cdot \mathds{1}\{N_h^k(s_h^k,a_h^k)=0\}\le SAH.
    \label{eq:first-visit-term-ucbq}
\end{align}

\item Regarding the third term on the right-hand side of~\eqref{eq:one-step-regret-recursion-ucbq}, observe that for each  $(s,a)$, the visitation count takes the values
$0,1,\ldots,N_h^{K+1}(s,a)-1$ over its successive visits. Thus, it follows from Lemma~\ref{lemma:opt_model_free} (see \eqref{eq:beta-t-upper-bound-ucbq}) that
\begin{align}
\sum_{k=1}^K \beta_{t(k,h)} &=
\sum_{k=1}^K \beta_{N_h^k(s_h^k,a_h^k)}
= \sum_{s,a}\sum_{n=1}^{N_h^{K+1}(s,a)} \beta_n
\le C\sum_{s,a}\sum_{n=1}^{N_h^{K+1}(s,a)}\sqrt{\frac{H^3\iota}{n}}
\notag\\
&\overset{\mathrm{(i)}}{\le} 2C\sqrt{H^3\iota}\sum_{s,a}\sqrt{N_h^{K+1}(s,a)} \nonumber\\
&\overset{\mathrm{(ii)}}{\le}  
2C\sqrt{H^3\iota}\sqrt{\sum_{s,a}1}\sqrt{\sum_{s,a}N_h^{K+1}(s,a)}
\overset{\mathrm{(iii)}}{=} 2C\sqrt{H^3SAK\iota},
\label{eq:bonus-sum-ucbq}
\end{align}
where (i) uses the elementary inequality $\sum_{n=1}^N \frac{1}{\sqrt{n}}
\leq  2\sqrt{N}$,  (ii) follows from the Cauchy-Schwarz inequality, and (iii) relies on the fact that $\sum_{s,a}N_h^{K+1}(s,a)= K$.

\item
We next control the second term on the right-hand side of
\eqref{eq:one-step-regret-recursion-ucbq}. For each
state-action pair $(s,a)$, let
$\tau_{s,a,h}(i)$ 
denote the episode index of the $i$-th visit to $(s,a)$ at step $h$.
Thus, whenever $N_h^k(s_h^k,a_h^k)\geq 1$, we have
$k_i(k,h)=\tau_{s_h^k,a_h^k,h}(i)$. 
For brevity, define
\[
\Delta_{h+1}^{\ell}
\coloneqq 
\big(V_{h+1}^{\ell}-V_{h+1}^{\star}\big)
(s_{h+1}^{\ell}) \geq 0,
\]
where the nonnegativity is due to optimism (see Lemma~\ref{lemma:opt_model_free}). 
For any given $(s,a,h)$, the contribution made by the $i$-th visit
to $(s,a,h)$ appears in every later episode $k$ for which
$(s_h^k,a_h^k)=(s,a)$; in particular, if the $i$-th visit occurs in episode
$\tau_{s,a,h}(i)$, then its coefficient in a later visit in episode $k$ is
$\eta_i^{N_h^k(s,a)}$. 
Therefore, regrouping the summation according to the episode in which each
term is generated, we obtain
\begin{align}
&\sum_{k=1}^K\sum_{i=1}^{t(k,h)}
\eta_i^{t(k,h)}
\big(V_{h+1}^{k_i(k,h)}-V_{h+1}^{\star}\big)
(s_{h+1}^{k_i(k,h)}) \nonumber\\
&\quad =
\sum_{s,a}\sum_{i=1}^{N_h^{K+1}(s,a)}
\Delta_{h+1}^{\tau_{s,a,h}(i)}
\sum_{k:\,(s_h^k,a_h^k)\,=\,(s,a), \,
N_h^k(s,a)\,\geq\, i}
\eta_i^{N_h^k(s,a)} \nonumber\\
&\quad \leq
\left(1+\frac{1}{H}\right)
\sum_{s,a}\sum_{i=1}^{N_h^{K+1}(s,a)}
\Delta_{h+1}^{\tau_{s,a,h}(i)},
\label{eq:regroup-phi-ucbq-12}
\end{align}
where the last inequality is valid since $\Delta_{h+1}^{\ell}\geq 0$, and according to \eqref{eq:properties-learning-rates-tail}, 
\[
\sum_{k:\,(s_h^k,a_h^k)\,=\,(s,a), \,
N_h^k(s,a)\,\geq\, i}
\eta_i^{N_h^k(s,a)}
\leq
\sum_{N=i}^{\infty}\eta_i^N
=1+\frac{1}{H}. 
\]
Indeed, this is a key benefit of the learning rate schedule \eqref{eq:learning_rate_ucbq}: although an estimation error may
propagate through many future updates, its
total effective weight remains well controlled.

Further, observe that 
$\big\{(s,a,i):1\leq i\leq N_h^{K+1}(s,a)\big\}$ 
are in one-to-one correspondence with the $K$ visits made at step $h$, and therefore, 
\[
\sum_{s,a}\sum_{i=1}^{N_h^{K+1}(s,a)}
\Delta_{h+1}^{\tau_{s,a,h}(i)}
=
\sum_{k=1}^K
\big(V_{h+1}^{k}-V_{h+1}^{\star}\big)(s_{h+1}^{k}).
\]
Combine it with \eqref{eq:regroup-phi-ucbq-12} to reach
\begin{align}
\sum_{k=1}^K\sum_{i=1}^{t(k,h)} \eta_i^{t(k,h)}
\big(V_{h+1}^{k_i}-V_{h+1}^{\star}\big)(s_{h+1}^{k_i}) 
&\le \left(1+\frac{1}{H}\right)\sum_{k=1}^K \big(V_{h+1}^{k}-V_{h+1}^{\star}\big)(s_{h+1}^{k}) .
\label{eq:regroup-phi-ucbq}
\end{align}

\end{itemize}

\paragraph{Step 5: recursion across horizon.}
We now turn the bounds from Step 4 into a recursion over the horizon. Substituting~\eqref{eq:first-visit-term-ucbq}, \eqref{eq:bonus-sum-ucbq} and \eqref{eq:regroup-phi-ucbq} into \eqref{eq:one-step-regret-recursion-ucbq}, 
we reach
\begin{align}
\arxiv{&}\sum_{k=1}^K \big(V_h^k-V_h^{\pi^k}\big)(s_h^k) \arxiv{\nonumber\\}
&\le SAH+
\left(1+\frac{1}{H}\right)\sum_{k=1}^K\big(V_{h+1}^{k}-V_{h+1}^{\star}\big)(s_{h+1}^{k}) + 2C\sqrt{H^3SAK\iota}\nonumber\\
&\quad +\sum_{k=1}^K\big(V_{h+1}^{\star}-V_{h+1}^{\pi^k}\big)(s_{h+1}^k) \arxiv{\nonumber\\
&\quad }+\sum_{k=1}^K (P_{s_h^k,a_h^k,h}-P_h^k)\big(V_{h+1}^{\star}-V_{h+1}^{\pi^k}\big) \nonumber\\
&\le SAH+
\left(1+\frac{1}{H}\right)\sum_{k=1}^K\big(V_{h+1}^k-V_{h+1}^{\pi^k}\big)(s_{h+1}^k)
+ 2C\sqrt{H^3SAK\iota} \nonumber\\
&\quad +\sum_{k=1}^K (P_{s_h^k,a_h^k,h}-P_h^k)\big(V_{h+1}^{\star}-V_{h+1}^{\pi^k}\big),
\label{eq:recursive-regret-ucbq}
\end{align}
where we have used the fact that $ V^{\star}_{h+1}\geq V_{h+1}^{\pi}$. 
Iterating this inequality from $h=1$ to $h=H$ propagates the
next-step error all the way to the terminal step:  
\begin{align}
\arxiv{&} \sum_{k=1}^K\big(V_1^k-V_1^{\pi^k}\big)(s_1^k) \arxiv{ \nonumber\\}
&\le O\bigg (H^2SA+\sqrt{H^5SAK\iota} +\sum_{h=1}^H\sum_{k=1}^K
(P_{s_h^k,a_h^k,h}-P_h^k)\big(V_{h+1}^{\star}-V_{h+1}^{\pi^k}\big) \bigg),
\label{eq:regret-before-bonus-martingale-ucbq}
\end{align}
where the recursion terminates at step $H+1$ since $V_{H+1}^k=V_{H+1}^{\pi^k}=0$, and we have used the elementary inequality $(1+1/H)^H\le e$ to bound the accumulated multiplicative factor. 

It remains to bound the last term (i.e., the martingale term) in \eqref{eq:regret-before-bonus-martingale-ucbq}. 
Given that $\|V_{h+1}^{\star}-V_{h+1}^{\pi^k}\|_{\infty}\leq H$, the Azuma-Hoeffding inequality (Lemma~\ref{lem:azuma-hoeffding}) gives, with probability at least $1-\delta$,
\begin{equation}
    \left|\sum_{h=1}^H\sum_{k=1}^K 
    (P_{s_h^k,a_h^k,h}-P_h^k)\big(V_{h+1}^{\star}-V_{h+1}^{\pi^k}\big)\right|
    \le O( H\sqrt{HK\iota}).
    \label{eq:martingale-sum-ucbq}
\end{equation}
Substitution into~\eqref{eq:regret-before-bonus-martingale-ucbq} gives
\begin{equation}
\mathsf{Regret}(K)
\le O\left(\sqrt{H^5SAK\iota}+ H^2SA\right).
\label{eq:regret-K-ucbq-13579}
\end{equation}

\paragraph{Step 6: removing the lower-order term.}
The preceding argument yields the intermediate bound
\eqref{eq:regret-K-ucbq-13579}, which contains the additional term
$H^2SA$. We now show that this term can be absorbed into the leading
term.   
Since the regret per
episode is at most $H$, we always have  $\mathsf{Regret}(K)\leq HK$. If $K\le H^3SA\iota$, then 
\[
    \mathsf{Regret}(K)\leq HK \leq \sqrt{H^5SAK\iota}, 
\]
so the desired bound follows directly from the trivial regret bound. If, on the other hand, $K>H^3SA\iota$, then it follows from \eqref{eq:regret-K-ucbq-13579} that
\begin{equation}
\mathsf{Regret}(K)\le O\left(\sqrt{H^5SAK\iota}+H^2SA\right) = \widetilde{O}\big(\sqrt{SAH^5K}\big).
\end{equation}
Combining these two cases establishes
Theorem~\ref{thm-ucbq-regret}.

\section{Minimax lower bounds}
\label{sec:minimax-lower-bound-online}

Having established intriguing regret upper bounds for online
RL algorithms --- particularly for the model-based approach --- we now turn to the complementary question of whether these
guarantees are statistically optimal. More precisely, we seek to
understand the smallest regret that \emph{any} online RL algorithm can achieve
in the worst case over a suitable class of MDPs. Such minimax lower bounds provide fundamental performance benchmarks, allowing us to assess the statistical optimality of existing algorithms and determine whether the regret guarantees established in the previous sections can be further improved.

A line of research has been devoted to this question
(e.g., \citet{jaksch2010near,bartlett2009regal,osband2016lower,jin2018q,domingues2021episodic}),
building in part upon  techniques originally developed
for multi-armed bandits.  In what follows, we present a minimax regret lower bound, originally established in \citet{jin2018q,domingues2021episodic}, which holds uniformly over all values of $K$.

\begin{theorem}
\label{thm:minimax-lower-online}
Consider any $S \ge 2$, $A\geq 3$, $H\geq 2$, and $K\geq 1$. 
There exists a collection $\mathcal{C}_{\mathsf{mdp}}$ of finite-horizon nonstationary MDPs --- each with $S$ states, at most $A$ actions per state, and horizon length $H$ --- together with an initial state distribution $\rho\in \Delta(\cS)$, such that
\begin{align*}
\inf_{\widehat{\pi}}\sup_{\mathcal{M}\,\in\, \mathcal{C}_{\mathsf{mdp}}} \mathop{\mathbb{E}}\nolimits_{\mathcal{M}}\Bigg[\sum_{k=1}^K\big(V_1^{\star}(\rho) - V_1^{\pi_k}(\rho)\big)\Bigg]
\ge c_0\min\big\{\sqrt{H^3SAK},HK \big\}
\end{align*}
holds for some universal constant $c_0 > 0$, where the expectation is taken w.r.t.~the randomness under the MDP $\mathcal{M}$, while the infimum is over all online learning algorithms $\widehat{\pi}$. Here, $\widehat{\pi}$ generates a sequence of policies $\widehat{\pi}=\{\pi^k\}_{k=1}^K$, and each $\pi^k$ may depend arbitrarily on the trajectories collected before episode $k$ (but not on any observations from episode $k$ or later episodes).  
\end{theorem}
Since a regret of at most $HK$ is trivially achievable --- because the cumulative reward collected over $K$ episodes cannot exceed $HK$ --- we shall often drop the linear term and simply write
\begin{align}
%	\text{(minimax lower bound)} \qquad 
	\Omega\big(\sqrt{SAH^3K} \big) \qquad \text{if }K \geq SAH. 
%	\label{eq:minimax-lower-bound-2}
\end{align}
Importantly, when combined with the regret upper bound for the \mvp algorithm in Theorem~\ref{thm-mvp-regret}, this lower bound rigorously confirms the minimax optimality of model-based online RL (modulo some logarithmic factors).

Although the lower bound itself was originally established in
\citet{jin2018q,domingues2021episodic}, the proof presented here follows a
different route. Rather than adapting the existing arguments in \citet{jin2018q,domingues2021episodic}, we
develop the proof from first principles using information-theoretic
techniques. The argument
proceeds in two stages. We first introduce a closely related bandit problem and
use it to establish the minimax lower bound for a carefully constructed family
of two-state MDPs (Section~\ref{sec:proof-lower-bound-online-simple}). We then
lift this result to general $S$-state MDPs
(Section~\ref{sec:proof-lower-bound-online}). Besides providing an alternative
proof, this derivation makes explicit the information-theoretic connection
between multi-armed bandits and online RL.

\subsection{Analysis: lower bound for a simpler case with $S=2$}
\label{sec:proof-lower-bound-online-simple}

We begin by considering the simpler setting with
$S=2$. Besides being considerably easier to analyze, this construction
captures the essential statistical bottleneck of online RL. In
Section~\ref{sec:proof-lower-bound-online}, we will extend this
construction to general state spaces by combining multiple independent
copies of the two-state MDPs.

 To be precise, our goal in this subsection is to construct a family $\mathcal{C}_{\mathsf{mdp}}^{\mathsf{sub}}$ of MDPs --- each with $S=2$ states, at most $A$ actions per state ($A\geq 3$), and horizon length $H\geq 2$ --- as well as an initial state distribution $\rho$, such that
\begin{align}
&\inf_{\widehat{\pi}}\max_{\mathcal{M}\,\in\,\mathcal{C}_{\mathsf{mdp}}^{\mathsf{sub}}}\mathbb{E}_{\mathcal{M}}\Bigg[\sum_{k=1}^{K}\big(V_{1}^{\star}(\rho)-V_{1}^{\pi^{k}}(\rho)\big)\Bigg] 
%\notag\\
%&\qquad \geq \inf_{\widehat{\pi}} \frac{1}{|\mathcal{C}_{\mathsf{mdp}}^{\mathsf{sub}}|}\sum_{\mathcal{M}\in\mathcal{C}_{\mathsf{mdp}}^{\mathsf{sub}}}\mathbb{E}_{\mathcal{M}}\Bigg[\sum_{k=1}^{K}\big(V_{1}^{\star}(1)-V_{1}^{\pi^{k}}(1)\big)\Bigg] 
%\notag\\
%&\qquad
\geq c_{\mathsf{lb}}\min\big\{\sqrt{H^{3}AK},HK\big\}
\label{eq:regret-lower-bound-S2-online-proof}
\end{align}
holds for some universal constant $c_{\mathsf{lb}}>0$.

\paragraph{Construction of the MDPs.}
As before, we begin by constructing a collection $\mathcal{C}_{\mathsf{mdp}}^{\mathsf{sub}}$ of finite-horizon nonstationary MDPs. Each MDP $\mathcal{M}$ in this collection is specified by the following components.  
%
%$$\mathcal{M} = \big(\mathcal{S}, \mathcal{A}, H, \{P_h\}_{h = 1}^H, \{r_h\}_{h = 1}^{H}\big),$$  

\begin{itemize}
\item {\em State space.} The state space consists of 2 states:
$\mathcal{S} = \{0, 1\}.$

\item {\em Action space.} State 0 and state 1 are associated respectively with the following action space:
$$\mathcal{A}_0 = \{0\}
\qquad \text{and}\qquad
\mathcal{A}_1 = \{1, \ldots, A\}.$$

\item {\em Reward function.} The reward function $r=\{r_h\}_{1\leq h\leq H}$ is chosen such that, for each $1\leq h\leq H$,
\begin{align}
r_h(0, 0) = 0
\qquad\text{and}\qquad
r_h(1, a) = 1 ~\text{ for every }a\in \cA_1.
\label{eq:reward-lower-bound-online}
\end{align}

\item {\em Optimal actions and probability transition kernel.} 
Since the action space $\mathcal{A}_0$ associated with state 0 is a singleton, it suffices to specify the optimal action at state $1$ for each step $1 \leq h \leq H$. For an MDP $\mathcal{M}\in \mathcal{C}_{\mathsf{mdp}}^{\mathsf{sub}}$, we denote these optimal actions by
\begin{align}
a_1^{\star,\mathcal{M}}, \ldots, a_H^{\star,\mathcal{M}}\in \mathcal{A}_1
\label{eq:optimal-a-M-lower-bound-online}
\end{align}
with $a_h^{\star,\mathcal{M}}$ denoting the optimal action at state $1$ and step $h$. The probability transition kernel of $\mathcal{M}$, denoted by $P^{\mathcal{M}}$, is then defined by 
\begin{subequations}
\label{eq:transition-kernel-M-lower-bound-online}
\begin{align}
&P_h^{\mathcal{M}}(0 \mymid 0, 0) = 1, \\
&P_h^{\mathcal{M}}(0 \mymid 1, a) = q_{a_h^{\star,\mathcal{M}}}(a),\quad P_h^{\mathcal{M}}(1 \mymid 1, a) = 1-q_{a_h^{\star,\mathcal{M}}}(a)
\end{align}
\end{subequations}
for every $h\in [H]$ and $a\in \mathcal{A}_1$.  
By construction, state $0$ is absorbing and yields zero reward. Thus, the behavior of the MDP is largely determined by the dynamics at state 1. We will  specify the function $q_{a_h^{\star,\mathcal{M}}}(\cdot)$ next and verify that the resulting transition kernel is indeed consistent with the designated optimal actions \eqref{eq:optimal-a-M-lower-bound-online}. 

\item {\em Initial state distribution.}
In what follows, we take the initial state distribution $\rho$ to be concentrated on state $1$, namely, 
\begin{equation}
\rho(s) = \begin{cases}1, &\text{if }s=1; \\
0, &\text{else}.\end{cases}
\label{eq:rho-s-initial-dist-lower-bound-online}
\end{equation}

\end{itemize}

\noindent 
It remains to specify the functions $q_{a_h^{\star,\mathcal{M}}}(\cdot)$ appearing in \eqref{eq:transition-kernel-M-lower-bound-online}. For notational simplicity, consider an arbitrary sequence of optimal actions  $a_1^{\star},\ldots,a_H^{\star}$. For each $h\in [H]$, we define
\begin{subequations}
\label{eqn:prob-astar-a-1}    
\begin{align}
 \text{when }a_h^{\star}\neq 1: \qquad 
  &q_{a_h^{\star}}(a) =
  \begin{cases}
  \frac{1}{H}-\frac{\varepsilon_0}{H^2} & \text{if } a = a_h^{\star},\\
   \frac{1}{H}-\frac{\varepsilon_0}{2H^2} & \text{if } a = 1,\\
  \frac{1}{H}, & \text{else},
  \end{cases}
  \\
 \text{when }a_h^{\star}= 1:\qquad  
  & q_{a_h^{\star}}(a) =
  \begin{cases}
  \frac{1}{H}-\frac{\varepsilon_0}{2H^2} & \text{if } a = a_h^{\star}=1,\\
  \frac{1}{H}, & \text{else},
  \end{cases}
\end{align}
where the parameter $0<\varepsilon_0\leq H/2$ will be specified later (see \eqref{eq:epsilon0-lower-bound-online}). 
%
%\begin{align}
%\varepsilon_0 = \min\Bigg\{ \sqrt{\frac{c_0H^3A}{K}} ,H\Bigg\}
%\label{eq:epsilon0-lower-bound-online}
%\end{align}
%
%for some constant $c_0 > 0$ to be specified momentarily. 
%
\end{subequations}
The consistency of the resulting transition kernel with the designated optimal actions is immediate. Indeed, since state $0$ is absorbing and yields zero reward, maximizing the value function is equivalent to minimizing $q_{a_h^{\star}}(a)$ over $a\in\mathcal{A}_1$. By \eqref{eqn:prob-astar-a-1}, this minimum is attained uniquely at $a=a_h^{\star}$ for every step $h$.

We now specify the collection $\mathcal{C}_{\mathsf{mdp}}^{\mathsf{sub}}$, where we allow the optimal actions
to vary independently across steps. Specifically, for each
\[
\bm{i}=(i_1,\ldots,i_H)\in[A]^H,
\]
let $\mathcal{M}_{\bm{i}}$ denote the MDP constructed above with
\begin{equation}
	a_h^{\star,\mathcal{M}_{\bm{i}}}=i_h,
	\qquad 1\leq h\leq H.
	\label{eq:optimal-action-Mi-lower-bound-online}
\end{equation}
We then define
\begin{equation}
	\mathcal{C}_{\mathsf{mdp}}^{\mathsf{sub}}
	\coloneqq
	\big\{
	\mathcal{M}_{\bm{i}}:\bm{i}\in[A]^H
	\big\}
    \qquad \Longrightarrow \qquad |\mathcal{C}_{\mathsf{mdp}}^{\mathsf{sub}}|=A^H. 
	\label{eq:collection-Cmdp-lower-bound-online-simple}
\end{equation}

\paragraph{Proof strategy.}
The proof of the lower bound for this family proceeds in two main steps.
First, we exploit a key feature of the above construction to relate the
suboptimality gap of an arbitrary policy to its probability of selecting
suboptimal actions, thereby reducing the regret analysis to controlling
the probability of such action-selection errors. The following lemma
formalizes this connection; its proof is deferred to
Section~\ref{sec:proof-lem:Vh-star-Vhpi-gap-lower-bound-online-final}.
\begin{lemma}
\label{lem:Vh-star-Vhpi-gap-lower-bound-online-final}
Suppose that $H\geq 2$.
For every $\mathcal{M}\in\mathcal{C}_{\mathsf{mdp}}^{\mathsf{sub}}$
and every policy $\pi$, one has
\begin{align}
V_{1}^{\star}(1)-V_{1}^{\pi}(1)
&\geq
\sum_{h=1}^{H}
\frac{\varepsilon_{0}(H-h)}{16H^{2}}
\big(1-\pi_{h}(a_{h}^{\star,\mathcal{M}}\mymid 1)\big).
\label{eq:Vh-star-Vhpi-gap-lower-bound-online-final}
\end{align}
\end{lemma}
Second, we develop a bandit statistical testing argument to show that the probability
of selecting a suboptimal action remains bounded away from zero until
sufficiently many observations are collected.

\paragraph{A detour: a lower bound on a bandit problem.} 
To control the right-hand side of \eqref{eq:Vh-star-Vhpi-gap-lower-bound-online-final}, we take a brief detour to establish a lower bound in a closely related bandit setting. The formulation is the following. 
\begin{itemize}
    \item Consider a collection $\mathcal{C}_{\mathsf{bandit}}$ consisting of $A$ bandit instances, where each instance has $A$ arms. 
    
    \item For $i\in[A]$, let $\mathcal{B}_i$ denote the bandit instance whose unique optimal arm is $a^{\star}_{\mathcal{B}_i}=i$.  The mean reward of arm $a$ in  $\mathcal{B}_i$ is given by 
    \[
        r_{\mathcal{B}_i}(a) = 1- q_i(a),
    \]
    where $q_i(\cdot)$ is defined in \eqref{eqn:prob-astar-a-1}. More concretely,  pulling arm $a$ in round $t$ in the bandit instance $\mathcal{B}_i$ yields an independent Bernoulli reward as
    \[
        R_{a,t} = \begin{cases}
        1, & \text{with probability }1-q_i(a),\\
        0, &\text{else}. 
        \end{cases}
    \]

\end{itemize}
\noindent 
In what follows, let $\mathbb{P}_{\mathcal{B}_i}$ denote the probability distribution under the bandit instance $\mathcal{B}_i$. As before, the learner interacts with the bandit instance in a sequential manner; at each round $t$, the learner selects an arm based only on the observations collected prior to round $t$.

We now apply the lower bound in Lemma~\ref{lem:reverse-Fano-bound} to 
develop a round-wise lower bound for this collection of bandit instances. The proof is deferred to Section~\ref{sec:proof:lem:online-lb-testing}. 
\begin{lemma}
\label{lem:online-lb-testing}
Consider the collection $\mathcal{C}_{\mathsf{bandit}}$ of bandit instances defined above.  Suppose that $A\geq 3$, $H\geq2$, and
$0<\varepsilon_0\leq H/2$. 
Then for any round $t \le \frac{H^3A}{18 \varepsilon_0^2}$, one has
\begin{align*}
\inf_{\widehat{\pi}}\frac{1}{A} \sum_{i = 1}^A \mathbb{P}_{\mathcal{B}_i}\big(a_{t+1} \ne i \big)
\ge \frac{1}{3},
\end{align*}
where the infimum is taken over all online bandit algorithms $\widehat{\pi}$. Here, for a given algorithm $\widehat{\pi}$, we denote by $\{\pi^t\}_{t\ge 1}$ the resulting sequence of policies, where $\pi^{t}$ depends only on observations collected prior to round $t$; the arm $a_{t}$ is then selected according to $\pi^{t}$. 
\end{lemma}
In words, the lemma states that, at any round $t+1$, the probability of selecting a suboptimal action --- based on the observations collected through round $t$ --- is bounded away from zero. Importantly, this is a per-round guarantee that holds whenever $t$ is not too large. 
As will become clear shortly, it is a little more convenient to develop this result in terms of the average probability of error rather than the minimax probability of error.

\begin{remark}
The proof argument used to establish Lemma~\ref{lem:online-lb-testing} can also be adapted to establish the minimax optimality of the \UCB algorithm for the Gaussian stochastic bandit setting in Section~\ref{sec:bandits-UCB}, with the Bernoulli KL divergence replaced by its Gaussian counterpart.
\end{remark}

\paragraph{Regret analysis.} 
We now return to the two-state MDP construction and leverage the preceding bandit lower bound in Lemma~\ref{lem:online-lb-testing}. The key observation is that the information available for identifying the optimal action in the MDP has precisely the bandit structure studied above: under $\mathcal{M}$, whenever the learner takes action $a$ at state $1$, the resulting transition corresponds to a Bernoulli observation with mean $1-q_{a_h^{\star,\mathcal{M}}}(a)$, matching the reward distribution obtained by pulling arm $a$ in $\mathcal{B}_i$ if $a_h^{\star,\mathcal{M}}=i$. Hence, when choosing an action at any given step and episode, the learner faces the same statistical problem of identifying the unknown optimal action from sequential bandit feedback. Lemma~\ref{lem:online-lb-testing} can therefore be used to lower bound the probability of selecting a suboptimal action in the MDP.

Before proceeding, we first introduce, for convenience of analysis, an auxiliary action variable in order to distinguish the action selected by the policy in state $1$ from the action actually executed along the realized trajectory. For every episode $k\in[K]$ and step $h\in[H]$, let 
\begin{equation} \widetilde a_h^k \sim \pi_h^k(\cdot\mymid 1) \label{eq:counterfactual-action-online-lb} \end{equation} 
conditional on what happens before episode $k$, regardless of whether the realized state $s_h^k$ equals $1$. Thus, $\widetilde a_h^k$ is an auxiliary action used only in the
analysis. By construction,
\begin{align}
\mathbb{E}_{\mathcal{M}}
\left[
1-\pi_h^k(a_h^{\star,\mathcal{M}}\mymid1)
\right]
=
\mathbb{P}_{\mathcal{M}}
\left(
\widetilde a_h^k\neq a_h^{\star,\mathcal{M}}
\right).
\label{eq:counterfactual-action-error-online-lb}
\end{align}
Conditional on the optimal actions at steps other than $h$, reaching
state $1$ at step $h$ is independent of
$a_h^{\star,\mathcal{M}}$, and the trajectory following the step-$h$
transition provides no further information about it. Treating episodes
that do not reach state $1$ as dummy bandit rounds, the learner can
therefore be viewed as a bandit algorithm run for $k-1$ rounds before
episode $k$.

Now consider any online learning algorithm
$\widehat{\pi}$. In view of
\eqref{eq:Vh-star-Vhpi-gap-lower-bound-online-final} and
\eqref{eq:counterfactual-action-error-online-lb},  one has
\begin{align}
&\mathbb{E}_{\mathcal{M}}\Bigg[
\sum_{k=1}^{K}
\big(
V_1^\star(1)-V_1^{\pi^k}(1)
\big)
\Bigg]
\arxiv{\notag\\
&\qquad}\geq
\sum_{k=1}^{K}\sum_{h=1}^{H}
\frac{\varepsilon_0(H-h)}{16H^2}
\mathbb{P}_{\mathcal{M}}
\left(
\widetilde a_h^k
\neq
a_h^{\star,\mathcal{M}}
\right)
\label{eq:regret-counterfactual-action-online-lb}
\end{align}
for every
$\mathcal{M}\in\mathcal{C}_{\mathsf{mdp}}^{\mathsf{sub}}$. 
It follows that
\begin{align}
\arxiv{&}\inf_{\widehat{\pi}}
\max_{\bm{i}\in[A]^H}
\mathbb{E}_{\mathcal{M}_{\bm{i}}}
\Bigg[
\sum_{k=1}^{K}
\big(
V_1^\star(1)-V_1^{\pi^k}(1)
\big)
\Bigg]
\arxiv{ \notag\\ }
&\geq
\inf_{\widehat{\pi}}
\frac{1}{A^H}
\sum_{\bm{i}\in[A]^H}
\mathbb{E}_{\mathcal{M}_{\bm{i}}}
\Bigg[
\sum_{k=1}^{K}
\big(
V_1^\star(1)-V_1^{\pi^k}(1)
\big)
\Bigg]
\notag\\
&\geq
\sum_{k=1}^{K}
\sum_{h=1}^{H}
\frac{\varepsilon_0(H-h)}{16H^2}
\inf_{\widehat{\pi}}
\frac{1}{A^H}
\sum_{\bm{i}\in[A]^H}
\mathbb{P}_{\mathcal{M}_{\bm{i}}}
\left(
\widetilde a_h^k\neq i_h
\right).
\label{eq:two-state-product-prior-bound}
\end{align}

Next, we seek to lower bound \eqref{eq:two-state-product-prior-bound}. For each $h\in[H]$, write
\[
\bm{i}_{-h}
\coloneqq
(i_1,\ldots,i_{h-1},i_{h+1},\ldots,i_H).
\]
Fix any $\bm{i}_{-h}\in[A]^{H-1}$. Conditional on $\bm{i}_{-h}$, the
preceding construction induces a valid bandit algorithm for identifying
$i_h$ from at most $k-1$ informative observations. Therefore,
applying Lemma~\ref{lem:online-lb-testing} with $t=k-1$ gives
\begin{align}
\frac{1}{A}
\sum_{i_h=1}^{A}
\mathbb{P}_{\mathcal{M}_{(i_h,\bm{i}_{-h})}}
\left(
\widetilde a_h^k\neq i_h
\right)
\geq
\frac{1}{3},
\label{eq:conditional-coordinate-testing-online-lb}
\end{align}
provided that
$
k-1
\leq
\frac{AH^3}{18\varepsilon_0^2}$. 
Averaging \eqref{eq:conditional-coordinate-testing-online-lb} over
$\bm{i}_{-h}\in[A]^{H-1}$ and then taking the infimum over
$\widehat{\pi}$ yields
\begin{align}
\inf_{\widehat{\pi}}
\frac{1}{A^H}
\sum_{\bm{i}\in[A]^H}
\mathbb{P}_{\mathcal{M}_{\bm{i}}}
\left(
\widetilde a_h^k\neq i_h
\right)
\geq
\frac{1}{3}.
\label{eq:coordinatewise-testing-lower-bound-online}
\end{align}
Substituting \eqref{eq:coordinatewise-testing-lower-bound-online} into
\eqref{eq:two-state-product-prior-bound} reveals that: if $
K-1
\leq
\frac{AH^3}{18\varepsilon_0^2}
$, then
\begin{align}
&\inf_{\widehat{\pi}}
\max_{\bm{i}\in[A]^H}
\mathbb{E}_{\mathcal{M}_{\bm{i}}}
\Bigg[
\sum_{k=1}^{K}
\big(
V_1^\star(1)-V_1^{\pi^k}(1)
\big)
\Bigg] \arxiv{
\notag\\
&\quad} \geq
\frac{1}{3}
\sum_{k=1}^{K}
\sum_{h=1}^{H}
\frac{\varepsilon_0(H-h)}{16H^2}=
\frac{\varepsilon_0K(H-1)}{96H}
\geq
\frac{\varepsilon_0K}{192}.
\label{eq:regret-minimax-lower-bound-proof-simple}
\end{align}

To finish up,  choosing 
\begin{align}
\varepsilon_0 = \min\Bigg\{ \sqrt{\frac{H^3A}{18K}} , \frac{H}{2}\Bigg\},
\label{eq:epsilon0-lower-bound-online}
\end{align}
clearly satisfies $K-1\le \frac{AH^3}{18\varepsilon_0^2}$ and $\varepsilon_0\leq H/2$. Substituting this choice into \eqref{eq:regret-minimax-lower-bound-proof-simple}, we arrive at
\begin{align}
\arxiv{&}\inf_{\widehat{\pi}}\max_{\mathcal{M}\,\in\,\mathcal{C}_{\mathsf{mdp}}^{\mathsf{sub}}}\mathbb{E}_{\mathcal{M}}\Bigg[\sum_{k=1}^{K}\big(V_{1}^{\star}(1)-V_{1}^{\pi^{k}}(1)\big)\Bigg] 
\arxiv{\notag\\ }
& \arxiv{\qquad} \geq \inf_{\widehat{\pi}} \frac{1}{|\mathcal{C}_{\mathsf{mdp}}^{\mathsf{sub}}|}\sum_{\mathcal{M}\,\in\,\mathcal{C}_{\mathsf{mdp}}^{\mathsf{sub}}}\mathbb{E}_{\mathcal{M}}\Bigg[\sum_{k=1}^{K}\big(V_{1}^{\star}(1)-V_{1}^{\pi^{k}}(1)\big)\Bigg] 
\notag\\
&\arxiv{\qquad} \geq\frac{\varepsilon_{0}K}{192}\asymp\min\big\{\sqrt{H^{3}AK},HK\big\}.
\label{eq:regret-lower-bound-S2-online-proof-135}
\end{align}
This taken together with the choice of $\rho$ (cf.~\eqref{eq:rho-s-initial-dist-lower-bound-online}) establishes the advertised lower bound \eqref{eq:regret-lower-bound-S2-online-proof} for the case with $S=2$.

\subsection{Analysis: lower bound for general $S$ (proof of Theorem~\ref{thm:minimax-lower-online})}
\label{sec:proof-lower-bound-online}

We now demonstrate how the regret lower bound established in
Section~\ref{sec:proof-lower-bound-online-simple} for $S=2$ can be
lifted to general $S\geq 2$. The key idea is to embed $S/2$ independent
copies of the two-state hard instances into a single MDP. Since each
episode starts in exactly one of these copies, the $K$ trajectories
are effectively distributed across $S/2$ independent learning
problems, allowing us to apply the two-state lower bound separately
to each sub-MDP. Throughout the proof, we assume, for simplicity, that
$S$ is even.

\paragraph{Construction of the MDPs and notation.}
We begin by constructing a family $\mathcal{C}_{\mathsf{mdp}}$ of MDPs. 
\begin{itemize}
    \item Each MDP $\mathcal{M}$ in this family is composed of $S/2$ disjoint sub-MDPs, each containing 2 states.

    \item For each $1\leq j\leq S/2$, the $j$-th sub-MDP of $\mathcal{M}$, denoted by $\mathcal{M}^{\mathsf{sub},j}$, has state space $\{j,j+S/2\}$ and is a relabeled copy of an MDP in $\mathcal{C}_{\mathsf{mdp}}^{\mathsf{sub}}$ introduced in Section~\ref{sec:proof-lower-bound-online-simple}, with states $1$ and $0$ therein corresponding to $j$ and $j+S/2$, respectively.

   \item More precisely, we define $\mathcal{C}_{\mathsf{mdp}}$ to
    consist of all MDPs obtained by choosing each of the $S/2$
    sub-MDPs independently from
    $\mathcal{C}_{\mathsf{mdp}}^{\mathsf{sub}}$. Equivalently, up to the state relabeling described above, 
    \begin{equation}
        \mathcal{C}_{\mathsf{mdp}}
        \coloneqq
        \big(
        \mathcal{C}_{\mathsf{mdp}}^{\mathsf{sub}}
        \big)^{S/2}.
        \label{eq:defn-C-mdp-general-S}
    \end{equation}
    Thus, each
    $\mathcal{M}\in\mathcal{C}_{\mathsf{mdp}}$ can be identified with
    a tuple
    \begin{align}
        \mathcal{M}
        =
        \big(
        \mathcal{M}^{\mathsf{sub},1},
        \ldots,
        \mathcal{M}^{\mathsf{sub},S/2}
        \big),
        \qquad
        \mathcal{M}^{\mathsf{sub},j}
        \in\mathcal{C}_{\mathsf{mdp}}^{\mathsf{sub}}.
        \label{eq:M-notation-Msub-j}
    \end{align}
    %
   % and, in particular, $|\mathcal{C}_{\mathsf{mdp}}| =|\mathcal{C}_{\mathsf{mdp}}^{\mathsf{sub}}|^{S/2} =A^{S/2}$.
    
   \item The initial state distribution $\rho\in \Delta(\cS)$ is taken to be
    \begin{align}
    \label{eq:defn-rho-general-S-lower-bound-pf-online}
        \rho(s) = \begin{cases}2/S, & \text{if }1\leq s\leq S/2, \\
        0, & \text{else}.\end{cases}
    \end{align}
    
\end{itemize}

\paragraph{Decomposition across the sub-MDPs.}
Let
\[
J_k\coloneqq s_1^k\in[S/2],
\qquad
\bm{J}\coloneqq(J_1,\ldots,J_K),
\qquad
K_j\coloneqq\sum_{k=1}^K\mathds{1}\{J_k=j\}.
\]
Thus, $J_k$ identifies the sub-MDP visited in episode $k$, and $K_j$
is the number of episodes assigned to the $j$-th sub-MDP. Since the
initial states are drawn independently from $\rho$, the random vector
$\bm{J}$ is independent of the identities of all sub-MDPs.

We now place the uniform product prior on
$\mathcal{C}_{\mathsf{mdp}}$; namely, the sub-MDPs
$\{\mathcal{M}^{\mathsf{sub},j}\}_{1\leq j\leq S/2}$ are drawn
independently and uniformly from
$\mathcal{C}_{\mathsf{mdp}}^{\mathsf{sub}}$. Fix an initial-state
sequence $\bm{J}$ and a component $j$. Under this product prior, the
behavior of $\widehat{\pi}$ during the $K_j$ episodes assigned to
component $j$ induces a valid randomized $K_j$-episode algorithm for
the two-state problem. Indeed, the interactions with all other
components can be incorporated into the internal randomization of this
induced algorithm: their identities are drawn independently from the
product prior, and their trajectories can be simulated between
successive visits to component $j$. Although these simulated
interactions may depend on the observations previously collected from
component $j$, they contain no additional information about the
identity of $\mathcal{M}^{\mathsf{sub},j}$.

Consequently, the two-state lower bound
\eqref{eq:regret-lower-bound-S2-online-proof-135} applies conditionally
on $\bm{J}$. In particular, for every $j\in[S/2]$,
\begin{align}
&\frac{1}{|\mathcal{C}_{\mathsf{mdp}}|}
\sum_{\mathcal{M}\,\in\,\mathcal{C}_{\mathsf{mdp}}}
\mathbb{E}_{\mathcal{M}}
\Bigg[
\sum_{k:J_k=j}
\big(
V_1^\star(j)-V_1^{\pi^k}(j)
\big)
\,\Bigm|\,\bm{J}
\Bigg]
\arxiv{\notag\\
&\qquad}\geq
\frac{\varepsilon_0K_j}{192}
\mathds{1}\left\{
K_j\leq
\frac{AH^3}{18\varepsilon_0^2}
\right\}.
\label{eq:conditional-subMDP-regret-lower-bound}
\end{align}
Moreover, since $J_k\sim\rho$ independently of the policy $\pi^k$
selected before episode $k$, for every
$\mathcal{M}\in\mathcal{C}_{\mathsf{mdp}}$ one has
\begin{align}
\mathbb{E}_{\mathcal{M}}
\left[
V_1^\star(J_k)-V_1^{\pi^k}(J_k)
\right]
=
\mathbb{E}_{\mathcal{M}}
\left[
V_1^\star(\rho)-V_1^{\pi^k}(\rho)
\right].
\label{eq:expected-realized-initial-state-regret}
\end{align}
Summing \eqref{eq:conditional-subMDP-regret-lower-bound} over the
components and averaging over $\bm{J}$ yield
\begin{align}
\arxiv{&} \inf_{\widehat{\pi}}
\max_{\mathcal{M}\,\in\,\mathcal{C}_{\mathsf{mdp}}}
\mathbb{E}_{\mathcal{M}}
\Bigg[
\sum_{k=1}^K
\big(
V_1^\star(\rho)-V_1^{\pi^k}(\rho)
\big)
\Bigg]
\arxiv{\notag\\}
&\geq
\inf_{\widehat{\pi}}
\frac{1}{|\mathcal{C}_{\mathsf{mdp}}|}
\sum_{\mathcal{M}\,\in\,\mathcal{C}_{\mathsf{mdp}}}
\mathbb{E}_{\mathcal{M}}
\Bigg[
\sum_{k=1}^K
\big(
V_1^\star(J_k)-V_1^{\pi^k}(J_k)
\big)
\Bigg]
\notag\\
&\geq
\mathbb{E}_{\bm{J}}
\Bigg[
\sum_{j=1}^{S/2}
\frac{\varepsilon_0K_j}{192}
\mathds{1}\left\{
K_j\leq
\frac{AH^3}{18\varepsilon_0^2}
\right\}
\Bigg].
\label{eq:minimax-regret-decompose-sub-MDPs-online-lower-bound-000}
\end{align}

In what follows, we look at two cases separately.

\paragraph{Case 1: $K\geq  S/2$. }
We begin with the regime $K\geq S/2$. Continue the bound \eqref{eq:minimax-regret-decompose-sub-MDPs-online-lower-bound-000} to derive
\begin{align}
 & \inf_{\widehat{\pi}}\max_{\mathcal{M}\,\in\,\mathcal{C}_{\mathsf{mdp}}}\mathbb{E}_{\mathcal{M}}\Bigg[\sum_{k=1}^{K}\big(V_{1}^{\star}(\rho)-V_{1}^{\pi_{k}}(\rho)\big)\Bigg]  \notag\\
 & \geq\mathop{\mathbb{E}}\limits _{s_{1}^{1},\cdots,s_{1}^{K}\overset{\mathrm{i.i.d.}}{\sim}\rho}\Bigg[\sum_{j=1}^{S/2}\frac{\varepsilon_{0}K_{j}}{192}\Bigg]-\frac{\varepsilon_{0}}{192}\mathop{\mathbb{E}}\limits _{s_{1}^{1},\cdots,s_{1}^{K}\overset{\mathrm{i.i.d.}}{\sim}\rho}\Bigg[\sum_{j=1}^{S/2}K_{j}\mathds{1}\bigg\{ K_{j}>\frac{AH^{3}}{18\varepsilon_{0}^{2}}\bigg\}\Bigg]\nonumber \\
 & =\frac{\varepsilon_{0}K}{192}-\frac{\varepsilon_{0}}{192}\sum_{j=1}^{S/2}\mathop{\mathbb{E}}\limits _{s_{1}^{1},\cdots,s_{1}^{K}\overset{\mathrm{i.i.d.}}{\sim}\rho}\Bigg[K_{j}\mathds{1}\bigg\{ K_{j}>\frac{AH^{3}}{18\varepsilon_{0}^{2}}\bigg\}\Bigg].
 \label{eq:minimax-regret-decompose-sub-MDPs-online-lower-bound-246}
\end{align}
Since $K_j\sim\mathsf{Binomial}(K,2/S)$,  Lemma~\ref{lem:tail-bound-sum-Bernoulli} tells us that: if $\varepsilon_{0}^{2}\leq\frac{SAH^{3}}{1800K}$, then
\begin{align*}
\arxiv{&} \mathop{\mathbb{E}}\Bigg[K_{j}\mathds{1}\bigg\{ K_{j}>\frac{AH^{3}}{18\varepsilon_{0}^{2}}\bigg\}\Bigg] \book{&}  \leq\mathop{\mathbb{E}}\Bigg[K_{j}\mathds{1}\bigg\{ K_{j}>\frac{100K}{S}\bigg\}\Bigg]\\
 & \arxiv{\qquad} =\frac{100K}{S}\mathbb{P}\bigg\{ K_{j}>\frac{100K}{S}\bigg\}+\int_{100K/S}^{\infty}\mathbb{P}\big\{ K_{j}>\tau\big\}\mathrm{d}\tau\\
 & \arxiv{\qquad} \leq\frac{100K}{S}\exp\left(-\frac{19K}{S}\right)+2\int_{100K/S}^{\infty}\exp(-0.19\tau)\mathrm{d}\tau\\
 & \arxiv{\qquad} \leq\frac{111K}{S}\exp\left(-\frac{19K}{S}\right)\leq\frac{0.01K}{S},
\end{align*}
where the second line relies on the basic property $$\mathbb{E}\big[X\mathds{1}\{X> x\}\big] = x \mathbb{P}(X>x) + \int_{x}^{\infty} \mathbb{P}(X>\tau) \mathrm{d}\tau.$$ Substitution into \eqref{eq:minimax-regret-decompose-sub-MDPs-online-lower-bound-246} shows that: if $\varepsilon_{0}^{2}\leq\frac{SAH^{3}}{1800K}$ and $\varepsilon_0 \leq  H/2$, then
\begin{align*}
\inf_{\widehat{\pi}}\max_{\mathcal{M}\,\in\,\mathcal{C}_{\mathsf{mdp}}}\mathbb{E}_{\mathcal{M}}\Bigg[\sum_{k=1}^{K}\big(V_{1}^{\star}(\rho)-V_{1}^{\pi_{k}}(\rho)\big)\Bigg] & \geq\frac{\varepsilon_{0}K}{192}-\frac{\varepsilon_{0} (S/2)}{192}\cdot \frac{0.01K}{S} \arxiv{\notag\\
&}=\frac{0.995\varepsilon_{0}K}{192}.
\end{align*}

\paragraph{Case 2: $K< S/2$. } We now turn to the regime $K<S/2$, where it suffices to focus on states that are visited exactly once.

To begin with, observe that if $\varepsilon_0 \leq H/2$, then 
\[
\frac{AH^3}{18\varepsilon_0^2}
\geq \frac{2AH}{9}
\geq \frac{4}{3}>1,
\]
provided that $A\geq3$ and $H\geq2$. Hence, every $j$ obeying $K_j=1$ satisfies  $\frac{AH^3}{18\varepsilon_0^2} > K_j$. Taking this collectively with \eqref{eq:minimax-regret-decompose-sub-MDPs-online-lower-bound-000}, we see that
\begin{align}
\arxiv{ &} \inf_{\widehat{\pi}}\max_{\mathcal{M}\,\in\,\mathcal{C}_{\mathsf{mdp}}}\mathbb{E}_{\mathcal{M}}\Bigg[\sum_{k=1}^{K}\big(V_{1}^{\star}(\rho)-V_{1}^{\pi_{k}}(\rho)\big)\Bigg] \arxiv{ \nonumber \\}
& \arxiv{ \quad} \geq\mathop{\mathbb{E}}\limits _{s_{1}^{1},\cdots,s_{1}^{K}\overset{\mathrm{i.i.d.}}{\sim}\rho}\Bigg[\sum_{j=1}^{S/2}\frac{\varepsilon_{0}K_{j}}{192}\mathds{1}\bigg\{ K_{j}\leq\frac{AH^{3}}{18\varepsilon_{0}^{2}}\bigg\}\Bigg] \notag\\
 & \arxiv{\qquad} \geq  \mathop{\mathbb{E}}\limits _{s_{1}^{1},\cdots,s_{1}^{K}\overset{\mathrm{i.i.d.}}{\sim}\rho}\Bigg[\sum_{j=1}^{S/2}\frac{\varepsilon_{0}K_{j}}{192}\mathds{1}\big\{ K_{j}=1\big\}\Bigg] \notag\\
& = \frac{\varepsilon_0}{192} \sum_{j=1}^{S/2} \mathbb{P}(K_j=1).
 \label{eq:minimax-regret-decompose-sub-MDPs-online-lower-bound-case2}
\end{align}
Recognizing that $K_j\sim\mathsf{Binomial}(K,2/S)$, one can deduce that
\begin{align*}
\sum_{j=1}^{S/2}\mathbb{P}(K_j=1)
&=
\frac{S}{2}
\cdot
K\frac{2}{S}
\bigg(1-\frac{2}{S}\bigg)^{K-1}=
K\bigg(1-\frac{2}{S}\bigg)^{K-1}
\geq \frac{K}{4},
\end{align*}
where the last inequality follows from $K<S/2$. Consequently, we arrive at
\begin{equation}
\inf_{\widehat{\pi}}
\max_{\mathcal{M}\,\in\,\mathcal{C}_{\mathsf{mdp}}}
\mathbb{E}_{\mathcal{M}}
\Bigg[
\sum_{k=1}^K
\big(
V_1^\star(\rho)-V_1^{\pi^k}(\rho)
\big)
\Bigg]
\geq
\frac{\varepsilon_0K}{768}.
\label{eq:general-S-K-small-lb}
\end{equation}

\paragraph{Putting all this together.} 
To finish up, we take
\[
\varepsilon_{0}=\min\bigg\{\sqrt{\frac{SAH^{3}}{1800K}}, \frac{H}{2}\bigg\}
\]
in the above bounds (for both Case 1 and Case 2) to conclude that
\begin{align*}
\inf_{\widehat{\pi}}\max_{\mathcal{M}\,\in\,\mathcal{C}_{\mathsf{mdp}}}\mathbb{E}_{\mathcal{M}}\Bigg[\sum_{k=1}^{K}\big(V_{1}^{\star}(\rho)-V_{1}^{\pi_{k}}(\rho)\big)\Bigg] & \gtrsim\min\big\{\sqrt{SAH^{3}K},HK\big\}.
\end{align*}

\section{Appendix A: proof of auxiliary lemmas in Section~\ref{sec:proof-of-thm-ucbvi}}

\subsection{Proof of Lemma~\ref{lemma:opt}} \label{sec:proof-lemma:opt}
We begin by introducing a scalar-valued function $f$, which facilitates the subsequent analysis: 
 \begin{align}
	 f(p,v,n) \coloneqq \langle p, v\rangle + \max\Bigg\{c_1\sqrt{\frac{\mathsf{Var}_p(v)\log \frac{1}{\delta'}} {n}} , \frac{c_1^2H\log \frac{1}{\delta'}}{n} \Bigg\},
     \label{eq:monotone-function-f-pvn-ucbvi}
 \end{align}
 defined for any probability vector $p=[p_s]_{1\leq s\leq S}\in \Delta^{S}$, any nonnegative vector $v=[v_s]_{1\leq s\leq S}\in \mathbb{R}^S$ obeying $\|v\|_{\infty}\leq H $, and any positive integer $n$. A key property of this function, which we rely on in this proof, is its monotonicity with respect to $v$, namely, 
\begin{equation}
	f(p,v,n)  \text{ is non-decreasing in each entry of }v.
	\label{eq:nondecreasing-f-property}
\end{equation}
The proof of this property is postponed to the end of this subsection.

We now present the proof of  Lemma~\ref{lemma:opt} using the property \eqref{eq:nondecreasing-f-property}. 
Consider any tuple $(h,k,s,a)$, and we divide into two cases. 

\paragraph{Case 1: $N_{h}^k(s,a) \leq 2$.}  
In this situation, the update rule \eqref{eq:updateqv-ucbvi} in conjunction with the bonus \eqref{eq:bh-defn-Bernstein-ucbvi} yields the following trivial bounds:
$$
	 Q_h^k(s,a)=H \geq Q_h^{\star}(s,a) \qquad \text{and} \qquad V_h^k(s)
     \geq Q_h^k(s,a) =H\geq V_h^{\star}(s) ,
$$ 
provided that the constant $c_2$ in \eqref{eq:bh-defn-Bernstein-ucbvi}  exceeds $c_2\geq 2$. 

\paragraph{Case 2: $N_{h}^k(s,a) > 2$.}
Suppose now, by way of induction, that $Q_{h+1}^k\geq Q_{h+1}^{\star}$, which necessarily implies that $V_{h+1}^k \geq V_{h+1}^{\star}$. 
If $Q_h^k(s,a)=H$, then $Q_h^k(s,a)\geq Q_h^{\star}(s,a)$ holds trivially. Therefore, it suffices to restrict attention to those state-action pairs obeying $Q_h^k(s,a)<H$ in the sequel.

In view of the update rule \eqref{eq:updateqv-ucbvi} and the bonus \eqref{eq:bh-defn-Bernstein-ucbvi}, when $Q_h^k(s,a)<H$, we have
\begin{align}
\arxiv{&} Q_{h}^{k}(s,a) \book{&} =r_h(s,a)+\big\langle\widehat{P}_{s,a,h}^{k},V_{h+1}^{k}\big\rangle \arxiv{\nonumber
 \\ & \qquad \qquad }+c_{1}\sqrt{\frac{\mathsf{Var}_{\widehat{P}_{s,a,h}^{k}}\big(V_{h+1}^{k}\big)\log\frac{1}{\delta'}}{N_{h}^{k}(s,a)}}+c_{2}\frac{H\log\frac{1}{\delta'}}{N_{h}^{k}(s,a)}  \nonumber\\
 & \arxiv{\quad } \geq r_h(s,a)+ \frac{(c_2-c_1^2)H\log\frac{1}{\delta'}}{N_{h}^{k}(s,a)}+f\big(\widehat{P}_{s,a,h}^{k},V_{h+1}^{k},N_{h}^{k}(s,a)\big)\nonumber\\
 &\arxiv{\quad} \geq r_h(s,a) 
 %+ \frac{(c_2-c_1^2)H\log\frac{1}{\delta'}}{N_{h}^{k}(s,a)}
 +f\big(\widehat{P}_{s,a,h}^{k},V_{h+1}^{\star},N_{h}^{k}(s,a)\big)
	\label{eq:Qhk-sa-LB1}
 \end{align}
as long as $c_2\geq c_1^2$, 
where the last inequality results from the claim \eqref{eq:nondecreasing-f-property} and the property $V_{h+1}^k \geq V_{h+1}^{\star}$.
Moreover, applying the empirical Bernstein inequality in Lemma \ref{empirical bernstein}  yields 
\begin{align*}
 & \mathbb{P}\Bigg\{ \Big|\big\langle\widehat{P}_{s,a,h}^{k}-P_{s,a,h},\,V_{h+1}^{\star}\big\rangle\Big|>2\sqrt{\frac{\mathsf{Var}_{\widehat{P}_{s,a,h}^{k}}\big(V_{h+1}^{\star}\big)\log \frac{1}{\delta'}}{N_{h}^{k}(s,a)}}+\frac{14H\log \frac{1}{\delta'}}{3N_h^{k}(s,a)}\Bigg\} \nonumber\\
 & \leq\mathbb{P}\Bigg\{ \Big|\big\langle\widehat{P}_{s,a,h}^{k}-P_{s,a,h},\,V_{h+1}^{\star}\big\rangle\Big|>\sqrt{\frac{2\mathsf{Var}_{\widehat{P}_{s,a,h}^{k}}\big(V_{h+1}^{\star}\big)\log \frac{1}{\delta'}}{N_{h}^{k}(s,a)-1}}+\frac{\frac{7}{3}H\log \frac{1}{\delta'}}{N_{h}^{k}(s,a)-1}\Bigg\} \arxiv{ \notag\\
 & } \leq2\delta' .
%	\label{eq_lemma1_ref.5+ref1}	
\end{align*}
This in turn implies that, with probability exceeding $1-2\delta'$, 
\begin{align*}
f\big(\widehat{P}_{s,a,h}^{k},V_{h+1}^{\star},N_{h}^{k}(s,a)\big) & \geq \big\langle P_{s,a,h},V_{h+1}^{\star}\big\rangle+\big\langle\widehat{P}_{s,a,h}^{k}-P_{s,a,h},V_{h+1}^{\star}\big\rangle \\
 & ~~~~ +\frac{c_1}{2}\sqrt{\frac{\mathsf{Var}_{\widehat{P}_{s,a,h}^{k}}\big(V_{h+1}^{\star}\big)\log\frac{1}{\delta'}}{N_{h}^{k}(s,a)}}+\frac{c_1^2H\log\frac{1}{\delta'}}{2N_{h}^{k}(s,a)} \notag\\
 & \geq \big\langle P_{s,a,h},V_{h+1}^{\star}\big\rangle+\bigg(\frac{c_1^2}{2} -\frac{14}{3}\bigg)\frac{H\log\frac{1}{\delta'}}{N_{h}^{k}(s,a)} \\
 & ~~~~ +\bigg(\frac{c_1}{2}-2\bigg)\sqrt{\frac{\mathsf{Var}_{\widehat{P}_{s,a,h}^{k}}\big(V_{h+1}^{\star}\big)\log\frac{1}{\delta'}}{N_{h}^{k}(s,a)}} \notag\\
 &\geq\big\langle P_{s,a,h},V_{h+1}^{\star}\big\rangle,
\end{align*}
where the last line holds as long as $c_1\geq 4$. 
Substitution into \eqref{eq:Qhk-sa-LB1} then yields that, with probability at least $1-2\delta'$,  
$$	
Q_h^k(s,a)\geq r_h(s,a)+ \big\langle P_{s,a,h}, V_{h+1}^{\star} \big\rangle= Q_h^{\star}(s,a),
$$
where the identity is due to the Bellman equation~\eqref{eq:Bellman-matrix-finite}. 

To finish up, taking the union bound over all $(s,a,h,k)\in \cS\times \cA\times [H] \times [K]$ concludes the proof.

\paragraph{Putting all this together.} 
With the above two cases in place,  one can invoke standard induction arguments to deduce that: with probability at least $1-4SAHK\delta'$, one has $Q_h^k(s,a)\geq Q_h^{\star}(s,a)$ 
and $V_h^k = \max_a Q_h^k(s,a)\geq \max_a Q_h^{\star}(s,a)= V_h^{\star}(s)$ for every $(s,a,h,k)$, thus completing the proof.

\begin{proof}[Proof of property~\eqref{eq:nondecreasing-f-property}]
Fix any index $1\leq s\leq S$, and consider $f$ as a function of the $s$-th entry $v_s$ while freezing $p, n$ and all other entries of $v$. 
We begin with the following observations:  
\begin{itemize}
\item $f$ is a continuous function w.r.t.~$v$; 
\item $f$ is differentiable in $v_s$, except for at most two possible points that make $c_1\sqrt{\frac{\mathsf{Var}_p(v)\log \frac{1}{\delta'}} {n}}  = c_1^2 \frac{H\log \frac{1}{\delta'}}{n} $ (due to the presence of non-differentiable points of the max operator and the quadratic nature of $\mathsf{Var}_p(v)$). 

\end{itemize}
By virtue of the continuity of $f$, establishing that $f$ is non-decreasing in $v_s$ reduces to verifying that $\frac{\partial f(p,v,n)}{\partial v_s} \geq 0$ at all differentiable points.

To this end, we compute the partial derivative at any differentiable point $v_s$ as follows:
 \begin{align}
	\frac{\partial f(p,v,n)}{\partial v_s}  &=p_s+ c_1\mathds{1}\Bigg\{c_1\sqrt{\frac{\mathsf{Var}_p(v)\log\frac{1}{\delta'}}{n}}\geq c_1^2\frac{H\log\frac{1}{\delta'}}{n}\Bigg\}  \arxiv{ \notag\\
    &\qquad\quad } \cdot  \frac{p_s\big(v_s-\langle p,v\rangle\big)\sqrt{\log\frac{1}{\delta'}}}{\sqrt{n\mathsf{Var}_p(v)}}\nonumber\\
 & =p_s+\mathds{1}(\mathcal{A}_v)
 %\Bigg\{\sqrt{n\mathsf{Var}_p(v)\log\frac{1}{\delta'}}\geq c_1 H\log\frac{1}{\delta'}\Bigg\} \notag\\
 %&\qquad \quad 
 \cdot\frac{c_1H\log\frac{1}{\delta'}}{\sqrt{n\mathsf{Var}_p(v)\log\frac{1}{\delta'}}}\cdot\frac{p_s\big(v_s-\langle p,v\rangle\big)}{H} ,
 \label{eq:derivative-f-calculation-middle1}
\end{align}
where we introduce the event $\mathcal{A}_v \coloneqq \big\{\sqrt{n\mathsf{Var}_p(v)\log\frac{1}{\delta'}}\geq c_1 H\log\frac{1}{\delta'}\big\}$. We now divide into two cases.
\begin{itemize}
\item 
If $v_s-\langle p,v\rangle \geq 0$, then it is trivially seen that $\frac{\partial f(p,v,n)}{\partial v_s} \geq 0$.
\item If $v_s-\langle p,v\rangle < 0$, then one can continue the derivation in \eqref{eq:derivative-f-calculation-middle1} and use the definition of $\mathcal{A}_v$ to reach
\begin{align*}
    \frac{\partial f(p,v,n)}{\partial v_s}  &\geq  p_s+p_s\frac{\big(v_s-\langle p,v\rangle\big)}{H} = p_s \frac{H+v_s - \langle p, v \rangle}{H} \geq 0, 
\end{align*}
where the last inequality holds since $v_s\geq 0$ and $\langle p,v\rangle \leq H$. 
\end{itemize}
Combining these two cases immediately establishes property~\eqref{eq:nondecreasing-f-property}. 
\end{proof}

\subsection{Proof of Lemma~\ref{lem:bound-W1-W2-ucbvi-proof}}
\label{sec:proof-lemma:bound-W1-W2-ucbvi-proof}

\paragraph{Bounding the quantity $W_1$.}
% and $W_3$
%
We begin by applying Bernstein's inequality (cf.~Lemma~\ref{lem:Bernstein-prelim}) in conjunction with a union bound over all tuples $(s,a,h,k) \in \mathcal{S}\times\mathcal{A}\times[H]\times[K]$. More specifically, since $\|V_{h+1}^{\star}\|_{\infty} \leq H$, it holds with probability at least $1-4SAHK\delta'$ that
\begin{equation}
\Big|\Big\langle\widehat{P}_{s,a,h}^{k}-P_{s,a,h},V_{h+1}^{\star}\Big\rangle\Big|\leq\sqrt{\frac{2\mathsf{Var}_{P_{s,a,h}}(V_{h+1}^{\star})\log\frac{1}{\delta'}}{N_{h}^{k}(s,a)}}+\frac{H\log\frac{1}{\delta'}}{N_{h}^{k}(s,a)}
\label{eq:bernstein-one-term-P-sahk-ucbvi}
\end{equation}
simultaneously for all $(s,a,h,k)\in \cS\times \cA\times [H]\times [K]$. 
%
%Clearly, \eqref{eq:bernstein-one-term-P-sahk-ucbvi} also immediately implies that
%%
%%
%\begin{equation*}
%\max\Big\{\Big\langle\widehat{P}_{s,a,h}^{k}-P_{s,a,h},V_{h+1}^{\star}\Big\rangle, 0\Big\}
%\leq\sqrt{\frac{2\mathsf{Var}_{P_{s,a,h}}(V_{h+1}^{\star})\log\frac{1}{\delta'}}{N_{h}^{k}(s,a)}}+\frac{H\log\frac{1}{\delta'}}{N_{h}^{k}(s,a)}.
%\end{equation*}
%
Using this relation to bound each summand in $W_1$, we obtain 
\begin{align}
W_{1}  & 
	%\le W_3
	\leq\sum_{k=1}^{K}\sum_{h=1}^{H}\sqrt{\frac{2\mathsf{Var}_{P_{s_{h}^{k},a_{h}^{k},h}}(V_{h+1}^{\star})\log\frac{1}{\delta'}}{N_{h}^{k}(s_{h}^{k},a_{h}^{k})}}+\sum_{k=1}^{K}\sum_{h=1}^{H}\frac{H\log\frac{1}{\delta'}}{N_{h}^{k}(s_{h}^{k},a_{h}^{k})}\nonumber \\
 & \leq\sqrt{\sum_{k,h}\frac{2\log\frac{1}{\delta'}}{N_{h}^{k}(s_{h}^{k},a_{h}^{k})}}\sqrt{\sum_{k,h}\mathsf{Var}_{P_{s_{h}^{k},a_{h}^{k},h}}\big(V_{h+1}^{\star}\big)}\nonumber+H\log\frac{1}{\delta'}\sum_{k,h}\frac{1}{N_{h}^{k}(s_{h}^{k},a_{h}^{k})}\nonumber\nonumber \\
 & \le 2\sqrt{SAHT_{6}^{\mathsf{opt}}(\log_{2}K)\log\frac{1}{\delta'}}
%\sqrt{\sum_{k,h}\mathsf{Var}_{P_{s_{h}^{k},a_{h}^{k},h}}\big(V_{h+1}^{\star}\big)}\nonumber\nonumber \\
 %& \qquad
 +2SAH^{2}(\log_{2}K)\log\frac{1}{\delta'}
 \label{eq:W1-ub-ucbvi-T1}
\end{align}
with probability greater than $1-4SAHK\delta'$,  
where the second line invokes the Cauchy-Schwarz inequality, and the last inequality makes use of Lemma~\ref{lem:sum-inv-Nhk-ucbvi} and the definition~\eqref{eq:defn-T6-varstar-proof} of $T_{6}^{\mathsf{opt}}$.

To finish up, note that for any two random variables $X$ and $Y$,
$$\mathsf{Var}(X+Y)\leq 2\mathsf{Var}(X)+2\mathsf{Var}(Y)\leq 2\mathsf{Var}(X)+2\mathbb{E}[Y^2].
$$ 
This combined with the property $V_{h+1}^{k}\geq V_{h+1}^{\star} \geq V_{h+1}^{\pi^k}$ (see Lemma~\ref{lemma:opt}) gives rise to
\begin{align*}
\mathsf{Var} _{P_{s_h^k,a_h^k,h}}\big( V_{h+1}^{\star} \big) 
&\leq 2 \mathsf{Var} _{P_{s_h^k,a_h^k,h}}\big( V_{h+1}^{\pi^k}\big) + 2\Big\langle P_{s_h^k,a_h^k,h}, \big( V_{h+1}^{\star}-V_{h+1}^{\pi^k} \big)^2\Big\rangle \notag\\
& \leq 2 \mathsf{Var} _{P_{s_h^k,a_h^k,h}}\big( V_{h+1}^{\pi^k}\big) + 2\Big\langle P_{s_h^k,a_h^k,h}, \big( V_{h+1}^k-V_{h+1}^{\pi^k} \big)^2\Big\rangle,
\end{align*}
which taken collectively with the definition \eqref{eq:defn-T56-proof} results in
\begin{subequations}
	\label{eq:T6opt-ub-T6-ucb-T6-diff}
\begin{align}
	T_{6}^{\mathsf{opt}} &\leq 2T_6^{\mathsf{low}} + 2T_{6}^{\mathsf{diff}} \\
	\Longrightarrow \qquad \sqrt{ T_{6}^{\mathsf{opt}} } &\leq \sqrt{2T_6^{\mathsf{low}}} + \sqrt{2T_{6}^{\mathsf{diff}}}.
\end{align}
\end{subequations}
This together with \eqref{eq:W1-ub-ucbvi-T1} and the definition of $\iotab$ (cf.~\eqref{eq:assumption-K-proof-B}) establishes the claimed bound
\[
	W_{1}  \le \frac{1}{8}\sqrt{\iotab SAHT_{6}^{\mathsf{low}}} + \frac{1}{8}\sqrt{\iotab SAHT_{6}^{\mathsf{diff}}}+ \frac{1}{8}\iotab SAH^{2}.
\]

\begin{remark}
Repeating the above arguments for \eqref{eq:T6opt-ub-T6-ucb-T6-diff} also yields
\begin{subequations}
	\label{eq:T6ucb-ub-T6-low-T6-diff}
\begin{align}
	T_{6}^{\mathsf{ucb}} &\leq 2T_6^{\mathsf{low}} + 2T_{6}^{\mathsf{diff}} \\
	\Longrightarrow \qquad \sqrt{ T_{6}^{\mathsf{ucb}} } &\leq \sqrt{2T_6^{\mathsf{low}}} + \sqrt{2T_{6}^{\mathsf{diff}}},
\end{align}
\end{subequations}
which will prove useful in the subsequent analysis.  
\end{remark}

\paragraph{Bounding the quantity $W_2$.}
We now turn our attention to the term $W_2$ in \eqref{eq:decompose-past-work}. 
Setting $$\Delta_{h+1}^{k}\coloneqq V_{h+1}^{k}-V_{h+1}^{\star}$$ for notational convenience and denoting by $v(s)$ the $s$-th entry of a vector $v$, 
we derive
\begin{align}
 & \Big\langle\widehat{P}_{s,a,h}^{k}-P_{s,a,h},V_{h+1}^{k}-V_{h+1}^{\star}\Big\rangle
 =
 \sum_{s'\in\mathcal{S}}\Big(\widehat{P}_{s,a,h}^{k}(s')-P_{s,a,h}(s')\Big)\Delta_{h+1}^{k}(s') \notag\\
 & \quad \overset{\text{(i)}}{=} \sum_{s'\in\mathcal{S}}\Big(\widehat{P}_{s,a,h}^{k}(s')-P_{s,a,h}(s')\Big)\Delta_{h+1}^{k}(s')  \arxiv {\notag\\
 &\qquad\qquad } -\Bigg\{\sum_{s'\in\mathcal{S}}\Big(\widehat{P}_{s,a,h}^{k}(s')-P_{s,a,h}(s')\Big)\Bigg\}\big\langle P_{s,a,h},\Delta_{h+1}^{k}\big\rangle \notag\\
 & \quad=\sum_{s'\in\mathcal{S}}\Big(\widehat{P}_{s,a,h}^{k}(s')-P_{s,a,h}(s')\Big)\Big(\Delta_{h+1}^{k}(s')-\big\langle P_{s,a,h},\Delta_{h+1}^{k}\big\rangle\Big) \notag\\
 &\quad\leq\sum_{s'\in\mathcal{S}}\Big|\widehat{P}_{s,a,h}^{k}(s')-P_{s,a,h}(s')\Big|\cdot\Big|\Delta_{h+1}^{k}(s')-\big\langle P_{s,a,h},\Delta_{h+1}^{k}\big\rangle\Big|,
 \label{eq:fluc-V-diff-ucbvi-expand}
\end{align}
where (i) follows since $\sum_{s'}\widehat{P}_{s,a,h}^{k}(s')=\sum_{s'} P_{s,a,h}(s')=1$. 
The Bernstein inequality (cf.~Lemma~\ref{lem:Bernstein-prelim}) combined with the union bound tells us that, with probability greater than $1-4S^2AHK\delta'$, 
\begin{align}
\Big|\widehat{P}_{s,a,h}^{k}(s')-P_{s,a,h}(s')\Big| & =\Big|\Big\langle\widehat{P}_{s,a,h}^{k}-P_{s,a,h},\, e_{s'}\Big\rangle\Big| \notag\\
&\leq\sqrt{\frac{2\mathsf{Var}_{P_{s,a,h}}(e_{s'})\log\frac{1}{\delta'}}{N_{h}^{k}(s,a)}}+\frac{2\log\frac{1}{\delta'}}{3N_{h}^{k}(s,a)}\nonumber \\
 & \leq\sqrt{\frac{2P_{s,a,h}(s')\log\frac{1}{\delta'}}{N_{h}^{k}(s,a)}}+\frac{\log\frac{1}{\delta'}}{N_{h}^{k}(s,a)}
 \label{eq:Bernstein-every-entry-ucbvi-Pk}
\end{align}
holds simultaneously for all $(s,a,s',h,k)\in \cS\times \cA\times \cS\times [H]\times [K]$, where the last inequality uses the fact that $\mathsf{Var}_{P_{s,a,h}}(e_{s'})=P_{s,a,h}(s')(1-P_{s,a,h}(s'))\leq P_{s,a,h}(s')$. Substitution into \eqref{eq:fluc-V-diff-ucbvi-expand} then leads to 
\begin{align}
%\eqref{eq:fluc-V-diff-ucbvi-expand} & 
& \Big\langle\widehat{P}_{s,a,h}^{k}-P_{s,a,h},V_{h+1}^{k}-V_{h+1}^{\star}\Big\rangle \notag\\
&\arxiv{\quad} \leq\sum_{s'\in\mathcal{S}}\Bigg(\sqrt{\frac{2P_{s,a,h}(s')\log\frac{1}{\delta'}}{N_{h}^{k}(s,a)}}+\frac{\log\frac{1}{\delta'}}{N_{h}^{k}(s,a)}\Bigg)\Big|\Delta_{h+1}^{k}(s')-\big\langle P_{s,a,h},\Delta_{h+1}^{k}\big\rangle\Big|\notag\\
 & \arxiv{\quad} \overset{\text{(ii)}}{\leq} \sqrt{\frac{2S\log\frac{1}{\delta'}}{N_{h}^{k}(s,a)}}\sqrt{\sum_{s'\in\mathcal{S}}P_{s,a,h}(s')\Big|\Delta_{h+1}^{k}(s')-\big\langle P_{s,a,h},\Delta_{h+1}^{k}\big\rangle\Big|^{2}} \arxiv{ \notag\\
 & \quad\qquad} +\sum_{s'\in\mathcal{S}}\frac{\log\frac{1}{\delta'}}{N_{h}^{k}(s,a)}\Big|\Delta_{h+1}^{k}(s')-\big\langle P_{s,a,h},\Delta_{h+1}^{k}\big\rangle\Big|\notag\\
 & \arxiv{\quad}   \overset{\text{(iii)}}{\leq}\sqrt{\frac{2S\log\frac{1}{\delta'}}{N_{h}^{k}(s,a)}}\sqrt{\mathsf{Var}_{P_{s,a,h}}\big(V_{h+1}^{k}-V_{h+1}^{\star}\big)}+\frac{SH\log\frac{1}{\delta'}}{N_{h}^{k}(s,a)},
 \notag
\end{align}
where (ii) follows from the Cauchy-Schwarz inequality, and (iii) results from the definition of $\mathsf{Var}_{P_{s,a,h}}(V)$ and the fact that $\Delta_{h+1}^k (s) \in [0, H]$ for all $s\in \cS$ (since both $V_{h+1}^k (s)$ and $V_{h+1}^{\star} (s)$ lie in $[0,H]$).  
Repeating similar arguments as in \eqref{eq:W1-ub-ucbvi-T1}, we arrive at
\begin{align}
W_{2} & \le 2\sqrt{S^{2}AH(\log_{2}K)\log\frac{1}{\delta'}}\sqrt{\sum_{k,h}\mathsf{Var}_{P_{s_{h}^{k},a_{h}^{k},h}}\big(V_{h+1}^{k}-V_{h+1}^{\star}\big)} \arxiv{ \nonumber\\
 & \qquad} +2S^{2}AH^{2}(\log_{2}K)\log\frac{1}{\delta'} 
 \label{eq:W2-ub-ucbvi-proof} \\
	& \le \frac{1}{8} \sqrt{\iotab S^{2}AHT_{6}^{\mathsf{diff}}} + \frac{1}{8} \iotab S^{2}AH^{2} \notag
\end{align}
with probability at least $1-4S^2AHK\delta'$, where we recall the definition of $T_{6}^{\mathsf{diff}}$ in \eqref{eq:defn-T6-vardif-proof}. Here, the last line follows since, according to the property $V_{h+1}^{k}\geq V_{h+1}^{\star}\geq  V_{h+1}^{\pi^{k}}$ (see Lemma~\ref{lemma:opt}), 
\[
\mathsf{Var}_{P_{s,a,h}}\big(V_{h+1}^{k}-V_{h+1}^{\star}\big)\leq\Big\langle P_{s,a,h},\big(V_{h+1}^{k}-V_{h+1}^{\star}\big)^{2}\Big\rangle\leq\Big\langle P_{s,a,h},\big(V_{h+1}^{k}-V_{h+1}^{\pi^{k}}\big)^{2}\Big\rangle,
\]
\begin{equation}
\Longrightarrow\qquad\sum_{k,h}\mathsf{Var}_{P_{s_{h}^{k},a_{h}^{k},h}}\big(V_{h+1}^{k}-V_{h+1}^{\star}\big)\leq T_{6}^{\mathsf{diff}}.
	\label{eq:Var-Vk-Vstar-diff-T6diff-ucbvi}
\end{equation}
%

%together with the definition~\eqref{eq:defn-T6-vardif-proof} of $T_{6}^{\mathsf{diff}}$ establishes the advertised bound on $W_2$. 

\paragraph{Bounding the quantity $W_3$.} 
Akin to the decomposition~\eqref{eq:decompose-past-work}, we split $W_3$ into two components: 
\begin{align}
	&
    %\sum_{k,h} \Big\langle \widehat{P}^{k}_{s_h^k,a_h^k,h}-P_{s_h^k,a_h^k,h},\big(V_{h+1}^k\big)^2 \Big\rangle
    W_3
    =
    \underset{\eqqcolon\, W_4}{\underbrace{ \sum_{k,h} \Big\langle \widehat{P}^{k}_{s_h^k,a_h^k,h}-P_{s_h^k,a_h^k,h}, \big(V_{h+1}^{\star}\big)^2 \Big\rangle }}
	\arxiv{ \nonumber\\
	&\qquad\qquad }
	+ \underset{\eqqcolon\, W_5}{\underbrace{\sum_{k,h} \Big\langle\widehat{P}^{k}_{s_h^k,a_h^k,h}-P_{s_h^k,a_h^k,h}, \big(V_{h+1}^k\big)^2  -\big(V_{h+1}^{\star}\big)^2\Big\rangle }}.
	\label{eq:decompose-past-work-W5}
\end{align}
The resulting quantities $W_4$ and $W_5$ can then be controlled using arguments analogous to those employed for $W_1$ and $W_2$, respectively. Additionally, it is helpful to record the following basic variance bound, quoted from \citet[Lemma~30]{chen2021implicit}. 
\begin{lemma}\label{lemma:sqv}
	Let $X$ be a random variable, and denote by $C_{\max}$ the largest possible magnitude of $X$. 
	Then we have $\mathsf{Var}(X^2)\leq 4 C_{\max}^2 \mathsf{Var}(X)$ and $\mathsf{Var}\big((X-\mathbb{E}[X])^2\big)\leq  C_{\max}^2 \mathsf{Var}(X)$.
\end{lemma}
\begin{proof}[Proof of Lemma~\ref{lemma:sqv}]
From the elementary fact $\mathsf{Var}(U+V)\leq 2\mathsf{Var}(U)+2\mathsf{Var}(V)$, we can derive
\begin{align*}
\mathsf{Var}(X^{2}) & \leq2\mathsf{Var}\big((X-\mathbb{E}[X])X\big)+2\mathsf{Var}\big(\mathbb{E}[X]X\big)\\
 & \leq2\mathbb{E}\big[(X-\mathbb{E}[X])^{2}X^{2}\big]+2(\mathbb{E}[X])^{2}\mathsf{Var}(X)\\
 & \leq2C_{\max}^{2}\mathbb{E}\big[(X-\mathbb{E}[X])^{2}\big]+2C_{\max}^{2}\mathsf{Var}(X)=4C_{\max}^{2}\mathsf{Var}(X).
\end{align*}
Similarly, we have
\begin{align*}
\mathsf{Var}\big( (X-\mathbb{E}[X])^{2}\big) & \leq \mathbb{E}\big[(X-\mathbb{E}[X])^4\big]
  \leq C_{\max}^{2}\mathbb{E}\big[(X-\mathbb{E}[X])^{2}\big]=C_{\max}^{2}\mathsf{Var}(X)
\end{align*}
as claimed. 
\end{proof}

Regarding the term $W_4$ in \eqref{eq:decompose-past-work-W5}, repeating the arguments in \eqref{eq:W1-ub-ucbvi-T1} yields, with probability greater than $1-4SAHK\delta'$, that 
\begin{align}
W_4
 & \le 2\sqrt{SAH(\log_{2}K)\log\frac{1}{\delta'}} \sqrt{\sum_{k,h}\mathsf{Var}_{P_{s_{h}^{k},a_{h}^{k},h}} \big( (V_{h+1}^{\star})^2\big)}\arxiv{ \notag\\
 &\qquad  }
 +2SAH^3(\log_{2}K)\log\frac{1}{\delta'} \notag\\
 & \le 4\sqrt{SAH^3T_6^{\mathsf{opt}}(\log_{2}K)\log\frac{1}{\delta'}}  
 +2SAH^3(\log_{2}K)\log\frac{1}{\delta'} \notag\\
	& \le \frac{1}{16}\sqrt{\iotab SAH^{3}T_{6}^{\mathsf{low}}}+ \frac{1}{16}\sqrt{\iotab SAH^{3}T_{6}^{\mathsf{diff}}}+ \frac{1}{16} \iotab SAH^{3}
 .\notag
% \label{eq:W5-ub-proof-123-ucbvi}
\end{align}
Here, the last inequality follows from \eqref{eq:T6opt-ub-T6-ucb-T6-diff}, whereas the penultimate inequality applies Lemma~\ref{lemma:sqv} and the property $\|V_{h+1}^{\star}\|_{\infty}\leq H$ to obtain
\[
\mathsf{Var}_{P_{s_{h}^{k},a_{h}^{k},h}} \big( (V_{h+1}^{\star})^2\big)
\leq 4H^2 \mathsf{Var}_{P_{s_{h}^{k},a_{h}^{k},h}} \big( V_{h+1}^{\star}\big).
\]
%

 %Denoting $\widetilde{\Delta}_{h+1}^k = \big(V_{h+1}^{k}\big)^2-\big(V_{h+1}^{\star}\big)^2$ for notational simplicity, one can 

  When it comes to the term $W_5$ in \eqref{eq:decompose-past-work-W5}, repeat the arguments for \eqref{eq:W2-ub-ucbvi-proof} to derive
\begin{align*}
&W_5
\le 2\sqrt{S^{2}AH(\log_{2}K)\log\frac{1}{\delta'}}\sqrt{\sum_{k,h}\mathsf{Var}_{P_{s,a,h}}\Big(\big(V_{h+1}^{k}\big)^2-\big(V_{h+1}^{\star}\big)^2\Big)} \arxiv{ \nonumber\\
 & \qquad } +2S^{2}AH^{3}(\log_{2}K)\log\frac{1}{\delta'}.
 %\label{eq:fluc-V-diff-ucbvi-expand-W6}
 \end{align*}
Furthermore, recalling the property $H\geq V_{h+1}^{k}(s)\geq V_{h+1}^{\star}(s) \geq V_{h+1}^{\pi^k}(s)\geq 0$ that holds for all $s\in \cS$ (see Lemma~\ref{lemma:opt}), one can deduce that
\begin{align*}
\mathsf{Var}_{P_{s,a,h}}\Big(\big(V_{h+1}^{k}\big)^{2}-\big(V_{h+1}^{\star}\big)^{2}\Big) & \leq\Big\langle P_{s,a,h},\Big(\big(V_{h+1}^{k}\big)^{2}-\big(V_{h+1}^{\star}\big)^{2}\Big)^{2}\Big\rangle\\
 & \leq\Big\langle P_{s,a,h},\Big(\big(V_{h+1}^{k}\big)^{2}-\big(V_{h+1}^{\pi^{k}}\big)^{2}\Big)^{2}\Big\rangle\\
 & =\Big\langle P_{s,a,h},\Big(\big(V_{h+1}^{k}-V_{h+1}^{\pi^{k}}\big)\circ\big(V_{h+1}^{k}+V_{h+1}^{\pi^{k}}\big)\Big)^{2}\Big\rangle\\
 & \leq 4H^{2}\Big\langle P_{s,a,h},\big(V_{h+1}^{k}-V_{h+1}^{\pi^{k}}\big)^{2}\Big\rangle.
\end{align*}
Taking this inequality together with the definition~\eqref{eq:defn-T56-proof}, we arrive at
\begin{align*}
W_5
	&\le 4\sqrt{S^{2}AH^3T_6^{\mathsf{diff}}(\log_{2}K)\log\frac{1}{\delta'}} +2S^{2}AH^{3}(\log_{2}K)\log\frac{1}{\delta'} \notag\\
	&\leq \frac{1}{16} \sqrt{\iotab S^{2}AH^{3}T_{6}^{\mathsf{diff}}}+ \frac{1}{16}\iotab S^{2}AH^{3}.
 \end{align*}

 Substituting the preceding bounds on $W_4$ and $W_5$ into \eqref{eq:decompose-past-work-W5}, we conclude that
 \begin{align}
W_3 & \leq W_4+W_5 \leq \frac{1}{8}\sqrt{\iotab SAH^{3}T_{6}^{\mathsf{low}}}+\frac{1}{8}\sqrt{\iotab S^{2}AH^{3}T_{6}^{\mathsf{diff}}}+\frac{1}{8} \iotab S^{2}AH^{3}.
	%\notag\\
% & \leq  4\sqrt{SAH^3(T_6^{\mathsf{low}}+T_6^{\mathsf{diff}})(\log_{2}K)\log\frac{1}{\delta'}}  
% +SAH^3(\log_{2}K)\log\frac{1}{\delta'} \notag \\
% &\qquad +\sqrt{8S^{2}AH^3T_6^{\mathsf{diff}}(\log_{2}K)\log\frac{1}{\delta'}} +S^{2}AH^{3}(\log_{2}K)\log\frac{1} {\delta'}\notag\\
% & \leq 4\sqrt{SAH^3T_6^{\mathsf{low}}(\log_{2}K)\log\frac{1}{\delta'}}  
% +8\sqrt{S^{2}AH^3T_6^{\mathsf{diff}}(\log_{2}K)\log\frac{1}{\delta'}}
% \notag\\
% &\qquad +2S^{2}AH^{3}(\log_{2}K)\log\frac{1} {\delta'}
 \notag
% \label{eq:W5-ub-proof-123-ucbvi}
\end{align}
%
%as claimed. 
%\begin{align}
% & \Big\langle\widehat{P}_{s,a,h}^{k}-P_{s,a,h}, \big(V_{h+1}^{k}\big)^2-\big(V_{h+1}^{\star}\big)^2\Big\rangle \notag\\
%&\quad\leq\sqrt{\frac{2S\log\frac{1}{\delta'}}{N_{h}^{k}(s,a)}}\sqrt{\mathsf{Var}_{P_{s,a,h}}\Big(\big(V_{h+1}^{k}\big)^2-\big(V_{h+1}^{\star}\big)^2\Big)}+\frac{SH^2\log\frac{1}{\delta'}}{N_{h}^{k}(s,a)}.
% \label{eq:fluc-V-diff-ucbvi-expand-W6}
%\end{align}
%

\subsection{Proof of Lemma~\ref{lem:bound-T234}}\label{app:pflem:bound-T234}

\paragraph{Bounding the term $T_2$.} 
We begin by analyzing $T_2$ (cf.~\eqref{eq:decomposition}). Recalling the definition of the bonus term \eqref{eq:bh-defn-Bernstein-ucbvi}, we have
\begin{align}
T_{2} & =\sum_{k=1}^{K}\sum_{h=1}^{H}b_{h}^{k}(s_{h}^{k},a_{h}^{k})\arxiv{ \nonumber\\
 & }= c_1\sum_{k,h}\sqrt{\frac{\mathsf{Var}_{\widehat{P}_{s_{h}^{k},a_{h}^{k},h}^{k}}\big(V_{h+1}^{k}\big)\log\frac{1}{\delta'}}{N_{h}^{k}(s_{h}^{k},a_{h}^{k})}}
 +c_2\sum_{k,h}\frac{H\log\frac{1}{\delta'}}{N_{h}^{k}(s_{h}^{k},a_{h}^{k})}.
	\label{eq:T2-decomposition-main1}
\end{align}
Applying the Cauchy-Schwarz inequality to the first sum in \eqref{eq:T2-decomposition-main1} yields
\begin{align}
T_{2} & \leq c_1\sqrt{\sum_{k,h}\frac{\log\frac{1}{\delta'}}{N_{h}^{k}(s_{h}^{k},a_{h}^{k})}}\sqrt{\sum_{k,h}\mathsf{Var}_{\widehat{P}_{s_{h}^{k},a_{h}^{k},h}^{k}}\big(V_{h+1}^{k}\big)} \arxiv{ \nonumber\\
	& \qquad } +c_2H \bigg(\log\frac{1}{\delta'} \bigg)\sum_{k,h}\frac{1}{N_{h}^{k}(s_{h}^{k},a_{h}^{k})}\nonumber\\
	&= c_1\sqrt{\sum_{k,h}\frac{\log\frac{1}{\delta'}}{N_{h}^{k}(s_{h}^{k},a_{h}^{k})}} \sqrt{T_5}+c_2H\bigg(\log\frac{1}{\delta'} \bigg) \sum_{k,h}\frac{1}{N_{h}^{k}(s_{h}^{k},a_{h}^{k})}\nonumber\\
 & \leq c_1\sqrt{2SAH(\log_{2}K)\Big(\log\frac{1}{\delta'}\Big)T_{5}}
+2c_2SAH^{2}(\log_{2}K)\log\frac{1}{\delta'} \notag\\
	&\le \frac{1}{2}\sqrt{ SAHT_{5} \iotab}+\frac{1}{2} SAH^{2} \iotab,
	\notag
\end{align}
where the second relation uses the definition \eqref{eq:defn-T5-proof} of $T_5$, the penultimate line invokes the visitation count property~\eqref{eq:sum-inv-Nhk-ucbvi}, and the last line holds as long as $\iotab \geq 8\max\{c_1^2,c_2\}(\log_{2}K)\log\frac{1}{\delta'}$.

\paragraph{Bounding the term $T_3$.} 
Next, we bound the term $T_3$ defined in \eqref{eq:decomposition}. Recall that, by definition, the value function estimate $V_{h+1}^{k}$ is computed prior to the start of episode $k$. Consequently, conditional on the state-action pair $(s_{h}^{k},a_{h}^{k})$ and the history up to the beginning of episode $k$, the next-state indicator vector $e_{s_{h+1}^{k}}$ is statistically independent of $V_{h+1}^{k}$ and has conditional mean $ P_{s_{h}^{k},a_{h}^{k},h}$. Furthermore, since all entries of $V_{h+1}^{k}$ lie within $[0, H]$, one has the deterministic bound:
\[
\big| \big\langle P_{s_{h}^{k},a_{h}^{k},h}-e_{s_{h+1}^{k}},V_{h+1}^{k}\big\rangle \big| \leq H.
\]
Additionally, it is easily seen that $T_6^{\ucb}$ (cf.~\eqref{eq:defn-T6-proof}) obeys the trivial upper bound 
\begin{align}
T_6^{\ucb} \leq  \sum_{k=1}^K \sum_{h=1}^H  \big\langle P_{s_h^k,a_h^k,h}, \big( V_{h+1}^k \big)^2\big\rangle \leq 
 KH^3.
	\label{eq:T6-ucb-trivial-UB-12345}
\end{align}

Therefore, using the definition of $T_6^{\ucb}$ in \eqref{eq:defn-T6-proof} and its trivial upper bound \eqref{eq:T6-ucb-trivial-UB-12345}, 
one can invoke the Freedman inequality (see Lemma~\ref{lem:Freedman-prelim} with $K_0=\log_2 (KH)$) to show that, with probability at least $1-2\delta'\log_2 (KH)$, 
\begin{align*}
 |T_3|  &\leq \sqrt{ 8 \max\{T_6^{\ucb}, H^2\} \log \frac{1}{\delta'} } + \frac{4}{3}H\log \frac{1}{\delta'} \notag\\
 &
	\leq  \sqrt{  8T_6^{\ucb} \log \frac{1}{\delta'}  } +\sqrt{8}H\log \frac{1}{\delta'}  + \frac{4}{3}H\log \frac{1}{\delta'} \notag\\
 &
	\leq  \sqrt{  8T_6^{\ucb} \log \frac{1}{\delta'}  } + 5H\log \frac{1}{\delta'} \notag\\
	&\leq  4\sqrt{  T_6^{\mathsf{low}} \log \frac{1}{\delta'}  } + 4\sqrt{  T_6^{\mathsf{diff}} \log \frac{1}{\delta'}  } + 5H\log \frac{1}{\delta'} ,
    %\label{eq:boundt3-proof}
\end{align*}
 where the second line uses the elementary inequality $\sqrt{\max\{a,b\}}\le \sqrt{a} + \sqrt{b}$, and the last inequality arises from \eqref{eq:T6ucb-ub-T6-low-T6-diff}. 
 %Furthermore, repeating the same arguments as in \eqref{eq:T6opt-ub-T6-ucb-T6-diff} reveals that
 %%
%\begin{equation}
 %   T_{6} \leq 2T_6^{\mathsf{low}} + 2T_{6}^{\mathsf{diff}},
 %   \label{eq:T6opt-ub-T6-ucb-T6-diff-up}
%\end{equation}
%
%which together with the above bound on $T_3$ and the fact $\sqrt{a+b}\le \sqrt{a} + \sqrt{b}$ readily yields
%%
%\begin{align*}
% |T_3|  &\leq  4\sqrt{  T_6^{\mathsf{low}} \log \frac{1}{\delta'}  } + 4\sqrt{  T_6^{\mathsf{diff}} \log \frac{1}{\delta'}  } + 5H\log \frac{1}{\delta'} .
%    %\label{eq:boundt3-proof}
%\end{align*}
%

\paragraph{Bounding the term $T_4$.} 
We now turn attention to the term $T_4$, i.e., 
\begin{align}
T_4 =  \sum_{k=1}^K \bigg( \underset{\eqqcolon\, E_k}{\underbrace{\sum_{h=1}^H r_h(s_h^k,a_h^k) - V_1^{\pi^k}(s_1^k) }} \bigg),
	\label{eq:T4-defn-repeat}
\end{align}
Note that conditional on $\pi^k$,  $E_k$ is a zero-mean random variable bounded in magnitude by $H$. An application of the Azuma-Hoeffding inequality (cf.~Lemma~\ref{lem:azuma-hoeffding}) guarantees that, with probability exceeding $1- 2\delta' $, 
\begin{align}
	|T_4|=\left| \sum_{k=1}^K E_k \right|
	&\leq \sqrt{2KH^2\log \frac{1}{\delta'}} .
	%\label{eq:bdt4_1}
    \notag
\end{align}

\subsection{Proof of Lemma~\ref{lem:bound-T56}} \label{sec:pflem:bound-T56}

To facilitate the analysis, we introduce the following auxiliary quantity:
\begin{align}
T_{8} & \coloneqq\sum_{k=1}^{K}\sum_{h=1}^{H}\Big\langle P_{s_{h}^{k},a_{h}^{k},h}-e_{s_{h+1}^{k}},\big(V_{h+1}^{k}\big)^{2}\Big\rangle.
\label{eq:defn-T8-proof}
\end{align}
Since this quantity will appear in the upper bounds on several other terms, we begin by deriving an upper bound for $T_8$.

\paragraph{Bounding the term $T_8$.} 
%
%Finally, note that the above bounds depend on the term $T_8$ (cf.~\eqref{eq:defn-T8-proof}), 
%which we would like to cope with now. 
%
According to Lemma~\ref{lemma:sqv}, $\mathsf{Var}(X^2)\leq 4 H^2\mathsf{Var}(X)$ holds for any random variable $X$ with support on $[-H,H]$. This implies that
\begin{align}
	%\sum_{k,h}
	\mathsf{Var}_{P_{s_{h}^{k},a_{h}^{k},h}}\Big(\big(V_{h+1}^{k}\big)^{2}\Big) \leq 4H^2 \mathsf{Var}_{P_{s_{h}^{k},a_{h}^{k},h}}\big(V_{h+1}^{k}\big). 
	%4KH^5. 
	\label{eq:Var-P-Vk-square-H4}
\end{align}
Therefore, invoking Freedman's inequality (see Lemma~\ref{lem:Freedman-prelim} with $K_0=\log_2 (KH)$) together with the trivial bound $\mathsf{Var}_{P_{s_{h}^{k},a_{h}^{k},h}}\big(\big(V_{h+1}^{k}\big)^{2}\big) \leq H^4$ reveals that, with probability at least $1-2\delta'\log_2(KH)$, 
\begin{align}
	|T_{8}|&\leq \sqrt{8\max\bigg\{ \sum_{k,h}\mathsf{Var}_{P_{s_{h}^{k},a_{h}^{k},h}}\Big(\big(V_{h+1}^{k}\big)^{2}\Big), H^4 \bigg\}\log\frac{1}{\delta'}}+\frac{4}{3} H^{2}\log\frac{1}{\delta'} \notag\\
	&	\leq \sqrt{8\max\bigg\{ \sum_{k,h}\mathsf{Var}_{P_{s_{h}^{k},a_{h}^{k},h}}\Big(\big(V_{h+1}^{k}\big)^{2}\Big), H^4 \bigg\}\log\frac{1}{\delta'}}+\sqrt{8 H^{4}\log\frac{1}{\delta'} } \arxiv{ \notag\\
	&\qquad } + \frac{4}{3} H^{2}\log\frac{1}{\delta'} \notag\\
    &\leq \sqrt{8 \sum_{k,h}\mathsf{Var}_{P_{s_{h}^{k},a_{h}^{k},h}}\Big(\big(V_{h+1}^{k}\big)^{2}\Big)\log\frac{1}{\delta'}}+5 H^{2}\log\frac{1}{\delta'} \label{eq:boundt8-intermediate}\\
    &\leq \sqrt{32 H^2 \sum_{k,h}\mathsf{Var}_{P_{s_{h}^{k},a_{h}^{k},h}}\big(V_{h+1}^{k}\big)\log\frac{1}{\delta'}}+5 H^{2}\log\frac{1}{\delta'} \notag\\  
    &= \sqrt{32H^{2}T_{6}^{\ucb}\log\frac{1}{\delta'}}+5H^{2}\log\frac{1}{\delta'},
	\label{eq:boundt8}
\end{align}
where the second line is valid since $\sqrt{\max\{a,b\}}\leq \sqrt{a}+\sqrt{b}$,  the penultimate line follows from \eqref{eq:Var-P-Vk-square-H4}, and the last identity uses the definition \eqref{eq:defn-T6-proof} of $T_{6}^{\ucb}$.

%Substitution into \eqref{eq:boundt5-intermediate} and \eqref{eq:boundt6-intermediate} establishes \eqref{eq:boundt56}. 
%

\paragraph{Bounding the term $T_5$ (cf.~\eqref{eq:defn-T5-proof}).} 
Regarding the term $T_5$, expanding the variance terms gives
\begin{align}
T_{5}  %&=\sum_{k=1}^{K}\sum_{h=1}^{H}\mathsf{Var}_{\widehat{P}^k_{s_h^k,a_h^k,h}} \big( V_{h+1}^k \big) \notag\\
&=\sum_{k=1}^{K}\sum_{h=1}^{H}\left(\Big\langle\widehat{P}_{s_{h}^{k},a_{h}^{k},h}^{k},\big(V_{h+1}^{k}\big)^{2}\Big\rangle-\big(\big\langle\widehat{P}_{s_{h}^{k},a_{h}^{k},h}^{k},V_{h+1}^{k}\big\rangle\big)^{2}\right)\nonumber\\
 & =\sum_{k=1}^{K}\sum_{h=1}^{H}\Big\langle\widehat{P}_{s_{h}^{k},a_{h}^{k},h}^{k}-P_{s_{h}^{k},a_{h}^{k},h},\big(V_{h+1}^{k}\big)^{2}\Big\rangle \arxiv{ \notag\\
 &\qquad}  +\sum_{k=1}^{K}\sum_{h=1}^{H}\Big\langle P_{s_{h}^{k},a_{h}^{k},h}-e_{s_{h+1}^{k}},\big(V_{h+1}^{k}\big)^{2}\Big\rangle \nonumber\\
 & \qquad +\sum_{k=1}^{K}\sum_{h=1}^{H}\big(V_{h+1}^{k}(s_{h+1}^{k})\big)^{2}-\sum_{k=1}^{K}\sum_{h=1}^{H}\big(\big\langle\widehat{P}_{s_{h}^{k},a_{h}^{k},h}^{k},V_{h+1}^{k}\big\rangle\big)^{2}\notag\\
 & =  W_3+T_{8}+\sum_{k=1}^{K}\sum_{h=1}^{H}\big(V_{h+1}^{k}(s_{h+1}^{k})\big)^{2}-\sum_{k=1}^{K}\sum_{h=1}^{H}\big(\big\langle\widehat{P}_{s_{h}^{k},a_{h}^{k},h}^{k},V_{h+1}^{k}\big\rangle\big)^{2}\notag\\ 
 & \leq W_3+T_{8}+\sum_{k=1}^{K}\sum_{h=2}^{H}\big(V_{h}^{k}(s_{h}^{k})\big)^{2}-\sum_{k=1}^{K}\sum_{h=2}^{H}\big(\big\langle\widehat{P}_{s_{h}^{k},a_{h}^{k},h}^{k},V_{h+1}^{k}\big\rangle\big)^{2},
 \label{eq:boundt5-intermediate-159}
\end{align}
where $W_3$ and $T_8$ are defined in \eqref{eq:defn-T7-proof} and \eqref{eq:defn-T8-proof}, respectively. 
Here, in the last line we reindex the first sum (by replacing $h+1$ with $h$) and make use of the fact that $V_{H+1}^k=0$.

To continue, we make the observation that
\begin{align*}
\arxiv{&} \big(V_{h}^{k}(s_{h}^{k})\big)^{2}-\big(\big\langle\widehat{P}_{s_{h}^{k},a_{h}^{k},h}^{k},V_{h+1}^{k}\big\rangle\big)^{2} \arxiv{ \notag\\}
	&\arxiv{\qquad} =\Big(V_{h}^{k}(s_{h}^{k})+\big\langle\widehat{P}_{s_{h}^{k},a_{h}^{k},h}^{k},V_{h+1}^{k}\big\rangle\Big)\Big(V_{h}^{k}(s_{h}^{k})-\big\langle\widehat{P}_{s_{h}^{k},a_{h}^{k},h}^{k},V_{h+1}^{k}\big\rangle\Big)\\
 &\arxiv{\qquad} \leq2H\max\left\{ V_{h}^{k}(s_{h}^{k})-\big\langle\widehat{P}_{s_{h}^{k},a_{h}^{k},h}^{k},V_{h+1}^{k}\big\rangle,\,0\right\} \\
 & \arxiv{\qquad} = 2H\max\left\{ Q_{h}^{k}(s_{h}^{k},a_{h}^{k})-\big\langle\widehat{P}_{s_{h}^{k},a_{h}^{k},h}^{k},V_{h+1}^{k}\big\rangle,\,0\right\},
\end{align*}
where the inequality holds since $0\leq V_h^k(s)\leq H$ for all $s\in \cS$, and the last identity holds since, due to the greedy policy choice, $V_{h}^{k}(s_{h}^{k})=\max_a Q_{h}^{k}(s_{h}^{k},a) = Q_{h}^{k}(s_{h}^{k},a_{h}^{k})$. Substitution back into \eqref{eq:boundt5-intermediate-159} yields
\begin{align}
%\eqref{eq:boundt5-intermediate-159} 
T_5 
& \leq W_3+T_{8}+2H\sum_{k=1}^{K}\sum_{h=2}^{H}\max\Big\{ Q_{h}^{k}(s_{h}^{k},a_{h}^{k})-\big\langle\widehat{P}_{s_{h}^{k},a_{h}^{k},h}^{k},V_{h+1}^{k}\big\rangle,0\Big\}\nonumber\\
 & \leq W_3+T_{8}+2H\sum_{k=1}^{K}\sum_{h=2}^{H}b_{h}^{k}(s_{h}^{k},a_{h}^{k})+2H\sum_{k=1}^{K}\sum_{h=2}^{H}r_{h}(s_{h}^{k},a_{h}^{k})
 \notag
 %\label{eq:boundt5-intermediate-13}
 \\
 & \leq W_3+T_{8}+2HT_{2}+2KH^{2},
\notag
% \label{eq:boundt5-intermediate}
\end{align}
where the second line is a direct consequence of the update rule \eqref{eq:updateqv-ucbvi}, 
and the last line uses the definition \eqref{eq:boundt2} of $T_2$ combined with the assumption that $r_h(s,a)\leq 1$.  Taking this collectively with \eqref{eq:boundt8} establishes the claimed bound~\eqref{eq:boundt5}.

\paragraph{Bounding the term $T_6^{\mathsf{low}}$ (cf.~\eqref{eq:defn-T6-varpi-proof}).}
%

%Now consider a sample trajectory $(s_1,a_1,\dots,s_H,a_H)$ generated by executing policy $\pi$ in the MDP. Then it follows from Lemma~\ref{lem:cum-var-ub-Hsquare-ucbvi-mean} that
%
%\begin{align}
%\sum_{h=1}^{H}\mathop{\mathbb{E}}\limits\Big[\mathsf{Var}_{P_{s_{h},a_{h},h}}\big(V_{h+1}^{\pi}\big)\Big] 
%& =\sum_{h=1}^{H}\mathop{\mathbb{E}}\limits\Bigg[\Bigg(\Big[V_{h+1}^{\pi}(s_{h+1})-V_{h}^{\pi}(s_{h})+r_{h}\big(s_{h},a_{h}\big)\Big]\Bigg)^{2}\Bigg] \notag\\
% & \overset{\text{(i)}}{=}\mathop{\mathbb{E}}\limits\Bigg[\Bigg(\sum_{h=1}^{H}\Big[V_{h+1}^{\pi}(s_{h+1})-V_{h}^{\pi}(s_{h})+r_{h}\big(s_{h},a_{h}\big)\Big]\Bigg)^{2}\Bigg] \notag\\
% & \overset{\text{(ii)}}{=}\mathop{\mathbb{E}}\limits\Bigg[\Bigg(\sum_{h=1}^{H}r_{h}\big(s_{h},a_{h}\big)-V_{1}^{\pi}(s_{1})\Bigg)^{2}\Bigg] \notag\\
% & \overset{\text{(iii)}}
%\leq H^{2},
% \label{eq:cum-var-ub-Hsquare-ucbvi}
%\end{align}
%
%where the expectation is taken over the randomness of the sample trajectory. 
%Here, (i) follows from the Markov property of the trajectory, which ensures that the cross terms vanish; (ii) follows from telescoping together with the condition $V_{H+1}^{\pi}=0$; and (iii) holds because both $\sum_{h=1}^{H} r_h(s_h,a_h)$ and $V_1^{\pi}(s_1)$ reside within the interval $[0,H]$.
%
%Returning to our setting, 
Let $\mathcal{F}_k$ denote the $\sigma$-field generated by all events prior to episode $k$, and define
\[
    U_k \coloneqq \sum_{h=1}^H\mathsf{Var} _{P_{s_h^k,a_h^k,h}}\big( V_{h+1}^{\pi^k} \big).
\]
It then follows directly from Lemma~\ref{lem:cum-var-ub-Hsquare-ucbvi-mean} that
\[
\mathbb{E}[U_k \mid \mathcal{F}_k] \leq H^2. 
\]
Moreover, since $\|V_{h+1}^{\pi}\|_{\infty}\le H$ for every policy $\pi$, we trivially have $U_k \le H^3$. Applying the Azuma--Hoeffding inequality (Lemma~\ref{lem:azuma-hoeffding}) therefore yields that, with probability at least $1-2\delta'$,
\[
    T_6^{\mathsf{low}}=\sum_{k=1}^K U_k \leq 
    \mathbb{E} \bigg[\sum_{k=1}^K U_k \bigg] + \sqrt{2H^6K \log\frac{1}{\delta'}}
    \leq KH^2 + \sqrt{2H^6K \log\frac{1}{\delta'}}. 
\]

%\begin{align}
%        T_{6}^{\mathsf{low}} &\coloneqq \sum_{k=1}^K \sum_{h=1}^H\mathsf{Var} _{P_{s_h^k,a_h^k,h}}\big( V_{h+1}^{\pi^k} \big) ,
%\end{align}

\paragraph{Bounding the term $T_6^{\mathsf{diff}}$ (cf.~\eqref{eq:defn-T6-vardif-proof}).}
Given that $H\geq V_{h+1}^k(s)\geq V_{h+1}^{\star}(s)\geq V_{h+1}^{\pi^k}(s)\geq 0$ for all $s\in \cS$ (see Lemma~\ref{lemma:opt}), we can derive
\begin{align}
        	T_{6}^{\mathsf{diff}} &= \sum_{k=1}^K \sum_{h=1}^H \Big\langle P_{s_h^k,a_h^k,h}, \big( V_{h+1}^k-V_{h+1}^{\pi^k} \big)^2\Big\rangle\notag\\
            &\leq H\sum_{k=1}^K \sum_{h=1}^H \Big\langle P_{s_h^k,a_h^k,h},  V_{h+1}^k-V_{h+1}^{\pi^k} \Big\rangle \notag\\
            &=H \underset{\eqqcolon\,J_1}{\underbrace{ \sum_{k=1}^K \sum_{h=1}^H \Big\langle P_{s_h^k,a_h^k,h}-e_{s_{h+1}^k},  V_{h+1}^k-V_{h+1}^{\pi^k} \Big\rangle }}
         \arxiv{   \notag\\
            &\qquad} +
            H\underset{\eqqcolon\, J_2}{\underbrace{\sum_{k=1}^K \sum_{h=1}^H \Big(  V_{h+1}^k\big(s_{h+1}^k\big)-V_{h+1}^{\pi^k}\big(s_{h+1}^k\big) \Big) }}, 
            \label{eq:T6-diff-expand-ucbvi}
\end{align}
leaving us with two terms to control.

\begin{itemize}
	\item
The first term $J_1$ in \eqref{eq:T6-diff-expand-ucbvi} can be controlled using the same argument as for $T_8$. Repeating the arguments in \eqref{eq:boundt8-intermediate} tells us that, with probability at least $1-2\delta'\log_2(KH)$, 
\begin{align}
	|J_1| 
    &\leq \sqrt{8 \sum_{k,h}\mathsf{Var}_{P_{s_{h}^{k},a_{h}^{k},h}^{k}}\big(V_{h+1}^{k}-V_{h+1}^{\pi^k}\big)\log\frac{1}{\delta'}}+5 H\log\frac{1}{\delta'} \notag\\  
	&\leq \sqrt{8 \sum_{k,h} \Big\langle {P_{s_{h}^{k},a_{h}^{k},h}^{k}}, \big(V_{h+1}^{k}-V_{h+1}^{\pi^k}\big)^2 \Big\rangle\log\frac{1}{\delta'}}+5 H\log\frac{1}{\delta'} \notag\\
	&= \sqrt{8 T_{6}^{\mathsf{diff}}\log\frac{1}{\delta'}}+5 H\log\frac{1}{\delta'} \notag\\
	& \leq \frac{1}{2H} T_{6}^{\mathsf{diff}} + 4H\log\frac{1}{\delta'} + 5 H\log\frac{1}{\delta'}  \notag\\
	&= \frac{1}{2H} T_{6}^{\mathsf{diff}} + 9 H\log\frac{1}{\delta'},
	\label{eq:boundt8-J1}
\end{align}
where the penultimate inequality applies the AM-GM inequality. 

	\item 
When it comes to the second term $J_2$ in \eqref{eq:T6-diff-expand-ucbvi}, it follows from \eqref{eq:suboptimality-h-ub1-ucbvi}
 that, with probability at least $1-9S^2AH^2K\delta'$, 
\begin{align}
	|J_2| 
	&\leq H T_0^{\mathsf{ub}}.
\end{align}
\end{itemize}
 The above two bounds taken collectively demonstrate that
 \[
	T_6^{\mathsf{diff}} \leq HJ_1+ HJ_2 \leq  \frac{1}{2} T_6^{\mathsf{diff}} + 9 H^2\log\frac{1}{\delta'} + H^2T_0^{\mathsf{ub}},
\]
thus implying that, with probability at least $1-10S^2AH^2K\delta'$, 
 \[
	T_6^{\mathsf{diff}} \leq    2H^2T_0^{\mathsf{ub}} + 18 H^2\log\frac{1}{\delta'}.
\]

\paragraph{Establishing the bounds~\eqref{eq:T56-low-simpler-ub-ucbvi}.} 
Under the condition~\eqref{eq:assumption-K-proof-ucbvi}, it follows from \eqref{eq:boundtpi} that
\begin{align*}
	T_{6}^{\mathsf{low}}\leq KH^{2}+\sqrt{2H^{6}K\log\frac{1}{\delta'}}\leq2KH^{2}.
	%\label{eq:T6-low-simpler-ub-ucbvi}
\end{align*}

Moreover, combining \eqref{eq:boundt5} with Lemma~\ref{lem:bound-W1-W2-ucbvi-proof} and \eqref{eq:T6ucb-ub-T6-low-T6-diff} (an inequality in the proof of Lemma~\ref{lem:bound-W1-W2-ucbvi-proof} in Section~\ref{sec:proof-lemma:bound-W1-W2-ucbvi-proof}) leads to
\begin{align*}
T_{5} & \leq2HT_{2}+2KH^{2}+\frac{1}{4}\sqrt{ SAH^{3}T_{6}^{\mathsf{low}} \iotab}+\frac{1}{4}\sqrt{ S^{2}AH^{3}T_{6}^{\mathsf{diff}} \iotab}+\frac{1}{4} S^{2}AH^{3} \iotab\notag\\
 & \overset{(\text{i})}{\leq} 2HT_{2}+2KH^{2}+\frac{1}{4}\sqrt{2 SAH^{5}K\iotab}+\frac{1}{4}\sqrt{ S^{2}AH^{3}T_{6}^{\mathsf{diff}} \iotab}+\frac{1}{4} S^{2}AH^{3}\iotab\notag\\
 & \overset{(\text{ii})}{\leq}\sqrt{ SAH^{3}T_{5} \iotab}+3KH^{2}+ \frac{5}{4}  SAH^{3} \iotab + \frac{1}{4}\sqrt{ S^{2}AH^{3}T_{6}^{\mathsf{diff}} \iotab}+\frac{1}{4} S^{2}AH^{3}\iotab\\
 & \overset{(\text{iii})}{\leq}\frac{1}{2}T_{5}+3KH^{2}+\frac{1}{4}\sqrt{ S^{2}AH^{3}T_{6}^{\mathsf{diff}} \iotab}+ 2 S^{2}AH^{3} \iotab,
\end{align*}
and as a result, 
\begin{align*}
\Longrightarrow \qquad 
	T_{5} & \leq6KH^{2}+\frac{1}{2}\sqrt{ S^{2}AH^{3}T_{6}^{\mathsf{diff}} \iotab}+ 4 S^{2}AH^{3} \iotab.	
	%\label{eq:T5-low-simpler-ub-ucbvi}
\end{align*}
Here, (i) follows from \eqref{eq:T6-low-simpler-ub-ucbvi}, (ii) invokes Lemma~\ref{lem:bound-T234} and the AM-GM inequality $\frac{1}{4}\sqrt{2 SAH^{5}K \iotab}\leq KH^2+ \frac{1}{4}  SAH^{3} \iotab$, while (iii) applies the AM-GM inequality $\sqrt{ SAH^{3}T_{5} \iotab}\leq \frac{1}{2}T_{5}+\frac{1}{2} SAH^{3} \iotab$.

\section{Appendix B: proof of auxiliary lemmas in Section~\ref{sec:proof-of-thm-mvp}}

\subsection{Proof of Lemma~\ref{lemma:key2}}\label{app:pfkey2}
Recalling the set~$\mathcal{C}$ defined in \eqref{eq:defn-C-choice}, it suffices to establish \eqref{eq:cardinality-C-ub-mvp}. To this end, set   
\begin{equation*}
	M = \log_2 K \qquad \text{and} \qquad N = SAH.
\end{equation*}
We begin by introducing the following collections of ``distinct'' profiles:
%\begin{subequations}
\begin{align*}
	\mathcal{C}^{\mathsf{distinct}}(l) 
	&\coloneqq \Big\{ \mathcal{I} = \{\mathcal{I}^1,\ldots,\mathcal{I}^l\} \mid \mathcal{I}^{1}\leq \cdots \leq\mathcal{I}^{l}  ,  \mathcal{I}^{\tau}\in \{0,1,\cdots,M\}^N  \arxiv{ \notag\\
	&\qquad\qquad\qquad\qquad } \text{ and } \mathcal{I}^{\tau}\neq \mathcal{I}^{\tau+1}  ~(\forall\tau) 
 \Big\} ;
	\\
	\mathcal{C}^{\mathsf{distinct}} &\coloneqq \bigcup_{l\geq 1}\mathcal{C}^{\mathsf{distinct}}(l). 
\end{align*}
%\end{subequations}
% 
In words, $\mathcal{C}^{\mathsf{distinct}}(l)$ consists of all non-decreasing paths of length $l$ in $\{0,1,\ldots,M\}^N$ whose points are all distinct, while $\mathcal{C}^{\mathsf{distinct}}$ collects all such paths of arbitrary length.

We first relate $|\mathcal{C}|$ to $\big|\mathcal{C}^{\mathsf{distinct}}\big|$. Define the projection operator $\mathsf{Proj}:\mathcal{C}\to \mathcal{C}^{\mathsf{distinct}}$ 
that maps each $\mathcal{I}\in \mathcal{C}$ to the path $\mathcal{I}^{\mathsf{distinct}}$ obtained by removing all repeated elements from $\mathcal{I}$. For each $\mathcal{I}^{\mathsf{distinct}}\in\mathcal{C}^{\mathsf{distinct}}$, consider its inverse image
\[
	\mathcal{B}(\mathcal{I}^{\mathsf{distinct}}) \coloneqq \big\{ \mathcal{I}\in \mathcal{C} \mid \mathsf{Proj}(\mathcal{I})=\mathcal{I}^{\mathsf{distinct}} \big\}.
\]
Since $\mathcal{I}^{\mathsf{distinct}}$ is a non-decreasing path with distinct points in $\{0,1,\ldots,M\}^N$, its length is at most $MN+1$; indeed, along such a path, the sum of the $N$ coordinates strictly increases at every step and ranges from $0$ to at most $MN$. Consequently, the cardinality of $\mathcal{B}(\mathcal{I}^{\mathsf{distinct}})$ is bounded by the number of nonnegative integer solutions to
\begin{align*}
	\sum_{i=1}^{MN+1} x_i = K \qquad \text{and} \qquad x_i\in \mathbb{N} ~~\text{for all } 1\leq i \leq MN+1\nonumber.
\end{align*}
Elementary combinatorial arguments then reveal that 
\begin{align*}
	\big| \mathcal{B}(\mathcal{I}^{\mathsf{distinct}})  \big | \leq 
\left( \begin{array}{c} K+MN \\ MN
\end{array}\right) 
%= \frac{(K+MN)!}{(MN)! K!}
\leq (K+MN)^{MN}\leq (2K)^{MN}
\end{align*}
for each $\mathcal{I}^{\mathsf{distinct}}$, 
provided that $K\geq MN=SAH\log_2K$. 
Summing over all $\mathcal{I}^{\mathsf{distinct}}\in\mathcal{C}^{\mathsf{distinct}}$ gives
\begin{equation}
	|\mathcal{C}|\leq 
    \sum_{\mathcal{I}^{\mathsf{distinct}}\in \mathcal{C}^{\mathsf{distinct}}} \big| \mathcal{B}(\mathcal{I}^{\mathsf{distinct}})  \big |
    \le \big|\mathcal{C}^{\mathsf{distinct}}\big|\cdot (2K)^{MN}. 
	\label{eq:C-Cdistinct}
\end{equation}

It remains to bound $|\mathcal{C}^{\mathsf{distinct}}|$. 
To do so, let us first look at the set $\mathcal{C}^{\mathsf{distinct}}(MN+1)$, 
as each path in $\mathcal{C}^{\mathsf{distinct}}$ cannot have length more than $MN+1$.  
For each $\mathcal{I}^{\mathsf{distinct}} = \{\widetilde{\mathcal{I}}^1,\widetilde{\mathcal{I}}^2,\ldots, \widetilde{\mathcal{I}}^{MN+1}\}\in \mathcal{C}^{\mathsf{distinct}}(MN+1)$, 
we necessarily have
\begin{itemize}
	\item	$\widetilde{\mathcal{I}}^1 = [0,0,\ldots,0]^{\top}$ and $\widetilde{\mathcal{I}}^{MN+1}=[M,M,\ldots, M]^{\top}$;
	\item for each $1\leq \tau \leq MN$, $\widetilde{\mathcal{I}}^{\tau}$ and $\widetilde{\mathcal{I}}^{\tau+1}$ differ in exactly one element (i.e., their Hamming distance is 1). 
\end{itemize}
In other words, each $\mathcal{I}^{\mathsf{distinct}}\in
\mathcal{C}^{\mathsf{distinct}}(MN+1)$ can be viewed as an $MN$-step path from
$[0,0,\ldots,0]^{\top}$ to $[M,M,\ldots,M]^{\top}$, with each step incrementing one of the $N$ coordinates.  Since there are at most $N$ choices at each step,  
\begin{align*}
	\big|\mathcal{C}^{\mathsf{distinct}}(MN+1)\big|\leq N^{MN}.
\end{align*}
Finally, every path $\mathcal{I}^{\mathsf{distinct}}\in\mathcal{C}^{\mathsf{distinct}}$ can be extended to some maximal path
$\widetilde{\mathcal{I}}^{\mathsf{distinct}}\in
\mathcal{C}^{\mathsf{distinct}}(MN+1)$. For each such maximal path, there are at most $2^{MN+1}$ ways to select a subsequence of its points. It thus follows that 
\begin{align*}
	\big|\mathcal{C}^{\mathsf{distinct}}\big| \leq  2^{MN+1} \big|\mathcal{C}^{\mathsf{distinct}}(MN+1)\big|\leq (2N)^{MN+1}, 
\end{align*}
which taken collectively with \eqref{eq:C-Cdistinct} establishes the advertised result
\begin{align*}
	|\mathcal{C}|\leq (2K)^{MN} \big|\mathcal{C}^{\mathsf{distinct}}\big|
	%\leq (2K)^{MN} 2^{MN+1} \big|\mathcal{C}^{\mathsf{distinct}}(MN+1)\big|
	\leq (4KN)^{MN+1} \leq (4KN)^{MN+1}.
\end{align*}

\subsection{Proof of Lemma~\ref{lemma:uniform}}\label{app:pfuniform}
We begin by fixing a total profile $\mathcal{I}\in\mathcal{C}$, an integer $l$ satisfying $2\leq l\leq\log_2K+1$, and a feasible sequence $\{X_{h,s,a}\}$. Recall that
\begin{itemize}
\item[(i)] $\widehat{P}_{s,a,h}^{(l)}$ is constructed from
the $l$-th batch of data, which consists of $2^{l-2}$ independent samples from $\mathcal{D}^{\mathsf{expand}}$ (see Definition~\ref{def:filt2});

\item[(ii)] each $X_{h+1,s,a}$ is a deterministic mapping of $\mathcal{I}$ and the empirical transition models associated with steps $h'\in[h+1,H]$.
\end{itemize}
Consequently,
for each $(s,a,h)$,
the inner product 
$\big\langle
\widehat{P}_{s,a,h}^{(l)}-P_{s,a,h},
X_{h+1,s,a}
\big\rangle$
can be expressed as the sum of $2^{l-2}$ centered random variables,
each bounded in magnitude by $H/2^{l-2}$. Thus, in the sum of interest  $\sum_{s,a,h}\big\langle \widehat{P}_{s,a,h}^{(l)}-P_{s,a,h}, X_{h+1,s,a} \big\rangle$, by ordering these
random variables backward in $h$ (from $H$ to $1$),
and arbitrarily over $(s,a)$ and the samples within each batch, we obtain
a martingale difference sequence with total conditional variance
\[
W = \frac{1}{2^{l-2}}
\sum_{s,a,h}
\mathsf{Var}_{P_{s,a,h}}\big(X_{h+1,s,a}\big).
\]
Additionally, since $\|X_{h+1,s,a}\|_{\infty}\leq H$, we have
the crude upper bound
\[
    W\leq \frac{SAH^3}{2^{l-2}} \eqqcolon \sigma^2. 
\]
We can therefore apply Lemma~\ref{lem:Freedman-prelim} with
$K_0=\log_2(SAHK)$. Using $2^{l-2}\leq K/2$ together with the elementary
inequality $\sqrt{\max\{a,b\}}\leq\sqrt{a}+\sqrt{b}$, we obtain, with
probability at least $1-\delta'$,
\begin{align}
	&\sum_{s,a,h}
	\big\langle
	\widehat{P}_{s,a,h}^{(l)}-P_{s,a,h},
	X_{h+1,s,a}
	\big\rangle
	\notag\\
	&\qquad
	\leq
	\sqrt{
		\frac{8}{2^{l-2}}
		\sum_{s,a,h}
		\mathsf{Var}_{P_{s,a,h}}
		\big(X_{h+1,s,a}\big)
		\log\frac{2\log_{2}(SAHK)}{\delta'}
	}
	\arxiv{\notag\\
	&\qquad\qquad}
	+\frac{4H}{2^{l-2}}
	\log\frac{2\log_{2}(SAHK)}{\delta'}.
	\label{eq:xx1-aux}
\end{align}

We next make this bound uniform over all admissible choices. For any profile $\mathcal{I}$, Assumption~\ref{assumption-X-mvp} ensures that each $X_{h,s,a}$ has at most $K+1$ possible choices. Hence, the number of feasible sequences $\{X_{h,s,a}\}_{(s,a,h)\in\mathcal{S}\times\mathcal{A}\times[H]}$ is at most $(K+1)^{SAH}\leq (2K)^{SAH}$. 
Moreover, Lemma~\ref{lemma:key2} ensures that the number of admissible total profiles $\mathcal{I}$ is at most $(4SAHK)^{2SAH\log_2K}$. Taking a union bound over all profiles, all feasible sequences, and all $2\leq l\leq\log_2K+1$, and replacing $\delta'$ in \eqref{eq:xx1-aux} by
$$\frac{\delta'}{\big((4SAHK)^{2SAH\log_{2}K} (2K)^{SAH}\log_2K\big)},$$ 
we conclude that, with probability at least $1-\delta'$, 
\begin{align}
 & %\Bigg|
	\sum_{s,a,h}\big\langle \widehat{P}_{s,a,h}^{(l)}-P_{s,a,h}, X_{h+1,s,a} \big\rangle
	%\Bigg|
	\notag\\
	& \leq\sqrt{\frac{8}{2^{l-2}}\sum_{s,a,h}\mathsf{Var}_{P_{s,a,h}}\big(X_{h+1,s,a}\big)B_0}+\frac{4H}{2^{l-2}} B_0
    %\left(2SAH\log_{2}K\log(4SAHK)+SAH\log (2K)+\log\frac{3\log_{2}^{2}(SAHK)}{\delta'}\right)
    \notag\\
 & \leq\sqrt{\frac{8}{2^{l-2}}\sum_{s,a,h}\mathsf{Var}_{P_{s,a,h}}\big(X_{h+1,s,a}\big)\left(6SAH\log_{2}^{2}K+\log\frac{1}{\delta'}\right)}
	\arxiv{ \notag\\
    &\qquad} +\frac{4H}{2^{l-2}}\left(6SAH\log_{2}^{2}K+\log\frac{1}{\delta'}\right) 
    \notag\\
    & \leq\sqrt{\frac{SAH\iotab}{2^{l}\log_2K}\sum_{s,a,h}\mathsf{Var}_{P_{s,a,h}}\big(X_{h+1,s,a}\big)}
	\arxiv{ \notag\\
    &\qquad} +\frac{SAH^2\iotab}{2^{l}\log_2K}
	\label{eq:xx1-aux-123}
\end{align}
holds simultaneously for all $\mathcal{I}\in \mathcal{C}$, all $2\leq l\leq\log_{2}K+1$, and all feasible sequences $\{X_{h,s,a}\}_{(s,a,h)\in\mathcal{S}\times\mathcal{A}\times[H]}$. Here, we recall the definition of $\iotab$ in \eqref{eq:assumption-K-proof-B-mvp}, and in the second line above we set 
\[
B_0 = 2SAH(\log_{2}K)\log(4SAHK)+SAH\log (2K)+\log\frac{2\log_{2}^{2}(SAHK)}{\delta'}.
\]

It remains to establish the corresponding bound for the positive parts.
To this end, recall that $0\in\mathcal{X}_{h+1,\mathcal{I}}$.
Given any feasible sequence $\{X_{h,s,a}\}$, define another feasible
sequence $\{\widetilde{X}_{h+1,s,a}\}$ by
\[
\widetilde{X}_{h+1,s,a}
=
\begin{cases}
X_{h+1,s,a},
& \text{if }
\big\langle
\widehat{P}_{s,a,h}^{(l)}-P_{s,a,h},
X_{h+1,s,a}
\big\rangle\geq 0,\\
0,
& \text{otherwise}.
\end{cases}
\]
By construction,
\begin{align*}
&\sum_{s,a,h}
\max\Big\{
\big\langle
\widehat{P}_{s,a,h}^{(l)}-P_{s,a,h},
X_{h+1,s,a}
\big\rangle,0
\Big\}
=
\sum_{s,a,h}
\big\langle
\widehat{P}_{s,a,h}^{(l)}-P_{s,a,h},
\widetilde{X}_{h+1,s,a}
\big\rangle.
\end{align*}
Since \eqref{eq:xx1-aux-123} holds uniformly over all feasible
sequences, it applies in particular to
$\{\widetilde{X}_{h+1,s,a}\}$. 
This yields the desired positive-part bound and completes the proof.

%Finally, recalling our assumption $0\in \mathcal{X}_{h+1,\mathcal{I}}$, we see that  for every total profile $\mathcal{I}$ and its associated feasible sequence  $\{X_{h,s,a}\}$, 
%%
%\begin{align*}
%	&\sum_{s,a,h}\max\Big\{\big\langle\widehat{P}_{s,a,h}^{(l)}-P_{s,a,h},X_{h+1,s,a}\big\rangle,0\Big\}  \notag\\
%    &\qquad \in\bigg\{ \sum_{s,a,h}\big\langle\widehat{P}_{s,a,h}^{(l)}-P_{s,a,h},\widetilde{X}_{h+1,s,a}\big\rangle \,\Big|\, \widetilde{X}_{h+1,s,a}\in\mathcal{X}_{h+1,\mathcal{I}}, \forall (s,a,h)\bigg\} 
%\end{align*}
%%
%holds true. Consequently, the uniform upper bound on the right-hand side of \eqref{eq:xx1-aux-123} continues to be a valid upper bound on  $\sum_{s,a,h}\max\big\{\big\langle\widehat{P}_{s,a,h}^{(l)}-P_{s,a,h},X_{h+1,s,a}\big\rangle,0\big\}$. This completes the proof. 

\subsection{Proof of Lemma~\ref{lemma:decouple}}\label{app:pfdecouple}

We begin with the following claim, whose proof is deferred to the end of this subsection.   
\begin{claim}
	\label{claim:PV-l-UB}
	With probability exceeding $1-\delta'$, 
\begin{align}
 & \sum_{s,a,h}\Big\langle\widehat{P}_{s,a,h}^{(l)}-P_{s,a,h},V_{h+1}^{k_{l,j,s,a,h}}\Big\rangle \arxiv{\notag\\
 & \qquad}\leq\sqrt{\frac{SAH\iotab}{2^{l}\log_2K}\sum_{s,a,h}\mathsf{Var}_{P_{s,a,h}}\big(V_{h+1}^{k_{l,j,s,a,h}}\big)}  +\frac{SAH^2\iotab}{2^{l}\log_2K}
	\label{eq:claim-PV-l-UB}
\end{align}
holds simultaneously for all $l=1,\ldots,\log_{2}K$ and all $j=1,\ldots,2^{l-1}$.  Here, we recall that $k_{l,j,s,a,h}$ denotes the
episode index corresponding to the $(2^{l-1}+j)$-th visit to $(s,a,h)$
during the online learning process.
\end{claim}

Assuming the validity of  Claim~\ref{claim:PV-l-UB} for the moment, we combine it with
the decomposition \eqref{eq:PV-sum-decompose} to obtain
\begin{align*}
 & \sum_{k=1}^{K}\sum_{h=1}^{H}\Big\langle\widehat{P}_{s_{h}^{k},a_{h}^{k},h}^{k}-P_{s_{h}^{k},a_{h}^{k},h},V_{h+1}^{k}\Big\rangle \arxiv{ \notag\\
 &} \leq\sum_{l=1}^{\log_{2}K}\sum_{j=1}^{2^{l-1}}\sum_{s,a,h}\Big\langle\widehat{P}_{s,a,h}^{(l)}-P_{s,a,h},V_{h+1}^{k_{l,j,s,a,h}}\Big\rangle+SAH^{2}\\
 & \leq\sum_{l=1}^{\log_{2}K}\sum_{j=1}^{2^{l-1}}\sqrt{\frac{SAH\iotab}{2^{l}\log_2K}\sum_{s,a,h}\mathsf{Var}_{P_{s,a,h}}\big(V_{h+1}^{k_{l,j,s,a,h}}\big)}\arxiv{  \\
 & \qquad\qquad } +\sum_{l=1}^{\log_{2}K}\sum_{j=1}^{2^{l-1}}\frac{SAH^2\iotab}{2^{l}\log_2K}+SAH^{2}\\
 & \overset{\mathrm{(i)}}{\leq}\sqrt{SAH\iotab\sum_{l=1}^{\log_{2}K}\sum_{j=1}^{2^{l-1}}\sum_{s,a,h}\mathsf{Var}_{P_{s,a,h}}\big(V_{h+1}^{k_{l,j,s,a,h}}\big)}+\frac{1}{2}SAH^{2}\iotab + SAH^2\\
 & \overset{\mathrm{(ii)}}{\leq}\sqrt{SAH\iotab \sum_{k=1}^{K}\sum_{h=1}^{H}\mathsf{V}_{P_{s_{h}^{k},a_{h}^{k},h}}\big(V_{h+1}^{k}\big)}  \arxiv{\notag\\
 &\qquad \qquad} +SAH^{2}\iotab.
\end{align*}
Here, (i) follows from the
Cauchy-Schwarz inequality;  
and (ii) is valid due to our notational convention that $V_{h+1}^k=0$ ($\forall k > K$) and the identity
\begin{align}
 \arxiv{&}  \sum_{k=1}^{K}\sum_{h=1}^{H}\mathsf{Var}_{P_{s_{h}^{k},a_{h}^{k},h}}\big(V_{h+1}^{k}\big) \book{&} =\sum_{l=1}^{\log_{2}K}\sum_{s,a,h}\sum_{j=1}^{2^{l-1}}\mathsf{Var}_{P_{s,a,h}}\big(V_{h+1}^{k_{l,j,s,a,h}}\big)   \notag\\
 &\qquad +\sum_{k=1}^{K}\sum_{h=1}^{H}\mathds{1}\Big\{ N_{h}^{k,\mathsf{all}}(s_{h}^{k},a_{h}^{k})=0\Big\}\mathsf{Var}_{P_{s_{h}^{k},a_{h}^{k},h}}\big(V_{h+1}^{k}\big).\nonumber
\end{align}
This proves the desired bound on $\sum_{k,h}\big\langle\widehat{P}_{s_{h}^{k},a_{h}^{k},h}^{k}-P_{s_{h}^{k},a_{h}^{k},h},V_{h+1}^{k}\big\rangle$, 
provided that Claim~\ref{claim:PV-l-UB} is valid.

Before proceeding to the proof of Claim~\ref{claim:PV-l-UB}, 
we note that the same argument also controls
the two related quantities $\sum_{k,h}\max\big\{\big\langle\widehat{P}_{s_{h}^{k},a_{h}^{k},h}^{k}-P_{s_{h}^{k},a_{h}^{k},h},V_{h+1}^{k}\big\rangle,0\big\}$ 
and $\sum_{k,h}\big\langle\widehat{P}_{s_{h}^{k},a_{h}^{k},h}^{k}-P_{s_{h}^{k},a_{h}^{k},h},\big(V_{h+1}^{k}\big)^{2}\big\rangle$; we omit the nearly identical derivations for brevity. 
In particular, the latter quantity can be shown to satisfy 
\begin{align*} 
	\arxiv{&} \sum_{k=1}^{K}\sum_{h=1}^{H}\Big\langle\widehat{P}_{s_{h}^{k},a_{h}^{k},h}^{k}-P_{s_{h}^{k},a_{h}^{k},h},\big(V_{h+1}^{k}\big)^{2}\Big\rangle  \arxiv{\\}
 & \leq\sqrt{\frac{SAH\iotab}{4}\sum_{k=1}^{K}\sum_{h=1}^{H}\mathsf{Var}_{P_{s_{h}^{k},a_{h}^{k},h}}\Big(\big(V_{h+1}^{k}\big)^{2}\Big)} \arxiv{\notag\\
 &\qquad} +SAH^{3}\iotab\\
 & \leq \sqrt{SAH^3\iotab \sum_{k=1}^{K}\sum_{h=1}^{H}\mathsf{Var}_{P_{s_{h}^{k},a_{h}^{k},h}}\big(V_{h+1}^{k}\big)} \arxiv{\notag\\
 &\qquad} +SAH^{3}\iotab,
\end{align*}
where the last inequality follows from Lemma~\ref{lemma:sqv} and
the fact that $0\leq V_{h+1}^{k}(s)\le H$ for all $s\in\mathcal{S}$.

\begin{proof}[Proof of Claim~\ref{claim:PV-l-UB}]

To apply Lemma~\ref{lemma:uniform}, we need to construct the collections
$\{\mathcal{X}_{h,\mathcal{I}}\}$ so that they contain the value
estimates $\{V_h^k\}$ generated by the actual online process.  
To this end, for each
fixed total profile $\mathcal{I}\in\mathcal{C}$, we introduce an
auxiliary version of \mvp that operates on the expanded sample set
$\mathcal{D}^{\mathsf{expand}}$ while following the prescribed total profile
$\mathcal{I}$.  
Specifically, for each $\mathcal{I} \in \mathcal{C}$ (cf.~\eqref{eq:defn-C-choice}), consider the updates in Algorithm~\ref{alg:main-fixed-profile} operating upon the expanded sample set $\mathcal{D}^{\mathsf{expand}}$.  
\begin{algorithm}[ht]
	\DontPrintSemicolon
	\caption{Monotonic Value Propagation for a given total profile $\mathcal{I}\in \mathcal{C}$ ($\mathtt{MVP}(\mathcal{I})$) \label{alg:main-fixed-profile}}
	
	\textbf{initialization: } set $V_{H+1}^{k,\mathcal{I}}(s)\leftarrow H$ for all $s\in \mathcal{S}$ and $1\leq k\leq K$. \\
	\For{$k=1,2,\ldots,K$} {
			\For{$h=H,H-1,...,1$} {
				\For{$(s,a)\in \mathcal{S}\times \mathcal{A}$} { \vspace{-4ex}
			\begin{align*} 
						j & \leftarrow I_{s,a,h}^k,~~ n \leftarrow 2^{j-2}, \\
						b_h(s,a) &\leftarrow c_1 \sqrt{\frac{   \mathsf{ Var}_{\widehat{P}^{(j)}_{s,a,h} }\big(V_{h+1}^{k,\mathcal{I}}\big) \log \frac{1}{\delta'}  }{ \max\{n,1 \} }}+c_2\frac{H\log \frac{1}{\delta'}}{ \max\{n,1\}  },  
						\\
						Q_h^{k,\mathcal{I}}(s,a) &\leftarrow \min\Big\{    r_h(s,a)+\langle \widehat{P}^{(j)}_{s,a,h}, V_{h+1}^{k,\mathcal{I}} \rangle +b_h(s,a)    ,H\Big\}, \\
						V^{k,\mathcal{I}}_{h}(s) &\leftarrow \max_{a}Q^{k,\mathcal{I}}_{h}(s,a).
					\end{align*}
					\vspace{-3ex}
				}
			}
	}
\end{algorithm}

Now for each $h\in[H]$ and $\mathcal{I}\in\mathcal{C}$, define
\begin{align}
	\mathcal{X}_{h,\mathcal{I}} \coloneqq \Big\{ V_h^{k,\mathcal{I}} \mid 1\leq k\leq K\Big\} \cup \{0\}, \qquad \forall h\in [H] \text{ and }\mathcal{I}\in \mathcal{C}. 
	\label{eq:X-I-ours}
\end{align}
By construction, the collections
$\{\mathcal{X}_{h,\mathcal{I}}\}$ satisfy
Assumption~\ref{assumption-X-mvp}.   
Hence, Lemma~\ref{lemma:uniform}
implies that
\begin{align}
 & \sum_{s,a,h}\Big\langle\widehat{P}_{s,a,h}^{(l)}-P_{s,a,h},X_{h+1,s,a}\Big\rangle \arxiv{\notag\\
 & \qquad}\leq\sqrt{\frac{SAH\iotab}{2^{l}\log_2K}\sum_{s,a,h}\mathsf{Var}_{P_{s,a,h}}\big(X_{h+1,s,a}\big)} +\frac{SAH^2\iotab}{2^{l}\log_2K}
	\label{eq:claim-PV-l-UB-temp}
\end{align}
simultaneously for all $l=1,\ldots,\log_{2}K$, all $\mathcal{I}\in \mathcal{C}$, and all sequences $\{X_{h,s,a}\}$ obeying $X_{h,s,a}\in \mathcal{X}_{h,\mathcal{I}}$, $\forall (s,a,h)$.

It remains to specialize this uniform bound to the profile $\mathcal{I}^{\mathsf{true}}$ generated by
the actual online process.  Under the coupling between $\mathcal{D}^{\mathsf{expand}}$ and $\mathcal{D}^{\mathsf{orig}}$ described after Definition~\ref{def:filt2}, the value function estimates produced by the online process coincide with those generated by the auxiliary algorithm under the profile $\mathcal{I}^{\mathsf{true}}$, and hence
\begin{equation}
	V_{h}^{k} = V_{h}^{k,\mathcal{I}^{\mathsf{true}}} \in \mathcal{X}_{h,\mathcal{I}^{\mathsf{true}}},\qquad 1\leq k \leq K. 
	\label{eq:Vk-Itrue}
\end{equation}
The claim then follows directly from \eqref{eq:Vk-Itrue} and the uniform bound \eqref{eq:claim-PV-l-UB-temp}. 
\end{proof}

\subsection{Proof of Lemma~\ref{lem:bound-T6-ucb-mvp}}
\label{sec:pflem:bound-T6-ucb-mvp}

%
%Applying an analogous decomposition as in the analysis of $T_5$ (see Chapter~\ref{sec:pflem:bound-T56}) to $T_6^{\ucb}$, we can readily derive

We begin by decomposing  $T_6^{\ucb}$ (cf.~\eqref{eq:defn-T6-proof-mvp}) as follows
\begin{align}
T_{6}^{\ucb}
& 
=\sum_{k=1}^{K}\sum_{h=1}^{H}\Big\langle P_{s_{h}^{k},a_{h}^{k},h},\big(V_{h+1}^{k}\big)^{2}\Big\rangle-\sum_{k=1}^{K}\sum_{h=1}^{H}\Big(\big\langle P_{s_{h}^{k},a_{h}^{k},h},V_{h+1}^{k}\big\rangle\Big)^{2}\nonumber\\
& =\sum_{k=1}^{K}\sum_{h=1}^{H}\Big\langle P_{s_{h}^{k},a_{h}^{k},h}-e_{s_{h+1}^{k}},\big(V_{h+1}^{k}\big)^{2}\Big\rangle+\sum_{k=1}^{K}\sum_{h=1}^{H}\big(V_{h+1}^{k}(s_{h+1}^{k})\big)^{2} \arxiv{ \notag\\
 &\qquad } -\sum_{k=1}^{K}\sum_{h=1}^{H}\Big(\big\langle P_{s_{h}^{k},a_{h}^{k},h},V_{h+1}^{k}\big\rangle\Big)^{2}\nonumber\\
 & \overset{\mathrm{(i)}}{=}\sum_{k=1}^{K}\sum_{h=1}^{H}\Big\langle P_{s_{h}^{k},a_{h}^{k},h}-e_{s_{h+1}^{k}},\big(V_{h+1}^{k}\big)^{2}\Big\rangle+\sum_{k=1}^{K}\sum_{h=2}^{H}\big(V_{h}^{k}(s_{h}^{k})\big)^{2} \arxiv{ \notag\\
 &\qquad } -\sum_{k=1}^{K}\sum_{h=1}^{H}\Big(\big\langle P_{s_{h}^{k},a_{h}^{k},h},V_{h+1}^{k}\big\rangle\Big)^{2}\nonumber\\
 & \overset{\mathrm{(ii)}}{\le} T_{8}+2H\sum_{k=1}^{K}\sum_{h=1}^{H}\max\Big\{ V_{h}^{k}(s_{h}^{k})-\big\langle P_{s_{h}^{k},a_{h}^{k},h},V_{h+1}^{k}\big\rangle,0\Big\} \notag\\
 & \overset{\mathrm{(iii)}}{=} T_{8}+2H\sum_{k=1}^{K}\sum_{h=1}^{H}\max\Big\{ Q_{h}^{k}(s_{h}^{k},a_{h}^{k})-\big\langle P_{s_{h}^{k},a_{h}^{k},h},V_{h+1}^{k}\big\rangle,0\Big\}.
 \notag
 \end{align}
Here, (i) replaces $h+1$ with $h$ in the second sum and uses $V_{H+1}^k=0$; in (ii), we invokes the definition of $T_8$ in
\eqref{eq:defn-T8-proof} together with the elementary inequality
\begin{align*}
\arxiv{&} \big(V_{h}^{k}(s_{h}^{k})\big)^{2}-\Big(\big\langle P_{s_{h}^{k},a_{h}^{k},h},V_{h+1}^{k}\big\rangle\Big)^{2} \arxiv{ \\}
&\arxiv{ \qquad } =\Big(V_{h}^{k}(s_{h}^{k})-\big\langle P_{s_{h}^{k},a_{h}^{k},h},V_{h+1}^{k}\big\rangle\Big)\Big(V_{h}^{k}(s_{h}^{k})+\big\langle P_{s_{h}^{k},a_{h}^{k},h},V_{h+1}^{k}\big\rangle\Big)\\
 & \arxiv{\qquad} \leq2H\max\Big\{ V_{h}^{k}(s_{h}^{k})-\big\langle P_{s_{h}^{k},a_{h}^{k},h},V_{h+1}^{k}\big\rangle,0\Big\};
\end{align*}
and (iii) uses the greedy update  $V_h^k(s_h^k)=\max_a Q_h^k(s_h^k,a)=Q_h^k(s_h^k,a_h^k)$.

The remaining issue is that the update rule
\eqref{eq:updateqv-mvp} is expressed in terms of the empirical transition
kernel $\widehat P^k$, whereas the preceding display still involves the true
transition kernel $P$. To address this, we incorporate $\widehat{P}^k_{s_{h}^{k},a_{h}^{k},h}$ into the max operator to yield 
 \begin{align}
 T_6^{\ucb}& \leq T_{8}+2H\sum_{k=1}^{K}\sum_{h=1}^{H}\max\Big\{ Q_{h}^{k}(s_{h}^{k},a_{h}^{k})-\big\langle\widehat{P}^k_{s_{h}^{k},a_{h}^{k},h},V_{h+1}^{k}\big\rangle,0\Big\}   \notag\\
 &\qquad~ +2H\sum_{k=1}^{K}\sum_{h=1}^{H}\max\Big\{\big\langle\widehat{P}_{s_{h}^{k},a_{h}^{k},h}^{k}-P_{s_{h}^{k},a_{h}^{k},h},V_{h+1}^{k}\big\rangle,0\Big\}\nonumber\\
 & \leq T_{8}+2H\sum_{k=1}^{K}\sum_{h=1}^{H}b_{h}^{k}(s_{h}^{k},a_{h}^{k})+2H\sum_{k=1}^{K}\sum_{h=1}^{H}r_h(s_{h}^{k},a_{h}^{k})+2HW_4
 %\label{eq:boundt6-intermediate-135}
 \notag\\
 & \leq T_{8}+2HT_{2}+2KH^2+2HW_4,
 %\label{eq:boundt6-intermediate}
\end{align}
where the second inequality follows from the update rule
\eqref{eq:updateqv-mvp} together with the definition of $W_4$ in
\eqref{eq:defn-W_4-proof-mvp}.  Combining this inequality with the bound
\eqref{eq:boundt8} on $T_8$ gives
\begin{align*}
T_{6}^{\ucb}  &\leq 2HT_{2}+6KH^{2}
	+\sqrt{32H^{2}T_{6}^{\ucb}\log\frac{1}{\delta'}}+3H^{2}\log\frac{1}{\delta'}
	+2HW_4 \notag\\
    &\le 2HT_{2}+6KH^{2}
	+\frac{1}{2}T_{6}^{\ucb} + 16H^{2}\log\frac{1}{\delta'}+3H^{2}\log\frac{1}{\delta'}
	+2HW_4,
\end{align*}
where the last line applied the AM-GM inequality. Rearranging terms and using the definition \eqref{eq:assumption-K-proof-B-mvp} of $\iotab$ yield, with probability exceeding $1-4SAHK\delta'$,  
\begin{align*}
T_{6}^{\ucb} 
%& \leq 4HT_{2}
%+ 4HW_4
%    + 12KH^{2} +38H^2\log\frac{1}{\delta'} \notag\\
    & \leq 4HT_{2}
+ 4HW_4
    + KH^{2} \iotab.
    %\label{eq:T6-ub-analysis-mvp}
\end{align*}

\subsection{Solving the inequalities in \eqref{eq:T123456-ub-analysis-mvp}}
\label{sec:solve-all-inequalities-mvp}

We now solve the intertwined inequalities in \eqref{eq:T123456-ub-analysis-mvp}. The strategy is to successively eliminate the
auxiliary quantities $T_6^{\ucb}$ and $T_5$, after which the remaining bounds
follow immediately.

We first tackle $T_6^{\ucb}$. Substituting \eqref{eq:T1-W4-ub-analysis-mvp} into
\eqref{eq:T6-ub-analysis-mvp} and invoking the AM-GM inequality give
\begin{align*}
T_{6}^{\ucb} & \leq4HT_{2}+4H\sqrt{T_{6}^{\ucb}SAH\iotab}+4SAH^{3}\iotab+KH^{2}\iotab\\
 & \leq4HT_{2}+\frac{1}{2}T_{6}^{\ucb}+8SAH^{3}\iotab+4SAH^{3}\iotab+KH^{2}\iotab,
\end{align*}
which together with a little algebra yields
\begin{align}
T_{6}^{\ucb} & \leq8HT_{2}+24SAH^{3}\iotab+2KH^{2}\iotab.\label{eq:T6-W4-ub-analysis-mvp}
\end{align}

Next, we turn to $T_5$. It follows from \eqref{eq:T5-ub-analysis-mvp} and \eqref{eq:W3-ub-analysis-mvp}
that
\begin{align*}
T_{5} & \leq2H\sqrt{T_{6}^{\ucb}SAH\iotab}+2HT_{2}+SAH^{3}\iotab+KH^{2}\iotab\\
 & \leq\frac{1}{2}T_{6}^{\ucb}+2SAH^{3}\iotab+2HT_{2}+SAH^{3}\iotab+KH^{2}\iotab\\
 & \leq4HT_{2}+12SAH^{3}\iotab+KH^{2}\iotab+2SAH^{3}\iotab+2HT_{2}+SAH^{3}\iotab+KH^{2}\iotab,
\end{align*}
where the second inequality arises from the AM-GM inequality, and
the third inequality results from \eqref{eq:T6-W4-ub-analysis-mvp}.
Combine terms to arrive at
\begin{align}
T_{5} & \leq6HT_{2}+15SAH^{3}\iotab+2KH^{2}\iotab.\label{eq:T5-T6-ub-analysis-mvp}
\end{align}
In addition, it is also seen from \eqref{eq:T2-ub-analysis-mvp} and the AM-GM inequality that
\begin{equation}
T_{2}\le\frac{1}{12H}T_{5}+3SAH^{2}\iotab+\frac{1}{2}SAH^{2}\iotab,
\label{eq:T2-T5-ub-analysis-mvp}
\end{equation}
which combined with \eqref{eq:T5-T6-ub-analysis-mvp} reveals that
\begin{align}
T_{5} & \leq\frac{1}{2}T_{5}+36SAH^{3}\iotab+2KH^{2}\iotab, \arxiv{\notag \\}
\qquad \Longrightarrow\qquad T_{5} \arxiv{&} \leq 72SAH^{3}\iotab+4KH^{2}\iotab.\label{eq:T5-T6-T2-ub-analysis-mvp}
\end{align}

Armed with inequality \eqref{eq:T5-T6-T2-ub-analysis-mvp}, we can
immediately combine it with \eqref{eq:T2-ub-analysis-mvp} and \eqref{eq:T6-W4-ub-analysis-mvp}
to yield
\begin{align}
T_{2} & \lesssim\sqrt{SAH^{3}K}\iotab+SAH^{2}\iotab,\label{eq:T2-final-ub-analysis-mvp}\\
T_{6}^{\ucb} & \lesssim SAH^{3}\iotab+KH^{2}\iotab.\label{eq:T6-final-ub-analysis-mvp}
\end{align}
Substitution into \eqref{eq:T1-W4-ub-analysis-mvp} and \eqref{eq:T3-ub-analysis-mvp} then leads to
\begin{align*}
T_{1} + |T_{3}| & \lesssim\sqrt{SAH^{3}K}\iotab+SAH^{2}\iotab.
\end{align*}
Putting these two bounds collectively with \eqref{eq:T2-final-ub-analysis-mvp}
and \eqref{eq:T4-ub-analysis-mvp}, we reach
\begin{align*}
\mathsf{Regret}(K) &\leq T_{1}+T_{2}+T_{3}+T_{4} \arxiv{\notag\\
&} \lesssim\sqrt{SAH^{3}K}\iotab+SAH^{2}\iotab\asymp \sqrt{SAH^{3}K}\iotab,
\end{align*}
provided that $K\geq SAH$. 

%\paragraph{Proof of Lemma~\ref{lem:online-lb-testing}}
%

\section{Appendix C: proof of auxiliary lemmas in Section~\ref{sec:model-free-onlineRL}}

\subsection{Proof of Lemma~\ref{lemma:property_lr}}
\label{sec:proof-properties-learning-rates}

Let us first prove \eqref{eq:properties-learning-rates-square}. In view of the definitions~\eqref{eq:learning_rate_ucbq} and \eqref{equ:learning_rate_notation}, for any $1\le n\le N$,
\begin{align}
\eta_{n}^{N} & =\frac{H+1}{H+n}\prod_{i=n+1}^{N}\frac{i-1}{H+i}=\frac{(H+1)\prod_{i=n}^{N-1}i}{\prod_{i=n}^{N}(H+i)}\\
 & =\frac{H+1}{H+N}\prod_{i=n}^{N-1}\frac{i}{H+i}\le\frac{H+1}{H+N}\le\frac{2H}{N}.
\end{align}
This combined with the fact $\sum_{n=1}^N\eta_n^N=1$ (see \eqref{eq:sum-eta-n-N}) implies that
\begin{equation}
    \sum_{n=1}^N(\eta_n^N)^2
    \le \left(\max_{n:1\le n\le N}\eta_n^N\right)\sum_{n=1}^N\eta_n^N
    \le \frac{2H}{N}.
\end{equation}

We next verify \eqref{eq:properties-learning-rates-tail}. For any given $N\geq n\ge 1$, recall that
\begin{equation}
\eta_n^N=\frac{H+1}{H+n}a_N
\qquad \text{with }
    a_N \defn \begin{cases}\prod_{i=n+1}^N\frac{i-1}{H+i}, & \text{if }N>n, \\
    1, & \text{if }N=n.
    \end{cases}
\end{equation}
It is easily seen that
\begin{equation}
    a_N=\frac{H+N}{H}a_N-\frac{H+N+1}{H}a_{N+1},
\end{equation}
and hence the partial sum of $a_N$ telescopes (for any $L\geq n$):
\begin{equation}
    \sum_{N=n}^{L}a_N
    =\frac{H+n}{H}-\frac{H+L+1}{H}a_{L+1}.
\end{equation}
Recognizing that the last term $\frac{H+L+1}{H}a_{L+1}$ vanishes as $L\to\infty$, we arrive at
\begin{equation}
\sum_{N=n}^{\infty}\eta_n^N
=\sum_{N=n}^{\infty}
\frac{H+1}{H+n}a_N
=\frac{H+1}{H+n}\cdot \frac{H+n}{H}=1+\frac{1}{H}.
\end{equation}

It remains to establish property~\eqref{eq:properties-learning-rates-sqrt}. When $N =1$, this property holds trivially. 
Now suppose that $N\geq 2$. Making use of the basic relation $\eta_n^{N} = (1-\eta_N) \eta_n^{N-1}$ for all $1\leq n \leq N-1$, 
we observe
\begin{equation}
  \sum_{n=1}^{N}\frac{\eta_n^{N}}{\sqrt n} = \frac{\eta_N}{\sqrt N} +  (1-\eta_N) \sum_{n=1}^{N-1} \frac{\eta_n^{N-1}}{\sqrt n}.
	\label{eq:basic-identity-sum-eta}
\end{equation}
We prove the property by induction. Suppose for the moment that the property holds for $N-1$, i.e., 
\begin{equation}
	\frac{1}{\sqrt{N-1}} \le \sum_{n=1}^{N-1}\frac{\eta_n^{N-1}}{\sqrt n} \le \frac{2}{\sqrt{N-1}} .
	\label{eq:induction-hypothesis-N-1}
\end{equation}
Then it is readily seen from \eqref{eq:basic-identity-sum-eta} that
\begin{align}
  	\sum_{n=1}^{N}\frac{\eta_n^{N}}{\sqrt n}&  = \frac{\eta_N}{\sqrt N} +  (1-\eta_N) \sum_{n=1}^{N-1} \frac{\eta_n^{N-1}}{\sqrt n} \nonumber \\ 
	& \geq \frac{\eta_N}{\sqrt N} + \frac{1-\eta_N}{\sqrt{N -1}} \geq \frac{\eta_N}{\sqrt N} + \frac{1-\eta_N}{\sqrt N} =  \frac{1}{\sqrt N}, 
\end{align}
where the first inequality comes from \eqref{eq:induction-hypothesis-N-1}. Similarly,
\begin{align*}
	\sum_{n=1}^{N}\frac{\eta_{n}^{N}}{\sqrt n}&=\frac{\eta_{N}}{\sqrt N}+(1-\eta_{N})\sum_{n=1}^{N-1}\frac{\eta_{n}^{N-1}}{\sqrt n} \\
    & \leq\frac{\eta_{N}}{\sqrt N}+\frac{2(1-\eta_{N})}{\sqrt{N-1}} =\frac{H+1}{\sqrt N(H+N)}+\frac{2\sqrt{N-1}}{H+N}\\
    &\leq\frac{H+1}{\sqrt N(H+N)}+\frac{2\sqrt N}{H+N}\\
    & =\frac{1}{\sqrt N}\left(\frac{H+1}{H+N}+\frac{2N}{H+N}\right) \leq\frac{2}{\sqrt N},
\end{align*}
where the last inequality follows since $H\geq 1$. This completes the induction and finishes the proof.

\subsection{Proof of Lemma~\ref{lemma:opt_model_free}}
\label{sec:proof-lemma:opt_model_free}

Consider any given $(s,a,h)\in\cS\times\cA\times[H]$. 
 When $t=0$, the claim follows directly from $Q_h^k(s,a)=H$ (as no update has yet occurred),
$0\le Q_h^\star(s,a)\le H$ and $\eta_0^0=1$ (cf.~\eqref{equ:learning_rate_notation-eta_0n}). Therefore, it suffices to focus on the case $t\ge1$ in the remainder of this proof.

As before, let $k_i$ denote the index of the episode in which $(s,a)$ is visited at step $h$ for the $i$-th time. As can be readily seen, the sequence
\begin{equation}
    Z_i\defn (  P_h^{k_i}-P_{s,a,h})V_{h+1}^{\star}
\end{equation}
forms a martingale difference sequence satisfying  $|Z_i|\le H$. Therefore, by applying the Azuma-Hoeffding inequality (cf.~Lemma~\ref{lem:azuma-hoeffding}) and a union bound over all $(s,a,h)$ and all visit counts $1\le t\le K$, we can show that, with probability at least $1-\delta$,
\begin{equation}
\left|\sum_{i=1}^t\eta_i^t Z_i\right|
\le C_1H\sqrt{\iota\sum_{i=1}^t(\eta_i^t)^2}
\le C_2\sqrt{\frac{H^3\iota}{t}}
\label{eq:concentration-q-ucbq}
\end{equation}
holds simultaneously for all $(s,a,h,t)$, where $C_1,C_2>0$ are some universal constants, and the last inequality is due to \eqref{eq:properties-learning-rates-square}. On the same event, if the constant $c_{\mathsf{b}}$ in \eqref{eq:hoeffding_ucbq} is chosen sufficiently large, then it follows from \eqref{eq:properties-learning-rates-sqrt} that
\begin{equation}
\sum_{i=1}^t \eta_i^t b_h^{k_i}(s,a)
=c_{\mathsf{b}}\sqrt{H^3\iota}\sum_{i=1}^t\frac{\eta_i^t}{\sqrt{i}}
\ge c_{\mathsf{b}}\sqrt{\frac{H^3\iota}{t}} \geq 
\left|\sum_{i=1}^t\eta_i^t Z_i\right|.
\label{eq:bonus-dominates-noise-ucbq}
\end{equation}
Similar arguments combined with \eqref{eq:properties-learning-rates-sqrt}   also give
\begin{equation} \label{eq:beta_t}
%\beta_t=
2\sum_{i=1}^t\eta_i^t b_h^{k_i}(s,a)
\le 4c_{\mathsf{b}}\sqrt{\frac{H^3\iota}{t}}.
\end{equation}
Therefore, we can readily  establish~\eqref{eq:opt-upper-q-bound} by combining \eqref{eq:q-recursion-model-free}, \eqref{eq:bonus-dominates-noise-ucbq} and \eqref{eq:beta_t}.

Next, we prove the optimism property \eqref{eq:Q-optimism} by induction. The property holds trivially for the base case with $k=0$ due to the choice of initialization. Now, assume that  \eqref{eq:Q-optimism} holds for all episodes up to $k$. In view of the recursion \eqref{eq:q-recursion-model-free} and \eqref{eq:bonus-dominates-noise-ucbq}, one has
\begin{align*}
\big(Q_h^k-Q_h^{\star}\big)(s,a)
&\ge \eta_0^t\big(H-Q_h^{\star}(s,a)\big)
+\sum_{i=1}^t\eta_i^t\big(V_{h+1}^{k_i}-V_{h+1}^{\star}\big)(s_{h+1}^{k_i}) \ge 0.
\end{align*}
It then follows that
\begin{equation*}
\max_{a\in\cA}Q_h^k(s,a)\ge \max_{a\in\cA}Q_h^{\star}(s,a)=V_h^{\star}(s),
\end{equation*}
and the operation $V_h^k(s)=\min\{H,\max_aQ_h^k(s,a)\}$ preserves the inequality since $V_h^{\star}(s)\le H$. This completes the induction and hence the proof.

\section{Appendix D: proof of auxiliary lemmas in Section~\ref{sec:minimax-lower-bound-online}}
\label{sec:proof:lemmas-lower-bound-online-RL}

\subsection{Proof of Lemma~\ref{lem:Vh-star-Vhpi-gap-lower-bound-online-final}}
\label{sec:proof-lem:Vh-star-Vhpi-gap-lower-bound-online-final}

We begin by establishing several basic properties of the constructed MDPs, which will then be used to prove the desired lower bound.

\paragraph{Basic properties of these MDPs.}

We now isolate several basic properties that hold for an arbitrary MDP $\mathcal{M}\in \mathcal{C}_{\mathsf{mdp}}^{\mathsf{sub}}$. For notational convenience, we suppress the dependence of quantities (e.g., $Q_h^{\pi}$, $V_h^{\pi}$, $d_h^{\pi}$, $a_h^{\star,\mathcal{M}}$, $P_h^{\mathcal{M}}$) on the underlying MDP $\mathcal{M}$, as long as it is clear from the context.

\begin{itemize}
\item {\em Value functions and Q-functions.}
 Given that state 0 is absorbing with zero  rewards, one has
\begin{subequations}
\begin{align}
\label{eq:Qhpi-always-zero-lower-bound-online}
Q_h^{\pi}(0, 0) &= 0\qquad \text{for all }1\le h\leq H. 
\end{align}
Also, since $V_{H+1}^{\pi}(0)=V_{H+1}^{\pi}(1)=0$, we see from \eqref{eq:reward-lower-bound-online} that
\begin{align}
V_{H}^{\pi}(1) &=  \sum_{a\in \cA_1}\pi_H(a\mymid 1) r_H\big(1,a\big)= 1,\label{eq:VH-pi-lower-bound-online}\\
V_{h}^{\pi}(0) &= r_h(0,0)= 0\qquad \text{for all }h.
\end{align}
Moreover, for any $1\leq h\leq H$ and any $a\in \cA_1$, the Bellman equation (cf.~\eqref{eq:Bellman-matrix-finite-pi}) together with \eqref{eq:reward-lower-bound-online} and \eqref{eq:Qhpi-always-zero-lower-bound-online} gives  
\begin{align}
Q_h^{\pi}(1, a) &= r_h(1,a)+ P_h(1\mymid 1,a) V_{h+1}^{\pi}(1) + P_h(0\mymid 1,a) V_{h+1}^{\pi}(0) \notag\\
&= 1 + P_h(1\mymid 1,a) V_{h+1}^{\pi}(1).
\label{eq:Qh-1a-iteration-lower-bound-online}
\end{align}
\end{subequations}
According to the construction \eqref{eq:transition-kernel-M-lower-bound-online} and \eqref{eqn:prob-astar-a-1}, we have 
\begin{align}
P_h(1\mymid 1,a) \geq 1-1/H\qquad \text{for all }a\in \cA_1,
\label{eq:PhM-1a-lower-bound-online}
\end{align}
which combined with \eqref{eq:Qh-1a-iteration-lower-bound-online} immediately leads to
\begin{align*}
Q_h^{\pi}(1, a)& \geq  1 + (1-1/H) V_{h+1}^{\pi}(1)\qquad \text{for all }a\in \cA_1, \notag\\
&\Longrightarrow \quad V_h^{\pi}(1) \ge 1 + (1-1/H)V_{h+1}^{\pi}(1). 
\end{align*}
Iterating this recursion and using $V_H^{\pi}(1)=1$ (cf.~\eqref{eq:VH-pi-lower-bound-online}) yield
\begin{align}
V_h^{\pi}(1) &\ge 1 + (1-1/H)\Big\{1 + (1-1/H)V_{h+2}^{\pi}(1)\Big\} \geq \cdots \notag\\
&\ge \sum_{i = 0}^{H-h} (1-1/H)^i = \frac{1 - (1-1/H)^{H-h+1}}{1/H} 
\geq \frac{1 - e^{-\frac{H-h+1}{H}}}{1/H}
\notag\\
&\ge \frac{\frac{1}{2}\frac{H-h+1}{H}}{1/H}
= \frac{H-h+1}{2},
\label{eq:Vh-pi-lower-bound-online}
\end{align}
where we have used the elementary inequalities $1-x\leq \exp(-x)$ and $1-\exp(-x)\geq x/2$ for $0\leq x\leq 1$.  Additionally, it also follows directly from \eqref{eq:Qh-1a-iteration-lower-bound-online} that
\begin{align}
V_h^{\star}(1)=Q_h^{\star}(1, a_h^{\star}) &= 1 + P_h(1\mymid 1,a_h^{\star}) V_{h+1}^{\star}(1).
\label{eq:Qh-opt-1a-iteration-lower-bound-online}
\end{align}
%

%\item {\em Occupancy distributions.}
%%
%We next lower bound the occupancy distribution. Under the initial state distribution~\eqref{eq:rho-s-initial-dist-lower-bound-online}, the occupancy distribution associated with any policy $\pi$ (see the definition in~\eqref{eq:dh-pi-defn}) satisfies
%%
%\begin{align*}
%d_h^{\pi}(1; \rho) &= \sum_{a \in \mathcal{A}_1}
%  d_{h-1}^{\pi}(1, a; \rho) P_{h-1}(1\mymid 1,a) \notag\\  
%&  \geq (1-1/H)\sum_{a \in \mathcal{A}_1}
%  d_{h-1}^{\pi}(1, a; \rho) = (1-1/H)d_{h-1}^{\pi}(1; \rho),
%\end{align*}
%
%where the inequality results from \eqref{eq:PhM-1a-lower-bound-online}. 
%Recursively applying this bound yields 
%
%\begin{align}
%d_h^{\pi}(1; \rho) &\geq (1-1/H)^{h-1} d_1^{\pi}(1; \rho) 
%= (1-1/H)^{h-1} 
%\arxiv{\notag\\
%&}
%\ge (1-1/H)^H \geq 1/4
%\label{eq:dh-rho-initial-lower-bound-online}
%\end{align}
%
%for each $h\in [H]$,  where we use $(1-1/H)^{H}\ge 1/4$ for any $H\geq 2$.  
%
\end{itemize}

\paragraph{Lower bounding the suboptimality gap.} 
Now, we analyze the suboptimality gap associated with policy $\pi$. Towards this, we first use \eqref{eq:Qh-1a-iteration-lower-bound-online} and \eqref{eq:Qh-opt-1a-iteration-lower-bound-online} to demonstrate that  
\begin{align}
 & V_{h}^{\star}(1)-V_{h}^{\pi}(1)  \notag\\
 & =1+P_{h}(1\mymid1,a_{h}^{\star})V_{h+1}^{\star}(1)-\bigg\{1+\mathop{\mathbb{E}}\limits _{a_{h}\sim\pi_{h}(\cdot\mymid1)}\big[P_{h}(1\mymid1,a_{h})V_{h+1}^{\pi}(1)\big]\bigg\}\notag\\
 & =\mathop{\mathbb{E}}\limits _{a_{h}\sim\pi_{h}(\cdot\mymid1)}\Big[P_{h}(1\mymid1,a_{h}^{\star})V_{h+1}^{\star}(1)-P_{h}(1\mymid1,a_{h})V_{h+1}^{\pi}(1)\Big]\notag\\
 & =\mathop{\mathbb{E}}\limits _{a_{h}\sim\pi_{h}(\cdot\mymid1)}\Big[P_{h}(1\mymid1,a_{h}^{\star})\big(V_{h+1}^{\star}(1)-V_{h+1}^{\pi}(1)\big)\Big]\arxiv{\notag\\}
\arxiv{ & \qquad} +\mathop{\mathbb{E}}\limits _{a_{h}\sim\pi_{h}(\cdot\mymid1)}\Big[\big(P_{h}(1\mymid1,a_{h}^{\star})-P_{h}(1\mymid1,a_{h})\big)V_{h+1}^{\pi}(1)\Big]\notag\\
 & \ge \bigg(1-\frac{1}{H}\bigg)\big(V_{h+1}^{\star}(1)-V_{h+1}^{\pi}(1)\big)+\frac{\varepsilon_{0}}{2H^{2}}V_{h+1}^{\pi}(1)\mathop{\mathbb{E}}\limits _{a_{h}\sim\pi_{h}(\cdot\mymid1)}\Big[\mathds{1}(a_{h}\ne a_{h}^{\star})\Big]\notag\\
 &\geq\bigg(1-\frac{1}{H}\bigg)\big(V_{h+1}^{\star}(1)-V_{h+1}^{\pi}(1)\big)+\frac{\varepsilon_{0}(H-h)}{4H^{2}}\big(1-\pi_{h}(a_{h}^{\star}\mymid1)\big),
 \label{eq:Vh-star-Vhpi-gap-lower-bound-online-simple}
\end{align}
where the last relation follows from \eqref{eq:Vh-pi-lower-bound-online}, and the penultimate inequality holds true since, according to \eqref{eq:transition-kernel-M-lower-bound-online} and \eqref{eqn:prob-astar-a-1}, 
\begin{align*}
P_{h}(1\mymid1,a_{h}^{\star}) & \geq1-1/H,\\
P_{h}(1\mymid1,a_{h}^{\star})-P_{h}(1\mymid1,a_{h}) & \geq\frac{\varepsilon_{0}}{2H^{2}}\mathds{1}(a_{h}\ne a_{h}^{\star}). 
\end{align*}
Applying the above property \eqref{eq:Vh-star-Vhpi-gap-lower-bound-online-simple} recursively, we arrive at
\begin{align*}
V_{1}^{\star}(1)-V_{1}^{\pi}(1) & \geq\bigg(1-\frac{1}{H}\bigg)^{H}\big(V_{H+1}^{\star}(1)-V_{H+1}^{\pi}(1)\big) \arxiv{ \notag\\ }
\arxiv{ & \qquad }+\sum_{h=1}^{H}\bigg(1-\frac{1}{H}\bigg)^{h}\frac{\varepsilon_{0}(H-h)}{4H^{2}}\big(1-\pi_{h}(a_{h}^{\star}\mymid1)\big) \notag\\
 & =\sum_{h=1}^{H}\bigg(1-\frac{1}{H}\bigg)^{h}\frac{\varepsilon_{0}(H-h)}{4H^{2}}\big(1-\pi_{h}(a_{h}^{\star}\mymid1)\big) \notag\\
 &\geq \sum_{h=1}^{H}\frac{\varepsilon_{0}(H-h)}{16H^{2}}\big(1-\pi_{h}(a_{h}^{\star}\mymid1)\big),
% \label{eq:Vh-star-Vhpi-gap-lower-bound-online-final-proof}
\end{align*}
where the last inequality holds since $(1-1/H)^h\geq (1-1/H)^H \geq 1/4$ if $H\geq 2$. Consequently, this lower bound hinges upon how often policy $\pi$ selects suboptimal actions at each step.

\subsection{Proof of Lemma~\ref{lem:online-lb-testing}}
\label{sec:proof:lem:online-lb-testing}

For notational convenience, let $a_t$ denote the arm selected at round $t$ and let
$R_t=R_{a_t,t}$ denote the corresponding observed reward. Define
\[
x_t \coloneqq \{a_1,R_1,\ldots,a_t,R_t\}
\]
to be the trajectory observed up to the end of round $t$, and let
$\mathbb{P}_{\mathcal{B}_i}^{(t)}$ denote the distribution of $x_t$
under the bandit instance $\mathcal{B}_i$. We also denote the conditional
reward distribution under $\mathcal{B}_i$ by
\begin{equation}
    P_i(R \mymid a) \coloneqq  \begin{cases}
       1-q_i(a), & \text{if }R=1, \\
       q_i(a), & \text{if }R=0.
    \end{cases}
    \label{eq:reward-distribution-Pi-bandit-lb-online}
\end{equation}

Consider any online algorithm $\widehat{\pi}$ that generates a
sequence of policies $\{\pi^t\}_{t\geq 1}$, where the action at round $t$
depends only on the past trajectory $x_{t-1}$. With a slight abuse of
notation, we write
\[
     \pi^t(\cdot \mymid x_{t-1})
\]
for the action-selection distribution at round $t$. Here,  $\pi^t(\cdot \mid \cdot)$ denotes the underlying algorithmic rule, while the realized action-selection distribution at round $t$ relies on the observed history $x_{t-1}$. Under the bandit instance $\mathcal{B}_i$, the distribution of the trajectory $x_t$ admits the following factorization: 
\begin{align*}
\mathbb{P}^{(t)}_{\mathcal{B}_i}(x_t)
= \prod_{\tau=1}^t \pi^{\tau}(a_{\tau}\,|\,x_{\tau-1})P_{i}(R_{\tau}\,|\,a_{\tau}).
\end{align*}
where $P_i(\cdot\mymid a_t)$ is defined in \eqref{eq:reward-distribution-Pi-bandit-lb-online}. 
This implies that
\[
\frac{\mathbb{P}^{(t)}_{\mathcal{B}_1}(x_t)}
     {\mathbb{P}^{(t)}_{\mathcal{B}_i}(x_t)}
=
\prod_{\tau=1}^{t}
\frac{P_1(R_{\tau}\mymid a_{\tau})}
     {P_i(R_{\tau}\mymid a_{\tau})},
\]
meaning that the likelihood ratio decomposes into a product of per-round reward likelihood ratios. 
Consequently, the chain rule for KL divergence gives
\begin{align}
\mathsf{KL}\big(
    \mathbb{P}_{\mathcal{B}_1}^{(t)}
    \,\big\|\,
    \mathbb{P}_{\mathcal{B}_i}^{(t)}
\big)
=
\sum_{\tau=1}^{t}
\mathbb{E}_{\mathcal{B}_1}
\left[
    \mathsf{KL}\big(
        P_1(\cdot\mymid a_{\tau})
        \,\big\|\,
        P_i(\cdot\mymid a_{\tau})
    \big)
\right],
\label{eq:KL-factorize-bandit-lower-bound-online-proof}
\end{align}
where the expectation is taken with respect to the randomness of
$a_{\tau}$ under $\mathcal{B}_1$.

Next, we bound each KL divergence term appearing in the summation on the right-hand side of~\eqref{eq:KL-factorize-bandit-lower-bound-online-proof}. Suppose that
$\varepsilon_0\leq H/2$ and $H\geq 2$. Recalling that the conditional reward distributions \eqref{eq:reward-distribution-Pi-bandit-lb-online} are Bernoulli and letting $\mathsf{KL}(p\,\|\,q)$ denote the KL divergence between Bernoulli distributions with means $p$ and $q$, respectively, we can invoke Lemma~\ref{lem:KL-quadratic-lower-bound} to show the following:  
\begin{itemize}
\item If $a=i$, then it holds that 
\begin{align*}
\mathsf{KL}\big(P_{1}(\cdot\,|\,a)\,\|\,P_{i}(\cdot\,|\,a)\big) & =\mathsf{KL}\big(q_{1}(a)\,\|\,q_{i}(a)\big)\leq\frac{\big(q_{i}(a)-q_{1}(a)\big)^{2}}{q_{i}(a)\big(1-q_{i}(a)\big)} \arxiv{ \\
 &} \leq\frac{(\varepsilon_{0}/H^{2})^{2}}{1/(4H)}=\frac{4\varepsilon_{0}^{2}}{H^{3}}.
\end{align*}
Here, the last line is a consequence of the properties (see~\eqref{eqn:prob-astar-a-1})
\begin{align*}
\big|q_{i}(a)-q_{1}(a)\big| & \leq \frac{\varepsilon_{0}}{H^{2}}, 
\qquad q_i(a)
\geq 
\frac{1}{H}-\frac{\varepsilon_0}{H^2}
\geq \frac{1}{2H},
\qquad
1-q_i(a)\geq 1-\frac{1}{H} \geq\frac{1}{2}.
\end{align*}
\item If $a\neq i$, then $q_i(a)=q_1(a)$ (see~\eqref{eqn:prob-astar-a-1})
and, as a consequence,
\[\mathsf{KL}\big(P_{1}(\cdot\,|\,a)\,\|\,P_{i}(\cdot\,|\,a)\big)=0.\]
\end{itemize}
These two properties about the KL divergence taken together yield
\[
\sum_{i=1}^{A}\mathsf{KL}\big(P_{1}(\cdot\,|\,a)\,\|\,P_{i}(\cdot\,|\,a)\big)
=\mathsf{KL}\big(P_{1}(\cdot\,|\,a)\,\|\,P_{a}(\cdot\,|\,a)\big)\leq\frac{4\varepsilon_{0}^{2}}{H^{3}},
\]
which combined with \eqref{eq:KL-factorize-bandit-lower-bound-online-proof} leads to
\begin{align}
\frac{1}{A}\sum_{i=1}^{A}\mathsf{KL}\big(\mathbb{P}_{\mathcal{B}_{1}}^{(t)}\parallel\mathbb{P}_{\mathcal{B}_{i}}^{(t)}\big) & =\frac{1}{A}\sum_{\tau=1}^{t}\mathbb{E}_{\mathcal{B}_{1}}\Bigg[\sum_{i=1}^{A}\mathsf{KL}\big(P_{1}(\cdot\,|\,a_{\tau})\,\|\,P_{i}(\cdot\,|\,a_{\tau})\big)\Bigg] \arxiv{ \nonumber \\
 &}  \leq\sum_{\tau=1}^{t}\frac{4\varepsilon_{0}^{2}}{AH^{3}}=\frac{4\varepsilon_{0}^{2}t}{AH^{3}}.
 \label{eq:KL-sum-over-A-ub-bandit-lb-online}
\end{align}
With \eqref{eq:KL-sum-over-A-ub-bandit-lb-online} in place, applying
Lemma~\ref{lem:reverse-Fano-bound} then reveals that
\begin{align*}
\inf_{\widehat{\pi}}
\frac{1}{A}
\sum_{i=1}^{A}
\mathbb{P}_{\mathcal{B}_i}(a_{t+1}\neq i)
&\geq
1-\frac{1}{A}
-
\sqrt{
    \frac{A-1}{2A^2}
    \sum_{i=2}^{A}
    \mathsf{KL}\big(
        \mathbb{P}_{\mathcal{B}_1}^{(t)}
        \,\big\|\,
        \mathbb{P}_{\mathcal{B}_i}^{(t)}
    \big)
}\\
&\geq
1-\frac{1}{A}
-
\sqrt{
    \frac{2(A-1)\varepsilon_0^2t}{A^2H^3}
} \\
& \geq 1-\frac{1}{A}
-\sqrt{\frac{A-1}{9A}} \geq \frac{1}{3},
\end{align*}
provided that $t \leq \frac{A H^3}{18 \varepsilon_0^2}$ and $A\geq 3$. This concludes the proof.

\section{Notes}

Multi-armed bandit problems have deep roots in the statistics literature. First formulated by \citet{thompson1933likelihood}, the bandit framework, along with the notion of regret, was later popularized by Herbert Robbins (e.g., the seminal work of \citet{robbins1952some}). The use of upper confidence bounds in bandit problems was initially suggested by \citet{lai1985asymptotically,lai1987adaptive}. Subsequent research established asymptotic guarantees for UCB-style algorithms (e.g., \citet{lai1987adaptive,katehakis1995sequential,agrawal1995sample,burnetas1996optimal}), while non-asymptotic regret analyses were later developed by, among others, \citet{auer2002finite,bubeck2012regret,russo2013eluder,cappe2013kullback}. By adaptively tuning the exploration bonuses, one can further obtain anytime guarantees for UCB-style algorithms (i.e., without requiring prior knowledge of the time horizon) and remove the extra logarithmic factor in the regret bound \citep{auer2002finite,audibert2009minimax,degenne2016anytime}. We refer the interested reader to \citet{lattimore2020bandit} for a comprehensive discussion of the historical development of bandit algorithms and related bibliographical remarks.

 %see also \citet{filippi2010optimism,bourel2020tightening,talebi2018variance} for other variants of UCRL.  

 %see discussion in \citet{azar2017minimax,jin2018q}.  
%\citet{fruit2020improved} developed an improved regret bound for UCRL2 by using empirical Bernstein-style bounds, which however was still suboptimal by a factor of at least $\sqrt{HS}$ when specialized to  episodic finite-horizon MDPs. 

The optimism principle has subsequently become a central design paradigm in online RL. Building on ideas that originated in the bandit literature, a large body of work has developed optimistic algorithms for sequential learning and decision-making in unknown environments (e.g., \citet{auer2006logarithmic,jaksch2010near,bartlett2009regal,filippi2010optimism,dann2015sample,dann2017unifying,azar2017minimax,jin2018q,dann2019policy,zanette2019tighter,bourel2020tightening}). One of the earliest contributions in this direction is the work of \citet{jaksch2010near}, which proposed the \texttt{UCRL2} algorithm for weakly communicating MDPs. The key idea is to construct confidence regions for the transition kernel and use them to identify policies that are optimistic w.r.t.~the set of plausible MDPs. This \texttt{UCRL2} algorithm builds upon earlier optimistic methods, such as the \texttt{UCRL} algorithm proposed by \citet{auer2006logarithmic}. When applied to episodic finite-horizon MDPs, however, the regret guarantees obtained via the \texttt{UCRL2} analysis are not minimax optimal, exhibiting a gap of at least a factor of $\sqrt{H^{2}S}$ relative to the minimax limit.

%;  

In comparison to \texttt{UCRL2} and its variants, a more sample-efficient paradigm is to construct upper confidence bounds for the optimal Q-functions and value functions and use these confidence estimates to guide exploration. This paradigm has led to a sequence of sample-efficient algorithms for online RL, including both model-based methods (e.g., \citet{azar2017minimax,zhang2020reinforcement,menard2021ucb,zhang2025settling}) and model-free methods (e.g., \citet{jin2018q,bai2019provably,yang2021q,zhang2025q}). When combined with confidence bounds derived from empirical Bernstein inequalities, which exploit variance information, model-based algorithms based on this paradigm can attain asymptotically optimal regret guarantees. For model-free RL, however, variance-aware bonuses alone are insufficient to achieve asymptotically optimal regret, which motivates the development of more sophisticated variance-reduction techniques 
\citep{zhang2020almost,li2021breaking,li2023breaking,zhang2026regret} (originally referred to as reference-advantage decomposition by  \citet{zhang2020almost}). 
Determining how to optimally leverage optimism to achieve minimax-optimal regret without incurring high burn-in cost remained an open problem until the work \citet{zhang2024settling,zhang2025settling}, which established the minimax optimality of model-based online RL for any sample size.  In contrast, the most sample-efficient model-free algorithm currently known, due to \citet{li2021breaking}, still incurs a burn-in sample size on the order of  $SA\mathsf{poly}(H)$.

Beyond the setting studied in this chapter, a variety of related problem formulations have been investigated, motivated by different practical considerations. One prominent example is reward-free (or reward-agnostic) exploration, where the goal is to collect data during an exploration phase without knowledge of the reward function, while still enabling near-optimal performance once the reward function is subsequently specified (e.g., \citet{jin2020reward,zanette2020provably,zhang2020task,zhang2020nearly,wagenmaker2022reward,li2024minimax,wang2020reward,zhang2021reward}). Another direction is the study of ``horizon-free RL,'' which studies finite-horizon {\em stationary} MDPs under the assumption that the total reward accumulated within each episode is bounded by 1; the objective is to design algorithms whose regret bounds almost do not scale with the horizon length $H$,  thereby mitigating the adverse effect of long horizon \citep{jiang2018open,zanette2019tighter,zhang2021reinforcement,li2022settling,zhang2024horizon,wang2025model}. There has also been a line of work
that focuses on reducing the switching cost, namely the number of times the learner changes its sampling policy throughout the learning process, motivated by practical applications in which policy deployment is costly  \citep{bai2019provably,qiao2022near,qiao2022sample,huang2022towards,zhao2024nearly,zhang2026deployment}. 
Additionally, low-regret algorithms for constrained MDPs have been developed in, e.g., \citet{efroni2020exploration,brantley2020constrained,ding2021provably,liu2021learning,bura2022dope,wei2022triple,wei2023provably}. 
Furthermore, the efficient use of the optimism principle under low-complexity function approximation --- particularly linear function approximation --- has itself become a major research direction; we refer the reader to, e.g., \citet{yang2020reinforcement,jin2020provably,zanette2020learning,du2020good,yang2020function,pmlr-v108-modi20a,ayoub2020model,wang2020reinforcement,zanette2021exponential,wang2021optimism,weisz2021exponential,wang2021exponential,li2021sample,zhou2021nearly,jin2021bellman,du2021bilinear,yin2022efficient,he2023nearly} and the references therein for details.

Beyond confidence-interval-based methods, several alternative approaches to incentivizing exploration have also been extensively studied in both theory and practice. One notable approach is {\em posterior sampling} \citep{osband2013more,osband2017why}, with representative examples including Thompson sampling \citep{gopalan2015thompson,agrawal2017posterior,ouyang2017learning,agrawal2026qlearning}, feel-Good Thompson sampling \citep{zhang2022feel}, and learning rate randomization \citep{tiapkin2023model,wang2025provably}. Another line of work promotes exploration by regularizing the maximum likelihood estimate of the model or value functions toward more optimistic versions, such as \citet{hung2021reward,mete2021reward,yang2025exploration,liu2023maximize}. Randomized exploration, when suitably designed has also been shown to achieve efficient exploration \citep{xiong2022near}.    Collectively, these developments showcase a diverse array of algorithmic ideas available for balancing exploration and exploitation. 

%% file: chapters/bandits.tex
\section{A detour: the \UCB algorithm for multi-armed bandits}
\label{sec:bandits-UCB}

Before embarking on the design and analysis of online RL algorithms, we take a brief detour to study multi-armed bandits, a special case of online RL where state transitions are absent or $S=1$. In this setting, there is no transition kernel to estimate; instead, we consider the scenario where the observed rewards are noisy, so that model uncertainty arises solely from the stochasticity of the rewards. As it turns out, several key ideas developed in the multi-armed bandit literature pave the way for the design of efficient online RL algorithms. 

\subsection{Setting: stochastic bandits} 

Consider a standard stochastic bandit model, where an agent interacts sequentially with the unknown environment over $T$ rounds. The key components of this bandit model are described as follows.     
\begin{itemize}
	\item {\em Actions.} The agent has access to a set of $A$ arms (i.e., actions), denoted by $\mathcal{A} = \{1, \dots, A\}$. One can picture a gambler standing in front of a row of $A$ slot machines, each with its own arm; at each round, the gambler needs to decide which machine to play.

	\item {\em Mean rewards.} Each arm $a \in \mathcal{A}$ is associated with a time-invariant mean reward $r(a) \in [0,1]$. In the gambling analogy, each slot machine has its own payout rate. The optimal mean reward is denoted by
	\begin{subequations}
    \label{eq:defn-rstar-astar-bandits}
	\begin{equation}
		r^{\star} \coloneqq \max _{a\in \mathcal{A}} r(a) ,
		\label{eq:defn-rstar-bandits}
	\end{equation}
	with the optimal arm defined correspondingly as
	\begin{equation}
		a^{\star} \coloneqq \arg \max _{a\in \mathcal{A}}\, r(a).
		\label{eq:defn-astar-bandits}
	\end{equation}
	\end{subequations}
	Crucially, these mean rewards are {\em unknown} to the agent and can only be learned through sequential interaction with the environment over time.

	\item {\em Realized rewards.} 
		At each round $t$, nature generates a noisy reward $R_{a,t}$ independently for each arm $a\in \mathcal{A}$, given by
		\begin{align}
			R_{a,t} = r(a) + z_{a,t}, \qquad \text{with } z_{a,t} \overset{\mathrm{i.i.d.}}{\sim} \mathcal{N}(0, 1).
			\label{eq:assumption-Gaussian-reward-bandit}
		\end{align}
		While we assume i.i.d.~standard Gaussian noise for simplicity of exposition, this assumption can be significantly relaxed without affecting the key algorithmic principles and insights (see, e.g., \citet{lattimore2020bandit}).

	\item {\em Observations (bandit feedback).}  The agent operates under bandit feedback, meaning that it does not observe the full collection $\{R_{a,t}\}_{a \in \mathcal{A}}$ of noisy rewards at each round. Instead, at round $t$, the agent selects a single arm $a_t \in \mathcal{A}$ --- potentially based on previous actions and observations --- and observes only the corresponding realized reward $R_{a_t, t}$.

%	\item {\em Policy.} 
%		As before, the action selection rule (or policy) in round $t$ is denoted by $\pi_t\in \Delta(\mathcal{A})$, so that action $a_t$ is drawn from $a_t\sim \pi_t$. We refer to $\pi=\{\pi_t\}_{1\leq t\leq T}$ as the policy of the agent over the entire learning process. 

\end{itemize}

Naturally, the agent would aim to maximize its cumulative payoff over $T$ rounds. To evaluate the performance of a bandit algorithm (which determines how actions are selected over time), we introduce the (cumulative) regret as  
\begin{subequations}
\begin{equation} \label{eq:regret_bandit}
	\mathsf{Regret}(T) \coloneqq Tr^{\star} -  \sum_{t=1}^T r(a_t)  ,
\end{equation}
which measures the total expected loss in reward relative to always playing the optimal arm. 
Here, we employ the mean reward $r(a_t)$ rather than the realized reward $R_{a_t,t}$ to marginalize out observation noise, which ensures non-negativity of regret. Notably, $\mathsf{Regret}(T)$ is itself a random variable, as the sequence of actions $\{a_t\}$ is a function of the history of noisy rewards.

Moreover, the regret admits a simple yet useful decomposition. Let $N_T(a)$ denote the number of times arm $a$ is pulled over $T$ rounds. Then   the regret can be expressed in terms of $\{N_T(a)\}$ and the suboptimality gaps:
\begin{equation}
\label{eqn:regret-decomposition-ucb}
	\mathsf{Regret}(T) = \sum_{t=1}^T \Delta(a_t)  
	= \sum_{a\in \mathcal{A}} \Delta(a) N_T(a) \qquad \text{with }\Delta(a)\coloneqq r^{\star} - r(a). 
\end{equation}
\end{subequations}
Given that the mean rewards are unknown {\em a priori}, the agent must balance exploring undersampled arms with exploiting those currently deemed favorable.  Despite its apparent simplicity, this classical bandit formulation already encapsulates a central challenge of sequential decision-making under uncertainty: the interplay between learning, decision-making, and data collection in real time.

\subsection{Optimism in the face of uncertainty} 
We now introduce a core design principle underlying a broad family of bandit algorithms. 
On the one hand, if an arm $a$ has been pulled only a small number of times, the corresponding estimate of its mean reward $r_a$ can be highly volatile and therefore unreliable; this motivates playing the arm more often in order to improve the accuracy of the reward estimate. On the other hand, if the estimated reward of arm $a$ is lower than that of other arms, this arm is likely suboptimal, suggesting that it should be pulled less frequently. In summary, 
\begin{itemize}
	\item[(i)] high uncertainty (i.e., low confidence) in an arm's reward estimate can incentivize further exploration of that arm;

	\item[(ii)] a large reward estimate encourages more frequent pulls of that arm. 
\end{itemize}  
One strategy to unify these two competing principles is to introduce an optimistic estimate of the form: 
\begin{equation*}
	\widehat{r}(a) = \text{empirical reward estimate for arm }a ~+~ \text{its uncertainty} , 	
\end{equation*}
and then select the arm with the highest value of $\widehat{r}(a)$, provided that this quantity is constructed appropriately. Since $\widehat{r}(a)$ acts as an optimistic estimate of the reward $r(a)$ (in the sense that it intentionally biases the estimate upward to account for uncertainty), this idea is commonly referred to as the {\em optimism principle} in the face of uncertainty, originally introduced by \citet{lai1985asymptotically}.

While conceptually natural, this paradigm raises a fundamental question: how should the uncertainty bonus be calibrated relative to the empirical reward estimate? Evidently, the choice of this trade-off is crucial for achieving optimal regret and is, in general, far from obvious.

\subsection{The upper confidence bound (\UCB) algorithm}

Somewhat surprisingly, a simple yet elegant approach --- known as the upper confidence bound (\UCB)  algorithm, originally developed by \citet{lai1985asymptotically,lai1987adaptive} --- is sufficient to achieve the optimal exploration-exploitation trade-off.  

%in two phases. \yc{do we need to separate the two phases? Since $N_a(t)=0$ at the beginning, all arms will be pulled once at the beginning.}
%
%\begin{itemize}
%	\item[(i)] The first phase (i.e., when $t \leq A$) serves as an initialization stage, during which each arm is pulled once and the corresponding noisy reward is observed. This phase guarantees that every arm has at least one observation, allowing the algorithm to form an initial estimate of each mean reward. 

%	\item[(ii)] In the second phase, 

More precisely, the \UCB algorithm operates by selecting, at each round $t$, the arm with the largest upper confidence bound on its mean reward, where these confidence bounds are constructed using the samples observed up to and including time $t$. Specifically, for each round $t$, the agent pulls the arm
\begin{equation}
	a_t = \arg\max_{a\in \mathcal{A}} r_{t-1}^{\ucb}(a),
\end{equation}
where $r_{t}^{\ucb}(a)$ represents the UCB estimate for arm $a$, defined as
\begin{subequations}
\begin{equation}
	r_{t}^{\ucb}(a)= \overline{r}_{t}(a)+\sqrt{\frac{2\log\frac{T^2}{\delta}}{N_{t}(a)}}. 
% \begin{cases}
% 	\widehat{r}_{a}(t)+\sqrt{\frac{2\log\frac{1}{\delta}}{N_{a}(t)}}, & \text{if }N_{a}(t)>0.\\
% 	\infty, & \text{else}.
% \end{cases}
\label{eq:UCB-reward-bandit-alg}
\end{equation} 
Here, $0<\delta<1$ is some confidence parameter (whose role will become clear shortly), $N_{t}(a)$ denotes the number of times arm $a$ has been pulled up to round $t$, and $\overline{r}_{t}(a)$ represents the empirical mean reward of arm $a$ computed from all observations collected so far:
\begin{align}
	\overline{r}_{t}(a) = \frac{1}{N_{t}(a)} \sum_{i=1}^t R_{a,i} \mathds{1}\big(\text{arm } a \text{ is pulled at round }i\big).
\end{align}
\end{subequations}
%
%\end{itemize}
%
The complete procedure of this algorithm is summarized in Algorithm~\ref{alg:main-ucb-bandits}.

The specific form of the uncertainty term $\sqrt{\frac{2\log(T^2/\delta)}{N_{t}(a)}}$ in \eqref{eq:UCB-reward-bandit-alg} --- often dubbed the exploration bonus --- is motivated by the use of concentration inequalities to quantify statistical uncertainty. To make this precise, recall from standard Gaussian tail bounds that a Gaussian random variable $Z\sim \mathcal{N}(\mu,\sigma^2)$ satisfies (see, e.g., \citet[Chapter~2]{vershynin2018high})
\begin{align*}
	\mathbb{P} \big(Z \leq \mu - \tau \big) \leq \exp\bigg(-\frac{\tau^2}{2\sigma^2}\bigg)
\end{align*}
for any $\tau \geq 0$, and as a result, 
\begin{align*}
	\mathbb{P} \bigg(Z  + \sigma \sqrt{2\log \frac{1}{\delta}} \geq \mu  \bigg) > 1- \delta 
	%\label{eq:Gaussian-deviation-standard}
\end{align*}
for any $0<\delta < 1$. 
Now observe that, under the Gaussian reward model \eqref{eq:assumption-Gaussian-reward-bandit}, one has  $\overline{r}_t(a) - r(a)= \mathcal{N}\big(0, \frac{1}{N_{t}(a)}\big).$
Applying the above Gaussian tail bound
%\eqref{eq:Gaussian-deviation-standard} 
thus yields 
\begin{subequations}
		\label{eq:Gaussian-deviation-standard-to-reward-forward-reverse}
\begin{align}
	\mathbb{P} \big(r^{\ucb}_t(a) \geq r(a) \big) = 
	\mathbb{P} \Bigg(\overline{r}_t(a) + \sqrt{\frac{2\log\frac{T^2}{\delta}}{N_{a}(t)}} \geq r(a)  \Bigg) > 1- \frac{\delta}{T^2}, 
	\label{eq:Gaussian-deviation-standard-to-reward}
\end{align}
and similarly, 
\begin{align} 
	\mathbb{P} \Bigg(r(a) + \sqrt{\frac{2\log\frac{T^2}{\delta}}{N_{t}(a)}} \geq \overline{r}_t(a)  \Bigg) > 1- \frac{\delta}{T^2}.
	\label{eq:Gaussian-deviation-standard-to-reward-reverse}
\end{align}
\end{subequations}
Consequently, the exploration bonus serves as a high-probability upper bound on the reward estimation error, while $r_t^{\mathsf{ucb}}(a)$ acts as a statistically valid upper confidence bound on the unknown mean reward $r(a)$ --- hence the name of the algorithm. This interpretation makes clear that the \UCB algorithm makes decision according to an optimistic, yet statistically plausible, view of their mean rewards.

\begin{algorithm}[t]
	\DontPrintSemicolon
\caption{The upper confidence bound (\UCB) algorithm for online stochastic bandits\label{alg:main-ucb-bandits}}
%\begin{algorithmic}[ht]

	\textbf{input:} number of rounds $T$, confidence parameter $\delta$. \\

    	\textbf{initialize} $N(a) \leftarrow 0$, $r^{\mathsf{sum}}(a) 
		\leftarrow 0$, and $\overline{r}(a) 
		\leftarrow 0$ for all $a\in \cA$. \\

	%\textbf{initialize }  
	%$c_1=\frac{460}{9}$, $c_2 = 2\sqrt{2}$, $c_3=\frac{544}{9}$.  
	
%	{\color{blue}\tcp{phase 1: pull each arm once.}}
%	\For{$t=1,\ldots,A$} {
%		set $a\leftarrow t$; pull arm $a$ at round $t$, and receive reward $R_{a,t}$. \\
%		set $N_a \leftarrow 1$, $r^{\mathsf{sum}}_{a} 
%		\leftarrow R_{a,t}$, and $\overline{r}_{a} 
%		\leftarrow R_{a,t}$.
%	}
	
	{\color{blue}\tcp{pull arms based on UCBs.}}
	\For{$t=1,\ldots,T$} {
	{\color{blue}\tcc{add bonus.}}
		compute $a_t \in  \arg\max_{a\in \mathcal{A}} 	r^{\mathsf{ucb}}(a)$, where 
		\begin{equation*}
			r^{\ucb}(a) \leftarrow \overline{r}(a)+\sqrt{\frac{2\log\frac{T^2}{\delta}}{N(a)}}. 
%\label{eq:UCB-reward-bandit-alg-block}
\end{equation*} 

		pull arm $a_t$ at round $t$, and receive reward $R_{a_t,t}$. \\
		set $N(a_t) \leftarrow N(a_t)+ 1$, $r^{\mathsf{sum}}(a_t) 
		\leftarrow r^{\mathsf{sum}}(a_t) + R_{a_t,t}$, and $\overline{r}(a_t) 
		\leftarrow r^{\mathsf{sum}}(a_t) / N(a_t)$.
	}

%\end{algorithmic}
\end{algorithm}

\subsection{Theoretical guarantees}

To rigorously demonstrate the efficacy of the \UCB algorithm, we present its regret guarantees in the following theorem; the proof is provided in Section~\ref{sec:proof-thm:bandit-pure-ucb}.   
\begin{theorem}
\label{thm:bandit-pure-ucb}
	Consider any $0<\delta<1$, and suppose that $T\geq A \geq 2$. Recall that $\Delta(a) = r^{\star} - r(a) \in [0,1]$ for all $a\in \cA$. Then the expected regret of the \UCB algorithm (see Algorithm~\ref{alg:main-ucb-bandits}) satisfies
	\begin{subequations}
%	%
	\begin{equation}
		\mathbb{E}[\mathsf{Regret}(T)] \leq 8 \sqrt{AT\log \bigg( \frac{T^2}{\delta} \bigg)}  .
		\label{eq:thm-bandit-general-regret}
	\end{equation}
	In addition, it also holds that
	\begin{align}
		\mathbb{E}[\mathsf{Regret}(T)] \leq 10 \log \bigg(\frac{T^2}  {\delta} \bigg) \cdot \sum_{a:\,\Delta(a)>0}   \frac{1}{\Delta(a) }.
	\label{eq:regret-ub-instance-bandit-proof}
\end{align} 
	\end{subequations}
\end{theorem}
\begin{remark}
    These results can also be stated as high-probability regret bounds \citep{lattimore2020bandit}. However, since bandits are not the primary focus of this monograph and are introduced mainly to motivate certain algorithmic ideas, we restrict our attention to the expected regret bounds for brevity.
\end{remark}

This theorem provides two complementary results: a worst-case regret bound \eqref{eq:thm-bandit-general-regret} that holds uniformly over all problem instances, and a more refined instance-dependent bound \eqref{eq:regret-ub-instance-bandit-proof} that explicitly captures the role of the suboptimality gaps $\{\Delta(a)\}_{a\in \cA}$. A few remarks about these regret guarantees are in order.
\begin{itemize}
	\item Assuming that the number of arms is $A=O(1)$, then the worst-case regret bound \eqref{eq:thm-bandit-general-regret}  simplifies to
		$$ \mathbb{E}[\mathsf{Regret}(T)]  \leq \widetilde{O}\big( \sqrt{T} \big).$$
		In the language of online learning \citep{shalev2012online}, this implies that \UCB is a \emph{no-regret} learning algorithm, meaning that the average regret per round, $\frac{1}{T} \mathbb{E}[\mathsf{Regret}(T)]$, vanishes asymptotically as $T \to \infty$. In other words, although the agent may initially incur losses while exploring uncertain arms, its long-term performance eventually approaches that of the optimal arm. Encouragingly, the regret bound $\widetilde{O}(\sqrt{AT})$ in \eqref{eq:thm-bandit-general-regret} is minimax-optimal up to logarithmic factors, revealing that the optimism principle achieves essentially optimal worst-case performance. In fact, the logarithmic factor in \eqref{eq:thm-bandit-general-regret} can be shaved through more refined algorithmic and analysis techniques (see \citet{audibert2009minimax}). See, e.g., \citet[Section~15.2]{lattimore2020bandit} for a matching lower bound.

	\item The instance-dependent regret bound \eqref{eq:regret-ub-instance-bandit-proof}, which is often tighter than the worst-case bound \eqref{eq:thm-bandit-general-regret}, reveals the intrinsic difficulty of a bandit instance through the suboptimality gaps. In particular, this regret bound scales inversely with the {\em harmonic mean} of the gaps: 
$$ 
		\mathbb{E}[\mathsf{Regret}(T)] \leq \widetilde{O}\bigg( \frac{  A }{\mathsf{harmonic}\text{-}\mathsf{mean}(\{\Delta(a)\}_{a\neq a^{\star}})} \bigg),
$$
where we define, for any sequence of positive numbers $\{x_i\}_{1\leq i\leq n}$,
$$\mathsf{harmonic}\text{-}\mathsf{mean}\big(\{x_i\}_{1\leq i\leq n}\big) \coloneqq \frac{n}{x_1^{-1}+\cdots+x_n^{-1}}.
$$
		For instance, when $\Delta(a)\asymp 1$ holds for all suboptimal arms and when $A=O(1)$, the expected regret bound reduces to 
		$$ \mathbb{E}[\mathsf{Regret}(T)]  \leq  \widetilde{O}( 1 ), $$
		illustrating that, for favorable problem instances, the regret can be considerably smaller than the worst-case $\widetilde{O}(\sqrt{T})$ scaling. 

\end{itemize}

\paragraph{Intuitive explanation.}
Before presenting the proof of Theorem~\ref{thm:bandit-pure-ucb}, we take a moment to provide some intuition for the regret bounds. Recall the regret decomposition in \eqref{eqn:regret-decomposition-ucb}:
\begin{align}
\mathbb{E}[\mathsf{Regret}(T)] & =  \sum_{a\in \mathcal{A}} \Delta(a) \mathbb{E}\left[ N_T(a) \right],
	\label{eq:regret-decomp-bandit-proof}
\end{align}
where, by a slight abuse of notation, $N_T(a)$ denotes the total number of pulls of arm $a$ over the $T$ rounds. 
This suggests controlling $\mathbb{E}[N_T(a)]$ for each arm $a$. As it turns out, a key step of the proof is to show that, for any suboptimal arm $a$, 
$$
\mathbb{E}\left[ N_T(a) \right]\leq \widetilde{O}\Big(\big(\Delta(a)\big)^{-2}\Big).
$$ 
This scaling is intuitive. If $N_T(a)$ becomes significantly larger than $\big(\Delta(a)\big)^{-2}$, then the empirical estimate $\overline{r}(a)$ concentrates sharply around the true mean $r(a)$, making it distinguishable from that of the optimal arm $a^\star$. More precisely, 
recall that the reward estimation error $\overline{r}_a-r(a) \sim \mathcal{N}(0,1/N_T(a))$ and is therefore on the order of $1/\sqrt{N_T(a)}$; once this uncertainty is much smaller than the suboptimality gap $\Delta(a)=r^{\star}-r(a)$, the learner can reliably identify arm $a$ as suboptimal and will no longer select it. Consequently, the summand $\Delta(a) \mathbb{E}[N_T(a)]$ in \eqref{eq:regret-decomp-bandit-proof} obeys
\[
    \Delta(a) \mathbb{E}[N_T(a)] \le \Delta(a) \widetilde{O}\Big(\big(\Delta(a)\big)^{-2}\Big) = \widetilde{O}\big(1/\Delta(a)\big).
\]
Summing over all arms suggests the claimed instance-dependent regret bound \eqref{eq:regret-ub-instance-bandit-proof}.

\subsection{Analysis: proof of Theorem~\ref{thm:bandit-pure-ucb}}
\label{sec:proof-thm:bandit-pure-ucb}

%Let us recall the following Hoeffding's inequality, which plays an important role in our analysis. 

%\begin{theorem}[Hoeffding's inequality]
%Let $X_1, X_2,\ldots, X_n$ be independent random variables satisfying $a_i \leq X_i \leq b_i$. Then for all $\delta\geq 0$, 
%\begin{align*}
%\mathbb{P}\left(  \left|\sum_{i=1}^n X_i - \mathbb{E}\Big[\sum_{i=1}^n X_i  \Big] \right|\geq \varepsilon \right) \leq 2\exp\left( -\frac{2\varepsilon^2}{\sum_{i=1}^n (b_i -a_i)^2} \right).
%\end{align*}
%\end{theorem}

%Setting $a_i =0$, $b_i =1$, and $\mathbb{E}[X_i] = \mu$, we obtain
%\begin{align*}
% \mathbb{P}\left(  \left|\frac{1}{n}\sum_{i=1}^n X_i - \mu   \right|  \geq \varepsilon \right)& \leq 2\exp\left( -2n \varepsilon^2 \right) \\
%\Longrightarrow  \quad \left| \frac{1}{n} \sum_{i=1}^n X_i - \mu   \right|  & \leq \sqrt{\frac{\log (2/\delta)}{ 2n}} \qquad \mbox{with prob.}\; 1 - \delta.
%\end{align*}

To begin with, we observe that the first $A$ rounds constitute an initialization phase, during which each arm is pulled once and the corresponding noisy reward is observed. Indeed, if an arm $a$ has never been selected, then $N(a)=0$ and hence $r^{\ucb}(a)=\infty$. Consequently, the algorithm always selects an unsampled arm whenever one exists, so that each arm must be pulled exactly once during the first $A$ rounds,  yielding an initial empirical mean estimate for every arm.

In what follows, we prove Theorem~\ref{thm:bandit-pure-ucb} with the aid of the regret decomposition \eqref{eq:regret-decomp-bandit-proof}, focusing on $t\geq A$.  For notational clarity, we use $N_{t}(a)$, $r^{\ucb}_t(a)$, and $\overline{r}_{t}(a)$  to denote the values of the running variables  $N(a)$, $r^{\ucb}(a)$, and $\overline{r}(a)$, respectively,  at the end of round $t$.

%\paragraph{Step 1: decomposing regret.} 

%Intuitively, once the number of pulls $N_a$ becomes sufficiently large, the reward estimation error diminishes, making it increasingly unlikely for arm $a$ to be selected if its mean reward is strictly below the optimal mean reward $r^\star$ (cf.~\eqref{eq:defn-rstar-bandits}). This observation forms the key idea for controlling $N_a$ of suboptimal arms. 
%
%To be precise, 

\paragraph{Step 1: connecting $N_{t}(a)$ with the arm-pulling probability.} 
Consider any given $a\in \cA$. 
We first identify a condition under which the UCB estimate $r^{\ucb}_t(a)$ of a suboptimal arm $a$ is smaller than that of the optimal arm $a^{\star}$ (cf.~\eqref{eq:defn-astar-bandits}), and hence arm $a$ will not be pulled at the beginning of round $t+1$. More specifically, if
\begin{equation}
	N_{t}(a) >  \frac{8 \log \frac{T^2}{\delta}}{\big(\Delta(a)\big)^2 }   
	\label{eq:Nat-condition-proof-bandit}
\end{equation}
holds for $t\geq A$, then with probability at least $1-2\delta / T^2$, 
\begin{align*} 
	r_{t}^{\mathsf{ucb}}(a) &= \overline{r}_{t}(a) + \sqrt{\frac{2\log \frac{T^2}{\delta}}{N_t(a)}}  \overset{\text{(i)}}{\leq}  r(a) +2 \sqrt{\frac{2\log \frac{T^2}{\delta}}{N_t(a)}}
	 \overset{\text{(ii)}}{<} r(a^{\star}) \overset{\text{(iii)}}{\leq}
	 %\overline{r}_{a^{\star},t} + \sqrt{\frac{2\log \frac{T^2}{\delta}}{N_{a^{\star}}(t)}} =
	 r_{t}^{\mathsf{ucb}}(a^\star) ,
\end{align*}
where (i) results from \eqref{eq:Gaussian-deviation-standard-to-reward-reverse}, (ii) is a consequence of Condition~\eqref{eq:Nat-condition-proof-bandit}, and (iii) follows from \eqref{eq:Gaussian-deviation-standard-to-reward}. 
As a result, for any $t\geq A$ and any given $a\in \cA$, 
\begin{align}
	\arxiv{&} \mathbb{P}\left(a_{t+1} =a, N_t(a) > \frac{8 \log \frac{T^2}{\delta}}{\big(\Delta(a)\big)^2 } \right) \arxiv{\notag\\}
	&\arxiv{\qquad} = \mathbb{P}\left(a_{t+1} =a \mid N_t(a) > \frac{8 \log \frac{T^2}{\delta}}{\big(\Delta(a)\big)^2 } \right) 
	\mathbb{P}\left( N_{t}(a) > \frac{8 \log \frac{T^2}{\delta}}{\big(\Delta(a)\big)^2 } \right) \notag\\
	&\arxiv{\qquad} \leq \mathbb{P}\left(a_{t+1} =a \mid N_t(a) > \frac{8 \log \frac{T^2}{\delta}}{\big(\Delta(a)\big)^2 } \right)
	\leq \frac{2\delta}{T^2}. 
	\label{eq:prob-suboptimal-a-N-cond-bandit}
\end{align}

\paragraph{Step 2: deriving an instance-dependent bound on $\mathbb{E}[N_T(a)]$.}
Equipped with the properties derived in Step 1, we can demonstrate that, for any arm $a$ with $\Delta(a)>0$, 
\begin{align}
	\mathbb{E}\left[ N_T(a) \right]  &=  1+ \mathbb{E}\left[ \sum_{t=A}^{T-1} \mathds{1}(a_{t+1} =a) \right] \notag\\
& = 1+ { \mathbb{E}\left[ \sum_{t=A}^{T-1} \mathds{1}\left(a_{t+1} =a, N_{t}(a) \leq \frac{8 \log \frac{T^2}{\delta}}{\big(\Delta(a)\big)^2 } \right)\right]} \arxiv{\notag\\}
\arxiv{&\qquad} + { \mathbb{E}\left[ \sum_{t=A}^{T-1} \mathds{1}\left(a_{t+1} =a, N_{t}(a) > \frac{8 \log \frac{T^2}{\delta}}{\big(\Delta(a)\big)^2 } \right) \right] } \notag\\
& \leq 1 +  \frac{8 \log \frac{T^2}{\delta}}{\big(\Delta(a)\big)^2 } + { \sum_{t=A}^{T-1} \mathbb{P}\left(a_{t+1} =a, N_{t}(a) > \frac{8 \log \frac{T^2}{\delta}}{\big(\Delta(a)\big)^2 } \right)  } \notag\\
& \leq 1+ \frac{8 \log \frac{T^2}{\delta}}{\big(\Delta(a)\big)^2 }+ \sum_{t=A}^{T-1}  { \frac{2\delta}{T^2} }  \notag\\
& \leq \frac{8 \log \frac{T^2}{\delta}}{\big(\Delta(a)\big)^2 } +  2. 
	\label{eq:instance-dependent-ub-Na-proof-bandit}
\end{align}
Here, the first inequality follows because arm $a$ is pulled exactly once during the first $A$ rounds, and thereafter the event
\[
	\left\{a_{t+1}=a,\; N_t(a)\le \frac{8\log(T^2/\delta)}{\big(\Delta(a)\big)^2}\right\}
\]
can occur at most $\frac{8\log(T^2/\delta)}{(\Delta(a))^2}$ times, as each occurrence increments $N_t(a)$ by one. The penultimate line in \eqref{eq:instance-dependent-ub-Na-proof-bandit} results from \eqref{eq:prob-suboptimal-a-N-cond-bandit}, whereas the last inequality in \eqref{eq:instance-dependent-ub-Na-proof-bandit} holds since $\delta\leq 1$ and $T\geq A\geq 2$.

\paragraph{Step 3: establishing the instance-dependent regret bound.} 
Substituting \eqref{eq:instance-dependent-ub-Na-proof-bandit} into the regret decomposition \eqref{eq:regret-decomp-bandit-proof} establishes the advertised instance-dependent regret bound (i.e., \eqref{eq:regret-ub-instance-bandit-proof}): 
\begin{align*}
	\mathbb{E}[\mathsf{Regret}(T)] 
	%&= \sum_{a:\, \Delta(a) >0} \Delta(a) \mathbb{E}[N_T(a)]
	%\notag\\
	&\leq \sum_{a:\, \Delta(a) >0} \Delta(a)  \left( \frac{8 \log \frac{T^2}{\delta}}{\big(\Delta(a)\big)^2 } + 2  \right) \leq \sum_{a:\Delta(a)>0}   \frac{10 \log \frac{T^2}{\delta}}{\Delta(a) } ,
	%\label{eq:regret-ub-instance-bandit-proof}
\end{align*} 
provided that $\Delta(a) \leq 1$ holds for all $a\in \cA$.

\paragraph{Step 4: establishing the worst-case regret bound.} 
With the instance-dependent bound \eqref{eq:instance-dependent-ub-Na-proof-bandit} on $\mathbb{E}\left[ N_T(a) \right]$ in place, we can proceed to establish the claimed worst-case regret bound \eqref{eq:thm-bandit-general-regret}. Considering any threshold $\Delta>0$ (to be specified shortly),  we make the following observations. 
\begin{itemize}
	\item For any arm $a$ with $\Delta(a) \geq \Delta$ (i.e., large suboptimality gap), it follows from \eqref{eq:instance-dependent-ub-Na-proof-bandit} that
		$$ \mathbb{E}\left[ N_T(a) \right] \leq \frac{8 \log \frac{T^2}{\delta}}{\big(\Delta(a)\big)^2 } +  2.$$

	\item For the set of arms with  $\Delta(a) < \Delta$ (i.e., small suboptimality gap), it holds trivially that
		$$\sum_{a} \mathbb{E}\left[ N_T(a) \right] \leq T. $$
\end{itemize}
Combine these bounds with the regret decomposition \eqref{eq:regret-decomp-bandit-proof} to reach
\begin{align*}
\mathbb{E}[\mathsf{Regret}(T)] 
	%&= \sum_{a:\, \Delta(a) >0} \Delta(a) \mathbb{E}[N_T(a)]
	%\notag\\  
	& \leq {\sum_{a: \, \Delta(a) \geq \Delta} \Delta(a)\left( \frac{8 \log \frac{T^2}{\delta}}{\big(\Delta(a)\big)^2 } +2 \right) }+ \Delta {\sum_{a: \, \Delta_a < \Delta}  \mathbb{E}\left[ N_T(a) \right]} \\
& \leq \frac{8 A\log \frac{T^2}{\delta}}{\Delta }  + 2A +  T \Delta ,
\end{align*}
where the last line also uses the assumption that $\Delta(a) \leq 1$ for each $a\in \mathcal{A}$. 
Choosing $\Delta = \sqrt{\frac{4A\log (T^2/\delta)}{T}}$, we arrive at
$$ 
	\mathbb{E}[\mathsf{Regret}(T)]\leq 6 \sqrt{AT\log \frac{T^2}{\delta}} + 2 A \leq 8 \sqrt{AT\log \frac{T^2}{\delta}},
$$
provided that $T\geq A$. This completes the proof.

%To establish Theorem~\ref{thm:bandit-instance-ucb}, it relies crucially on proving the following characterization for the number of arm-pulls for each arm
%\begin{align}
%\label{eqn:arm-pull-gap-dept}
%    \mathbb{E}\left[N_a \right]\leq \frac{4\log T}{\Delta_a^2} +  8.
%\end{align}
%Putting this together with relation~\eqref{eqn:regret-decomposition-ucb}
%\begin{equation}
%	\mathbb{E}[\mathsf{Regret}(T)] = \sum_{t=1}^T \mathbb{E}[\Delta_{a_t}]
%	= \sum_{a\in \mathcal{A}} \Delta_a \mathbb{E}[N_a]
%    %
%    \leq \frac{4\log T}{\Delta_a} +  8\Delta_a.
%\end{equation}

%To summarize, for any arm $a$ with $\Delta_a>0$, it holds that
%$$\mathbb{E}\left[ N_{a} \right] \leq \frac{4\log T}{\Delta_a^2} +8.$$

%This completes the proof of relation~\eqref{eqn:arm-pull-gap-dept} and thus the proof of Theorem~\ref{thm:bandit-instance-ucb}.

% \begin{lemma}[bounding the number of pulls of suboptimal arms]

% \end{lemma}

% \medskip

% \textbf{Proof of Theorem 2:}
% \begin{align*}
% R_T  =  \sum_{i=1}^n \Delta_i \mathbb{E}\left[ T_{i,T} \right] & \leq   \sum_{\Delta_i >0 }  \Delta_i \left( \frac{4\log T}{\Delta_i^2} +8  \right) \\
% & = \sum_{\Delta_i >0 }  \left(  \frac{4\log T}{\Delta_i}  + 8\Delta_i \right).
% \end{align*}

%% file: chapters/offlineRL.tex
\chapter{Offline RL}
\label{chapter:offline-RL}

Despite the promise of online RL, the limited capability of online data collection in various real-world applications presents a fundamental bottleneck to carrying the success of RL over to broader scenarios. For example, in applications such as clinical trials, autonomous systems, and online advertising, real-time data acquisition is expensive, high-stakes, and/or time-consuming. To circumvent this bottleneck, a plausible idea is to make more effective use of previously collected data,
since historical datasets may contain information that transfers to new tasks. For instance, the state-transition dynamics encountered in past tasks may sometimes partially resemble those arising in new tasks. 
The potential of this data-driven paradigm has been explored and recognized in a diverse array of contexts,  including but not limited to robotic manipulation \citep{ebert2018visual}, autonomous driving \citep{diehl2021umbrella}, and healthcare \citep{tang2021model}; see \citet{levine2020offline,prudencio2022survey} for recent surveys.  The subfield of RL based solely on historical data, without further interaction with the environment, is now commonly referred to as {\em offline RL} or {\em batch RL} \citep{lange2012batch,levine2020offline}.

\section{Problem formulation and challenges}

To facilitate a precise technical exposition, we begin by introducing the settings and objectives that will serve as the primary focus of this chapter. 
We then discuss the key challenges that arise in these settings, together with some important problem parameters introduced in response to these challenges.

\paragraph{Data generation mechanism.} 
The effectiveness and potential of offline RL hinge critically upon the quality of the historical dataset, denoted throughout by $\mathcal{D}$. For concreteness, this chapter focuses primarily on an independent sampling model tailored to discounted infinite-horizon MDPs to describe the sample-generating process underlying $\mathcal{D}$.

To be precise, suppose the learner has access to a batch dataset $\mathcal{D} = \{(s_i,a_i,s_i')\}_{1\leq i\leq N}$ comprising $N$ sample transitions. These samples are generated according to an unknown distribution $\myrho \in \Delta(\cS\times \cA)$ and the true transition kernel $P$ of the underlying MDP, namely, 
		\begin{align}
			(s_i,a_i) \overset{\mathrm{ind.}}{\sim} \myrho 
			\quad \text{and} \quad
			s_i' \overset{\mathrm{ind.}}{\sim} P(\cdot \mymid s_i, a_i),
			\qquad  1\leq i\leq N.
			\label{eq:sampling-offline-inf-iid}
		\end{align}
This coincides with the sampling model proposed in the seminal work \citet{rashidinejad2021bridging}. In practical applications, such samples may have been generated by executing one or more ``behavior policies'' in the MDP, which motivates the notation $\mathsf{b}$ in the distribution $d^{\mathsf{b}}$. 
As in previous chapters, we assume for simplicity that the exact reward functions are available to the learner; handling unknown rewards introduces only moderate additional technical challenges.

\begin{remark}
The independent sampling model can often be extended to more realistic data collection mechanisms, such as the setting in which the learner observes a single Markovian trajectory; see, e.g., \citet{yan2022efficacy}. We focus on independent sampling here to streamline the exposition. 
\end{remark}

\paragraph{Goal.} With the batch dataset $\mathcal{D}$ in hand, the goal of offline RL in this setting is to learn, in a data-efficient manner, a policy $\widehat{\pi}$ that attains near-optimal value with respect to a prescribed test state distribution $\rho\in \Delta(\cS)$ (which is {\em not} known to the learner {\em a priori}). More precisely, for a prescribed precision level $\varepsilon$, we seek to identify an $\varepsilon$-optimal policy $\widehat{\pi}$ satisfying
\begin{align}
	V^\star(\rho) - V^{\widehat{\pi}}(\rho) \leq \varepsilon
\end{align}
with high probability.  A desirable offline RL algorithm would achieve this target accuracy using as few samples from $\mathcal{D}$ as possible. 
As in Chapters~\ref{chapter:generative-model} and \ref{chapter:online-RL}, particular emphasis is placed on attaining minimal sample complexity uniformly across the full range of accuracy levels, namely for all $\varepsilon \in \big(0,\frac{1}{1-\gamma}\big]$.

% For the most part, the historical dataset is generated by a behavior policy that differs from the desirable optimal policy. 

\paragraph{Challenges.} 
Unlike online RL, offline RL precludes active interaction with the environment to collect additional data. Consequently, the learner must rely entirely on the static dataset $\mathcal{D}$. The statistical efficiency of offline RL is therefore fundamentally dictated by how informative this historical dataset is for learning a near-optimal policy. Below, we highlight two fundamental challenges arising from this restricted data generation mechanism, both of which must be carefully addressed in order to achieve statistically efficient offline RL.

\begin{itemize}

\item {\em Distribution shift.}
For the most part,  the historical dataset is generated by one or more behavior policies that differ from the target optimal policy. As a result, the state-action distribution induced by the target policy may deviate substantially from the sampling distribution $d^{\mathsf{b}}$ underlying the dataset $\mathcal{D}$. A central challenge in offline RL is therefore how to effectively leverage past data despite this mismatch between the data distribution and the one induced by the optimal policy.

\item {\em Limited data coverage.} 
	Ideally, one would like the dataset $\mathcal{D}$ to contain sufficiently many samples for every state-action pair, so that the performance of all candidate policies can be accurately evaluated. In practice, however, such uniform coverage is often unrealistic, since the learner has no control over the past data collection process. Fortunately, uniform coverage may be unnecessary, given that the ultimate objective is only to identify a single near-optimal policy rather than accurately evaluate every policy.  

\end{itemize}

%\noindent 
%All in all, effectively coping with distribution shift and limited data coverage is central to the feasibility and statistical efficiency of offline RL, and forms a major focus of this chapter.

\paragraph{Key parameters: single-policy concentrability.}
To quantitatively capture the distribution shift between the target occupancy measure and the underlying data distribution, we introduce a key quantity previously introduced in \citet{rashidinejad2021bridging}.
\begin{definition}[Single-policy concentrability for infinite-horizon MDPs] 
\label{assumption:concentrate-infinite-simple}
For any given initial distribution $\rho\in \Delta(\cS)$, the single-policy concentrability coefficient underlying the batch dataset $\mathcal{D}$ (see \eqref{eq:sampling-offline-inf-iid}) is defined as
\begin{align}
	\Cstar \coloneqq \max_{(s, a) \in \cS \times \cA}\, \frac{d^{\star}(s, a; \rho)}{\myrho(s, a)} , 
	\label{eq:concentrate-infinite-simple}
\end{align}
where $d^{\star}$ represents the discounted occupancy distribution induced by the optimal deterministic policy $\pi^{\star}$ (see \eqref{eq:infinite-d-star-def}), and the ratio 0/0 is interpreted as 0.  
	%The smallest quantity $C^{\star}$ obeying \eqref{eq:concentrate-infinite-simple} is called the . 
\end{definition}
\begin{remark}
\label{rem:Cstar_lower_bound_infinite}
It is immediate to see that $C^{\star} _{\rho} \geq 1$, since
\[
C^{\star} _{\rho} = \max_{(s, a) \in \cS \times \cA}\, \frac{d^{\star}(s, a; \rho)}{\myrho(s, a)} 
\geq \frac{\sum_{s,a}d^{\star}(s, a; \rho)}{\sum_{s,a}\myrho(s, a)} = \frac{1}{1}=1.
\]
\end{remark}
In words, $C^{\star} _{\rho}$ quantifies the distribution mismatch between the target occupancy distribution $d^{\star}$ (induced by the optimal policy starting from the initial distribution $\rho$) and the underlying data distribution $d^{\mathsf{b}}$ through the maximum density ratio. 
Since the value function in \eqref{eq:defn-Vpi-inf} is defined in a discounted manner, the occupancy distribution $d^{\star}$ is likewise chosen to be discounted. When $C^{\star} _{\rho}$ is close to $1$, the batch dataset $\mathcal{D}$ may be viewed as containing near-expert data, meaning that its induced distribution closely aligns with that induced by the optimal policy and therefore reflects, to some extent, how an expert makes decisions. In this favorable scenario, the offline RL problem becomes intimately connected to the problem of ``imitation learning,'' where it may suffice to simply imitate the expert's behavior.

Moreover, $C^{\star} _{\rho}$ is referred to as a ``single-policy'' concentrability coefficient because it concerns only the optimal policy $\pi^{\star}$. This assumption is substantially weaker than a uniform coverage requirement, or more precisely, the all-policy concentrability assumption adopted in earlier work, e.g., \citet{munos2007performance,farahmand2010error,fan2019theoretical,chen2019information}. Note that the latter requires a uniform density-ratio bound over all policies and therefore demands the dataset to be highly exploratory, or equivalently, to provide adequate coverage of all state-action pairs. 

Given that $C^{\star} _{\rho}$ is introduced to capture the challenges arising from both distribution mismatch and limited data coverage, we will treat it as a key problem parameter throughout the remainder of this chapter, and develop both sample complexity upper bounds and minimax lower bounds primarily in terms of $C^{\star} _{\rho}$.

%Similar to the finite-horizon counterpart, 
%$C^{\star}$ employs the largest density ratio to measure the distribution mismatch;
%it concerns the behavior policy vs.~a single policy $\pi^{\star}$, and does not require uniform coverage of the state-action space (namely, it suffices to cover the part reachable by $\pi^{\star}$). 
%Given that the value function \eqref{eq:defn-Vpi-inf} is defined in a discounted fashion, 
%the two distributions to be compared are chosen to be the discounted occupancy distributions. 
%As before, we further introduce a slightly modified version of $C^{\star}$ as follows. 

%
%Assumption~\ref{assumption:concentrate-finite},
%we introduce a certain concentrability assumption below to capture the distribution shift between the desired distribution and the one induced by the behavior policy. 
%

\begin{remark}
A refined version of $C^{\star} _{\rho}$, called the {\em single-policy clipped concentrability} coefficient \citep{li2024settling},  can be obtained via clipping: 
\begin{align}
C^{\star}_{\rho,\mathsf{clipped}} \coloneqq \max_{(s, a) \in \cS \times \cA}\frac{\min\big\{d^{\star}(s, a; \rho), \frac{1}{S}\big\}}{\myrho(s, a)} .
\label{eq:concentrate-infinite}
\end{align}
	Clearly, one has $C^{\star}_{\rho,\mathsf{clipped}} \leq C^{\star} _{\rho}$. In addition, unlike $C^{\star} _{\rho}$, which always satisfies $C^{\star} _{\rho}\geq 1$, the clipped coefficient can be as low as $1/S$, a property that can sometimes lead to improved statistical guarantees, especially when $\myrho$ is close to a uniform distribution, such as the generative model setting. 
	As it turns out, many of the upper and lower bounds developed in this chapter continue to hold with $C^{\star} _{\rho}$ replaced by $C^{\star}_{\rho,\mathsf{clipped}}$; see \citet{li2024settling} for details. 
\end{remark}

\section{A detour: the \LCB algorithm for multi-armed bandits}

 Before turning to offline RL in full generality, it is instructive to first examine a much simpler setting that already captures two central challenges: distribution mismatch and limited data coverage. To build intuition, we begin with the classical multi-armed bandit problem to introduce the principle of pessimism in the face of uncertainty --- a principle that will play a central role more broadly in offline RL. 

\subsection{Setting: offline stochastic bandits}
The basic setup of the multi-armed bandit problem has been introduced in Section~\ref{sec:bandits-UCB}. We briefly recap the setting and describe the corresponding offline data-generation mechanism.
\begin{itemize}
    \item {\em Bandits.} Consider an $A$-armed stochastic bandit setting, where each arm $a\in \mathcal{A}=\{1,\dots,A\}$ is associated with a mean reward $r_a\in [0,1]$.  The optimal mean reward and the corresponding optimal arm are denoted by $r^{\star}$ and $a^{\star}$, respectively (see \eqref{eq:defn-rstar-astar-bandits}).  

    \item {\em Offline data.} The offline dataset $\cD$ consists of $N$ samples $\{(a_i, R_i)\}_{i=1}^{N}$ generated independently according to some joint distribution. More concretely, each action $a_i$ is sampled as
    \begin{subequations}
    \begin{align}
        a_i \overset{\mathrm{i.i.d.}}{\sim}d^{\mathsf{b}},
        \qquad i=1,\cdots,N
    \end{align}
    for some action distribution $d^{\mathsf{b}}\in \Delta(\mathcal{A})$,  after which the corresponding (noisy) reward $R_i$ is generated according to
    \begin{align}
        %\mathbb{E}[R_t \mid a_t = a] = r_a. 
        R_i = r_{a_i} + z_i
        \qquad \text{with }z_i \overset{\mathrm{i.i.d.}}{\sim} \mathcal{N}(0,1).
    \end{align}
    \end{subequations}

\item {\em Single-policy concentrability.}
Let $d^{\star}$ denote the distribution induced by the optimal policy, i.e., always pulling the optimal arm $a^{\star}$, 
\begin{subequations}
\label{assumption:MAB_general}
\begin{equation}
    d^{\star}(a) \coloneqq \begin{cases}
        1, \quad & \text{if }a=a^{\star},\\
        0, & \text{else}. 
    \end{cases}
\end{equation}
Then the single-policy concentrability coefficient defined in~\eqref{eq:concentrate-infinite-simple} reduces to
\begin{align}
    C^{\star} \coloneqq \max_{a\in \cA} \frac{d^{\star}(a)}{d^{\mathsf{b}}(a)}=\frac{1}{d^{\mathsf{b}}(a^{\star})}.
    \label{eq:defn-Cstar-bandits}
\end{align}
\end{subequations}
Here, we drop the dependence on $\rho$ in the notation $C^{\star} _{\rho}$, since the bandit setting contains only a single state. 
\end{itemize}

%In the offline setting, the offline dataset $\cD$ consists of tuples $\{(a_t, R_t)\}_{t=1}^{N}$ sampled independently from some joint distribution, and the mean reward of arm $a\in \mathcal{A}$ is $r_a \coloneqq \mathbb{E}[R_t \mid a_t = a]$. Denote the marginal distribution of action $a_t$ as $ d^{\mathsf{b}} $.  

\paragraph{Goal.} 
Armed with the offline dataset $\mathcal{D}$, our goal is to learn an arm $\widehat{a}$ that achieves a suboptimality gap
\begin{equation*}
    %\mathbb{E}_{\cD}[
    r(a^\star) - r(\widehat{a})
    %]
\end{equation*}
as small as possible. Note that this notion of suboptimality also plays a central role in online multi-armed bandits through the definition of regret; see \eqref{eq:regret_bandit} in Section~\ref{sec:bandits-UCB}.

%In the language of regret minimization, this corresponds to a regret equal to $T \mathbb{E}_{\cD}[r(a^\star) - r(\widehat{a})]$ if $\widehat{a}$ is played for $T$ rounds. 

\subsection{Pessimism in the face of uncertainty}

We now introduce some core ideas that enable statistically efficient offline learning in bandit problems. In contrast to online bandits, which rely heavily on the optimism principle (see Section~\ref{sec:bandits-UCB}) to guide exploration, 
offline settings do not permit active exploration. As a result, optimism-based approaches are inherently inapplicable, necessitating a fundamentally different set of ideas.

\paragraph{A first attempt: the empirical best arm.}
A natural strategy for identifying the optimal arm is to select the one with the highest empirical mean reward. Denoting by $N_a$ the number of times arm $a$ is pulled in $\cD$, the empirical reward estimate for each arm $a\in \cA$ is given by
\begin{subequations}
\label{eq:empirical-best-arm-offline}
\begin{equation}
    \overline{r}_a  \coloneqq\begin{dcases}
    \frac{1}{N_a}\sum_{t=1}^{N} R_{t} \ind \{a_t = a\},
     & \text{if }N_a > 0,\\
    0, & \text{otherwise},
    \end{dcases}
\end{equation}
which coincides with the maximum likelihood estimator of the mean reward under the Gaussian noise assumption. 
The empirical best arm is then defined as
\begin{equation}
a^{\mathsf{emp}} \coloneqq \arg\max_{a\in \cA} \overline{r}_a.
\end{equation}
\end{subequations}

Although intuitive, this empirical best-arm estimator can perform poorly when certain arms are only infrequently sampled in the dataset, making their mean rewards difficult to estimate reliably. In fact, one can construct examples showing that even in the favorable regime where $C^\star \approx 1$, selecting the empirical best arm may still incur a non-vanishing suboptimality gap regardless of how large the sample size becomes. This issue arises from the high statistical uncertainty associated with poorly explored suboptimal arms, which can lead to over-estimation of their rewards with non-vanishing probability. One concrete example is given below. 
\begin{example}
\label{example:offline-bandits}
Suppose that $\mathcal{A}=\{1,2\}$, and that the corresponding mean rewards are given by $r_1=1$ and $r_2=0$. Assume that $a_i = 1$ for all $i=1,\dots,N-1$, and $a_{N}=2$. Then for any $N\ge 2$, we have
\[
    \mathbb{E}[r^{\star} -r_{a^{\mathsf{emp}}}] \geq 0.075.
\]
\end{example}
\begin{remark}
In this example, the optimal arm is arm $1$, which is sampled $N-1$ times out of a total of $N$ observations. Consequently,  the corresponding $C^{\star}$ is exceedingly close to $1$.
\end{remark}
\begin{proof}[Proof of Example~\ref{example:offline-bandits}]
Under the reward model, the empirical reward estimates satisfy  $\overline{r}_1\sim\mathcal{N}(1,\frac{1}{N-1})$ and $\overline{r}_2\sim\mathcal{N}(0,1)$, and hence 
$$
\mathbb{P}(\overline{r}_1 < 1) = 0.5 \qquad 
\text{and} \qquad
\mathbb{P}(\overline{r}_2 \geq 1) \geq 0.15. 
$$
It then follows from the statistical independence between $\overline{r}_1$ and $\overline{r}_2$ that 
\begin{align*}
 \mathbb{E}[r_1-r_{a^{\mathsf{emp}}}] = \mathbb{P}(\overline{r}_2>\overline{r}_1) 
 \geq \mathbb{P} (\overline{r}_1 < 1) \mathbb{P}(\overline{r}_2 \ge 1) \geq 0.075
\end{align*}
as claimed. 
\end{proof}

\paragraph{The lower confidence bound (\LCB) algorithm.} 
As illustrated in Example~\ref{example:offline-bandits}, the main issue with the empirical best-arm estimator is the systematic over-estimation of the rewards of poorly sampled arms, caused by large variance. A natural strategy toward mitigating this issue is therefore to properly adjust the empirical reward estimates by accounting for variance information.

This idea leads to the principle of pessimism (or conservatism) in the face of uncertainty: arms with fewer samples should be treated more conservatively, since their empirical rewards are more volatile and less reliable. As a concrete method, one can penalize the empirical reward estimate by a penalty function $b_a$ that decreases as the sample count $N_a$ (i.e., the number of times arm $a$ is sampled) increases. Rather than selecting the arm with the largest empirical reward, the pessimism principle instead recommends choosing the arm according to the lower confidence bound:
\begin{equation*}
a^{\lcb} = \arg\max_{a\in \cA} r_{a}^{\lcb}
\qquad 
\text{with } 
   r_{a}^{\lcb} = \overline{r}_{a} - b_a. 
\end{equation*}
A concrete choice of the penalty function is 
\begin{align}
b_a = \sqrt{\frac{2\log\big(\frac{4A}{\delta}\big)}{N_{a}}}
\label{eq:penalty-bandit-illustrate}
\end{align}
with a prescribed confidence parameter $\delta \in (0,1)$, 
which serves as a confidence width derived from standard Gaussian tail bounds and resembles the exploration bonus (cf.~\eqref{eq:UCB-reward-bandit-alg}) adopted in the UCB algorithm. 
Intuitively, the quantity $r^{\lcb}_a$ can be viewed as a lower confidence bound on the true reward $r_a$, motivating the name \LCB algorithm. The full procedure is summarized in Algorithm~\ref{alg:MAB-LCB}.

Careful readers may notice that the \LCB algorithm bears some resemblance to the \UCB algorithm presented in Section~\ref{sec:bandits-UCB}. However, whereas \UCB adds an exploration bonus to encourage exploration in online settings, the \LCB approach subtracts a confidence penalty to guard against overestimation in the absence of exploration.

\begin{remark}
As a sanity check, when applying Algorithm~\ref{alg:MAB-LCB} to Example~\ref{example:offline-bandits}, we can readily verify that $r_1^{\lcb} \approx r_1=1$ for large enough $N$, and $r_2^{\lcb} \le r_2=0$. Consequently, the \LCB algorithm selects arm 1 (the optimal arm) and rules out arm 2 with high probability. Part of the reason this approach works well in the present example is that the algorithm effectively ``trusts'' the preference of the offline dataset: since the optimal arm is sampled frequently in the data, the estimate of $r_1$ becomes highly reliable, allowing the algorithm to confidently distinguish it from the suboptimal arm. 
\end{remark}

\begin{algorithm}[t]
\DontPrintSemicolon
\caption{The lower confidence bound (\LCB) algorithm for multi-armed bandits}
\label{alg:MAB-LCB}
% \begin{algorithmic}[1]

\textbf{input:} batch dataset $\cD = \{(a_i, R_i)\}_{i=1}^N$, number of arms $A$, confidence parameter $\delta \in (0,1)$. \\

\For{$a \in \cA$}{
set $N_a \leftarrow \sum_{i=1}^N \ind\{a_i = a\}$. {\color{blue}\tcc{visitation count.}} 
%\If{$N_a = 0$}{
%set $\overline{r}_a \leftarrow 0$ and $b_a \leftarrow 1$. %}
%\Else{
compute empirical  reward 
$$\overline{r}_a =\begin{cases} 
\frac{1}{N_a}\sum_{i=1}^{N}R_{i} \ind \{a_{i} = a\}, & \mbox{if}~ N_a > 0, \\
0, & \mbox{otherwise}.
\end{cases}  
$$ 

set $b_a \leftarrow  \sqrt{\frac{2\log(4A/\delta)}{N_a}}$. {\color{blue}\tcc{LCB penalty.}}
}
%}
\textbf{output:} $a^{\lcb} = \arg\max_a r_a^{\lcb}$, where $r_a^{\lcb}=\overline{r}_a - b_a$.
% \end{algorithmic}
\end{algorithm}

\subsection{Theoretical guarantees}

We now turn to the statistical guarantees of the \LCB algorithm. As it turns out, the pessimism principle described above proves highly effective in coping with the challenges inherent to offline settings. This is formalized in the following theorem, originally established by \citet{rashidinejad2021bridging}.

\begin{theorem}
%[\citet{rashidinejad2021bridging}]
\label{thm:MAB_LCB_upper}
Suppose that the sample size satisfies $N \geq 8 C^\star \log (2/\delta)$. Then with probability at least $1-\delta$, the action $a^{\lcb}$ returned by the \LCB algorithm (cf.~Algorithm~\ref{alg:MAB-LCB}) with penalty function \eqref{eq:penalty-bandit-illustrate} achieves 
\begin{equation}\label{eq:expect-MAB}
    %\E_\cD[
    r^\star - r_{a^{\lcb}}
    %] 
    %\lesssim 
    %\min\Bigg\{ 
\le 4\sqrt{\frac{C^{\star}\log(4A/\delta)}{N }}.
    %,\, 1\Bigg\}. 
\end{equation}
\end{theorem}

Before embarking on the proof of this result, we make several remarks. To achieve any target accuracy $0<\varepsilon<1$ (in the sense that $r^\star - r_{a^{\lcb}}\leq \varepsilon$), it suffices for the \LCB algorithm to use
\begin{equation}
	\widetilde{O}\bigg( \frac{C^{\star}}{\varepsilon^2} \bigg) 
\end{equation}
samples from the offline dataset $\mathcal{D}$. This stands in sharp contrast to the empirical best-arm estimator, which in certain scenarios may be fundamentally incapable of achieving $o(1)$ statistical error. This theorem therefore provides rigorous evidence that the pessimism principle can significantly outperform the vanilla empirical approach. 

Notably, the sample complexity of the \LCB algorithm scales linearly with the single-policy concentrability coefficient, without requiring this coefficient to be known {\em a priori}. This unveils that the statistical efficiency of the pessimistic approach automatically adapts to the intrinsic quality of the offline dataset, as characterized by $C^{\star}$, rather than the number of arms $A$.

% From Theorem~\ref{thm:MAB_LCB_upper}, one can see that the \lcb algorithm achieves a sub-optimality gap on the order of $\sqrt{(\log N)/ N}$ that avoids the constant error that encountered by the empirical best arm.  This demonstrates the benefit of pessimism over the vanilla approach. Intuitively, the \LCB approach applies larger penalties to the actions that are less observed.  Even if we have actions with huge fluctuations in their respective empirical rewards due to a small number of samples, the penalty term helps to rule them out. 

\paragraph{Intuitive explanation.} 
Prior to delving into the proof of Theorem~\ref{thm:MAB_LCB_upper}, it is instructive to first develop some intuition for the result. 
Suppose, for simplicity, that every suboptimal arm $a$ satisfies $r_a< r^{\star}-\varepsilon$, so that achieving $\varepsilon$-accuracy amounts to ensuring that any suboptimal arm is unlikely to be selected.

Recall from the definition \eqref{eq:defn-Cstar-bandits} that $C^\star = 1/d^{\mathsf b}(a^\star)$. Consequently, the number of observations collected for the optimal arm $a^\star$ scales as
\[
N_{a^{\star}} \asymp Nd^{\mathsf{b}}(a^{\star}) \asymp N/C^{\star}.
\]
As a result, when the sample size $N$ is sufficiently large, both the empirical estimate $\overline r_{a^\star}$ and the corresponding lower confidence bound $r_{a^\star}^{\lcb}$ concentrate tightly around the true reward:
\[
\overline{r}_{a^{\star}} \approx r^{\lcb}_{a^{\star}} 
= r_{a^{\star}} - O\bigg(\frac{1}{\sqrt{N_{a^{\star}}}} \bigg)
= r^{\star} - O\Bigg(\sqrt{\frac{C^{\star}}{N}}\Bigg), 
\]
where logarithmic factors are omitted for simplicity.  
Therefore, the lower confidence bound $r_{a^\star}^{\lcb}$ can be statistically separated from the mean reward $r_a$ of any suboptimal arm --- and hence also from its lower confidence bound $r_a^{\lcb}$ --- provided that
\[
\sqrt{\frac{C^{\star}}{N}} \ll \varepsilon 
\qquad \Longleftrightarrow \qquad
N\gg \frac{C^{\star}}{\varepsilon^2}. 
\]
This heuristic captures the key intuition underlying the analysis.

\subsection{Analysis: proof of Theorem~\ref{thm:MAB_LCB_upper}}
\label{app:proof_MAB_LCB_upper}

We begin by establishing two high-probability events. First, akin to the application of the Gaussian tail bounds~\eqref{eq:Gaussian-deviation-standard-to-reward-forward-reverse} in the online setting, one can show that
\begin{equation}\label{eq:bonus-sandwich-MAB}
    r^{\lcb}_a = \overline{r}_a - b_a \leq r_a \leq \overline{r}_a + b_a \qquad \text{for all } 1\leq a\leq A,
\end{equation}
with probability at least $1-\delta/2$. This guarantees that $r^{\lcb}_a$ is indeed a pessimistic estimate of the true mean reward $r_a$. Second, the concentration bound in Lemma~\ref{lem:tail-bound-sum-Bernoulli} tells us that 
\begin{equation}\label{eq:N-lower-bound-MAB}
    N_{a^\star} \geq \mathbb{E}[N_{a^\star}]  - \frac{1}{2}N d^{\mathsf{b}} (a^\star) = \frac{1}{2}N d^{\mathsf{b}} (a^\star)
\end{equation}
with probability at least $1 - \exp(-N d^{\mathsf{b}} (a^\star) / 8)$. Consequently, if $N\geq 8C^{\star}\log(2/\delta)$, then \eqref{eq:N-lower-bound-MAB} holds with probability exceeding $1-\delta/2$.

Conditioning on the intersection of the two events above, we proceed to compare the optimal arm \(a^\star\) with the pessimistic choice \(a^{\lcb}\). It is readily seen from \eqref{eq:bonus-sandwich-MAB} that 
\[
r_{a^\star}\leq\overline{r}_{a^\star}+b_{a^\star}=\overline{r}_{a^\star}-b_{a^\star}+2b_{a^\star}.
\]
In view of the definition of $a^{\lcb}$, we have $\overline{r}_{a^\star}-b_{a^\star}\leq\overline{r}_{a^{\lcb}}-b_{a^{\lcb}}$,
which combined with  \eqref{eq:bonus-sandwich-MAB} again yields
\begin{align}
r_{a^\star}\leq\overline{r}_{a^{\lcb}}-b_{a^{\lcb}}+2b_{a^\star}\leq r_{a^{\lcb}}+2b_{a^\star}.
\label{eq:gap-astar-lcb-bandit}
\end{align}

Therefore, to bound the suboptimality gap $r_{a^\star} - r_{a^{\lcb}}$, it suffices to control the penalty term $b_{a^\star}$, which reduces to lower bounding $N_{a^\star}$. By virtue of~\eqref{eq:N-lower-bound-MAB} and the definition  $C^{\star}=1/d^{\mathsf{b}}(a^{\star})$ (cf.~\eqref{eq:defn-Cstar-bandits}), we have 
\[
N_{a^\star}\geq\frac{1}{2}N d^{\mathsf{b}} (a^\star)= \frac{N}{2C^\star}>0,
\]
which further implies that
\[
b_{a^\star}=\sqrt{\frac{2\log(4A/\delta)}{N_{a^\star}}}\leq 
\sqrt{\frac{4C^{\star}\log(4A/\delta)}{N }}. 
\]
Substituting this into \eqref{eq:gap-astar-lcb-bandit}, we arrive at
\begin{equation*}
r_{a^\star} - r_{a^{\lcb}}\leq 2b_{a^\star}
\le 4\sqrt{\frac{C^{\star}\log(4A/\delta)}{N }}
%\label{eq:clean-event-consequence-MAB}
\end{equation*}
with probability at least $1-\delta$. This establishes Theorem~\ref{thm:MAB_LCB_upper}.

\section{Model-based offline RL}
\label{sec:VI-LCB-infinite-horizon}

Thus far, we have demonstrated the effectiveness of the pessimism principle in the simpler setting of offline multi-armed bandits. This naturally motivates extending the pessimism principle to offline RL, with the goal of addressing the challenges of distribution mismatch and limited data coverage in much more general offline settings.

As it turns out, despite the different sources of uncertainty in offline bandits and offline RL --- namely, noisy rewards in the former and unknown transition models in the latter --- the pessimism principle can be incorporated into both model-based and model-free approaches in a principled manner. Given the minimax optimality of model-based methods in both RL with a generative model and online RL, it is natural to first investigate model-based offline RL algorithms and their statistical optimality, which will be the focus of this section.

\subsection{Algorithmic ideas}

A natural approach to incorporating pessimism into offline RL is to penalize the value estimates of state-action pairs that receive poor coverage in the offline dataset $\mathcal{D}$. Intuitively, when a state-action pair is rarely observed, the corresponding value estimate is subject to high uncertainty and should therefore be treated with caution. Pessimistic value estimation addresses this challenge by systematically lowering value estimates in poorly covered regions of the state-action space, with the amount of penalization calibrated according to the degree of uncertainty.

%As we will shortly demonstrate, a proper use of pessimism yields several provable benefits: (i) it enables sample complexity guarantees that adapt to the degree of distribution shift; and (ii) unlike approaches requiring uniform data coverage, it only necessitates sufficient coverage of the portion of the state-action space reachable under the target policy.

In what follows, we introduce a model-based offline RL algorithm that integrates lower confidence bounds into the value estimation procedure,  building on ideas originally developed in \citet{jin2021pessimism,rashidinejad2021bridging}. Commonly referred to as \vilcb, this algorithm performs value iteration over the empirical MDP through a pessimistic Bellman operator. We now describe its core algorithmic components.

\begin{itemize}

\item {\em Model estimation.} As in the setting of model-based RL with a generative model discussed in Chapter~\ref{chapter:generative-model}, our first step is to build an empirical estimate $\widehat{P}$ of the true transition kernel $P$ underlying the MDP.  Recall that the offline dataset contains $N$ independent sample transitions $\{(s_i, a_i, s_i')\}_{i = 1}^N$ (see \eqref{eq:sampling-offline-inf-iid}). 
For each state-action pair $(s,a)$, denote by 
\begin{subequations}
\label{eq:empirical-N-P-infinite}
\begin{align}
N_{s,a}  \coloneqq \sum_{i = 1}^{N} \ind\big((s_i, a_i) = (s, a)\big)
	\label{eq:defn-Nsa-infinite}
\end{align}
the number of times $(s, a)$ appears in the dataset. Based on these visitation counts, we define the empirical transition kernel $\widehat{P}$ by
\begin{align}
	\widehat{P}(s'\mymid s,a) = 
	\begin{cases} \frac{1}{N_{s,a} } \sum\limits_{i=1}^N \mathds{1} \big\{ (s_i, a_i, s_i') = (s,a,s') \big\}, & \text{if } N_{s,a}  > 0 \\
		\frac{1}{S}, & \text{else}
	\end{cases}  
	\label{eq:empirical-P-infinite}
\end{align}
\end{subequations}
for each $(s,a,s')\in \cS\times \cA \times \cS$. 
The uniform assignment in the case $N_{s,a}=0$ is introduced merely to ensure that 
$\widehat{P}(\cdot \mid s,a)$ remains a valid probability distribution even for unseen state-action pairs.

\item {\em The pessimistic Bellman operator.} 
At the core of the \vilcb algorithm lies a modified Bellman operator whose fixed point produces conservative value estimates.  
Recall that, given a transition kernel $P$, 
the classical Bellman operator  $\mathcal{T}(\cdot): \mathbb{R}^{SA}\rightarrow \mathbb{R}^{SA} $ is defined as follows:  
for any input vector $Q \in \mathbb{R}^{SA}$ and its induced value vector $V \in \mathbb{R}^S$ (i.e., $V(s) \coloneqq \max_a Q(s, a)$ for all $s$), 
\begin{align}
	\mathcal{T}(Q)(s, a) \coloneqq r(s, a) + \gamma P_{s, a}V \qquad\text{for all }(s, a)\in\cS\times\cA .
	\label{eq:classical-Bellman-operator-inf}
\end{align}

To incorporate pessimism into the value estimation procedure, we replace the true transition kernel \(P\) with the empirical kernel \(\widehat{P}\) (cf.~\eqref{eq:empirical-P-infinite}) and introduce an additional penalty term that accounts for statistical uncertainty. Specifically, we define the pessimistic Bellman operator \(\Tpess\) such that, for every $(s, a)\in\cS\times\cA$,  
\begin{align}
	\Tpess (Q)(s, a) \coloneqq \max\Big\{r(s, a) + \gamma\widehat{P}_{s, a} V - b(s, a; V) , 0\Big\}.
	\label{eq:empirical-Bellman-infinite}
\end{align}
Here, $b(s, a; V)$ stands for a penalty function designed to capture the uncertainty associated with estimating the Bellman update at \((s,a)\), whose precise form will be specified momentarily.  The \vilcb algorithm then seeks to compute the fixed point of this pessimistic Bellman operator.

\item {\em Modified value iterations.} 
A natural scheme for computing the fixed point of the pessimistic Bellman operator $\Tpess$ (cf.~\eqref{eq:empirical-Bellman-infinite}) is to employ a value-iteration-style iterative procedure. Specifically, analogous to standard value iteration~\eqref{eq:value_iteration}, we repeatedly apply the operator \(\Tpess\) starting from the initialization \(\widehat{Q}_0 = 0\). For each iteration \(\tau = 1,2,\ldots,\tau_{\max}\), we update 
\begin{align}
	\widehat{Q}_{\tau}(s, a) & = \Tpess\big(\widehat{Q}_{\tau-1}\big)(s, a) \nonumber \\
	& = \max\Big\{r(s, a) + \gamma\widehat{P}_{s, a}\widehat{V}_{\tau-1} - b\big(s, a; \widehat{V}_{\tau-1} \big), 0\Big\},
	%\qquad 
	\label{eq:VI-LCB-iterations-basic-inf}
\end{align}
where for all $s\in \cS$, 
\[
\widehat{V}_{\tau-1}(s)
\coloneqq
\max_a \widehat{Q}_{\tau-1}(s,a).
\]
After $\tau_{\max}$ iterations, the algorithm outputs $\widehat{Q}= \widehat{Q}_{\tau_{\max}}$ as the final Q-estimate. 
The resulting policy estimate \(\widehat{\pi}\) is then obtained greedily with respect to \(\widehat{Q}\), namely, 
\begin{align}
	\widehat{\pi}(s) \in \arg\max_a \widehat{Q}(s, a) \qquad \text{for all }s\in \cS.
\end{align}

\end{itemize}

\noindent 
The whole algorithm is summarized in Algorithm~\ref{alg:vi-lcb-infinite}.  

Let us briefly provide some high-level intuition underlying this algorithmic design. When a state-action pair $(s,a)$ is insufficiently covered in the offline dataset $\mathcal{D}$ (that is, when the visitation count $N_{s,a}$ is small), the corresponding estimate 
$\widehat{Q}_{\tau}(s,a)$ suffers from high variance and therefore becomes unreliable; if not carefully handled, such uncertainty can propagate through the value iterations and adversely affect the overall value estimation process. To mitigate this issue, the \vilcb algorithm introduces a penalty function $b(s,a; \widehat{V}_{\tau-1})$ derived from suitable lower confidence bounds, which effectively downweights poorly visited state-action pairs and limits their negative influence on the learned value function.  Importantly, when the coverage coefficient $\Cstar$ is small, suppressing these poorly covered regions does not necessarily introduce significant bias in estimating the optimal value function. For instance, when the behavior policy $\pib$ generating $\mathcal{D}$ aligns closely with the optimal policy $\pi^{\star}$, the poorly covered regions typically correspond to suboptimal actions that are rarely visited under $\pib$, making it acceptable to assign conservative estimates to such pairs.

\subsection{Choices of penalty functions}
As one might expect, the properties of the fixed point of $\Tpess$ --- and consequently the statistical performance of \vilcb --- depend critically on the choice of the penalty function. The role of $b(s,a; V)$ is to help construct a lower confidence bound for the true Bellman update $ r(s,a)+\gamma \langle P_{s,a}, V\rangle$ based on its empirical counterpart (see \eqref{eq:VI-LCB-iterations-basic-inf})
\[
    r(s,a)+\gamma \langle \widehat{P}_{s,a}, V\rangle.
\]
Accordingly, the penalty term $b(s,a; V)$ should be chosen to characterize the confidence width of the estimation error of the form 
\[
    \big\langle \widehat{P}_{s,a} - P_{s,a},  V \big\rangle,
\]
where $V\in \mathbb{R}^S$ is some vector obeying $V(s)\in \big[0,\frac{1}{1-\gamma}\big]$. 
Similar to the design of the exploration bonus in \ucbvi (see Section~\ref{sec:choice-bonus-online}), there are at least two commonly used choices of penalty functions for this purpose.
\begin{itemize}
    \item {\em Hoeffding-style penalty.}
    The first type of penalty functions, adopted for instance by \citet{rashidinejad2021bridging}, is motivated by Hoeffding-type concentration inequalities together with the boundedness property $Q^{\star}(s,a)\in \big[0,\frac{1}{1-\gamma}\big]$. Specifically, one can choose
	\begin{equation*}
		b(s,a; V) =\widetilde{\Theta} \bigg( \sqrt{\frac{1}{(1-\gamma)^2N_{s,a}}} \bigg),
	\end{equation*}
		which admits a relatively simple form and is often amenable to analysis. 
        %than its Bernstein-style counterpart. 
        However, since it does not leverage variance information, the resulting statistical guarantees are generally suboptimal.

    \item {\em Bernstein-style penalty.}
   Similar to the online RL setting, employing a Bernstein-style penalty to characterize the confidence width often leads to improved statistical performance, owing to its ability to properly exploit the underlying variance structure of the MDP of interest. More specifically, the Bernstein-style penalty typically takes the form
\begin{align*}
	b(s,a; V) 
	=\widetilde{\Theta} 
	\Bigg( \sqrt{\frac{   \mathsf{Var}_{\widehat{P}_{s,a} }(V)   }{ N_{s,a} }}
						+ \frac{1}{ (1-\gamma)N_{s,a}   }  \Bigg) ,
    &
	%\label{eq:bh-defn-Bernstein-ucbvi-motivate}
\end{align*}
where the variance term appearing in the Bernstein inequality is replaced by its empirical counterpart in order to make the penalty computable. Compared with the Hoeffding-style counterpart, this choice yields a substantially sharper confidence width by adapting to the local variability of the value function. As we shall justify shortly, such variance-aware penalization enables \vilcb to achieve the minimax-optimal sample complexity for offline RL.

\end{itemize}

\noindent In this section, we focus on the following concrete choice of Bernstein-style penalty functions:
\begin{align}
	%\forall (s,a) \in \cS\times \cA:
	%\quad 
	b(s, a; V) \defn \min \Bigg\{  \max \Bigg\{ & \sqrt{\frac{\cb\log\frac{N}{(1-\gamma)\delta}}{N_{s,a} }\mathsf{Var}_{\widehat{P}_{ s, a}} (V )} , \arxiv{ \nonumber\\ 
	 & } \,\frac{2\cb \log\frac{N}{(1-\gamma)\delta}}{(1-\gamma)N_{s,a} } \Bigg\} ,\, \frac{1}{1-\gamma}  \Bigg\} +\frac{5}{N} 
	\label{def:bonus-Bernstein-infinite}
\end{align}
for every $(s,a)\in \cS\times \cA$, 
where $\cb> 0$ is some numerical constant, and $\delta \in (0,1)$ is some confidence parameter. 
When $N_{s,a}=0$, we adopt the convention
$b(s,a;V)=1/(1-\gamma)+5/N$. 
As usual,  $\mathsf{Var}_{\widehat{P}_{s,a}} (V)$ denotes the variance of the vector $V\in \mathbb{R}^S$ under the empirical transition distribution \(\widehat{P}_{s,a}\); see~\eqref{eq:defn-Var-Psa-V}.

\begin{algorithm}[t]
	\textbf{input:} dataset $\mathcal{D}$, discount factor $\gamma$, reward function $r$, confidence parameter $\delta$, max iteration number $\tau_{\max}$. \\
	\textbf{initialize} $\widehat{Q}_{0}=0$, $\widehat{V}_0=0$.  \\
{\color{blue}\tcc{model estimation based on offline dataset.}}
	construct the empirical transition kernel $\widehat{P}$ according to \eqref{eq:empirical-N-P-infinite}. \\
{\color{blue}\tcc{pessimistic value iteration.}}
   \For{$\tau =1,2,\cdots, \taumax $}
	{
		
		\For{$s\in \cS, a\in \cA$}{
			compute penalty term $b\big(s,a; \widehat{V}_{\tau-1}\big)$ according to \eqref{def:bonus-Bernstein-infinite}. \\
            $\widehat{Q}_{\tau}(s, a) \leftarrow \max\big\{r(s, a) + \gamma\widehat{P}_{s, a}\widehat{V}_{\tau-1} - b\big(s, a; \widehat{V}_{\tau-1}\big), 0\big\}$. \\
		}
	\For{$s\in \cS$}{
		 $\widehat{V}_{\tau}(s) \leftarrow \max_a \widehat{Q}_{\tau}(s,a)$. \label{alg:infinite-q-update}
	}
	}
	\textbf{output:} $\widehat{\pi}$ s.t.~$\widehat{\pi}(s) \in \arg\max_a \widehat{Q}_{\taumax}(s,a)$ for every $s\in \cS$. 
	\caption{Value iteration with lower confidence bounds (\vilcb) for discounted infinite-horizon MDPs}
 \label{alg:vi-lcb-infinite}
\end{algorithm}

%shall be detailed towards the end of Chapter~\ref{sec:theory-infinite-horizon}. 

\subsection{Computational efficiency}

Given the iterative nature of the \vilcb algorithm in Algorithm~\ref{alg:vi-lcb-infinite}, a natural question concerns the number of iterations required for convergence, as this directly determines the computational complexity of the algorithm.

As it turns out, the \vilcb algorithm enjoys a rapid convergence guarantee.  
In particular, the pessimistic Bellman operator \(\Tpess\) inherits a \(\gamma\)-contraction property (to be established shortly), which implies that the iterates \(\{\widehat{Q}_\tau\}_{\tau\geq 0}\) converge geometrically fast to the fixed point \(\widehat{Q}_{\mathsf{pe}}^\star\). More precisely, we have the following result. 
\begin{lemma} \label{lem:monotone-contraction}
	Suppose $\widehat{Q}_0=0$. Then the iterates of Algorithm~\ref{alg:vi-lcb-infinite} obey
	\begin{align}
		\widehat{Q}_{\tau} \leq \widehat{Q}_{\mathsf{pe}}^{\star} 
		\quad \text{and}\quad
		\big\|\widehat{Q}_{\tau} - \widehat{Q}_{\mathsf{pe}}^{\star} \big\|_{\infty} \le  \frac{\gamma^{\tau}}{1-\gamma} 
		\qquad
		\text{for all }\tau \geq 0,
	\end{align}
	where $\widehat{Q}_{\mathsf{pe}}^{\star}$ denotes the unique fixed point of $\Tpess$. 
	Thus, choosing $\taumax \geq \frac{\log\frac{N}{1-\gamma}}{\log (1/\gamma)}$ guarantees that
	\begin{align}
		\big\|\widehat{Q}_{\taumax} - \widehat{Q}_{\mathsf{pe}}^{\star} \big\|_{\infty} \le  \frac{1}{N}. 
		\label{eq:taumax-Q-converge}
	\end{align}
\end{lemma}
The proof of this lemma is deferred to Section~\ref{sec:proof-lem:monotone-contraction}. According to this lemma, $\widehat{Q}_{\tau}$ becomes an $\varepsilon$-accurate approximation of $\widehat{Q}_{\mathsf{pe}}^{\star}$ --- in the sense that $\big\|\widehat{Q}_{\tau} - \widehat{Q}_{\mathsf{pe}}^{\star} \big\|_{\infty} \le \varepsilon$ --- after at most $\frac{1}{1-\gamma} \log\frac{1}{\varepsilon(1-\gamma)}$ iterations.

\subsection{Theoretical guarantees}

We now turn to the statistical guarantees of the \vilcb algorithm. As we shall see, when equipped with the Bernstein-style penalty function in~\eqref{def:bonus-Bernstein-infinite}, the \vilcb algorithm described in Algorithm~\ref{alg:vi-lcb-infinite} is guaranteed to produce an \(\varepsilon\)-optimal policy using a minimal number of offline samples. This is formalized in the theorem below, originally established by \citet{li2024settling}.
\begin{theorem}[\citet{li2024settling}]\label{thm:infinite-offline}
Suppose  $\gamma \in [\frac{1}{2},1)$.  Consider any $0<\delta <1$ and $ \varepsilon \in \big(0, \frac{1}{1-\gamma} \big]$, and let $\rho\in \Delta(\cS)$ be any test distribution over the state space. Assume that the number of iterations satisfies $\taumax \geq \frac{1}{1-\gamma}\log\frac{N}{1-\gamma}$. 
Then with probability at least $1-2\delta,$  the policy $\widehat{\pi}$ returned by \vilcb (see Algorithm~\ref{alg:vi-lcb-infinite}) with penalty function \eqref{def:bonus-Bernstein-infinite} obeys
\begin{align}
	V^{\star}(\rho) - V^{\widehat{\pi}}(\rho) \leq \varepsilon,
\end{align}
provided that $\cb>0$ is some sufficiently large numerical constant and the total sample size exceeds 
\begin{align}
	N\geq\frac{c_1 S\Cstar  \log\frac{NS }{(1-\gamma)\delta}}{(1-\gamma)^{3}\varepsilon^{2}} 
	\label{eq:N-range-epsilon2-inf-thm}
	%+\frac{c_{1}  S\Cstar \log\frac{N }{(1-\gamma)\delta}}{(1-\gamma)^{2}\varepsilon}	
\end{align}
for some large enough universal constant $c_1>0$.  
Here, we recall that $\Cstar\geq 1$ has been defined in Definition~\ref{assumption:concentrate-infinite-simple}. 
%
%In addition, the above result continues to hold if $\Cstar$ is replaced with $C^{\star}$ (introduced in Definition~\ref{assumption:concentrate-infinite-simple}).
\end{theorem}

We now briefly discuss the implications of Theorem~\ref{thm:infinite-offline}.  Under the Bernstein-style penalty, the \vilcb algorithm achieves \(\varepsilon\)-optimality with a sample complexity no greater than
\begin{equation}
	\widetilde{O}\bigg( \frac{ S\Cstar  }{(1-\gamma)^3\varepsilon^2}  \bigg). 
	\label{eq:sample-size-Bernstein-infinite}
\end{equation}
Notably, this sample size can be substantially smaller than the ambient dimension \(S^2A\) of the transition kernel \(P\), in which case accurately estimating the full model becomes information-theoretically impossible. Nevertheless, Theorem~\ref{thm:infinite-offline} demonstrates that pessimistic model-based methods can still achieve statistically efficient offline learning even in regimes where reliable model estimation is unattainable.

Importantly, the statistical guarantees in Theorem~\ref{thm:infinite-offline} scale linearly with the single-policy concentrability coefficient $\Cstar$, a metric quantifying the degree of distribution mismatch between the behavior policy generating the offline dataset and the target optimal policy. We emphasize again that Algorithm~\ref{alg:vi-lcb-infinite} requires no prior knowledge of $\Cstar$, making the method fully adaptive to the quality and coverage of the dataset.

It is also instructive to compare the sample complexity in~\eqref{eq:sample-size-Bernstein-infinite} with the minimax sample complexity $\widetilde{\Theta}\big(\frac{SA}{(1-\gamma)^3\varepsilon^2}\big)$ under a generative model  (as established in Chapter~\ref{chapter:generative-model}). In the favorable scenarios where $\Cstar\leq \widetilde{O}(1)$,  the model-based offline approach therefore enjoys an 
$\widetilde{O}
(A)$ sample size saving relative to the generative-model setting when controlling the suboptimality gap $V^{\star}(\rho)-V^{\widehat{\pi}}(\rho)$ w.r.t.~the initial state distribution $\rho$. At a high level, \vilcb achieves this improvement by effectively ``trusting'' the offline dataset $\mathcal{D}$:  it primarily relies on regions of the state-action space that are well covered by $\mathcal{D}$, while acting conservatively in poorly covered regions. Intuitively, when  $\mathcal{D}$ is generated by a behavior policy with expert-level performance --- so that $\Cstar$ becomes small --- the frequently visited state-action pairs are precisely those achieving high-quality decision-making, making it advantageous to place greater emphasis on these well-sampled regions.

Furthermore, as will be corroborated by a matching minimax lower bound established in Section~\ref{sec:minimax-lower-bound-offline}, the \vilcb algorithm achieves minimax-optimal sample complexity up to logarithmic factors. In particular, the sample complexity bound in~\eqref{eq:N-range-epsilon2-inf-thm} is valid over the entire accuracy regime $\varepsilon \in \big( 0, \frac{1}{1-\gamma} \big]$, implying that {\em no additional burn-in sample size} is needed to attain sample optimality. Establishing such a statistical guarantee is especially challenging in the sample-starved regime. 

Finally, we remark on a technical aspect of Algorithm~\ref{alg:vi-lcb-infinite}. The algorithm reuses the entire offline dataset across all iterations, a feature that is crucial for attaining optimal sample complexity. Noteworthily, while such sample reuse is natural from a practical perspective, it introduces substantial technical challenges for the analysis by creating intricate statistical dependencies across iterations. By contrast, some earlier work, such as \citet{rashidinejad2021bridging}, assumed the use of fresh samples at each iteration in order to simplify the analysis, albeit at the expense of higher sample complexity.

\subsection{Analysis: proof of Theorem~\ref{thm:infinite-offline}}
\label{sec:proof:thm:infinite-offline}

In this subsection, we outline the proof of Theorem~\ref{thm:infinite-offline}. We begin with several preliminary results and then organize the main proof argument into a few steps.

As in Lemma~\ref{lem:monotone-contraction}, we denote by $\widehat{Q}_{\mathsf{pe}}^{\star}$ the unique fixed point of $\Tpess$ (cf.~\eqref{eq:empirical-Bellman-infinite}) throughout this subsection, and let $\widehat{V}_{\mathsf{pe}}^{\star}$ represent its induced value function (i.e., $\widehat{V}_{\mathsf{pe}}^{\star}(s)=\max_{a}\widehat{Q}_{\mathsf{pe}}^{\star}(s,a)$ for all $s\in \cS$). Also, let $\widehat{Q}=\widehat{Q}_{\taumax}$ (resp.~$\widehat{V}=\widehat{V}_{\taumax}$) indicate the final Q-estimate (resp.~value estimate) in Algorithm~\ref{alg:vi-lcb-infinite}. 
In addition,  we introduce the transition matrix $P^{\star} \in \mathbb{R}^{S\times S}$ and the penalty vector $b^{\star} \in \mathbb{R}^{S\times 1}$, whose $s$-th row and $s$-th entry are given respectively by
\begin{align}
	\big[ P^{\star} \big]_{s,\cdot} \defn P_{s, \pi^{\star}(s)} 
	\qquad \text{and} \qquad 
	b^{\star}(s) \defn b\big(s, \pi^{\star}(s); \widehat{V}\big)
	\label{eq:P-b-star-infinite}
\end{align}
for all $s\in\cS$. Moreover, we define the occupancy distribution vector $d^\star_{\rho}\in \Delta(\cS)$ whose $s$-th entry is given by  
\begin{equation}
    d^\star_{\rho}(s) \defn d^\star(s; \rho). 
    \label{eq:defn-dstar-rho-vector-offline}
\end{equation}

\paragraph{Preliminary facts.}

In view of Lemma~\ref{lem:monotone-contraction}, the final Q-estimate $\widehat{Q}$ in Algorithm~\ref{alg:vi-lcb-infinite} satisfies
\begin{align} 
\label{eq:Q-condition-Hoeffding}
	\big\| \widehat{Q} - \widehat{Q}^{\star}_{\mathsf{pe}} \big\|_{\infty} \le \frac{1}{N}.
\end{align}
Moreover, the definition \eqref{eq:empirical-Bellman-infinite} of the pessimistic Bellman operator and the
initialization $\widehat Q_0=0$ ensure that (alternatively, see the proof of Lemma~\ref{lem:monotone-contraction}) 
\[
0\leq \widehat Q,\widehat Q_{\mathsf{pe}}^\star
\leq \frac{1}{1-\gamma} 1_{SA},
\qquad
0\leq \widehat V,\widehat V_{\mathsf{pe}}^\star
\leq \frac{1}{1-\gamma} 1_S.
\]
The following lemma records a few direct consequences of \eqref{eq:Q-condition-Hoeffding}; the proof is deferred to Section~\ref{proof:lem:offline-prelim}.
\begin{lemma}
	\label{lem:offline-prelim}
	For every $(s,a) \in \cS \times \cA$, it holds that
	\begin{subequations}
	\begin{align}
	\big\| \widehat{V} - \widehat{V}^{\star}_{\mathsf{pe}} \big\|_{\infty}  & \leq \frac{1}{N}, \label{eq:Vhar-Vstarhat-diff-1/T-inf}\\
	\big|b\big(s,a; \widehat{V}_{\mathsf{pe}}^{\star}\big)-b\big(s,a; \widehat{V}\big) \big|  &\leq  
	\frac{2}{N}.\label{infinite-bpe-to-bV}
	\end{align}
	\end{subequations}
\end{lemma}
In words, Lemma~\ref{lem:offline-prelim} shows that terminating pessimistic value
iteration after finitely many iterations introduces only a small ``optimization'' error. It also shows that evaluating the penalty at the resulting iterate and at the fixed-point value changes the penalty by at most $2/N$. Together, these properties facilitate the analysis of the computable $\widehat V$ by linking it to the fixed point $\widehat V_{\mathsf{pe}}^\star$.

The next lemma provides a lower bound on $N_{s,a}$, the total number of
sample transitions collected for each state-action pair $(s,a)$. Its proof
is deferred to Section~\ref{proof:lem:sample-split-infinite}.
\begin{lemma} 
	\label{lem:sample-split-infinite}
	Consider any $\delta\in(0,1)$.  With probability exceeding $1-\delta$, the quantities $\{N_{s,a} \}$ defined in \eqref{eq:defn-Nsa-infinite} satisfy
	\begin{align}
		\max\bigg\{N_{s,a} , \frac{2}{3}\infs \log\frac{SN}{\delta}\bigg\} &\ge \frac{ N \myrho(s, a) }{12}
		\label{eq:samples-infinite-LB}
	\end{align}
	simultaneously for all $(s, a) \in \cS \times \cA$. 
\end{lemma}
In words, this lemma relates the visitation counts $N_{s,a}$ to the data
distribution $\myrho$. Accordingly, throughout the proof, unless stated otherwise, we work on the high-probability event
\begin{align}
	\forall (s,a) \in \cS\times \cA:
	\qquad \max\bigg\{N_{s,a} ,\, \frac{2}{3}\infs \log\frac{SN }{\delta}\bigg\} &\ge \frac{ N\myrho(s, a) }{12} .
	\label{eq:assumption-Nsa-LB-infinite}
\end{align}
We also note that, under the sampling model \eqref{eq:sampling-offline-inf-iid}, the sample
transitions used to construct $\widehat{P}$ are conditionally independent
given the visitation counts $\{N_{s,a}\}$.

With these preliminaries in place, we proceed to the main proof, which is organized into four steps.

\paragraph{Step 1: Bernstein-style inequalities and leave-one-out analysis.}

The first step is to obtain sharp concentration bounds for the quantity
$\big(\widehat{P}_{s,a}-P_{s,a}\big)\widehat{V}$. 
The main challenge is that $\widehat{P}$ and $\widehat{V}$ are statistically
dependent, since the same samples are reused throughout the iterations of
Algorithm~\ref{alg:vi-lcb-infinite}. 
To address this dependence issue, we employ a
leave-one-out argument, similar to that used in the generative-model
analysis in Section~\ref{sec:sharp-analysis-thm:sample-compl-MB-generative-inf}. This yields the following Bernstein-style concentration bounds; the proof is postponed to Section~\ref{proof:lem:Bernstein-infinite}.

\begin{lemma} \label{lem:Bernstein-infinite}
	Suppose that $\gamma \in [1/2, 1)$, and consider any $\delta\in (0,1)$. 
	With probability at least $1-\delta$, 
	\begin{subequations} \label{eq:Bernstein-infinite-all}
		\begin{align}
			\big|\big(\widehat{P}_{ s, a} - P_{s, a}\big)\widetilde{V} \big| 
			&\le \sqrt{\frac{8\log\frac{4N}{(1-\gamma)\delta}}{N_{s,a} }\mathsf{Var}_{\widehat{P}_{s,a}}\hspace{-0.5ex}(\widetilde{V})}+\frac{26\log\frac{4N}{(1-\gamma)\delta}}{(1-\gamma)N_{s,a} } 
			\label{eq:bonus-Bernstein-infinite} \\
			\mathsf{Var}_{\widehat{P}_{s, a}}( \widetilde{V} ) &\le 2\mathsf{Var}_{P_{s, a}}(\widetilde{V}) 
			+ \frac{24\log\frac{4N}{(1-\gamma)\delta}}{(1-\gamma)^2N_{s,a} } \label{eq:empirical-var-infinite}
		\end{align}
	\end{subequations}
	hold simultaneously for all $(s, a) \in \cS \times \cA$   
	and all $\widetilde{V}$ satisfying $\|\widetilde{V} - \widehat{V}^{\star}_{\mathsf{pe}} \|_{\infty} \le \frac{1}{N}$ and $\|\widetilde{V}\|_{\infty}\leq \frac{1}{1-\gamma}$.
\end{lemma}
%
%\begin{proof}
%	At a high level, the proof constructs, for each state $s \in \cS$, a finite collection of auxiliary MDPs $\{\widehat{\mathcal{M}}^{s,u}\}$ with the following two properties. 	First, each $\widehat{\mathcal{M}}^{s,u}$ is constructed without using any sample transition originating from the state $s$, and is therefore statistically independent of any $\widehat{P}_{s,a}$. 	Second, at least one MDP in this collection is sufficiently close to the original true MDP in terms of the resulting value function. 	These leave-one-out auxiliary MDPs allow us to control $\big(\widehat{P}_{s,a}-P_{s,a}\big)\widetilde{V}$ by first exploiting the statistical independence between $\widehat{P}_{s,a}$ and $\{\widehat{\mathcal{M}}^{s,u}\}$, and then transferring the resulting concentration bound back to the original true MDP. 	See Section~\ref{proof:lem:Bernstein-infinite} for the formal proof. 
%\end{proof}

In the remainder of the proof, we work on the high-probability event on
which both~\eqref{eq:Bernstein-infinite-all} and  \eqref{eq:assumption-Nsa-LB-infinite} hold. Combining \eqref{eq:bonus-Bernstein-infinite} with the definition~\eqref{def:bonus-Bernstein-infinite} of $b(s,a;V)$ yields
\begin{align}
	\Big|\big(\widehat{P}_{ s, a} - P_{s, a}\big)\widetilde{V} \Big|  + \frac{5}{N}
	\leq b\big(s,a; \widetilde{V} \big) 
	\label{eq:infitnie-bernstein-b-bound-diff}
\end{align}
for all $\widetilde{V}$ obeying $\big\|\widetilde{V} - \widehat{V}^{\star}_{\mathsf{pe}} \big\|_{\infty} \le \frac{1}{N}$ and $\|\widetilde{V}\|_{\infty}\leq \frac{1}{1-\gamma}$, provided that the constant $\cb$ in~\eqref{def:bonus-Bernstein-infinite} is sufficiently large. It is worth noting that, thus far, the bound \eqref{eq:bonus-Bernstein-infinite} has been derived only for state-action pairs $(s,a)$ obeying $N_{s,a} >0$. 
For any pair $(s,a)$ with $N_{s,a} =0$, both
$\widehat{P}_{s,a}\widetilde{V}$ and $P_{s,a}\widetilde{V}$ belong to
$[0,1/(1-\gamma)]$. Consequently,
\[
\Big|\big(\widehat{P}_{s,a}-P_{s,a}\big)\widetilde{V}\Big|
\leq \frac{1}{1-\gamma},
\]
and hence inequality~\eqref{eq:infitnie-bernstein-b-bound-diff} therefore holds
trivially in this case as well. 

\begin{remark}
The variance-dependent form of the  concentration bound in Lemma~\ref{lem:Bernstein-infinite} plays a crucial role in the subsequent analysis. Although
$\widetilde{V}$ may be statistically dependent on $\widehat{P}$, the leave-one-out technique decouples this dependence and
yields a sharp Bernstein-style bound without requiring a large burn-in
sample size.  This argument is the offline analog of the leave-one-out
analysis developed for the generative model in
Section~\ref{sec:sharp-analysis-thm:sample-compl-MB-generative-inf}.
\end{remark}

\paragraph{Step 2: showing that $\widehat{Q}(s,a)$ is a lower bound on $Q^{\widehat{\pi}}(s,a)$.}
We next show that $\widehat{Q}(s,a)$ (resp.~$\widehat{V}(s)$) is a ``pessimistic'' estimate of $Q^{\widehat{\pi}}(s,a)$ (resp.~$V^{\widehat{\pi}}(s)$). The argument combines two ingredients: the incorporation of the penalty term makes the empirical
Bellman updates conservative, while Lemma~\ref{lem:Bernstein-infinite}
ensures that the estimation error of the transition kernel is well-controlled and dominated by the penalty term.

More precisely, we first recall from Lemma~\ref{lem:monotone-contraction} that $\widehat{Q}^{\star}_{\mathsf{pe}}$ is the unique fixed point of the pessimistic Bellman operator and satisfies
\begin{align}
	\widehat{Q}^{\star}_{\mathsf{pe}} (s,a) = \max\Big\{r(s, a) + \gamma\widehat{P}_{s, a}\widehat{V}^{\star}_{\mathsf{pe}} - b\big(s, a; \widehat{V}^{\star}_{\mathsf{pe}}\big) , 0\Big\}
	\label{eq:hat-Qstar-pe-Bellman-123}
\end{align}
for all $(s,a)\in \cS\times \cA$. We divide the state-action pairs into two cases.
\begin{itemize}
	\item {\em Case 1: }$\widehat Q_{\mathsf{pe}}^\star(s,a)=0$.
In view of Lemma~\ref{lem:monotone-contraction},
\[
0\leq \widehat Q(s,a)
\leq \widehat Q_{\mathsf{pe}}^\star(s,a)=0,
\]
and therefore, $\widehat Q(s,a)=0\leq Q^{\widehat\pi}(s,a)$.

	\item {\em Case 2: } $\widehat{Q}^{\star}_{\mathsf{pe}} (s,a) >0$.
	We first make the observation that
	\begin{align}
		\widehat{Q} (s,a) 
		&\overset{\text{(i)}}{\le} \widehat{Q}^{\star}_{\mathsf{pe}} (s,a) 
		\overset{\text{(ii)}}{=} r(s,a) - b\big(s, a; \widehat{V}^{\star}_{\mathsf{pe}}\big) + \gamma\widehat{P}_{s,a}\widehat{V}^{\star}_{\mathsf{pe}}   \nonumber \\
		& \leq r(s,a) - b\big(s, a; \widehat{V}^{\star}_{\mathsf{pe}}\big) + \gamma \widehat{P}_{s,a} \widehat{V}  + \gamma \big\|\widehat{P}_{s, a}\big\|_1 \big\|\widehat{V} - \widehat{V}_{\mathsf{pe}}^\star\big\|_\infty \notag\\
		& \overset{\text{(iii)}}{\le} r(s,a) - b\big(s, a; \widehat{V}^{\star}_{\mathsf{pe}}\big) + \gamma \widehat{P}_{s,a} \widehat{V} + \frac{1}{N} \notag\\
		& \leq r(s,a) - b\big(s,a; \widehat{V} \big) + \gamma P_{s,a} \widehat{V} + \frac{1}{N}  
		+ \gamma \big|\big(\widehat{P}_{s, a} - P_{s, a}\big)\widehat{V}\big| \arxiv{ \nonumber \\
		& \qquad } + \big|b\big(s,a; \widehat{V}_{\mathsf{pe}}^{\star}\big)-b\big(s,a; \widehat{V}\big) \big|
		\notag \\
		& \overset{\text{(iv)}}{\le} r(s,a) + \gamma P_{s,a}\widehat{V}. 
		\label{eq:infinite-q-upper2}
	\end{align}
	Here, (i) follows from Lemma~\ref{lem:monotone-contraction}; 
	(ii) uses the fixed point characterization \eqref{eq:hat-Qstar-pe-Bellman-123} and the assumption that $\widehat{Q}^{\star}_{\mathsf{pe}} (s,a) >0$; (iii) utilizes \eqref{eq:Vhar-Vstarhat-diff-1/T-inf} and $\|\widehat{P}_{s,a}\|_{1}=1$;
	and (iv) follows from \eqref{infinite-bpe-to-bV} as well as \eqref{eq:infitnie-bernstein-b-bound-diff}. 
	Combining \eqref{eq:infinite-q-upper2} with the Bellman equation $Q^{\widehat{\pi}}=r+\gamma P V^{\widehat{\pi}}$ gives
	\begin{align}
		Q^{\widehat{\pi}}(s, a) - \widehat{Q}(s, a) 
		&\geq  r(s,a) + \gamma P_{s,a}V^{\widehat{\pi}}
		-  \big(r(s,a) + \gamma P_{s,a}\widehat{V} \big) \notag\\
		&= \gamma P_{s,a} \big( V^{\widehat{\pi}}
		-  \widehat{V} \big).
		\label{eq:Case2-Q-Qhat-V-Vhat-inf}
	\end{align}
	Denote 
    \[
    (s',a') = \arg\min_{(s,a)} \big\{ Q^{\widehat{\pi}}(s, a)-\widehat{Q}(s, a) \big\}, 
    \]
    and suppose for the moment that 
    \begin{align}
    Q^{\widehat{\pi}}(s', a')-\widehat{Q}(s', a')<0.
    \label{eq:assumption-Qhat-pihat-negative-offline}
    \end{align}
    Note that $(s',a')$ must belong to Case 2, as we have already shown that $Q^{\widehat{\pi}}(s, a)-\widehat{Q}(s, a)\geq0$ for any $(s,a)$ in Case 1. Hence, we can derive
	\begin{align}
		\min_{s, a}\big\{Q^{\widehat{\pi}}(s, a) - \widehat{Q}(s, a)\big\} 
		&= Q^{\widehat{\pi}}(s', a')-\widehat{Q}(s', a') \nonumber\\
		&\overset{\text{(i)}}{\ge} \gamma P_{s', a'} \big(V^{\widehat{\pi}} - \widehat{V} \big) 
		\overset{\text{(ii)}}{\ge} \gamma\min_{s}\big\{V^{\widehat{\pi}}(s) - \widehat{V}(s)\big\} \notag\\
		&= \gamma\min_{s} \big\{Q^{\widehat{\pi}}\big( s, \widehat{\pi}(s) \big) - \widehat{Q}\big( s, \widehat{\pi}(s) \big) \big\}  \notag\\
		&\geq \gamma\min_{s, a}\big\{Q^{\widehat{\pi}}(s, a) - \widehat{Q}(s, a)\big\}.
		\label{eq:min-Q-lower-bound-gamma-infinite}
	\end{align}
	Here (i) follows from \eqref{eq:Case2-Q-Qhat-V-Vhat-inf}, while (ii) holds since $P_{s,a}\in\Delta(\cS)$. 
	Given that $\gamma\in(0,1)$, \eqref{eq:min-Q-lower-bound-gamma-infinite} can only hold if $\min_{s, a} \{Q^{\widehat{\pi}}(s, a) - \widehat{Q}(s, a)\} \geq 0$, which contradicts the previous assumption \eqref{eq:assumption-Qhat-pihat-negative-offline}. Hence, in Case 2 as well, we have 
	$Q^{\widehat{\pi}}(s,a)\geq \widehat{Q}(s,a)$.
\end{itemize}
Combining the two cases, we conclude that
\begin{equation}
\label{eq:infinite-q-upper3}
	Q^{\widehat{\pi}}(s, a) \geq  \widehat{Q}(s, a)\qquad \text{for all }(s,a)\in \cS\times \cA.  
\end{equation}
As an immediate consequence, for all $s\in\cS$ we have
\begin{equation}
	V^{\star}(s) \geq V^{\widehat{\pi}}(s) =  Q^{\widehat{\pi}}\big(s, \widehat{\pi}(s)\big) \geq 
	\widehat{Q}\big(s, \widehat{\pi}(s)\big) = \widehat{V}(s). 
	\label{eq:Vhat-pi-V-pihat-order-infinite}
\end{equation}

\paragraph{Step 3: bounding the value estimation error.} 
We now turn to the value estimation error $V^{\star}(s) - V^{\widehat{\pi}}(s)$. Towards this, 
let us first make note of the following lower bound on $\widehat{V}$: for every $s\in\cS$,
\begin{align}
	\widehat{V}(s) 
	&= \max_a \widehat{Q}(s, a) \ge \widehat{Q}\big(s, \pi^{\star}(s)\big) 
	\overset{\mathrm{(i)}}{\ge} \widehat{Q}^{\star}_{\mathsf{pe}}\big(s, \pi^{\star}(s)\big) - \frac{1}{N}  \notag\\
	&\overset{\mathrm{(ii)}}{\ge} r\big(s, \pi^{\star}(s)\big) - b\big(s, \pi^{\star}(s); \widehat{V}^{\star}_{\mathsf{pe}}\big) + \gamma \widehat{P}_{s, \pi^\star(s)}\widehat{V}_{\mathsf{pe}}^\star - \frac{1}{N} \notag\\
	&= r\big(s, \pi^{\star}(s)\big) - b\big(s, \pi^{\star}(s); \widehat{V}^{\star}_{\mathsf{pe}}\big) + \gamma \widehat{P}_{s, \pi^\star(s)} \widehat{V} - \frac{1}{N} 
	- \gamma \widehat{P}_{s, \pi^\star(s)}  \big( \widehat{V} - \widehat{V}_{\mathsf{pe}}^\star\big )\notag\\
	& \overset{\mathrm{(iii)}}{\ge} r\big(s, \pi^{\star}(s)\big) - b\big(s, \pi^{\star}(s); \widehat{V}^{\star}_{\mathsf{pe}}\big) + \gamma \widehat{P}_{s, \pi^\star(s)} \widehat{V} - \frac{2}{N} \notag \\
	& \ge r\big(s, \pi^{\star}(s)\big) - b\big(s, \pi^{\star}(s); \widehat{V} \big) + \gamma P_{s, \pi^\star(s)} \widehat{V} - \frac{2}{N}  \notag\\
	&\qquad - \gamma \big|\big(\widehat{P}_{s, \pi^\star(s)} - P_{s, \pi^\star(s)}\big)\widehat{V}\big| 
	- \big|b\big(s,\pi^{\star}(s); \widehat{V}_{\mathsf{pe}}^{\star}\big)-b\big(s,\pi^{\star}(s); \widehat{V}\big) \big|
	\notag \\
	&\overset{\mathrm{(iv)}}{\ge} r\big(s, \pi^{\star}(s)\big) - 2b\big(s, \pi^{\star}(s); \widehat{V}\big) + \gamma P_{s, \pi^\star(s)}\widehat{V}. 
	\label{eq:infinite-v-lower}
\end{align}
Here, (i) follows from \eqref{eq:Q-condition-Hoeffding}; 
(ii) utilizes the fixed point property \eqref{eq:hat-Qstar-pe-Bellman-123}; 
(iii) holds because, according to Lemma~\ref{lem:offline-prelim}, 
\[
\widehat{P}_{s, \pi^\star(s)}  \big( \widehat{V} - \widehat{V}_{\mathsf{pe}}^\star\big )
\leq \big\|\widehat{P}_{s, \pi^\star(s)}\big\|_1 
\big\|\widehat{V} - \widehat{V}_{\mathsf{pe}}^\star\big\|_\infty 
\leq \frac{1}{N};
\]
and (iv) follows from Lemma~\ref{lem:offline-prelim} and \eqref{eq:infitnie-bernstein-b-bound-diff}. Combining \eqref{eq:infinite-v-lower} with the Bellman optimality equation
\begin{align}
	V^{\star}(s) = r\big(s, \pi^{\star}(s)\big) + \gamma P_{s, \pi^\star(s)}V^{\star}, 
	\label{eq:infinite-Bellman-opt}
\end{align}
we arrive at
\begin{align}
	\rho^{\top}\big(V^{\star}-\widehat{V}\big) 
	&= \sum_{s\in\cS} \rho(s) \big( V^{\star}(s) - \widehat{V} (s) \big) \notag \\
	&\le \sum_{s\in\cS} \rho(s) 
	\Big\{ \gamma P_{s,\pi^\star(s)}\big(V^\star-\widehat{V}\big)
	+ 2b\big(s, \pi^{\star}(s); \widehat{V}\big) \Big\} \notag\\
	&= \gamma\sum_{s\in\cS}\rho(s)P_{s,\pi^{\star}(s)}\big(V^{\star}-\widehat{V}\big)
	+2\sum_{s\in\cS}\rho(s)b\big(s, \pi^{\star}(s); \widehat{V}\big) \notag \\
	& = \gamma \rho^\top P^{\star}\big(V^{\star}-\widehat{V}\big)+ 2\rho^{\top}b^{\star},
\end{align}
where we recall the definition of $P^{\star} $ and $b^{\star} $ in \eqref{eq:P-b-star-infinite}.

 Since this relation holds for any distribution $\rho \in \Delta(\cS)$, applying it recursively yields
\begin{align}
	\rho^{\top}\big(V^{\star}-\widehat{V}\big) 
	& \leq  \big( \gamma \rho^\top P^{\star} \big) \big(V^{\star}-\widehat{V}\big)+ 2\rho^{\top}b^{\star}\notag\\
	& \leq  \gamma \big( \gamma \rho^\top P^{\star} \big) P^{\star} \big(V^{\star}-\widehat{V}\big)
	+2 \big(\gamma  \rho^{\top} P^{\star} \big)  b^{\star}
	+2\rho^{\top}b^{\star}\notag\\
	& = \gamma^2 \rho^\top \big( P^{\star}\big)^{2}\big(V^{\star}-\widehat{V}\big)
	+2\gamma  \rho^{\top} P^{\star} b^{\star}
	+2\rho^{\top}b^{\star}\notag\\
	& \leq\cdots \leq \lim_{i\rightarrow \infty}\gamma^{i} \rho^\top \big( P^{\star}\big)^{i}\big(V^{\star}-\widehat{V}\big) 
	+ 2\rho^{\top} \bigg\{\sum_{i=0}^{\infty}\gamma^{i}\big(P^{\star}\big)^{i}\bigg\} b^{\star} 
	\notag\\
	& \overset{\mathrm{(i)}}{=} 2\rho^{\top} \bigg\{\sum_{i=0}^{\infty}\gamma^{i}\big(P^{\star}\big)^{i}\bigg\} b^{\star} 
	= 2\rho^{\top}\big(I-\gamma P^{\star}\big)^{-1}b^{\star} \notag\\
	& \overset{\mathrm{(ii)}}{=} \frac{2}{1-\gamma} \langle d_{\rho}^{\star}, b^{\star} \rangle. 	\label{eqn:brahms-reduction}
\end{align}
Here, to see why (i) holds, it suffices to note that 
$
\lim_{i \to \infty} \gamma^i \rho^\top \big(P^\star\big)^i\big(V^\star -\widehat{V}\big) = 0
$
since $\gamma^i\to 0$ and $\|\rho^\top \big(P^\star\big)^i \|_1 =1$ for all $i\geq 0$;
and (ii) follows from the definition of the discounted occupancy measure $d^{\star}_{\rho}$ in \eqref{eq:defn-dstar-rho-vector-offline} and \eqref{eq:infinite-d-star-def}, equivalently expressed in matrix-vector form as
\begin{align}
	\big(d_{\rho}^\star\big)^\top = (1-\gamma)  \sum_{t=0}^\infty\gamma^t  \rho^\top \big(P^\star\big)^t =  (1-\gamma) \rho^\top \big(I-\gamma P^{\star}\big)^{-1}.
	\label{eq:infinite-d-matrix-form}
\end{align}
Combine \eqref{eqn:brahms-reduction} with \eqref{eq:Vhat-pi-V-pihat-order-infinite} to reach
\begin{align}
	\big\langle \rho, V^{\star} - V^{\widehat{\pi}}\big\rangle 
	\le \big\langle \rho, V^{\star} - \widehat{V}\big\rangle 
	\leq \frac{2\big\langle d_{\rho}^{\star}, b^{\star} \big\rangle}{1-\gamma}. 
	\label{eq:Vall-Bernstein-infinite}
\end{align}
In words, the bound in \eqref{eq:Vall-Bernstein-infinite} reduces the remaining statistical challenge to controlling the single quantity $\langle d_\rho^\star,b^\star\rangle$, which is the main focus of the remainder of the proof.

\paragraph{Step 4: using single-policy concentrability to control $\langle d_{\rho}^\star, b^\star \rangle$.}
We find it helpful to partition the state space $\cS$ into the following two disjoint subsets: 
\begin{subequations}
	\label{eq:defn-Ssmall-Slarge-inf}
	\begin{align}
		\cS^{\mathsf{small}} \coloneqq \bigg\{ s\in \cS \mid  N \myrho \big(s, \pi^{\star}(s) \big) \leq 8\infs \log \frac{NS}{(1-\gamma)\delta} \bigg\}; \label{eq:defn-Ssmall-inf}\\
		\cS^{\mathsf{large}} \coloneqq \bigg\{ s\in \cS \mid  N \myrho \big(s, \pi^{\star}(s) \big) > 8\infs\log \frac{NS}{(1-\gamma)\delta} \bigg\}. \label{eq:defn-Slarge-inf}
	\end{align}
\end{subequations}

\paragraph{Step 4.1: contributions from $\cS^{\mathsf{small}}$.} 
		First, consider any state $s \in \cS^{\mathsf{small}}$. In view of the definition of $\Cstar$ in \eqref{eq:concentrate-infinite-simple} and the definition of $\cS^{\mathsf{small}}$,
		\begin{align*}
			d^{\star}_{\rho}(s) 
			\le \Cstar  \myrho \big(s, \pi^{\star}(s) \big) 
			\le \frac{8 \Cstar \infs \log \frac{NS}{(1-\gamma)\delta}}{N}.
		\end{align*}
		Combining this bound with the following property (a direct consequence of the definition \eqref{def:bonus-Bernstein-infinite})
		\begin{align}
			b^{\star}(s) = b\big(s, \pi^{\star}(s); \widehat{V} \big) \leq \frac{1}{1-\gamma} + \frac{5}{N},
		\end{align}
		we obtain
		\begin{align}
	       \sum_{s\,\in\,\cS^{\mathsf{small}}}	d^{\star}_{\rho}(s) b^{\star}(s) 
			&\leq \sum_{s\,\in\,\cS^{\mathsf{small}}} \bigg( \frac{8 \Cstar \infs \log \frac{NS}{(1-\gamma)\delta}}{(1-\gamma)N} +   d_{\rho}^{\star}(s) \frac{5}{N} \bigg) \nonumber \\
			& \leq  \frac{8 S \Cstar \infs \log \frac{NS}{(1-\gamma)\delta}}{(1-\gamma)N} +  \frac{5}{N},
			\label{eq:infinite-bernstein-small-result}
		\end{align}
        where the last inequality is valid since $\sum_{s\in \cS^{\mathsf{small}}} d_{\rho}^{\star}(s) \leq 1$.

\paragraph{Step 4.2: contributions from $\cS^{\mathsf{large}}$.}
		Next, consider any state $s \in \cS^{\mathsf{large}}$. By the definition \eqref{def:bonus-Bernstein-infinite} of $b(s,a;V)$, we have
		\begin{align}
			b^{\star}(s) 
			&\le \sqrt{\frac{\cb\log\frac{N }{(1-\gamma)\delta}}{N_{s, \pi^{\star}(s)}}
				\mathsf{Var}_{\widehat{P}_{s, \pi^{\star}(s)}}\big(\widehat{V}\big)} 
			+  \frac{2\cb\log\frac{N }{(1-\gamma)\delta}}{(1-\gamma)N_{s, \pi^{\star}(s)}} + \frac{5}{N}\notag\\
			& \leq \sqrt{\frac{2\cb\log\frac{N }{(1-\gamma)\delta}}{N_{s, \pi^{\star}(s)}}
				\mathsf{Var}_{P_{s, \pi^{\star}(s)}}\big(\widehat{V}\big)} 
			+  \frac{4\cb\log\frac{N }{(1-\gamma)\delta}}{(1-\gamma)N_{s, \pi^{\star}(s)}}, 
			\label{eq:infinite-bernstein-large-result-middle}
		\end{align}
		provided that $\cb$ is sufficiently large. Here, the last inequality follows from the fact that $N \geq N_{s,a} $ and the following property:
		\[
		\mathsf{Var}_{\widehat{P}_{s, \pi^{\star}(s)}}\big(\widehat{V}\big) 
		\leq 2\mathsf{Var}_{P_{s, \pi^{\star}(s)}}\big(\widehat{V}\big) 
		+ \frac{24\log\frac{4N}{(1-\gamma)\delta}}{(1-\gamma)^2N_{s, \pi^{\star}(s)}},
		\]
        which is a direct consequence of Lemmas~\ref{lem:offline-prelim} and~\ref{lem:Bernstein-infinite}. 
        
        Moreover, it is seen that
		\begin{align}
			\frac{1}{N_{s, \pi^{\star}(s)}} 
			&\overset{\mathrm{(i)}}{\le} \frac{12}{N\myrho \big(s, \pi^{\star}(s)\big)} 
			\overset{\mathrm{(ii)}}{\le} \frac{12\Cstar}{N d_{\rho}^{\star}(s)} ,
			\label{eq:infinite-N-to-d-bound}
		\end{align}
		where (i) follows from \eqref{eq:assumption-Nsa-LB-infinite} and the definition of $\cS^{\mathsf{large}}$, and (ii) arises from~\eqref{eq:concentrate-infinite-simple}. 
        Given that states satisfying $d_{\rho}^\star(s)=0$ make no contribution to
$\langle d_{\rho}^\star,b^\star\rangle$, it suffices to restrict attention to
\[
\cS^{\mathsf{large}}_{+}
\coloneqq
\Big\{
s\in\cS^{\mathsf{large}}:d_{\rho}^\star(s)>0
\Big\}.
\]
        Substituting the above bound \eqref{eq:infinite-N-to-d-bound} into \eqref{eq:infinite-bernstein-large-result-middle} gives
		\begin{align}
			b^{\star}(s) 
			&\leq \sqrt{\frac{24\cb  \Cstar \log\frac{N }{(1-\gamma)\delta}}{N} } \alpha_1(s) + \frac{48\cb \Cstar \log\frac{N  }{(1-\gamma)\delta}}{(1-\gamma)N}  \alpha_2(s),
			\label{eq:infinite-bernstein-large-result}
		\end{align}
		where
		\[
		\alpha_1(s) \coloneqq \frac{1}{\sqrt{d_{\rho}^{\star}(s)}} \sqrt{\mathsf{Var}_{P_{s, \pi^{\star}(s)}}\big(\widehat{V}\big)}, \qquad \alpha_2(s) \coloneqq \frac{1}{d_{\rho}^{\star}(s)} .
		\]
		We now control the weighted sums of $\alpha_1$ and $\alpha_2$. 
        \begin{itemize}
        \item Regarding $\alpha_1$, the Cauchy-Schwarz inequality tells us that
		\begin{align}
			\sum_{s\,\in\,\cS^{\mathsf{large}}_{+}}d_{\rho}^{\star}(s) \alpha_1(s) 
			& = \sum_{s\,\in\,\cS^{\mathsf{large}}_{+}} \sqrt{d_{\rho}^{\star}(s)  \mathsf{Var}_{P_{s, \pi^{\star}(s)}}\big(\widehat{V}\big)} \nonumber \\
			& \leq
			\sqrt{S} \cdot \sqrt{\sum_{s\,\in\,\cS^{\mathsf{large}}_{+}}  d_{\rho}^{\star}(s)  
				\mathsf{Var}_{P_{s, \pi^{\star}(s)}}\big(\widehat{V}\big)}.   
			\label{eq:infinite-bernstein-large-result0}
		\end{align}
		To further bound the right-hand side of \eqref{eq:infinite-bernstein-large-result0}, we resort to the following lemma, whose proof is provided in Section~\ref{proof:lem:sum-d-Var-UB}.
		\begin{lemma} \label{lem:sum-d-Var-UB}
			It holds that
			\begin{align}
				\sum_{s\in\cS} d_{\rho}^{\star}(s) \mathsf{Var}_{P_{s, \pi^{\star}(s)}}\big(\widehat{V}\big) 
				\le \frac{2}{\gamma^2(1-\gamma)} + \frac{4}{\gamma^2(1-\gamma)}\big\langle d_{\rho}^{\star}, b^{\star}\big\rangle.
				\label{eq:sum-d-Var-UB-123-infinite}
			\end{align}
		\end{lemma}
\begin{remark}
At a high level, Lemma~\ref{lem:sum-d-Var-UB} controls the weighted sum of the variance terms through a Bellman recursion, yielding the optimal horizon dependence. Its proof follows a variance-propagation argument similar in some sense to that used in the generative-model analysis in Section~\ref{sec:sharp-analysis-thm:sample-compl-MB-generative-inf}, which is essential for obtaining sharp horizon dependence. 
\end{remark}
		Taken together, Lemma~\ref{lem:sum-d-Var-UB} and inequality~\eqref{eq:infinite-bernstein-large-result0} yield
		\begin{align}
			\sum_{s\,\in\,\cS^{\mathsf{large}}_+}
d_{\rho}^\star(s)\alpha_1(s)
			& \leq
			2 \sqrt{\frac{2S}{1-\gamma}} 
			+4 \sqrt{\frac{S}{1-\gamma}\big\langle d_{\rho}^{\star}, b^{\star}\big\rangle},
			\label{eq:infinite-bernstein-large-result1}
		\end{align}
		where we have used $\gamma \geq 1/2$. 
        
        \item In addition, the weighted sum of $\alpha_2$ satisfies
		\begin{align}
			\sum_{s\,\in\,\cS_+^{\mathsf{large}}}d_{\rho}^{\star}(s) \alpha_2(s)
			\leq S, 
			\label{eq:infinite-bernstein-large-result2}
		\end{align}
		where we used $\alpha_2(s)=1/d_{\rho}^\star(s)$ and
$|\cS^{\mathsf{large}}_+|\leq S$. 
\end{itemize}
Combining \eqref{eq:infinite-bernstein-large-result1} and \eqref{eq:infinite-bernstein-large-result2} with \eqref{eq:infinite-bernstein-large-result}, we obtain
		\begin{align}
			\sum_{s\,\in\,\cS^{\mathsf{large}}}d_{\rho}^{\star}(s) b^{\star}(s)
			& = 
\sum_{s\,\in\,\cS^{\mathsf{large}}_+}
d_{\rho}^\star(s)b^\star(s) \notag\\
&\leq \sqrt{\frac{384 \cb S\Cstar \log\frac{N  }{(1-\gamma)\delta}}{(1-\gamma)N} 
				\big\langle d_{\rho}^{\star}, b^{\star}\big\rangle}
			\arxiv{ \notag\\
			& \qquad } +21 \sqrt{\frac{\cb S \Cstar \log\frac{N }{(1-\gamma)\delta}}{(1-\gamma)N} },  
			\label{eq:infinite-bernstein-large-result-all}
		\end{align}
		as long as 
		\begin{align}
			N\geq\frac{c_1 S\Cstar \log\frac{NS }{(1-\gamma)\delta}}{1-\gamma}  
			\label{eq:N-condition-basic-burnin-proof-inf}
		\end{align}
        for some large enough constant  $c_1$,  
        where \eqref{eq:N-condition-basic-burnin-proof-inf} is a consequence of \eqref{eq:N-range-epsilon2-inf-thm} and the condition $\varepsilon \leq 1/(1-\gamma)$.
		 
\paragraph{Step 4.3: putting all this together.}
Combining the preceding bounds for $\cS^{\mathsf{small}}$ and $\cS^{\mathsf{large}}$ in \eqref{eq:infinite-bernstein-small-result} and \eqref{eq:infinite-bernstein-large-result-all}, respectively, we arrive at
\begin{align*}
	\arxiv{&}\big\langle d_{\rho}^{\star}, b^\star \big\rangle  
	\book{&}=\sum_{s\,\in\,\cS^{\mathsf{large}}}d_{\rho}^{\star}(s)b^{\star}(s) + \sum_{s\,\in\,\cS^{\mathsf{small}}}d_{\rho}^{\star}(s)b^{\star}(s) \\
	&\arxiv{\qquad} \overset{\mathrm{(i)}}{\leq}  \sqrt{\frac{384 \cb S\Cstar \log\frac{N  }{(1-\gamma)\delta}}{(1-\gamma)N} \big\langle d_{\rho}^{\star}, b^{\star}\big\rangle} +34 \sqrt{\frac{\cb S \Cstar \log\frac{N }{(1-\gamma)\delta}}{(1-\gamma)N} } \\
	&\arxiv{\qquad} \overset{\mathrm{(ii)}}{\leq} \frac{1}{2} \big\langle d_{\rho}^{\star}, b^{\star}\big\rangle + 192\frac{\cb S\Cstar \log\frac{NS  }{(1-\gamma)\delta}}{(1-\gamma)N}  + 34\sqrt{\frac{\cb S\Cstar \log\frac{NS  }{(1-\gamma)\delta}}{(1-\gamma)N}  } \\
	&\arxiv{\qquad} \overset{\mathrm{(iii)}}{\leq} \frac{1}{2} \big\langle d_{\rho}^{\star}, b^{\star}\big\rangle + 35\sqrt{\frac{\cb S\Cstar \log\frac{NS  }{(1-\gamma)\delta}}{(1-\gamma)N}  } .
\end{align*}
Here, (i) and (iii) follow from the sample size condition \eqref{eq:N-condition-basic-burnin-proof-inf}, provided that $\cb$ and $c_1$ are sufficiently large; 
%. Step (i) also uses the fact that $\Cstar \geq 1$ (see Remark~\ref{rem:Cstar_lower_bound_infinite}). Step 
and (ii) follows from the AM-GM inequality. Rearranging terms, we are left with
\begin{align*}
	\big\langle d_{\rho}^{\star}, b^{\star} \big\rangle 
	& \leq 70\sqrt{\frac{\cb S\Cstar \log\frac{NS  }{(1-\gamma)\delta}}{(1-\gamma)N}},
\end{align*}
which combined with~\eqref{eq:Vall-Bernstein-infinite} gives
\begin{align}
	\big\langle \rho, V^{\star} - V^{\widehat{\pi}}\big\rangle 
	\leq \frac{2\big\langle d_{\rho}^{\star}, b^{\star} \big\rangle}{1-\gamma} 
	\leq 140\sqrt{\frac{\cb S\Cstar \log\frac{NS  }{(1-\gamma)\delta}}{(1-\gamma)^3N}} .
\end{align}
This taken collectively with the sample size condition \eqref{eq:N-range-epsilon2-inf-thm} establishes the advertised result in Theorem~\ref{thm:infinite-offline}, provided that $c_1$ is sufficiently large.

\section{Model-free offline RL}
\label{sec:model-free-offline-RL}

Thus far, we have demonstrated the effectiveness of the pessimism principle amid uncertainty in model-based offline RL in Section~\ref{sec:VI-LCB-infinite-horizon}. Similar ideas can also be incorporated into model-free methods to address the challenges arising in the offline setting. Since such model-free approaches are generally statistically suboptimal, we provide only a brief introduction here, focusing primarily on the key ideas and underlying intuition rather than pursuing a treatment as detailed as that of the model-based counterpart. As before, for ease of presentation, our discussion focuses on the independent data model in~\eqref{eq:sampling-offline-inf-iid}; the extension to the Markovian data setting can be found in~\citet{yan2022efficacy}.

\subsection{Algorithm}

We now describe how the pessimism principle can be incorporated into Q-learning, leading to the algorithm known as \emph{pessimistic Q-learning}, or simply \emph{\LCBQ}, proposed by \citet{shi2022pessimistic,yan2022efficacy}. As its name suggests, \LCBQ can be viewed as the offline counterpart of the \ucbq algorithm introduced for online RL in Section~\ref{sec:ucbvi-intro}.

Similar to the model-based counterpart, a central idea of \LCBQ is to incorporate lower confidence bounds into the  iterative Q-learning procedure to account for uncertainty arising from limited data coverage.  
Specifically, throughout its execution, the algorithm maintains ``pessimistic'' estimates $Q^{\lcb}$ and $V^{\lcb}$ for the optimal Q-function and value function, respectively.  At iteration $t$, the learner accesses the $t$-th sample $(s_{t}, a_{t}, s_t')$ from the offline dataset $\mathcal{D}$. For notational convenience, let $n \coloneqq n(s_{t},a_{t})$ 
represent the number of visits to the state-action pair $(s_t,a_t)$ up to and including time $t$. The vanilla Q-learning update is modified by incorporating a penalty term as follows:
\begin{align*}
Q^{\lcb}\left(s_{t},a_{t}\right) & \leftarrow \left(1-\eta_{n}\right)Q^{\lcb}\left(s_{t},a_{t}\right) +\eta_{n}\Big\{ r(s_{t}, a_{t}) + \gamma V^{\lcb}(s_t') -b_n \Big\},
\end{align*}
where $\eta_n$ and $b_n$ stand for the corresponding learning rate and penalty term used at the current iteration, respectively. The resulting estimates are called ``pessimistic'' because the penalty term deliberately shifts the Q-value updates downward, producing conservative estimates in the face of uncertainty. 
The corresponding value estimate is then updated according to
\begin{equation*}
V^{\lcb}(s_{t}) \leftarrow \max\Big\{ V^{\lcb}(s_{t}),\, \max_{a\in\cA} Q^{\lcb}(s_{t}, a) \Big\},
\end{equation*}
where the additional maximization operation is introduced to preserve the monotonicity of the value updates. This procedure is run for a total number of $T$ iterations.

Clearly, the choices of the learning rates and the penalty terms play a pivotal role in determining both the convergence behavior and the statistical efficiency of this pessimistic variant of Q-learning. For concreteness, we adopt choices that closely resemble those used by the online \ucbq algorithm (see Section~\ref{sec:model-free-onlineRL}). Specifically, given a confidence level $\delta \in (0,1)$, define the effective horizon
\begin{subequations}
\label{eq:stepsize-penalty-offlineQ}
\begin{equation}
H \coloneqq \bigg\lceil \frac{4}{1-\gamma}\log\frac{SAT}{\delta} \bigg\rceil,
\label{eq:defn-H-lcbq-stepsize-proof}
\end{equation}
which corresponds to the effective horizon $(1-\gamma)^{-1}$ inflated by a logarithmic factor.  
We then choose the learning rate
\begin{equation} \label{eq:lr_offlineQ}
\eta_n = \frac{H +1}{H +n},
\end{equation}
as well as the Hoeffding-style penalty term 
\begin{equation} \label{eq:penalty-offlineQ}
b_n =c_{\mathsf{b}}\sqrt{\frac{H \log\left(SAT/\delta\right)}{n  \left(1-\gamma\right)^{2}}}, 
\end{equation}
\end{subequations}
where $c_{\mathsf{b}}>0$ is some universal constant.

The complete procedure is summarized in Algorithm~\ref{alg:q-infinite-lcb}.

\begin{remark}
In Algorithm~\ref{alg:q-infinite-lcb}, the policy iterate \(\widehat{\pi}\) is updated to be greedy w.r.t.~the current Q-estimate \(Q^{\lcb}\) only at the visited state \(s_t\), and only when the corresponding value estimate strictly increases. This particular policy-update rule is adopted primarily to facilitate the analysis; other choices may also be used.
\end{remark}

\begin{algorithm}[t]
	%\begin{algorithmic}[1] 
    \DontPrintSemicolon
	\textbf{input:} offline dataset $\mathcal{D}$, discount factor $\gamma$, reward function $r$, learning rates $\{\eta_t\}$, max iteration number $T$. \\
    \textbf{initialize} $Q^{\lcb}(s,a) \leftarrow 0 $, $V^{\lcb}(s)  \leftarrow 0$, $n(s,a) \leftarrow0$ for all $(s,a) \in \cS\times \cA$, and set
$\widehat{\pi}:\cS\to\cA$ as an arbitrary deterministic policy. \\
%\STATE  \textbf{initialization:} $Q_0=0$. 
   \For{$t=1,2,\cdots,T$}{
  %  \For{$(s,a)\in \cS\times \cA$}{
  {\color{blue}\tcc{draw the $t$-th sample from $\mathcal{D}$.}} 
	   draw a fresh sample $(s_{t},a_{t}) \sim d^{\mathsf{b}}$ and $s_{t}'  \sim  P(\cdot \mymid s_{t},a_{t})$. \\
      % {\color{blue}\tcc{stochastic approximation.}}
       set $ n(s_{t},a_{t}) \leftarrow n(s_{t},a_{t})+1$, and $n \leftarrow n(s_{t},a_{t})$. \\
	   compute $\eta_n$ and $b_n$ according to %\eqref{eq:lr_offlineQ} and 
       \eqref{eq:stepsize-penalty-offlineQ}. \\
{\color{blue}\tcc{pessimistic Q-learning updates.}}
	   update $Q$-estimate according to
       %\eqref{eq:q-learning-update-rule}.  
\begin{align}
Q^{\lcb}\left(s_{t},a_{t}\right)&  \leftarrow  \left(1-\eta_{n}\right)Q^{\lcb}\left(s_{t},a_{t}\right) \arxiv{ \nonumber \\
& \quad\quad\quad} +\eta_{n}\Big\{ r(s_{t}, a_{t}) + \gamma V^{\lcb}(s_t') -b_n \Big\},
\label{eq:Q-lcb-update-rule-Q}
\end{align}
set $V_{\mathsf{old}}^{\lcb} \leftarrow V^{\lcb}$, and update
\begin{equation}
V^{\lcb}(s_{t}) \leftarrow \max\Big\{V_{\mathsf{old}}^{\lcb}(s_{t}),\,\max_{a\in\cA}Q^{\lcb}(s_{t},a)\Big\}. 
\label{eq:Q-lcb-update-rule-V}
\end{equation}

\If{$V^{\lcb}(s_t)>V_{\mathsf{old}}^{\lcb}(s_t)$}{
    update 
    $\widehat{\pi}(s_t)\leftarrow
    \arg\max_{a\in\cA}Q^{\lcb}(s_t,a)$, 
    where ties are broken according to an arbitrary deterministic rule.
}
    }
   % }
   \textbf{output:} policy estimate $\widehat{\pi}$.
   %s.t.~$\widehat{\pi}(s) = \arg\max_{a\in \cA}{Q}^{\lcb}(s,a)$ for all $s\in \cS$. 
   %\end{algorithmic} 
    \caption{Pessimistic Q-learning (\LCBQ) for discounted infinite-horizon MDPs}
 \label{alg:q-infinite-lcb}
\end{algorithm}

\subsection{Theoretical guarantees}

Under the aforementioned choices of learning rate and penalty, the \LCBQ algorithm described in Algorithm~\ref{alg:q-infinite-lcb} achieves the following, albeit suboptimal, statistical guarantee. This result was originally established by \citet{yan2022efficacy}. 
\begin{theorem}[Yan et al. (2022)]\label{theorem:lcbq} 
Consider any $0<\delta<1$ and any $ \varepsilon \in \big(0, \frac{1}{1-\gamma} \big]$. Let $\rho\in \Delta(\cS)$ be any given test distribution over the state space. With probability exceeding $1-\delta$, the policy
$\widehat{\pi}$ returned by the \LCBQ algorithm in Algorithm~\ref{alg:q-infinite-lcb} with learning rates and penalty in \eqref{eq:stepsize-penalty-offlineQ} satisfies
\begin{equation*}
V^{\star}(\rho)-V^{\widehat{\pi}}(\rho)\leq 
\varepsilon ,
\end{equation*}
provided that $c_{\mathsf{b}}>0$ (cf.~\eqref{eq:penalty-offlineQ}) is sufficiently large and the total number of iterations exceeds
\begin{equation}
T\geq\frac{c_{2}S\Cstar\log\frac{SAT}{(1-\gamma)\delta}}{\left(1-\gamma\right)^{5}\varepsilon^{2}}\label{eq:sample-size-lcbq-T-thm}
\end{equation}
for some large enough universal constant $c_2>0$. Here, 
$\Cstar\geq 1$ is the single-policy concentrability coefficient in Definition~\ref{assumption:concentrate-infinite-simple}. 
\end{theorem}

Since each iteration of \LCBQ accesses only a single sample from the offline dataset, the iteration complexity in~\eqref{eq:sample-size-lcbq-T-thm} established in Theorem~\ref{theorem:lcbq} also coincides with the algorithm's sample complexity.  
Moreover, if a Bernstein-style penalty is employed in place of the Hoeffding-style penalty, the sample complexity of \LCBQ can be further improved to
\begin{align} \widetilde{O}\Bigg( \frac{S\Cstar}{ (1-\gamma)^4 \varepsilon^2 }\Bigg) 
\label{eq:sample-size-LCBQ-bernstein}
\end{align}
for sufficiently small $\varepsilon$. When the offline data are near-expert, i.e., $\Cstar \le  \widetilde{O}(1)$, the resulting sample complexity of \LCBQ with Bernstein-style penalty reduces to
$$ \widetilde{O}\left( \frac{S}{ (1-\gamma)^4 \varepsilon^2 }\right), $$
which is, in general, substantially smaller than the sample complexity guarantees established for Q-learning under a generative model in Chapter~\ref{chapter:generative-model-MF}. It is important to note, however, that this improvement pertains only to the suboptimality gap with respect to the initial distribution \(\rho\), namely \(V^\star(\rho)-V^{\widehat{\pi}}(\rho)\), rather than holding uniformly over all entries of \(V^\star - V^{\widehat{\pi}}\).

Finally, we remark that the above statistical guarantees can be further improved through the use of variance reduction techniques. In particular, when combined with the pessimism principle, variance-reduced model-free algorithms can attain a sample complexity of
$$ \widetilde{O}\Bigg( \frac{S\Cstar}{ (1-\gamma)^3 \varepsilon^2 }\Bigg) 
$$
as $\varepsilon \rightarrow 0$, thereby achieving asymptotically optimal sample complexity. This improvement, however, comes at the expense of a large burn-in cost. We refer the interested reader to \citet{shi2022pessimistic,yan2022efficacy} for further details on such variance-reduced, model-free offline algorithms.

\subsection{Analysis: proof of Theorem~\ref{theorem:lcbq}}
\label{sec:analysis-model-free-offline-RL}

This subsection presents the proof of Theorem~\ref{theorem:lcbq}. 

\paragraph{Notation and auxiliary iterates.}
We adopt the following convenient notation: 
\begin{itemize}
\item $(s_t, a_t, s_t')$: the state-action-state transition tuple sampled at the $t$-th iteration;

\item $Q_t(s,a)$ and $V_t(s)$ denote, respectively, $Q^{\lcb}\left(s,a\right)$ and $V^{\lcb}\left(s\right)$ at the end of the $t$-th iteration.

\item $n_t(s,a)$: the cumulative number of visits to $(s,a)$ among the first $t$ iterations;

\item $k_i(s,a)$: the time index at which $(s,a)$ is visited for the $i$-th time.

\end{itemize}
To facilitate the presentation of the analysis, we let $\pi_t$ denote the (deterministic) policy iterate $\widehat{\pi}$ maintained by the end of iteration $t$; more precisely, let $\pi_0:\cS\to\cA$ be the initial policy iterate, and define the subsequent policy iterate recursively such that 
\begin{equation}
\pi_t(s)=
\begin{cases}
\arg\max_{a\in\cA}Q_t(s,a), & \text{if }s=s_{t}\text{ and }V_t(s)>V_{t-1}(s)\\
\pi_{t-1}(s), & \text{otherwise}
\end{cases}
\label{eq:analysis-policy-lcbq}
\end{equation}
for every $s\in \cS$. Clearly, the final policy returned by Algorithm~\ref{alg:q-infinite-lcb} is $\widehat{\pi}=\pi_T$.

We also use the notation $\eta_i^n$ introduced in
\eqref{eq:learning-rate-notation-all-ucbq}. The corresponding learning-rate
properties established in Lemma~\ref{lemma:property_lr} continue to hold in
the present setting. Moreover, a straightforward induction shows that
\begin{equation}
0\le V_t(s)\le \frac{1}{1-\gamma},\qquad \forall s\in\cS\text{ and } 0\le t\le T.
\label{eq:value-bounded-lcbq}
\end{equation}
Finally, throughout the proof, we use the shorthand notation
\begin{align}
    \iota \coloneqq \log\frac{SAT}{\delta}.
    \label{eq:iota-defn-lcbq-proof}
\end{align}
%

 %We emphasize that the policy sequence \(\{\pi_t\}\) is introduced solely as an analytical device and is not used by the algorithm itself; its main purpose is to help certify the pessimism properties of the monotone value estimates.

\paragraph{Step 1: recursion on Q-estimation errors. } 
To begin with, the lemma below establishes useful recursive relations concerning the Q-estimation errors. 
\begin{lemma} 
\label{lemma:Q-recursion-pessimistic-Q}
With probability at least $1-\delta$, one has
\begin{align}
Q^\star(s,a)-Q_t(s,a)
&\le \gamma\sum_{i=1}^{n_t(s,a)}\eta_i^{n_t(s,a)}P_{s,a}\big(V^\star-V_{k_i(s,a)-1}\big)
+\beta_{n_t(s,a)}
\label{eq:qstar-recursion-lcbq} \\
Q^\pi(s,a)-Q_t(s,a)
&\geq
\gamma\sum_{i=1}^{n_t(s,a)}
\eta_i^{n_t(s,a)}
P_{s,a}\big(V^\pi-V_{k_i(s,a)-1}\big)
\label{eq:uniform-policy-pessimism-recursion}
\end{align}
simultaneously for all $(s,a)\in \cS \times \cA$, all $1\leq t\leq T$, and all deterministic policy $\pi$, provided that the constant $c_{\mathsf{b}}>0$ in~\eqref{eq:penalty-offlineQ} is large enough. Here and below, we define 
\begin{equation} \label{eq:beta-bar-beta-lcbq}
\beta_n\coloneqq \begin{cases} 3c_{\mathsf b}\sqrt{\frac{H\iota}{n(1-\gamma)^2}}, & \text{if }n\geq 1, \\
+\infty, & \text{if }n=0,
\end{cases}
\end{equation}
and adopt the convention that $\sum_{i=1}^{0}a_i=0$ for any sequence $\{a_i\}$. 
\end{lemma}

In words, the above recursion singles out two components that play a crucial role in the Q-estimation error: an error term propagated from value-estimation errors at previous visits to the same state-action pair, and a residual term that, via the Azuma–Hoeffding inequality, bounds the uncertainty arising from finite samples. The proof of Lemma~\ref{lemma:Q-recursion-pessimistic-Q} is deferred to Section~\ref{sec:proof-lemma-Q-recursion-pessimistic}.

\paragraph{Step 2: pessimism.} 
Next, we establish the pessimism properties of the Q- and value estimates, showing that they are bounded above by the corresponding optimal Q- and value functions. This result can be viewed as the model-free counterpart of the pessimism property established for the model-based approach (see Step 2 in the proof of Theorem~\ref{thm:infinite-offline} in Section~\ref{sec:proof:thm:infinite-offline}). The formal statement is given in the following lemma, whose proof is provided in Section~\ref{sec:proof-pessimistic-Q}.

\begin{lemma} 
\label{lemma:pessimistic-Q}
With probability exceeding $1-\delta$, one has
\begin{equation} \label{eq:pessimistic-lcbq}
Q_t(s,a) \leq Q^{\star}(s,a) \qquad \text{and} \qquad V_t(s) \leq V^{\pi_t}(s) \leq V^{\star}(s)
\end{equation}
simultaneously for all $(s,a)\in \cS\times \cA$ and all $1\leq t\leq T$.  
\end{lemma}

From now on, we define, for each $0\leq t\leq T$, the value estimation error:
\begin{subequations}
\begin{equation}
\Delta_t \coloneqq V^\star-V_t.
\label{eq:Delta-chi-offlineQ}
\end{equation}
According to \eqref{eq:value-bounded-lcbq} and Lemma~\ref{lemma:pessimistic-Q}, with probability at least $1-\delta$ one has 
\begin{equation}
0\leq \Delta_t(s) \leq \frac{1}{1-\gamma},\qquad \forall s\in \cS \text{ and }0\leq t\leq T.
\label{eq:Delta-chi-offlineQ-bound}
\end{equation}
\end{subequations}

\paragraph{Step 3: decomposing the suboptimality gap and introducing auxiliary sequences.}
Building on the recursion and pessimism properties established in the
preceding steps, we now decompose the suboptimality gap into several terms that will be controlled separately. We begin with the following upper
bound: 
\begin{align}
V^\star(\rho)-V^{\widehat{\pi}}(\rho)
&=  \left\langle \rho,V^\star-V^{\pi_T}\right\rangle
\le \left\langle \rho, \Delta_T\right\rangle
\le \frac{1}{T}\sum_{t=1}^{T}\left\langle \rho,\Delta_t\right\rangle 
\label{eq:performance-to-alpha0}
\end{align}
with $\Delta_t$ defined in~\eqref{eq:Delta-chi-offlineQ},  where the first inequality invokes the pessimism property in Lemma~\ref{lemma:pessimistic-Q}, and the second inequality comes from the monotonicity of $V_t$ in $t$. In view of the recursion~\eqref{eq:qstar-recursion-lcbq} established in Lemma~\ref{lemma:Q-recursion-pessimistic-Q}, we can deduce that
\begin{align}
\sum_{t=1}^{T}\left\langle \rho,\Delta_t\right\rangle 
& \leq \sum_{t=1}^{T}\sum_{s\in\cS}  \rho(s) \min\left\{Q^{\star}\big(s,\pi^{\star}(s)\big) - Q_t\big(s, \pi^{\star}(s)\big) , \frac{1}{1-\gamma} \right\}  \nonumber\\
&\leq \underbrace{ \gamma \sum_{t=1}^{T}\sum_{s\in\cS}  \rho(s) \sum_{i=1}^{n_t(s, \pi^{\star}(s))} \eta_i^{n_t(s,\pi^{\star}(s))} P_{s,\pi^{\star}(s)} \Delta_{k_i(s, \pi^{\star}(s))-1} }_{\eqqcolon\, \zeta}   \nonumber \\
& \qquad  +   \sum_{t=1}^{T}\sum_{s\in\cS} \rho(s) \min\left\{ \beta_{n_t(s,\pi^{\star}(s))}, \frac{1}{1-\gamma} \right\},\label{eq:first_step_rewrite}
\end{align}
where $\beta_n$ has been defined in~\eqref{eq:beta-bar-beta-lcbq}. 
Here, the first relation follows by combining \eqref{eq:value-bounded-lcbq} and the facts $V^{\star}(s) = Q^{\star}\big(s, \pi^{\star}(s)\big)$ and  $V_t(s) \geq \max_a Q_t(s,a) \geq Q_t(s, \pi^{\star}(s))$, while the last inequality arises from Lemma~\ref{lemma:Q-recursion-pessimistic-Q}. This leaves us with two terms to control: the first reflects the
propagation of value-estimation errors through the dynamics induced by the optimal policy,
whereas the second captures the cumulative effect
of the penalty terms used to compensate for finite-sample uncertainty.

To analyze these terms, we introduce several auxiliary quantities. For each $j\ge 0$, set
\begin{equation}
 d_j(s,a)\coloneqq \big[\rho^{\top}(P_{\pi^\star})^j\big](s)\ind\big\{a=\pi^\star(s) \big\},
\label{eq:dj-offlineQ}
\end{equation}
where we recall the definition of $P_{\pi}$ in~\eqref{eqn:ppivq}. 
In addition, for every $j\ge0$, set 
\begin{subequations}\label{eq:five-sequence-def-offlineQ}
\begin{align}
\alpha_j
& \coloneqq \left[\gamma\left(1+\frac{1}{H}\right)^{3}\right]^j\sum_{t=1}^{T} \rho^{\top}(P_{\pi^\star})^j \Delta_t, 
\label{eq:alpha-j-offlineQ}\\
\theta_j
&\coloneqq \left[\gamma\left(1+\frac{1}{H}\right)^{3}\right]^j\sum_{t=1}^{T}\sum_{s\in\cS}\big[\rho^{\top}(P_{\pi^\star})^j\big](s)
\min\left\{ \beta_{n_{t}\left(s,\pi^{\star}\left(s\right)\right)},\frac{1}{1-\gamma}\right\},
\label{eq:theta-j-offlineQ}\\
\xi_j
&\coloneqq \left[\gamma\left(1+\frac{1}{H}\right)^{3}\right]^{j+1} \rho^{\top} (P_{\pi^\star})^{j+1}\Delta_0,
\label{eq:xi-j-offlineQ}\\
\psi_j
&\coloneqq
\gamma\left[\gamma\left(1+\frac{1}{H}\right)^{3}\right]^j\sum_{t=1}^{T}
\Bigg\{
\sum_{(s,a)\in\cS\times \cA}d_j(s,a)
\sum_{i=1}^{n_t(s,a)}\eta_i^{n_t(s,a)}P_{s,a}\Delta_{k_i(s,a)-1}
\nonumber\\
&\hspace{3em}
-\left(1+\frac1H\right)
\frac{d_j(s_t,a_t)}{\myrho(s_t,a_t)}
\sum_{i=1}^{n_t(s_t,a_t)}\eta_i^{n_t(s_t,a_t)}P_{s_t,a_t}\Delta_{k_i(s_t,a_t)-1}
\Bigg\},
\label{eq:psi-j-offlineQ}\\
\phi_j
&\coloneqq
\gamma\left[\gamma\left(1+\frac{1}{H}\right)^{3}\right]^j\left(1+\frac1H\right)^2
\sum_{t=1}^{T}
\Bigg\{
\frac{d_j(s_t,a_t)}{\myrho(s_t,a_t)}P_{s_t,a_t}\Delta_{t-1}
\arxiv{ \nonumber\\
&\hspace{6em} }
-\left(1+\frac1H\right)
\sum_{(s,a)\in\cS\times \cA}d_j(s,a)P_{s,a}\Delta_{t-1}
\Bigg\}.
\label{eq:phi-j-offlineQ}
\end{align}
\end{subequations}
Equipped with the above set of notation, the original error bound~\eqref{eq:first_step_rewrite} can be expressed as
\begin{equation} \label{eq:alpha0-zeta-theta-offlineQ}
\alpha_0\le \zeta +\theta_0.
\end{equation}
In the subsequent analysis, we will recursively propagate $\zeta$ through the dynamics induced by
the optimal policy, which gives rise to five types of
contributions that we bound separately.

Before proceeding, we briefly interpret the quantities defined above. At a high level, the first three sequences describe the propagation of
the main error terms through the dynamics induced by the optimal policy:
\begin{itemize}
\item Each
$\alpha_j$ (cf.~\eqref{eq:alpha-j-offlineQ}) measures the cumulative value estimation error after propagation through $j$ steps of the MDP dynamics induced by the optimal policy; 

\item Each $\theta_j$ (cf.~\eqref{eq:theta-j-offlineQ}) is the penalty-related counterpart of
    $\alpha_j$, reflecting the influence of the penalties
    along the same dynamics;

\item Each $\xi_j$ (cf.~\eqref{eq:xi-j-offlineQ}) accounts for the initialization error after
    $j+1$ such steps, reflecting the time shift introduced
    in the recursive analysis.
\end{itemize}
The remaining two sequences arise from replacing population-level
quantities under the offline data distribution $\myrho$ with their
empirical counterparts. More specifically,
\begin{itemize}
    \item Each $\psi_j$ (cf.~\eqref{eq:psi-j-offlineQ}) captures the
    discrepancy between the value estimation error accumulated over the visits to each
    state-action pair and its corresponding importance-weighted
    contribution evaluated at the sampled state-action pairs. Thus,
    $\psi_j$ arises from regrouping the recursive error according to
    state-action visits.

    \item Each $\phi_j$ (cf.~\eqref{eq:phi-j-offlineQ}) captures the statistical fluctuation in the one-step propagation of the value estimation error.
\end{itemize}
%

%As we shall see momentarily, the terms $\psi_j$ and $\phi_j$ will be controlled using suitable concentration inequalities.

\paragraph{Step 4: recursion along the dynamics.}
The decomposition \eqref{eq:alpha0-zeta-theta-offlineQ} leaves us with two terms to control. We begin with $\zeta$, and recursively propagate its error along the
dynamics induced by the optimal policy. Recalling the definition \eqref{eq:psi-j-offlineQ} of $\psi_0$ and $d_0(s,a)=\rho(s)\ind\{a=\pi^\star(s)\}$ (see \eqref{eq:dj-offlineQ}), one has
\begin{align}
\zeta
&=
\gamma\left(1+\frac1H\right)
\sum_{t=1}^{T}
\frac{d_0(s_t,a_t)}{\myrho(s_t,a_t)}
\sum_{i=1}^{n_t(s_t,a_t)}
\eta_i^{n_t(s_t,a_t)}
P_{s_t,a_t}\Delta_{k_i(s_t,a_t)-1}
+\psi_0.
\label{eq:zeta0-to-sample-offlineQ}
\end{align}
Roughly speaking, $\zeta$ is an importance-weighted, sample-based accumulation of propagated value estimation errors. Our strategy is to repeatedly replace such sample-based quantities by their one-step
population counterpart under the optimal dynamics.

To do so, let us first regroup the above sum according to visits to each state-action pair. Consider any $(s,a)$, whose contribution to the sum in~\eqref{eq:zeta0-to-sample-offlineQ} --- up to some multiplicative factor --- is given by
\begin{align}
&\sum_{t=1}^{T}
\ind\{(s_t,a_t)=(s,a)\}
\sum_{i=1}^{n_t(s,a)}
\eta_i^{n_t(s,a)}
P_{s,a}\Delta_{k_i(s,a)-1}
\nonumber\\
&\quad=
\sum_{m=1}^{n_T(s,a)}
\sum_{i=1}^{m}
\eta_i^m
P_{s,a}\Delta_{k_i(s,a)-1}
=
\sum_{i=1}^{n_T(s,a)}
\Bigg(\sum_{m=i}^{n_T(s,a)}\eta_i^m\Bigg)
P_{s,a}\Delta_{k_i(s,a)-1}.
\label{eq:visit-regrouping-zeta0-offlineQ}
\end{align}
It is seen from Lemma~\ref{lemma:property_lr} that
\[
\sum_{m=i}^{n_T(s,a)}\eta_i^m
\le
\sum_{m=i}^{\infty}\eta_i^m
\le 1+\frac1H,
\]
which combined with the fact that $\Delta_t\ge0$ (see~\eqref{eq:Delta-chi-offlineQ-bound}) gives
\begin{align}
\eqref{eq:visit-regrouping-zeta0-offlineQ}
&\le
\left(1+\frac1H\right)
\sum_{i=1}^{n_T(s,a)}
P_{s,a}\Delta_{k_i(s,a)-1}
= \left(1+\frac1H\right) \sum_{t=1}^{T}
\ind\{(s_t,a_t)=(s,a)\}
P_{s,a}\Delta_{t-1}.
\label{eq:zeta0-visit-bound-offlineQ}
\end{align}
Combining \eqref{eq:visit-regrouping-zeta0-offlineQ} and
\eqref{eq:zeta0-visit-bound-offlineQ} and summing over $(s,a)$, we reach
\begin{align*}
&\sum_{t=1}^{T}
\frac{d_0(s_t,a_t)}{\myrho(s_t,a_t)}
\sum_{i=1}^{n_t(s_t,a_t)}
\eta_i^{n_t(s_t,a_t)}
P_{s_t,a_t}\Delta_{k_i(s_t,a_t)-1}
\arxiv{\nonumber\\
&\quad}\le
\left(1+\frac1H\right)
\sum_{t=1}^{T}
\frac{d_0(s_t,a_t)}{\myrho(s_t,a_t)}
P_{s_t,a_t}\Delta_{t-1}.
%\label{eq:zeta0-grouped-offlineQ}
\end{align*}
Substitution into
\eqref{eq:zeta0-to-sample-offlineQ} leads to
\begin{align}
\zeta
&\le
\gamma\left(1+\frac1H\right)^2
\sum_{t=1}^{T}
\frac{d_0(s_t,a_t)}{\myrho(s_t,a_t)}
P_{s_t,a_t}\Delta_{t-1}
+\psi_0 \notag \\
%\label{eq:zeta0-intermediate-offlineQ}
&\leq 
\gamma\left(1+\frac1H\right)^3
\sum_{t=1}^{T}
\sum_{(s,a)\in\cS\times\cA}
d_0(s,a)P_{s,a}\Delta_{t-1}
+\phi_0 +\psi_0,
\label{eq:zeta0-phi-application-offlineQ} 
\end{align}
where the last relation arises from the definition of $\phi_0$ (cf.~\eqref{eq:phi-j-offlineQ}).

We now connect \eqref{eq:zeta0-phi-application-offlineQ} with $\alpha_1$ (defined in \eqref{eq:alpha-j-offlineQ}). Since $\Delta_t\ge0$ (see Lemma~\ref{lemma:pessimistic-Q}), one has
\begin{align}
\sum_{t=1}^{T}
\sum_{(s,a)}
d_0(s,a)P_{s,a}\Delta_{t-1}
&=
\sum_{(s,a)}d_0(s,a)P_{s,a}\Delta_0
+
\sum_{t=1}^{T-1}
\sum_{(s,a)}d_0(s,a)P_{s,a}\Delta_t
\nonumber\\
&\le
\rho^\top P_{\pi^\star}\Delta_0
+
\sum_{t=1}^{T}
\sum_{(s,a)}
d_0(s,a)P_{s,a}\Delta_t,
\label{eq:zeta0-time-shift-offlineQ}
\end{align}
where the last line is valid since
\begin{align*}
\sum_{(s,a)}d_0(s,a)P_{s,a}\Delta_0
&= 
\sum_{(s,a)}\rho(s)\mathds{1}\big\{a=\pi^{\star}(s)\big\} P_{s,a}\Delta_0
\arxiv{\notag\\
& }=
\sum_{s}\rho(s) P_{s,\pi^{\star}(s)}\Delta_0 =
\rho^\top P_{\pi^\star}\Delta_0.
\end{align*}
Consequently, we can use the definition of $\xi_0$ (cf.~\eqref{eq:xi-j-offlineQ}) to show that
\begin{align}
&\gamma\left(1+\frac1H\right)^3
\sum_{t=1}^{T}
\sum_{(s,a)}
d_0(s,a)P_{s,a}\Delta_{t-1}
\arxiv{\nonumber\\
&\quad}\le
\alpha_1
+
\gamma\left(1+\frac1H\right)^3
\rho^\top P_{\pi^\star}\Delta_0
\arxiv{\nonumber\\
&\quad}=
\alpha_1+\xi_0.
\label{eq:zeta0-alpha1-xi0-offlineQ}
\end{align}
Substitution into~\eqref{eq:zeta0-phi-application-offlineQ} leads to
\begin{equation*}
\zeta
\le
\alpha_1+\xi_0+\phi_0+\psi_0,
%\label{eq:zeta0-recursion-offlineQ}
\end{equation*}
which combined with~\eqref{eq:alpha0-zeta-theta-offlineQ} yields
\begin{equation}
\alpha_0
\le
\alpha_1+\xi_0+\theta_0+\psi_0+\phi_0.
\label{eq:alpha0-recursion-offlineQ}
\end{equation}

Repeating similar calculations --- with $\rho^{\top}(P_{\pi^\star})^j$ in place of $\rho^{\top}$ --- allows us to derive recursively that, for all $j\ge0$,
\begin{equation}
\alpha_j\le \alpha_{j+1}+\xi_j+\theta_j+\psi_j+\phi_j.
\label{eq:alpha-recursion-offlineQ}
\end{equation}
Iterating \eqref{eq:alpha-recursion-offlineQ} leads to
\begin{equation}
\alpha_0\le \alpha+\xi+\theta+\psi+\phi,
\label{eq:alpha0-five-terms-offlineQ}
\end{equation}
where we define
\begin{align}
\alpha&\coloneqq\limsup_{j\to\infty}\alpha_j,
&
\xi&\coloneqq\sum_{j=0}^{\infty}\xi_j,
&
\theta&\coloneqq\sum_{j=0}^{\infty}\theta_j,
&
\psi&\coloneqq\sum_{j=0}^{\infty}\psi_j,
&
\phi&\coloneqq\sum_{j=0}^{\infty}\phi_j.
\label{eq:five-terms-def-offlineQ}
\end{align}

\paragraph{Step 5: bounding the terms separately.}
It remains to bound the five terms in \eqref{eq:five-terms-def-offlineQ} separately. Before continuing, we find it helpful to introduce the notation
\begin{align}
\kappa \coloneqq \gamma \left( 1+\frac{1}{H} \right)^3 < 1, 
\label{eq:defn-kappa-proof-lcbq}
\end{align}
where the inequality follows immediately from the choice~\eqref{eq:defn-H-lcbq-stepsize-proof} of $H$. 
Moreover, we single out the following lemma that will be used multiple times throughout the remaining proof. 
\begin{lemma}
\label{lemma:useful}
Consider any $0<\delta<1$, and recall the definition of $H$ and $\kappa$ in \eqref{eq:defn-H-lcbq-stepsize-proof} and \eqref{eq:defn-kappa-proof-lcbq}, respectively. For any vector $W\in\mathbb{R}^{S}$ with nonnegative entries, we have
\begin{equation}
\sum_{j=0}^{\infty}\kappa ^{j} \rho^{\top}(P_{\pi^{\star}})^{j}W \lesssim\frac{1}{1-\gamma}\big\langle d_{\rho}^{\star},W\big\rangle +\frac{\delta}{ST^{4}\left(1-\gamma\right)}\left\Vert W\right\Vert _{\infty}.\label{eq:useful}
\end{equation}
\end{lemma}
Informally speaking, Lemma~\ref{lemma:useful} links the geometrically weighted propagation of a
nonnegative object --- along the dynamics induced by the optimal policy --- to its weighted average (weighted by the discounted occupancy distribution). The proof of this lemma is provided in Section~\ref{sec:proof-lemma:useful-lcbq-step5}.

With Lemma~\ref{lemma:useful} in place, we can now bound the terms in~\eqref{eq:five-terms-def-offlineQ}. The resulting bounds are summarized
in the following lemma.
\begin{lemma}
\label{lem:lcbq-alpha-xi-psi-phi-bound}
With probability at least $1-\delta$, the terms in~\eqref{eq:five-terms-def-offlineQ} satisfy
\begin{align}
\alpha=0,\quad\xi\lesssim\frac{1}{(1-\gamma)^{2}},\quad\theta\lesssim\sqrt{\frac{\Cstar ST\iota^{2}}{(1-\gamma)^{5}}},\quad\psi+\phi\lesssim\frac{\Cstar\iota^{2}}{(1-\gamma)^{3}}.
\end{align}
\end{lemma}
At a high level, $\alpha$ is controlled by geometric contraction, while $\xi$ is bounded by means of Lemma~\ref{lemma:useful}. The term $\theta$ is handled by combining concentration of the visitation counts with the single-policy concentrability assumption. Finally, $\psi$ is controlled using martingale concentration together
with a one-step time-shift decomposition, whereas $\phi$ follows
directly from martingale concentration because its summands are
predictable. The proof is deferred to Section~\ref{sec:proof:lem:lcbq-alpha-xi-psi-phi-bound}.

\paragraph{Step 6: putting all this together.}
Recall that $\Cstar\geq 1$. 
Putting \eqref{eq:alpha-bound-offlineQ}, \eqref{eq:xi-bound-offlineQ}, \eqref{eq:theta-bound-offlineQ}, and \eqref{eq:psi-phi-bound-offlineQ} together with \eqref{eq:alpha0-five-terms-offlineQ}, we arrive at 
\begin{align}
\alpha_0
&\lesssim
\sqrt{\frac{\Cstar S T\iota^2}{(1-\gamma)^5}}
+
\frac{1}{(1-\gamma)^2}+
\frac{\Cstar \iota^2}{(1-\gamma)^3}
 \asymp \sqrt{\frac{\Cstar S T\iota^2}{(1-\gamma)^5}},
%\label{eq:alpha0-bound-final-offlineQ}
\end{align}
where the last relation arises from the condition on $T$ in
\eqref{eq:sample-size-lcbq-T-thm}. Therefore, by
\eqref{eq:performance-to-alpha0} and the definition of $\alpha_0$ in
\eqref{eq:alpha-j-offlineQ}, one arrives at
\[
V^\star(\rho)-V^{\widehat{\pi}}(\rho) \leq \frac{1}{T}\alpha_0 \lesssim 
\sqrt{\frac{\Cstar S \iota^2}{(1-\gamma)^5T}}\leq \varepsilon,
\]
provided that the universal constant $c_2$ in the condition~\eqref{eq:sample-size-lcbq-T-thm} is sufficiently large.

\section{Minimax lower bounds}
\label{sec:minimax-lower-bound-offline}

While we have previously highlighted the sample optimality of the \vilcb algorithm, a matching lower bound is still needed in order to rigorously establish its minimax optimality. We present such a result below, which serves as a benchmark for assessing the optimality or suboptimality of the offline RL algorithms developed in this chapter and more broadly in the literature. 

\begin{theorem}
\label{thm:infinite-lwoer-bound}
Consider any $(\gamma,S,\Cstar,\varepsilon)$ satisfying
$\gamma\in\left[\frac{2}{3},1\right)$, 
$S>12$, $\Cstar> 2$,
and $\varepsilon\leq\frac{1}{672(1-\gamma)}$. There exist a class of MDPs
$\{\mathcal{M}_{\theta}:\theta\in\Theta\}$ --- each with $S$ states and 2 actions --- a test state
distribution $\rho\in \Delta(\cS)$, and a batch dataset with $N$ independent samples
whose single-policy concentrability coefficient lies within the range $[\Cstar,2\Cstar]$, 
such that
\begin{align}
\inf_{\widehat{\pi}}\max_{\theta\in\Theta}
\mathbb{P}_{\mathcal{M}_{\theta}}
\left(
V_\theta^\star(\rho)
-
V_\theta^{\widehat{\pi}}(\rho)
>
\varepsilon
\right)
\geq \frac{1}{4},
\label{eq:infinite-lower-bound-theorem}
\end{align}
provided that
\[
N\leq
\frac{c_2S\Cstar}
{(1-\gamma)^3\varepsilon^2}
\]
for some universal numerical constant $c_2>0$. Here, the infimum is
taken over all policy estimators $\widehat{\pi}$, and
$\mathbb{P}_{\mathcal{M}_{\theta}}$ denotes the probability under
$\mathcal{M}_{\theta}$.
\end{theorem}
\begin{remark}
A sharp minimax lower bound under the single-policy clipped concentrability coefficient
$C^{\star}_{\rho,\mathsf{clipped}}$, defined in~\eqref{eq:concentrate-infinite},
was established in \citet{li2024settling}. Theorem~\ref{thm:infinite-lwoer-bound}
instead concerns the single-policy concentrability coefficient $\Cstar$.
Although the proof presented here is inspired by that work, it differs from
the proof therein in several important respects.
\end{remark}

In words, Theorem~\ref{thm:infinite-lwoer-bound} asserts that no algorithm
can reliably identify an $\varepsilon$-optimal policy unless the sample
size is at least on the order of
\[
\Omega\left(
\frac{S\Cstar}{(1-\gamma)^3\varepsilon^2}
\right).
\]
As before, the proof relies on standard information-theoretic techniques, in particular
a reduction to a multiple-hypothesis testing problem and an application
of Fano's lemma. Taken together with
Theorem~\ref{thm:infinite-offline}, this lower bound confirms the
minimax optimality of model-based offline RL algorithms in terms of
sample complexity, up to logarithmic factors.

\subsection{Analysis: construction of hard problem instances}
\label{sec:analysis-lower-bound-offline-part1}

In this subsection, we construct a collection of hard instances that will
be used in Section~\ref{sec:analysis-lower-bound-offline-part2} to establish
the minimax lower bound in Theorem~\ref{thm:infinite-lwoer-bound}. We assume throughout that
\begin{equation}
	\frac{2}{3}\leq \gamma < 1 \qquad \text{and} \qquad \frac{224(1-\gamma)\varepsilon}{\gamma} \leq \frac{1}{2}.
	%\varepsilon \leq \frac{\gamma}{28(1-\gamma)}. 
	\label{eq:infinite-epsilon-assumption}
\end{equation}

\paragraph{Construction of the hard MDPs.}
We introduce a class of MDPs
\begin{align}
\left\{ \mathcal{M}_{\theta} = (\mathcal{S}, \mathcal{A}, P_{\theta}, r, \gamma) \mid \theta \in \Theta \subseteq  \{0,1\}^{S-1} \right\},
\label{eq:set-MDP-Mtheta-lower-bound-offline}
\end{align}
parameterized by $\theta \in \Theta$, where $\Theta$ is a subset in $\{0,1\}^{S-1}$ to be specified shortly. Each MDP $\mathcal{M}_{\theta}$ consists of $S$ states and 2 actions:
\begin{align*}
	\cS = \{0, 1, \ldots, S-1\} \qquad \text{and} \qquad  \mathcal{A} = \{0, 1\}.
\end{align*}
To facilitate the construction and its subsequent analysis, we introduce
the following state distribution:
\begin{align}  \label{infinite-mu-assumption}
\mu(s) = \frac{1}{C(S-1)}\mathds{1}\{s > 0\} + \bigg(1 - \frac{1}{C}\bigg)\mathds{1}\{s = 0\}
, \qquad 
\forall s\in \cS,
\end{align}
where $C> 2$ is some quantity to be specified later (in fact, as we will show shortly in
\eqref{eq:expression-Cstar-LB-inf}, $C$ is within a factor of two of the
single-policy concentrability coefficient $\Cstar$). 
With this distribution in place, we define the transition kernel $P_{\theta}$ of the MDP $\mathcal{M}_\theta$ as follows:
\begin{align}
	P_\theta(s^{\prime} \mymid s, a) = \left\{ \begin{array}{ll}
		p\mathds{1}\{s^{\prime} = s\} + (1-p)\mu(s^{\prime}), & \text{if } s>0, a=\theta_s, \\
		q\mathds{1}\{s^{\prime} = s\} + (1-q)\mu(s^{\prime}), & \text{if } s>0, a=1-\theta_s, \\
		\mathds{1}\{s^{\prime} = 0\}, & \text{if } (s, a) = (0, 0), \\ 
		(2\gamma-1)\mathds{1}\{s^{\prime} = 0\} + 2(1-\gamma)\mu(s^{\prime}), & \text{if } (s, a) = (0, 1).
		% \gamma\mathds{1}\{s^{\prime} = s\} + (1-\gamma)\mu(s^{\prime}) & \text{for}   & s > 1 .
	\end{array}\right.
	\label{eq:defn-Ptheta-defn-inf}
\end{align}
Here, $p$ and $q$ are two parameters chosen as
\begin{align}
	\label{eq:infinite-p-q}
	p = \gamma + \frac{224(1-\gamma)^2\varepsilon}{\gamma},\qquad q = \gamma - \frac{224(1-\gamma)^2\varepsilon}{\gamma},
\end{align}
which, under the assumption \eqref{eq:infinite-epsilon-assumption}, satisfy
\begin{equation}
	p > q \geq \gamma - \frac{1-\gamma}{2} \geq \frac{1}{2}.
	\label{eq:p-q-relation-LB-inf}
\end{equation}
%
% It is clear from the construction that, if the MDP is initialized at either state $0$ or state $1$, then it never leaves the state subset $\{0,1\}$.  
The reward function for each MDP $\mathcal{M}_\theta$ is taken to be
\begin{align}
	r(s, a) = \left\{ \begin{array}{ll}
		1, & \text{if } s > 0, \\
		\frac{4}{7}, & \text{if } (s, a) = (0, 0), \\ 
		0, & \text{if } (s, a) = (0, 1).
		% 0 & \text{if } s > 1 .
	\end{array}\right.
	\label{eq:defn-r-theta-defn-inf}
\end{align}
%
% Thus, the reward at state $0$ is higher than that at the other states. 

Finally, we describe our choice of the parameter set $\Theta \subseteq  \{0, 1\}^{S-1}$. By virtue of the Gilbert-Varshamov lemma \citep{gilbert1952comparison}, 
one can construct a set $\Theta \subseteq  \{0, 1\}^{S-1}$ such that
\begin{equation}
	|\Theta| \ge e^{(S-1)/8}
	\qquad \text{and} \qquad 
	\|\theta - \widetilde{\theta}\|_1 \ge \frac{S-1}{8}
	\label{eq:property-Theta} 
\end{equation}
for every $\theta,\widetilde{\theta}\in \Theta$ obeying $\theta \ne \widetilde{\theta}$. 
In other words, the set $\Theta$ can be constructed to contain an  exponentially large number of pairwise well-separated vectors, a property that will be important in the ensuing analysis.

\paragraph{Value functions and optimal policies.}
We next compute the value functions of the constructed MDPs and identify their optimal policies. 
For notational clarity, for the MDP $\mathcal{M}_{\theta}$, 
we denote by $\pi_{\theta}^{\star}$ its optimal policy,  
and let $V_{\theta}^{\pi}$ and $V_{\theta}^{\star}$ denote,
respectively, the value function of policy $\pi$ and the optimal
value function in $\mathcal{M}_{\theta}$.   
The following lemma collects several useful properties of these value functions and optimal policies. The proof of this lemma is deferred to Section~\ref{proof:lemma:value-policy-LB-inf}. 
\begin{lemma}
	\label{lemma:value-policy-LB-inf}
	Consider any $\theta\in \{0,1\}^{S-1}$. One has
    \begin{subequations}
	\begin{align}
		V_\theta^\pi(s)
		&\le \frac{1 + \gamma \big(1-x_{\pi,\theta}(s)\big)\big(C^{-1}V^{\star}_{\theta}(s)+(1-C^{-1})V^{\star}_{\theta}(0)\big)}{1 - \gamma x_{\pi,\theta}(s)},
        \label{eq:infinite-value-0-expression}
	\end{align}
    for every policy $\pi$ and every $s>0$, 
	where we define 
    \begin{equation}
    x_{\pi,\theta}(s) \coloneqq p\pi(\theta_s\mymid s) + q\pi(1-\theta_s\mymid s).
    \label{eq:defn-x-pi-theta-s-lb-proof-lcbq}
    \end{equation}
    \end{subequations}
    Also, equality in~\eqref{eq:infinite-value-0-expression} is attained when $\pi = \pi^{\star}$.
	In addition, the optimal policy $\pi_{\theta}^{\star}$ and the optimal value function satisfy, for every $s > 0$
\begin{subequations}
\label{eq:optimal-pi-0-inf-135}
	\begin{align}
		\pi_\theta^\star(0 \mymid 0) = 1, \quad  
		%\\
		\pi^{\star}_{\theta}(\theta_s\mymid s) = 1,
	\end{align}
    and
    \begin{align}
    V^{\star}_{\theta}(0)= \frac{4}{7(1-\gamma)}
    \quad \text{and} \quad 
    V^{\star}_{\theta}(s) = \frac{1 + \gamma(1-p)(1-C^{-1})V^{\star}_{\theta}(0)}{1-\gamma\big(p + (1-p)C^{-1}\big)}.
    \end{align}
\end{subequations}
\end{lemma}

\paragraph{Construction of the batch dataset.}
Given any constructed MDP $\mathcal{M}_\theta$, we now generate a dataset consisting of $N$ i.i.d.~samples $\{(s_i,a_i,s_i')\}_{1\leq i\leq N}$ according to \eqref{eq:sampling-offline-inf-iid},  
where the data distribution satisfies
\begin{align}
	\myrho(s) =  \mu(s), \qquad
	\myrho(s,a) =  \mu(s)/2, \qquad \forall (s,a)\in\cS\times \cA, 
	\label{eq:db-sa-inf-LB}
\end{align}
with $\mu$ defined in \eqref{infinite-mu-assumption}. This distribution
can be realized in $\mathcal{M}_{\theta}$ by executing the behavior policy
$\pi^{\mathsf b}$ from the initial state distribution $\rhob$ given by
\begin{align}
\rhob(s)=\mu(s),
\qquad
\pib(a\mymid s)=\frac12,
\qquad
\forall (s,a)\in\cS\times\cA.
\label{eq:defn-rhob-pib-lcb-lb}
\end{align}
Indeed, as will be shown in Section~\ref{sec:proof:eq:db-Cstar-inf-LB},
$\mu$ is stationary under this behavior policy, which yields the
occupancy distribution in \eqref{eq:db-sa-inf-LB}.

Moreover, we choose the test distribution $\rho$ as
\begin{align}
	\rho(s) = 
	\begin{cases} 
		\frac{1}{S-1}, \quad &\text{if }s>0, \\
		0, &\text{if }s=0.
	\end{cases}
	\label{eq:rho-defn-inf-LB}
\end{align}
Then, as we shall justify in Section~\ref{sec:proof:eq:db-Cstar-inf-LB}, the single-policy concentrability coefficient $\Cstar$ of the dataset w.r.t.~the constructed MDP $\mathcal{M}_{\theta}$ satisfies
\begin{align}
	C < C_{\rho}^{\star} < 2C. 
	\label{eq:expression-Cstar-LB-inf}
\end{align}

\subsection{Analysis: establishing the minimax lower bound}
\label{sec:analysis-lower-bound-offline-part2}

Equipped with the construction of hard instances in Section~\ref{sec:analysis-lower-bound-offline-part1}, we now establish the claimed minimax
lower bound.  
Under the test distribution $\rho$ defined in
\eqref{eq:rho-defn-inf-LB}, the suboptimality gap reduces to
\[
\big\langle \rho, V_\theta^{\star} - V_\theta^{\widehat{\pi}}\big\rangle
= \frac{1}{S-1}\sum_{s=1}^{S-1}\big(V_\theta^{\star}(s) - V_\theta^{\widehat{\pi}}(s)\big),
\]
where $\widehat{\pi}$ denotes a policy estimate computed from the batch dataset. Thus, it suffices to control the aggregate suboptimality over the 
states $s>0$. In what follows, we let $\mathbb{P}_{\theta}$ denote the probability distribution under the MDP $\mathcal{M}_{\theta}$.

\paragraph{Step 1: converting $\widehat{\pi}$ into a parameter estimate for $\theta$.} 
At a high level, this step seeks to reduce policy learning to estimating the parameter $\theta$. Towards this end, consider first an arbitrary policy $\pi$. We claim that
\begin{align}
	\arxiv{&} \big\langle \rho, V_\theta^{\star} - V_\theta^{\pi}\big\rangle 
	&=
    \frac{1}{S-1}\sum_{s=1}^{S-1} 
     \big(V_\theta^{\star}(s) - V_\theta^{\pi}(s)\big) \arxiv{\notag\\ &}
    \geq \frac{24\varepsilon}{S-1}\sum_{s=1}^{S-1}\big( 1 - \pi(\theta_s \mymid s) \big),
	\label{eq:infinite-delta-key-auxiliary}
\end{align}
whose proof is postponed to Section~\ref{sec:proof:eq:db-Cstar-inf-LB}. Since the optimal action at state $s$ is $\theta_s$ (see \eqref{eq:optimal-pi-0-inf-135} in Lemma~\ref{lemma:value-policy-LB-inf}), the claim~\eqref{eq:infinite-delta-key-auxiliary} indicates that every
state where $\pi$ chooses a suboptimal action contributes a nontrivial
amount to the overall suboptimality gap.

Now suppose that a policy estimator $\widehat{\pi}$ achieves
\begin{align}
\mathbb{P}_{\theta} \big\{  \big\langle \rho, V_\theta^{\star} - V_\theta^{\widehat{\pi}} \big\rangle \leq \varepsilon  \big\} \geq \frac{3}{4}.
\label{eq:sub-optimality-gap-proof-34-lb-offline}
\end{align}
Then~\eqref{eq:infinite-delta-key-auxiliary} and $\pi_{\theta}^{\star}(\theta_s\mymid s)=1$ (see Lemma~\ref{lemma:value-policy-LB-inf}) imply that, with probability at least $3/4$,
\begin{align}
\sum_{s=1}^{S-1}\big\|\widehat{\pi}(\cdot \mymid s) - \pi^{\star}_{\theta}(\cdot \mymid s)\big\|_1 
=
2\sum_{s=1}^{S-1}\big( 1 - \widehat{\pi}(\theta_s \mymid s) \big)
\leq \frac{S-1}{12}.
\label{eq:pihat-pistar-gap-theta-lb-lcbq}
\end{align} 
Let us convert $\widehat{\pi}$ into an estimator of $\theta$ by defining
\begin{align}
	\widehat{\theta}=\arg \min_{\overline{\theta} \in\Theta} ~\sum_{s=1}^{S-1} \big\|\widehat{\pi}(\cdot\mymid s) - \pi^{\star}_{\overline{\theta}}(\cdot\mymid s)\big\|_1,
	\label{eq:defn-theta-hat-inf-LB}
\end{align}
with ties broken deterministically. 
As it turns out, $\widehat{\theta}$ is a good estimator of $\theta$ in the sense that
\begin{align}
    \mathbb{P}_{\theta}\big(\widehat{\theta} = \theta\big) 
	\geq \frac{3}{4}.
	\label{eq:P-theta-accuracy-inf}
\end{align}
To justify~\eqref{eq:P-theta-accuracy-inf}, note that for any $\widetilde{\theta} \in \Theta$ with $\widetilde{\theta}  \neq \theta$, one necessarily has
\begin{align}
	\arxiv{&} \sum_{s=1}^{S-1} \big \|\widehat{\pi}(\cdot\mymid s) - \pi^{\star}_{\widetilde{\theta}}(\cdot\mymid s) \big\|_1 \arxiv{ \notag\\ }
	&\arxiv{\qquad} \geq   \sum_{s=1}^{S-1}  \big\|\pi^{\star}_{\theta}(\cdot\mymid s) - \pi^{\star}_{\widetilde{\theta}}(\cdot\mymid s)\big\|_1 - \sum_{s=1}^{S-1} \big\| \widehat{\pi}(\cdot\mymid s) 
	-   \pi^{\star}_{\theta}(\cdot\mymid s)\big\|_1 \notag\\
    &\arxiv{\qquad} = 2\big\|\theta - \widetilde{\theta}\big\|_1 - \sum_{s=1}^{S-1}  \big\|\widehat{\pi}(\cdot\mymid s) - \pi^{\star}_{\theta}(\cdot\mymid s)\big\|_1  \notag\\
    &\arxiv{\qquad} \geq  \frac{S-1}{4} - \frac{S-1}{12} > \frac{S-1}{12} , 
    \label{eq:infinite-tilde-theta}
\end{align}
where the first inequality holds by the triangle inequality, 
the penultimate line arises since $\pi^{\star}_{\theta}(\theta_s \mymid s)=1$ (see Lemma~\ref{lemma:value-policy-LB-inf}), and the last line comes from the properties \eqref{eq:property-Theta} about $\Theta$. It then follows that
\begin{align*}
    \mathbb{P}_{\theta}\big(\widehat{\theta} = \theta\big) 
	\geq \mathbb{P}_{\theta}\left( \sum_{s=1}^{S-1}\big\|\widehat{\pi}(\cdot \mymid s) - \pi^{\star}_{\theta}(\cdot \mymid s)\big\|_1 \leq \frac{S-1}{12}\right) \geq  \frac{3}{4},
\end{align*}
as claimed.

Thus far, we have shown that a policy estimator satisfying~\eqref{eq:sub-optimality-gap-proof-34-lb-offline} yields an estimator of
$\theta$ satisfying~\eqref{eq:P-theta-accuracy-inf}. In the ensuing steps, we will show that \eqref{eq:P-theta-accuracy-inf} is information-theoretically impossible without more samples, thus leading to contradiction.

\paragraph{Step 2: probability of error in testing multiple hypotheses.}
We now turn attention to the task of recovering $\theta$, which is a $|\Theta|$-ary hypothesis testing problem. Consider the minimax probability of error
\begin{equation}
p_{\mathrm e}
\coloneqq
\inf_{\psi}
\max_{\theta\in\Theta}
\mathbb{P}_{\theta}(\psi\neq\theta),
\label{eq:error-prob-two-hypotheses-finite-LB}
\end{equation}
where the infimum is taken over all tests $\psi$ (constructed based on the batch dataset). Let $\mu^{\mathsf b,\theta}$ denote the distribution of one
observed transition $(s_i,a_i,s_i')$ under $\mathcal M_\theta$, meaning that
\[
\mu^{\mathsf b,\theta}(s,a,s')
=
\myrho(s,a)P_\theta(s'\mymid s,a).
\]
As in the lower-bound analyses in previous chapters, we connect the
difficulty of this testing problem with the KL divergences between the data distributions induced by different hypotheses.

Consider any reference hypothesis $\theta_0\in\Theta$. Since the $N$ samples are
independently generated, applying
Fano's lemma (i.e., Lemma~\ref{lem:fano}(ii)) with
$M=|\Theta|-1$ yields
\begin{align}
p_{\mathrm e}
&\geq
\frac{\log|\Theta|-\log 2}{\log(|\Theta|-1)}
-
\frac{N}{|\Theta|\log(|\Theta|-1)}
\sum_{\theta\in\Theta\setminus\{\theta_0\}}
\mathsf{KL}\big(
\mu^{\mathsf b,\theta}
\,\big\|\,
\mu^{\mathsf b,\theta_0}
\big)
\notag\\
&\geq
\frac{\log|\Theta|-\log 2}{\log(|\Theta|-1)}
-
\frac{N}{\log(|\Theta|-1)}
\max_{\theta\neq\widetilde\theta}
\mathsf{KL}\big(
\mu^{\mathsf b,\theta}
\,\big\|\,
\mu^{\mathsf b,\widetilde\theta}
\big),
\label{eq:Fano-LB-infinite}
\end{align}
where the first inequality uses the tensorization property of KL divergence
for independent samples (see Lemma~\ref{lem:KL-tensorize}). It remains to bound the one-sample KL divergence between $\mu^{\mathsf b,\theta}$ and $\mu^{\mathsf b,\widetilde{\theta}}$.

\paragraph{Step 3: bounding the one-sample KL divergence.}
Recall from \eqref{eq:db-sa-inf-LB} that $\myrho(s,a)=\mu(s)/2$. By the factorization of the one-sample
distribution, we have
\begin{align}
\mathsf{KL}\big(
\mu^{\mathsf b,\theta}
\,\big\|\,
\mu^{\mathsf b,\widetilde\theta}
\big)
&=
\sum_{s\in\cS}\sum_{a\in\{0,1\}}
\myrho(s,a)
\mathsf{KL}\big(
P_\theta(\cdot\mymid s,a)
\,\big\|\,
P_{\widetilde\theta}(\cdot\mymid s,a)
\big)
\notag\\
&=
\frac{1}{2C(S-1)}
\sum_{s=1}^{S-1}\sum_{a\in\{0,1\}}
\mathsf{KL}\big(
P_\theta(\cdot\mymid s,a)
\,\big\|\,
P_{\widetilde\theta}(\cdot\mymid s,a)
\big),
\label{eq:single-sample-KL-infinite}
\end{align}
where the last equality follows since the transition kernels at state
$0$ do not depend on $\theta$. 
The lemma below bounds the KL divergences appearing in
\eqref{eq:single-sample-KL-infinite}; the proof is provided in Section~\ref{sec:proof:lem:transition-KL-bound-infinite}.

\begin{lemma}
\label{lem:transition-KL-bound-infinite}
Consider any $\theta,\widetilde{\theta}\in\Theta$ and any $s>0$. 
If $\theta_s=\widetilde{\theta}_s$, then for every
$a\in\{0,1\}$,
\[
\mathsf{KL}\big(
P_\theta(\cdot\mymid s,a)
\,\big\|\,
P_{\widetilde{\theta}}(\cdot\mymid s,a)
\big)=0.
\]
If $\theta_s\neq\widetilde{\theta}_s$, then for every
$a\in\{0,1\}$,
\begin{align}
\mathsf{KL}\big(
P_\theta(\cdot\mymid s,a)
\,\big\|\,
P_{\widetilde{\theta}}(\cdot\mymid s,a)
\big)
&\leq
c_3(1-\gamma)^3\varepsilon^2,
\label{eq:KL-single-transition-bound-infinite}
\end{align}
where $c_3>0$ is some universal constant. 
\end{lemma}

Combining Lemma~\ref{lem:transition-KL-bound-infinite} with~\eqref{eq:single-sample-KL-infinite} reveals that
\begin{align}
\mathsf{KL}\big(
\mu^{\mathsf b,\theta}
\,\big\|\,
\mu^{\mathsf b,\widetilde{\theta}}
\big)
&\leq
\frac{c_3(1-\gamma)^3\varepsilon^2}
{2C(S-1)}
\sum_{s=1}^{S-1}
\mathds{1}\{\theta_s\neq\widetilde{\theta}_s\}
\sum_{a\in\{0,1\}}1
\notag\\
&\leq
\frac{c_3(1-\gamma)^3\varepsilon^2}{C}
\leq \frac{2c_3(1-\gamma)^3\varepsilon^2}{\Cstar}, 
\label{eq:KL-bound-infinite}
\end{align}
where the last inequality follows since $C_\rho^\star<2C$ (cf.~\eqref{eq:expression-Cstar-LB-inf}).

\paragraph{Step 4: putting all pieces together.}
Consequently, if
$N
\leq
\frac{c_2C_\rho^\star S}
{(1-\gamma)^3\varepsilon^2}$
for some sufficiently small constant $c_2>0$, then
substituting~\eqref{eq:KL-bound-infinite} into
\eqref{eq:Fano-LB-infinite} and using~\eqref{eq:property-Theta}
imply, for $S>12$, that
\[
p_{\mathrm e}>\frac14. 
\]
On the other hand, if a policy estimator satisfying
\eqref{eq:sub-optimality-gap-proof-34-lb-offline} existed for every
$\theta\in\Theta$, then Step~1 would produce an estimator
$\widehat{\theta}$ satisfying
\[
\max_{\theta\in\Theta}
\mathbb{P}_{\theta}(\widehat{\theta}\neq\theta)
\leq\frac14,
\]
thus resulting in contradiction. This proves the claimed
minimax lower bound.

\section{Extension: finite-horizon nonstationary MDPs}
\label{sec:finite-horizon-offline}

We now provide a brief discussion of how the pessimism principle can be applied to finite-horizon nonstationary MDPs, covering both model-based and model-free approaches. Since the key ideas largely parallel those developed for the infinite-horizon setting, our treatment here is intentionally concise. We refer the interested reader to \citet{li2024settling,shi2022pessimistic} for more comprehensive treatments and detailed analyses of the model-based and model-free approaches, respectively.

\subsection{Problem setting}

We begin by describing the problem setup.

\paragraph{Offline dataset.}
Suppose that we are given access to a batch dataset $\mathcal{D}$, containing a collection of $K$ independent sample trajectories generated by a behavior policy $\pib=\{\pib_h\}_{1\leq h\leq H}$ starting from an initial state distribution $\rhob$. 
More specifically, the $k$-th sample trajectory ($1 \le k \le K$) is a sequence
\begin{align}
	\big( s_1^k, a_1^k, s_2^k, a_2^k, \ldots, s_H^k, a_H^k, s_{H+1}^k \big),
	\label{eq:finite-sequence-k}
\end{align}
generated by executing the behavior policy $\pib$ in the underlying MDP according to
\begin{align}\label{eq:finite-batch-size-def}
s_1^k \sim \rhob,
\quad a_h^k \sim \pib_h(\cdot\mymid s_h^k), 
\quad\text{and}\quad 
	s_{h+1}^k \sim P_h(\cdot\mymid s_h^k, a_h^k)
\end{align}
for every $1\leq h\leq H$. 
In addition to the dataset $\mathcal{D}$, the learner is assumed to have full access to the reward function.

For notational convenience, we introduce the shorthand notation for the occupancy measures induced by the behavior policy $\pib$: $\forall (s,a,h)\in \cS\times \cA \times [H]$,
\begin{equation}
	\myrho_h(s) \coloneqq  d_h^{\pib}(s; \rhob) \quad \text{and} \quad \myrho_h(s, a) \coloneqq d_h^{\pib}(s,a; \rhob). 
	\label{eq:dhb-finite}
\end{equation}
Note that the initial state distribution $\rhob$ underlying the offline data set does not necessarily coincide with the test distribution $\rho$ used for performance evaluation.

\paragraph{Key parameter.}
Akin to Definition~\ref{assumption:concentrate-infinite-simple}, 
we introduce below the notion of single-policy concentrability tailored to finite-horizon MDPs.
%
%coefficient to characterize the distribution shift between the target distribution and the distribution induced by the behavior policy. 
 %
\begin{definition}[Single-policy concentrability for finite-horizon MDPs]
\label{assumption:concentrate-finite}
For any given initial state distribution $\rho\in \Delta(\cS)$, the single-policy concentrability coefficient underlying the batch dataset $\mathcal{D}$ is defined as
\begin{align}
	\Cstar \coloneqq \max_{(s, a, h) \in \cS \times \cA \times [H]}\frac{d_h^{\star}(s, a; \rho)}{\myrho_h(s, a)} ,
	\label{eq:concentrate-finite}
\end{align}
where the convention $0/0 = 0$ is adopted.
\end{definition}
Here, $\Cstar$ quantifies the distribution shift between the target distribution and the distribution induced by the behavior policy through the maximum density ratio between their corresponding occupancy distributions. As before, this condition does not require uniform coverage of the entire state-action space; it often suffices for the dataset to adequately cover the regions reachable under the optimal policy $\pi^\star$. 

\paragraph{Goal.} 
Given the offline dataset $\mathcal{D}$, the objective is to efficiently learn a policy $\widehat{\pi}$ that attains near-optimal statistical performance. More specifically, for a prescribed test distribution $\rho \in \Delta(\cS)$ and a target accuracy level $\varepsilon\in (0, H]$, we aim to design a learning algorithm that guarantees, with high probability,
\[
	V_1^{\star}(\rho) - V_1^{\widehat{\pi}}(\rho) \leq \varepsilon,
\]
while requiring as few samples from the dataset as possible.

\subsection{Model-based algorithm}

As in the discounted infinite-horizon setting, minimax-optimal sample complexity in the offline data setting \eqref{eq:finite-batch-size-def} is achieved by model-based methods. In the sequel, we briefly describe a simple pessimistic variant of the model-based algorithm.

%Suppose for the moment that we have access to a dataset $\mathcal{D}_0$ containing $N$ sample transitions $\{(s_i, a_i, h_i, s_i')\}_{i=1}^{N}$, where $(s_i, a_i, h_i, s_i')$ denotes the transition from state $s_i$ at step $h_i$ to state $s_i'$ in the next step when action $a_i$ is taken.  

\paragraph{The \vilcb algorithm for finite-horizon MDPs.} 
The algorithm first constructs an empirical transition kernel $\widehat{P}$ from the offline dataset $\mathcal{D}$. 
Based on this estimated model, the \vilcb algorithm maintains estimates of the value functions $\{\widehat{V}_h^{\lcb}\}_{1\leq h\leq H}$ and Q-functions $\{\widehat{Q}_h^{\lcb}\}_{1\leq h\leq H}$, and proceeds backward from $h=H$ to $h=1$ in the spirit of classical dynamic programming,  with the terminal value  $\widehat{V}_{H+1}^{\lcb}=0$ \citep{xie2021policy,jin2021pessimism}. Specifically, the updates take the form:
\begin{subequations}
\begin{align}
	Q_h^{\lcb}(s, a) = \max\Big\{r_h(s, a) + \widehat{P}_{s,a,h} V^{\lcb}_{h+1}   - b_h(s, a; V^{\lcb}_{h+1}), 0\Big\} ,
	\label{eq:VI-LCB-finite}
\end{align}
where $\widehat{P}_{s,a,h}$ denotes the empirical estimate of $P_{s,a,h}$ (cf.~\eqref{eq:transition_vector}), 
\begin{equation}
	V^{\lcb}_{h+1}(s) = \max_{a\in \cA} Q^{\lcb}_{h+1}(s, a),
	\label{eq:defn-widehat-V-h+1-finite}
\end{equation}
\end{subequations}
and $b_h(s, a; V) \geq 0$ denotes some penalty function that is decreasing in $N_h(s,a)$, the number of times the tuple $(s,a,h)$ is sampled. The output policy $\widehat{\pi}$ is then chosen greedily with respect to the estimated Q-functions:
\begin{align}
	\forall(s,h)\in \cS\times [H]: \qquad
	\widehat{\pi}_h(s) \in \arg\max_a Q^{\lcb}_h(s, a).
\end{align}
In essence, the \vilcb algorithm performs value iteration on the empirical model $\widehat{P}$, while incorporating lower-confidence penalty terms ${b_h(s,a;V)}$ into each Bellman update to enforce pessimism.

To more effectively capture the variance structure of the underlying MDP, it is, as before, advantageous to adopt a Bernstein-style penalty. Concretely, for all $(s,a,h)\in \cS\times \cA\times [H]$, 
\begin{align*}
	b_h(s, a; V) = \widetilde{\Theta} \Bigg(\sqrt{\frac{\mathsf{Var}_{\widehat{P}_{s,a,h}}(V)}{N_h(s, a)}} +   \frac{H}{N_h(s, a)}  \Bigg).
	%\label{def:bonus-Bernstein-finite}
\end{align*}
Here, $\mathsf{Var}_{\widehat{P}_{s,a,h}}(V)$ denotes the variance of the vector $V\in\mathbb{R}^S$ under the distribution $\widehat{P}_{s,a,h}$.

It is worth noting that, in order to establish provable sample optimality, \citet{li2024settling} further developed a refined variant of the above model-based offline RL method, based on a two-fold sample splitting procedure that helps decouple complicated statistical dependency arising along each sample trajectory. We omit these technical details for brevity and refer interested readers to \citet{li2024settling} for in-depth discussion.

%Note that we choose $\widehat{P}$ as opposed to $P$ (i.e., $\mathsf{Var}_{P_{h,s,a}} \big(\widehat{V}_{h+1}\big)$) in the variance term, mainly because we have no access to the true  transition kernel $P$. 

% \begin{algorithm}[t]
% \DontPrintSemicolon
% 	\textbf{input:} dataset $\mathcal{D}_0$; reward function $r$; target success probability $1-\delta$. \\
% 	% $N_h(s,a)$:  number of sample transitions in $\mathcal{D}_0$ for each $(s,a,h)\in \cS\times \cA\times [H]$; 
% 	\textbf{initialization:} $\widehat{V}_{H+1}=0$. \\

%    \For{$h=H,\cdots,1$}
% 	{
% 		compute the empirical transition kernel $\widehat{P}_h$. \\
% 		\For{$s\in \cS, a\in \cA$}{
% 			compute the penalty term $b_h(s,a)$ according to \eqref{def:bonus-Bernstein-finite}. \\
% 			set $\widehat{Q}_h(s, a) = \max\big\{r_h(s, a) + \widehat{P}_{h, s, a} \widehat{V}_{h+1} - b_h(s, a), 0\big\}$. \\ 
% 		}
% 		%\blue{O.} \\ 
% 		\For{$s\in \cS$}{
% 			set $\widehat{V}_h(s) = \max_a \widehat{Q}_h(s, a)$ and $\widehat{\pi}_h(s) \in \arg\max_{a} \widehat{Q}_h(s,a)$.
% 		}
% 	}

% 	\textbf{output:} $\widehat{\pi}=\{\widehat{\pi}_h\}_{1\leq h\leq H}$. 
% 	\caption{Offline value iteration with LCB (VI-LCB) for finite-horizon MDPs.}
%  \label{alg:vi-lcb-finite}
% \end{algorithm}

\paragraph{Theoretical guarantees.} 
The following theorem establishes the sample complexity guarantee for a variant of the aforementioned \vilcb algorithm for finite-horizon MDPs, paralleling the guarantee in Theorem~\ref{thm:infinite-offline} for discounted infinite-horizon MDPs. 

\begin{theorem}[\citet{li2024settling}] \label{thm:finite-VI-LCB}
Consider any $\varepsilon \in (0, H]$ and any $0<\delta <1$. Let $\rho\in \Delta(\cS)$ be any given state distribution. Then, there exists a variant of the above \vilcb algorithm such that, with probability at least $1-\delta$, the policy $\widehat{\pi}$ returned by the algorithm satisfies
\begin{align}
%\sum_s d_h^{\star}(s) \big(V^{\star}_h(s) - V^{\pi}_h(s)\big) \le 8\cb\sqrt{\frac{H^4SC^{\star}\log\frac{HSA}{\delta}}{T}}
	V^{\star}_1(\rho) - V^{\widehat{\pi}}_1(\rho) \leq \varepsilon 
\end{align}
as long as the total number of sample trajectories in $\mathcal{D}$ exceeds 
\begin{align}\label{eq:K-complexity-finite}
	K\geq \frac{c_{\mathsf{k}}H^3S\Cstar\log\frac{KH}{\delta}}{\varepsilon^2} 
\end{align}
for some sufficiently large numerical constant $c_{\mathsf{k}}>0$. Here,  $\Cstar$ is introduced in Definition~\ref{assumption:concentrate-finite}.
%Here, $c_{\mathsf{k}}>0$ is some sufficiently large constant.
\end{theorem}

Given that each sample trajectory has length $H$, the total number of samples (namely, $KH$) required by model-based offline RL algorithms to achieve $\varepsilon$-accuracy can be as low as
\begin{equation}
  \widetilde{O}\bigg( \frac{ H^4S\Cstar  }{\varepsilon^2} \bigg). 
	\label{eq:sample-size-Bernstein}
\end{equation}
This sample complexity can be proven to be minimax-optimal up to logarithmic factors \citep{li2024settling}. Importantly, the bound in \eqref{eq:K-complexity-finite} is valid for any $\varepsilon \in (0,H]$, indicating that the algorithm achieves sample-optimal performance without imposing any burn-in sample size requirement.

\subsection{Model-free algorithm}

Next, we briefly describe how the \LCBQ algorithm introduced in Section~\ref{sec:model-free-offline-RL} can be adapted to the finite-horizon setting.

\paragraph{The \LCBQ algorithm for finite-horizon MDPs.} For each trajectory $k=1,\ldots, K$, the \LCBQ algorithm updates a single entry in the Q-estimate at each step $h$ using the observed transition tuple $(s_h^k, a_h^k, s_{h+1}^k)$. In particular, it employs a modified Q-learning update equipped with a pessimistic penalty term as follows: 
\begin{align*}
   % \label{equ:lcb-q-update}
Q_h^{\lcb}(s_h^k, a_h^k) \leftarrow  (1-\eta_n )Q_h^{\lcb}(s_h^k, a_h^k) + \eta_n \Big\{ r_h(s_h^k,a_h^k) +  V^{\lcb}_{h+1}(s_{h+1}^k)  - b_n \Big\}. 
\end{align*}
Here, $\eta_n$ denotes the learning rate, depending on the number of times $n $ that the state-action pair $(s_h^k,a_h^k)$ has been visited at step $h$, with a common choice being $\eta_n = \frac{H+1}{H+n}$. The penalty term $b_n>0$ is chosen to measure the uncertainty of the corresponding Q-estimate and implement pessimism in the face of uncertainty; for instance, a Hoeffding-style penalty term takes the form
\begin{equation} \label{eq:lcb-q-penalty}
b_n = \widetilde{\Theta} \Bigg( \sqrt{\frac{H^3}{n}} \Bigg) .
\end{equation}
The algorithm then outputs a policy estimate $\widehat{\pi}$ obtained by acting greedily with respect to the Q-function estimate $Q^{\lcb}$.

The resulting algorithm is a {\em single-pass} procedure, requiring only one sweep through the offline dataset. It enjoys the following statistical guarantees.

% \begin{algorithm}[t]
% \DontPrintSemicolon
% 	\textbf{input:} dataset $\mathcal{D}_0$; reward function $r$; target success probability $1-\delta$. \\
% 	% $N_h(s,a)$:  number of sample transitions in $\mathcal{D}_0$ for each $(s,a,h)\in \cS\times \cA\times [H]$; 
% 	\textbf{initialization:} $\widehat{V}_{H+1}=0$. \\
% \yc{to update}
% \For{$k = 1, \ldots, K$}{
%    \For{$h=H,\cdots,1$}
% 	{
% 		sample one . \\
% 		\For{$s\in \cS, a\in \cA$}{
% 			compute the penalty term $b_h(s,a)$ according to \eqref{def:bonus-Bernstein-finite}. \\
% 			set $\widehat{Q}_h(s, a) = \max\big\{r_h(s, a) + \widehat{P}_{h, s, a} \widehat{V}_{h+1} - b_h(s, a), 0\big\}$. \\ 
% 		}
% 		%\blue{O.} \\ 
% 		\For{$s\in \cS$}{
% 			set $\widehat{V}_h(s) = \max_a \widehat{Q}_h(s, a)$ and $\widehat{\pi}_h(s) \in \arg\max_{a} \widehat{Q}_h(s,a)$.
% 		}
% 	}
% }
% 	\textbf{output:} $\widehat{\pi}=\{\widehat{\pi}_h\}_{1\leq h\leq H}$. 
% 	\caption{Pessimistic Q-learning for finite-horizon MDPs.}
%  \label{alg:lcbq}
% \end{algorithm}

\begin{theorem}[\citet{shi2022pessimistic}]\label{thm:lcbq}
  Consider any $\delta \in (0,1)$ and any $\varepsilon \in (0,H]$. Let $\rho\in \Delta(\cS)$ be any given state distribution.  
   With probability at least $1-\delta$, the policy $\widehat{\pi}$ returned by the \LCBQ algorithm described above with the Hoeffding-style penalty~\eqref{eq:lcb-q-penalty} obeys
   \begin{align}
%\sum_s d_h^{\star}(s) \big(V^{\star}_h(s) - V^{\pi}_h(s)\big) \le 8\cb\sqrt{\frac{H^4SC^{\star}\log\frac{HSA}{\delta}}{T}}
	V^{\star}_1(\rho) - V^{\widehat{\pi}}_1(\rho) \leq \varepsilon ,
\end{align}
with the proviso that the total number of sample trajectories exceeds
\begin{align}\label{equ:lcb-result}
K  &\geq c_\mathsf{a} \frac{H^5 S \Cstar \log^3\frac{SAK}{\delta}}{\varepsilon^2}
\end{align}
for some sufficiently large numerical constant $c_\mathsf{a}>0$. Here, $\Cstar$ is introduced in Definition~\ref{assumption:concentrate-finite}. 
\end{theorem}

Compared with its model-based counterpart, the \LCBQ algorithm equipped with a Hoeffding-style penalty may incur an additional factor of $H^2$ in sample complexity, rendering it statistically suboptimal. Comprehensive discussions of this pessimistic model-free approach can be found in \citet{shi2022pessimistic}.

\section{Appendix A: proof of auxiliary lemmas in Section~\ref{sec:VI-LCB-infinite-horizon}}

\subsection{Proof of Lemma~\ref{lem:monotone-contraction}}
\label{sec:proof-lem:monotone-contraction}

For notational convenience, for any $Q, Q_1, Q_2 \in \mathbb{R}^{SA}$, define their induced value functions by
\begin{align*}
	V(s) \coloneqq \max_a Q(s,a), \quad 
	V_1(s) \coloneqq \max_a Q_1(s,a), \quad V_2(s) \coloneqq \max_a Q_2(s,a) 
\end{align*}
for all $s\in \cS$. 
Unless otherwise noted, we assume throughout this subsection that
\[
Q(s,a), Q_1(s,a), Q_2(s,a) \in \Big[0, \frac{1}{1-\gamma}\Big] \qquad \text{for all }(s,a)\in \cS\times \cA.
\]

In the sequel, we first establish two useful properties of the operator $\Tpess$ (cf.~\eqref{eq:empirical-Bellman-infinite}). To this end, we find it helpful to define its unclipped counterpart
$\tildeTpess$ such that: for any $(s,a)$, 
\[
\tildeTpess(Q)(s,a)
\coloneqq r(s,a)-b(s,a;V)+\gamma \widehat P_{s,a}V ,
\]
then according to the definition~\eqref{eq:empirical-Bellman-infinite}, 
\[
\Tpess(Q)(s,a)
=
\max\big\{\tildeTpess(Q)(s,a),0\big\}.
\]

\paragraph{Proof of monotonicity.} We begin by establishing the monotonicity of the unclipped operator
$\tildeTpess$, namely, 
\begin{equation}
	Q\le \widetilde Q
	\qquad\Longrightarrow\qquad
	\tildeTpess(Q)\le \tildeTpess(\widetilde Q).
	\label{eq:proof-monotone-tilde-Tpess}
\end{equation}
Since the mapping $x\mapsto \max\{x,0\}$ is nondecreasing, this immediately
implies that $\Tpess$ is also monotone.

To prove \eqref{eq:proof-monotone-tilde-Tpess}, consider any given $(s,a)\in\cS\times\cA$, and regard
$\tildeTpess(Q)(s,a)$ as a function of $V$. It suffices to show that this function is coordinate-wise nondecreasing in $V$. Recall the definition of $b(s,a;V)$ in \eqref{def:bonus-Bernstein-infinite}, namely,
\[
b(s, a; V) \defn \min \bigg\{  \max \{ b_1, b_2 \} ,\frac{1}{1-\gamma}  \bigg\} +\frac{5}{N} 
\]
where
\[
b_1 \coloneqq \sqrt{\frac{\cb\log\frac{N}{(1-\gamma)\delta}}{N_{s,a} }\mathsf{Var}_{\widehat{P}_{ s, a}} (V )} \qquad \text{and} \qquad 
b_2 \coloneqq \frac{2\cb \log\frac{N}{(1-\gamma)\delta}}{(1-\gamma)N_{s,a} } .
\]
For any given $(s,a)$, we look at the following three cases separately.  
	\begin{itemize}
		\item {\em Case 1:} $\max \{b_1,b_2\} > \frac{1}{1-\gamma}$. In this case, the penalty term simplifies to
		\[
		b(s,a; V) = \frac{1}{1-\gamma} + \frac{5}{N},
		\]
		which is independent of $V$. Hence, for any $s'\in\cS$,
		\[
		\frac{\partial \tildeTpess(Q)(s,a)}{\partial V(s')}
		=
		\gamma \widehat P(s'\mid s,a)\ge 0 .
		\]
		
		\item {\em Case 2:}   $b_1<b_2<\frac{1}{1-\gamma}$. In this case, we have
		\[
b(s,a;V)
=
\frac{2\cb\log\frac{N}{(1-\gamma)\delta}}
{(1-\gamma)N_{s,a}}
+\frac{5}{N},
		\]
		which is again independent of $V$. Thus the same calculation yields the same result as Case 1.

		\item {\em Case 3:}   $b_2< b_1<\frac{1}{1-\gamma}$. As an immediate consequence of $b_2 < b_1$, one has
		\begin{equation}
			\mathsf{Var}_{\widehat P_{s,a}}(V)
			>
			\frac{4\cb\log\frac{N}{(1-\gamma)\delta}}
			{(1-\gamma)^2N_{s,a} } .
			\label{eq:proof-var-lower-bound}
		\end{equation}
		In this case, the penalty term depends on $V$ as follows: 
		\[
		b(s,a; V) = \sqrt{\frac{\cb\log\frac{N}{(1-\gamma)\delta}}{N_{s,a} }\mathsf{Var}_{\widehat{P}_{s,a}}(V)} + \frac{5}{N}. 
		\]
		For any $s'\in\cS$, differentiating the penalty term gives
		\begin{align*}
			\frac{\partial b(s,a;V)}{\partial V(s')}
			&=
			\sqrt{
				\frac{\cb \log\frac{N}{(1-\gamma)\delta}}{N_{s,a} }
			}
			\frac{
				\widehat P(s'\mid s,a)
				\big(V(s')-\widehat P_{s,a}V\big)
			}{
				\sqrt{\mathsf{Var}_{\widehat P_{s,a}}(V)} 
			} \\
			&\overset{\text{(i)}}{\leq}
			\frac{1-\gamma}{2}
			\widehat P(s'\mid s,a)
			\max\big\{ V(s')-\widehat P_{s,a}V, 0\big\}  \\
			&\overset{\text{(ii)}}{\leq}
			\frac{1}{2}\widehat P(s'\mid s,a)
			\overset{\text{(iii)}}{\leq}
			\gamma \widehat P(s'\mid s,a),
		\end{align*}
        where (i) follows from \eqref{eq:proof-var-lower-bound}, (ii) holds since $V(s')-\widehat P_{s,a}V \leq \frac{1}{1-\gamma}$, 
		and (iii) results from $\gamma\ge 1/2$. Consequently,
		\[
		\frac{\partial \tildeTpess(Q)(s,a)}{\partial V(s')}
		=
		\gamma \widehat P(s'\mid s,a)
		-
		\frac{\partial b(s,a;V)}{\partial V(s')}
		\ge 0 .
		\]
\end{itemize}
The only points not covered by the preceding three cases lie on the
switching boundaries
\[
b_1=b_2
\qquad\text{or}\qquad
\max\{b_1,b_2\}=\frac{1}{1-\gamma}.
\]
At these points, the penalty function may not be differentiable.
Nevertheless, $\tildeTpess(Q)(s,a)$ is continuous and piecewise
differentiable as a function of $V$. Along any coordinate-wise line
segment, its derivative is nonnegative at every differentiability
point, by the three cases above. It follows that the function is
nondecreasing along the entire line segment, including across the
switching boundaries. Therefore, $\tildeTpess(Q)(s,a)$ is
coordinate-wise nondecreasing in $V$.

Finally, $Q\le \widetilde Q$ implies 
$V(s) \leq \widetilde{V}(s)$ 
for every $s\in\cS$, which proves \eqref{eq:proof-monotone-tilde-Tpess}.

\paragraph{Proof of $\gamma$-contraction.} We next prove that $\Tpess$ is a $\gamma$-contraction. Consider any $Q_1, Q_2 \in \mathbb{R}^{SA}$,  and set
$c=\|Q_1-Q_2\|_{\infty}$. Then we have
\begin{align*}
	\tildeTpess(Q_1)-\tildeTpess(Q_2)
	& \overset{\text{(i)}}{\leq}
	\tildeTpess(Q_2+c1_{SA})-\tildeTpess(Q_2) \\
	& \overset{\text{(ii)}}{=} \gamma c \widehat P 1_{S} = \gamma\|Q_1-Q_2\|_{\infty}1_{SA}.
\end{align*}
Here, (i) follows from $Q_1 \le Q_2+c 1_{SA}$ and the monotonicity of $\tildeTpess$;  (ii) holds since the penalty $b(s,a;V)$ (cf.~\eqref{def:bonus-Bernstein-infinite}) is invariant under constant shifts of the value function $V$, as $\mathsf{Var}_{\widehat{P}_{s, a}}(V) = \mathsf{Var}_{\widehat{P}_{s, a}}(V + c\cdot 1_S)$. Clearly, this inequality continues to hold if we swap $Q_1$ and $Q_2$. As a result, we arrive at
\[
\big\|
\tildeTpess(Q_1)-\tildeTpess(Q_2)
\big\|_{\infty}
\le
\gamma \|Q_1-Q_2\|_{\infty}.
\]
Since the mapping $x\mapsto \max\{x,0\}$ is $1$-Lipschitz, it follows immediately that
\begin{equation}
	\big\|\Tpess(Q_1)-\Tpess(Q_2)\big\|_{\infty}
	\le
	\gamma\|Q_1-Q_2\|_{\infty}.
	\label{eq:proof-Tpess-contraction}
\end{equation}

\paragraph{Fixed point of $\Tpess$.} We now turn to the existence and uniqueness of the fixed point of $\Tpess$. For any
$0\le Q\le \frac{1}{1-\gamma}1_{SA}$, the definition \eqref{eq:empirical-Bellman-infinite} of $\Tpess$ ensures that $$0\le \Tpess(Q)\le \frac{1}{1-\gamma}1_{SA}.$$ Define the following sequence recursively:
\[
Q^{(0)}=0 \qquad \text{and} \qquad Q^{(t+1)}= \Tpess(Q^{(t)}) \quad \text{for } t\geq 0.
\]
By the contraction property \eqref{eq:proof-Tpess-contraction}, the standard proof for the Banach fixed-point theorem (e.g., \citet[Theorem 1]{agarwal2001fixed}) guarantees that $Q^{(t)}$ converges to some point $Q^{(\infty)}$ as $t\rightarrow \infty$. Clearly, $Q^{(\infty)}$ is a fixed point of $\Tpess$ obeying $0\leq Q^{(\infty)} \leq \frac{1}{1-\gamma} 1_{SA}$.

Moreover, suppose that there exists another point $\widetilde{Q}$ obeying $\widetilde{Q}= \Tpess \big( \widetilde{Q} \big)$, 
which clearly satisfies $\widetilde{Q}\geq 0$. 
If $\|\widetilde{Q}\|_{\infty} > \frac{1}{1-\gamma}$, then 
\begin{align*}
	\big\| \widetilde{Q} \big\|_{\infty} &= \big\| \Tpess \big( \widetilde{Q} \big) \big\|_{\infty}
	\leq \| r \|_{\infty} + \gamma \| \widehat{P} \|_1 	\big\| \widetilde{Q} \big\|_{\infty} 
	\leq 1 + \gamma \big\| \widetilde{Q} \big\|_{\infty} \\
	& < (1-\gamma) \big\| \widetilde{Q} \big\|_{\infty} + \gamma \big\| \widetilde{Q} \big\|_{\infty} = \big\| \widetilde{Q} \big\|_{\infty} ,
\end{align*}
resulting in contradiction. Hence, every fixed point must satisfy $0 \leq \widetilde{Q} \leq \frac{1}{1-\gamma}1_{SA}$. 
The $\gamma$-contraction property \eqref{eq:proof-Tpess-contraction} implies that
\[
\big\| \widetilde{Q} - Q^{(\infty)} \big\|_{\infty}
= \big\| \Tpess \big( \widetilde{Q} \big) - \Tpess \big( Q^{(\infty)} \big) \big\|_{\infty}
\leq \gamma \big\| \widetilde{Q} - Q^{(\infty)} \big\|_{\infty}. 
\]
Given that $\gamma < 1$, it follows immediately that $\widetilde{Q} = Q^{(\infty)}$, 
thereby confirming the uniqueness of the fixed point of $\Tpess$. From now on, we shall denote this unique fixed point by $\widehat{Q}_{\mathsf{pe}}^{\star}$.

\paragraph{Convergence properties.} 
We now establish the advertised convergence properties of the iterates. Since
$Q_0=0\le \widehat Q_{\mathsf{pe}}^{\star}$, the monotonicity of $\Tpess$ gives
\[
Q_1
=
\Tpess(Q_0)
\le
\Tpess(\widehat Q_{\mathsf{pe}}^{\star})
=
\widehat Q_{\mathsf{pe}}^{\star}.
\]
Repeat the same argument to yield
\[
Q_{\tau}\le \widehat Q_{\mathsf{pe}}^{\star}
\qquad\text{for all }\tau\ge 0.
\]
Furthermore, applying the contraction property \eqref{eq:proof-Tpess-contraction} recursively gives
\begin{align}
	\big\|Q_{\tau}-\widehat Q_{\mathsf{pe}}^{\star}\big\|_{\infty}
	&=
	\big\|
	\Tpess(Q_{\tau-1})-\Tpess(\widehat Q_{\mathsf{pe}}^{\star})
	\big\|_{\infty}  \le
	\gamma
	\big\|Q_{\tau-1}-\widehat Q_{\mathsf{pe}}^{\star}\big\|_{\infty} \nonumber\\
	&\le \cdots \le
	\gamma^{\tau}
	\big\|Q_0-\widehat Q_{\mathsf{pe}}^{\star}\big\|_{\infty} \le
	\frac{\gamma^{\tau}}{1-\gamma}, \label{eq:Q-convergence-proof-135}
\end{align}
where the last inequality follows from $Q_0=0$ and
$\|\widehat Q_{\mathsf{pe}}^{\star}\|_{\infty}\le 1/(1-\gamma)$. 

Finally, the remaining claim~\eqref{eq:taumax-Q-converge} follows immediately by taking the right-hand side of \eqref{eq:Q-convergence-proof-135} to be no larger than $1/N$. This completes the proof of this lemma.

\subsection{Proof of Lemma~\ref{lem:offline-prelim}} \label{proof:lem:offline-prelim}

With~\eqref{eq:Q-condition-Hoeffding} in place, it is straightforward to obtain
\begin{align*}
	\big\| \widehat{V} - \widehat{V}^{\star}_{\mathsf{pe}} \big\|_{\infty} 
	= \max_{s}\left| \max_a \widehat{Q}(s,a)- \max_a \widehat{Q}^{\star}_{\mathsf{pe}}(s,a)\right| 
	\le 
	\big\| \widehat{Q} - \widehat{Q}^{\star}_{\mathsf{pe}} \big\|_{\infty} \le \frac{1}{N},
\end{align*}
as claimed in~\eqref{eq:Vhar-Vstarhat-diff-1/T-inf}.

Next, we turn to the claim~\eqref{infinite-bpe-to-bV}. When $N_{s,a}=0$, the claim follows immediately from the convention
$b(s,a;V)=1/(1-\gamma)+5/N$. We may therefore assume below that
$N_{s,a}>0$. Given the proximity of $\widehat{V}$ and $\widehat{V}^{\star}_{\mathsf{pe}}$, 
we can apply Lemma~\ref{lem:variance-perturbation-var-P-V} to bound the difference of the corresponding variance terms as follows:
\begin{align}
	\big|\mathsf{Var}_{\widehat{P}_{s,a}}\big(\widehat{V}_{\mathsf{pe}}^{\star}\big) - \mathsf{Var}_{\widehat{P}_{s,a}}\big(\widehat{V}\big)\big| &
	\leq \frac{4}{1-\gamma}\big\| \widehat{V} - \widehat{V}^{\star}_{\mathsf{pe}} \big\|_{\infty}  \leq \frac{4}{(1-\gamma)N}. \label{infinite-Vpe-to-V}
\end{align}
In view of the definition of $b(s,a; V)$ in \eqref{def:bonus-Bernstein-infinite}, we know that
if at least one of the following variance terms is not too small, i.e., 
\begin{equation}
	\max\Big\{ \mathsf{Var}_{\widehat{P}_{s, a}}\big(\widehat{V}_{\mathsf{pe}}^{\star}\big) , \mathsf{Var}_{\widehat{P}_{s, a}}\big(\widehat{V}\big) \Big\} 
	\geq \frac{4\cb \log\frac{N }{(1-\gamma)\delta}}{(1-\gamma)^2 N_{s,a} }, 
	\label{eq:max-variance-term-LB-inf}
\end{equation}
then we have
\begin{align*}
	\big|b\big(s,a; \widehat{V}_{\mathsf{pe}}^{\star}\big)-b\big(s,a; \widehat{V}\big) \big| &\leq   \sqrt{\frac{\cb\log\frac{N  }{(1-\gamma)\delta}}{N_{s,a} }}\left|\sqrt{\mathsf{Var}_{\widehat{P}_{s, a}}\big(\widehat{V}_{\mathsf{pe}}^{\star}\big)} - \sqrt{\mathsf{Var}_{\widehat{P}_{s, a}}\big(\widehat{V}\big)}\right| \\
	&= 
	\sqrt{\frac{\cb\log\frac{N  }{(1-\gamma)\delta}}{N_{s,a} }}\frac{\big| \mathsf{Var}_{\widehat{P}_{s,a}}\big(\widehat{V}_{\mathsf{pe}}^{\star}\big)-\mathsf{Var}_{\widehat{P}_{s,a}}\big(\widehat{V}\big) \big| }{\sqrt{\mathsf{Var}_{\widehat{P}_{s,a} }\big(\widehat{V}_{\mathsf{pe}}^{\star}\big)}+\sqrt{\mathsf{Var}_{\widehat{P}_{s,a}}\big(\widehat{V}\big)}}  \\	
	&  \overset{\mathrm{(i)}}{\leq} \frac{1-\gamma}{2} \Big| \mathsf{Var}_{\widehat{P}_{s,a}}\big(\widehat{V}_{\mathsf{pe}}^{\star}\big)-\mathsf{Var}_{\widehat{P}_{s,a}}\big(\widehat{V}\big) \Big| 
	\overset{\mathrm{(ii)}}{\leq}  
	\frac{2}{N} ,
\end{align*}
where (i) results from \eqref{eq:max-variance-term-LB-inf}, and (ii) holds due to \eqref{infinite-Vpe-to-V}. 
Additionally, if condition~\eqref{eq:max-variance-term-LB-inf} is not satisfied, then one clearly has
$b\big(s,a; \widehat{V}_{\mathsf{pe}}^{\star}\big) = b\big(s,a; \widehat{V}\big)$ according to the definition~\eqref{def:bonus-Bernstein-infinite}. In summary, in all cases the advertised bound~\eqref{infinite-bpe-to-bV} holds true.

\subsection{Proof of Lemma~\ref{lem:sample-split-infinite}}
\label{proof:lem:sample-split-infinite}

For notational convenience, define
\[
L\coloneqq \log\frac{SN}{\delta}.
\]
We divide the proof into two cases.
\begin{itemize}
\item 
First, consider the case where 
$N\myrho(s,a)<8L$. In this case,
\[
\frac{N\myrho(s,a)}{12}
<
\frac{2}{3}L,
\]
and hence~\eqref{eq:samples-infinite-LB} holds deterministically.

\item 
It remains to consider the collection of state-action pairs: 
\begin{align}
\mathcal{N}_{\mathsf{large}}
\coloneqq
\left\{
(s,a)\in\cS\times\cA:
N\myrho(s,a)\geq 8L
\right\}.
\end{align}
For any fixed $(s,a)\in\mathcal{N}_{\mathsf{large}}$,
the count $N_{s,a}$ is a binomial random variable with mean
$N\myrho(s,a)$. Then invoking the Chernoff-type bound in Lemma~\ref{lem:tail-bound-sum-Bernoulli} yields
\begin{align}
\mathbb{P}\left(
N_{s,a}<\frac{N\myrho(s,a)}{2}
\right)
&\leq
\exp\left(-\frac{N\myrho(s,a)}{8}\right)
\leq
\exp(-L)
=
\frac{\delta}{SN}.
\label{eq:count-lower-tail-offline}
\end{align}
Moreover, since $\sum_{s,a}\myrho(s,a)=1$, the cardinality of
$\mathcal{N}_{\mathsf{large}}$ satisfies
\begin{align}
\left|\mathcal{N}_{\mathsf{large}}\right|
\cdot \frac{8L}{N}
\leq
\sum_{(s,a)\,\in\,\mathcal{N}_{\mathsf{large}}}\myrho(s,a)
\leq 1,
\end{align}
and as a result, 
\[
\left|\mathcal{N}_{\mathsf{large}}\right|
\leq \frac{N}{8L}\leq N.
\]
Applying a union bound to \eqref{eq:count-lower-tail-offline}, we conclude
that, with probability at least $1-\delta$,
\[
N_{s,a}
\geq
\frac{N\myrho(s,a)}{2}
\]
simultaneously for all
$(s,a)\in\mathcal{N}_{\mathsf{large}}$.

\end{itemize}

\noindent 
Combining the two cases establishes~\eqref{eq:samples-infinite-LB}.

\subsection{Proof of Lemma~\ref{lem:Bernstein-infinite}}\label{proof:lem:Bernstein-infinite}
If $N_{s, a} = 0$, then the desired inequalities hold trivially. It therefore suffices to focus on the case where $N_{s, a} > 0$.
We begin by recording a key Bernstein-style result, whose proof is deferred to Section~\ref{sec:lem:Bernstein-infinite-proof}. 
\begin{lemma} 
\label{lem:Bernstein-infinite-proof}
	Consider any $\delta\in(0,1/8)$, and any fixed pair $(s,a)\in \cS\times \cA$ with $N_{s,a}>0$. Let $V\in \mathbb{R}^S$ be any vector independent of $\widehat{P}_{s,a}$ satisfying $0\leq V\leq \frac{1}{1-\gamma}1_S$. 
	With probability at least $1-8\delta$, one has
	\begin{subequations}
		\begin{align}
			\big|\big(\widehat P_{s,a}-P_{s,a}\big)V\big|
&\leq
\sqrt{
\frac{4\mathsf{Var}_{\widehat P_{s,a}}(V)
\log\frac{1}{\delta}}
{N_{s,a}}
}
+
\frac{5\log\frac{1}{\delta}}
{(1-\gamma)N_{s,a}},
			\label{eq:bonus-Bernstein-infinite-proof} \\
			\mathsf{Var}_{\widehat{P}_{s, a}} ( V ) &\le 2\mathsf{Var}_{P_{s, a}}\big(V\big) + \frac{5\log\frac{1}{\delta }}{3(1-\gamma)^2 N_{s, a}}.
			\label{eq:empirical-var-infinite-proof}
		\end{align}
	\end{subequations}
\end{lemma}

With this concentration result in place, we now proceed to prove
Lemma~\ref{lem:Bernstein-infinite}.  Our argument employs the
leave-one-out decoupling technique --- similar to that introduced in
Section~\ref{sec:sharp-analysis-thm:sample-compl-MB-generative-inf}
 --- and proceeds in several steps.

\paragraph{Step 1: constructing auxiliary state-absorbing MDPs.}  
For each state $s \in \cS$ and each scalar $u\geq 0$, we construct an auxiliary MDP $\widehat{\mathcal{M}}^{s, u}$ that coincides with the empirical MDP $\widehat{\mathcal{M}}=(\cS,\cA,\widehat{P},r,\gamma)$ except in state $s$. 
More precisely, the transition kernel of $\widehat{\mathcal{M}}^{s, u}$, denoted by $P^{s,u}$, is defined as
\begin{align*}
	P^{s,u}(\,\widetilde{s} \mymid s, a) &= \mathds{1}(\widetilde{s} = s) \qquad\quad &&\text{for all }(\,\widetilde{s},a) \in \cS\times \cA, \\
	P^{s,u}(\cdot\mymid s^{\prime}, a) &= \widehat{P}(\cdot\mymid s^{\prime}, a)
	\qquad &&\text{for all }(s^{\prime}, a) \in \cS \times \cA \text{ and }s^{\prime}\neq s; 
\end{align*}
and the reward function of $\widehat{\mathcal{M}}^{s,u}$, denoted by $r^{s,u}$, is defined as
\begin{align*}
	r^{s,u}(s, a) &= u \quad\quad\quad~&&\text{for all }a \in \cA, \\
	r^{s,u}(s^{\prime}, a) &= r(s^{\prime}, a) \qquad&&\text{for all }(s^{\prime}, a) \in \cS \times \cA \text{ and }s^{\prime}\neq s.
\end{align*}
Thus, $\widehat{\mathcal M}^{s,u}$ makes state $s$ absorbing and replaces its
reward by $u$, thereby removing any dependence on the empirical
transition vectors $\{\widehat P_{s,a}\}_{a\in\cA}$.

Further, we define the pessimistic Bellman operator $\Tpess^{s,u}$ associated with the auxiliary MDP $\widehat{\mathcal{M}}^{s,u}$ such that, for every $(s,a)\in \cS\times \cA$, 
\begin{align}
	\Tpess^{s,u} (Q)(s, a) \coloneqq \max\Big\{r^{s,u}(s, a) + \gamma P^{s,u}_{s,a} V - b^{s,u}(s, a; V) 
	,\, 0\Big\},
	\label{eq:empirical-Bellman-infinite-Lemma8}
\end{align}
where the penalty term is chosen as
\begin{align*}
	b^{s,u}(s, a; V) \defn \min \Bigg\{  \max \Bigg\{ & \sqrt{\frac{\cb\log\frac{N}{(1-\gamma)\delta}}{N_{s, a}}\mathsf{Var}_{P^{s,u} (\cdot \mymid s, a)} (V )} , \arxiv{ \\ 
	& \qquad} \,\frac{2\cb \log\frac{N}{(1-\gamma)\delta}}{(1-\gamma)N_{s, a}} \Bigg\} ,\, \frac{1}{1-\gamma}  \Bigg\} +\frac{5}{N}.
\end{align*}

\paragraph{Step 2: connecting empirical and auxiliary MDPs.} 
Set
\begin{align}
	u^{\star} = (1-\gamma) \widehat{V}_{\mathsf{pe}}^{\star}(s) + \min \Bigg\{ \frac{2\cb \log\frac{N}{(1-\gamma)\delta}}{(1-\gamma)\max_a N_{s,a}}, \frac{1}{1-\gamma} \Bigg\} 
	+\frac{5}{N}.
    \label{eq:choice-ustar-offline-mb-proof}
\end{align}
We claim that $\Tpess^{s,u^{\star}}$ has a fixed point
$\widehat{Q}_{s,u^{\star}}^{\star}$ whose induced value function is
$\widehat{V}_{\mathsf{pe}}^{\star}$. To establish this claim, we first verify the following two properties.
%
%{\setlength\leftmargini {1.1em} 
	\begin{itemize}
		\item Consider any $a\in \cA$. Since $P^{s,u}(\cdot \mymid s,a)$ has only one nonzero entry, which equals $1$, we have
		$ \mathsf{Var}_{P^{s,u} (\cdot \mymid s, a)} (V ) = 0$ for every $V$ and every $u$. Therefore,
		\begin{align}
			b^{s,u}(s, a; V) = \min \Bigg\{ \frac{2\cb \log\frac{N }{(1-\gamma)\delta}}{(1-\gamma)N_{s, a}} ,\, \frac{1}{1-\gamma}  \Bigg\} +\frac{5}{N} .
			\label{eq:defn-bsu-sa-inf}
		\end{align}
		Consequently, it follows that, in state $s$, 
		\begin{align}
			&\max_a\Big\{ r^{s,u^{\star}}(s,a)-b^{s,u^{\star}}\big(s,a; \widehat{V}_{\mathsf{pe}}^{\star}\big)+\gamma\big\langle P^{s,u^{\star}}(\cdot\mymid s,a),\,\,\widehat{V}_{\mathsf{pe}}^{\star}\big\rangle\Big\}  \nonumber \\
			&\qquad =\max_a\Big\{u^{\star}-b^{s,u^{\star}}\big(s,a;\widehat{V}_{\mathsf{pe}}^{\star}\big)+\gamma\widehat{V}_{\mathsf{pe}}^{\star}(s)\Big\} \notag\\
			&\qquad=u^{\star}-\min_a b^{s,u^{\star}}\big(s,a;\widehat{V}_{\mathsf{pe}}^{\star}\big)+\gamma\widehat{V}_{\mathsf{pe}}^{\star}(s)\notag\\
			&\qquad=(1-\gamma)\widehat{V}_{\mathsf{pe}}^{\star}(s)+\gamma\widehat{V}_{\mathsf{pe}}^{\star}(s) =\widehat{V}_{\mathsf{pe}}^{\star}(s), 
			\label{eq:Bellman-su-first-s-inf}
		\end{align}
		where the third identity follows from the choice \eqref{eq:choice-ustar-offline-mb-proof} of $u^{\star}$ together with \eqref{eq:defn-bsu-sa-inf}.

		\item Next, consider any $s'\neq s$ and any $a\in \cA$. By construction of the auxiliary MDP, we have
		\begin{align}
			&\max\Big\{ r^{s,u^{\star}}(s',a)-b^{s,u^{\star}}\big(s',a; \widehat{V}_{\mathsf{pe}}^{\star}\big)+\gamma\big\langle P^{s,u^{\star}}(\cdot\mymid s',a),\,\,\widehat{V}_{\mathsf{pe}}^{\star}\big\rangle, \,0\Big\} \notag\\
			& \qquad =\max\Big\{ r(s',a)-b\big(s',a; \widehat{V}_{\mathsf{pe}}^{\star}\big)+\gamma\big\langle\widehat{P}(\cdot\mymid s',a),\,\widehat{V}_{\mathsf{pe}}^{\star}\big\rangle, \,0\Big\} \arxiv{ \nonumber \\
			& \qquad } = \widehat{Q}_{\mathsf{pe}}^{\star}(s',a),
			\label{eq:Bellman-su-non-s-a-inf}
		\end{align}
		where the last equality uses the fact that $\widehat{Q}_{\mathsf{pe}}^{\star}$ represents a fixed point of $\Tpess$ (cf.~\eqref{eq:empirical-Bellman-infinite}). 
	\end{itemize}
The preceding two identities suggest that
$\widehat{V}_{\mathsf{pe}}^\star$ satisfies the pessimistic Bellman
equations of the auxiliary MDP.  
We now formalize it by constructing the corresponding fixed point. 
Specifically, define the candidate Q-function $\widehat Q_{s,u^\star}^\star$ as follows: 
\begin{align}
\widehat Q_{s,u^\star}^\star(s',a)
\coloneqq
\begin{cases}
\displaystyle
\max\left\{
u^\star
-b^{s,u^\star}\big(s,a;\widehat V_{\mathsf{pe}}^\star\big)
+\gamma\widehat V_{\mathsf{pe}}^\star(s),
\,0
\right\},
& \text{if }s'=s,\\[2ex]
\widehat Q_{\mathsf{pe}}^\star(s',a),
& \text{if }s'\neq s.
\end{cases}
\label{eq:candidate-Q-su-star}
\end{align}
Denote by $\widehat V_{s,u^\star}^\star$ the value function induced by
this candidate Q-function. For every $s'\neq s$, its definition gives
\begin{subequations}
\label{eq:V-hat-s-ustar-equal-V-pestar}
\begin{align}
\widehat V_{s,u^\star}^\star(s')
=
\max_a\widehat Q_{\mathsf{pe}}^\star(s',a)
=
\widehat V_{\mathsf{pe}}^\star(s').
\end{align}
In state $s$, property~\eqref{eq:Bellman-su-first-s-inf} together with the
nonnegativity of $\widehat V_{\mathsf{pe}}^\star(s)$ implies that
\begin{align}
\widehat V_{s,u^\star}^\star(s)
&=
\max_a\widehat Q_{s,u^\star}^\star(s,a) \notag\\
&=
\max\left\{
\max_a
\left[
u^\star
-b^{s,u^\star}
\big(s,a;\widehat V_{\mathsf{pe}}^\star\big)
+\gamma\widehat V_{\mathsf{pe}}^\star(s)
\right],
\,0
\right\} \notag\\
&=
\widehat V_{\mathsf{pe}}^\star(s).
\end{align}
\end{subequations}
Consequently, we arrive at
$
\widehat V_{s,u^\star}^\star
=
\widehat V_{\mathsf{pe}}^\star$.

It remains to verify that the candidate Q-function is indeed a fixed point. In state $s$, the first
case in \eqref{eq:candidate-Q-su-star}, together with
$P^{s,u^\star}(s'\mymid s,a)=\mathds{1}\{s'=s\}$, gives
\[
\widehat Q_{s,u^\star}^\star(s,a)
=
\Tpess^{s,u^\star}
\big(\widehat Q_{s,u^\star}^\star\big)(s,a).
\]
For every state $s'\neq s$, it follows from~\eqref{eq:Bellman-su-non-s-a-inf} that
\[
\widehat Q_{s,u^\star}^\star(s',a)
=
\Tpess^{s,u^\star}
\big(\widehat Q_{s,u^\star}^\star\big)(s',a).
\]
Therefore, we have justified that
\[
\widehat Q_{s,u^\star}^\star
=
\Tpess^{s,u^\star}
\big(\widehat Q_{s,u^\star}^\star\big),
\]
which proves that $\widehat{Q}_{s,u^{\star}}^{\star}$ is a fixed point of the auxiliary
pessimistic Bellman operator and that its induced value function is precisely
$\widehat V_{\mathsf{pe}}^\star$.

\paragraph{Step 3: building a covering set.}
Fix any $(s,a)\in\cS\times\cA$ with $N_{s,a}>0$. We discretize the
possible range of $u^\star$ using a grid with resolution $1/N$, so that the concentration bound in Lemma~\ref{lem:Bernstein-infinite-proof} can be
applied uniformly over a finite collection of auxiliary MDPs.

Define
\[
u_{\mathsf{max}}
\coloneqq
\min\Bigg\{
\frac{2\cb\log\frac{N}{(1-\gamma)\delta}}
     {(1-\gamma)\max_{a'}N_{s,a'}},
\frac{1}{1-\gamma}
\Bigg\}
+\frac{5}{N}+1,
\]
and construct the covering set
\begin{subequations}
\begin{align}
\mathcal{U}_{\textsf{cover}}
\coloneqq
\left\{
\frac{i}{N}:
1\leq i\leq \left\lfloor Nu_{\mathsf{max}}\right\rfloor
\right\}.
\label{eq:defn-Ucover-infinite-Bernstein}
\end{align}
Given that $N\geq 5$ and $\gamma\geq 1/2$, its cardinality can be readily bounded by
\begin{align}
\big|\mathcal{U}_{\textsf{cover}}\big| 
\leq
Nu_{\mathsf{max}}
\leq
\frac{N}{1-\gamma}+5+N
\leq
\frac{2N}{1-\gamma}. 
\label{eq:defn-Ucover-infinite-Bernstein-cardinality}
\end{align}
\end{subequations}
We first verify that this grid contains a point within $1/N$ of $u^\star$.
By the definition \eqref{eq:choice-ustar-offline-mb-proof} of $u^\star$,
one has $u^\star\geq 5/N$. 
Moreover, since 
$(1-\gamma)\widehat V_{\mathsf{pe}}^\star(s)\leq 1$, it is readily seen that
\begin{align}
u^\star
&\leq
1+
\min\Bigg\{
\frac{2\cb\log\frac{N}{(1-\gamma)\delta}}
     {(1-\gamma)\max_{a'}N_{s,a'}},
\frac{1}{1-\gamma}
\Bigg\}
+\frac{5}{N}
=u_{\mathsf{max}}.
\end{align}
Therefore, there exists $u_0\in\mathcal U_{\textsf{cover}}$ such that
\begin{align}
|u_0-u^\star|\leq\frac{1}{N}.
\label{eq:u0-ustar-cover}
\end{align}

The choice of $u_{\mathsf{max}}$ also ensures that the fixed points
associated with the auxiliary MDPs remain well bounded. Indeed, in
state $s$, property~\eqref{eq:defn-bsu-sa-inf} implies that, for every
$u\in\mathcal{U}_{\textsf{cover}}$ and every action $a'$,
\[
    u-b^{s,u}(s,a';V) \leq u_{\mathsf{max}} -b^{s,u}(s,a';V) \leq 1.
\]
Since the immediate rewards in states $s'\neq s$ also lie in $[0,1]$,
the operator $\Tpess^{s,u}$ maps
\[
\left\{Q\in\mathbb{R}^{SA}:0\leq Q\leq
\frac{1}{1-\gamma}1_{SA}\right\}
\]
into itself. Given that $\Tpess^{s,u}$ is a $\gamma$-contraction, it has a
unique fixed point $\widehat Q_{s,u}^{\star}$ satisfying
\[
0\leq \widehat Q_{s,u}^{\star}
\leq \frac{1}{1-\gamma}1_{SA}.
\]
In what follows, we shall denote by $\widehat{V}^{\star}_{s,u}$ the value function induced by $\widehat{Q}^{\star}_{s,u}$.

For each $u\in\mathcal{U}_{\textsf{cover}}$, construct the auxiliary MDP
$\widehat{\mathcal{M}}^{s,u}$ as in Step~1. Since
$\widehat{V}_{s,u}^{\star}$ is independent of $\widehat{P}_{s,a}$ (conditional on the visitation counts $\{N_{s,a}\}$), applying
Lemma~\ref{lem:Bernstein-infinite-proof} and taking a union bound over
$u\in\mathcal{U}_{\textsf{cover}}$ (using the cardinality bound \eqref{eq:defn-Ucover-infinite-Bernstein-cardinality}) yields that: for each fixed $(s,a)$ with
$N_{s,a}>0$, with probability at least $1-\delta/N$ one has
\begin{subequations}
	\label{eq:Hoeffding-infinite-net-Bernstein-all}
	\begin{align}
		\Big|\big(\widehat{P}_{s,a}-P_{s,a}\big)\widehat{V}_{s,u}^{\star}\Big| & \le\sqrt{\frac{4\log\frac{16N^{2}}{(1-\gamma)\delta}}{N_{s,a}}\mathsf{Var}_{\widehat{P}_{s,a}}\big(\widehat{V}_{s,u}^{\star}\big)}+\frac{5\log\frac{16N^{2}}{(1-\gamma)\delta}}{(1-\gamma)N_{s,a}} \label{eq:Hoeffding-infinite-net-Bernstein-all-1}\\
		\mathsf{Var}_{\widehat{P}_{s,a}}\big(\widehat{V}_{s,u}^{\star}\big) & \le 2\mathsf{Var}_{P_{s,a}}\big(\widehat{V}_{s,u}^{\star}\big)+\frac{5\log\frac{16N^{2}}{(1-\gamma)\delta}}{3(1-\gamma)^{2}N_{s,a}} 
		\label{eq:Hoeffding-infinite-net-Bernstein-all-2}
	\end{align}
\end{subequations}
simultaneously for all $u\in \mathcal{U}_{\textsf{cover}}$.  
Since there are at most $N$ state-action pairs satisfying $N_{s,a}>0$, a further union bound over all these pairs reveals that, with probability at least
$1-\delta$, the bounds in~\eqref{eq:Hoeffding-infinite-net-Bernstein-all} hold uniformly for all $(s,a)\in \cS\times \cA$ satisfying $N_{s,a}>0$ and all $u$ from the associated covering set $\mathcal{U}_{\textsf{cover}}$.

\paragraph{Step 4: a covering argument.} 
In this step, we work on the high-probability event \eqref{eq:Hoeffding-infinite-net-Bernstein-all}.
By the construction in
Step~3, there exists some $u_0 \in \mathcal{U}_{\textsf{cover}}$ such that $| u_0 - u^{\star} | \leq 1/N$.
By the construction of the MDP $\widehat{\mathcal{M}}^{s, u}$ and the operator in \eqref{eq:empirical-Bellman-infinite-Lemma8}, 
we have
\[
\big\| \Tpess^{s,u_0}(Q) - \Tpess^{s,u^{\star}}(Q) \big\|_{\infty} \leq \big| u_0 - u^{\star} \big| \leq \frac{1}{N}
\]
for any $Q\in \mathbb{R}^{SA}$. Consequently, the $\gamma$-contraction property of the operator gives
\begin{align*}
	\big\|\widehat{Q}_{s,u_{0}}^{\star}-\widehat{Q}_{s,u^{\star}}^{\star}\big\|_{\infty} 
	&=\big\|\Tpess^{s,u_{0}}\big(\widehat{Q}_{s,u_{0}}^{\star}\big)-\Tpess^{s,u^{\star}}\big(\widehat{Q}_{s,u^{\star}}^{\star}\big)\big\|_{\infty}\\
	& \leq\big\|\Tpess^{s,u^{\star}}\big(\widehat{Q}_{s,u_{0}}^{\star}\big)-\Tpess^{s,u^{\star}}\big(\widehat{Q}_{s,u^{\star}}^{\star}\big)\big\|_{\infty} \arxiv{ \\
	&\qquad } +\big\|\Tpess^{s,u_{0}}\big(\widehat{Q}_{s,u_{0}}^{\star}\big)-\Tpess^{s,u^{\star}}\big(\widehat{Q}_{s,u_{0}}^{\star}\big)\big\|_{\infty}\\
	& \leq\gamma\big\|\widehat{Q}_{s,u_{0}}^{\star}-\widehat{Q}_{s,u^{\star}}^{\star}\big\|_{\infty}+\frac{1}{N}.
\end{align*}
Rearranging terms immediately gives
\begin{subequations}
\label{eq:Qhat-Vhat-ustar-u-gap-offline}
\begin{align}
\big\|\widehat{Q}_{s,u_{0}}^{\star}-\widehat{Q}_{s,u^{\star}}^{\star}\big\|_{\infty} \leq \frac{1}{(1-\gamma)N},
\end{align}
and hence
\begin{align}
\big\|\widehat{V}_{s,u_{0}}^{\star}-\widehat{V}_{s,u^{\star}}^{\star}\big\|_{\infty} 
\leq \big\|\widehat{Q}_{s,u_{0}}^{\star}-\widehat{Q}_{s,u^{\star}}^{\star}\big\|_{\infty} \leq \frac{1}{(1-\gamma)N}. 
\end{align}
\end{subequations}

We next use this perturbation bound to compare the corresponding variance terms. Specifically, 
\begin{align*}
	 &  \mathsf{Var}_{P_{s,a}}\big(\widehat{V}_{s,u_{0}}^{\star}\big)-\mathsf{Var}_{P_{s,a}}\big(\widehat{V}_{s,u^{\star}}^{\star}\big)   \nonumber \\
	& \quad \overset{\text{(i)}}{\leq} P_{s,a}\Big(\big(\widehat{V}_{s,u_{0}}^{\star} - \big\langle P_{s,a},\widehat{V}_{s,u^{\star}}^{\star}\big\rangle 1 \big) \circ\big(\widehat{V}_{s,u_{0}}^{\star}  - \big\langle  P_{s,a},\widehat{V}_{s,u^{\star}}^{\star}\big\rangle 1 \big) \Big)  
    \notag\\
    &\qquad
    - P_{s,a}\Big( \big(\widehat{V}_{s,u^{\star}}^{\star} - \big\langle P_{s,a},\widehat{V}_{s,u^{\star}}^{\star} \big\rangle  1 \big) \circ\big(\widehat{V}_{s,u^{\star}}^{\star} - \big\langle  P_{s,a}, \widehat{V}_{s,u^{\star}}^{\star} \big\rangle 1\big)\Big) \nonumber \\
	& \quad = P_{s,a}\Big(\big(\widehat{V}_{s,u_{0}}^{\star} - \big\langle  P_{s,a}, \widehat{V}_{s,u^{\star}}^{\star} \big\rangle 1 + \widehat{V}_{s,u^{\star}}^{\star} - \big\langle  P_{s,a}, \widehat{V}_{s,u^{\star}}^{\star} \big\rangle 1\big) \circ\big(\widehat{V}_{s,u_{0}}^{\star} - \widehat{V}_{s,u^{\star}}^{\star}\big)\Big) \nonumber \\
	& \quad \leq \frac{2}{1-\gamma}\big\|P_{s,a} \big\|_1 \big\| \widehat{V}_{s,u_{0}}^{\star}-\widehat{V}_{s,u^{\star}}^{\star}\big\|_{\infty} \\
	& \quad =\frac{2}{1-\gamma}\big\|\widehat{V}_{s,u_{0}}^{\star}-\widehat{V}_{s,u^{\star}}^{\star}\big\|_{\infty}
	\overset{\text{(ii)}}{\leq}\frac{2}{(1-\gamma)^{2}N}.
\end{align*}
Here, (i) follows from a direct use of $\mathbb{E}[X] = \arg\min_c \mathbb{E}[(X-c)^2]$, so that
\begin{align*}
	\mathsf{Var}_{P_{s,a}}\big(\widehat{V}_{s,u_{0}}^{\star}\big) & =P_{s,a}\Big(\big(\widehat{V}_{s,u_{0}}^{\star}-\big\langle P_{s,a},\widehat{V}_{s,u_{0}}^{\star}\big\rangle1\big)\circ\big(\widehat{V}_{s,u_{0}}^{\star}-\big\langle P_{s,a},\widehat{V}_{s,u_{0}}^{\star}\big\rangle1\big)\Big)\\
 & \leq P_{s,a}\Big(\big(\widehat{V}_{s,u_{0}}^{\star}-\big\langle P_{s,a},\widehat{V}_{s,u^{\star}}^{\star}\big\rangle1\big)\circ\big(\widehat{V}_{s,u_{0}}^{\star}-\big\langle P_{s,a},\widehat{V}_{s,u^{\star}}^{\star}\big\rangle1\big)\Big),
\end{align*}
while (ii) uses the bounds $0 \le \widehat{V}_{s,u_{0}}^{\star}, \widehat{V}_{s,u^{\star}}^{\star} \leq \frac{1}{1-\gamma} 1_S$. 
The above inequality continues to hold if we swap $\widehat{V}_{s,u_{0}}^{\star}$ and $\widehat{V}_{s,u^{\star}}^{\star}$. As a result, we arrive at
\begin{align}
	\Big|\mathsf{Var}_{P_{s,a}}\big(\widehat{V}_{s,u_{0}}^{\star}\big)-\mathsf{Var}_{P_{s,a}}\big(\widehat{V}_{s,u^{\star}}^{\star}\big)\Big| \leq\frac{2}{(1-\gamma)^{2}N}.
	\label{eq:Var-perturbation-V-su-inf}
\end{align}
The same argument also applies with $P_{s,a}$ replaced by $\widehat{P}_{s,a}$, yielding
\begin{align}
	\Big|\mathsf{Var}_{\widehat{P}_{s,a}}\big(\widehat{V}_{s,u_{0}}^{\star}\big)-\mathsf{Var}_{\widehat{P}_{s,a}}\big(\widehat{V}_{s,u^{\star}}^{\star}\big)\Big| \leq\frac{2}{(1-\gamma)^{2}N}.
	\label{eq:Var-perturbation-V-su-inf-Phat}
\end{align}

With these perturbation bounds in place, we can use the triangle inequality to show that
\begin{align}
	\Big|\big(\widehat{P}_{s,a}-P_{s,a}\big)\widehat{V}_{\mathsf{pe}}^{\star}\Big|  
	&\overset{\text{(i)}}{=}\Big|\big(\widehat{P}_{s,a}-P_{s,a}\big)\widehat{V}_{s,u^{\star}}^{\star}\Big| \notag \\
	&\leq\Big|\big(\widehat{P}_{s,a}-P_{s,a}\big)\widehat{V}_{s,u_{0}}^{\star}\Big|+\Big|\big(\widehat{P}_{s,a}-P_{s,a}\big)\big(\widehat{V}_{s,u^{\star}}^{\star}-\widehat{V}_{s,u_{0}}^{\star}\big)\Big| \notag\\
	& \overset{\text{(ii)}}{\leq}\Big|\big(\widehat{P}_{s,a}-P_{s,a}\big)\widehat{V}_{s,u_{0}}^{\star}\Big|+\frac{2}{N(1-\gamma)} \notag\\
	& \overset{\text{(iii)}}{\leq}\sqrt{\frac{4\log\frac{16N^{2}}{(1-\gamma)\delta}}{N_{s,a}}\mathsf{Var}_{\widehat{P}_{s,a}}\big(\widehat{V}_{s,u^{\star}}^{\star}\big)}
	+\frac{10\log\frac{16N^{2}}{(1-\gamma)\delta}}{(1-\gamma)N_{s,a}}.
	\label{eq:Phat-sa-Vsa-Vpe-bound-proof-123}
\end{align}
Here, (i) arises from \eqref{eq:V-hat-s-ustar-equal-V-pestar}; (ii) follows since, according to \eqref{eq:Qhat-Vhat-ustar-u-gap-offline}, 
\begin{align*}
	\Big|\big(\widehat{P}_{s,a}-P_{s,a}\big)\big(\widehat{V}_{s,u^{\star}}^{\star}-\widehat{V}_{s,u_{0}}^{\star}\big)\Big|
	&\leq\big(\big\|\widehat{P}_{s,a}\big\|_{1}+\big\| P_{s,a}\big\|_{1}\big)\big\|\widehat{V}_{s,u^{\star}}^{\star}-\widehat{V}_{s,u_{0}}^{\star}\big\|_{\infty}\\
	& \leq\frac{2}{N(1-\gamma)};
\end{align*}
and (iii) comes from \eqref{eq:Var-perturbation-V-su-inf-Phat} and \eqref{eq:Hoeffding-infinite-net-Bernstein-all-1}, which implies
\begin{align*}
	\Big|\big(\widehat{P}_{s,a}-P_{s,a}\big)\widehat{V}_{s,u_{0}}^{\star}\Big| 
	&\leq \sqrt{\frac{4\log\frac{16N^2}{(1-\gamma)\delta}}{N_{s,a}}\mathsf{Var}_{\widehat{P}_{s,a}}\big(\widehat{V}_{s,u_0}^{\star}\big)}
	+\frac{5\log\frac{16N^2}{(1-\gamma)\delta}}{(1-\gamma)N_{s,a}} \\
	&\leq \sqrt{\frac{4\log\frac{16N^2}{(1-\gamma)\delta}}{N_{s,a}}\mathsf{Var}_{\widehat{P}_{s,a}}\big(\widehat{V}_{s,u^\star}^{\star}\big)}
	+\frac{5\log\frac{16N^2}{(1-\gamma)\delta}}{(1-\gamma)N_{s,a}} \arxiv{ \\ 
	& \qquad} +\sqrt{\frac{8\log\frac{16N^2}{(1-\gamma)\delta}}{(1-\gamma)^{2}N\cdot N_{s,a}}} \\
	& \leq \sqrt{\frac{4\log\frac{16N^2}{(1-\gamma)\delta}}{N_{s,a}}\mathsf{Var}_{\widehat{P}_{s,a}}\big(\widehat{V}_{s,u^\star}^{\star}\big)}
	+\frac{8\log\frac{16N^2}{(1-\gamma)\delta}}{(1-\gamma)N_{s,a}},
\end{align*}
where the last step uses $N\geq N_{s,a}$. Similarly, one can also use \eqref{eq:V-hat-s-ustar-equal-V-pestar} to obtain
\begin{align}
	\mathsf{Var}_{\widehat{P}_{s,a}}\big(\widehat{V}_{\mathsf{pe}}^{\star}\big) 
	&= \mathsf{Var}_{\widehat{P}_{s,a}}\big(\widehat{V}_{s,u^{\star}}^{\star}\big) \nonumber \\
	&\leq\mathsf{Var}_{\widehat{P}_{s,a}}\big(\widehat{V}_{s,u_{0}}^{\star}\big)
	+\Big|\mathsf{Var}_{\widehat{P}_{s,a}}\big(\widehat{V}_{s,u^{\star}}^{\star}\big)-\mathsf{Var}_{\widehat{P}_{s,a}}\big(\widehat{V}_{s,u_{0}}^{\star}\big)\Big| \notag\\
	& \overset{(\mathrm{i})}{\leq}2\mathsf{Var}_{P_{s,a}}\big(\widehat{V}_{s,u_{0}}^{\star}\big)
	+\frac{5\log\frac{16N^{2}}{(1-\gamma)\delta}}{3(1-\gamma)^{2}N_{s,a}}
	+\frac{2}{(1-\gamma)^{2}N} \notag\\
	& \leq2\mathsf{Var}_{P_{s,a}}\big(\widehat{V}_{s,u^{\star}}^{\star}\big)
	+2\Big|\mathsf{Var}_{P_{s,a}}\big(\widehat{V}_{s,u^{\star}}^{\star}\big)-\mathsf{Var}_{P_{s,a}}\big(\widehat{V}_{s,u_{0}}^{\star}\big)\Big| \arxiv{ \nonumber\\
	&\qquad} +\frac{5\log\frac{16N^{2}}{(1-\gamma)\delta}}{3(1-\gamma)^{2}N_{s,a}}
	+\frac{2}{(1-\gamma)^{2}N} \notag\\
	& \overset{(\mathrm{ii})}{\leq}2\mathsf{Var}_{P_{s,a}}\big(\widehat{V}_{\mathsf{pe}}^{\star}\big)
	+\frac{6}{(1-\gamma)^{2}N}
	+\frac{5\log\frac{16N^{2}}{(1-\gamma)\delta}}{3(1-\gamma)^{2}N_{s,a}} \notag\\
	& \overset{(\mathrm{iii})}{\leq}2\mathsf{Var}_{P_{s,a}}\big(\widehat{V}_{\mathsf{pe}}^{\star}\big)
	+\frac{23\log\frac{16N^{2}}{(1-\gamma)\delta}}{3(1-\gamma)^{2}N_{s,a}},
	\label{eq:Var-Vpe-hat-diff-proof-135}
\end{align}
where (i) results from \eqref{eq:Hoeffding-infinite-net-Bernstein-all-2} and \eqref{eq:Var-perturbation-V-su-inf-Phat}, 
(ii) follows from \eqref{eq:Var-perturbation-V-su-inf}, 
and (iii) uses $N\geq N_{s,a}$.

\paragraph{Step 5: extending the bounds to $\widetilde{V}$.}
Now consider any $\widetilde{V}$ satisfying
\[
\big\|\widetilde{V}-\widehat{V}_{\mathsf{pe}}^{\star}\big\|_{\infty}
\leq \frac{1}{N} \qquad \text{and}
\qquad
\|\widetilde{V}\|_{\infty}\leq \frac{1}{1-\gamma}.
\]
For notational convenience, define
\[
\ell \coloneqq \log\frac{4N}{(1-\gamma)\delta}.
\]
Since $(1-\gamma)\delta\leq 1$, we have
\begin{align}
\log\frac{16N^{2}}{(1-\gamma)\delta}
\leq
2\log\frac{4N}{(1-\gamma)\delta}
=2\ell.
\label{eq:log-comparison-infinite}
\end{align}
Invoking \eqref{eq:Phat-sa-Vsa-Vpe-bound-proof-123} and using
\eqref{eq:log-comparison-infinite}, we obtain
\begin{align}
\Big|\big(\widehat{P}_{s,a}-P_{s,a}\big)
\widehat{V}_{\mathsf{pe}}^{\star}\Big|
&\leq
\sqrt{
    \frac{8\ell}{N_{s,a}}
    \mathsf{Var}_{\widehat{P}_{s,a}}
    \big(\widehat{V}_{\mathsf{pe}}^{\star}\big)
}
+\frac{20\ell}{(1-\gamma)N_{s,a}}.
\label{eq:Psa-Vpe-simplified}
\end{align}
Moreover, the triangle inequality gives
\begin{align}
\Big|\big(\widehat{P}_{s,a}-P_{s,a}\big)\widetilde{V}\Big|
&\leq
\Big|\big(\widehat{P}_{s,a}-P_{s,a}\big)
\widehat{V}_{\mathsf{pe}}^{\star}\Big|
+
\Big|\big(\widehat{P}_{s,a}-P_{s,a}\big)
\big(\widetilde{V}-\widehat{V}_{\mathsf{pe}}^{\star}\big)\Big|
\notag\\
&\leq
\Big|\big(\widehat{P}_{s,a}-P_{s,a}\big)
\widehat{V}_{\mathsf{pe}}^{\star}\Big|
+\frac{2}{N},
\label{eq:Psa-Vtilde-triangle}
\end{align}
where the second inequality follows from
\[
\|\widehat{P}_{s,a}\|_1=\|P_{s,a}\|_1=1
\quad\text{and}\quad
\big\|\widetilde{V}-\widehat{V}_{\mathsf{pe}}^{\star}\big\|_\infty
\leq \frac{1}{N}.
\]
The same variance-perturbation argument used to establish
\eqref{eq:Var-perturbation-V-su-inf-Phat} yields
\begin{align}
\Big|
\mathsf{Var}_{\widehat{P}_{s,a}}
\big(\widehat{V}_{\mathsf{pe}}^{\star}\big)
-
\mathsf{Var}_{\widehat{P}_{s,a}}
\big(\widetilde{V}\big)
\Big|
\leq
\frac{2}{(1-\gamma)^2N},
\label{eq:Var-perturbation-Vtilde}
\end{align}
and as a result,
\begin{align}
\sqrt{
\mathsf{Var}_{\widehat{P}_{s,a}}
\big(\widehat{V}_{\mathsf{pe}}^{\star}\big)
}
&\leq
\sqrt{
\mathsf{Var}_{\widehat{P}_{s,a}}
\big(\widetilde{V}\big)
}
+
\sqrt{\frac{2}{(1-\gamma)^2N}}.
\label{eq:sqrt-Var-perturbation-Vtilde}
\end{align}
Combining
\eqref{eq:Psa-Vpe-simplified},
\eqref{eq:Psa-Vtilde-triangle} and
\eqref{eq:sqrt-Var-perturbation-Vtilde}, we arrive at
\begin{align}
\Big|\big(\widehat{P}_{s,a}-P_{s,a}\big)\widetilde{V}\Big|
&\leq
\sqrt{
    \frac{8\ell}{N_{s,a}}
    \mathsf{Var}_{\widehat{P}_{s,a}}
    \big(\widetilde{V}\big)
}
+
\sqrt{
    \frac{16\ell}
    {(1-\gamma)^2N N_{s,a}}
}
+
\frac{20\ell}{(1-\gamma)N_{s,a}}
+
\frac{2}{N} \notag\\
&\leq
\sqrt{
    \frac{8\ell}{N_{s,a}}
    \mathsf{Var}_{\widehat{P}_{s,a}}
    \big(\widetilde{V}\big)
}
+
\frac{26\ell}{(1-\gamma)N_{s,a}},
\label{eq:Psa-hat-perturb-13579}
\end{align}
thereby establishing~\eqref{eq:bonus-Bernstein-infinite}.

The other desired
inequality, \eqref{eq:empirical-var-infinite}, follows analogously by
combining \eqref{eq:Var-Vpe-hat-diff-proof-135} with the variance
perturbation bound \eqref{eq:Var-perturbation-Vtilde}.

\subsection{Proof of Lemma~\ref{lem:sum-d-Var-UB}}\label{proof:lem:sum-d-Var-UB}

Recall the definition of $P^{\star}$ and $b^{\star}$ in \eqref{eq:P-b-star-infinite}. 
We first make the observation that
\begin{align}
	\big(\widehat{V} \circ \widehat{V}\big) - \big(\gamma P^{\star}\widehat{V}) \circ \big(\gamma P^{\star}\widehat{V}) 
	&= \big(\widehat{V} - \gamma P^{\star}\widehat{V}\big) \circ \big(\widehat{V} + \gamma P^{\star}\widehat{V}\big) \notag\\
	& \overset{\mathrm{(i)}}{\le} \big(\widehat{V} - \gamma P^{\star}\widehat{V} + 2 b^\star \big) \circ \big(\widehat{V} + \gamma P^{\star}\widehat{V}\big) \notag \\
	&\overset{\mathrm{(ii)}}{\le} \frac{2}{1-\gamma}\big(\widehat{V} - \gamma P^{\star}\widehat{V} + 2 b^\star \big). 
	\label{eq:infinite-extra-lemma-au1}
\end{align}
Here, (i) arises from $b^\star \geq 0$ and $\widehat{V} + \gamma P^{\star}\widehat{V} \geq 0$; (ii) follows from
\[
\|\widehat{V} + \gamma P^{\star}\widehat{V}\|_\infty 
\leq 2\|\widehat{V}\|_\infty 
\leq \frac{2}{1-\gamma},
\]
together with the fact that $\widehat{V} - \gamma P^{\star}\widehat{V} + 2 b^\star \geq 0$, which has been verified in \eqref{eq:infinite-v-lower}.  
With this observation in mind, we can demonstrate that
\begin{align*}
	\arxiv{&} \sum_{s} d^{\star}(s) \mathsf{Var}_{P_{s, \pi^{\star}(s)}}\big(\widehat{V}\big) 
	\book{&} \overset{\mathrm{(i)}}{=} 
	\left\langle d^{\star}, P^{\star}\big(\widehat{V} \circ \widehat{V}\big) - \big(P^{\star}\widehat{V}\big) \circ \big(P^{\star}\widehat{V}\big)\right\rangle \\
	&\arxiv{ \qquad} \overset{\mathrm{(ii)}}{\le} 
	\left\langle d^{\star}, P^{\star}\big(\widehat{V} \circ \widehat{V}\big) 
	-\frac{1}{\gamma^2} \widehat{V} \circ \widehat{V} 
	+ \frac{2}{\gamma^2(1-\gamma)}
	\big(\widehat{V} - \gamma P^{\star}\widehat{V} + 2 b^\star \big)\right\rangle \\
	& \arxiv{\qquad} \overset{\mathrm{(iii)}}{\le} 
	\left\langle d^{\star}, P^{\star}\big(\widehat{V} \circ \widehat{V}\big) 
	-\frac{1}{\gamma} \widehat{V} \circ \widehat{V} 
	+ \frac{2}{\gamma^2(1-\gamma)} \big( (I - \gamma P^{\star})\widehat{V} + 2 b^\star \big) \right\rangle \notag\\
	& \arxiv{\qquad} =
	d^{\star\top}\big(I-\gamma P^{\star}\big)
	\left\{ -\frac{1}{\gamma}\widehat{V}\circ\widehat{V}
	+\frac{2}{\gamma^{2}(1-\gamma)}\widehat{V}\right\} 
	+\frac{4}{\gamma^{2}(1-\gamma)}\langle d^{\star},b^{\star}\rangle \\
	& \arxiv{\qquad} \overset{\mathrm{(iv)}}{\le} 
	(1-\gamma)\rho^{\top} 
	\left\{ -\frac{1}{\gamma} \widehat{V} \circ \widehat{V} 
	+ \frac{2}{\gamma^2(1-\gamma)}\widehat{V} \right\} 
	+ \frac{4}{\gamma^2(1-\gamma)}\left\langle d^{\star}, b^{\star}\right\rangle \\
	&\arxiv{\qquad} \leq 
	\frac{2}{\gamma^2} \rho^{\top}  \widehat{V} 
	+ \frac{4}{\gamma^2(1-\gamma)}\left\langle d^{\star}, b^{\star}\right\rangle  \overset{\mathrm{(v)}}{\le} 
	\frac{2}{\gamma^2(1-\gamma)} 
	+ \frac{4}{\gamma^2(1-\gamma)}\left\langle d^{\star}, b^{\star}\right\rangle. 
\end{align*}
Here, (i) follows from the variance identity \eqref{eq:matrix-VarP-V-expression}; 
(ii) arises from \eqref{eq:infinite-extra-lemma-au1}; 
(iii) uses $\gamma<1$; 
(iv) is a direct consequence of the matrix-vector representation \eqref{eq:infinite-d-matrix-form}; 
and (v) follows from $\|\rho\|_1=1$ and $\|\widehat{V}\|_\infty \leq 1/(1-\gamma)$.

\subsection{Proof of Lemma~\ref{lem:Bernstein-infinite-proof}}
\label{sec:lem:Bernstein-infinite-proof}

In this subsection, we prove Lemma~\ref{lem:Bernstein-infinite-proof} using the concentration
inequalities developed in Section~\ref{sec:concentration}. We consider any given $(s,a)$ obeying $N_{s,a}>0$.

\paragraph{Proof of inequality~\eqref{eq:bonus-Bernstein-infinite-proof}.}
If $0< N_{s,a} < 5 \log \frac{1}{\delta}$, then we can immediately see that 
\begin{align*}
	\left|\big(\widehat{P}_{s, a} - P_{s, a}\big)V\right| 
	\leq \|V\|_{\infty} \leq \frac{1}{1-\gamma} \leq \frac{5 \log\frac{1}{\delta}}{(1-\gamma) N_{s,a}},
\end{align*}
and hence the claim~\eqref{eq:bonus-Bernstein-infinite-proof} is valid. As a result, it suffices to look at the case where
\begin{align}
	N_{s,a} \geq  5 \log \frac{1}{\delta}.
	\label{eq:Nhsa-lower-bound-infinite}
\end{align}
Applying the empirical Bernstein inequality (cf.~Lemma~\ref{empirical bernstein}) yields, with probability exceeding $1-4\delta$, 
\begin{align*}
	\big|\big(\widehat{P}_{s,a}-P_{s,a})V\big| & \leq\sqrt{\frac{2\log\frac{1}{\delta}}{N_{s,a}-1}}\sqrt{\mathsf{Var}_{\widehat{P}_{s,a}}(V)}+\frac{7\|V\|_{\infty}\log\frac{1}{\delta}}{3(N_{s,a}-1)} \\
	& \leq \sqrt{\frac{4\mathsf{Var}_{\widehat{P}_{s,a}}(V)\log\frac{1}{\delta}}{N_{s,a}}}+\frac{5 \log\frac{1}{\delta}}{(1-\gamma) N_{s,a}},
\end{align*}
where the last inequality uses $\|V\|_{\infty}\leq \frac{1}{1-\gamma}$ as well as $N_{s,a}-1 \geq N_{s,a}/2$ under the condition~\eqref{eq:Nhsa-lower-bound-infinite}.

\paragraph{Proof of inequality~\eqref{eq:empirical-var-infinite-proof}.}
Defining $\overline{V} = V - (P_{s, a}V) 1_S$, one can invoke the Bernstein inequality (cf.~Lemma~\ref{lem:Bernstein-prelim}) to show that, with probability at least $1-4\delta$,
\begin{align*}
	\left|\big(P_{s, a} - \widehat{P}_{s, a}\big)\big(\overline{V} \circ \overline{V}\big)\right| 
	&\le \sqrt{\frac{2\mathsf{Var}_{P_{s, a}}\big(\overline{V} \circ \overline{V}\big)\log\frac{1}{\delta}}{N_{s, a}}} + \frac{2\| \overline{V} \circ \overline{V} \|_{\infty} \log\frac{1}{\delta}}{3N_{s, a}} \nonumber\\
	&\le \sqrt{\frac{2\mathsf{Var}_{P_{s, a}}(V)\log\frac{1}{\delta}}{(1-\gamma)^2N_{s, a}}} + \frac{2\log\frac{1}{\delta}}{3(1-\gamma)^2N_{s, a}}	,
\end{align*}
where the second inequality results from Lemma~\ref{lemma:sqv} and the fact that $\|V\|_{\infty}\leq \frac{1}{1-\gamma}$. In addition, we observe that
\begin{align*} 
\mathsf{Var}_{P_{s,a}}(V) & =P_{s,a}\big(\overline{V}\circ\overline{V}\big)=\big(P_{s,a}-\widehat{P}_{s,a}\big)\big(\overline{V}\circ\overline{V}\big)+\widehat{P}_{s,a}\big(\overline{V}\circ\overline{V}\big)\nonumber\\
 & =\big(P_{s,a}-\widehat{P}_{s,a}\big)\big(\overline{V}\circ\overline{V}\big)+\mathsf{Var}_{\widehat{P}_{s,a}}(V)+\big|\widehat{P}_{s,a}\overline{V}\big|^{2}\\
 & =\big(P_{s,a}-\widehat{P}_{s,a}\big)\big(\overline{V}\circ\overline{V}\big)+\mathsf{Var}_{\widehat{P}_{s,a}}(V)+\big|\big(\widehat{P}_{s,a}-P_{s,a}\big)V\big|^{2}.
\end{align*}
Taking the above two equations collectively yields, with probability at least $1-4\delta$, 
\begin{align*}
	\mathsf{Var}_{P_{s, a}} (V) & \ge -\left|\big(P_{s,a}-\widehat{P}_{s,a}\big)\big(\overline{V}\circ\overline{V}\big)\right|+\mathsf{Var}_{\widehat{P}_{s,a}}(V) \nonumber\\
	& \ge-\sqrt{\frac{2\mathsf{Var}_{P_{s, a}}\big(V\big)\log\frac{1}{\delta}}{(1-\gamma)^2N_{s,a}}}-\frac{2\log\frac{1}{\delta}}{3(1-\gamma)^2N_{s,a}}+\mathsf{Var}_{\widehat{P}_{s,a}}(V).
\end{align*}
Rearranging terms and applying the AM-GM inequality, we have
\begin{align*}
	\mathsf{Var}_{\widehat{P}_{s,a}}(V) & \leq\mathsf{Var}_{P_{s, a}}(V)+ 2\sqrt{\frac{\mathsf{Var}_{P_{s, a}}\big(V\big)\log\frac{1}{\delta}}{(1-\gamma)^2N_{s,a}}}+\frac{2\log\frac{1}{\delta}}{3(1-\gamma)^2N_{s,a}}\\
	& \leq 2\mathsf{Var}_{P_{s, a}}(V)+\frac{5\log\frac{1}{\delta}}{3(1-\gamma)^2N_{s,a}} .
\end{align*}

\section{Appendix B: proof of auxiliary lemmas in Section~\ref{sec:analysis-model-free-offline-RL}}
\label{sec:proof-model-free-offline-RL}

\subsection{Proof of Lemma~\ref{lemma:Q-recursion-pessimistic-Q}}
\label{sec:proof-lemma-Q-recursion-pessimistic}

Consider any fixed $(s,a)\in\cS\times\cA$ and any fixed
$1\leq t\leq T$. For notational simplicity, let $n$ denote the
number of visits to $(s,a)$ up to iteration $t$, and let $k_i$ denote
the iteration at which $(s,a)$ is visited for the $i$-th time. For each
$1\leq i\leq n$, define the indicator vector
$P_{k_i}\in\{0,1\}^{1\times S}$ associated with the next state observed
during the $i$-th visit by
\begin{equation}
P_{k_i}(s')\coloneqq\mathds{1}\{s'=s_{k_i}'\},
\qquad \forall s'\in\cS.
\label{eq:defn-P-ki-vector-lcbq-proof}
\end{equation}

If $n=0$, then $Q_t(s,a)=0$, and the claims~\eqref{eq:qstar-recursion-lcbq} and \eqref{eq:uniform-policy-pessimism-recursion} follow immediately. It therefore suffices to focus on the case $n>0$ below. To unroll the pessimistic Q-iterates, one can apply the update rule~\eqref{eq:Q-lcb-update-rule-Q} recursively and employ the initialization $Q_0 =0$ to show that
\begin{align}
Q_t(s,a) &=\sum_{i=1}^{n}\eta_i^n\Big\{r(s,a)+\gamma V_{k_i-1}(s_{k_i}')-b_i \Big\} \notag\\
&=r(s,a)+\sum_{i=1}^{n}\eta_i^n\Big\{\gamma P_{k_i}V_{k_i-1}-b_i \Big\},
\label{eq:unroll-q-lcbq}
\end{align}
where we recall the definition of $\eta_i^n$ in \eqref{eq:learning-rate-notation-all-ucbq} and have used property~\eqref{eq:sum-eta-n-N} in the last line. Combine this with the Bellman equation $Q^{\star}(s,a)=r(s,a)+\gamma P_{s,a}V^{\star}$ to yield
\begin{align}
\big(Q^\star-Q_t\big)(s,a)
&= \gamma P_{s,a} V^{\star} -  \sum_{i=1}^{n}\eta_i^n\Big\{\gamma P_{k_i}V_{k_i-1} -b_i \Big\} \nonumber \\
& = \gamma\sum_{i=1}^{n}\eta_i^n P_{s,a}\big(V^\star-V_{k_i-1}\big)
 \arxiv{ \nonumber\\
&\quad } +\gamma\sum_{i=1}^{n}\eta_i^n (P_{s,a}-P_{k_i})V_{k_i-1}   +\sum_{i=1}^{n}\eta_i^n b_i,
\label{eq:qstar-recursion-all-actions-lcbq}
\end{align}
where the second line again relies on property~\eqref{eq:sum-eta-n-N}. Similarly, taking~\eqref{eq:unroll-q-lcbq} together with the Bellman equation $Q^{\pi}(s,a)=r(s,a)+\gamma P_{s,a}V^{\pi}$ gives
\begin{align}
\big(Q^\pi-Q_t\big)(s,a)
&=
\gamma\sum_{i=1}^{n}\eta_i^n
P_{s,a}\bigl(V^\pi-V_{k_i-1}\bigr)
\arxiv{\nonumber\\
&\quad}+
\gamma\sum_{i=1}^{n}\eta_i^n
(P_{s,a}-P_{k_i})V_{k_i-1}
+
\sum_{i=1}^{n}\eta_i^n b_i.
\label{eq:policy-pessimism-decomposition}
\end{align}
These leave us with several terms to control. 
\begin{itemize}
\item 
We first control the second term on the right-hand side of~\eqref{eq:qstar-recursion-all-actions-lcbq} (or equivalently, the second term on the right-hand side of~\eqref{eq:policy-pessimism-decomposition}). Since the visitation count
$n$ is random, we would like to establish a bound that holds simultaneously for all
possible values of $n$. To do so, invoke the Azuma-Hoeffding inequality (see Lemma~\ref{lem:azuma-hoeffding}) and take the union bound over all $1\leq \ell \leq T$ to show that, with probability at least $1-\delta/(SAT)$,
\begin{align*}
\left|\gamma\sum_{i=1}^{\ell}\eta_{i}^{\ell}(P_{s,a}-P_{k_{i}})V_{k_{i}-1}\right| & \leq\sqrt{2\sum_{i=1}^{\ell}\big(\eta_{i}^{\ell}\big)^{2}\frac{1}{(1-\gamma)^{2}}\log\frac{2SAT^2}{\delta}}\notag\\
 & \leq2\sqrt{\frac{H}{\ell(1-\gamma)^{2}}\log\frac{2SAT^2}{\delta}}
 %\label{eq:transition-martingale-lcb-ell}
\end{align*}
holds simultaneously for all $1\leq \ell \leq T$, where the last inequality results from \eqref{eq:properties-learning-rates-square}. Recognizing the trivial fact that $n\in [T]$, we can take $\ell=n$ in the preceding bound to obtain
\begin{align}
\left|\gamma\sum_{i=1}^{n}\eta_{i}^{n}(P_{s,a}-P_{k_{i}})V_{k_{i}-1}\right| 
 & \leq2\sqrt{\frac{H}{n(1-\gamma)^{2}}\log\frac{2SAT^2}{\delta}}
 \label{eq:transition-martingale-lcb}
\end{align}
with probability exceeding $1-\delta/(SAT)$.

\item We now turn to the last term of~\eqref{eq:qstar-recursion-all-actions-lcbq} (or equivalently, the last term of~\eqref{eq:policy-pessimism-decomposition}). By an argument analogous to that used in the online setting (see Lemma~\ref{lemma:opt_model_free} and its proof), we have
\begin{equation} 
c_{\mathsf b}\sqrt{\frac{H\iota}{n(1-\gamma)^2}}
\le
\sum_{i=1}^{n}\eta_i^n b_i
\le
2c_{\mathsf b}\sqrt{\frac{H\iota}{n(1-\gamma)^2}}.
\label{eq:weighted-penalty-lcb}
\end{equation}
\end{itemize}
Substitution into~\eqref{eq:qstar-recursion-all-actions-lcbq} and \eqref{eq:policy-pessimism-decomposition} then gives, with probability at least $1-\delta/(SAT)$, 
\begin{subequations}
\label{eq:Q-star-Qt-gap-lcbq-2469}
\begin{align}
\big(Q^\star-Q_t\big)(s,a) &\leq \gamma\sum_{i=1}^{n}\eta_i^n P_{s,a}\big(V^\star-V_{k_i-1}\big) + \beta_n,
\label{eq:Q-star-Qt-gap-lcbq-2469-i} \\
\big(Q^{\pi}-Q_t\big)(s,a) &\geq \gamma\sum_{i=1}^{n}\eta_i^n P_{s,a}\big(V^{\pi}-V_{k_i-1}\big) ,
\label{eq:Q-star-Qt-gap-lcbq-1358}
\end{align}
\end{subequations}
provided that $c_{\mathsf{b}}$ is sufficiently large. Importantly, \eqref{eq:Q-star-Qt-gap-lcbq-1358} holds simultaneously for all deterministic policy $\pi$, regardless of whether it is dependent on the data samples.

Taking the union bound over all $1\leq t\leq T$ and all $(s,a)\in \cS\times \cA$, we see that with probability at least $1-\delta$, the inequalities in \eqref{eq:Q-star-Qt-gap-lcbq-2469} hold simultaneously for all $1\leq t\leq T$, all $(s,a)\in \cS\times \cA$, and all deterministic policy $\pi$.  This concludes the proof.

\subsection{Proof of Lemma~\ref{lemma:pessimistic-Q} }
\label{sec:proof-pessimistic-Q}

To begin with, we single out an immediate, useful consequence of Lemma~\ref{lemma:Q-recursion-pessimistic-Q}: taking
$\pi=\pi^\star$ and using
$Q^{\pi^\star}=Q^\star$ and $V^{\pi^\star}=V^\star$ in Lemma~\ref{lemma:Q-recursion-pessimistic-Q} lead to
\begin{align}
Q^\star(s,a)-Q_t(s,a)
\geq
\gamma\sum_{i=1}^{n_t(s,a)}
\eta_i^{n_t(s,a)}
P_{s,a}\bigl(V^\star-V_{k_i(s,a)-1}\bigr)
\label{eq:pessimistic-Q-lower-recursion}
\end{align}
simultaneously for all $(s,a)\in\cS\times\cA$ and
$1\leq t\leq T$, on an event of probability at least $1-\delta$.

We first prove that $Q_t\leq Q^\star$ and $V_t\leq V^\star$
by induction. The claim is immediate for the base case with
$t=0$, since $Q_0=0$ and $V_0=0$. Suppose for the moment that
\[
Q_\tau\leq Q^\star
\qquad\text{and}\qquad
V_\tau\leq V^\star
\]
hold for every $\tau<t$, so that $V_{k_i(s,a)-1}\leq V^\star$
 for every $i\leq n_t(s,a)$. Hence, the right-hand side of
\eqref{eq:pessimistic-Q-lower-recursion} is nonnegative, which readily implies that
\[
Q_t(s,a)\leq Q^\star(s,a)
\qquad\text{for all }(s,a)\in\cS\times\cA.
\]
With the monotone update of value estimates in mind, we further obtain
\[
V_t(s)
\leq
\max\Bigl\{V_{t-1}(s),\max_a Q^\star(s,a)\Bigr\}
=
V^\star(s).
\]
This completes the induction, and establishes $Q_t\leq Q^\star$ and $V_t\leq V^\star$ for all $0\leq t\leq T$.

It remains to prove that $V_t\leq V^{\pi_t}$. Consider any
$t\in\{1,\ldots,T\}$. Consider first any state $s$ obeying $V_t(s)>0$, and define
\[
\tau_s
\coloneqq
\max\Bigl\{u\in\{1,\ldots,t\}:
V_u(s)>V_{u-1}(s)\Bigr\}.
\]
Since the value iterates are nondecreasing in $t$, we have
\[
V_{\tau_s}(s)=V_t(s).
\]
Moreover, the policy update rule~\eqref{eq:analysis-policy-lcbq} indicates that
\[
\pi_t(s)=\arg\max_{a\in\cA}Q_{\tau_s}(s,a)
\qquad\text{and}\qquad
V_t(s)=Q_{\tau_s}\bigl(s,\pi_t(s)\bigr).
\]
With this relation in mind, 
applying the recursion
\eqref{eq:uniform-policy-pessimism-recursion} in Lemma~\ref{lemma:Q-recursion-pessimistic-Q} with
$\pi=\pi_t$, $a=\pi_t(s)$, and $t=\tau_s$ yields
\begin{align}
V^{\pi_t}(s)
-V_t(s)
&= Q^{\pi_t}\bigl(s,\pi_t(s)\bigr)
-Q_{\tau_s}\bigl(s,\pi_t(s)\bigr) \notag\\
&\geq
\gamma\sum_{i=1}^{n_{\tau_s}(s,\pi_t(s))}
\eta_i^{n_{\tau_s}(s,\pi_t(s))}
P_{s,\pi_t(s)}
\bigl(V^{\pi_t}-V_{k_i(s,\pi_t(s))-1}\bigr).
\label{eq:policy-pessimism-recursion-lcbq}
\end{align}
Now, let us isolate the quantity
\[
m\coloneqq
\min_{s\in\cS}
\bigl\{V^{\pi_t}(s)-V_t(s)\bigr\}.
\]
In view of the fact $k_i(s,\pi_t(s))\leq \tau_s\leq t$ (for all $i\leq n_{\tau_s}(s,\pi_t(s))$) and the monotonicity of 
$V_u$ in $u$, 
\[
V^{\pi_t}-V_{k_i(s,\pi_t(s))-1}
\geq
V^{\pi_t}-V_t
\geq m\,1_S.
\]
Recognizing that $P_{s,\pi_t(s)}$ is a probability distribution vector and
$\sum_{i=1}^{n}
\eta_i^{n}=1$ for any integer $n\geq 1$,  
we can see from \eqref{eq:policy-pessimism-recursion-lcbq} that,
for every state $s$ satisfying $V_t(s)>0$,
\[
V^{\pi_t}(s)-V_t(s)\geq \gamma m.
\]

Moreover, for any state $s$ satisfying $V_t(s)=0$, we already know that
$V^{\pi_t}(s)-V_t(s)\geq 0$. 
Thus, in all cases, we have 
\[
m\geq \min\{0,\gamma m\}.
\]
If $m<0$, then the above condition simplifies to
$m\geq\gamma m$, which is impossible because $\gamma<1$. Therefore, one necessarily has $m\geq0$, thereby establishing that
$V_t\leq V^{\pi_t} \leq V^{\star}$. 

\subsection{Proof of Lemma~\ref{lemma:useful}}
\label{sec:proof-lemma:useful-lcbq-step5}

Recall the definition of $\kappa<1$ in \eqref{eq:defn-kappa-proof-lcbq}. 
Fix an integer $K\geq 1$. Since $W$ has nonnegative entries and
$\rho^\top(P_{\pi^\star})^j$ is a probability distribution vector, we can split the series at $K$ and obtain
\begin{align}
\sum_{j=0}^{\infty}
\kappa^j
\rho^\top(P_{\pi^\star})^j W 
&=
\sum_{j=0}^{K-1}
\kappa^j
\rho^\top(P_{\pi^\star})^j W
+
\sum_{j=K}^{\infty}
\kappa^j
\rho^\top(P_{\pi^\star})^j W
\nonumber\\
&\leq
\left(1+\frac{1}{H}\right)^{3K}
\sum_{j=0}^{K-1}
\gamma^j
\rho^\top(P_{\pi^\star})^j W
+
\sum_{j=K}^{\infty}
\kappa^j\|W\|_\infty \notag\\
&\leq
\left(1+\frac{1}{H}\right)^{3K}
\sum_{j=0}^{\infty}
\gamma^j
\rho^\top(P_{\pi^\star})^j W
+
\frac{\kappa^K}{1-\kappa}\|W\|_\infty.
\label{eq:useful-split}
\end{align}
Regarding the first term on the right-hand side of \eqref{eq:useful-split}, it is seen from the
definition of the discounted occupancy distribution (cf.~\eqref{eq:defn-dpi-occupancy-discounted-dist} and \eqref{eq:defn-dpi-occupancy-discounted-dist-opt}) that
\[
d_\rho^\star
=
(1-\gamma)
\sum_{j=0}^{\infty}
\gamma^j
\rho^\top(P_{\pi^\star})^j,
\]
and as a result,
\begin{align}
\sum_{j=0}^{\infty}
\gamma^j
\rho^\top(P_{\pi^\star})^j W
\arxiv{\nonumber\\
&}=
\frac{1}{1-\gamma}
\langle d_\rho^\star,W\rangle.
\label{eq:useful-main-term}
\end{align}
Substituting it into~\eqref{eq:useful-split} and taking $K=H$ (so that $\big(1+\frac{1}{H}\big)^{3K}\leq e^3$), we reach
\begin{align}
\sum_{j=0}^{\infty}
\kappa^j
\rho^\top(P_{\pi^\star})^j W
&\leq
\left(1+\frac{1}{H}\right)^{3K}
\frac{\langle d_\rho^\star,W\rangle}{1-\gamma}
+
\frac{\kappa^K}{1-\kappa}\|W\|_\infty \notag\\
&\lesssim
\frac{\langle d_\rho^\star,W\rangle}{1-\gamma}
+
\frac{\kappa^K}{1-\kappa}\|W\|_\infty.
\label{eq:useful-before-K-choice}
\end{align}

With regards to the last term in \eqref{eq:useful-before-K-choice}, it follows from the elementary fact $\log(1-x)\leq -x$ for $x\in(0,1)$ and the choice~\eqref{eq:defn-H-lcbq-stepsize-proof} of $H$ that
\[
\gamma^H
=
\exp\!\big(H\log\gamma\big)
\leq
\exp\!\big(-H(1-\gamma)\big)
\leq
\frac{\delta}{ST^4}.
\]
Moreover, a little algebra reveals that $1-\kappa
\gtrsim 1-\gamma.$
As a consequence, 
\[
\frac{\kappa^H}{1-\kappa}
\leq
\frac{\gamma^H(1+1/H)^{3H}}{1-\kappa}
\lesssim
\frac{\delta}{ST^4(1-\gamma)}.
\]
Taking this bound together with \eqref{eq:useful-before-K-choice}, we conclude the proof.

\subsection{Proof of Lemma~\ref{lem:lcbq-alpha-xi-psi-phi-bound}}
\label{sec:proof:lem:lcbq-alpha-xi-psi-phi-bound}

Throughout this proof, we adopt the notation $\kappa$ defined in~\eqref{eq:defn-kappa-proof-lcbq}.

\paragraph{Bounding $\alpha$.}
To begin with, recall from~\eqref{eq:defn-kappa-proof-lcbq} that $\kappa<1$. Recognizing that $ \|\Delta_t\|_{\infty}\le (1-\gamma)^{-1}$ (see~\eqref{eq:Delta-chi-offlineQ-bound}) and that $\rho^{\top}(P_{\pi^\star})^j$ forms a probability distribution vector, we can derive
\begin{align}
\alpha
& =\limsup_{j\to\infty}\kappa^j\sum_{t=0}^{T-1} \rho^{\top}(P_{\pi^\star})^j\Delta_t \arxiv{ \nonumber \\
&} \le \frac{T}{1-\gamma}\limsup_{j\to\infty}\kappa^j=0.
\label{eq:alpha-bound-offlineQ}
\end{align}

\paragraph{Bounding $\xi$.}  
Regarding the term~$\xi$, one has
\begin{align}
\xi
&=\sum_{j=0}^{\infty}\kappa^{j+1}\rho^{\top}(P_{\pi^\star})^{j+1}\Delta_0 \arxiv{\nonumber \\
&} \lesssim
\frac{1}{1-\gamma}\big\langle d_\rho^\star,\Delta_0\big\rangle
+\frac{\delta}{ST^4(1-\gamma)}\|\Delta_0\|_\infty
\lesssim \frac{1}{(1-\gamma)^2},
\label{eq:xi-bound-offlineQ}
\end{align}
where the first inequality arises from Lemma~\ref{lemma:useful}, and the last inequality holds since $\|\Delta_0\|_{\infty}\leq \frac{1}{1-\gamma}$.

\paragraph{Bounding $\theta$.}
By the definition~\eqref{eq:theta-j-offlineQ} of $\theta_j$ and Tonelli's theorem, one can interchange the
sums over $j$ and $t$ to obtain
\begin{align}
\theta
&=
\sum_{t=1}^{T}\sum_{s\in\cS}
\left[
\sum_{j=0}^{\infty}
\kappa^j
\big[\rho^\top(P_{\pi^\star})^j\big](s)
\right]
W_t(s),
\label{eq:theta-rearranged-offlineQ}
\end{align}
where we define the nonnegative vector $W_t\in \mathbb{R}^S$ such that
\begin{align}
W_t(s)\coloneqq
\min\left\{
\beta_{n_t(s,\pi^\star(s))},\frac{1}{1-\gamma}
\right\}, \qquad \forall s\in \cS.
\label{eq:defn-Wt-s-proof-theta-bound-lcbq}
\end{align}
Applying Lemma~\ref{lemma:useful} (with $W$ taken to be $W_t$) and using the property $\|W_t\|_{\infty}\leq \frac{1}{1-\gamma}$, we reach
\begin{align}
\theta
&\lesssim
\frac{1}{1-\gamma}
\sum_{t=1}^{T}\sum_{s\in\cS}
d_\rho^\star(s)
W_t(s)
+
\frac{1}{ST^4(1-\gamma)}
\sum_{t=1}^{T}\frac{1}{1-\gamma}
\nonumber\\
&\lesssim
\frac{1}{1-\gamma}
\sum_{t=1}^{T}\sum_{s\in\cS}
d_\rho^\star(s)
W_t(s)
+
\frac{1}{T^3(1-\gamma)^2}.
\label{eq:theta-useful-lemma-offlineQ}
\end{align}

To further control $W_t(s)$ in
\eqref{eq:defn-Wt-s-proof-theta-bound-lcbq}, we need a lower bound on
the visitation count $n_t\bigl(s,\pi^\star(s)\bigr)$ in order to bound $\beta_{n_t(s,\pi^\star(s))}$. Towards this end, applying Lemma~\ref{lem:sample-split-infinite} with sample
size $t$ and confidence level $\delta/T$, followed by a union bound over
$t\in[T]$, shows that, with probability at least $1-\delta$, 
\begin{equation}
\max\left\{
n_t(s,a),\,
\frac{4}{3}\iota
\right\}
\geq
\frac{t\myrho(s,a)}{12}
\label{eq:prefix-count-lower-bound-lcbq}
\end{equation}
holds simultaneously
for all $t\in[T]$ and all $(s,a)\in\cS\times\cA$. We now claim that 
\begin{align}
W_t(s)
&\lesssim
\sqrt{
\frac{H\iota}
{t\myrho\big(s,\pi^\star(s)\big)(1-\gamma)^2}
},
\qquad \forall s\in \cS,~t\in [T].
\label{eq:prefix-count-beta-bound-lcbq}
\end{align}
To verify this claim, it suffices to divide into two cases. 
\begin{itemize}
\item 

If
$t\myrho\big(s,\pi^\star(s)\big)> 16\iota$, then combining~\eqref{eq:prefix-count-lower-bound-lcbq} with the definition~\eqref{eq:beta-bar-beta-lcbq} of $\beta_n$ reveals that 
\[
n_t\big(s,\pi^\star(s)\big)
\geq
\frac{t\myrho\big(s,\pi^\star(s)\big)}{12},
\]
from which \eqref{eq:prefix-count-beta-bound-lcbq} follows. 
\item Otherwise, the clipping operation in the definition~\eqref{eq:defn-Wt-s-proof-theta-bound-lcbq} of $W_t$ gives
\[
\min\left\{
\beta_{n_t(s,\pi^\star(s))},\frac{1}{1-\gamma}
\right\}
\leq \frac{1}{1-\gamma},
\]
thereby validating~\eqref{eq:prefix-count-beta-bound-lcbq} in this case. 
\end{itemize}
Substituting \eqref{eq:prefix-count-beta-bound-lcbq} directly into
\eqref{eq:theta-useful-lemma-offlineQ} and using the elementary inequality
$\sum_{t=1}^{T}\frac{1}{\sqrt{t}}\leq 2\sqrt{T}$ 
give
\begin{align}
\theta 
&\lesssim
\sum_{s\in\cS}d_\rho^\star(s)
\sqrt{
\frac{HT\iota}
{(1-\gamma)^4\myrho\big(s,\pi^\star(s)\big)}
}
+
\frac{1}{T^3(1-\gamma)^2} \notag\\
&\overset{\mathrm{(i)}}{\lesssim}
\sum_{s\in\cS}
\sqrt{
\frac{\Cstar d_\rho^\star(s)HT\iota}
{(1-\gamma)^4}
}
+
\frac{1}{T^3(1-\gamma)^2} \notag\\
&\overset{\mathrm{(ii)}}{\lesssim}
\sqrt{
\frac{\Cstar S H T\iota}{(1-\gamma)^4}
}
+
\frac{1}{T^3(1-\gamma)^2}
\overset{\mathrm{(iii)}}{\asymp} 
\sqrt{
\frac{\Cstar S T\iota^{2}}{(1-\gamma)^5}.
}
\label{eq:theta-bound-offlineQ}
\end{align}
Here, (i) uses the definition of the single-policy concentrability
coefficient (see Definition~\ref{assumption:concentrate-infinite-simple}); (ii) invokes the Cauchy-Schwarz inequality, as well as the fact that  $\sum_s d_\rho^\star(s)=1$; and (iii) results from 
$H\asymp \iota/(1-\gamma)$ 
(see \eqref{eq:defn-H-lcbq-stepsize-proof}) and the condition on $T$ in~\eqref{eq:sample-size-lcbq-T-thm}.

\paragraph{Bounding $\psi$ and $\phi$.}
It remains to control the terms $\psi$ and $\phi$. Recall the definition of $d_j(s,a)$ (resp.~$H$) in \eqref{eq:dj-offlineQ} (resp.~\eqref{eq:defn-H-lcbq-stepsize-proof}). Define
\begin{equation}
\widetilde d(s,a)
\coloneqq
\gamma\sum_{j=0}^{\infty}
\left[\gamma\left(1+\frac{1}{H}\right)^3\right]^j d_j(s,a),
\label{eq:dtilde-offlineQ}
\end{equation}
which, according to Lemma~\ref{lemma:useful} (by taking $W$ to be the indicator function corresponding to state $s$), satisfies
\begin{equation}
\widetilde d\bigl(s,\pi^\star(s)\bigr)
\lesssim
\frac{d_\rho^\star(s)}{1-\gamma}
+
\frac{\delta}{ST^4(1-\gamma)},
\quad \forall s\in \cS.
\label{eq:dtilde-bound-offlineQ}
\end{equation}
If one further defines 
\begin{equation}
X_t=(s_t,a_t)\qquad \text{and}\qquad 
X=(s,a)
\end{equation}
and set
\begin{subequations}
\label{eq:ft-gt-def-offlineQ}
\begin{align}
f_t(s,a)
&\coloneqq
\frac{\widetilde d(s,a)}{\myrho(s,a)}
\sum_{i=1}^{n_t(s,a)}
\eta_i^{n_t(s,a)}
P_{s,a}\Delta_{k_i(s,a)-1},
\label{eq:ft-def-offlineQ}\\
g_t(s,a)
&\coloneqq
\frac{\widetilde d(s,a)}{\myrho(s,a)}
P_{s,a}\Delta_{t-1},
\label{eq:gt-def-offlineQ}
\end{align}
\end{subequations}
then summing $\psi_j$ and $\phi_j$ over $j$ and rearranging terms yield the expression
\begin{subequations}
\begin{align}
\psi
&=
\sum_{t=1}^{T}
\left\{
\mathop{\mathbb{E}}\limits_{X\sim \myrho}[f_t(X)]
-
\left(1+\frac{1}{H}\right)f_t(X_t)
\right\},
\label{eq:psi-ft-rewrite-offlineQ}\\
\phi
&=
\left(1+\frac{1}{H}\right)^2
\sum_{t=1}^{T}
\left\{
g_t(X_t)
-
\left(1+\frac{1}{H}\right)
\mathop{\mathbb{E}}\limits_{X\sim \myrho}[g_t(X)]
\right\}. 
\label{eq:phi-gt-rewrite-offlineQ}
\end{align}
\end{subequations}
Before proceeding, we note that for the functions in~\eqref{eq:ft-gt-def-offlineQ}, 
it suffices to focus on the state-action pairs $\big(s,\pi^\star(s)\big)$ ($1\leq s\leq S)$, since 
\begin{align}
\widetilde d(s,a)=0
\qquad \Longrightarrow \qquad f_t(s,a)=g_t(s,a)=0 
\qquad \big(\text{for }a\neq \pi^{\star}(s)\big).
\label{eq:ft-gt-zero-case-lcbq}
\end{align}
Moreover, in view of the learning-rate property~\eqref{eq:sum-eta-n-N} and $0\leq \Delta_t \leq \frac{1}{1-\gamma} 1_S$ (cf.~\eqref{eq:Delta-chi-offlineQ-bound}), one can see from~\eqref{eq:dtilde-bound-offlineQ} and the single-policy concentrability
coefficient (see Definition~\ref{assumption:concentrate-infinite-simple}) that
\begin{equation}
\label{eq:fg-envelope-offlineQ-B-defn}
0\leq
f_t\bigl(s,\pi^\star(s)\bigr), g_t\bigl(s,\pi^\star(s)\bigr)
\lesssim  \frac{\Cstar}
{(1-\gamma)^2}
+
\frac{\delta}
{ST^4(1-\gamma)^2\myrho\big(s,\pi^\star(s)\big)} .
\end{equation}

We first bound $\psi$. Define $f_0(s,a)=0$ for every $(s,a)\in\cS\times\cA$. Adding and subtracting
$f_{t-1}(X_t)$ for $t\geq 2$ gives
\begin{align}
\psi
&=
\underbrace{
\sum_{t=2}^{T}
\left\{
\mathop{\mathbb{E}}\limits_{X\sim \myrho}[f_{t-1}(X)]
-
\left(1+\frac1H\right)f_{t-1}(X_t)
\right\}
}_{\eqqcolon\,\kappa_1}
\arxiv{\notag\\}
\arxiv{&\quad}+
\underbrace{
\sum_{t=2}^{T}
\left\{
\mathop{\mathbb{E}}\limits_{X\sim \myrho}[f_t(X)]
-
\mathop{\mathbb{E}}\limits_{X\sim \myrho}[f_{t-1}(X)]
\right\}
}_{\eqqcolon\,\kappa_2}
\notag\\
&\quad+ \left(1+\frac1H\right)
\underbrace{
 \sum_{t=1}^{T}
\big\{
f_{t-1}(X_t)-f_t(X_t)
\big\} 
}_{\eqqcolon\,\kappa_3}
\arxiv{\notag\\}
\arxiv{&\quad} + 
\mathop{\mathbb{E}}\limits_{X\sim \myrho}[f_1(X)]
.
\label{eq:psi-kappa-decomposition-offlineQ}
\end{align}
We are left with four terms to control, which we tackle separately in the sequel. 
\begin{itemize}

\item 
To facilitate the analysis of $\kappa_1$, we separate well-covered and poorly covered pairs.
Define
\begin{align}
\mathcal{I}
\coloneqq
\left\{
(s,a)\in\cS\times\cA:
\myrho(s,a)\geq \frac{\delta}{ST}
\right\},
\label{eq:defn-I-set-kapp1-lcbq}
\end{align}
and write, for each $t$, 
\[
f_t=f_t^{\mathcal{I}}+f_t^{\mathcal{I}^c},
\qquad
f_t^{\mathcal{J}}(s,a)
\coloneqq f_t(s,a)\ind\{(s,a)\in\mathcal{J}\},\quad \mathcal{J}\in \{\mathcal{I},\mathcal{I}^{\mathrm{c}}\}.
\]
Accordingly, one can use the nonnegativity of $f_t$ to derive and decompose
\begin{align}
\kappa_{1} & \leq\sum_{t=2}^{T}\Bigg\{\mathop{\mathbb{E}}\limits _{X\sim\myrho}\big[f_{t-1}^{\mathcal{I}}(X)+f_{t-1}^{\mathcal{I}^{\mathrm{c}}}(X)\big]-\left(1+\frac{1}{H}\right)f_{t-1}^{\mathcal{I}}(X_{t})\Bigg\}\nonumber \\
 & =\left(1+\frac{1}{H}\right)\sum_{t=2}^{T}\bigg\{\underset{\eqqcolon\,Z_{t}^{\mathcal{I}}}{\underbrace{\mathop{\mathbb{E}}\limits _{X\sim\myrho}\big[f_{t-1}^{\mathcal{I}}(X)\big]-f_{t-1}^{\mathcal{I}}(X_{t})}}\bigg\}\nonumber \\
 & \quad-\frac{1}{H}\underset{\eqqcolon\,A_{\mathcal{I}}}{\underbrace{\sum_{t=2}^{T}\mathop{\mathbb{E}}\limits _{X\sim\myrho}\big[f_{t-1}^{\mathcal{I}}(X)\big]}}+\sum_{t=2}^{T}\mathop{\mathbb{E}}\limits _{X\sim\myrho}\big[f_{t-1}^{\mathcal{I}^{\mathrm{c}}}(X)\big].
 \label{eq:kappa1-upper-bound-zt-aI}
\end{align}
In the sequel, 
let $\mathcal{F}_{t-1}$ denote the information available before
$X_t$ is sampled. Since $f_{t-1}$ is determined by the data observed
up to iteration $t-1$, it is $\mathcal{F}_{t-1}$-measurable. It is also self-evident that 
the summands $\{Z_t^{\mathcal{I}}\}$ in~\eqref{eq:kappa1-upper-bound-zt-aI} form a martingale-difference
sequence. We now cope with each term in \eqref{eq:kappa1-upper-bound-zt-aI} separately.

\begin{itemize}
\item 

We first control the contribution from the well-covered state-action
pairs. By~\eqref{eq:fg-envelope-offlineQ-B-defn} and the definition~\eqref{eq:defn-I-set-kapp1-lcbq} of $\mathcal I$, we have the following magnitude bound: 
\begin{align}
f_{t-1}^{\mathcal I}\big(s,\pi^\star(s)\big)
&\lesssim
\frac{\Cstar}{(1-\gamma)^2}
+
\frac{\delta}
{ST^4(1-\gamma)^2\myrho\big(s,\pi^\star(s)\big)}
\notag\\
&\lesssim
\frac{\Cstar}{(1-\gamma)^2}
+
\frac{1}{T^3(1-\gamma)^2}
\asymp
\frac{\Cstar}{(1-\gamma)^2}.
\label{eq:f-well-covered-bound-offlineQ}
\end{align}
This taken together with the nonnegativity of $f_t$ further reveals that
\begin{align}
\mathbb E\left[
\big(Z_t^{\mathcal I}\big)^2
\mid \mathcal F_{t-1}
\right]
&\leq
\mathbb E\left[
\big(f_{t-1}^{\mathcal I}(X_t)\big)^2
\mid \mathcal F_{t-1}
\right] \arxiv{\notag\\
&}\lesssim
\frac{\Cstar}{(1-\gamma)^2}\,
\mathop{\mathbb{E}}\limits_{X\sim\myrho}
\big[f_{t-1}^{\mathcal I}(X)\big],
\label{eq:kappa1-conditional-variance-bound-offlineQ}
\end{align}
and as a result, 
\begin{align}
\sum_{t=2}^T\mathbb E\left[
\big(Z_t^{\mathcal I}\big)^2
\mid \mathcal F_{t-1}
\right] \lesssim 
\frac{\Cstar}{(1-\gamma)^2} A_{\mathcal I}
.
\label{eq:sum-cond-var-kappa1-lcbq}
\end{align}
According to \eqref{eq:f-well-covered-bound-offlineQ}, this cumulative variance also admits a deterministic upper bound
\[
\eqref{eq:sum-cond-var-kappa1-lcbq}
\lesssim
\frac{(\Cstar)^2 T}{(1-\gamma)^4}.
\]
With the preceding properties and the condition \eqref{eq:sample-size-lcbq-T-thm} on $T$ in place, Freedman's inequality
(Lemma~\ref{lem:Freedman-prelim}) tells us that
\[
\left|
\sum_{t=2}^{T}Z_t^{\mathcal I}
\right|
\lesssim
\sqrt{\frac{\Cstar}{(1-\gamma)^2} A_{\mathcal I}\iota}
+
\frac{\Cstar}{(1-\gamma)^2} \iota
\]
with probability at
least $1-\delta$. This in turn implies that
\begin{align}
\left(1+\frac1H\right)
\sum_{t=2}^{T}Z_t^{\mathcal I}
-\frac1H A_{\mathcal I} 
&\lesssim
\sqrt{\frac{\Cstar}{(1-\gamma)^2}  A_{\mathcal I}\iota}
+
\frac{\Cstar\iota}{(1-\gamma)^2} 
-
\frac1H A_{\mathcal I} 
\notag\\
&\lesssim
\frac1H A_{\mathcal I} + \frac{\Cstar H\iota}{(1-\gamma)^2} 
+
\frac{\Cstar\iota}{(1-\gamma)^2} 
-
\frac1H A_{\mathcal I}
\asymp \frac{\Cstar H\iota}{(1-\gamma)^2}, 
\label{eq:kappa1-well-covered-bound-offlineQ}
\end{align}
where we have applied the AM-GM inequality.

\item 
It remains to control the contribution from $\mathcal{I}^{\mathrm{c}}$. 
Again restricting attention to the pairs $\big(s,\pi^\star(s)\big)$ (in view of~\eqref{eq:psi-kappa-decomposition-offlineQ}), we can derive
\begin{align}
\mathop{\mathbb{E}}\limits_{X\sim\myrho}
\big[f_{t-1}^{\mathcal{I}^{\mathrm{c}}}(X)\big]
&=
\sum_{s:\,
\myrho(s,\pi^\star(s))<\delta/(ST)}
\widetilde d\bigl(s,\pi^\star(s)\bigr)
\sum_{i=1}^{n_{t-1}(s,\pi^\star(s))}
\eta_i^{n_{t-1}(s,\pi^\star(s))}
P_{s,\pi^\star(s)}
\Delta_{k_i(s,\pi^\star(s))-1}
\notag\\
&\leq
\frac{1}{1-\gamma}
\sum_{s:\,
\myrho(s,\pi^\star(s))<\delta/(ST)}
\widetilde d\bigl(s,\pi^\star(s)\bigr).
\label{eq:E-Ic-f-ub-123}
\end{align}
where the last line makes use of the learning-rate property~\eqref{eq:sum-eta-n-N}, $\|\Delta_t\|_{\infty}\leq \frac{1}{1-\gamma}$ (see~\eqref{eq:Delta-chi-offlineQ-bound}), and the fact that $P_{s,\pi^{\star}(s)}$ is a probability vector. 
Using \eqref{eq:dtilde-bound-offlineQ} and the concentrability
condition (see Definition~\ref{assumption:concentrate-infinite-simple}), we obtain
\begin{align}
\sum_{s\in\cS:\,
\myrho(s,\pi^\star(s))<\delta/(ST)}
\widetilde d\big(s,\pi^\star(s)\big)
&\lesssim
\frac{1}{1-\gamma}
\sum_{s\in\cS:\,
\myrho(s,\pi^\star(s))<\delta/(ST)}
\left\{ d_\rho^\star(s)
+
\frac{\delta}{ST^4} \right\}
\notag\\
&\lesssim
\frac{1}{1-\gamma}
\sum_{s\in\cS:\,
\myrho(s,\pi^\star(s))<\delta/(ST)}
\left\{ \Cstar \myrho\big(s,\pi^{\star}(s)\big) 
+
\frac{\delta}{ST^4} \right\}
\notag\\
&\lesssim
\frac{\Cstar\delta}{T(1-\gamma)}
+
\frac{\delta}{T^4(1-\gamma)}.
\end{align}
Substitution into~\eqref{eq:E-Ic-f-ub-123} gives
\begin{align}
\sum_{t=2}^T \mathop{\mathbb{E}}\limits_{X\sim\myrho}
\big[f_{t-1}^{\mathcal{I}^{\mathrm{c}}}(X)\big]
&\lesssim
\frac{\Cstar\delta}{1-\gamma}
+
\frac{\delta}{T^3(1-\gamma)^2}
\lesssim
\frac{\Cstar}{(1-\gamma)^2}.
\label{eq:kappa1-poorly-covered-bound-offlineQ}
\end{align}
\end{itemize}
Combining
\eqref{eq:kappa1-well-covered-bound-offlineQ} and
\eqref{eq:kappa1-poorly-covered-bound-offlineQ}, we conclude that,
with probability at least $1-\delta$,
\begin{align}
\kappa_1
&\lesssim
\frac{\Cstar H\iota}{(1-\gamma)^2}
+
\frac{\Cstar }{(1-\gamma)^2}
\asymp 
\frac{\Cstar \iota^2}{(1-\gamma)^3}
.
\label{eq:kappa1-bound-offlineQ}
\end{align}

\item 
The sum in the second term $\kappa_2$ telescopes, and hence
\begin{align}
\kappa_2
&=
\mathop{\mathbb{E}}\limits_{X\sim\myrho}[f_T(X)]
-
\mathop{\mathbb{E}}\limits_{X\sim\myrho}[f_1(X)]
\arxiv{\notag\\
&}\leq
\mathop{\mathbb{E}}\limits_{X\sim\myrho}[f_T(X)]
\lesssim
\frac{\Cstar}{(1-\gamma)^2}.
\label{eq:kappa2-bound-offlineQ}
\end{align}
Here, the first inequality comes from the nonnegativity of $f_t(\cdot)$, while the last inequality follows by combining \eqref{eq:fg-envelope-offlineQ-B-defn} with   Definition~\ref{assumption:concentrate-infinite-simple}.

\item
Next, we turn to the term $\kappa_3$. For each state $s\in\cS$, let
\[
k_{s,1}<k_{s,2}<\cdots<k_{s,n_s}
\]
denote the iterations at which the state-action pair
$\bigl(s,\pi^\star(s)\bigr)$ is visited, where $n_s$ is the total number
of such visits. For a fixed pair
$\bigl(s,\pi^\star(s)\bigr)$, the quantity
$f_t\bigl(s,\pi^\star(s)\bigr)$ remains unchanged between consecutive
visits, because both $n_t\bigl(s,\pi^\star(s)\bigr)$ and the visit
indices $k_i\bigl(s,\pi^\star(s)\bigr)$ remain unchanged. Hence, the
sum over visit times telescopes. Specifically, using $f_0(s,a)=0$, we
have
\[
\sum_{m=1}^{n_s}
\left[
f_{k_{s,m}-1}\bigl(s,\pi^\star(s)\bigr)
-
f_{k_{s,m}}\bigl(s,\pi^\star(s)\bigr)
\right]
=
-f_{k_{s,n_s}}\bigl(s,\pi^\star(s)\bigr)
\leq 0.
\]
Since $f_t(s,a)=0$ whenever
$a\neq\pi^\star(s)$ (cf.~\eqref{eq:ft-gt-zero-case-lcbq}), summing over all states gives
\begin{equation}
\kappa_3
= \sum_{s\in \cS} \sum_{m=1}^{n_s}
\left[
f_{k_{s,m}-1}\bigl(s,\pi^\star(s)\bigr)
-
f_{k_{s,m}}\bigl(s,\pi^\star(s)\bigr)
\right] \leq 0.
\label{eq:kappa3-bound-offlineQ}
\end{equation}

\item It remains to bound
$\mathop{\mathbb{E}}_{X\sim\myrho}[f_1(X)]$.  
At the first iteration, only the pair $X_1$ has been visited. This means 
$f_1(s,a)=0$ for every $(s,a)\neq X_1$, and according to the definition~\eqref{eq:ft-def-offlineQ},
\[
\mathop{\mathbb{E}}\limits_{X\sim\myrho}\big[f_1(X)\big]
=
\myrho(X_1)f_1(X_1)
=
\widetilde d(X_1)P_{X_1}\Delta_0.
\]
If $a_1\neq \pi^\star(s_1)$, then
$\widetilde d(X_1)=0$ (see~\eqref{eq:ft-gt-zero-case-lcbq}), so the right-hand side vanishes. Otherwise, one has
$X_1=\big(s_1,\pi^\star(s_1)\big)$, and it follows from~\eqref{eq:dtilde-bound-offlineQ} that
\[
\widetilde d(X_1)
\lesssim
\frac{d_\rho^\star(s_1)}{1-\gamma}
+
\frac{\delta}{ST^4(1-\gamma)}.
\]
Combine this with the basic fact $\|\Delta_0\|_\infty\leq 1/(1-\gamma)$ (see~\eqref{eq:Delta-chi-offlineQ-bound}) to yield
\begin{align}
\mathop{\mathbb{E}}\limits_{X\sim\myrho}\big[f_1(X)\big]
&\leq 
\frac{1}{1-\gamma}\widetilde d(X_1)
\lesssim
\frac{1}{1-\gamma}
\Bigg\{
\frac{d_\rho^\star(s_1)}{1-\gamma}
+
\frac{\delta}{ST^4(1-\gamma)}
\Bigg\} \lesssim
\frac{1}{(1-\gamma)^2}.
\label{eq:kappa34-bound-offlineQ}
\end{align}
\end{itemize}
Combine
\eqref{eq:kappa1-bound-offlineQ}-\eqref{eq:kappa34-bound-offlineQ}
to demonstrate that
\begin{align}
\psi
&\lesssim
\frac{\Cstar \iota^2}{(1-\gamma)^3}
+
\frac{\Cstar }{(1-\gamma)^2} + \frac{1}{(1-\gamma)^2}
\asymp \frac{\Cstar \iota^2}{(1-\gamma)^3}
.
\label{eq:psi-bound-offlineQ}
\end{align}

The term $\phi$ can be handled analogously using Freedman's inequality, but more simply, because
$g_t$ is already determined by the data available before $X_t$ is
sampled. We omit the details for brevity.
To conclude, we have
\begin{align}
\psi+\phi
&\lesssim
\frac{\Cstar \iota^2}{(1-\gamma)^3}. 
\label{eq:psi-phi-bound-offlineQ}
\end{align}

%\frac{\Cstar\iota^2}{(1-\gamma)^3} + \frac{\Cstar S\iota}{(1-\gamma)^2}.

\section{Appendix C: proof of auxiliary lemmas in Section~\ref{sec:minimax-lower-bound-offline}}
\label{sec:proof-lower-bound-offline}

\subsection{Proof of Lemma~\ref{lemma:value-policy-LB-inf}} \label{proof:lemma:value-policy-LB-inf}

We first establish a useful symmetry property: 
\begin{align}
\label{eq:value-symmetry-s-lower-bound-lcbq}
V_\theta^\star(s)=V_\theta^\star(s'),
\qquad \forall s,s'>0.
\end{align}
To do so, define
$W(V)\coloneqq \sum_{u\in\cS}\mu(u)V(u)$.
For any $s>0$, the optimal Bellman equation gives
\[
V_\theta^\star(s)
=
1+\gamma\max\big\{
pV_\theta^\star(s)+(1-p)W(V_\theta^\star),\,
qV_\theta^\star(s)+(1-q)W(V_\theta^\star)
\big\}.
\]
This equation has the same form at every positive state: the dependence
on $\theta_s$ only relabels the two actions and disappears under the
maximization. Equivalently, the optimal Bellman operator is invariant
under any permutation of the positive states. Since this Bellman operator is a contraction, it has a unique fixed
point, which must therefore be
invariant under such permutations. This proves
\eqref{eq:value-symmetry-s-lower-bound-lcbq}.

We now analyze the value function at state $s > 0$ under an arbitrary policy $\pi$. The Bellman equation tells us that
\begin{align}
	\arxiv{&} V_\theta^\pi(s) 
	\book{&}= \mathop{\mathbb{E}}\limits_{a\sim\pi(\cdot\mymid s)}
	\bigg[r(s,a)+\gamma \sum_{s'} P_{\theta}(s' \mymid s,a)V_{\theta}^{\pi}(s')\bigg] \nonumber \\
	&\arxiv{~~}\overset{\mathrm{(i)}}{\le} 1 + \gamma \pi(\theta_s \mymid s) \Big[ p V_\theta^\pi(s) + (1-p)\big(C^{-1}V_\theta^\star(s) + (1-C^{-1}) V_\theta^\star(0)\big)\Big] \nonumber \\
	& ~~\qquad + \gamma \pi(1-\theta_s \mymid s) \Big[q V_\theta^\pi(s) + (1-q)\big(C^{-1}V_\theta^\star(s) + (1-C^{-1}) V_\theta^\star(0)\big)\Big] \nonumber \\
	& \arxiv{~~}\overset{\mathrm{(ii)}}{=} 1 + \gamma \Big[ x_{\pi,\theta}(s) V_\theta^\pi(s) + \big(1-x_{\pi,\theta}(s)\big)\big(C^{-1}V_\theta^\star(s) + (1-C^{-1}) V_\theta^\star(0)\big)\Big]. 
	\label{eq:infinite-Value-0}
\end{align}
Here, (i) uses
$V_\theta^\pi \le V_\theta^\star$ and \eqref{eq:value-symmetry-s-lower-bound-lcbq}, while in (ii) we define
\begin{align}
    x_{\pi,\theta}(s)
=
p\,\pi(\theta_s\mid s)
+
q\,\pi(1-\theta_s\mid s)
= q+ (p-q) \pi(\theta_s\mymid s).
	\label{eq:infinite-x-h-proof}
\end{align}
Rearranging terms in \eqref{eq:infinite-Value-0} gives the desired relation \eqref{eq:infinite-value-0-expression}.
Moreover, the equality is attained when we take $\pi = \pi^{\star}$.

We next analyze the value function at state $0$. 
If $\pi(0\mymid 0)=1$, then 
\begin{equation}
	V_{\theta}^{\pi}(0)=\frac{4}{7}+\gamma V_{\theta}^{\pi}(0)
	\qquad\Longrightarrow\qquad 
	V_{\theta}^{\pi}(0)=\frac{4}{7(1-\gamma)}.
    \label{eq:V-theta-pi-0-upper-bound-lb-lcbq-245}
\end{equation}
For any policy $\pi$ with $\pi(0\mymid 0)<1$, we claim that 
\begin{align}
V_{\theta}^{\pi}(0)<\frac{4}{7(1-\gamma)}.
\label{eq:V-theta-pi-0-upper-bound-lb-lcbq}
\end{align}
To verify this, observe that if $V_{\theta}^{\pi}(0)\ge\frac{4}{7(1-\gamma)}$, then we can see from the Bellman equation that
\begin{align}
	V_\theta^{\pi}(0) 
    &=
\pi(0\mid 0)
\left(\frac{4}{7}+\gamma V_\theta^\pi(0)\right)
\arxiv{\notag\\
&\quad}+
\pi(1\mid 0)\gamma
\left[
(2\gamma-1)V_\theta^\pi(0)
+2(1-\gamma)\sum_{s'\in\cS}\mu(s')V_\theta^\pi(s')
\right]
\notag\\ 
	&\le \pi(0 \mymid 0) \left( \frac{4}{7} + \gamma V_\theta^{\pi}(0)\right) + \pi(1 \mymid 0) \gamma \Big[ 2C^{-1} \arxiv{ \nonumber\\
	& \qquad } + \big((2\gamma - 1) + 2(1-\gamma) (1-C^{-1})\big) V_\theta^{\pi}(0)  \Big] \notag\\
	&= \frac{4}{7} \pi(0 \mymid 0)  + \pi(0 \mymid 0)  \gamma V_\theta^{\pi}(0) + \pi(1 \mymid 0) \gamma V_\theta^{\pi}(0) + \pi(1 \mymid 0) \gamma \left[ \frac{2}{C}-  \frac{2(1-\gamma)}{C}V_\theta^{\pi}(0) \right]\nonumber\\
	&\le \frac{4}{7}\pi(0 \mymid 0) + \gamma V_\theta^{\pi}(0) + \frac{6\gamma}{7C} \pi(1 \mymid 0) \nonumber\\
	&< \frac{4}{7} + \gamma V_\theta^{\pi}(0),
	\label{eq:Vpi-expansion-1-infinite-246}
\end{align}
where the first inequality follows from the elementary fact $V_\theta^{\pi}(s) \leq \frac{1}{1-\gamma}$, the second inequality is valid if $V_{\theta}^{\pi}(0)\ge\frac{4}{7(1-\gamma)}$, and the last inequality holds under the condition that $C>3/2$. This relation~\eqref{eq:Vpi-expansion-1-infinite-246}, however, contradicts the assumption that $V_{\theta}^{\pi}(0)\ge\frac{4}{7(1-\gamma)}$, which in turn validates~\eqref{eq:V-theta-pi-0-upper-bound-lb-lcbq}.   
Taken collectively, these two facts~\eqref{eq:V-theta-pi-0-upper-bound-lb-lcbq-245} and \eqref{eq:V-theta-pi-0-upper-bound-lb-lcbq} show that the optimal policy and the optimal value at state $0$ satisfy
\begin{equation}
	\pi^{\star}_{\theta}(0\mymid 0) = 1 
	\qquad \text{and} \qquad 
	V^{\star}_{\theta}(0)= \frac{4}{7(1-\gamma)}.
    \label{eq:Vtheta-0-proof-lower-bound-lcbq}
\end{equation}

It remains to characterize the optimal action at each state $s>0$. Define
\begin{subequations}
\label{eq:proof-defn-Bs-gs-lcbq}
\begin{align}
B_s&\coloneqq
C^{-1}V_{\theta}^{\star}(s)
+(1-C^{-1})V_{\theta}^{\star}(0),\\
g_s(x)&\coloneqq
B_s+
\frac{1-(1-\gamma)B_s}{1-\gamma x}.
\end{align}
\end{subequations}
For every policy $\pi$, it follows from~\eqref{eq:infinite-value-0-expression} that
\begin{align}
V_{\theta}^{\pi}(s)
\leq
g_s\bigl(x_{\pi,\theta}(s)\bigr),
\label{eq:V-pi-s-ub-in-terms-of-gs}
\end{align}
while equality is attained by $\pi=\pi_{\theta}^{\star}$. Moreover, it is seen from~\eqref{eq:Vtheta-0-proof-lower-bound-lcbq} and $C>1$ that
$B_s<\frac{1}{1-\gamma}$,
and therefore
\[
g_s'(x)
=
\frac{\gamma\bigl[1-(1-\gamma)B_s\bigr]}
{(1-\gamma x)^2}
>0.
\]
Finally, from the definition~\eqref{eq:infinite-x-h-proof}, we know that $x_{\pi,\theta}(s)$ 
is strictly increasing in $\pi(\theta_s\mymid s)$ because $p>q$.
Consequently, the upper bound~\eqref{eq:infinite-value-0-expression} on $V_{\theta}^{\pi}(s)$ is maximized
when $\pi(\theta_s\mymid s)=1$, which together with the fact $x_{\pi,\theta}(s)\leq p$ indicates that, for every policy $\pi$,
\[
V_{\theta}^{\pi}(s)
\leq
g_s\bigl(x_{\pi,\theta}(s)\bigr)
\leq
g_s(p).
\]
Note that the policy that chooses action $\theta_s$ deterministically at state $s$ satisfies
$x_{\pi,\theta}(s)=p$, and the optimal policy attains equality in the
first inequality. Hence, we arrive at
\begin{equation}
V_{\theta}^{\star}(s)=g_s(p),
\label{eq:V-s-star-equal-gsp-lcbq}
\end{equation}
which is possible only if
$\pi_{\theta}^{\star}(\theta_s\mymid s)=1$.

\subsection{Proof of Lemma~\ref{lem:transition-KL-bound-infinite}}
\label{sec:proof:lem:transition-KL-bound-infinite}

If $\theta_s=\widetilde{\theta}_s$, then the transition kernels
coincide at $(s,a)$ for both actions, and the first claim follows
immediately.

Now suppose that $\theta_s\neq\widetilde{\theta}_s$. Without loss of
generality, assume that $\theta_s=0$ and
$\widetilde{\theta}_s=1$. Let $s'$ denote the next state, sampled according to
$P_\theta(\cdot\mid s,a)$ or $P_{\widetilde{\theta}}(\cdot\mid s,a)$,
depending on the underlying hypothesis. Define
\[
\alpha\coloneqq 1-\frac{1}{C(S-1)}.
\]
We first consider action $a=0$. Under the two hypotheses, the
probabilities of the event $\{s'=s\}$ are
\[
u\coloneqq
\alpha p+\frac{1}{C(S-1)}
\qquad\text{and}\qquad
v\coloneqq
\alpha q+\frac{1}{C(S-1)},
\]
respectively. Conditioned on $s'\neq s$, the two transition kernels induce the same
conditional distribution of $s'$. Therefore, one has
\[
\mathsf{KL}\big(
P_\theta(\cdot\mymid s,0)
\,\big\|\,
P_{\widetilde{\theta}}(\cdot\mymid s,0)
\big)
=
\mathsf{KL}\big(\mathsf{Bern}(u)\,\|\,\mathsf{Bern}(v)\big);
\]
Applying
Lemma~\ref{lem:KL-quadratic-lower-bound} then yields
\begin{align}
\mathsf{KL}\big(
P_\theta(\cdot\mymid s,0)
\,\big\|\,
P_{\widetilde{\theta}}(\cdot\mymid s,0)
\big)
&\leq
\frac{(u-v)^2}{u(1-v)}
=
\frac{\alpha^2(p-q)^2}
{\left(\alpha p+\frac{1}{C(S-1)}\right)\alpha(1-q)}
\notag\\
&\leq
\frac{\alpha^2(p-q)^2}{p\,\alpha(1-p)}
\leq
\frac{(p-q)^2}{p(1-p)},
\label{eq:KL-single-action-bound-infinite-proof}
\end{align}
where we have used $u\geq p$, $1-q\geq 1-p$, and $\alpha\leq 1$.
For action $a=1$, the roles of $p$ and $q$ are interchanged. Repeating
the same argument gives
\begin{align}
\mathsf{KL}\big(
P_\theta(\cdot\mymid s,1)
\,\big\|\,
P_{\widetilde{\theta}}(\cdot\mymid s,1)
\big)
&\leq
\frac{(p-q)^2}{q(1-p)}
\leq
\frac{2(p-q)^2}{1-p},
\label{eq:KL-single-action-one-bound-infinite-proof}
\end{align}
where the last inequality uses $q\geq 1/2$.

From our parameter choice~\eqref{eq:infinite-p-q}, we have
\[
p-q=\frac{448(1-\gamma)^2\varepsilon}{\gamma}.
\]
Moreover, the assumption
\eqref{eq:infinite-epsilon-assumption} implies
\[
p\geq\gamma,
\qquad
1-p
=
1-\gamma-\frac{224(1-\gamma)^2\varepsilon}{\gamma}
\geq \frac{1-\gamma}{2}.
\]
Consequently, for each $a\in\{0,1\}$, one has
\begin{align}
\mathsf{KL}\big(
P_\theta(\cdot\mymid s,a)
\,\big\|\,
P_{\widetilde{\theta}}(\cdot\mymid s,a)
\big)
&\leq
c_3(1-\gamma)^3\varepsilon^2
\end{align}
for some universal constant $c_3>0$, 
provided that $\gamma\geq 2/3$. 

\subsection{Proof of auxiliary properties}
\label{sec:proof:eq:db-Cstar-inf-LB}

\paragraph{Proof of \eqref{eq:db-sa-inf-LB}.}
With a slight abuse of notation, consider an MDP trajectory $\{(s_t,a_t)\}_{t\geq 0}$ initialized from $s_{0}\sim \rhob = \mu$. 
Under the behavior policy $\pib$ in~\eqref{eq:defn-rhob-pib-lcb-lb}, we have
\begin{align*}
	\mathbb{P}\left\{ s_{1}=0\right\}  
	& =\mu(0)\left\{ \frac{1}{2} P_{\theta}(0\mymid0,0)+\frac{1}{2} P_{\theta}(0\mymid0,1) \right\}  \arxiv{ \\
	&\qquad } +\sum_{s=1}^{S-1}\mu(s)\left\{ \frac{1}{2} P_{\theta}(0\mymid s,0)+\frac{1}{2}P_{\theta}(0\mymid s,1)\right\} \\
	& =\mu(0)\left\{ \gamma+(1-\gamma)\mu(0)\right\}  +\sum_{s=1}^{S-1}\mu(s)(1-\gamma)\mu(0) \\
	& =\mu(0),
\end{align*}
where the last identity follows from $\sum_{s=0}^{S-1}\mu(s)=1$. 
According to the symmetry, one obtains $\mathbb{P}\{s_{1}=s\} = \frac{1-\mu(0)}{S-1} = \mu(s)$, and hence $s_1\sim \mu$. 
Repeating the same argument shows that $\mu$ is a stationary distribution of the state process under $\pib$, so that $s_t \sim \mu$ for all $t\geq 0$. 
Consequently,
\[
\myrho(s) = (1-\gamma) \mathbb{E}\left[\sum_{t=0}^\infty \gamma^t \mathbb{P}\big(s_t=s\mid s_{0}\sim \rhob;\pib\big)\right]
= \mu(s), \qquad \forall s\in\cS.
\]
In addition, since $\pib(a\mymid s)=1/2$ for all $(s,a)\in\cS\times\cA$, we have
\begin{align}
	\myrho(s,a)=\myrho(s) \pib(a\mymid s)= \mu(s)/2.
\end{align}

\paragraph{Proof of \eqref{eq:expression-Cstar-LB-inf}.}
Consider the MDP $\mathcal{M}_\theta$, whose optimal policy $\pi_\theta^{\star}$ satisfies $\pi_\theta^{\star}(\theta_s \mymid s) = 1$ for $s>0$ by Lemma~\ref{lemma:value-policy-LB-inf}. 
With a slight abuse of notation, let $\{(s_t,a_t)\}_{t\geq 0}$ denote a trajectory generated under the policy $\pi_{\theta}^{\star}$, namely, $a_t\sim \pi_{\theta}^{\star}(\cdot \mymid s_t)$. 
Then it is seen that
\begin{align*}
d^{\star}(0) &= (1-\gamma)\sum_{t=0}^\infty \gamma^t \mathbb{P}\big(s_t=0\mid s_{0}\sim \rho;\pi_{\theta}^{\star}\big) \\
&\le (1-\gamma)\sum_{t=0}^\infty \gamma^t(1-\gamma^t) < \frac{1}{2}.
\end{align*}
Here, the last line is valid since
\begin{align*}
\arxiv{&} \mathbb{P}\big(s_t=0\mid s_{0}\sim \rho;\pi_{\theta}^{\star}\big)\arxiv{ \notag\\} 
&\arxiv{\quad}= \mathbb{P}\big(s_{t-1}=0\mid s_{0}\sim \rho;\pi_{\theta}^{\star}\big) + (1-C^{-1})(1-p)\mathbb{P}\big(s_{t-1}>0\mid s_{0}\sim \rho;\pi_{\theta}^{\star}\big) \\
&\arxiv{\quad}\le \gamma\mathbb{P}\big(s_{t-1}=0\mid s_{0}\sim \rho;\pi_{\theta}^{\star}\big) + 1-\gamma \\
&\arxiv{\quad}\le (1-\gamma)\sum_{k=0}^{t-1}\gamma^k = 1-\gamma^t, 
\end{align*}
% %
where the first inequality uses
$(1-C^{-1})(1-p)\leq 1-p\leq 1-\gamma$ (since $C>1$ and $p\geq\gamma$).  
By symmetry, combining the above bound with \eqref{eq:db-sa-inf-LB} yields
\begin{align*}
\frac{d^{\star}(s, \theta_s)}{\myrho(s,\theta_s)} 
= \frac{2\big(1-d^{\star}(0)\big)}{(S-1)\mu(s)} \in (C, 2C),
\quad\forall s > 0.
\end{align*}
In addition, it is straightforward to verify that 
\[
\frac{d^{\star}(0,0)}{\myrho(0,0)}
\leq \frac{1}{\mu(0)}
\leq 2C,
\]
provided that $C>3/2$. 
Putting the preceding bounds together, we establish the advertised result.

\paragraph{Proof of~\eqref{eq:infinite-delta-key-auxiliary}.}
Recall the definition of $x_{\pi,\theta}(\cdot)$ in~\eqref{eq:defn-x-pi-theta-s-lb-proof-lcbq}, and  the definition of $B_s$ and $g_s(\cdot)$ in \eqref{eq:proof-defn-Bs-gs-lcbq}. We have already proven in~\eqref{eq:V-pi-s-ub-in-terms-of-gs}  and \eqref{eq:V-s-star-equal-gsp-lcbq} that
$$
V_\theta^\star(s)=g_s(p),\qquad V_\theta^\pi(s)
\leq g_s\bigl(x_{\pi,\theta}(s)\bigr),
$$
thus implying that
\begin{align*}
V_\theta^\star(s)-V_\theta^\pi(s)
&\geq
g_s(p)-g_s\bigl(x_{\pi,\theta}(s)\bigr) \arxiv{\\
&}=
\frac{\bigl[1-(1-\gamma)B_s\bigr]\,
\gamma\bigl(p-x_{\pi,\theta}(s)\bigr)}
{(1-\gamma p)\bigl(1-\gamma x_{\pi,\theta}(s)\bigr)}.
\end{align*}
Since $V_\theta^\star(s)\leq \frac{1}{1-\gamma}$ and
$V_\theta^\star(0)=\frac{4}{7(1-\gamma)}$ (see Lemma~\ref{lemma:value-policy-LB-inf}), one has
\[
(1-\gamma)B_s
\leq
C^{-1}+\frac47(1-C^{-1})
=
\frac47+\frac{3}{7C}
\leq \frac{11}{14},
\]
with the proviso that $C\geq 2$. Moreover, recall from~\eqref{eq:defn-x-pi-theta-s-lb-proof-lcbq} that  $q\leq x_{\pi,\theta}(s)\leq p$, and hence
\[
(1-\gamma p)\bigl(1-\gamma x_{\pi,\theta}(s)\bigr)
\leq
(1-\gamma p)(1-\gamma q)
\leq
(1-\gamma^2)^2
\leq 4(1-\gamma)^2,
\]
where we use the definition \eqref{eq:infinite-p-q} of $p$ and $q$. 
Finally, it follows from \eqref{eq:infinite-x-h-proof} and \eqref{eq:infinite-p-q} that
\[
\gamma\bigl(p-x_{\pi,\theta}(s)\bigr)
=
448(1-\gamma)^2\varepsilon
\bigl(1-\pi(\theta_s\mymid s)\bigr).
\]
Combining the preceding inequalities yields
\[
V_\theta^\star(s)-V_\theta^\pi(s)
\geq
24\varepsilon
\bigl(1-\pi(\theta_s\mymid s)\bigr).
\]
This concludes the proof of 
\eqref{eq:infinite-delta-key-auxiliary}.

\section{Notes}

While the core ideas underlying batch RL can be traced back more than two decades \citep{ernst2005tree,lange2012batch}, this subfield of RL has witnessed a remarkable resurgence since around 2020, largely under the banner of offline RL \citep{kumar2020conservative,levine2020offline,agarwal2020optimistic}. This renewed interest has catalyzed a substantial body of work across computer science, statistics, operations research, and related disciplines on the statistical and algorithmic foundations of offline RL,   motivated by the growing range of applications in which real-time data collection is expensive, safety-critical, or otherwise constrained. We refer the interested reader to \citet{jin2021pessimism,rashidinejad2021bridging,xie2021policy,yin2021optimal,ma2021conservative,uehara2021pessimistic,uehara2021finite,ren2021nearly,ma2022offline,sonabend2020expert,yin2021towards,xie2021bellman,yin2021near,shi2022pessimistic,lyu2022mildly,yan2022efficacy,zhan2022offline,zanette2022bellman,zhou2023offline,ye2023corruption,bhardwaj2023adversarial,li2024settling,yin2022near,yin2021nearVR,xu2024provably,zhu2024robust,zhou2024federated,jin2025policy,jiang2025offline} for some examples of recent developments, many of which pursue perspectives that are complementary to the ones presented in this chapter.

A recurring theme in offline RL is to identify the fundamental conditions under which a near-optimal policy can be learned from a historical dataset. Early analyses based on concepts of the concentrability coefficients typically require the offline data distribution to provide adequate coverage of all admissible policies, or equivalently, or equivalently, of all reachable state-action pairs  \citep{munos2007performance,farahmand2010error,le2019batch,fan2019theoretical,chen2019information,yin2021near,yin2021nearVR}. This condition, commonly referred to as the \emph{all-policy concentrability} requirement, is often overly restrictive in practice, particularly in high-dimensional settings where adequate coverage of the entire policy space is difficult to guarantee.  

Subsequent work sought to relax this stringent requirement by arguing that the offline dataset may only need to provide sufficient coverage of the state-action pairs visited by the target policy \citep{liu2020provably}. Building on this insight, \citet{jin2021pessimism,rashidinejad2021bridging} developed offline RL algorithms and analyses based on the notion of \emph{single-policy concentrability}, whereby the coverage requirement is imposed only along the state-action distribution induced by the target policy, as discussed in this chapter. For tabular MDPs, the full-range minimax sample complexity of offline RL under single-policy concentrability was first characterized by \citet{li2024settling}, who showed the optimality of model-based algorithms for both discounted infinite-horizon MDPs and finite-horizon nonstationary MDPs without imposing any burn-in cost. In addition, \citet{yin2021towards,yin2022near} established instance-dependent statistical guarantees for pessimistic model-based algorithms, revealing how the statistical complexity of offline RL is governed by the variance structure of the problem instance (albeit at the cost of a large burn-in sample size requirement). The more general offline RL setting with function approximation has received much attention as well; see, e.g.,  \citet{jin2021pessimism,zanette2021provable,yin2022near,xie2021bellman,di2024pessimistic}. 
More broadly, the pessimism idea has since found applications in other statistical tasks, such as learning optimal individualized decision rules \citep{jin2025policy} and learning optimal dynamic treatment regimes \citep{zhou2023optimizing}.

%From an algorithmic perspective, so far the chapter focuses on imposing the pessimism principle via penalizing the estimated values of under-covered state-action pairs with the aid of suitably constructed lower confidence bounds. Theoretical guarantees on model-based methods have been developed in \citet{rashidinejad2021bridging,jin2021pessimism,xie2021policy,yin2021towards} for tabular MDPs, and \citet{yin2022near,uehara2022representation} for MDPs with function approximation. In parallel, model-free methods are investigated in \citet{shi2022pessimistic,yan2022efficacy} for tabular MDPs, and \citet{zanette2021provable} for function approximation settings.

From an algorithmic perspective, the methods discussed in this chapter primarily implement the pessimism principle by penalizing the estimated values of under-covered state-action pairs with the aid of uncertainty quantification.  Alternative approaches have also been proposed that enforce pessimism through an adversarial learning perspective. For example, a line of work \citep{xie2021bellman,uehara2021pessimistic,zhan2022offline,wang2024achieving,panaganti2025bridging,cen2025vpo} has suggested searching for the \emph{least favorable model} ---that is, the model yielding the lowest value function --- among all models that are statistically consistent with the offline data, typically through regularized or constrained optimization formulations. 

While the discussion thus far has focused on learning a near-optimal policy from offline data, another major and closely related line of research is \emph{off-policy evaluation} (OPE), whose goal is to estimate the value of a target policy using data generated by a potentially different behavior policy. OPE has evolved into a rich research area in its own right, with deep connections to statistics, causal inference, semiparametric efficiency theory, and related fields. A comprehensive treatment of OPE falls beyond the scope of this chapter; we refer the interested reader to \citet{uehara2022review,li2014minimax,jiang2016doubly,thomas2016data,xie2019towards,uehara2020minimax,yang2020off,jiang2020minimax,duan2020minimax,liao2021off,ren2021nearly,yin2020asymptotically,yin2021near,kallus2022doubly,zhan2021off,xu2022quantile,hu2023off,sonabend2023semi,mehrabi2024off,shi2024off,bian2025off,qi2025distributional}, as well as the references therein, for detailed discussions.

%% file: chapters/cousins.tex
\chapter{Extensions}
\label{chapter:extensions}

 Thus far, we have focused on three canonical --- and arguably the most widely studied --- formulations of single-agent RL. Many modern applications, however, depart from these classical settings in important ways, giving rise to a rich collection of emerging frameworks that await further investigation. In this chapter, we briefly survey a few representative extensions along these lines, including distributionally robust RL, multi-agent RL, federated RL, statistical inference in RL, and RL with human feedback. Our treatment is necessarily brief and offers only a bird’s-eye view of these rapidly evolving topics. Rather than aiming for an exhaustive account, the purpose is to shed light on how the non-asymptotic perspective developed throughout the previous chapters continues to inform both the statistical principles and the algorithm designs in settings that extend beyond the basic MDP framework.

\input{chapters/robust_RL}

\input{chapters/marl.tex}

\input{chapters/federated_RL}

\input{chapters/inference.tex}

\input{chapters/RLHF}

\section{Notes}

\paragraph{Distributionally robust RL.} 
Beyond the discounted infinite-horizon setting \citep{panaganti2021sample,wang2023finite,shi2024distributionally,shi2026curious,jiao2026provably}, distributionally robust RL has also been studied for finite-horizon MDPs  \citep{xu2023improved,dong2022online,lu2024distributionally}, average-reward MDPs  \citep{chen2025sample,yang2026robust}, constrained RL \citep{zhang2024distributionally}, and multi-agent RL \citep{shi2025breaking,shi2024sample,qu2026distributionally}. 
Several other forms of uncertainty sets have been studied in the literature as well. For instance,
\citet{wang2021online} considered $R$-contamination uncertain sets and developed a provably robust Q-learning algorithm for the online setting, whereas \citet{clavier2024near} studied the $\ell_p$-based uncertainty sets under the generative model. Another widely studied class of uncertainty sets is based on KL divergence. In this context,  \citet{yang2021towards,panaganti2022sample,zhou2021finite,shi2022distributionally,xu2023improved,wang2023finite,blanchet2023double,wang2023sample,liang2023single,wang2024distributionally} investigated the statistical performance of various algorithms under several sampling mechanisms. 

Last but not least, distributionally robust RL has also been extended beyond the tabular setting. We refer the interested reader to \citet{badrinath2021robust,ramesh2023distributionally,panaganti2022robust,wang2024sample,liu2024distributionally,mandal2025finite} for recent developments involving function approximation.

  \paragraph{Markov games.}
Moving beyond the generative-model setting, a recent line of work has focused on the design and analysis of a class of online MARL algorithms known as \emph{V-learning}, which leverages online adversarial learning techniques and adaptive sampling to mitigate the curse of multi-agents  \citep{mao2022provably,song2021can,jin2021v}. We refer the interested reader to these papers for algorithmic details. Importantly, V-learning was the first framework to reveal the following striking statistical phenomenon: what ultimately matters in the sample complexity is the largest individual action space $\max_{1\leq i\leq m} A_i$, as opposed to the size of the joint action space $\prod_{i=1}^m A_i$. 
This insight leads to substantial improvements in data efficiency and helps overcome one of the principal obstacles in multi-agent RL. 

Despite this significant progress, the current theoretical guarantees for V-learning remain suboptimal in their horizon dependence. More broadly, the fundamental statistical limits of online learning in Markov games remain unsettled and appear to be considerably more subtle than their single-agent counterparts. For instance, the optimal horizon dependence is still unknown: it remains unclear whether the minimax sample complexity of equilibrium learning should exhibit the same $H^3$ dependence (in terms of the number of episodes required) that arises in single-agent online RL, or whether the multi-agent setting gives rise to fundamentally different horizon scaling. Progress on such questions has been hindered both by the challenge of developing sharper upper-bound analyses and by the difficulty of constructing tight lower bounds.

%Note, however, the state-of-the-art theoretical guarantees for V-learning suffer from suboptimal horizon dependency. In fact, it remains unclear how to simultaneously what the fundamental statistical limits would be in online Markov games; for instance, it is not clear what the optimal horizon scaling is and it is also challenging to construct tight lower bounds.  

%which remains suboptimal in terms of the horizon dependency, but considers the more challenging online setting. 
		%As a drawback of these works, the policy returned by V-learning is non-Markovian, an issue that has been recently addressed by \citet{daskalakis2022complexity} at the price of a much higher sample complexity.  
		%It is worth emphasizing that all these works assume the online exploration setting as opposed to the scenario with a generative model. 

In addition to online Markov games and the generative-model setting, the curse of multi-agency also manifests itself in offline MARL, where it has been addressed in work such as \citet{cui2022offline,yan2022model}. Moreover, while our discussion has focused primarily on competitive Markov games, several other important variants have received considerable attention, including Markov potential games \citep{song2021can,ding2022independent} and robust Markov games \citep{shi2024sample,shi2025breaking,jiao2024minimax,blanchet2023double}. 
The sample complexity of MARL under function approximation has likewise been investigated, with particular emphasis on mitigating the curse of multi-agents; see, for example,  \citet{dai2024refined,cui2023breaking,wang2023breaking}. 
Finally, \citet{yang2025incentivize,liu2023maximize} developed efficient exploration mechanisms for online MARL that avoid explicit uncertainty estimation while still achieving strong performance.

\paragraph{Federated RL.} 
A growing body of recent work has investigated the sample complexity of federated value-based algorithms in various settings, such as online RL \citep{zheng2024federated,zhang2025gap,zhang2026regret}, offline RL \citep{woo2024federated}, asynchronous Q-learning \citep{woo2023blessing}, average-reward RL \citep{jiao2026sample}, and heterogeneous RL \citep{zhang2024finite,wang2025on,yang2024federated}. Federated variants of model-based RL algorithms have also been studied; see, for example,  \citet{labbi2025federated,zhou2024federated}. In addition to policy learning, 
several recent studies \citep{liu2023distributed,tian2024one} established that a single round of communication is sufficient to achieve linear speedup of TD learning for {\em policy evaluation}. Note that these results do not contradict the communication lower bounds established by \citet{salgia2024sample}, which concern Q-learning for \emph{policy learning}. 

%The latter is more involved due to the nonlinearity of the Bellman optimality operator. 
%Specifically, if the operator whose fixed point is to be found is linear in the decision variable (e.g., the value function in TD learning) then the fixed point update only induces a variance term corresponding to the noise. However, if the operator is non-linear, then in addition to the variance term, we also obtain a \emph{bias} term in the fixed point update. While the variance term can be controlled with one-shot averaging, more frequent communication is necessary to ensure that the bias term is small enough. 

%Moreover, we mention a few  policy gradient algorithms for federated RL, including but not limited to \citet{lan2025asynchronous,labbi2025global}. \citet{yang2024federated} proposed federated natural policy gradient  and actor-critic algorithms with global convergence guarantees for multi-task RL when the agents communicate over a fixed graph topology without a central server.

\paragraph{Statistical inference in RL.}
Moving beyond the i.i.d.~data setting \eqref{eq:iid-samples} discussed in this chapter and in \citet{mou2020linear,wu2024statistical, bhandari2018finite,samsonov2024gaussian}, 
there has been substantial recent progress on statistical analysis and inference for Markov-chain-based RL algorithms driven by Markovian sample trajectories; see, e.g., \cite{bhandari2018finite,mou2020linear,fan2025statistical,li2021sample,ramprasad2023online,samsonov2024improved,li2023online,samsonov2024gaussian,srikant2025rates}.
The state-of-the-art finite-sample distributional guarantees for Polyak-Ruppert averaged TD learning (e.g., finite-sample Berry-Esseen bounds) have been developed by   \cite{wu2025uncertainty,srikant2025rates}.

Beyond settings in which samples are generated under a fixed policy, it is also important to study statistical inference in online environments, where data are collected adaptively through sequential interaction with the environment \citep{zhang2020inference,chen2021statistical,shen2024doubly}. In such scenarios, adaptive data collection induces intricate dependencies across observations, often invalidating classical inferential techniques developed for independent data. Motivated by these challenges, a recent line of work has investigated statistical inference in bandit problems with adaptively collected data \citep{kalvit2021closerlookworstcasebehavior,zhang2020inference,dimakopoulou2021online,hadad2021confidence, khamaru2024inference, han2024ucb, halder2025stable,guo2025statistical,fan2025statistical,han2025online,chang2026online}. 
For instance, one important insight emerging from this literature is that UCB-type algorithms can serve not only as efficient exploration strategies but also as mechanisms for valid statistical inference. Extending these ideas beyond bandits and developing principled inferential paradigms for online RL remains an important direction for future research.

\paragraph{RLHF.}
RLHF has been attracting considerable attention in recent years, and, in particular, the connection to the classical statistical ranking formulation has been explored \citep{zhu2023principled}. Besides the Bradley-Terry model, other ranking models have also been considered for RLHF, including but not limited to the Mallows model \citep{chen2025mallowspo}. Concurrent with \VPO, several papers have proposed similar regularization techniques to impose the principles of optimism and pessimism, respectively, for online and offline RLHF, e.g., \citet{xie2024exploratory,zhang2025selfexploring,liu2024provably}. Another popular approach to RLHF is via game-theoretic formulations, e.g., \citet{azar2024general}, which are less prone to reward hacking.
Finally, we refer the interested reader to \citet{liu2026reinforcement} for a recent survey of RLHF from a statistical perspective.

%The principles of optimism and pessimism are typically implemented via constructing confidence intervals, which have been demonstrated to be provably efficient in tabular settings as demonstrated in earlier chapters \citep{jin2018q,shi2022pessimistic}. Yet, these approaches have had limited success in conjunction with deep learning architectures \citep{gawlikowski2021survey}, and many empirical heuristics in turn lack theoretical validation \citep{kumar2020conservative}. Notwithstanding, alternative regularization schemes are developed for general function approximation settings with theoretical guarantees, such as Bellman-consistent pessimism for offline RL \citep{xie2021bellman}, and reward-biased exploration for online RL \citep{mete2021reward, liu2023maximize} and Markov games \citep{liu2023maximize,yang2025incentivize}. 
%VPO draws inspiration from reward-biased exploration \citep{kumar1982new,liu2019exploration,liu2023maximize,hung2021reward,mete2021reward,yang2025exploration} in the standard online RL literature, which has been further extended to game-theoretic formulations \citep{yang2025incentivize}.%, but significantly broadens its scope to the offline setting and RLHF for the first time.   

%% file: chapters/robust_RL.tex
\section{Distributionally robust RL}
\label{sec:DRL}

Despite the extensive study of standard RL, its practical deployment is often hindered by the sim-to-real gap and/or various forms of environmental uncertainty \citep{bertsimas2019adaptive}. For instance, a policy learned in an ideal, nominal environment may fail catastrophically when deployed in a setting subject to even small changes in task objectives or adversarial perturbations to the system dynamics  \citep{zhang2020robust,klopp2017robust,mahmood2018benchmarking}. As a result, robustness has emerged as a critical objective alongside reward maximization, particularly in high-stakes applications such as robotics, autonomous driving, clinical trials, and financial investments.

To address these challenges, 
distributionally robust RL \citep{iyengar2005robust,nilim2005robust,xu2012distributionally,bauerle2022distributionally,cai2016robust} has emerged as a natural yet versatile framework, drawing inspiration and insight from the broader literature on distributionally robust optimization and supervised learning \citep{rahimian2019distributionally,gao2020finite,bertsimas2018data,duchi2018learning,blanchet2019quantifying,chen2019distributionally,lam2019recovering}. The objective is to learn a policy that maintains strong performance even when the deployed environment deviates from the nominal one used for training, thereby enhancing robustness to environmental uncertainty and distributional shifts. In what follows, we briefly discuss, through the lens of sample complexity, whether and how the incorporation of distributional robustness affects the statistical bottlenecks in learning a desirable policy.

\subsection{Model and preliminaries}

In contrast to standard RL, which seeks to learn an optimal policy for a single MDP using samples generated from that MDP, distributionally robust RL aims to learn a policy that remains effective in the presence of model uncertainty. At a high level, using the same collection of samples, it seeks to optimize worst-case performance over a prescribed uncertainty set of transition kernels centered around the nominal kernel. We make this formulation more precise below. For expositional simplicity, we focus on the discounted infinite-horizon setting.

\paragraph{Distributionally robust MDPs (RMDPs).}
We now introduce the distributionally robust MDP (RMDP) tailored to the discounted infinite-horizon setting. Specifically, an RMDP is formally defined by a tuple
$$\cM_{\mathsf{rob}} = \left\{\cS,\cA, \gamma, \cU^{\ror}(P^\no), r \right\}, $$ 
where the state space $\cS=\{1,\cdots,S\}$, the action space $\cA=\{1,\cdots,A\}$, the discount factor $0<\gamma<1$, and the reward function $r: \cS\times \cA \rightarrow [0,1]$ remain identical to those in a standard MDP. The key distinction lies in the transition kernel:\footnote{Although uncertainty in the reward function can also be incorporated into the framework, we omit it here for simplicity and focus exclusively on uncertainty in the transition kernel, which constitutes the primary technical challenge.} rather than assuming a fixed transition kernel $P$, the actual transition kernel can deviate arbitrarily within a prescribed uncertainty set   $\cU^{\ror}(P^\no)$ centered around a {\em nominal} kernel $P^\no: \cS\times\cA \rightarrow \Delta(\cS)$. 

Compared with standard MDPs, RMDPs provide a more flexible modeling framework by allowing one to specify both the shape and size of the uncertainty set around the nominal transition kernel. To ensure computational tractability for solving RMDPs, however, the uncertainty set is often chosen to possess additional structure.  A common structure in the literature is that the uncertainty set decomposes into a product of independent local uncertainty subsets over states or state-action pairs, known respectively as \emph{$s$-rectangularity} and \emph{$(s,a)$-rectangularity} \citep{zhou2021finite,wiesemann2013robust,wang2023foundation}.

For ease of exposition, we focus on the $(s,a)$-rectangular setting. Specifically, let $\rho: \Delta(\cS) \times \Delta(\cS) \rightarrow \mathbb{R}_{\geq 0}$ 
be a distance-like metric,  and let $\ror>0$ represent a prescribed radius that controls the level of uncertainty. The uncertainty set is then defined as
\begin{subequations}
\label{eq:general-infinite-P-both}
\begin{align}\label{eq:general-infinite-P}
	\cU^{\ror}(P^\no) \defn \otimes_{s,a} \; \cU^{\ror}(P^{\no}_{s,a}),
\end{align}
where for each $(s,a)\in \cS\times \cA$, the local uncertainty set is given by
\begin{align}
	\cU^{\ror}(P^\no_{s,a}) \defn \left\{ P_{s,a} \in \Delta (\cS): \rho \big(P_{s,a}, P^{\nominal}_{s,a}\big) \leq \ror \right\},
\end{align}
\end{subequations}
with $P_{s,a}$ and $P_{s,a}^{\nominal}$ representing the row vectors of $P$ and $P^{\nominal}$ associated with the state-action pair $(s,a)$, respectively. 
In other words, the uncertainty is imposed in a decoupled manner across state-action pairs.

In particular, two popular instances of \eqref{eq:general-infinite-P-both} arise when the metric $\rho$ is chosen to be the total-variation (TV) distance, namely $\rho(\cdot,\cdot)=\mathsf{TV}(\cdot,\cdot)$ (see \eqref{eq:TV}), or the $\chi^2$ divergence, namely $\rho(\cdot,\cdot)=\chi^2(\cdot \,\|\, \cdot)$ (see \eqref{eq:defn-chi2}). These two choices will serve as the primary focus of this section. As we shall see shortly, these two choices of divergence metrics have markedly different statistical consequences when it comes to sample complexities.

%, 
%given respectively by
%
%\begin{align}
%\rho_{\TV}\left(P_{s,a}, P^0_{s,a}\right) & \defn  \frac{1}{2} \sum_{s'\in \cS} P^{\no}(s' \mymid s,a)\left| 1 - \frac{P(s' \mymid s,a)}{P^{\no}(s' \mymid s,a)}\right|, \label{eq:tv-distance} \\
%\rho_{\chi^2} \left(P_{s,a}, P^0_{s,a}\right) & \defn \sum_{s'\in \cS} P^{\no}(s' \mymid s,a)\left(1 - \frac{P(s' \mymid s,a)}{P^{\no}(s' \mymid s,a)}\right)^2. \label{eq:chi-squared-distance}
%\end{align}
%
%Note that $\rho_{\TV}\left(P_{s,a}, P^0_{s,a}\right)\in [0,1]$ and $\rho_{\chi^2} \left(P_{s,a}, P^0_{s,a}\right) \in [0,\infty)$ in general. As we shall see shortly, these two choices of divergence metrics result in drastically different messages when it comes to sample complexities.

\paragraph{Robust value functions and Q-functions.}
In an RMDP, the performance of a policy $\pi$ is evaluated according to its worst-case performance over all transition kernels in the uncertainty set. This leads to the notions of the {\em robust value function} $V^{\pi, \ror} $ and the {\em robust Q-function} $Q^{\pi,\ror}$ in $\cM_\rob$, defined respectively as
\begin{subequations}
\begin{align} \label{eq:robust-value-def}
  V^{\pi,\ror}(s) &\defn \inf_{P\,\in \, \unb^{\ror}(P^{\no})} V^{\pi,P} (s) \\  
	  Q^{\pi,\ror}(s,a) & \defn \inf_{P\,\in\, \unb^{\ror}(P^{\no})} Q^{\pi,P}(s,a)
\end{align}
\end{subequations}
for all $(s,a)\in \cS \times \cA$. Here, $V^{\pi,P}$ and $Q^{\pi,P}$ denote the value function and Q-function, respectively, of policy $\pi$ in the standard MDP with transition kernel $P$. Thus, $V^{\pi,\ror}$ and $Q^{\pi,\ror}$ quantify the worst-case discounted cumulative rewards attainable under $\pi$ when the transition dynamics are allowed to vary within the uncertainty set $\cU^{\ror}(P^\no)$.

\paragraph{Optimal robust policy and robust Bellman operator.}
As a fundamental generalization of the dynamic programming framework of standard MDPs,  there exists at least one
deterministic policy that maximizes the robust value function (resp.~robust Q-function) simultaneously across all states (resp.~state-action pairs) \citep{iyengar2005robust,nilim2005robust}. Accordingly, we denote such a policy (often referred to as an optimal robust policy) by $\pi^\star$, and let $V^{\star,\ror}$ and $Q^{\star,\ror}$ denote the corresponding  {\em optimal robust value function}  and {\em optimal robust Q-function}, respectively, which satisfy
\begin{subequations} \label{eq:mdp-value-Q}
\begin{align}
	\forall s \in \cS: \quad &V^{\star,\ror}(s) \defn V^{\pi^\star,\ror}(s) = \max_\pi V^{\pi,\ror}(s), \\
	\forall (s,a) \in \cS \times \cA: \quad &Q^{\star,\ror}(s,a) \defn Q^{\pi^\star,\ror}(s,a) = \max_\pi Q^{\pi,\ror}(s,a).
\end{align}
\end{subequations}

A key machinery in RMDPs is the robust counterpart of Bellman's optimality principle, 
embodied in the following {\em robust Bellman consistency equation} and {\em robust Bellman optimality equation}: 
\begin{subequations}\label{eq:bellman-equ-infinite-all}
\begin{align}
    Q^{\pi,\ror}(s,a) &= r(s,a) + \gamma\inf_{ P_{s,a} \,\in\, \unb^{\ror}(P^{\no}_{s,a})} \langle P_{s,a}, V^{\pi,\ror} \rangle, \label{eq:bellman-equ-pi-infinite}\\
    Q^{\star,\ror}(s,a) &= r(s,a) + \gamma\inf_{P_{s,a}\,\in\, \unb^{\ror}(P^{\no}_{s,a})} \langle P_{s,a}, V^{\star,\ror} \rangle \label{eq:bellman-equ-star-infinite}
\end{align}
\end{subequations}
for every $(s,a)\in \cS\times \cA$, 
where $\pi$ stands for any policy. 
Correspondingly, the robust Bellman operator \citep{iyengar2005robust,nilim2005robust}, denoted by $\cT^\ror(\cdot): \mathbb{R}^{SA} \rightarrow \mathbb{R}^{SA}$, is defined by
\begin{subequations}
\label{eq:robust_bellman}
\begin{align}
	%\forall (s,a)\in \cS\times \cA :\quad 
    \cT^\ror(Q)(s,a) \defn r(s,a) + \gamma \inf_{ P_{s,a} \,\in\, \unb^{\ror}(P^{\no}_{s,a})} \langle P_{s,a}, V \rangle
	\end{align}
for any input vector $Q\in \mathbb{R}^{SA}$ and every $(s,a)\in \cS\times \cA$, 
where, as usual,  
\begin{equation}\forall s\in \cS: \quad V(s) \defn \max_a Q(s,a).
\end{equation}
\end{subequations}

It is well known that $Q^{\star,\ror}$ is the unique fixed point of $\cT^\ror$. Consequently, the optimal robust value function and Q-function can be computed via \emph{distributionally robust value iteration} (\texttt{DRVI}), which is a direct generalization of the classical value iteration algorithm by recursively applying the robust Bellman operator starting from an arbitrary initialization. Analogous to standard value iteration,  \texttt{DRVI} converges geometrically fast due to the $\gamma$-contraction property of $\cT^\ror$  w.r.t.~the $\ell_\infty$ norm \citep{iyengar2005robust,nilim2005robust}.

\paragraph{Sampling mechanism and goal.} 
To keep the discussion concrete and avoid the complications arising from different data collection schemes, we restrict attention in this section to the generative model as described in Chapter~\ref{chapter:generative-model}, which already captures several key distinctions between the standard MDPs and their distributionally robust counterparts. 

Specifically, we draw $N$ independent samples for each state-action pair from the generative model according to the nominal transition kernel $P^{\nominal}$. That is, for every $(s,a)\in\cS\times\cA$, we collect
\begin{align*}
s^{(i)}(s,a)\overset{\mathrm{ind.}}{\sim} P^\no(\cdot \mymid s,a), \qquad i = 1, \cdots, N,
\end{align*}
resulting in a total sample size $N^{\mathsf{total}} := NSA$.

Given these samples, the objective is to learn an $\varepsilon$-optimal robust policy $\widehat{\pi}$ satisfying
\begin{align}
	\forall s\in \cS:\quad V^{\star, \sigma}(s) - V^{\widehat{\pi}, \sigma}(s) \leq \varepsilon.
\end{align}
Compared with the standard RL setting, the data are collected in exactly the same manner; the key difference lies in the performance criterion, which is now defined in terms of robust value functions.

Having formalized the problem, a natural question is to quantify the impact of robustness on sample complexity. The remainder of this section is devoted to discussing the statistical price of robustness and illustrating how it varies across different choices of uncertainty sets.

%
%\begin{itemize}
%    \item  For each $(s,a)\in \cS\times \cA$, draw $N$ independent samples from $P^{\nominal}(\cdot\mymid s,a)$, where $P^{\nominal}: \cS\times \cA\rightarrow \Delta(\cS)$ denotes the transition kernel of a ``nominal'' MDP. 
%\end{itemize}
%

%More concretely, imagine that one has access to a generative model  that draws samples from a MDP with a nominal transition kernel \citep{kearns1999finite}. 

% Standard RL aims to learn the optimal policy tailored to the nominal kernel, for which the minimax sample complexity limit has been fully settled \citep{azar2013minimax,li2020breaking}; see also Chapter~\ref{chapter:generative-model}. 

%This setting is frequently referred to as robust MDPs (RMDPs). Clearly, the RMDP framework helps ensure that the performance of the learned policy does not fail catastrophically as long as the sim-to-real gap is not overly large. 

%\subsection{Problem formulation}

%\paragraph{Sampling mechanism: a generative model.}

%\paragraph{Goal.} Given the collected samples, the task is to learn the robust optimal policy for the RMDP --- w.r.t.~some prescribed uncertainty set $\cU^\ror(P^\no)$ around the nominal kernel --- using as few samples as possible. Specifically, given some target accuracy level $\varepsilon>0$, the goal is to seek an $\varepsilon$-optimal robust policy $\widehat{\pi}$ obeying

\subsection{A model-based algorithm}

In this section, we focus on a model-based approach for RMDPs that attains state-of-the-art statistical guarantees. At a high level, this algorithm first constructs an empirical estimate of the nominal transition kernel using the collected samples and subsequently applies \DRVI to compute an optimal robust policy for the resulting empirical model.

\paragraph{Empirical nominal kernel.} The empirical nominal transition kernel $\widehat{P}^\no \in \mathbb{R}^{SA\times S}$ can be constructed from the empirical transition frequencies. Specifically, for each state-action pair $(s,a)\in \cS\times \cA$, compute
\begin{align}\label{eq:nominal_empirical}
	\forall s'\in \cS:\quad \widehat{P}^{\nominal}(s'\mymid s,a) \defn  \frac{1}{N} \sum\limits_{i=1}^N \mathds{1} \big\{ s' = s^{(i)}(s,a) \big\},
\end{align}
which coincides with the empirical model estimation procedure used in standard RL (see Chapter~\ref{chapter:generative-model}). Replacing the nominal transition kernel $P^\no$ with its empirical counterpart $\widehat{P}^\no$ yields the empirical RMDP
\[
\widehat{\cM}_{\mathsf{rob}} = \big\{\cS,\cA, \gamma, \cU^{\ror}(\widehat{P}^\no), r \big\}.
\]

As in the population setting, let $\widehat{V}^{\pi,\ror}$ and $\widehat{Q}^{\pi,\ror}$ denote the robust value function and robust Q-function of policy $\pi$ in $\widehat{\cM}_{\mathsf{rob}}$, respectively (analogous to~\eqref{eq:mdp-value-Q}). In addition, we denote the corresponding optimal robust policy as $\widehat{\pi}^{\star}$ and the optimal robust value function (resp.~optimal robust Q-function) as $\widehat{V}^{\star, \ror}$ (resp.~$\widehat{Q}^{\star, \ror}$) (analogous to \eqref{eq:bellman-equ-infinite-all}). The corresponding robust Bellman optimality equation is given by
\begin{align}
	&\widehat{Q}^{\star,\ror}(s,a) = r(s,a) + \gamma\inf_{P_{s,a}\,\in\, \unb^{\ror}(\widehat{P}^{\no}_{s,a})} \langle P_{s,a}, \widehat{V}^{\star,\ror} \rangle \label{eq:bellman-equ-star-infinite-est}
\end{align}
for all $(s,a)\in \cS\times \cA$.

Equipped with $\widehat{P}^\no$, we can define the empirical robust Bellman operator $\that^\ror$ such that: for any input vector $Q\in \mathbb{R}^{SA}$ and each $(s,a)\in \cS\times \cA$,
\begin{align}\label{eq:robust_bellman_empirical}
	    \that^\ror(Q)(s,a) \defn r(s,a) + \gamma \inf_{ P_{s,a}\, \in\, \unb^{\ror}(\widehat{P}^{\no}_{s,a})} \langle P_{s,a}, V\rangle,   
\end{align}
where $V(s) \defn \max_a Q(s,a)$ for all $s \in \cS$.

\paragraph{Distributionally robust value iteration.} 
To compute the fixed point of $\that^\ror(\cdot)$, we introduce distributionally robust value iteration (\DRVI). Starting from an initialization $\widehat{Q}_0 = 0$, the update rule at the $t$-th ($t\geq 1$) iteration is given by
\begin{align}
\widehat{Q}_t(s,a) &= \that^\ror \big(\widehat{Q}_{t-1} \big)(s,a) \notag\\
 &= r(s,a) + \gamma \inf_{ P_{s,a}\, \in \, \unb^{\ror}(\widehat{P}^{\no}_{s,a})} \langle P_{s,a}, \widehat{V}_{t-1}  \rangle 
 \label{eq:vi-iteration}
\end{align}
for all $ (s,a)\in  \cS\times \cA $, where $\widehat{V}_{t-1}(s) = \max_a \widehat{Q}_{t-1}(s,a)$ for all $s \in \cS$.
After running the update for $T$ iterations, we output the greedy policy $\widehat{\pi}$ induced by $\widehat{Q}_T$, namely,
\begin{align}
	 \forall s \in \cS: \quad \widehat{\pi}(s) = \arg\max_a \widehat{Q}_T(s,a).
\end{align}
The complete procedure is summarized in Algorithm~\ref{alg:cvi-dro-infinite}.
As in standard value iteration, the iterates $\big\{\widehat{Q}_t \big\}_{t\geq0}$ of \DRVI converge linearly to the unique fixed point $\widehat{Q}^{\star,\ror}$, 
owing to the $\gamma$-contraction property of the empirical robust Bellman operator $\that^\ror(\cdot)$.  

\begin{algorithm}[t]
\DontPrintSemicolon
	\textbf{input:} %empirical nominal transition kernel $\widehat{P}^{\no}$; 
    discount factor $\gamma$, reward function $r$, uncertainty level $\ror$, number of iterations $T$. \\ 
	\textbf{initialize } $\widehat{Q}_0(s,a)\leftarrow 0$, $\widehat{V}_0(s)\leftarrow 0$ for all $(s,a) \in \cS\times \cA$. \\
    \For{$(s,a)\in \cS\times \cA$}{
    draw $N$ independent samples $s^{(i)}(s,a) \overset{\text{i.i.d.}}{\sim} P^{\nominal}(\cdot\mymid s,a)$. \\
    compute $\widehat{P}^{\nominal}(\cdot \mymid s,a)$ based on \eqref{eq:nominal_empirical}. 
    }
   \For{$t = 1,2,\cdots, T$}
	{
		
		\For{$(s,a)\in\cS\times \cA$}{
			compute $\widehat{Q}_t(s, a)$ according to \eqref{eq:vi-iteration};

		}
		\For{$s\in \cS$}{
			compute $\widehat{V}_t(s) = \max_a \widehat{Q}_t(s, a)$;
		}
	}

	\textbf{output:} %$\widehat{Q}_T$, $\widehat{V}_T$ and 
    policy $\widehat{\pi}$ obeying $\widehat{\pi}(s) \defn \arg\max_a \widehat{Q}_T(s,a)$.
	\caption{Distributionally robust value iteration (\DRVI) for discounted infinite-horizon RMDPs}
 \label{alg:cvi-dro-infinite}
\end{algorithm} 

Nevertheless, a direct implementation of \eqref{eq:vi-iteration} can be computationally expensive, as it requires solving an optimization problem over an $S$-dimensional probability simplex for every state-action pair at each iteration. Fortunately, by strong duality \citep{iyengar2005robust}, the inner optimization problem in \eqref{eq:vi-iteration} admits an equivalent dual formulation involving only a \emph{scalar} dual variable. Consequently, each update in \DRVI can be carried out efficiently by solving a one-dimensional optimization problem. For concreteness, below we illustrate this idea for the two choices of $\rho$: TV distance and $\chi^2$ divergence. To this end, for any $V\in \mathbb{R}^S$ and any $\alpha\geq 0$, define the clipped vector $[V]_{\alpha}$ as 
\begin{align}
    \forall s\in \cS: \quad 
	[V]_{\alpha}(s) \defn \begin{cases} \alpha, & \text{if } V(s) > \alpha, \\
V(s), & \text{otherwise.}
\end{cases} \label{eq:V-alpha-defn}
\end{align}
The lemma below, which follows directly from \citet[Lemma~4.3]{iyengar2005robust}, provides an equivalent dual representation for the inner optimization problem in \eqref{eq:vi-iteration}. 
\begin{lemma}[Strong duality]
\label{lemma:dual-form}
Consider any given probability vector  $p^{\nominal}\in\Delta(\cS)$, any uncertainty level $\ror$, and the uncertainty set $\unb^{\ror}(p^{\nominal})$. Consider any vector $V\in \mathbb{R}^S$ obeying $  V \geq  {0}$, recall the definition of $[V]_\alpha$ in \eqref{eq:V-alpha-defn}, and set $V_{\min} = \min_s V(s)$, $V_{\max} = \max_s V(s)$. 
\begin{itemize}
    \item If $\rho$ is taken to be the TV distance, then
  \begin{align}
\arxiv{&}	\inf_{ p \,\in\, \unb^{\ror}(p^{\nominal})} \langle p, V\rangle  \arxiv{ \notag\\ &\quad} = 
	 \max_{\alpha\in [ V_{\min},  V_{\max}]} \left\{ \langle p^{\nominal}, \left[V\right]_{\alpha} \rangle - \ror \Big(\alpha - \min_{s'}\left[V\right]_{\alpha}(s') \Big)\right\} .\label{eq:vi-l1norm} 
\end{align}  

    \item If $\rho$ is taken to be the $\chi^2$ divergence, then
\begin{align}
 \arxiv{&}	\inf_{ p \,\in\, \unb^{\ror}(p^{\nominal})} \langle p, V\rangle  \arxiv{ \notag\\
&\quad }= \max_{\alpha\in [ V_{\min},  V_{\max}] } \left\{ \langle p^{\nominal}, [V]_\alpha \rangle - \sqrt{\sigma \mathsf{Var}_{p^{\nominal}}\left([V]_\alpha\right) } \right\},
\end{align}
where we recall the definition of $\mathsf{Var}_p(V)$ in \eqref{eq:matrix-VarP-V-expression}.  
\end{itemize}
\end{lemma}

\subsection{Theoretical guarantees: the case of TV distance}\label{sec:upper-L1}

We begin with the setting in which the uncertainty set is defined using the total-variation distance. The following theorem develops an upper bound on the sample complexity of \DRVI for identifying an $\varepsilon$-optimal robust policy.

\begin{theorem}[\citet{shi2024curious,shi2026curious}]\label{thm:l1-upper-bound} 
	Let the metric $\rho$ be the TV distance.  Consider any discount factor $\gamma \in [1/4,1 )$, 
 uncertainty level $\ror\in (0,1)$, $\delta \in (0,1)$, and $\varepsilon \in \big(0, \sqrt{1/\max\{1-\gamma, \ror\}} \,\big]$. 
	Let $\widehat{\pi}$ be the policy returned by Algorithm~\ref{alg:cvi-dro-infinite} after $T = C_1 \log \big( \frac{N}{1-\gamma}\big)$ iterations. 
	Then, with probability at least $1-\delta$, one has
\begin{align*}
	\forall s\in\cS: \quad V^{\star, \ror}(s) - V^{\widehat{\pi}, \ror}(s) \leq \varepsilon,
\end{align*}
provided that the number of samples per state-action pair obeys
\begin{align*}
	N \geq  \frac{C_2 }{ (1-\gamma)^2 \max\{1-\gamma, \ror\} \varepsilon^2}\log\left(\frac{SAN}{(1-\gamma)\delta}\right).
\end{align*}
Here, $C_1, C_2>0$ are some large enough universal constants.  
\end{theorem}

\begin{remark}\label{remark:tv-upper}
Note that Theorem~\ref{thm:l1-upper-bound} is not only valid when invoking Algorithm~\ref{alg:cvi-dro-infinite}. 
In fact, the theorem holds for any oracle planning algorithm (designed based on the empirical transitions $\widehat{P}^\no$) whose output policy $\widehat{\pi}$ obeys 
\begin{equation}\label{eq:oracle_requirement}
\big\| \widehat{V}^{\star,\ror} - \widehat{V}^{\widehat{\pi},\ror} \big\|_\infty \leq \widetilde{O}\bigg(\frac{(1-\gamma)^2 }{N} 
%\log\left(\frac{SAN}{(1-\gamma)\delta}\right)  
\bigg). 
\end{equation}
Thus, the statistical guarantee is largely independent of the specific planning algorithm employed, provided that the empirical planning error is sufficiently small. 
\end{remark}

Next, we present a matching minimax lower bound. Taken collectively, these upper and lower bounds characterize the sample complexity of learning RMDPs under TV-based uncertainty and establish the statistical optimality of the model-based algorithm \DRVI.
\begin{theorem}[\cite{shi2024curious,shi2026curious}]\label{thm:l1-lower-bound}
Let $\rho$ be the TV distance. 
Consider any tuple $(S, A, \gamma, \ror, \varepsilon)$ obeying $\ror\in (0, 1 - c_0]$, with $0 <c_0 \leq 1/8$ being any small enough positive constant. Suppose further that $\gamma\in [ 1/2, 1)$ and $\varepsilon \in \big(0, \frac{c_0}{256(1-\gamma)} \big]$. Then there exist two $\gamma$-discounted infinite-horizon RMDPs $\cM_0$ and $\cM_1$, an initial state distribution $\mu$, and a dataset with $N$ independent samples for each state-action pair over the nominal transition kernel (for $\cM_0$ and $\cM_1$ respectively), such that
	\[
	\inf_{\widehat{\pi}}\max\left\{ \mathbb{P}_{0}\big( V^{\star,\ror}(\mu)-V^{\widehat{\pi}, \ror}(\mu)>\varepsilon\big), \,
	\mathbb{P}_{1}\big( V^{\star,\ror }(\mu) - V^{\widehat{\pi},\ror}(\mu) >\varepsilon\big)\right\} \geq\frac{1}{8},
\]
provided that
\[N \leq \frac{c_0    \log 2  }{ 8192 (1-\gamma)^2 \max\{ 1 -\gamma, \sigma\}\varepsilon^2}.\]
Here, the infimum is over all estimators $\widehat{\pi}$, and $\mathbb{P}_{0}$ (resp.~$\mathbb{P}_{1}$) denotes the probability distribution
induced by the RMDP $\mathcal{M}_{0}$ (resp.~$\mathcal{M}_{1}$).
\end{theorem}

\paragraph{Implications.}
We next interpret the preceding upper and lower bounds and discuss their implications for the sample complexity of learning RMDPs when the uncertainty set is specified via the TV distance. 
\begin{itemize}
\item \emph{Minimax-optimal sample complexity.} Theorem~\ref{thm:l1-upper-bound} shows that \DRVI (or any oracle planning algorithm satisfying the condition in Remark~\ref{remark:tv-upper}) achieves $\varepsilon$-accuracy using
\begin{align}\label{eq:tv-final-samples}
\widetilde{O} \left(\frac{SA}{ (1-\gamma)^2 \max\{1-\gamma, \ror\} \varepsilon^2} \right)
\end{align}
samples in total. Combined with the lower bound in Theorem~\ref{thm:l1-lower-bound}, this establishes the minimax optimality of \DRVI (up to logarithmic factors) over a broad range of uncertainty levels $\ror$. 
Importantly,  the sample complexity scales linearly with the dimension of the state-action space and improves as the uncertainty level $\ror$ increases in the regime $\ror\gtrsim 1-\gamma$.

\item \emph{RMDPs can be easier to learn than standard MDPs under TV uncertainty.} Recall from Chapter~\ref{chapter:generative-model} that learning a standard $\gamma$-discounted MDP with access to a generative model requires 
%}
\begin{equation}\label{eq:standard-final-samples}
 \widetilde{O} \left(\frac{SA}{ (1-\gamma)^3 \varepsilon^2} \right)
\end{equation}
samples to achieve $\varepsilon$ accuracy. 
Comparing this with the RMDP sample complexity \eqref{eq:tv-final-samples}, 
we observe that learning RMDPs under TV uncertainty is at least as easy as --- if not easier than --- learning standard MDPs. In particular, when $\sigma \lesssim 1-\gamma$ is small, the sample complexity for learning RMDPs coincides with \eqref{eq:standard-final-samples} for learning standard MDPs, which is as expected since the RMDP approaches the standard MDP when $\sigma=0$. In contrast, when $1-\gamma \lesssim \sigma < 1 $, the sample complexity for learning RMDPs simplifies to
\begin{align*}
%\label{eq:tv-large-sigma}
\widetilde{O} \left(\frac{SA}{ (1-\gamma)^2   \ror  \varepsilon^2} \right) ,
\end{align*}
which improves upon the standard MDP counterpart by a factor of $\sigma/(1-\gamma)$. 

\end{itemize}

\subsection{Theoretical guarantees: the case of $\chi^2$ divergence}

We now turn to the setting in which the uncertainty set is specified by means of the $\chi^2$ divergence. As we shall see, the resulting sample complexity guarantees differ markedly from those obtained under TV-based uncertainty. The following theorem summarizes the corresponding upper bound.

 \begin{theorem}[\cite{shi2024curious,shi2026curious}]\label{thm:l2-upper-bound}
	Let the metric $\rho$ be the $\chi^2$ divergence. Consider any uncertainty level $\ror\in (0, \infty)$, $\gamma\in [1/4,1)$, $\delta \in(0,1)$, and $\varepsilon\in \big(0,\frac{1}{1-\gamma} \big]$. With probability at least $1-\delta$, the policy $\widehat{\pi}$ returned by  Algorithm~\ref{alg:cvi-dro-infinite} after $T = c_1 \log \big( \frac{N}{1-\gamma}\big)$ iterations satisfies
\begin{align*}
	\forall s\in\cS: \quad V^{\star, \ror}(s) - V^{\widehat{\pi}, \ror}(s) \leq \varepsilon,
\end{align*}
provided that the number of samples per state-action pair obeys
\begin{align*}
	N \geq \frac{c_2  }{(1-\gamma)^3 \varepsilon^2} \left(1+\frac{\max\{\sqrt{\ror}, \ror \} }{1-\gamma} \right) \log\left(\frac{SA N}{\delta}\right) .
\end{align*}
Here, $c_1, c_2>0$ are some large enough universal constants.  
\end{theorem}

\begin{remark}\label{remark:chi2-upper}
Analogous to  Remark~\ref{remark:tv-upper}, the sample complexity guarantee in Theorem~\ref{thm:l2-upper-bound} extends beyond \DRVI and remains valid for any oracle planning algorithm whose output policy $\widehat{\pi}$ satisfies
$\big\| \widehat{V}^{\star,\ror} - \widehat{V}^{\widehat{\pi},\ror} \big\|_\infty \leq O\Big(\frac{ \log(\frac{SAN}{(1-\gamma)\delta}) }{N^2} \Big)$.
\end{remark}

In addition, to gauge the tightness of Theorem~\ref{thm:l2-upper-bound} and pin down the sample complexity limit under $\chi^2$-based uncertainty, we next present a minimax lower bound.
\begin{theorem}[\cite{shi2024curious,shi2026curious}]\label{thm:chi2-lower-bound}
Let $\rho$ be the $\chi^2$ divergence. 
Consider any $(S, A, \gamma, \ror, \varepsilon)$ obeying  $\gamma\in [3/4, 1)$, $\ror \in (0,\infty)$, and
$\varepsilon \leq  \frac{c_3}{(1-\gamma)}$
for some small universal constant $c_3>0$.
Then there exist two discounted infinite-horizon RMDPs $\cM_0$ and $\cM_1$, an initial state distribution $\mu$, and a dataset with $N$ independent samples per $(s,a)$ pair over the nominal transition kernel (for $\cM_0$ and $\cM_1$ respectively), such that
\begin{align*}
	\inf_{\widehat{\pi}}\max\left\{ \mathbb{P}_{0}\big( V^{\star,\ror}(\mu)-V^{\widehat{\pi}, \ror}(\mu)>\varepsilon\big), \,
	\mathbb{P}_{1}\big( V^{\star,\ror }(\mu) - V^{\widehat{\pi},\ror}(\mu) >\varepsilon\big)\right\} \geq\frac{1}{8}, %\label{eq:lower-bound-chi2-all}
\end{align*}
provided that the number of samples per state-action pair obeys
\begin{align*} %\label{eq:final-chi2-lower-bound}
N \leq \frac{c_4}{(1-\gamma)^3  \varepsilon^2}\left(1+\frac{\ror}{1-\gamma} \right) 
\end{align*}
for some universal constant $c_4>0$.
\end{theorem}

\paragraph{Implications.}
We are now ready to discuss several important implications of the above theorems.

\begin{itemize}
\item \emph{Nearly tight sample complexity.} In order to achieve $\varepsilon$-accuracy for RMDPs under the $\chi^2$ divergence, Theorem~\ref{thm:l2-upper-bound} asserts that a total number of samples on the order of
\begin{align}\label{eq:chi2-final-samples}
 \widetilde{O} \left(\frac{SA}{(1-\gamma)^3 \varepsilon^2} \left(1+\frac{ \max\{\sqrt{\ror},  \ror \}}{1-\gamma} \right)\right)
\end{align}
is sufficient for \DRVI (or any other oracle planning algorithm as discussed in Remark~\ref{remark:chi2-upper}). Taking this together with the minimax lower bound in Theorem~\ref{thm:chi2-lower-bound} confirms that the sample complexity is optimal over a broad range of the uncertainty level $\sigma$ (when $\sigma \lesssim (1-\gamma)^2$ and $\sigma\gtrsim 1$). In particular,
\begin{itemize}
\item when $\sigma \lesssim (1-\gamma)^2$, the sample complexity upper bound $\widetilde{O} \left( \frac{SA }{  (1-\gamma)^3 \varepsilon^2} \right)$ is sharp and matches the minimax lower bound; 

\item  when $(1-\gamma)^2 \lesssim \sigma \lesssim 1$, the sample complexity upper bound is on the order of $\widetilde{O} \left( \frac{SA \sqrt{\sigma}}{  (1-\gamma)^4 \varepsilon^2} \right)$, leaving a gap of at most $\sqrt{\frac{1}{1-\gamma}}$ from the minimax lower bound;

\item  when $\sigma\gtrsim 1$, the sample complexity upper bound $\widetilde{O} \left( \frac{SA \sigma}{  (1-\gamma)^4 \varepsilon^2} \right)$ again matches the minimax lower bound.
\end{itemize}

\item \emph{RMDPs are harder to learn than standard MDPs under $\chi^2$ uncertainty.}
The minimax lower bound developed in Theorem~\ref{thm:chi2-lower-bound} reveals a striking contrast with the TV-distance case. Whereas TV uncertainty can make robust MDPs statistically easier to learn, $\chi^2$ uncertainty never reduces the sample complexity relative to standard MDPs. Indeed, several distinct regimes are worth highlighting:
\begin{itemize}
\item when $\sigma \lesssim 1-\gamma$, the lower bound reduces to
\begin{align*}
\widetilde{\Omega} \left( \frac{SA }{  (1-\gamma)^3 \varepsilon^2} \right),
\end{align*}
which coincides with the sample complexity of standard MDPs;

\item when $\sigma \gtrsim 1-\gamma$, the lower bound becomes
\begin{align*}
\widetilde{\Omega} \left( \frac{SA \sigma}{  (1-\gamma)^4 \varepsilon^2} \right),
\end{align*}
which exceeds the standard MDP sample complexity by a factor of $\ror/(1-\gamma)$.
\end{itemize}

\end{itemize}

\subsection{Why are the TV and $\chi^2$ cases markedly different?}

The preceding results reveal a surprising phenomenon: the sample complexity of robust RL depends critically on both the size and the geometry of the uncertainty set. To gain intuition, consider how estimation errors propagate through a single value-iteration update (cf.~\eqref{eq:vi-iteration}) from a fixed value function $V$, when the nominal transition kernel $P^{\nominal}$ is replaced by its empirical estimate $\widehat{P}^{\nominal}$. 
This leads to the following error terms in standard RL and robust RL, respectively:
	 \begin{align*}
    \text{standard RL:}   ~~  \delta_{\textsf{standard}} &= \Big| \big\langle P^{\nominal}_{s,a}, {V} \big\rangle - \big\langle \widehat{P}_{s,a}^{\nominal}, {V} \big\rangle \Big|  , \\
    \text{robust RL:}~~   ~~  \delta_{\textsf{robust}} &= \Big|\inf_{P_{s,a}\,\in\, \mathcal{U}^\sigma(P^{\nominal} )} \langle P_{s,a}, V \rangle  - \inf_{P_{s,a}\, \in\, \mathcal{U}^\sigma(\widehat{P}^{\nominal} )} \langle P_{s,a}, V \rangle \Big| .
  \end{align*}
As a key distinction, the estimation error in standard RL is linear in the model estimation error $P^{\nominal} - \widehat{P}^{\nominal}$, whereas in robust RL, it is governed by an additional inner optimization over the uncertainty set, resulting in a generally nonlinear dependence that often lacks a closed-form characterization. We provide some brief remarks regarding $\delta_{\textsf{robust}}$ under both TV distance and $\chi^2$ divergence. 
\begin{itemize}
	\item \emph{TV uncertainty.} 
    For uncertainty sets based on the TV distance, the dual representation in \eqref{eq:vi-l1norm} reveals that $\delta_{\textsf{robust}}$ behaves approximately linearly in $P^{\nominal}-\widehat{P}^{\nominal}$, much like $\delta_{\textsf{standard}}$. Intuitively, this is because TV-based uncertainty sets have a homogeneous shape across all states and exhibit an almost instance-independent structure (w.r.t.~uncertainty set center).  Moreover, when $\sigma\gtrsim 1-\gamma$, the robust value function of a fixed policy exhibits a substantially smaller dynamic range than its standard-RL counterpart. This favorable low-variance statistical property ultimately leads to improved data efficiency and explains why robust RL can require fewer samples than standard RL under TV uncertainty.

	\item \emph{$\chi^2$ uncertainty.} The situation is fundamentally different for uncertainty sets based on the $\chi^2$ divergence. In this case, $\delta_{\textsf{robust}}$ becomes highly nonlinear and can significantly amplify the model estimation error $P^{\nominal}-\widehat{P}^{\nominal}$. This effect stems from the heterogeneous structure of the $\chi^2$ uncertainty sets, which varies across states and depends strongly on the specific center of the set ($P^{\nominal}$ or $\widehat{P}^{\nominal}$).  As a result, small perturbations in the nominal transition kernel can induce substantially larger perturbations in the robust Bellman update. This amplification effect leads to the increased sample complexity observed under $\chi^2$ uncertainty.

     	\end{itemize}

%% file: chapters/marl.tex
\section{Competitive multi-agent RL}

The field of multi-agent reinforcement learning (MARL) studies how multiple agents interact and make decisions in a shared dynamic environment \citep{zhang2021multi}. Recent breakthroughs in game playing \citep{vinyals2019grandmaster,brown2019superhuman} and multi-robot control \citep{matignon2012coordinated} epitomize its broad applicability. In many applications, however, agents have conflicting objectives and must act strategically to maximize their individual utilities. Such competitive interactions are commonly modeled as {\em Markov games (MGs)} --- also known as {\em stochastic games} --- which provide a mathematically rigorous and versatile framework for studying competitive MARL \citep{shapley1953stochastic,littman1994markov}. 

In competitive settings, the goal is typically to learn certain equilibrium strategy profiles, most notably the Nash equilibria \citep{nash1950equilibrium}. A key challenge in modern applications is data efficiency: agents often have little prior knowledge of the environment and must learn effective strategies through interaction, often in high-dimensional environments. Despite significant progress, the sample complexity of learning equilibrium strategies in MARL was poorly understood until fairly recently, even in the fundamental setting of two-player zero-sum Markov games.

\subsection{Markov games: preliminaries}

In this section, we introduce the basic framework of Markov games, together with two important solution concepts: Nash equilibrium and coarse correlated equilibrium. For simplicity, we restrict our attention to finite-horizon, non-stationary Markov games. 

\paragraph{Markov games.} 
A finite-horizon, non-stationary {\em multi-agent general-sum Markov game}, denoted by $\mathcal{MG}=\big\{ \cS, \{\cA_i\}_{1 \le i \le m}, H, P, r \big\}$, models the interaction of $m$ agents in a shared dynamic environment,  where agents may have competing objectives.
The components of $\mathcal{MG}$ are defined as follows. 
\begin{itemize}
	\item $\cS=\{1,\cdots,S\}$: the state space of the shared environment, consisting of $S$ different states. 
    
	\item $\mathcal{A}_i=\{1,\cdots, A_i\}$: the action space of the $i$-th agent ($1\leq i\leq m$), consisting of $A_i$ actions.

    \item $\mathcal{A}$ and $\mathcal{A}_{-i}$: the joint action space of all agents and the joint action space excluding agent $i$, respectively, defined by
		\begin{equation}
			\mathcal{A} \coloneqq \mathcal{A}_1 \times \cdots \times \mathcal{A}_m; 
            \quad 
			\mathcal{A}_{-i} \coloneqq \prod_{j:j\neq i}\mathcal{A}_j ~~~ (1\leq i\leq m). 
		\end{equation}
		Throughout, we use  $\bm{a} \in \cA$ (resp.~$\bm{a}_{-i} \in \cA_{-i}$) to denote a joint action profile of all agents (resp.~all agents except agent $i$).  
		
	\item $H$: the horizon length. 

	\item $P=\{P_h\}_{1\leq h \leq H}$, with $P_h: \cS \times \cA \rightarrow \Delta(\cS)$: the probability transition kernel of $\mathcal{MG}$. 
		Specifically, for any $(s, \bm{a}, h,s')\in \cS\times \cA \times [H] \times \cS$, we let $P_h(s'\mymid s,\bm{a})$ indicate the probability of transitioning from state $s$ to state $s'$ at step $h$ when the joint action profile $\bm{a}$ is executed. 

	\item $r=\{r_{i,h}\}_{1\leq h \leq H, 1\leq i\leq m}$, with $r_{i,h}: \cS\times \cA \rightarrow [0,1]$: the (deterministic) reward function. 
		Specifically, for any $(s,\bm{a},h)\in \cS\times \cA \times [H]$, $r_{i,h}(s,\bm{a})$ stands for the immediate reward received by agent $i$ at step $h$ when the Markov game is in state $s$ and the joint action profile is $\bm a$.   
		Throughout, rewards are assumed to be normalized so that $r_{i,h}(s,\bm{a})\in [0,1]$ for any $(s,\bm{a},h,i)\in \cS\times \cA \times[H] \times [m]$.

\end{itemize}

\begin{figure}
\begin{center}
\includegraphics[width=\textwidth]{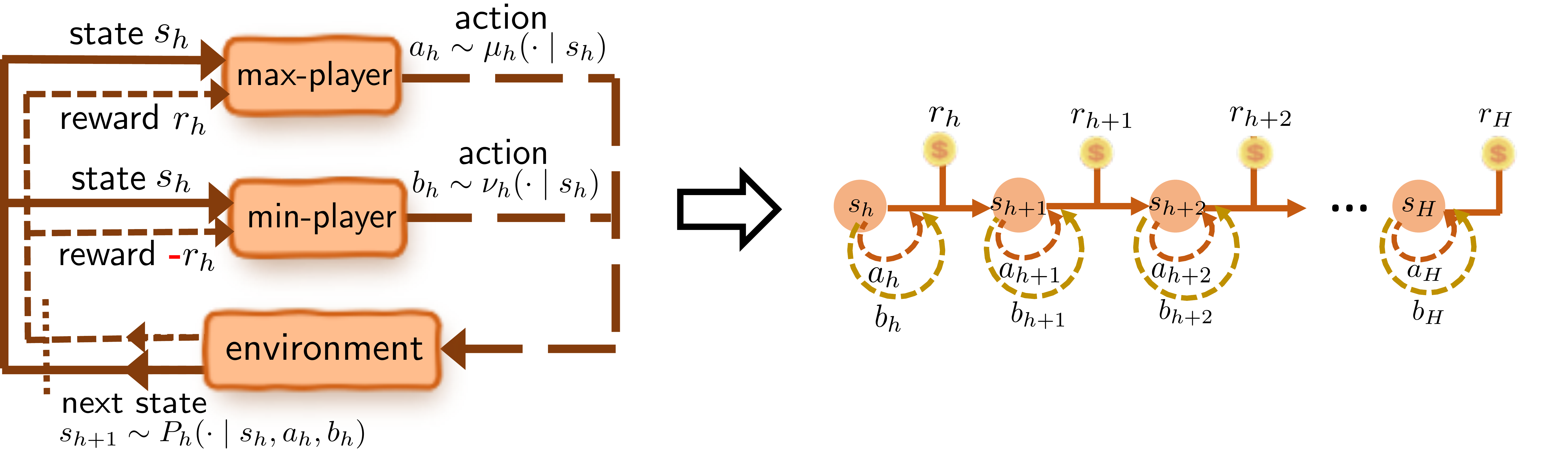} 
\end{center}	
	\caption{Illustration of a two-player zero-sum Markov game.
	\label{fig:MG}}

\end{figure}

\noindent 
An important special case is the \emph{two-player zero-sum Markov game}, which satisfies $$r_{2,h}=-r_{1,h} \qquad \text{for all } h\in [H],$$ 
denoted by $\mathcal{MG}=\big\{ \cS, \{\cA_1, \cA_2\}, H, P, r \big\}$. 
Following standard convention, we assume that $r_{1,h}\geq 0$ for all $h\in [H]$,\footnote{Since $r_{2,h}=-r_{1,h}\le 0$, the reward normalization assumption is violated for player 2. This can be remedied by adding a constant shift to $r_{2,h}$, which typically does not affect the equilibrium strategies or the learning problem.} 
and refer to player 1 and player 2 as the \emph{max-player} and \emph{min-player}, respectively. See Figure~\ref{fig:MG} for an illustration.

\paragraph{Markov policies.}
Having specified the dynamics of a Markov game, we now turn to the policies used by the agents. Throughout this section, we focus on the class of \emph{Markov policies}, under which each agent's action depends only on the current state $s$ and the current step number $h$, and not on the previously visited states of the game.

Specifically, 
let $\pi_i=\{\pi_{i,h}\}_{1\leq h\leq H}$ represent the policy of agent $i$.  
Here, $\pi_{i,h}(\cdot \mymid s) \in \Delta(\cA_i)$ for any $(s,h)\in \cS \times [H]$,  
where $\pi_{i,h}(a \mymid s)$ denotes the probability that agent $i$ selects action $a$ when the game is in state $s$ at step $h$.  
A joint Markov policy can be defined analogously: 
we let $\pi= (\pi_1,\ldots, \pi_m): \cS  \times [H] \rightarrow \Delta(\cA)$  represent a joint Markov policy of all agents, 
where the joint actions of all agents in state $s$ and step $h$ are chosen according to the joint distribution specified by $\pi_h(\cdot \mymid s) = (\pi_{1,h},\ldots, \pi_{m,h})(\cdot\mymid s) \in \Delta(\cA)$. 
For any given joint policy $\pi$, we employ $\pi_{-i}$ to represent the policies of all agents except agent $i$, 
and similarly use $\pi_{-i,h}$ to denote the corresponding policies at step $h$.

%All policies are assumed throughout to be Markovian, except our brief remarks on non-Markovian policies in Section~\ref{sec:main-results}. 

Additionally, a joint policy $\pi$ is said to be a {\em product policy} if the agents act independently under $\pi$; equivalently, this means the joint action distribution factorizes across agents.  
In this case, we write $\pi=\pi_1\times \cdots \times \pi_m$ to indicate that $\pi$ is a product policy.

\paragraph{Value functions.} 
Consider a Markovian trajectory $\{(s_h,\bm{a}_h)\}_{1\leq h\leq H}$, where $s_h\in \cS$ is the state and $\bm{a}_h\in \cA$ is the joint action profile at step $h$. 
For any given joint policy $\pi$ and any step $h\in[H]$, 
the value function $V_{i,h}^{\pi}: \cS \rightarrow \mathbb{R}$ of agent $i$ is defined as
\begin{equation}
    \forall s\in \cS,
    \qquad
	V_{i,h}^{\pi}(s)\coloneqq\mathbb{E}\left[\sum_{t=h}^{H}r_{i,t}\big(s_{t}, \bm{a}_{t}\big)\mid s_{h}=s\right], 
\label{eq:value-function-defn}
\end{equation}
where the expectation is taken over the Markovian trajectory with the $m$ agents jointly executing policy $\pi$ (conditional on $s_h=s$). Similarly, we define the Q-function $Q_{i,h}^{\pi}: \cS \times \cA \rightarrow \mathbb{R}$ of the $i$-th agent under policy $\pi$ as follows
\begin{equation*}
\forall (s,a)\in \cS\times \cA,\quad
	Q_{i,h}^{\pi}(s, \bm{a})\coloneqq\mathbb{E}\left[\sum_{t=h}^{H}r_{i,t}\big(s_{t}, \bm{a}_{t}\big)\mid s_{h}=s, \bm{a}_h = \bm{a} \right],
%\label{eq:Q-value-function-defn}
\end{equation*}
which differs from $V_{i,h}^{\pi}(s)$ only in that the action at step $h$ is fixed to be $\bm a$. 

In addition, consider the case where all agents except agent $i$ follow the joint policy $\pi_{-i}$, while agent $i$ {\em independently} executes a policy $\pi'_i$.  We denote by $V_{i,h}^{\pi'_i \times \pi_{-i}}$ the value function resulting from this joint policy $\pi'_i \times \pi_{-i}$. 
Optimizing over all policies $\pi'_i$ of agent $i$, we further define 
\begin{equation}
	\label{eq:defn-optimal-V}
	V_{i,h}^{\star,\pi_{-i}}(s)\coloneqq\max_{\pi'_i: \cS\times [H] \rightarrow \Delta(\mathcal{A}_i)} V_{i,h}^{\pi'_i \times \pi_{-i}}(s) ,
	\quad \forall (s,h,i) \in \cS \times [H] \times [m].  
\end{equation}
It is well known that there exists at least one policy, denoted by $\pi_i^{\star}\big( \pi_{-i} \big): \cS\times [H] \rightarrow \Delta(\mathcal{A}_i)$, 
that can simultaneously attain $V_{i,h}^{\star,\pi_{-i}}(s)$ for all $h\in [H]$ and all $s\in \cS$. 
Such a policy is often referred to as a \emph{best-response policy}. 
Importantly, the best-response policy $\pi_i^{\star}\big( \pi_{-i} \big)$ 
is the best among all policies of the $i$-th agent executed {\em independently} of $\pi_{-i}$.

Furthermore, if we freeze $\pi_{-i}$, then the Bellman optimality condition for agent $i$ can be expressed as \citep{bertsekas2017dynamic}
\begin{equation*}
	V_{i,h}^{\star,\pi_{-i}}(s)
	=\max_{a_i\in \cA_i}\left\{ \mathop{\mathbb{E}}\limits _{\bm{a}_{-i}\sim \pi_{-i,h}(\cdot | s)}\left[r_{i,h}(s,\bm{a})+\Big\langle P_h(\cdot\mymid s, \bm{a}),\,V_{i,h+1}^{\star,\pi_{-i}}\Big\rangle\right]\right\} 
	%\label{eq:Bellman-equation-max}
\end{equation*}
for all $(s,h,i) \in \cS \times [H] \times [m]$, where the joint action profile $\bm{a}$ is composed of agent $i$'s action $a_i$ and the remaining agents' actions $\bm a_{-i}$. The terminal value function is set to be $V_{i,H+1}^{\star,\pi_{-i}}=0$.

\paragraph{Equilibria of Markov games.}
In a multi-agent Markov game, each agent wishes to maximize its own value function. 
Since agents may have competing objectives, identifying appropriate equilibrium solutions has become a central theme in the study of Markov games. Two of the most widely studied solution concepts are the Nash equilibrium (NE) \citep{nash1950equilibrium} and the coarse correlated equilibrium (CCE) \citep{moulin1978strategically,aumann1987correlated}, which we briefly introduce below.
\begin{itemize}
	\item {\em Nash equilibrium (NE).} A product policy $\pi=\pi_1\times \cdots \times \pi_m$ is 
said to be a {\em (mixed-strategy) Nash equilibrium} of $\mathcal{MG}$ if the following holds: 
\begin{equation}
	V_{i, 1}^{\pi}(s)=V_{i, 1}^{\star,\pi_{-i}}(s),
	\qquad\forall (s,i)\in\cS\times [m].
	\label{eq:defn-Nash-E}
\end{equation}
In other words, conditional on the current policies of all other agents and the assumption that all agents take actions {\em independently}, no agent can improve its value by unilaterally deviating from its current policy. 
Thus, under a Nash equilibrium, every agent is playing a best response to the policies of the opponents.

	\item {\em Coarse correlated equilibrium (CCE).}  A joint policy $\pi $ is said to be a {\em coarse correlated equilibrium} of $\mathcal{MG}$ if
\begin{equation}
	V_{i, 1}^{\pi}(s) \geq V_{i, 1}^{\star,\pi_{-i}}(s),
	\qquad\forall (s,i)\in\cS\times [m].
	\label{eq:defn-CCE}
\end{equation}
Like a Nash equilibrium, a CCE guarantees that no agent can benefit from a unilateral deviation performed independently of the others. The key distinction is that a CCE permits correlation among the agents' actions, whereas a Nash equilibrium requires the joint policy to factorize across agents. Consequently, every Nash equilibrium is also a CCE, although the converse does not generally hold.

\end{itemize}

For general-sum Markov games, computing a Nash equilibrium is well known to be computationally intractable in general (specifically, PPAD-hard) \citep{daskalakis2013complexity,daskalakis2009complexity}. This computational barrier motivates the study of alternative equilibrium notions, such as the CCE, which can often be computed efficiently via standard mathematical programming techniques. Another widely studied notion is the correlated equilibrium (CE) \citep{aumann1987correlated,greenwald2003correlated,papadimitriou2008computing}, which sits between NE and CCE; we omit a detailed discussion of the CE in this chapter.

\subsection{Problem formulation}

In many applications, the underlying Markov game is unknown and must be learned from data. As in single-agent RL, MARL has been studied under a variety of data collection mechanisms. To keep the exposition focused and concise, this section restricts attention to the generative-model setting. While idealized, this framework already captures some challenges unique to multi-agent scenarios and provides a useful starting point for understanding other important settings, including online Markov games \citep{tian2021online,mao2022provably,jin2021v,song2021can,daskalakis2023complexity,cui2023breaking,feng2024improving} and offline Markov games \citep{cui2022offline,cui2022provably,yan2022model}.

\paragraph{Sampling from a generative model.} 
Throughout this section, we assume access to a generative model. In each query, the learner may choose an arbitrary state-action-step tuple $(s,\bm a,h)\in\cS\times\cA\times[H]$ and obtain an independent sample
\[
	s' \overset{\mathrm{ind.}}{\sim} P_h(\cdot \mymid s, \bm{a}). 
\]
Here and throughout the section, we use boldface $\bm{a}$ to emphasize that it represents a joint action profile of all agents.

While this sampling model resembles its counterpart in single-agent RL, the resulting statistical challenges can be markedly different. In particular, whereas a nearly uniform allocation of samples across state-action-step tuples is often sufficient in single-agent RL, data-efficient learning in Markov games frequently requires highly adaptive sampling strategies, as we shall illustrate momentarily.

%As before,  a generative model facilitates the query of arbitrary state-action-step combinations, which helps alleviate the sampling constraints arising in online episodic settings for exploration.

\paragraph{Goal.} 
In practice, it might be challenging to compute an exact equilibrium, and instead one would seek to find approximate solutions. Accordingly, the goal of this section is to learn an $\varepsilon$-approximate equilibrium (either NE or CCE) of $\mathcal{MG}=\big\{ \cS, \{\cA_i\}_{1 \le i \le m}, H, P, r \big\}$ using as few samples as possible, or equivalently, with a minimal number of calls to the generative model.

To formalize this objective, we need to define the $\varepsilon$-approximate equilibrium. For any policy $\pi$, define its suboptimality gap --- measured in an $\ell_{\infty}$-based manner --- by
\begin{subequations}
\label{eq:defn-NE-gap}
\begin{equation}
	\NEgap(\pi) \coloneqq \max_{s\in\cS} \, \NEgap(\pi;s),
	\label{eq:defn-NE-gap-final}
\end{equation}
where
\begin{equation}
	\NEgap(\pi;s) \, \coloneqq \max_{1\leq i\leq m}\Big\{ V_{i, 1}^{\star,\pi_{-i}}(s)-V_{i, 1}^{\pi}(s)\Big\} .
	\label{eq:defn-NE-gap-s}
\end{equation}
\end{subequations}
With this definition in place, a {\em product} policy $\pi_{\mathsf{NE}}=\pi_1\times \cdots \times \pi_m$ is said to be an $\varepsilon$-approximate NE --- or more concisely, $\varepsilon$-Nash --- if the resultant suboptimality gap obeys $\NEgap(\pi_{\mathsf{NE}}) \leq \varepsilon$. 
Similarly, a joint (and possibly correlated) policy  $\pi_{\mathsf{CCE}}$ is said to be an $\varepsilon$-approximate CCE --- or more concisely, $\varepsilon$-CCE --- if $\NEgap(\pi_{\mathsf{CCE}}) \leq \varepsilon$. 
A key objective is therefore to characterize the sample complexity required to learn such $\varepsilon$-approximate equilibria.

\subsection{A model-based algorithm}
\label{sec:model-based-native-MARL}

A straightforward and natural model-based approach to equilibrium learning in Markov games is the plug-in estimator \citep{zhang2020model-based,shi2024sample}, which first constructs an empirical estimate of the underlying Markov game from data samples and then computes an equilibrium of the empirical Markov game using classical planning methods. More specifically, the procedure consists of the following two stages. 
\begin{itemize}
	\item[1)] {\em Model estimation.} 
		For each state-joint-action-step tuple  $(s,\bm{a},h )\in \cS\times \cA \times [H]$, draw $N$ independent samples from the generative model: 
		$$
			s_h^{(i)}(s, \bm{a}) \overset{\mathrm{i.i.d.}}{\sim} P_h(\cdot \mymid s, \bm{a}), \qquad i=1,\ldots, N.
		$$ 
		Using these samples, construct an empirical transition kernel $\widehat{P}$ according to
		\begin{equation*}
			\forall s' \in \mathcal{S}, \qquad \widehat{P}_h(s^\prime \mymid s, \bm{a}) = \frac{1}{N}\sum_{i=1}^N \ind \big\{  s^{\prime} = s_h^{(i)}(s, \bm{a}) \big\} 
			%\label{eq:defn-empirical-P-marl}
		\end{equation*}
        for every $(s,\bm{a},h )\in \cS\times \cA \times [H]$. 
		That is, $\widehat P_h(s'\mid s,\bm a)$ records the empirical frequency with which the tuple $(s,\bm a,h)$ transitions to state $s'$. This leads to the empirical Markov game $\widehat{\mathcal{MG}}=\big\{ \cS, \{\cA_i\}_{1 \le i \le m}, H, \widehat{P}, r \big\}$.

	\item[2)] {\em Planning.} Compute either a Nash equilibrium $\widehat{\pi}_{\mathsf{NE}}$ or a coarse correlated equilibrium $\widehat{\pi}_{\mathsf{CCE}}$ of $\widehat{\mathcal{MG}}$ using an appropriate planning oracle. For instance, NE in two-player zero-sum Markov games can be efficiently computed by Nash value iteration \citep{shapley1953stochastic}, while CCE in general-sum Markov games can be obtained using linear-programming-based methods \citep{papadimitriou2008computing}, no-external-regret learning dynamics \citep{roughgarden2016twenty}, among others. 
	
\end{itemize}

The above plug-in estimator provides perhaps the most direct way of extending model-based algorithms for single-agent RL (see Section~\ref{sec:finite-horizon-generative}) to the problem of equilibrium learning. A natural question is therefore: how many samples are required for this approach to compute an approximate equilibrium? The following theorem provides an answer.
\begin{theorem}[\citet{zhang2020model-based,shi2024sample}] \label{thm:sample-complexity-plug-in-marl}
Consider any $\delta \in (0,1)$. Let $\widehat{\pi}_{\mathsf{NE}}$ and $\widehat{\pi}_{\mathsf{CCE}}$ denote, respectively, the NE and CCE returned by the planning oracle in the above model-based plug-in algorithm. Then for any $0<\varepsilon \leq \sqrt{H}$, one has
    \begin{align*}
 \mathsf{gap}(\widehat{\pi}_{\mathsf{NE}}) \leq \varepsilon
 \qquad \text{and} \qquad
  \mathsf{gap}(\widehat{\pi}_{\mathsf{CCE}}) \leq \varepsilon
	\end{align*}
with probability at least $1-\delta$, provided that
\begin{align*}
	N  \geq  \frac{ C_1  H^3   }{  \varepsilon^2} \log\left(\frac{SHN \prod_{1\leq i\leq m} A_i  }{\delta} \right)
\end{align*}
for some large enough universal constant $C_1>0$. Here, the metric $\mathsf{gap}(\cdot)$ is defined in \eqref{eq:defn-NE-gap}.  
\end{theorem}

 Since the Markov game contains a total of $SH\prod_{i=1}^{m}A_i$ state-joint-action-step combinations,
 Theorem~\ref{thm:sample-complexity-plug-in-marl} implies that the above model-based plug-in algorithm can learn either an $\varepsilon$-NE or an $\varepsilon$-CCE with a sample complexity of
\begin{align}
\widetilde{O}\bigg( \frac{ S H^4 \prod_{1\leq i\leq m} A_i    }{  \varepsilon^2}   \bigg). \label{eq:upper-bound-order}
\end{align}
It is worth emphasizing that this theorem concerns only the statistical aspect of the plug-in approach and does not address its computational complexity. Indeed, while the planning oracle can be implemented efficiently in certain settings --- such as two-player zero-sum Markov games --- it becomes computationally intractable in general-sum Markov games when tasked with computing NE.

While appealing for its simplicity, the model-based plug-in estimator introduced above turns out to be quite expensive and, in general, statistically suboptimal. The main bottleneck stems from its dependence on the size of the joint action space $$\prod_{1\leq i\leq m} A_i,$$ 
which grows exponentially with the number of agents $m$ when the individual action spaces are of comparable size. Consequently, the sample complexity deteriorates rapidly as the number of agents increases, thereby limiting the applicability of this plug-in approach in applications involving many participants. 
This phenomenon is commonly referred to as the \emph{curse of multi-agency}. A natural question is therefore whether one can circumvent this curse and develop algorithms whose sample complexity exhibits a substantially milder dependence on the size of the joint action space. As we shall see shortly, the answer is often affirmative.

%Combining this with the lower bound in \eqref{eq:sample-size-lower} of Theorem~\ref{thm:lower-marl} confirms that the sample complexity of the plug-in approach is optimal with respect to many salient factors, including $\varepsilon$, $S$ and $H$. Nonetheless, it leaves open an important question: {\em can we break the curse of multi-agents and remove the dependency on $\prod_{1\leq i\leq m} A_i $ in the sample complexity?}

\subsection{Breaking the curse of multi-agency}

Motivated by the goal of overcoming the curse of multi-agency in Markov games, a recent line of work \citep{mao2022provably,song2021can,jin2021v,li2022minimax} sought to design equilibrium learning algorithms that can achieve a significantly improved sample complexity than the plug-in estimator discussed in Section~\ref{sec:model-based-native-MARL}. We discuss one such algorithm, termed \texttt{Q-FTRL}, developed by \citet{li2022minimax} for the generative-model setting.

At a high level, \texttt{Q-FTRL} combines two powerful ideas from the RL and online learning literature: optimism in the face of uncertainty, implemented through upper confidence bounds (UCBs) as in online RL (see Chapter~\ref{chapter:online-RL}), and online adversarial learning via the Follow-the-Regularized-Leader (\texttt{FTRL}) framework \citep{shalev2012online}. A key feature of the algorithm is its use of adaptive sampling, which directs data collection toward statistically informative regions of the state-joint-action space from each agent's perspective independently, and thereby mitigates the curse of multi-agency. This stands in sharp contrast to the single-agent generative-model setting discussed in Chapter~\ref{chapter:generative-model}, where a uniform, non-adaptive allocation of samples is often sufficient.  More broadly, \texttt{Q-FTRL} reveals how ideas originating from online RL (see Chapter~\ref{chapter:online-RL}) can play a pivotal role even in the generative-model setting. 

Since the \texttt{Q-FTRL} algorithm is considerably more sophisticated than the model-based methods introduced in previous chapters, we omit full algorithmic details and refer the interested reader to \citet{li2022minimax} for a complete description. In the sequel, we present its statistical guarantees and discuss their implications.

\begin{theorem}[\citet{li2022minimax}]
\label{thm:Q-FTRL}
Consider any $\delta \in (0,1)$ and any $\varepsilon \in (0, H]$. There exists a model-based algorithm (i.e., \texttt{Q-FTRL}) that, with probability at least $1-\delta$, achieves the following.
\begin{itemize}
    \item For a two-player zero-sum Markov game, it can find a policy $\widehat{\pi}_{\mathsf{NE}}$ satisfying
      \begin{align*}
 \mathsf{gap}(\widehat{\pi}_{\mathsf{NE}}) &\leq \varepsilon, 
	\end{align*}   
    with a total sample complexity of
    $$ \widetilde{O}\bigg( \frac{SAH^4}{\varepsilon^2} \bigg).$$

        \item For a general-sum Markov game, it can find a policy $\widehat{\pi}_{\mathsf{CCE}}$ satisfying
      \begin{align*}
 \mathsf{gap}(\widehat{\pi}_{\mathsf{CCE}}) &\leq \varepsilon
	\end{align*}   
        with a total sample complexity of
    $$ \widetilde{O}\bigg( \frac{H^4S\sum_{i=1}^m A_i}{\varepsilon^2} \bigg).$$
\end{itemize}
\end{theorem}

Theorem~\ref{thm:Q-FTRL} 
demonstrates that \texttt{Q-FTRL} manages to break the curse of multi-agency. In contrast to the plug-in estimator described in Section~\ref{sec:model-based-native-MARL}, whose sample complexity is proportional to $\prod_{i=1}^{m}A_i$,  the sample complexity of \texttt{Q-FTRL} depends only on the total number of individual actions, i.e., $\sum_{i=1}^{m}A_i$. 
Consequently, when the action spaces of different agents are of comparable size, \texttt{Q-FTRL} reduces an exponential dependence on the number of agents to a linear one, resulting in a dramatic improvement in data efficiency.  
Another noteworthy feature is that such statistical guarantees hold uniformly over the entire accuracy range $\varepsilon\in(0,H]$, making the algorithm provably appealing even in sample-starved scenarios.

%establishes the sample complexity upper bounds for the Q-FTRL algorithm, which asserts that, when a generative model is available, the total number of samples (i.e., $KSH\sum_i A_i$) needed for Q-FTRL
%to work is  $$ \widetilde{O}\left( \frac{H^4S\sum_{i=1}^m A_i}{\varepsilon^2} \right)$$ for  learning an $\varepsilon$-NE (resp. $\varepsilon$-CCE) in two-player zero-sum (resp. multi-player general-sum) MGs.
%This reveals that what ultimately matters is the total number of individual actions (i.e., $\sum_{i=1}^m A_i$) as opposed to the total number $\prod_{i=1}^m A_i$ of possible joint actions. 

To assess the optimality of the above sample complexity guarantees, we next present a minimax lower bound for learning Markov games (w.r.t.~finding either $\varepsilon$-NE or $\varepsilon$-CCE) as follows.
\begin{theorem}
\label{thm:lower-marl} 
Consider any $m\geq 2$ and any $0<\varepsilon \leq c_1 H$ for some small enough constant $c_1>0$. 
Then one can construct a collection of $m$-player zero-sum Markov games $\{\mathcal{MG}_{\theta} \mid \theta \in \Theta\}$ with $S$ states, horizon $H$, and $A_i$ actions for the $i$-th player ($1\leq i\leq m$) such that
\begin{align*}
	\inf_{\widehat{\pi}} \max_{\theta\in\Theta} \mathbb{P}_{\mathcal{MG}_{\theta}}\left\{ \NEgap\big(\widehat{\pi}\big) > \varepsilon\right\} \geq \frac{1}{4},
\end{align*}
provided that the total sample size is smaller than
\begin{align}
	%N_{\mathsf{total}} \leq 
    \frac{c_2 H^4S \max_{1\leq i\leq m}A_i }{\varepsilon^2} \label{eq:sample-size-lower}
\end{align}
for some sufficiently small constant $c_2>0$. Here, the infimum is over all (joint) policy estimators $\widehat{\pi}$, and $\mathbb{P}_{\mathcal{MG}_{\theta}}$ denotes the probability measure under the Markov game $\mathcal{MG}_{\theta}$. 
\end{theorem}
%
%Crucially, the sample complexity scales with the size of the largest action space $\max_{1\leq i\leq m}A_i $, rather than the size of the joint action space $\prod_{1\leq i\leq m}A_i $, allowing scalability to a large number of agents in MARL. The lower bound is significantly smaller than the size of the transition kernel $ S^2 \prod_{1\leq i\leq m}A_i $ or the size of the Q-function $S \prod_{1\leq i\leq m}A_i$, raising interesting questions regarding how to develop algorithms to achieve the lower bound.

Combining Theorem~\ref{thm:Q-FTRL} with the minimax lower bound in Theorem~\ref{thm:lower-marl} establishes the minimax optimality of  \texttt{Q-FTRL} (up to logarithmic factors) when the number $m$ of agents is fixed. Notably, \texttt{Q-FTRL} is provably minimax optimal for learning Nash equilibria in the important special case of two-player zero-sum Markov games.%; in fact, it is currently the only algorithm known to attain this guarantee, further highlighting the statistical power of model-based approaches in equilibrium learning. 

An intriguing open question is whether the sample complexity of \texttt{Q-FTRL}, or related algorithms, can be further improved to match the lower bound \eqref{eq:sample-size-lower}. Since linear dependence on $\max_{1\le i\le m} A_i$ has already been shown to be attainable in online MARL via V-learning \citep{song2021can,jin2021v} --- albeit at the cost of suboptimal dependence on the horizon length --- we conjecture the lower bound \eqref{eq:sample-size-lower} to be tight (up to logarithmic factors) and hence captures the fundamental statistical limit of equilibrium learning in Markov games.

%% file: chapters/federated_RL.tex
\section{Federated RL}

Collecting, through a single agent, an enormous amount of data required by RL is often prohibitively time-consuming and, in many applications, infeasible. As a result, practical implementations of RL systems often deploy multiple agents to interact with the environment and gather data in parallel. 
This decentralized data-collection paradigm has motivated the development of distributed and federated RL algorithms, enabling agents to collaboratively learn an optimal policy 
without actually communicating the collected data to a centralized server. Ideally, such methods would achieve a linear speedup with respect to the number of participating agents.

This federated approach to RL has attracted growing attention, as it significantly reduces bandwidth requirements while offering improved privacy and security guarantees compared to fully centralized approaches. At the same time, the federated setting introduces new statistical and algorithmic challenges that are absent in classical single-agent RL.  In this section, we use federated Q-learning as a representative example to illustrate some of these challenges and discuss the fundamental statistical questions arising due to communication constraints.

\subsection{Federated synchronous Q-learning}

As a case study, we consider federated synchronous Q-learning, in which all state-action pairs are updated simultaneously, and each agent is assumed to have access to a generative model (or simulator). This provides a natural framework for studying the interplay between communication constraints and statistical efficiency in federated RL.

\paragraph{Problem setup.} 
Consider a federated RL setting with $K$ agents interacting with the same discounted infinite-horizon MDP $\mathcal{M}=\{\cS,\cA,P,r,\gamma\}$. We focus on the \emph{synchronous} setting, where each agent $k\in[K]$ has access to the generative model of the underlying MDP. At iteration $t$, for every state-action pair $(s,a)\in\cS\times\cA$, each agent $k$ can independently query the generative model and obtain a fresh sample
\begin{equation}\label{eq:sync_sampling}  
s_t^k(s,a) \overset{\text{ind.}}{\sim} P(\cdot\mymid s,a).
 \end{equation}
The agents perform local updates using their own samples and may periodically synchronize their estimates through a central server. The goal is to collaboratively learn the optimal Q-function $Q^\star$, while reducing the communication overhead associated with synchronizing the agents' local estimates.

\paragraph{Algorithm description: \fedsyncq.} We now describe a natural federated synchronous Q-learning algorithm, dubbed \fedsyncq, which alternates between local Q-learning updates and periodic synchronization across agents. The complete procedure is summarized in Algorithm~\ref{alg:dist-syncq-local}. 

The \fedsyncq algorithm initializes, at each agent $k \in [K]$, a local Q-function estimate $Q_0^k = Q_0$.  At the beginning of each iteration $t\geq 1$,  agent $k$ maintains a local Q-function estimate $Q_{t-1}^{k}$ together with the corresponding value function estimate $V_{t-1}^k$:
\begin{align*}	
  \forall s\in \cS: \qquad
  V_{t-1}^k(s) :=  \max_{a\in \cA} Q_{t-1}^k(s,a).
\end{align*} 
Each iteration $t$ performs the following two steps.
\begin{enumerate}  
\item {\em Local updates.} Each agent $k$ first updates \emph{all} entries of its local Q-function estimate $Q_{t-1}^k$ independently using the standard synchronous Q-learning update rule, yielding the following \emph{intermediate} Q-function estimate:
\begin{align} \label{eq:syncq-local}
   Q_{t-\frac{1}{2}}^k(s, a) &= (1-\eta) Q_{t-1}^k(s,a) + \eta \left( r(s,a) + \gamma   V_{t-1}^k\big(s_t^k(s,a) \big)  \right) 
\end{align}
for all $(s,a) \in \cS \times \cA$, where $s_t^k(s,a)$ is generated according to \eqref{eq:sync_sampling}, and $0<\eta<1$ is the learning rate.

\item {\em Periodic synchronization and averaging.} Every $\tau\geq 1$ iterations --- where $\tau$ denotes the synchronization period --- the agents synchronize their local Q-function estimates through a central server. At each synchronization round, the server averages the intermediate Q-function estimates $\{Q_{t-\frac{1}{2}}^k\}_{k=1}^K$ and broadcasts the resulting estimate back to all agents. Consequently, the updated Q-function estimate at the end of iteration $t$ can be expressed as
\begin{align} \label{eq:syncq-periodic}
 Q_{t}^k(s,a) =
  \begin{cases}
    \frac{1}{K} \sum_{k=1}^K Q_{t-\frac{1}{2}}^k(s,a),\quad  &~\text{if}~ t \equiv 0 \; (\text{mod}~ \tau)\\
    Q_{t-\frac{1}{2}}^k(s,a), &~\text{otherwise}
  \end{cases}
\end{align}
for all $(s,a) \in \cS \times \cA$.
\end{enumerate}
After $T$ iterations, the \fedsyncq algorithm outputs the averaged Q-function estimate
$${Q}_{T}=\frac{1}{K}\sum_{k=1}^K Q_T^k.$$ 
Without loss of generality, we assume that $T$ is divisible by the synchronization period $\tau$. The resulting communication complexity can be measured by $$C_{\mathsf{round}} = T/\tau,$$  
which corresponds to the total number of communication rounds.

\begin{algorithm}[t]
  \DontPrintSemicolon
  \textbf{inputs:} discount factor $\gamma$, reward function $r$, learning rate $0<\eta<1$, number of agents $K$, synchronization period $\tau$, number of iterations $T$, initial guess $Q_0$. \\
  \textbf{initialize }$Q_0^k \leftarrow Q_0$ for all $k$. \\
    \For{$t=1,\cdots,T$}{
    \For{$k =1,\cdots,K$}{    
    {draw  $s_t^k(s,a) \overset{\mathrm{ind.}}{\sim}  P(\cdot \mymid s,a)$ for every $(s,a)\in \cS\times \cA$. }
    
    {compute $Q_{t-\frac{1}{2}}^k$ according to \eqref{eq:syncq-local}.}
    
    {compute $Q_t^k$ according to \eqref{eq:syncq-periodic}.}
    }
    }
    
    \textbf{return:} ${Q}_{T}=\frac{1}{K}\sum_{k}Q_T^k$. 

  \caption{Federated synchronous Q-learning (\fedsyncq)}
  \label{alg:dist-syncq-local}
\end{algorithm}

\paragraph{Theoretical guarantees.}
We are now in a position to establish non-asymptotic performance guarantees for \fedsyncq. %Algorithm~\ref{alg:dist-syncq-local}.
\begin{theorem}[\cite{woo2023blessing}] \label{thm:dist-syncq-local}
Consider any given $\delta \in (0,1)$ and $\varepsilon\in \big(0, \frac{1}{1-\gamma}\big]$. 
 Suppose that the initialization satisfies $0 \leq Q_0  \le \frac{1}{1-\gamma}$,  
and the synchronization period $\tau$ obeys
  \begin{subequations}
\begin{align}\label{eq:sync_period}
 \tau \le  1+ \frac{1}{\eta} \min \left\{  \frac{1-\gamma}{8\gamma}, \, \frac{1}{K} \right\}. 
 \end{align}
There exists some sufficiently large (resp.~small) constant $c_T>0$ (resp.~$c_\eta>0$) such that, with probability at least $1-\delta$, Algorithm~\ref{alg:dist-syncq-local} achieves $\| Q_T - Q^{\star}\|_{\infty}\leq \varepsilon$, provided that %the sample size per agent $T$ and the learning rate $\eta$ satisfy
\begin{align}\label{eq:guarantee_sync}
  T
  &\ge  \frac{c_T}{K(1-\gamma)^5 \varepsilon^2} %(\log((1-\gamma)^2\varepsilon))^2 
  \log^3 \bigg({\frac{SAK T}{\delta \varepsilon (1-\gamma)^2}}\bigg), \\
 \eta & =  \frac{c_{\eta}        K (1-\gamma)^4 \varepsilon^2}{\log{\frac{SAK T}{\delta}}}. \label{eq:lr_sync}
\end{align}
\end{subequations}
\end{theorem}

\paragraph{Implications.} 
Theorem~\ref{thm:dist-syncq-local} implies that, to obtain an $\varepsilon$-accurate estimate of the optimal Q-function in an $\ell_\infty$ sense,  each agent requires at most
\begin{align}
\widetilde{O}\left( \frac{SA}{K(1-\gamma)^5 \varepsilon^2}   \right)
\label{eq:sample-size-per-agent-fedQ}
\end{align}
samples, where each agent draws $SA$ samples at each iteration. This theorem yields several implications as follows. 
\begin{itemize}
\item \textit{Linear speedup.} The per-agent sample complexity in \eqref{eq:sample-size-per-agent-fedQ} enjoys a linear improvement with respect to the number of agents $K$. For comparison, recall from Chapter~\ref{chapter:generative-model-MF} that the sharp sample complexity of single-agent synchronous Q-learning is  $\widetilde{\Theta}\left( \frac{SA}{(1-\gamma)^4 \varepsilon^2 } \right)$
for $\varepsilon \in (0,1)$. Consequently, the federated setting provably achieves faster learning whenever
$ K \gtrsim \frac{1}{1-\gamma} $
up to logarithmic factors, provided that $\varepsilon \in (0,1)$.

\item \textit{Communication efficiency.} A key feature of this federated setting is the use of periodic averaging to reduce communication costs. According to \eqref{eq:sync_period}, the above theory requires the synchronization period $\tau$ to scale inversely with the learning rate $\eta$, suggesting that more frequent communication may be needed to compensate for the greater discrepancy of local updates under larger learning rates. To gain further insight, consider the regime where $K\gtrsim \frac{1}{1-\gamma}$ and $\varepsilon \lesssim \frac{1}{K(1-\gamma)^2}$. Substituting the learning rate \eqref{eq:lr_sync} into the upper bound  \eqref{eq:sync_period} suggests a feasible synchronization period as
$$\tau =\widetilde{\Theta}\bigg( \frac{1}{K^2(1-\gamma)^4\varepsilon^2}\bigg).$$ 
This leads to a communication complexity no larger than
$$ C_{\mathsf{round}}  = \frac{T}{\tau} \le \widetilde{O}\bigg(  \frac{K}{1-\gamma} \bigg),$$
which is nearly independent of the target accuracy level $\varepsilon$.

\end{itemize} 

\noindent 
 Finally, we note that the communication efficiency of \fedsyncq can be further improved by employing more orchestrated communication schedules; see, e.g.,  \citet{woo2024federated,salgia2024sample}.

 \subsection{Trade-offs between statistical and communication efficiency}
 
 A central challenge in the design of federated RL algorithms, including federated Q-learning, is to address the natural tension between sample and communication complexities. 
 At one extreme, one may adopt a fully centralized approach that gathers all locally collected data at a central server and then runs a statistically optimal RL algorithm using all samples. While this na\"ive strategy trivially achieves the optimal sample complexity, it incurs a very large --- and often prohibitive --- communication cost. In contrast, several recently proposed federated RL algorithms (e.g., \citet{khodadadian22fedrlspeedup,woo2023blessing,woo2024federated}) operate in a more communication-efficient fashion. By reducing the frequency of synchronization between agents and a central server, these methods significantly lower communication costs relative to the fully centralized approach, albeit often at the expense of suboptimal sample complexity. 
 
 These results hint at the existence of a fundamental trade-off between sample and communication complexities in federated RL. To fully appreciate the statistical benefits from collaboration, one must understand the price of collaboration, i.e., the minimum amount of communication required to realize such gains. In what follows, we study this trade-off within the class of federated Q-learning algorithms with intermittent communication, which encompasses various communication-efficient schemes.

\paragraph{Intermittent-communication protocols.}
Consider a popular class of intermittent communication protocols in federated learning.  Under such protocols, agents communicate intermittently --- namely, only at selected synchronization rounds --- sharing their updated estimates of $\starQ$ either with one another or through a central server. Between communication rounds, each agent performs local stochastic-approximation-type updates using only its locally available data, much like a single-agent Q-learning algorithm. 
As such, intermittent communication provides a natural framework to extend single-agent Q-learning to the federated setting while reducing communication costs. This class of protocols has been extensively studied in federated learning and distributed optimization and has played a central role in establishing communication lower bounds for distributed stochastic optimization problems~\citep{Woodworth2021IntermittentComm}.

A generic federated Q-learning algorithm with intermittent communication is outlined in Algorithm~\ref{alg:general_fed_alg_with_inter_comm}. This generic algorithm is specified by five parameters:
\begin{itemize}
\item[(i)] the total number of updates $T$; 
\item[(ii)] the number of communication rounds $C_{\mathsf{round}} $; \item[(iii)] a step size schedule $ \{\eta_t\}_{t = 1}^T$; 
\item[(iv)] a communication schedule $\cC = \{t_r\}_{r= 1}^R$; 
\item[(v)] a batch size $B$. 
\end{itemize}
At iteration $t$, each agent $k$ computes $\{\widehat{\cT}_{b}(Q_{t-1}^{k})\}_{b = 1}^B$, 
a minibatch of i.i.d. sample Bellman operators $\widehat{\cT}_{b}(\cdot)$, $b=1,\ldots, B$, applied to its current estimate $Q_{t-1}^k$, using $B$ samples from the generative model for each $(s,a)$ pair, and obtains an intermediate local estimate using the Q-learning update rule as follows:
\begin{align}
    Q_{t-\frac{1}{2}}^k = (1- \eta_t)Q_{t-1}^k + \frac{\eta_t}{B} \sum_{b = 1}^B \widehat{\cT}_{b}(Q_{t-1}^{k}). \label{eqn:generic_algo_update}
\end{align}
Here, $\eta_t \in (0,1]$ denotes the learning rate at iteration $t$. The resulting intermediate estimates are synchronized and averaged according to the communication schedule $\cC = \{t_r\}_{r = 1}^{C_{\mathsf{round}}}$ (which consists of $C_{\mathsf{round}}$ synchronization rounds). Specifically, 
\begin{align}
    Q_{t}^k = \begin{cases} \frac{1}{K} \sum_{j = 1}^K Q_{t-\frac{1}{2}}^j,\quad  & \text{if } t \in \cC, \\ Q_{t-\frac{1}{2}}^k, & \text{otherwise.} \end{cases}
    \label{eqn:generic_algo_averaging}
\end{align}
Here, the averaging step may be replaced by any distributed mean estimation routine, including variants that include compression to control the bit-level costs.  Without loss of generality, we assume $t_{C_{\mathsf{round}}} = T$, i.e., the final iterates are always synchronized. Evidently, the number of samples taken per agent for each state-action pair is $N=BT$.

\begin{algorithm}[t]
    \caption{A generic federated Q-learning algorithm $\sA$}
    % \caption{A generic algorithm $\sA$}
    \label{alg:general_fed_alg_with_inter_comm}

    \textbf{input:} discount factor $\gamma$, reward function $r$, 
    number of iterations $T$, learning rates $\{\eta_t\}_{t = 1}^T$, number of synchronization rounds $R$,  synchronization schedule $\cC = \{t_r\}_{r = 1}^R$, batch size $B$. \\
    \textbf{initialize }$Q_0^k \leftarrow 0$ for all $1\leq k\leq K$. \\
        \For{$t = 1,2,\dots, T$}{
        \For{$k \in [K]$}{
        % \STATE Perform a Q-learning update
        % \STATE If $t \in  \cC$, average the local estimates
        {compute $Q_{t-\frac{1}{2}}^k$ according to \eqref{eqn:generic_algo_update}.}
        
        compute $Q_{t}^k$ according to \eqref{eqn:generic_algo_averaging}.
        
        }
        }
   
   \textbf{return} $\widehat{Q} = Q_T$.
\end{algorithm}

 \begin{theorem}
    Assume that $\gamma \in [5/6, 1)$, $S \geq 4$ and $A \geq 2$. Let $\sA$ be a federated Q-learning algorithm with intermittent communication, as described in Algorithm~\ref{alg:general_fed_alg_with_inter_comm}. Suppose that $\sA$ is run for $T \geq \max\{16,\frac{1}{1-\gamma}\}$ iterations with either a linearly rescaled learning rate schedule $\eta_t = \frac{1}{1 + c_{\eta}(1-\gamma)t}$ or a constant learning rate schedule $\eta_t = \eta$ for all $ 1\leq t \leq T$. If
    \begin{align*}
        C_{\mathsf{round}} \leq \frac{c_0}{(1- \gamma) \log^2 N} %\qquad C_{\textsf{bit}}(\sA; N) \leq \frac{c_1 SA}{(1- \gamma) \log^2 N}
    \end{align*}
    for some universal constant $c_0 > 0$, then for any choices of communication schedule, batch size $B$, $c_{\eta} > 0$ and $\eta \in (0,1)$, there exists an instance of MDP such that the output of $\sA$ obeys
    % and $d \geq 3$, there exists a $\gamma$-contractive operator $\cT:\R^d \to \R^d$ and a corresponding stochastic oracle $\widehat{\cT}$ satisfying Assumption~\ref{ass:operator_noise}, such that when $\sA$ is run for $T \geq 3$ steps with a batch size of $B \geq 1$ and any choice of communication schedule, $c_{\eta} > 0$ and $\eta \in (0,1)$, the error rate of $\sA$ satisfies
    \begin{align*}
\mathbb{E}\big[\big\|\widehat{Q}  - \starQ \big\|_{\infty}\big] \geq  \frac{C_{\gamma}}{\sqrt{N} \log^3 N}
    \end{align*}
    for all $K \geq 2$ and $N = BT$. Here, $C_{\gamma} > 0$ is a quantity that depends only on $\gamma$.
    \label{thm:lower_bound}
\end{theorem}

 The above theorem unveils a fundamental communication requirement for federated Q-learning with intermittent communication. Specifically, for any such algorithm to obtain \emph{any} statistical benefit from collaboration --- namely, for its error rate to improve as the number of agents increases --- it must perform at least $\widetilde{\Omega} \big( \frac{1}{1- \gamma } \big)$ rounds of communication and transmit $\widetilde{\Omega} \big( \frac{SA}{1- \gamma  }\big)$ bits of information per agent, both of which scale linearly with the effective horizon $\frac{1}{1-\gamma}$ of the underlying MDP. The above lower bound on the communication complexity also immediately applies to federated Q-learning algorithms that offer order-optimal sample complexity and thereby a linear speedup with respect to the number of agents.
Therefore, this theorem characterizes a fundamental converse for the statistical-communication trade-off in federated Q-learning.

The communication lower bound stems from the bias-variance trade-off underlying the convergence of Q-learning, which we discussed thoroughly in Section~\ref{sec:Q_learning_suboptimality}. In the fully centralized setting, this trade-off can be achieved through an appropriate choice of step sizes. In the federated setting, however, the communication schedule becomes an additional factor.
In particular, the local updates performed between two communication rounds induce a positive estimation bias that depends on the standard deviation of the iterates, a manifestation of the well-documented over-estimation issue in Q-learning (see, e.g.,  \citet{hasselt2010double} and the lower bound of Q-learning in Theorem~\ref{thm:LB-example-Q-generative}). Since this bias arises solely from local updates, it does not reflect any benefit of collaboration.  Communication mitigates this effect by averaging the local iterates across agents, thereby reducing the effective variance of the updates. When communication is too infrequent, however, the accumulated local bias dominates, and averaging of iterates is insufficient to offset the impact of the positive bias induced by the local steps. Consequently, the algorithm may fail to realize any statistical gains from collaboration when communication is too infrequent. 

 \begin{remark}[Tightness of the communication lower bound] 
\citet{salgia2024sample} developed a companion variance-reduced federated Q-learning algorithm that simultaneously achieves order-optimal sample and communication complexities, thereby establishing the tightness of the communication lower bound presented above.  
\end{remark} 
%Our argument is inspired by the analysis of Q-learning in~\cite{li2024q} and is based on analyzing the convergence of an intermittent communication algorithm on a specifically chosen ``hard'' MDP instance. 

% At one end of the spectrum lies the na\"ive approach of running a centralized algorithm with an optimal sample complexity after transferring and combining all the collected data at a central server. Such an approach trivially achieves the optimal sample complexity while suffering from a very high and prohibitive communication complexity. On the other hand, several recently proposed algorithms~\citep{khodadadian22fedrlspeedup,woo2023blessing,woo2024federated} operate in more practical regimes, offering significantly lower communication complexities when compared to the na\"ive approach at the cost of sub-optimal sample complexities. These results hint at the potential existence of an underlying trade-off between sample and communication complexities of federated RL algorithms. Therefore, it is of great interest to better understand this trade-off in the context of federated Q-learning by investigating the following fundamental question: {\em What is the minimum amount of communication required by a federated Q-learning algorithm to achieve any statistical benefit of collaboration?} 

%% file: chapters/inference.tex
\section{Statistical inference in RL}
\label{sec:UQ-RL}

% In Reinforcement Learning (RL), a discipline that underpins many recent machine learning breakthroughs, one central question is to evaluate with confidence the quality of a given policy, measured by its value function.  

%As RL is often modeled as decision-making in MDPs, the task of statistical inference needs to accommodate the online nature of the sampling mechanism. 

While most of this monograph has focused on learning value functions and near-optimal policies, many practical RL applications also require reliable measures of uncertainty to assess the quality of these estimates \citep{shi2022statistical,zhang2025estimation,shi2025statistical}.
%,zhou2026reinforcement 
For instance, constructing valid confidence intervals for value functions is often an important step in supporting downstream decision-making, not to mention its key role in the design of UCB- and LCB-based algorithms discussed in previous chapters. 
Achieving statistically efficient inference and uncertainty quantification typically requires establishing a distributional theory for the estimators of interest. Motivated by the high-dimensional nature of modern RL problems, a substantial body of recent work has focused on developing non-asymptotic distributional guarantees that remain valid well beyond classical asymptotic regimes. Moreover, since many RL algorithms produce estimates through iterative updates, statistical inference often needs to be developed in a manner that accommodates the online, iterative nature of the learning procedure.

We note that statistical inference for RL is a broad and rapidly growing area; see \citet{shi2025statistical} for a recent overview. 
In this section, rather than attempting a comprehensive treatment, we use TD learning as a case study to illustrate some of the key ideas in enabling statistical inference and non-asymptotic distributional theory for RL.  Recall that in Section~\ref{sec:model-free-policy-eval-sim}, we have established non-asymptotic convergence guarantees of synchronous TD learning and showed that its sample complexity in the generative-model setting achieves minimax-optimal rates \citep{li2024q}; such results, however, typically suppress pre-constants and logarithmic factors, both of which can have a substantial impact on the accuracy of the resulting confidence intervals. A complementary thread of work focused on the asymptotic behavior of TD learning, demonstrating that, when combined with Polyak-Ruppert averaging \citep{polyak1992acceleration,ruppert1988efficient}, the TD iterates converge to a Gaussian distribution in the asymptotic large-sample limit  (e.g., \citet{fort2015central,mou2020linear,borkar2025ode,li2023statistical}). 
While informative, such asymptotic results may not directly address finite-sample inference. In this section, we review some recent developments that extend these results to develop non-asymptotic distributional guarantees for the finite-sample regime.

%Consequently, these results were inadequate for statistically efficient finite-sample inference, thus motivating the development of non-asymptotic distributional theory that enables statistical inference for TD learning and other RL algorithms. In this section, we present some recent developments along this direction. 

%In this section, based on the TD learning algorithm, we design confidence regions and simultaneous confidence intervals for the value function in the tabular case, and discuss extensions of this approach to the linear coefficients of the value function when it is approximated by a linear combination of features. 

% \begin{figure*}[!ht]
% \begin{center}
% \begin{tabular}{cc}
% \includegraphics[width =\textwidth]{imgs/d3_alpha067_Delta1_asyVsemp.pdf}\\
% \end{tabular}
% \end{center}
% \caption{Emergence of the first entry of TD estimation error on an MDP with a 3-dimensional feature map.}
% % \yuting{if we don't have space, we can include 3 of them instead 4}
% \label{fig:iid-d3-alpha067-deltaT1-converge}
% \end{figure*}

% To achieve this goal, we describe refined high-dimensional Berry-Esseen bounds, which control the difference between the finite-sample and asymptotic distributions of the TD estimation error over the class of convex sets. 
% And we propose a plug-in estimator for the asymptotic covariance matrix, improving upon other methodologies proposed in the recent literature \citep{chen2020statistical,white2010interval,hanna2017bootstrapping,hao2021bootstrapping,ramprasad2022online, samsonov2024gaussian} in both computational and statistical efficiency. 

\subsection{Setting and algorithm}

%Although state-action-reward tuples are often generated sequentially along a Markovian trajectory, 
For concreteness, we focus on a simple variant of TD learning for estimating the value function $V^\pi$ of a fixed policy $\pi$ in a discounted infinite-horizon MDP $\mathcal{M}=(\cS,\cA,P,r,\gamma)$. Akin to Section~\ref{sec:model-free-policy-eval-sim}, we consider here an idealized setting in which the agent has access to a sequence of i.i.d.~samples; specifically, the data sequence consists of
\begin{subequations}
	\label{eq:iid-samples}
\begin{align}
	 \big\{(s_t,a_t,s_t',r_t)\big\}_{t=1}^T, 
\end{align}
where, for each $1\le t\le T$, the tuple $(s_t,a_t,s_t',r_t)$ is generated independently according to
\begin{equation}
    s_t \sim \mu, 
    \quad 
    a_t \sim \pi(\cdot\mymid s_t),\quad r_t = r(s_t,a_t), \quad s_t' \sim P(\cdot \mymid s_t, a_t)
\end{equation}
\end{subequations}
for some state distribution $\mu \in \Delta(\cS)$ satisfying $\lambda_0 = \min_{s \in S}\mu(s) >0$. In practice, the state-action-reward tuples are often generated sequentially along a Markovian trajectory; here, we adopt the i.i.d. setting to simplify the exposition and avoid additional technical complications. 
The TD learning variant considered in this section maintains an estimate $V_t:\cS\to\mathbb{R}$ of the target value function and utilizes one sample tuple in each iteration. At iteration $t$, it updates the value estimate according to
\begin{align}
\label{eqn:td-learning-inference-2}
	V_t (s_{t}) &= (1- \eta_t ) V_{t-1} (s_{t}) + \eta_t \big( r_t + \gamma   V_{t-1}(s_{t}') \big), 
\end{align}
where $0<\eta_t<1$ denotes the learning rate at iteration $t$.

As discussed in Section~\ref{sec:model-free-policy-eval-sim}, TD learning can be viewed as a stochastic approximation procedure for finding the fixed point of the Bellman operator. A classical technique for reducing the variability of the iterates and improving asymptotic statistical efficiency is Polyak-Ruppert averaging  \citep{polyak1992acceleration,ruppert1988efficient}, which forms the final estimate by averaging all iterates (e.g., \citet{mou2020linear,mou2023optimal,durmus2025finite,borkar2025ode,srikant2025rates,wu2024statistical}):
\begin{align}
	\label{eq:TD-averaging-inference}
	\overline{V}_T & =\frac{1}{T}\sum_{t=1}^{T} V_t. 
\end{align}
A key step toward statistical inference for TD learning is to characterize the distribution of the value estimation error 
\begin{equation}
\label{eq:defn-bar-deltat}
    \overline{\Delta}_T \coloneqq \overline{V}_T-V^{\pi}
\end{equation}
of the Polyak-Ruppert averaged TD estimator.

\subsection{Distributional theory of TD learning}

Classical theory in stochastic approximation \citep{polyak1992acceleration,fort2015central,borkar2025ode} established that, with an appropriate choice of the learning rate schedule, the averaged iterates of a broad class of stochastic approximation algorithms --- including TD learning --- converge asymptotically to a Gaussian distribution as $T\to\infty$ with all other problem parameters held fixed. Specializing to the averaged TD estimator, this asymptotic normality says
\begin{align}
\label{eqn:classic-clt}
\sqrt{T} \, \overline{{\Delta}}_T  \xrightarrow{\mathrm{d}} \mathcal{N}({0},{\Lambda}^\star),
\end{align}
where ${\Lambda}^\star$ is the associated (asymptotic) covariance matrix  defined as 
\begin{align}
\label{eq:defn-Lambdastar}
{\Lambda}^\star \coloneqq {A}^{-1} {\Gamma} {A}^{-\top},  %\\ 
% &{\Gamma} = \mathbb{E}[({A}_t {\theta}^\star - {b}_t)({A}_t {\theta}^\star - {b}_t)^\top]. 
\end{align}
with 
\begin{subequations}
\begin{align}
A  & \coloneqq  \mathsf{diag}(\mu) (I - \gamma P_\pi),  \\
\Gamma & \coloneqq \mathsf{diag}(\Gamma_{s,s}), \quad \mbox{with} \quad \Gamma_{s,s} = \mu(s) \mathop{\mathbb{E}}\limits_{s' \sim P_{\pi}(\cdot \mymid s)} \Big[\Big(V^{\pi}(s) - \big(r(s) + \gamma V^\pi(s')\big)\Big)^2\Big].
\end{align}
 \end{subequations}
Here, the definition of the matrix $ P_\pi$ can be found in~\eqref{eqn:ppivq}, $\mathsf{diag}(\mu)$ denotes a diagonal matrix with diagonal entries being $\{\mu(s)\}_{1\leq s\leq S}$, and $\Gamma$ is a diagonal matrix whose $s$-th diagonal entry is given by $\{\Gamma_{s,s}\}_{1\leq s\leq S}$. Intuitively, $\Gamma_{s,s}$ captures the variance of the TD error at each state, weighted by the state distribution $\mu$.  

It remains to strengthen this asymptotic guarantee by establishing a finite-sample, high-dimensional Gaussian approximation for $\overline{\Delta}_T$, which has attracted considerable recent attention (e.g., \citet{srikant2025rates,wu2024statistical,samsonov2024gaussian}). To this end, a natural tool is the Berry-Esseen bound, which provides an estimate of the distance between the distribution of an estimator and its Gaussian approximation. In high-dimensional settings, such bounds are particularly useful because they yield uniform guarantees over rich classes of sets (e.g., convex sets) \citep{raivc2019multivariate,kloeckner2019effective,shao2022berry,chen2010normal}, thereby enabling the construction of confidence regions with finite-sample validity. 
Throughout this subsection, we measure the discrepancy between two probability distributions $P$ and $Q$ using the convex distance
\begin{align}
\label{eqn:convex-distance}
	d_{\mathsf{cvx}}(P,Q) := \sup_{\mathcal{A} \in \mathscr{C}_d} |P(\mathcal{A}) - Q(\mathcal{A})|,
\end{align}
where $\mathscr{C}_d$ is the class of all convex sets in $\mathbb{R}^d$. Recalling the definition of the TV distance $\mathsf{TV}(P,Q)$ in \eqref{eq:TV}, we have $d_{\mathsf{cvx}}(P,Q) \leq \mathsf{TV}(P,Q).$
The following theorem provides a non-asymptotic Berry-Esseen bound (\cite{wu2024statistical}). 
\begin{theorem}\label{thm:Berry-Esseen-linear}
Consider the TD algorithm \eqref{eqn:td-learning-inference-2} with Polyak-Ruppert averaging~\eqref{eq:TD-averaging-inference}, independent samples \eqref{eq:iid-samples}, and decaying learning rates $\eta_t = \eta_0 t^{-2/3}$ for some $0<\eta_0< 1/4$. Then we have 
\begin{align*}
& d_{\mathsf{cvx}}\big(\sqrt{T}\,\overline{{\Delta}}_T,\mathcal{N}({0},{\Lambda}^\star)\big)\arxiv{ \\}
\arxiv{ & ~~~} \lesssim \Bigg(\sqrt{\frac{\eta_0 }{(1-\gamma) \lambda_0}} \mathsf{Tr}({\Gamma})\|{\Gamma}^{-1}\| + \frac{\sqrt{S \kappa({\Gamma})}}{(1-\gamma)\lambda_0\eta_0} \Bigg) T^{-\frac{1}{3}} + \widetilde{C} \; T^{-\frac{1}{2}}, 
\end{align*} 
% \fi
where $d_{\mathsf{cvx}}(\cdot,\cdot)$ is defined in \eqref{eqn:convex-distance}, $\kappa({\Gamma})$ is the condition number of $\Gamma$, $\lambda_0 = \min_{s \in S}\mu(s)$.  Here $\widetilde{C}>0$ is some problem-related quantity independent of $T$, and polynomial in salient problem parameters; see details in \cite{wu2024statistical}. 
\end{theorem}

This result establishes that $\sqrt{T}\,\overline{\Delta}_T$ is uniformly close in distribution, over the class of convex subsets of $\mathbb{R}^d$, to a standard Gaussian random vector with the asymptotic covariance \eqref{eq:defn-Lambdastar}. In this sense, it provides a finite-sample, high-dimensional analog of the classical central limit theorem and Berry-Esseen bound (e.g.,  \citet{polyak1992acceleration,fort2015central}). We note, however, that this guarantee becomes meaningful only when the sample size $T$ exceeds a polynomial function of the dimension, and the minimal sample size $T$ required to ensure such finite-sample validity remains an open question.

\subsection{Statistical inference}
\label{sec:inference}

The distributional theory in Theorem~\ref{thm:Berry-Esseen-linear} shows that the distribution of $\sqrt{T}\,\overline{\Delta}_T$ is well approximated by a Gaussian distribution with covariance matrix \(\Lambda^\star\) in finite samples, which naturally suggests a route to statistical inference through Gaussian confidence regions. The main obstacle is that the covariance matrix \(\Lambda^\star\) is unknown and must be estimated from data. We therefore begin by introducing an estimator of \(\Lambda^\star\), before showing how it can be used to construct finite-sample confidence sets.

\paragraph{Covariance matrix estimation.} 
Several approaches have been proposed for estimating covariance matrices in stochastic approximation and RL, such as \emph{plug-in estimation} \citep{chen2020statistical} and \emph{bootstrapping} \citep{white2010interval, hanna2017bootstrapping,hao2021bootstrapping,ramprasad2022online,zhu2023online,samsonov2024gaussian}. 
Here, we present a computationally efficient online plug-in estimator for \(\Lambda^\star\), as studied in  \cite{wu2024statistical}. 

Specifically, denoting by $e_{s}$ the $s$-th standard basis vector, we define
%
%\begin{subequations}
%\label{eq:defn-hat-Gamma}
\begin{align*}
\overline{{A}}_T & \coloneqq \frac{1}{T}\sum_{t=1}^{T} A_t \qquad \text{with } A_t \coloneqq e_{s_t}(e_{s_t} - \gamma e_{s'_t})^\top, 
\qquad \text{and} \\
\widehat{{\Gamma}}_T & \coloneqq \sum_{t=1}^T w_{t,T}({A}_t \overline{V}_T - b_t)({A}_t \overline{V}_T - b_t)^\top ~~\text{ with }  b_t \coloneqq r_t e_{s_t}
\end{align*}
%\end{subequations}
%
as estimators of ${A}$ and ${\Gamma}$, respectively. Here, the weights $\{w_{t,T}\}_{1 \leq t \leq T}$ in $\widehat{{\Gamma}}_T$ are chosen as 
\begin{align*}
w_{t,T} \coloneqq \frac{t}{\sum_{t=1}^T t} = \frac{2t}{T(T+1)};
\end{align*}
the rationale is that later iterations tend to be more accurate and therefore provide more reliable information about the covariance matrix.  We take
\begin{align}\label{eq:defn-hat-Lambda}
\widehat{{\Lambda}}_T = \overline{{A}}_T^{-1}\widehat{{\Gamma}}_T \big(\overline{{A}}_T^{-1}\big)^{\top} 
\end{align}
to be the final estimator of ${\Lambda}^\star$. 
Crucially, the estimator $\widehat{{\Lambda}}_T$ can be computed and updated in an online fashion.  

\paragraph{Confidence sets.} 
Equipped with the covariance estimator \(\widehat{\Lambda}_T\), we now describe a general procedure for constructing convex confidence sets for \(V^\pi\). Given a prescribed confidence level \(1-\alpha\in(0,1)\), it suffices to choose any convex set \(\mathcal C_{\alpha}\) that contains at least \(1-\alpha\) probability mass under the Gaussian distribution \(\mathcal N\big(\overline{V}_T,\widehat{\Lambda}_T/T\big)\); that is,
\begin{align}
\mathbb{P} \left(V \in \mathcal{C}_{\alpha} \mid \overline{V}_T,\widehat{{\Lambda}}_T \right) \geq 1-\alpha \label{eq:convex.Cdelta}
\end{align}
with $V \sim \mathcal{N}\big(\overline{V}_T,\widehat{{\Lambda}/T}\big)$. 
By the Gaussian approximation in Theorem~\ref{thm:Berry-Esseen-linear}, such sets serve as approximate confidence regions for \(V^\pi\). Natural choices of $\mathcal{C}_{\alpha}$ include, but not limited to, the following.   
\begin{itemize}
\item[(i)] {\em Hyper-rectangles}, yielding simultaneous confidence intervals of the form
\begin{align*}
%\label{eq:multi-CI}
\mathcal{C}_{\alpha} = \prod_{s=1}^{S} \big[\overline{V}_T(s) - \widehat{R}/\sqrt{T}, \overline{V}_T(s) + \widehat{R}/\sqrt{T} \big],
\end{align*}
where $\widehat{R}>0$ is the (random) quantile of the $\ell_{\infty}$ norm of a centered Gaussian random vectors with covariance $\widehat{{\Lambda}}_T/{T}$. This quantile can be calculated efficiently via simulations.  

\item[(ii)] \emph{Ellipsoids} or \emph{balls}, for example
\begin{align*}%\label{eq:ellipsoid-region}
\mathcal{C}_{\alpha} = \left\{V: T\, \big(V - \overline{V}_T\big)^{\top} \widehat{{\Lambda}}_T^{-1} \big(V - \overline{V}_T\big) \leq R^2\right\},
\end{align*}
where $R$ is the $1-\alpha$ quantile of the $\chi^2(d)$ distribution.

\end{itemize}

%% file: chapters/RLHF.tex
\section{RL with human feedback}
 \label{sec:rlhf}
 
Reinforcement learning from human feedback (RLHF) \citep{ziegler2019fine}, a paradigm to fine-tune large language models (LLMs), has been shown to be instrumental in significantly improving the helpfulness, truthfulness and controllability of LLMs  \citep{ouyang2022training}. Roughly speaking, there are two critical components of RLHF: (1) {\em reward modeling}, which maps human preference rankings into a quantitative reward function that can guide policy improvement; and (2) {\em RL fine-tuning}, which seeks to adjust LLM output to align with human preferences by leveraging the learned reward function, i.e., increasing the probability of preferred answers and decreasing the probability of unfavored answers.  

Evidently, the curation of preference data is instrumental in the performance of RLHF, which is commonly modeled as pairwise comparisons from a Bradley-Terry ranking model \citep{bradley1952rank}.
In particular, given a query $x$, human annotators choose a preferred answer from two candidate answers $y_1$ and $y_2$ generated by an LLM. Despite the simple form, collecting large-scale and high-quality preference data can be expensive and time-consuming. Depending on the availability of preference data, two paradigms of RLHF are considered: (1) \emph{offline} RLHF, where only a pre-collected preference dataset is available, possibly generated from a pre-trained LLM after supervised fine-tuning (SFT); and (2) \emph{online} RLHF, where additional preference data can be collected adaptively to improve alignment. 

While there has been significant work on the theoretical underpinnings of RLHF that seeks to uncover algorithmic improvements, significant challenges remain, particularly concerning how to incorporate estimates of reward {\em uncertainty} without explicitly estimating the reward, given that it is often desirable to only maintain the policy itself.

\subsection{Background}

In RLHF, a language model is described by a policy $\pi$, which generates an answer $y \in \mathcal{Y}$ given a prompt $x\in \mathcal{X}$ according to the conditional probability distribution $\pi(\cdot \mymid  x)$. The standard RLHF process consists of four stages: supervised fine-tuning (SFT), preference data generation, reward modeling, and RL fine-tuning. In the SFT stage, a language model $\pi_{\textsf{sft}}$ is obtained by fine-tuning a pre-trained LLM with supervised learning. The remaining stages continue training by leveraging the preference data, which we elaborate below.
 
\paragraph{Reward modeling from preference data.} An oracle (e.g., a human labeler or a scoring model) evaluates the quality of two answers $y_1$ and $y_2$ given prompt $x$ and reveals its preference. A widely used approach for modeling the probability of pairwise preferences is the Bradley–Terry model \citep{bradley1952rank}:
\begin{align} \label{eq:BT}
    \mathbb{P}(y_1\succ y_2|x) &= \frac{\exp\bigl(r^\star(x, y_1)\bigr)}{\exp\bigl(r^\star(x, y_1)\bigr)+\exp\bigl(r^\star(x, y_2)\bigr)} \arxiv{\nonumber \\
    &} = \sigma\bigl(r^\star(x, y_1) - r^\star(x, y_2)\bigr),
\end{align}
where $y_1\succ y_2$ indicates that $y_1$ is preferred over $y_2$, $r^\star: \mathcal{X}\times\mathcal{Y}\to \mathbb{R}$ is the ground truth reward function, and $\sigma : \mathbb{R} \to (0, 1)$ is the logistic function. A preference data sample is denoted by a tuple $(x, y_+, y_-)$, where $y_+$ (resp.  $y_{-}$) is the preferred (resp. unpreferred) answer in the comparison.

Given a preference dataset $\mathcal{D} = \bigl\{(x^i, y_+^i, y_-^i)\bigr\}$ composed of independent samples, one natural idea to estimate the reward function $r$ is via maximum likelihood estimation (MLE):
\begin{equation}
    r_{\mathsf{MLE}} = \arg\min_{r} \; \ell(r, \mathcal{D}),
    \label{eq:MLE}
\end{equation}
where $\ell(r, \mathcal{D})$ is the negative log-likelihood of $\mathcal{D}$, given as
\begin{equation}
     \ell(r, \mathcal{D}) \coloneqq -\sum_{(x^i, y_+^i, y_-^i)\in\mathcal{D}}{\log \sigma\bigl(r(x^i,y_+^i) - r(x^i,y_-^i)\bigr)}.
\end{equation}

\paragraph{RL fine-tuning.} With a reward model $r$ in hand, we seek to fine-tune the policy $\pi$ to achieve an ideal balance between the expected reward and its distance from an initial reference policy $\pi_{\mathsf{ref}}$, which is typically set the same as $\pi_{\textsf{sft}}$. This is  achieved by maximizing the KL-regularized value function $J(r, \pi)$, defined as
\begin{equation} \label{eq:KL_reward}
    J(r, \pi) \coloneqq \mathop\mathbb{E}\limits_{x \sim \rho, y \sim \pi(\cdot\mymid x)}\bigl[r(x, y) \bigr] - \beta\mathop\mathbb{E}\limits_{x \sim \rho}\Bigl[ \KL{\pi(\cdot\mymid x)}{\pi_{\mathsf{ref}}(\cdot\mymid x)}\Bigr],
\end{equation}
where $\KL{\pi_1}{\pi_2} $ is the KL divergence from $\pi_1$ to $\pi_2$, and $\beta>0$ is a regularization parameter. Consequently, the RL fine-tuned policy $\pi_r$ with respect to the reward $r$ satisfies  
\begin{equation}
    \pi_r \coloneqq \arg\max_{\pi} J(r, \pi),
\end{equation}
which admits a closed-form solution \citep{rafailov2023direct}, i.e.,
\begin{equation}
\forall (x, y) \in \mathcal{X}\times\mathcal{Y}: \qquad    \pi_r(y|x) = \frac{\pi_{\mathsf{ref}}(y\mymid x)\exp\bigl(r(x,y)/\beta\bigr)}{Z(r, x)}.
    \label{eq:RLHF-policy-closed-form}
\end{equation}
Here, $ Z(r, x)$ is a normalization factor given by
\begin{equation} \label{eq:normalization}
    Z(r, x) = \sum_{y' \in \mathcal{Y}}\pi_{\mathsf{ref}}(y'\mymid x)\exp\bigl(r(x,y')/\beta\bigr).
\end{equation}

\paragraph{Direct preference optimization.} 
The closed-form solution \eqref{eq:RLHF-policy-closed-form} allows us to write the reward function $r$ in turn as
\begin{equation}\label{eq:reward-policy-eq}
    r(x, y) = \beta \big(\log \pi_r(y\mymid x) - \log \pi_{\mathsf{ref}}(y\mymid x) + \log Z(r, x)\big).
\end{equation}
Plugging the above equation into the reward MLE~\eqref{eq:MLE}, we obtain the seminal formulation of direct preference optimization (\DPO) over the policy space \citep{rafailov2023direct}, 
\begin{align} \label{eq:DPO}
 \arxiv{ &}  \pi_{\mathsf{DPO}}  \arxiv{ \nonumber \\}
  &= \arg\min_{\pi} \; -\sum_{(x^i, y_+^i, y_-^i)\in\mathcal{D}}{\log \sigma\Biggl( \beta \left(\log \frac{\pi(y_+^i\mymid x^i)}{\pi_{\mathsf{ref}}(y_+^i\mymid x^i)}  - \log \frac{\pi(y_-^i\mymid x^i)}{\pi_{\mathsf{ref}}(y_-^i\mymid x^i)}  \right) \Biggr)} .
\end{align}
An important advantage of this formulation is that it completely avoids the explicit learning of the reward model, where the normalization factor  $Z(r, x)$ cancels out in \eqref{eq:DPO}, as it is formidable to compute when the number of responses $|\mathcal{Y}|$ is large.

\subsection{Value-incentivized preference optimization}
\label{sec:vpo}

A major caveat of the standard RLHF framework concerns the lack of accounting for reward uncertainty, which is known to be indispensable in the success of standard RL paradigms in both online and offline settings as seen repetitively in this monograph. This motivates the design of a principled mechanism that can be easily integrated into the RLHF pipeline, while bypassing the difficulties of explicit uncertainty estimation in LLMs.

In view of the suboptimality of naive MLE for reward estimation \citep{cesa2017boltzmann,rashidinejad2021bridging}, it is tempting to regularize the reward estimate towards those with higher returns \citep{kumar1982new,liu2019exploration} 
\begin{equation} 
% \label{eq:Jstar}
J^\star(r) = \max_{\pi} \, J(r,\pi),
\end{equation}
which measures the resulting value function for the given reward if one acts according to its optimal policy. However, in RLHF,  by the definition \eqref{eq:BT}, the reward function $r^\star$ is only identifiable up to a prompt-dependent global shift. Specifically, suppose that $r_1(x, y)$ and $r_2(x, y)$ are two reward functions that only differ by a prompt-dependent shift $c(x)$, i.e., $r_1(x, y) = r_2(x, y) + c(x)$. It follows that $r_1(x, y_1) - r_1(x, y_2) = r_2(x, y_1) - r_2(x, y_2)$, which means that the two reward functions are not distinguishable in terms of the log-likelihood function since $\ell(r_1, \cD) = \ell(r_2, \cD) $. However, their respective returns are shifted by $J^{\star}(r_1) = J^{\star}(r_2) +\mathbb{E}_{x\sim \rho}\bigl[c(x)\bigr]$, making $J^{\star}(r)$ improper to be used directly as a regularizer. To resolve this challenge, we introduce a calibrated value function with respect to a calibration policy $\pi_{\mathsf{cal}}$, given as 
\begin{align}
    J(r, \pi;\pi_{\mathsf{cal}}) &= \mathop\mathbb{E}\limits_{x \sim \rho, y \sim \pi(\cdot\mymid x)}\bigl[r(x, y)\bigr] - \mathop\mathbb{E}\limits_{x \sim \rho, y \sim \pi_{\mathsf{cal}}(\cdot\mymid x)}\bigl[r(x, y)\bigr]  \arxiv{\nonumber \\}
   \arxiv{ & \quad\quad \quad } - \beta\mathop\mathbb{E}\limits_{x \sim \rho}\Bigl[ \KL{\pi(\cdot\mymid x)}{\pi_{\mathsf{ref}}(\cdot\mymid x)}\Bigr].
    \label{eq:calibrated_value}
\end{align}
This calibrated value function is invariant to prompt-dependent shifts of the reward: if $r_1(x, y)=r_2(x, y)+c(x)$, then $J(r_1,\pi;\pi_{\mathsf{cal}})=J(r_2,\pi;\pi_{\mathsf{cal}})$ for any policy $\pi$. We further define the optimal calibrated value
\begin{equation}
J^\star(r;\pi_{\mathsf{cal}}) = \max_{\pi} \, J(r,\pi;\pi_{\mathsf{cal}}),
\label{eq:Jstar_calibrated}
\end{equation}
which measures the best KL-regularized improvement over the calibration policy under the reward model $r$. %Whenever it is clear from the context, we drop the dependency with $\pi_{\mathsf{cal}}$, and simply write $J^\star(r;\pi_{\mathsf{cal}})$ and $J^\star(r)$.

\citet{cen2025vpo}  proposes a regularized MLE of the Bradley-Terry model \eqref{eq:MLE} called value-incentivized preference optimization (\VPO), which  
appends the regularization term $J^\star(r;\pi_{\mathsf{cal}})$ to the negative log-likelihood  
\begin{equation}
    r_{\mathsf{VPO}} = \arg\min_{r } \; \bigl\{\ell(r, \mathcal{D}) - \mathsf{sign} \cdot \alpha \cdot J^\star(r;\pi_{\mathsf{cal}})\bigr\},
        \label{eq:RBMLE}
\end{equation}
incentivizing the algorithm to favor (resp. avoid) reward models with higher calibrated value  $J^\star(r;\pi_{\mathsf{cal}})$ in the online (resp. offline) setting. Here, $\alpha > 0$ is a constant controlling the strength of regularization, and $\mathsf{sign}$ is set to $1$ in the online setting and $-1$ in the offline setting, corresponding to the optimism and pessimism for online and offline RL, explained in earlier chapters.

At first glance, the objective function for \VPO \eqref{eq:RBMLE} does not immediately imply a computationally-efficient algorithm due to the presence of $J^\star(r;\pi_{\mathsf{cal}})$. However, by exploiting the same closed-form solution for the optimal policy given the reward in~\eqref{eq:RLHF-policy-closed-form}, and the reward representation inferred from the policy via \eqref{eq:reward-policy-eq}, 
we can explicitly express $J^\star(r;\pi_{\mathsf{cal}})$ as 
\begin{align}
J^\star(r;\pi_{\mathsf{cal}}) &= \mathop\mathbb{E}\limits_{x \sim \rho, y \sim \pi_r(\cdot|x)}\Bigl[r(x, y) - \beta \bigl(\log \pi_r(y|x) - \log\pi_{\mathsf{ref}}(y|x)\bigr)\Bigr]  \arxiv{ \nonumber \\}
   \arxiv{ & \qquad\qquad\qquad } - \mathop\mathbb{E}\limits_{x \sim \rho, y \sim \dnorm(\cdot|x)}\bigl[r(x, y)\bigr]\notag\\
    &= \beta\mathop\mathbb{E}\limits_{x \sim \rho}\bigl[\log Z(r, x)\bigr] - \mathop\mathbb{E}\limits_{x \sim \rho, y \sim \dnorm(\cdot|x)}\bigl[r(x, y)\bigr]\notag\\
    &\overset{\mathrm{(i)}}{=} \mathop\mathbb{E}\limits_{x \sim \rho, y \sim \dnorm(\cdot|x)}\bigl[\beta\log Z(r, x) - r(x, y)\bigr]\notag\\
    &\overset{\mathrm{(ii)}}{=}  -\beta \mathop\mathbb{E}\limits_{x \sim \rho, y \sim \dnorm(\cdot|x)}\bigl[ \log \pi_r(y|x) - \log\pi_{\mathsf{ref}}(y|x)\bigr], \label{eq:Jstar}
\end{align}
where (i) follows because the bracketed term is independent of $y$ (c.f. \eqref{eq:RLHF-policy-closed-form}) and (ii) follows by rearranging \eqref{eq:RLHF-policy-closed-form}.
Given this key ingredient, we can then rewrite \eqref{eq:RBMLE} to directly optimize the policy, in a flavor similar to \DPO, as
\begin{align}
\arxiv{ &}    \pi_{\mathsf{VPO}}  \arxiv{ \nonumber \\}
 &= \argmin_{\pi_r } \, \bigl\{\ell(r, \mathcal{D}) -  \mathsf{sign} \cdot \alpha \cdot J^\star(r;\pi_{\mathsf{cal}})  \bigr\}\notag\\
   &= \argmin_{\pi_r } \Biggl\{-\sum_{(x^i, y_+^i, y_-^i)\in\mathcal{D}} {\log \sigma\Biggl(\beta\log\frac{\pi_r(y_+^{i} |x^{i})}{\pi_{\mathsf{ref}}(y_+^{i}|x^{i})} - \beta\log\frac{\pi_r(y_-^{i}|x^{i})}{\pi_{\mathsf{ref}}(y_-^{i}|x^{i})}\Biggr)}   \notag\\
  &\qquad\qquad\qquad  + \mathsf{sign} \cdot\alpha \beta  \mathop\mathbb{E}\limits_{x \sim \rho, y \sim \dnorm(\cdot|x)}\bigl[\log \pi_r(y|x) - \log\pi_{\mathsf{ref}}(y|x)\bigr]\Biggr\} \nonumber \\
   &= \arg\min_{\pi} \Biggl\{-\sum_{(x^i, y_+^i, y_-^i)\in\mathcal{D}} {\log \sigma\Biggl(\beta\log\frac{\pi(y_+^{i} |x^{i})}{\pi_{\mathsf{ref}}(y_+^{i}|x^{i})} - \beta\log\frac{\pi(y_-^{i}|x^{i})}{\pi_{\mathsf{ref}}(y_-^{i}|x^{i})}\Biggr)} \notag\\
    &\qquad\qquad\qquad + \mathsf{sign} \cdot\alpha \beta  \mathop\mathbb{E}\limits_{x \sim \rho, y \sim \dnorm(\cdot|x)}\bigl[\log \pi (y|x) - \log\pi_{\mathsf{ref}}(y|x)\bigr]\Biggr\}.
    \label{eq:policy_RBMLE}
\end{align}

%\begin{remark}
Observing that in the last term $\mathop\mathbb{E}\limits_{x \sim \rho, y \sim \dnorm(\cdot|x)}\bigl[\log \pi (y|x) - \log\pi_{\mathsf{ref}}(y|x)\bigr] $ of \eqref{eq:policy_RBMLE}, the the reference policy $\pi_{\mathsf{ref}}(y|x)$  does not impact the optimization solution, we can change it to 
\[
\mathop\mathbb{E}\limits_{x \sim \rho, y \sim \dnorm(\cdot|x)}\bigl[\log \pi (y|x) - \log\dnorm(y|x)\bigr] = - \mathop\mathbb{E}\limits_{x \sim \rho}\Bigl[ \KL{\dnorm(\cdot|x)}{\pi(\cdot|x)}\Bigr] ,
\]
which amounts to adding a  KL regularization to the original \DPO formulation and offers an interesting interpretation as pushing $\pi$ against/towards $\dnorm$ in the online/offline settings, respectively, unveiling the role of reward calibration in RLHF. 
%\end{remark}

In what follows, we elaborate on the development of \VPO in the online setting and its possible improvement. For simplicity, we restrict attention to the linear approximation of the reward model.
\begin{assumption}[Linear Reward]
\label{assump:linear}
We parameterize the reward model by 
\begin{equation}
    r_\theta(x, y) = \innprod{\phi(x, y), \theta},\quad \forall (x, y)\in\mathcal{X}\times \mathcal{Y},
    \label{eq:linear_FA}
\end{equation} where $\phi: \mathcal{X}\times\mathcal{Y} \to \mathbb{R}^d$ is a fixed feature mapping and $\theta \in \mathbb{R}^d$ is the parameter vector. We assume that $\bigl\|\phi(x, y)\bigr\|_2\le 1$ for all $(x, y) \in \mathcal{X}\times\mathcal{Y}$, and that $r^\star(x, y)= \innprod{\phi(x, y), \theta^\star}$ for some $\theta^\star \in \mathbb{R}^d$.
\end{assumption}

\subsection{Theory and improvement for online RLHF}
The online RLHF procedure extends training by performing reward learning and policy learning iteratively, with a growing preference dataset collected by using the current policy. We use $\pi^{(t)}$ to denote the policy used in the $t$-th iteration, where the superscript $^{(t)}$ indicates iteration $t$ in the online setting.
 The $t$-th iteration of \VPO for online RLHF  consists of the following steps:
\begin{enumerate}%[leftmargin=*,nosep]
    \item \textbf{New preference data generation.} We sample a new prompt $x^{(t)}\sim\rho$ and two answers $y_1^{(t)}, y_2^{(t)}\sim\pi^{(t)}\bigl(\cdot|x^{(t)}\bigr)$, query the preference oracle and append $\bigl(x^{(t)}, y_+^{(t)}, y_-^{(t)}\bigr)$ to the preference dataset.
    \item \textbf{Reward learning.} We train a reward model with preference data $\mathcal{D}^{(t)} \coloneqq \Bigl\{\bigl(x^{(s)}, y_+^{(s)}, y_-^{(s)}\bigr)\Bigr\}_{s=1}^{t}$ by minimizing the regularized negative log-likelihood, i.e., 
    \begin{equation}
        r^{(t+1)} = 
    %    \begin{cases}
     %   \arg\min_{r\in\mathcal{R}} \ell(r, \mathcal{D}^{(t)}) & \texttt{MLE}\\
        \arg\min_{r } \; \Bigl\{\ell\bigl(r, \mathcal{D}^{(t)}\bigr) - \alpha \cdot J^\star(r;\pi_{\mathsf{cal}}) \Bigr\}. %& \texttt{Reward-biased MLE}
      %  \end{cases}
        \label{eq:RBMLE_online}
    \end{equation}
    \item \textbf{Policy learning.} This step trains the policy by solving the RL fine-tuning problem:
    \begin{equation}
        \pi^{(t+1)} = \arg\max_{\pi} J\bigl(r^{(t+1)}, \pi\bigr).
        \label{eq:RLHF-policy}
    \end{equation}
\end{enumerate}
\noindent In view of the development in Section~\ref{sec:vpo}, the reward learning and policy learning steps can be combined and instead, it is adequate to execute the policy update rule specified in \eqref{eq:policy_RBMLE} using the preference data  $\mathcal{D}^{(t)}$ availabel at the $t$-th iteration.
We summarize the detailed procedure in Algorithm \ref{alg:online_RBMLE}.

%\begin{figure}
\begin{algorithm}[t]
    \DontPrintSemicolon
       \textbf{initialization:}  $\pi^{(0)}$.  \\
       % \textbf{parameters:} learning rate $\eta_t$.
        % number of iterations $T$, 
      %\textbf{initialization:} $Q_0=0$. \\
       \For{$t=0,1,2,\cdots$}
        {
            Sample  $x^{(t)}\sim \rho$, $y_1^{(t)}, y_2^{(t)} \sim \pi^{(t)}\bigl(\cdot|x^{(t)}\bigr)$. \\
            Obtain the preference between $\bigl(x^{(t)}, y_1^{(t)}\bigr)$ and $\bigl(x^{(t)}, y_2^{(t)}\bigr)$ from some oracle. Denote the comparison outcome by $\bigl(x^{(t)}, y_+^{(t)}, y_-^{(t)}\bigr)$.\\
            Update policy $\pi$ as
            \small
            \begin{align}   \label{eq:online_RBMLE}
            \arxiv{ &}   \pi^{(t+1)} \arxiv{ \nonumber \\}
                &= \arg\min_{\pi  } \Biggl\{-\sum_{s=1}^{t}{\log \sigma\Biggl(\beta\log\frac{\pi(y_+^{(s)}|x^{(s)})}{\pi_{\mathsf{ref}}(y_+^{(s)}|x^{(s)})} - \beta\log\frac{\pi(y_-^{(s)}|x^{(s)})}{\pi_{\mathsf{ref}}(y_-^{(s)}|x^{(s)})}\Biggr)} \nonumber \\
                &\qquad \qquad\qquad + \alpha \beta  \mathop\mathbb{E}\limits_{x \sim \rho, y \sim \dnorm(\cdot|x)}\bigl[\log \pi(y|x) - \log\pi_{\mathsf{ref}}(y|x)\bigr]\Biggr\}.
            \end{align}
            \normalsize
        }
        \caption{\VPO for online RLHF}
        \label{alg:online_RBMLE}
    \end{algorithm}
%\end{center}

\paragraph{Theoretical guarantees.} Encouragingly, \VPO admits appealing theoretical guarantees under linear function approximation. The next theorem demonstrates that Algorithm \ref{alg:online_RBMLE} achieves $\widetilde{O}\bigl(\sqrt{T}\bigr)$ cumulative regret under mild assumptions. 
\begin{theorem}[\citet{cen2025vpo}]\label{thm_online}
Under Assumption~\ref{assump:linear}, let $r_{\theta^{(t)}}$ denote the corresponding reward model for $\pi^{(t)}$.    Assume that $\|\theta^\star\|_2 \le r_{\mathsf{max}}$ and $\bigl\|\theta^{(t)}\bigr\|_2 \le r_{\mathsf{max}}, \forall t \ge 0$ for some $r_{\mathsf{max}} > 0$.
     Then, with probability $1-\delta$, we have 
    \begin{align*}
        \mathsf{Regret} &\coloneqq \sum_{t=1}^{T} \Bigl[J^\star(r^\star;\pi_{\mathsf{cal}} ) - J\bigl(r^\star, \pi^{(t)};\pi_{\mathsf{cal}}\bigr) \Bigr] \le \widetilde{O}\Bigl(\exp\bigl(2r_{\mathsf{max}}+ r_{\mathsf{max}}/\beta\bigr)\sqrt{\kappa d T} \Bigr),
    \end{align*}
    with $\alpha = \dfrac{1}{\exp\bigl(2r_{\mathsf{max}}+r_{\mathsf{max}}/\beta\bigr)}\sqrt{\dfrac{T}{\kappa\min\{d\log T, T\}}}$ and $\kappa = \sup_{x,y}\dfrac{\dnorm(y|x)}{\pi_{\mathsf{ref}}(y|x)}$.
\end{theorem}
Theorem~\ref{thm_online} shows that \VPO --- without explicit uncertainty quantification --- achieves the same $\widetilde{O}\bigl(\sqrt{T}\bigr)$ regret for online RLHF as its counterparts in standard contextual bandits with scalar rewards and using UCB for exploration \citep{lattimore2020bandit}.

\paragraph{Adaptive calibration policy.}
The online procedure above fixes the calibration distribution $\dnorm$ throughout training. This fixed choice is useful for eliminating the shift ambiguity of the reward function inherent in the Bradley-Terry model, but it also determines which reward differences are emphasized by the optimistic reward estimate and can hinder the exploration. \citet{li2025towards} identifies two opportunities to improve upon \VPO:
\begin{itemize}
\item Instead of using a fixed $\dnorm$, one can adaptively update the calibration policy, in particular, using the current policy $\pi^{(t)}$;
\item Instead of drawing both responses following $\pi^{(t)}$, one can sample the responses following  $\pi^{(t-1)}$ and $\pi^{(t)}$, respectively.
\end{itemize}
%The calibrated value formulation in \eqref{eq:calibrated_value} separates the calibration policy from the policy being optimized, and therefore allows the calibration policy to vary over time. For any fixed $\pi_{\mathsf{cal}}$, the maximizer over $\pi$ still admits the closed-form expression in \eqref{eq:RLHF-policy-closed-form}, since the calibration term in \eqref{eq:calibrated_value} is independent of the policy being optimized. Thus, the calibration policy affects the optimistic bias used for reward learning, but does not change the RL fine-tuning solution associated with a fixed reward model.
%
%The choice of $\pi_{\mathsf{cal}}$ is nevertheless important for online exploration. If the algorithm draws both candidates from the current policy $\pi^{(t)}$, then the new comparison mainly reduces uncertainty between two samples favored by $\pi^{(t)}$, while the value incentive may be emphasizing uncertainty relative to the calibration policy. On the other hand, sampling one candidate from $\pi^{(t)}$ and the other from a fixed calibration policy can also be inefficient when the fixed calibration policy assigns most of its mass to low-reward answers: the preference oracle then almost always favors the candidate from $\pi^{(t)}$, yielding little information about the reward differences that matter for policy improvement.
This leads to a refinement of \VPO which in the $t$-th iteration, adaptively uses the current policy $\pi^{(t)}$ as the calibration policy:
\begin{align}
 \arxiv {&}   \pi^{(t+1)} \arxiv{ \nonumber \\}
    &= \argmin_{\pi  } \Biggl\{-\sum_{s=1}^{t}{\log \sigma\Biggl(\beta\log\frac{\pi(y_+^{(s)}|x^{(s)})}{\pi_{\mathsf{ref}}(y_+^{(s)}|x^{(s)})} - \beta\log\frac{\pi(y_-^{(s)}|x^{(s)})}{\pi_{\mathsf{ref}}(y_-^{(s)}|x^{(s)})}\Biggr)} \notag \\
    &\qquad \qquad\qquad + \alpha_t \beta  \mathop\mathbb{E}\limits_{x \sim \rho, y \sim \pi^{(t)}(\cdot|x)}\bigl[\log \pi(y|x) - \log\pi_{\mathsf{ref}}(y|x)\bigr]\Biggr\}.
    \label{eq:adaptive_policy_update}
\end{align}
In the case of multi-armed bandits, this yields a regret bound of order $T^{(\beta+1)/(\beta+2)}$ up to polynomial dependency with other problem parameters. When $\beta=0$, the rate reduces to the usual $\widetilde{O}\bigl(\sqrt{T}\bigr)$ form. In general, the slower exponent in $T$ reflects the trade-off between reward maximization and the KL regularization strength, while avoiding the exponential dependence on quantities such as $r_{\max}/\beta$ that can appear under fixed-calibration sampling protocols such as \VPO (cf.~Theorem~\ref{thm_online}).

%% file: chapters/conclusion.tex
\chapter{Concluding remarks}

 In this monograph, we have developed a non-asymptotic statistical perspective on reinforcement learning, focusing primarily on tabular MDPs under three canonical sampling mechanisms: access to a simulator, online exploration, and offline data. By emphasizing sharp dependencies of the statistical performance on salient problem parameters --- especially the dimension of the state-action space and the planning horizon --- our treatment highlights the critical role of the data-collection mechanism in shaping both the fundamental statistical limits and the algorithmic principles needed to attain them. 

Several recurring themes have emerged from the non-asymptotic viewpoints developed throughout this monograph, and we highlight a few of them below. First, obtaining sharp finite-sample guarantees requires exploiting the intrinsic Bellman structure of MDPs, rather than relying solely on generic concentration-of-measure arguments. Techniques such as variance-aware analyses, leave-one-out decoupling arguments, and carefully calibrated uncertainty quantification play a pivotal role in overcoming substantial sample-size barriers and enabling statistically optimal RL algorithms. Second, optimism and pessimism arise as complementary guiding principles: optimism drives efficient exploration in online RL, whereas pessimism guards against distribution shift and insufficient coverage in offline RL. In settings beyond these canonical sampling paradigms, the two seemingly opposing principles may even need to be carefully integrated in order to achieve statistical efficiency. Third, the comparison between model-based and model-free approaches reveals a persistent gap in statistical performance. Model-based methods, often coupled with carefully designed uncertainty quantification, remain the only class of approaches known to provably attain minimax-optimal sample complexity across the entire accuracy regime --- including the most sample-starved settings --- under the three canonical sampling mechanisms studied in this monograph. By contrast, classical model-free methods such as Q-learning may suffer from inherent statistical inefficiency and frequently require sophisticated algorithmic modifications even to achieve asymptotically optimal statistical guarantees.

Taken collectively, the developments covered in this monograph illustrate both the remarkable progress and the substantial challenges that continue to shape modern statistical RL. Ultimately, we hope that this monograph equips new researchers with some of the core conceptual and technical toolkits of the field, while motivating further theoretical and algorithmic investigations. 

Naturally, the scope of this monograph has been selective, and many important directions in modern RL fall outside its coverage. Below, we briefly highlight a few representative examples. 
\begin{itemize}

\item {\em Beyond discounted and episodic settings.} This monograph has focused primarily on discounted infinite-horizon MDPs and finite-horizon episodic MDPs. 
A few other formulations have also received considerable attention in the RL literature. One classic and practically important example is {\em average-reward RL}, where performance is measured by the long-run average reward rather than by a discounted or finite-horizon return \citep{puterman2014markov,bartlett2009regal,wan2021learning}. Average-reward formulations arise naturally in continuing tasks and long-running control systems and often require techniques that differ substantially from those used in discounted or episodic settings. Significant recent progress has been made in designing efficient algorithms and establishing non-asymptotic performance guarantees in the context of average-reward MDPs \citep{jin2021towards,liao2022batch,zhang2021finite,zurek2024spanbased,jin2024feasible,chen2026non,jiao2026sample}, although many fundamental statistical questions remain open.

%Although significant progress has been made in recent years for both model-based \citep{jin2021towards,zurek2024spanbased} and model-free methods \citep{jin2024feasible,jiao2026sample}, obtaining sharp finite-sample guarantees under different data collection mechanisms remains an active area of research.

\item {\em Policy optimization and actor-critic methods.}
Policy optimization constitutes another major paradigm in modern RL, which seeks to improve parameterized policies directly through optimization-based procedures  \citep{williams1992simple,sutton2000policy,kakade2002natural,peters2008natural}. Owing to their flexibility in accommodating rich policy parameterizations, policy-gradient methods and their variants (e.g., natural policy gradient methods, trust-region methods, and proximal policy optimization) have played a central role in the successes of deep RL \citep{sutton2000policy,mnih2015human,schulman2015trust,silver2016mastering,schulman2017proximal} and have attracted substantial theoretical attention (e.g., \citet{agarwal2019optimality,mei2020global,cen2022fast,li2023softmax,lan2023policy,zhan2023policy,bhandari2024global}). Actor-critic methods \citep{konda2000actor} further combine policy optimization with value estimation, using a critic to evaluate the current policy and guide its improvement. Despite their empirical effectiveness and statistical studies  \citep{xu2020improving,cai2020provably,yuan2023linear,chen2023finite,tan2025actor}, the statistical foundations of policy optimization and actor--critic methods remain considerably less developed than those of the model-based and value-based approaches discussed throughout this monograph.

\item {\em RL in continuous time and continuous spaces.}
While this monograph has focused on finite-state, finite-action MDPs, many important applications involve continuous state-action spaces and/or continuous-time dynamics  \citep{recht2019tour,van2012reinforcement,wang2020reinforcement,hazan2025research,guo2026introduction}. Such settings are intimately connected to stochastic control and introduce a range of statistical, computational, and modeling challenges. From a statistical perspective, finite-sample guarantees for both sample complexity and regret have been established for a few structured models, most notably linear-quadratic regulators (e.g., \citet{abbasi2011regret,dean2020sample,simchowitz2020naive,zheng2021sample,zhang2021regret,wang2021exact}). Note that rigorous learning guarantees in continuous domains often rely on additional structural assumptions, such as linearity, kernel structure, or other forms of tractable function approximation.

%Extending non-asymptotic statistical guarantees to broader classes of continuous-time and continuous-space problems remains a major research direction. 

\end{itemize}

Beyond the directions highlighted above, a number of other active research areas continue to shape the development of modern RL. We conclude by singling out several directions that have recently attracted significant attention and are likely to play an important role in the future development of the field.

\begin{itemize}

\item {\em Hybrid RL.}
While this monograph has addressed generative models, offline RL, and online RL as separate sampling mechanisms, 
it has omitted important settings in which an agent has simultaneous access to multiple data collection mechanisms. For example, an agent may be allowed to perform a limited amount of online exploration while also leveraging a pre-collected offline dataset, or it may be able to collect samples in real time while making extensive use of a simulator. Such hybrid RL settings have garnered significant recent attention from both theoretical and practical perspectives (e.g., \cite{xie2022role,song2023hybrid,li2024reward,wagenmaker2023leveraging,tan2024hybrid,panaganti2024model,qu2025hybrid}), offering insights that extend far beyond the studies of online RL, offline RL, and generative models in isolation.

%Perhaps most appealing is their flexibility in adopting various kinds of policy parameterizations (e.g.~a class of policies parameterized via deep neural networks),  which makes them remarkably powerful and versatile in contemporary RL. 

%Also, what has been left uncovered but is also worth mentioning is the actor-critic algorithms. Actor-critic algorithms have emerged as a foundational approach in RL, mitigating the deficiencies of both policy-based and value-based approaches \citep{sutton2018reinforcement, mnih2016asynchronousmethodsdeepreinforcement,haarnoja2018softactorcriticoffpolicymaximum，tan2025actor}. These methods combine two components: an actor, which directly learns and improves the policy, and a critic, which evaluates the policy's quality. Given their popularity, significant recent progress has been made in understanding their theoretical underpinnings and statistical efficiency \citep{cai2024provablyefficientexplorationpolicy, zhong2023theoreticalanalysisoptimisticproximal, sherman2024rateoptimalpolicyoptimizationlinear, liu2023optimisticnaturalpolicygradient}. 

\item {\em Towards more general function approximation.} 
The tabular theory developed in this monograph should be viewed as a starting point rather than an endpoint for understanding the statistical principles underlying RL. A central objective of modern RL theory is to understand learning and decision-making with rich function approximation (e.g., \citet{foster2021statistical,fan2020theoretical,chen2025unified,yang2025incentivize}). Existing theoretical guarantees beyond the tabular setting, however, typically rely on highly structured function classes, such as linear models, kernel methods, low-rank representations, or other settings that admit tractable statistical and computational analyses. Although these frameworks have yielded important conceptual and technical insights, they remain far from capturing the highly expressive function approximators used in modern practice, most notably deep neural networks. Bridging this gap will require fundamentally new ideas that go well beyond current theories of linear and low-complexity function approximation.

%\item {\em RL meets transfer learning.} 

\item {\em Partially observable MDPs (POMDPs).}
The MDP framework considered throughout this monograph assumes full observability of the underlying state. In many applications, however, agents must make decisions based only on noisy or partial observations, leading to partially observable MDPs (POMDPs) \citep{aastrom1965optimal,kaelbling1998planning}. The presence of unobserved states introduces substantial statistical and computational challenges, as optimal decisions may depend on the entire observation history rather than the current observation alone. Consequently, both learning, planning, and policy evaluation become significantly more difficult than in fully observable MDPs. From a statistical perspective, obtaining rigorous finite-sample guarantees often requires additional structural assumptions, such as observability, low-rank structure, or suitable mixing conditions. Despite remarkable recent progress (e.g., \citet{liu2022partially,liu2023optimistic,golowich2022learning,uehara2022provably,shi2022minimax,zhan2022pac,chen2023lower}), the statistical foundations of POMDPs remain considerably less mature than those of fully observable MDPs. 

\item {\em Transfer learning for RL from heterogeneous data.}
Another compelling frontier is the development of RL methods that can effectively leverage heterogeneous data sources. Classical offline RL frameworks often assume that historical data are drawn from a single homogeneous population. In contrast, real-world applications such as personalized medicine, recommendation systems, and e-commerce routinely involve data collected from diverse subpopulations. Naively pooling these multi-source datasets and treating them as arising from a common distribution often leads to suboptimal statistical performance, as it fails to exploit the latent similarities and structural relationships across sources.
Addressing this challenge requires new algorithmic and statistical foundations that can effectively integrate transfer learning into RL \citep{taylor2009transfer,chen2022transferred,zhu2023transfer,cai2024transfer,chai2025deep,miao2025reinforcement}. The goal is to enable RL agents to effectively leverage information across related domains, populations, or environments, thereby improving sample efficiency and generalization while maintaining strong statistical guarantees.

\item {\em The interplay between RL and diffusion models.}
Diffusion models \citep{lai2025principles,chen2024opportunities} have recently emerged as a powerful generative framework for RL and control, with applications ranging from diffusion policies for imitation learning and offline RL \citep{chi2025diffusion,ajay2023conditional} to diffusion-based planning and policy optimization \citep{janner2022planning}. Their ability to represent complex, multimodal action distributions makes them particularly appealing for high-dimensional decision-making problems. Conversely, RL and control have increasingly been employed to improve and fine-tune diffusion models themselves, enabling alignment with downstream objectives that are difficult to specify through conventional likelihood-based training alone \citep{black2024training,fan2023dpok,han2024stochastic,yang2026diffusion,zhao2025score,jiao2025connections,gao2025reward}. Despite substantial empirical progress, the statistical and algorithmic foundations of these developments remain significantly underdeveloped, which point to a rich and rapidly growing research agenda at the intersection of RL and generative AI.

%Despite empirical advances, however, the statistical and algorithmic foundations of diffusion-based RL remain largely underdeveloped. Fundamental questions include how to quantify uncertainty and distribution shift for diffusion policies, how estimation errors in the learned diffusion model affect downstream planning and learning performance, and how to establish finite-sample guarantees for learning and control with diffusion-based policies. More broadly, these questions point to a rich research agenda at the intersection of RL and modern generative AI.

\item {\em RL for reasoning in foundation models.} 
Recent years have witnessed a remarkable expansion of RL beyond its traditional applications in sequential decision-making. In particular, RL has emerged as a cornerstone of the post-training and alignment pipelines for foundation models, where it is increasingly deployed to enhance reasoning, planning, and other high-level cognitive capabilities \citep{ouyang2022training,guo2025deepseek}. Unlike the settings studied in this monograph, the underlying environments are often generated by large language models themselves, the action spaces consist of long sequences of tokens, and rewards may be sparse, delayed, and derived from automated verifiers and/or human feedback. These introduce a host of new statistical and algorithmic questions concerning exploration, credit assignment, reward design, and the interplay between RL training and model architecture. Developing a principled theoretical framework that explains when and how RL improves reasoning in foundation models has emerged as an important frontier of modern RL (see, e.g., \citet{kim2025metastable,shang2026errors,huang2026learning,foster2025good,zhao2025diffpo,ran2026outcome,chen2026outcome,yang2026agentic}).
\end{itemize}

%% file: chapters/ack.tex
\begin{acknowledgements}

This monograph will not be possible without the contributions and encouragement of our collaborators over the years, including but not limited to Changxiao Cai, Shicong Cen, Chen Cheng, Bo Dai, Hanjun Dai, Simon Du, Wei Fan, Jingchu Gai, Matthieu Geist, Yu Huang, Yuchen Jiao, Gauri Joshi, Jason Lee, Jiadong Liang, Yingbin Liang, Jason Lee, Cong Ma, Wanteng Ma, Eric Mazumdar, Jincheng Mei, Alessandro Rinaldo, Sudeep Salgia, Dale Schuurmans, Laixi Shi, Kevin Tan, Bingyan Wang, He Wang, Zixin Wen, Adam Wierman, Jiin Woo, Weichen Wu, Lin Xiao, Tong Yang, Yuepeng Yang, Wenhao Zhan, Zihan Zhang, and Yuchen Zhou.

We would also like to thank the generous support of multiple funding agencies. 
Y.~Chen has been supported in part by the Alfred P.~Sloan Research Fellowship, the Google Research Scholar Award, the AFOSR grants FA9550-19-1-0030 and FA9550-22-1-0198, the ONR grant N00014-22-1-2354, and the NSF grants CCF-2221009,
CCF-1907661, DMS-2014279, IIS-2218713 and IIS-2218773, and the Wharton AI \& Analytics Initiative's AI Research Fund. 
Y.~Chi has been supported in by part by the grants ONR  N00014-19-1-2404, AFRL FA8750-20-2-050, NSF CCF-2106778, CNS-2148212, and ECCS-2318441, and in part by funds from federal agencies and industry partners as specified in the NSF Resilient \& Intelligent NextG Systems (RINGS) program. J.~Fan has been supported in part by the ONR Grants N00014-25-1-2317, ONR N00014-22-1-2340, and N00014-19-1-2120, the NSF grants DMS-2210833, DMS-2053832, and DMS-2052926, and the NIH grant 2R01-GM072611-13.
G.~Li is supported in part by the Chinese University of Hong Kong Direct Grant for Research and the Hong Kong Research Grants Council ECS 24305724 and GRF 14307525.
Y.~Wei has been supported in part by the Google Research Scholar Award,  the NSF
grants CCF-2106778, DMS-2147546/2015447 and CAREER award DMS-2143215, and the Wharton Dean's Research Fund. This work is also supported in part by the NSF under Cooperative Agreement No.~2433450.

\end{acknowledgements}